\documentclass[12pt,oneside,english]{book}
\usepackage[T1]{fontenc}
\usepackage[latin9]{inputenc}
\usepackage[a4paper]{geometry}
\usepackage{array}
\usepackage{verbatim}
\usepackage{longtable}
\usepackage{float}
\usepackage{amsmath}
\usepackage{amssymb}
\usepackage{graphicx}
\usepackage{setspace}
\usepackage[section] {placeins}
\PassOptionsToPackage{normalem}{ulem}
\usepackage{ulem}
\makeatletter

\providecommand{\tabularnewline}{\\}
\floatstyle{ruled}
\newfloat{algorithm}{tbp}{loa}[chapter]
\providecommand{\algorithmname}{Algorithm}
\floatname{algorithm}{\protect\algorithmname}

\usepackage{geometry}
\usepackage{array}
\usepackage{float}
\usepackage{setspace}

\usepackage{times}

\usepackage{fancyheadings}
\renewcommand{\chaptermark}[1]{\markboth{#1}{}}
\renewcommand{\sectionmark}[1]{\markright{\thesection\ #1}}
\newcommand{\titlefont}{\bfseries \fontsize{22}{25pt} \linespread{1.2} \selectfont}

\date{}

\@ifundefined{showcaptionsetup}{}{%
 \PassOptionsToPackage{caption=false}{subfig}}
\usepackage{subfig}
\makeatother

\usepackage{babel}
\begin{document}
\begin{singlespace}
\begin{center}
\thispagestyle{empty}\includegraphics[width=1\columnwidth,keepaspectratio]{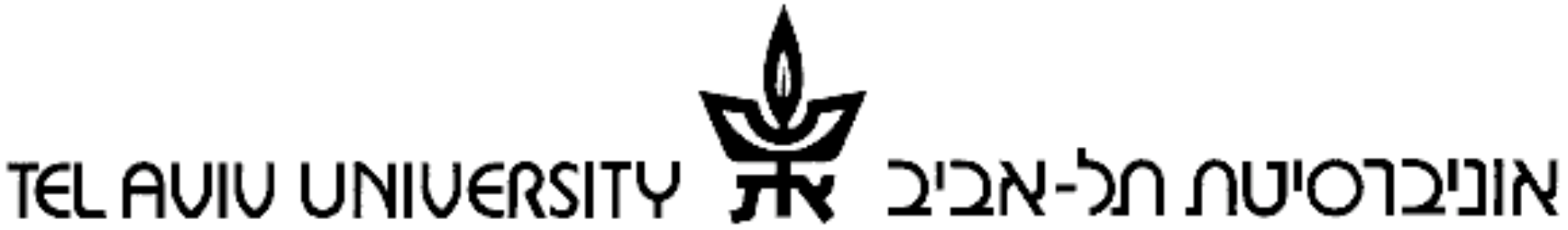}
\par\end{center}

\begin{center}
\parbox{\textwidth}{\large \centering 
Raymond and Beverly Sackler\\
Faculty of Exact Sciences\\
School of Computer Science}
\par\end{center}

\begin{center}
\vfill{}
\parbox{\textwidth}{\titlefont \centering Multi-Sensor Fusion via Reduction of Dimensionality}
\par\end{center}
\end{singlespace}

\begin{center}
\vfill{}

\par\end{center}

\begin{singlespace}
\begin{center}
\parbox{\textwidth}{\large \centering Thesis submitted for the degree of \\
\textbf {``Doctor of Philosophy''}
}\vfill{}

\par\end{center}

\begin{center}
{\large by}
\par\end{center}{\large \par}

\begin{center}
\textbf{\large Alon Schclar}
\par\end{center}{\large \par}

\begin{center}
\vfill{}

\par\end{center}

\begin{center}
\parbox{\textwidth}{\large \centering This thesis was carried out under the supervision of \\
\textbf {Prof. Amir Averbuch}
}\vfill{}

\par\end{center}

\begin{center}
\parbox{\textwidth}{\large \centering Submitted to the Senate of Tel-Aviv University\\
January, 2008
}
\par\end{center}
\end{singlespace}

\newpage{}

\thispagestyle{empty}~\newpage{}

\thispagestyle{empty}\vfill{}

\begin{center}
\parbox{\textwidth}{\vskip 11cm \Large \centering This thesis was carried out under the supervision of \\ Prof. Amir Averbuch}
\par\end{center}

\vfill{}
\newpage{}

\thispagestyle{empty}~\newpage{}

\begin{center}
\thispagestyle{empty}\vfill{}
\parbox{\textwidth}{\vskip 11cm \Large \centering \emph{Dedicated to my beloved family}}\vfill{}
\newpage{}
\par\end{center}

\thispagestyle{empty}~\newpage{}

\chapter*{Abstract}

Large high-dimensional datasets are becoming more and more popular
in an increasing number of research areas. Processing the high dimensional
data incurs a high computational cost and is inherently inefficient
since many of the values that describe a data object are redundant
due to noise and inner correlations. Consequently, the dimensionality,
i.e. the number of values that are used to describe a data object,
needs to be reduced prior to any other processing of the data. The
dimensionality reduction removes, in most cases, noise from the data
and reduces substantially the computational cost of algorithms that
are applied to the data.

In this thesis, a novel coherent integrated methodology is introduced
(theory, algorithm and applications) to reduce the dimensionality
of high-dimensional datasets. The method constructs a diffusion process
among the data coordinates via a random walk. The dimensionality reduction
is obtained based on the eigen-decomposition of the Markov matrix
that is associated with the random walk. The proposed method is utilized
for: (a) segmentation and detection of anomalies in hyper-spectral
images; (b) segmentation of multi-contrast MRI images; and (c) segmentation
of video sequences. 

We also present algorithms for: (a) the characterization of materials
using their spectral signatures to enable their identification; (b)
detection of vehicles according to their acoustic signatures; and
(c) classification of vascular vessels  recordings to detect hyper-tension
and cardio-vascular diseases.

The proposed methodology and algorithms produce excellent results
that successfully compete with current state-of-the-art algorithms.

\newpage{}

\thispagestyle{empty}~

\newpage{}

\thispagestyle{empty}\textbf{\Large Acknowledgments}\\

First and foremost, I wish to extend my deepest gratitude to my adviser
Prof. Amir Averbuch for his invaluable guidance throughout the preparation
of this thesis. Prof. Averbuch has been my adviser since I started
my M.Sc degree and he has been a source for inspiration, knowledge
and creativity, ever since. His patience and willingness to go far
and beyond for his students is nothing less than inspiring - setting
high standards to follow. Working with Prof. Averbuch has been an
honor and a privilege and I can only hope to continue and conduct
further research with him in the years to come. 

I am also in debt to Prof. Ronald R. Coifman for his insightful ideas
and enlightening discussions and especially for being my adviser at
Yale University during my fellowship there. His original ideas and
fruitful advise have been keystones in my voyage into this fascinating
research area.

I would like to thank Dr. Yosi Keller for his very helpful advice
and discussion during my stay at Yale University. Yosi is not only
a collaborator but also a friend who made, along with his charming
wife, Orit, tremendous efforts to welcome me at Yale. I am grateful
for both of them for their very warm welcome and hospitality.

Special thanks goes to Dr. Amit Singer and Dr. Yoel Shkolinsky for
their valuable critic.

I would like to thank the Fulbright Foundation (The United States-Israel
Educational Foundation) for granting me the \emph{Fulbright Doctoral
Dissertation Research Fellowship.} This distinguished grant, which
enabled me to carry out research at the Department of Applied Mathematics
and Computer Science at Yale University, had a crucial contribution
to my thesis.

And last but never the least, I am grateful to my family for their
constant support, encouragement and belief in me throughout my Ph.D
studies.

%Amir, Family, Fulbright, The authors would like to thank Ronald Coifman and Yosi Keller for their helpful suggestions and to Amit Singer and Yoel Shkolinsk for their valuable critic on the automatic derivation of epsilon.

\newpage{}

\thispagestyle{empty}~\newpage{}

\tableofcontents{}

\listoffigures

\newpage{}

\thispagestyle{empty}~\newpage{}

\listoftables

\listof{algorithm}{List of Algorithms}

\newpage{}

\thispagestyle{empty}~

\chapter*{List of abbreviations}

\renewcommand{\chaptermark}[1]{}
\renewcommand{\sectionmark}[1]{}
\lhead[\fancyplain{}{\bfseries LIST OF ABBREVIATIONS}]
   {\fancyplain{}{\bfseries LIST OF ABBREVIATIONS}}
\rhead[\fancyplain{}{\bfseries\leftmark}]
   {\fancyplain{}{\bfseries\thepage}}
\cfoot{}

\begin{flushleft}
\begin{longtable}{ll}
BGD & Background data\tabularnewline
BSDB & Background subtraction algorithm using diffusion bases\tabularnewline
CART & Classification and regression trees\tabularnewline
CCD & Charge-coupled device \tabularnewline
CPU & Central processing unit\tabularnewline
CSF & Cerebrospinal fluid\tabularnewline
DB & Diffusion bases\tabularnewline
DB1 & Signature database 1\tabularnewline
DB2 & Signature database 2\tabularnewline
DBG & Dynamic background\tabularnewline
DBSDB & Dynamic background subtraction algorithm using DB\tabularnewline
DFS & Depth first search\tabularnewline
DM & Diffusion maps\tabularnewline
DTI & Diffusion tensor imaging\tabularnewline
FIFO & First in first out\tabularnewline
FLAIR & Fluid level attenuated inversion recovery\tabularnewline
fMRI & Functional magnetic resonance imaging\tabularnewline
FOV & Field of view\tabularnewline
GMDS & Generalized multidimensional scaling\tabularnewline
HLLE & Hessian local linear embedding\tabularnewline
ID & Intrinsic dimensionality\tabularnewline
KDE & Kernel density estimation\tabularnewline
KPCA & Kernel principle component analysis\tabularnewline
LDA & Linear discriminant analysis\tabularnewline
LLE & Local linear embedding\tabularnewline
LTSA & Local tangent space analysis\tabularnewline
MD & Medial-dorsal nucleus\tabularnewline
MDB & Modified diffusion bases\tabularnewline
MDM & Modified diffusion maps\tabularnewline
MDS & Multidimensional scaling\tabularnewline
MgT & Magnetization transfer\tabularnewline
MinDist & Minimal distance\tabularnewline
MRI & Magnetic resonance imaging\tabularnewline
OIF & Optimum index factor\tabularnewline
PCA & Principle component analysis\tabularnewline
PCM & Pulse-code modulation\tabularnewline
RAM & Random access memory\tabularnewline
RGB & Red, Green and Blue\tabularnewline
ROI & Region of interest\tabularnewline
RSNOFP & Random search for a near-optimal footprint\tabularnewline
RTD & Real-time data\tabularnewline
SAM & Spectral angle mapper\tabularnewline
SBG & Static background\tabularnewline
SBSDB & Static background subtraction algorithm using DB\tabularnewline
SCM & Spectral correlation mapper\tabularnewline
SDK & Silhouette-driven K-means\tabularnewline
SFF & Spectral feature fitting\tabularnewline
SIM & Spectral identification method\tabularnewline
SNE & Stochastic neighbor embedding\tabularnewline
SNF & Sub-nuclei finder \tabularnewline
SPE & Stochastic proximity embedding\tabularnewline
SPGR & Spoiled gradient recalled echo\tabularnewline
SPS & Samples per second\tabularnewline
SR & Sampling rate\tabularnewline
STIR & Short tau inversion recovery\tabularnewline
SW & Sliding window\tabularnewline
WAV & Wavelength-averaged-version\tabularnewline
WPT & Wavelet packet transform\tabularnewline
WWG & Wavelength-wise global\tabularnewline
SOM & Self-organizing maps\tabularnewline
GTM & Generative topographic map\tabularnewline
RBF & Radial basis function\tabularnewline
AVIRIS & Airborne Visible InfraRed Imaging Spectrometer\tabularnewline
\end{longtable}
\par\end{flushleft}

\newpage{}

\chapter{Introduction\label{cha:INTRODUCTION}}

\begin{figure}[!t]
\begin{centering}
\includegraphics[%bb=102bp 300bp 480bp 550bp,
width=3in]{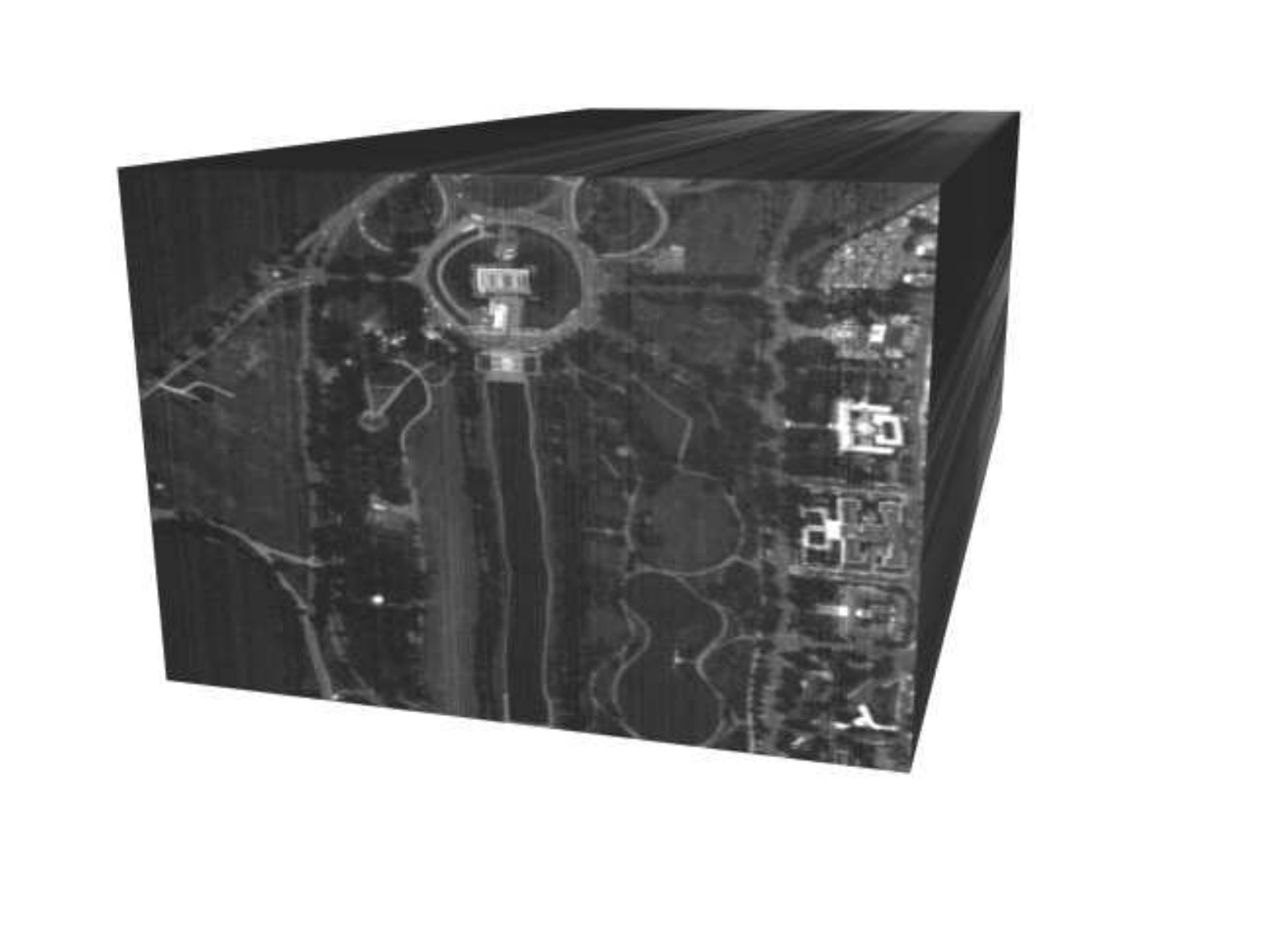}
\par\end{centering}

\caption{A satellite-generated hyper-spectral image of Washington DC mall, USA.\label{cap:Hyper-spectral-cube}}
\end{figure}

\begin{figure*}[!t]
\begin{centering}
\begin{tabular}{>{\centering}m{0.48\columnwidth}>{\centering}m{0.48\columnwidth}}
\multicolumn{1}{c}{\includegraphics[%bb=141bp 297bp 467bp 554bp,clip,
width=0.45\textwidth,height=0.48\columnwidth,keepaspectratio]{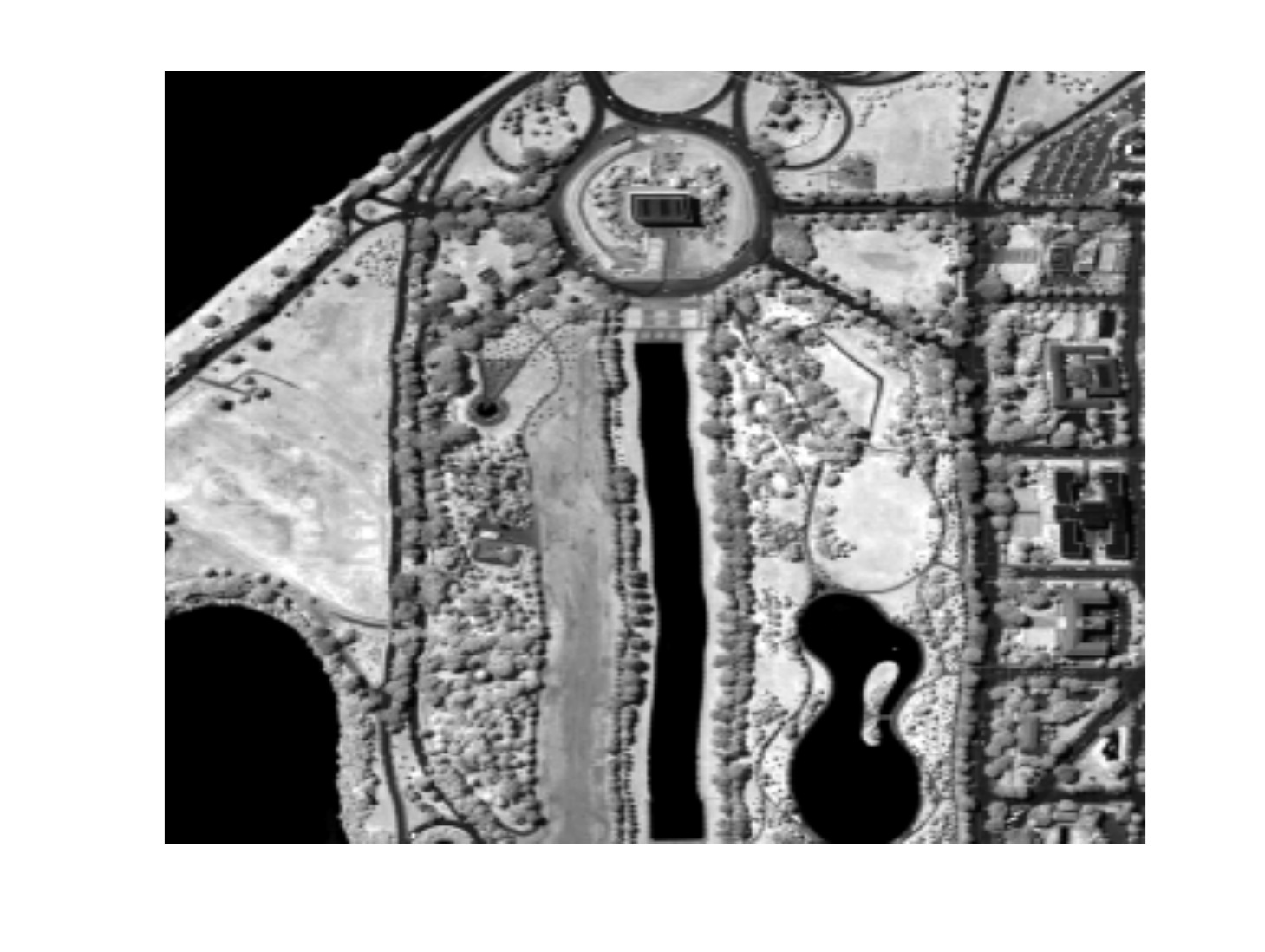}} & \multicolumn{1}{c}{\includegraphics[%bb=141bp 297bp 467bp 554bp,clip,
width=0.45\textwidth,height=0.48\columnwidth,keepaspectratio]{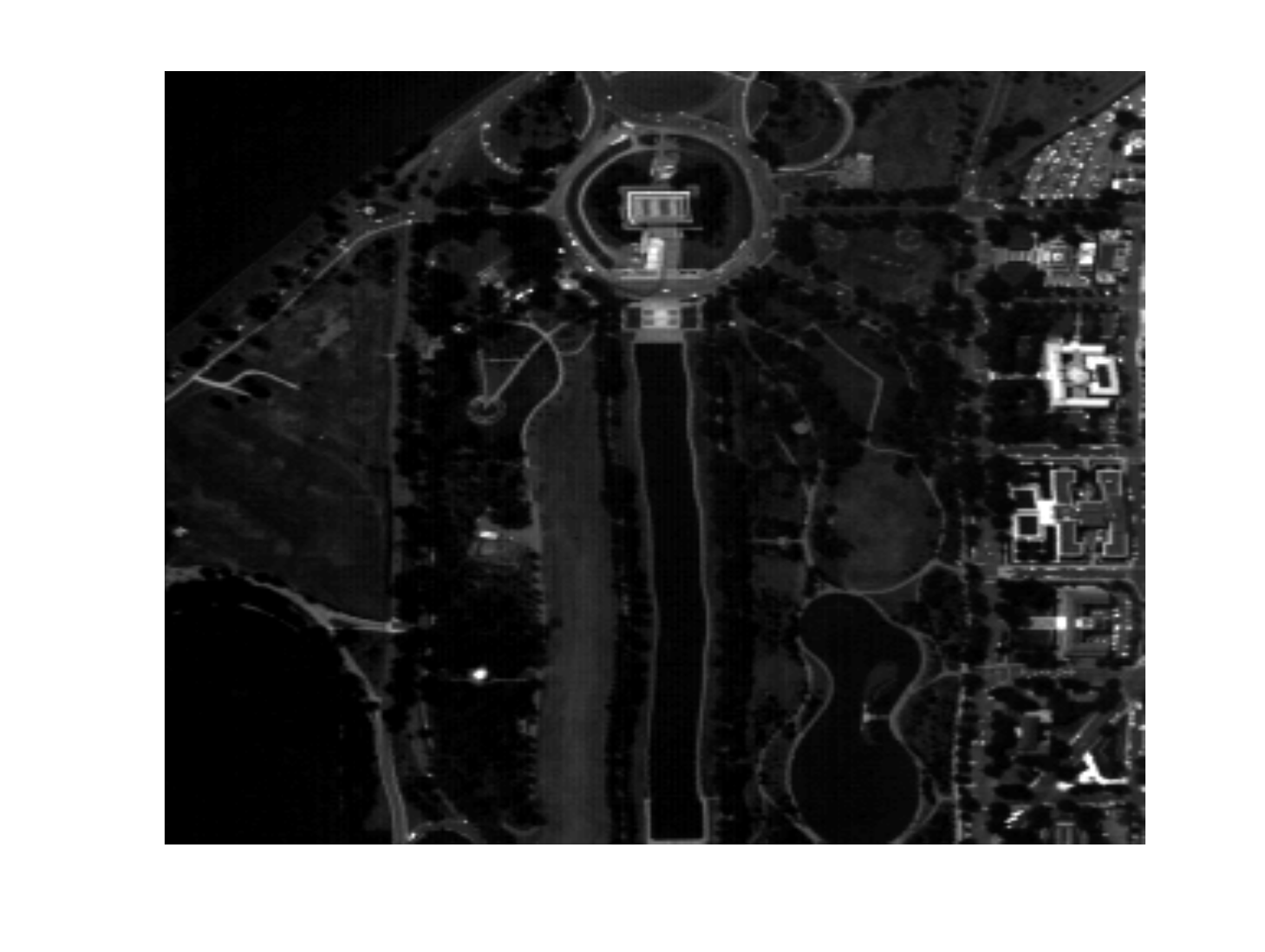}}\tabularnewline
\multicolumn{2}{c}{}\tabularnewline
\multicolumn{1}{c}{} & \multicolumn{1}{c}{}\tabularnewline
\end{tabular}
\par\end{centering}

\caption{The images of two wavelengths of Fig. 1.1. \label{fig:A-hyper-spectral-satellite}}
\end{figure*}

\renewcommand{\chaptermark}[1]{}
\renewcommand{\sectionmark}[1]{}
\lhead[\fancyplain{}{\bfseries INTRODUCTION}]
   {\fancyplain{}{\bfseries INTRODUCTION}}
\rhead[\fancyplain{}{\bfseries\leftmark}]
   {\fancyplain{}{\bfseries\thepage}}
\cfoot{}
The overflow of data is a critical contemporary challenge in many
areas such as information retrieval, biotechnology, textual search,
hyper-spectral sensing, classification etc. It is commonly manifested
by a high dimensional representation of data observations. A dimension
of a data observation is the number of values that describes it. A
simple example is an ordinary color image where each pixel has 3 values
that represent the red, green and blue values. In this example, the
dimensionality is low (equals to 3), however in hyper-spectral images,
the dimensionality can reach several hundreds (each value corresponds
to a different wavelength of the spectrum). Figure \ref{cap:Hyper-spectral-cube}
illustrates a hyper-spectral image, which is also referred as a hyper-spectral
cube (see Chapter \ref{cha:Introduction-to-Hyper-spectral}). The
image is a birds-eye view of Washington DC mall, USA, and its surroundings.
The images that correspond to two of the wavelengths are given in
Fig. \ref{fig:A-hyper-spectral-satellite}. Images such as those in
Figs. \ref{cap:Hyper-spectral-cube} and \ref{fig:A-hyper-spectral-satellite}
are common in remote sensing applications e.g. segmentation and anomalies
detection (see Chapter \ref{cha:Segmentation-and-Anomalies}).

\begin{figure*}[!t]
\begin{centering}
\begin{tabular}{>{\centering}m{0.48\textwidth}>{\centering}m{0.48\textwidth}}
\multicolumn{1}{c}{\includegraphics[%bb=141bp 297bp 467bp 554bp,clip,
width=0.45\textwidth,keepaspectratio]{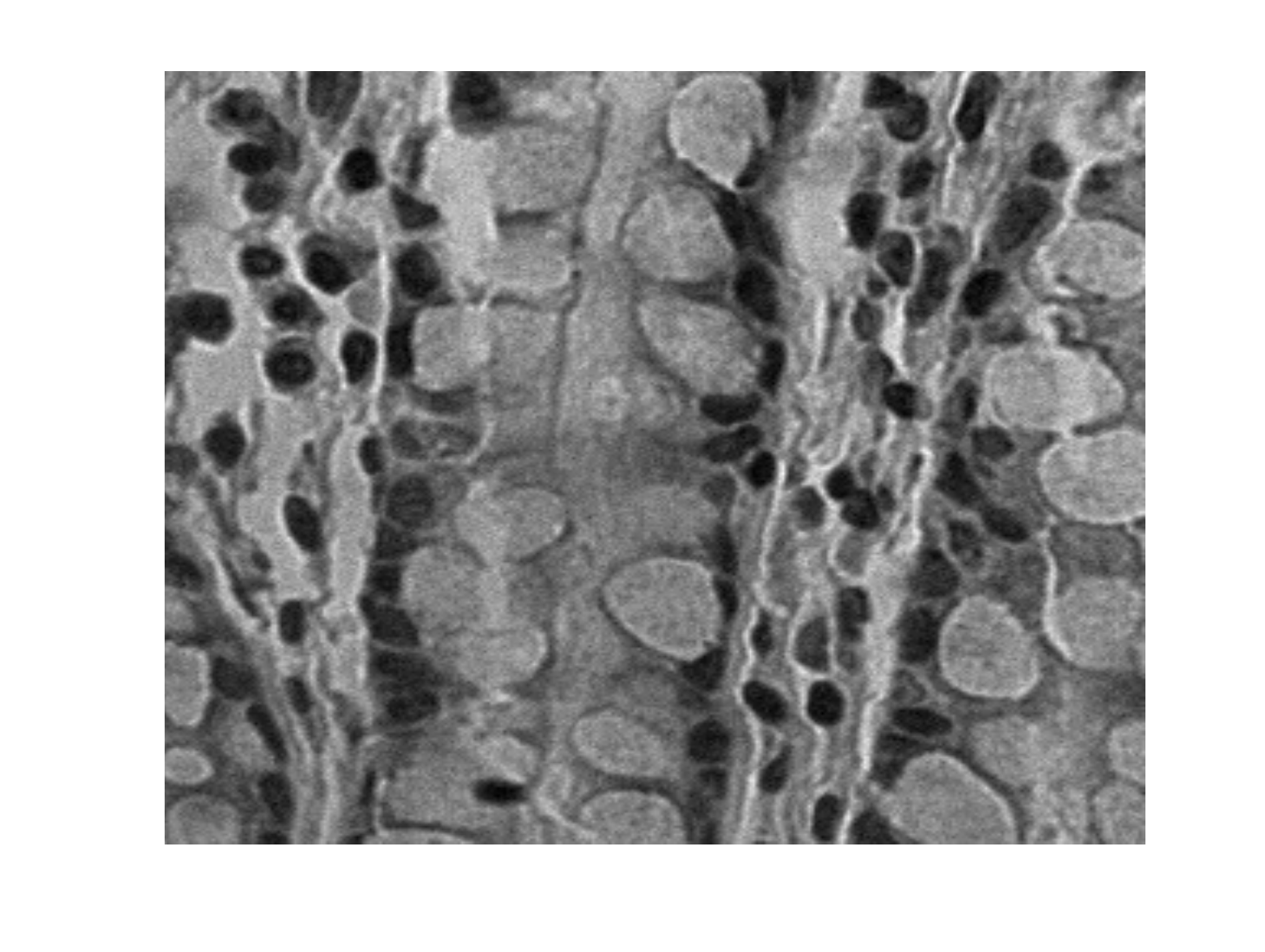}} & \multicolumn{1}{c}{}\tabularnewline
\multicolumn{2}{c}{}\tabularnewline
\multicolumn{1}{c}{} & \multicolumn{1}{c}{}\tabularnewline
\end{tabular}
\par\end{centering}

\caption{Two wavelengths of a hyper-spectral microscopy image of a healthy
human colon tissue.\label{fig:Two-wavelengths-of}}
\end{figure*}

Hyper-spectral images are also used in medical applications such as
detection of tissue anomalies. In this case a hyper-spectral microscope
is used to capture hyper-spectral images of tissue samples. These
images are analyzed in order to detect benign and malignant tumors
\cite{Mauro1,Mauro2,Mauro4}. Figure \ref{fig:Two-wavelengths-of}
shows two wavelengths of a hyper-spectral microscope-generated image
of a human colon tissue. A more comprehensive introduction on hyper-spectral
imagery, can be found in Chapter \ref{cha:Introduction-to-Hyper-spectral}.

Another area that uses high-dimensional data is \textbf{information
retrieval}. Applications in this area include text mining, search
engines (\emph{Google}\texttrademark   search is one typical example),
etc. In order to process text documents, they must be represented
as a vector \cite{LL06}. One possible high-dimensional representation
of a text document is a vector whose coordinates contain the number
of occurrences of common and uncommon words. 

\textbf{Financial data} such as credit transactions, loan payments
and transactions in banking monitoring systems are also examples for
high-dimensional data. Data-mining applications such as fraud detection
and credit-rate evaluation commonly use high-dimensional financial
data to infer behavior characteristics of clients. One representation
of high-dimensional financial-data is transaction amounts along time
periods whose length is predefined. 

The main problem of high dimensional data is the so called \emph{curse
of dimensionality}, which means that in a large number of algorithms
the complexity grows exponentially with the increase of the dimensionality
of the input data \cite{JL98}. Furthermore, in certain situations,
the number of observations is not sufficient to produce satisfactory
reduction of dimensionality \cite{5_Bellman61}.

Commonly, the acquiring sensor produces data whose dimensionality
is much higher than the actual degrees of freedom of the data. Unfortunately,
this phenomenon is usually unavoidable due to the inability (lack
of knowledge which values from the sensor are more important for the
task at hand) to produce a special sensor for each application. Consider
for example a task that separates red objects from green objects using
an off-the-shelf digital camera. In this case, a the camera will produce,
in addition to the red and green channels, a blue channel, which is
unnecessary for this task.

In order to efficiently process high-dimensional datasets, one must
first analyze their geometrical structure and detect the actual degrees
of freedom, which are manifested by a (small) number of parameters
that govern the structure of the dataset. This number is referred
to as the \emph{intrinsic dimension} (ID) of the dataset. Consequently,
the information that is conveyed by the dataset can be described by
a set of vectors whose dimension is equal to the ID of the original
dataset. 

There is a direct connection between the intrinsic dimensionality
of the data and the degrees of freedom of the observations. % However, this connection relates to the data observations and not to the data itself.
This is one of the key motivations for trying to reduce the dimensionality
of the observed data. Commonly, correlation exists between some subsets
of the observation variables. Hence, it is reasonable to try and find
a mapping that embeds the data into a low-dimensional space in which
the data is represented by a small number of \emph{uncorrelated} variables.
This embedding can additionally reveal hidden information in the original
data, even before subsequent algorithms are applied. The loss of information,
which is due to the dimensionality reduction, is not always a disadvantage,
on the contrary, the lost information is in many cases not essential,
for example information which is due to noise.

The process of finding a low-dimensional representation is called
\emph{dimensionality reduction.} Dimensionality reduction is also
referred to as \emph{manifold learning.} An ID that is lower than
the dimension of the ambient space, means that the dataset lies near
or on a manifold whose dimension is equal to the ID. \emph{Local}
dimensionality reduction techniques, which are introduced in Section
\ref{sub:Local-methods}, are better defined as manifold learning
techniques since they investigate local neighborhoods, which usually
have a simple structure, in order to reveal the global and usually
more complicated structure of the dataset.

One of the most important applications of dimensionality reduction
is visualization of high-dimensional datasets. This is facilitated
when the dimension of the reduced space is not greater than three.
The high-dimensional set is visualized by a plot of its low-dimensional
embedding. Since most dimensionality techniques are designed to produce
a low-dimensional embedding that preserves the geometry of the original
high-dimensional set, the visualized results provide valuable information
regarding the geometry of the dataset at hand.

\paragraph{Contribution of this thesis}

The contribution of this thesis is two-fold. First, a novel method
for dimensionality reduction - \emph{Diffusion Bases} - is introduced.
The method is based on the diffusion map dimensionality reduction
algorithm and the theoretical connection between the methods is described.
The applicative effectiveness of the diffusion bases method is demonstrated
for the segmentation of images that originate from various domains
- hyper-spectral imagery, multi-contrast MRI and video. 

Second, this thesis shows that dimensionality reduction is an effective
tool for solving problems from various domains, namely, identification
of materials via their spectral signatures, detection of vehicles
using their acoustic signatures and detection of vascular diseases
using acoustic recordings of blood vessels.

\paragraph{Structure of this thesis}

This thesis is composed of three parts. In the first part, we introduce
the novel Diffusion Bases methodology (theory, algorithms and applications)
for dimensionality reduction. Specifically, in Chapter \ref{cha:Dimensionality-reduction}
we give an in depth introduction to dimensionality reduction where
we provide a formal definition of the problem followed by a description
of the current state-of-the-art techniques for dimensionality reduction.
In Chapter \ref{cha:Diffusion-Maps} we describe in details the diffusion
maps technique \cite{CL_DM06} since it is closely connected to our
diffusion Bases (DB) dimensionality reduction scheme which we introduce
in Chapter \ref{cha:Diffusion-bases}. The DB algorithm explores the
variability among the \emph{coordinates} of the original data while
the DM explores local neighborhoods of points in the dataset. Both
algorithms use a random walk model. The DB algorithm uses the eigenvectors
of the corresponding Markov matrix as an orthonormal system and projects
the original data onto it to obtain the low-dimensional representation.
The DM algorithm, on the other hand, builds a different Markov matrix
whose eigenvectors constitute the low-dimensional representation.
In Chapter \ref{cha:Introduction-to-Hyper-spectral} we provide an
introduction to hyper-spectral imagery which includes the terminology,
concept, motivation and common applications in this area. Chapter
\ref{cha:Introduction-to-Hyper-spectral} is necessary for the understanding
of Chapters \ref{cha:Segmentation-and-Anomalies} and \ref{cha:Automatic-identification-of}.
Chapters \ref{cha:Segmentation-and-Anomalies}-\ref{cha:Video-segmentation-via}
include successful applications of the DB scheme. Specifically, in
Chapter \ref{cha:Segmentation-and-Anomalies}, the DB dimensionality
reduction scheme is used for segmentation of hyper-spectral images
and for the detection of anomalies in images of this type. In Chapter
\ref{cha:Neuronal-Tissues-Sub-Nuclei}, the DB scheme is incorporated
in an algorithm for segmentation of multi-contrast MRI images. Segmentation
of video sequences which uses the DB scheme is described in Chapter
\ref{cha:Video-segmentation-via}.

In the second part of this thesis, we present two methods for uniquely
identifying materials according to their spectral signatures. Given
a spectral signature of a material to be identified, the first method
seeks an exact match in a database of spectral signatures while the
second method takes into account noise and looks for an approximate
identification. Both methods reduce the dimensionality of the spectral
signatures by means of feature extraction.

In the third part, we introduce two methods for the detection and
classification of predefined events in acoustic signals. The first
method is tailored for the detection of vehicles in acoustic signals
that were recorded in various terrains where noise and background
sounds are present. The second method detects hyper-tension and cardio-vascular
diseases according to recordings of vascular vessels. Both methods
reduce the dimensionality of the recordings in order to extract features
which uniquely characterize the various events that need to be detected
and classified. 

This thesis is based on the following papers: \cite{mine-DB07}, \cite{mine-Neuronal07},
\cite{mine-video07}, \cite{mine-nati1}, \cite{mine-nati2}, \cite{mine-vehicle-wave07}
and \cite{mine-vascular07}. Table \ref{tab:Paper-Chapter-associations}
associates the chapters with the papers. Figure \ref{fig:The-chapter-flow}
illustrates the connections and dependencies between the chapters.
It indicates the order in which the chapters are to be read. For example,
in order to read Chapters 6-8, Chapters 1-5 have to be read first.

\begin{center}
\begin{table}[!h]
\caption{Paper-Chapter associations\label{tab:Paper-Chapter-associations}}

\centering{}%
\begin{tabular}{|c|c|}
\hline 
Paper & Chapter\tabularnewline
\hline 
\hline 
\cite{mine-DB07} & \ref{cha:Diffusion-bases}, \ref{cha:Segmentation-and-Anomalies}\tabularnewline
\hline 
\cite{mine-Neuronal07} & \ref{cha:Neuronal-Tissues-Sub-Nuclei}\tabularnewline
\hline 
\cite{mine-video07} & \ref{cha:Video-segmentation-via}\tabularnewline
\hline 
\cite{mine-nati1}, \cite{mine-nati2} & \ref{cha:Automatic-identification-of}\tabularnewline
\hline 
\cite{mine-vehicle-wave07} & \ref{cha:Wavelet-based-acoustic}\tabularnewline
\hline 
\cite{mine-vascular07} & \ref{cha:Classification-and-Detection-Vascular}\tabularnewline
\hline 
\end{tabular}
\end{table}

\par\end{center}

\begin{center}
\begin{figure}[!h]
\begin{centering}
\includegraphics[bb=50bp 580bp 555bp 800bp,clip,width=1\columnwidth]{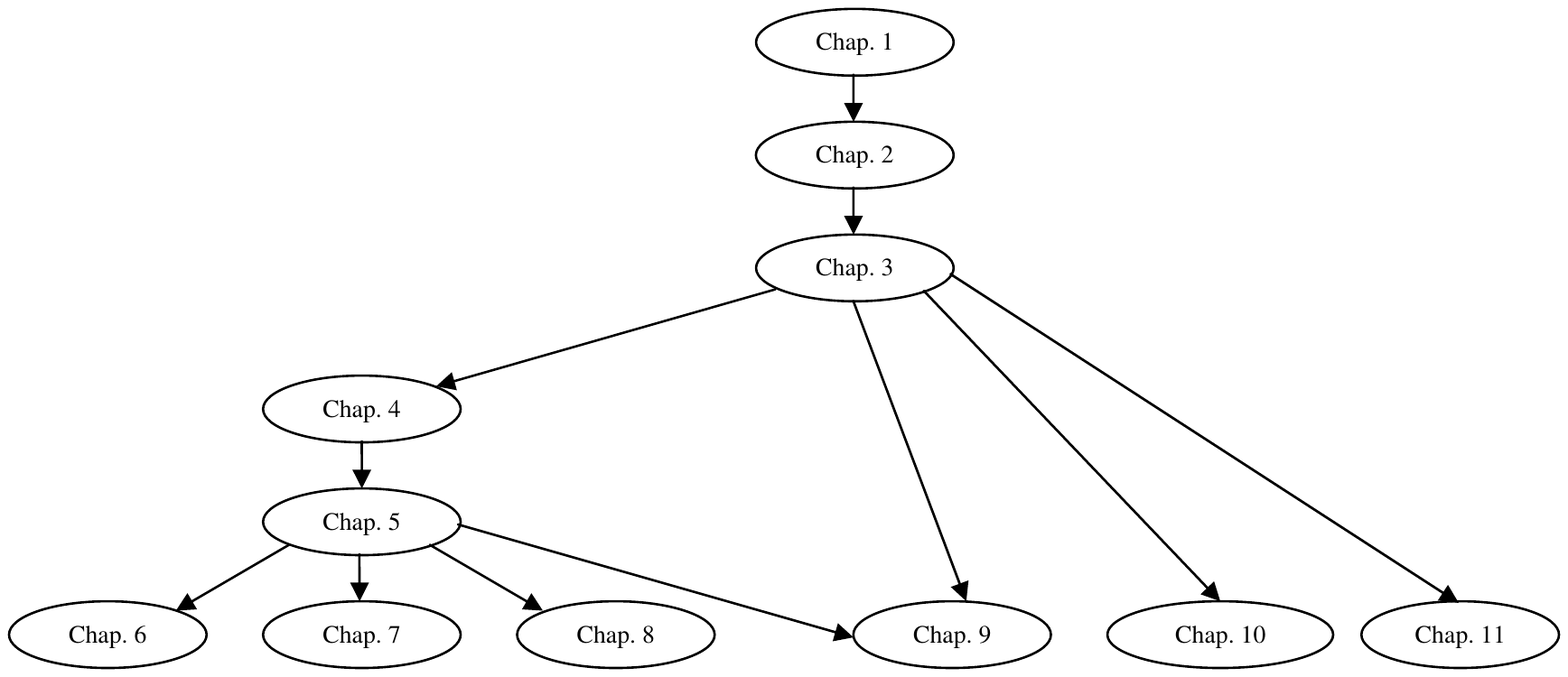}
\par\end{centering}

\caption{The chapter flow in this thesis.\label{fig:The-chapter-flow} }
\end{figure}

\par\end{center}

\newpage{}

\thispagestyle{empty}~

\chapter{Dimensionality Reduction\label{cha:Dimensionality-reduction}}

\renewcommand{\chaptermark}[1]{\markboth{#1}{}}
\renewcommand{\sectionmark}[1]{\markright{\thesection\ #1}}
\lhead[\fancyplain{}{\bfseries\thepage}]
   {\fancyplain{}{\bfseries\rightmark}}
\rhead[\fancyplain{}{\bfseries\leftmark}]
   {\fancyplain{}{\bfseries\thepage}}
\cfoot{}
In this chapter, we describe in detail the current state-of-the-art
techniques for dimensionality reduction. We focus on a set of methods
that we feel span most of the different approaches to tackle the dimensionality
reduction problem while providing references to other available methods.

\subsection*{Notation}

The following notation are used throughout this chapter. We denote
a original high-dimensional dataset as a set of column vectors 
\begin{equation}
\Gamma=\left\{ x_{i}\right\} _{i=1}^{N}\label{eq:gamma}
\end{equation}
 where $x_{i}\in\mathbb{R}^{n}$, $n$ is the (high) dimension and
$N$ is the size of the dataset. All dimensionality reduction methods
embed the vectors into a lower dimensional space $\mathbb{R}^{q}$
where $q\ll n$. Their output is a set of column vectors in the lower
dimensional space 
\begin{equation}
\widetilde{\Gamma}=\left\{ \widetilde{x}_{i}\right\} _{i=1}^{N},\,\widetilde{x}_{i}\in\mathbb{R}^{q}\label{eq:gamma_gal}
\end{equation}
where $q$ approximates the ID (Chapter \ref{cha:INTRODUCTION}) of
$\Gamma$. We refer to the vectors in the set $\widetilde{\Gamma}$
as the \emph{embedding} \emph{vectors}.

\section*{Previous work\label{sec:Previous-work}}

The general problem of dimensionality reduction has been extensively
researched. In their pioneering work, Johnson and Lindenstrauss \cite{JL84}
laid the theoretical foundations by proving the feasibility of dimension
reduction. Specifically, they showed that $N$ points in $N$ dimensional
space can almost always be projected onto a space of dimension $C\log N$
with control on the ratio of distances and the error (distortion).
Bourgain \cite{B85} showed that any metric space with $N$ points
can be embedded by a bi-Lipschitz map into an Euclidean space of $\log N$
dimension with a bi-Lipschitz constant of $\log N$. Various randomized
versions of this theorem are used for protein mapping \cite{LLTY00}
and for reconstruction of frequency sparse signals \cite{CRT06,D06}.

We classify current dimensionality reduction methods into two categories:
global and local methods. Although all methods seek to maintain global
properties of the dataset, they differ in the way they accomplish
this. Global methods (Section \ref{sub:Global-methods}) first extract
the properties they aim to maintain from the high-dimensional dataset
and then seek to embed the data into a low-dimensional space while
preserving the extracted properties for \emph{all} the high-dimensional
points. Local methods (Section \ref{sub:Local-methods}), on the other
hand, utilize the Newtonian paradigm according to which a global description
of a system can be derived by the aggregation of \emph{local} transitions. 

%A different classification approach of dimensionality reduction algorithms along with an extensive comparative review of current state-of-the-art dimensionality reduction techniques can be found in .

\section{\label{sub:Global-methods}Global methods}

In the following, we describe common available global methods for
dimensionality reduction.

\subsection{Principal Component Analysis (PCA)}

PCA \cite{PCA} finds a low-dimensional embedding of the data points
that best preserves their variance as measured in the high-dimensional
input space. As a preliminary step, the vectors in $\Gamma$ are centered
around the origin. Let $\overline{\Gamma}=\left\{ \overline{x}_{i}\right\} _{i=1}^{N}$
be the set $\Gamma$ centered at the origin where $\overline{x}_{i}\triangleq x_{i}-\frac{1}{N}\sum_{j=1}^{N}x_{j}$.
We define a $n\times N$ matrix $X$ whose columns are comprised of
the vectors $\left\{ \overline{x}_{i}\right\} _{i=1}^{N}$ 
\[
X=\left(\overline{x}_{1}|\overline{x}_{2}|\ldots|\overline{x}_{N}\right).
\]
 The covariance matrix $C_{\overline{\Gamma}}$ of the set $\overline{\Gamma}$
is computed by $C_{\overline{\Gamma}}=\frac{1}{N-1}XX^{T}.$ $C_{\overline{\Gamma}}$
is a symmetric $n\times n$ matrix that captures the correlations
between all possible pairs of measurements. An eigen-decomposition
of $C_{\overline{\Gamma}}$ is performed to produce a set of eigenvalues
$\lambda_{1}\ge\lambda_{2}\ge\ldots\ge\lambda_{n}$ and their corresponding
eigenvectors $\varphi_{1},\ldots,\varphi_{n}$. The eigenvector $\varphi_{1}$
gives the direction where the variance of the data is maximal. The
magnitude of this variance is given by $\lambda_{1}$. In a similar
way, the eigenvector $\varphi_{2}$ gives the direction of the second
largest variance $\lambda_{2}$, and so on.

The low dimensional vector $\widetilde{x}_{i}$, which is the embedding
of $x_{i}$, is given by the projection of $\overline{x}_{i}$ onto
the first $q$ eigenvectors $\varphi_{1},\ldots,\varphi_{q}$:

\[
\widetilde{x}_{i}\triangleq\left(\left\langle \overline{x}_{i},\varphi_{1}\right\rangle ,\left\langle \overline{x}_{i},\varphi_{2}\right\rangle ,\ldots,\left\langle \overline{x}_{i},\varphi_{q}\right\rangle \right)
\]
 where $\left\langle \cdot,\cdot\right\rangle $ denotes the inner
product operator.

One can find many applications spanning a wide range of domains that
successfully apply PCA. These include seismology \cite{PVMAPIM93},
face recognition \cite{Eigenfaces91}, coin classification \cite{CoinPCA05}
to name a few. 

The size of the covariance matrix depends on the dimensionality of
the vectors in $\Gamma$. Consequently, the computation of the covariance
matrix may be impossible when the dimensionality of $\Gamma$ is very
high. In such cases, one can use approximation techniques \cite{PCAapprox97}.
Furthermore, PCA is only able to discover the true structure of data
in case it lies on or near a linear subspace of the high-dimensional
input space \cite{MKB02}. This pitfall is crucial since many datasets
contain nonlinear structures.

\subsection{Kernel PCA (KPCA)}

KPCA \cite{KPCA98} is a generalization of PCA that is able to detect
non-linear structures. This ability relies on the \emph{kernel trick:}
any algorithm whose description involves only dot products and does
not require explicit usage of the variables can be extended to a non-linear
version by using Mercer kernels \cite{KPCA_book02}. When this principle
is applied to dimensionality reduction it means that non-linear structures
correspond to linear structures in high-dimensional spaces. 

In order to \emph{linearize} a structure, one must find a transformation
into a high-dimensional space which maps the original non-linear structure
to a high-dimensional linear structure. Finding this is not a trivial
task. Fortunately, this can be accomplished by applying a non-linear
kernel on the pair-wise inner products.

In KPCA, the covariance matrix is substituted by a pair-wise kernel
matrix 
\[
K=\left(k_{ij}\right)=k\left(x_{i},x_{j}\right),\, i,j=1,\ldots,N.
\]
According to the kernel trick we have $k\left(x_{i},x_{j}\right)=\left\langle \Phi\left(x_{i}\right),\Phi\left(x_{j}\right)\right\rangle $
where $\Phi\left(\cdot\right)$ is the linearizing transformation
to a high-dimensional space and $\left\langle \cdot,\cdot\right\rangle $
denotes the inner product operator. 

Centering of the points in the \emph{high-dimensional} space is achieved
by
\begin{equation}
k_{ij}=k_{ij}-\frac{1}{N}\sum_{r=1}^{N}k_{rj}-\frac{1}{N}\sum_{s=1}^{N}k_{is}+\frac{1}{N^{2}}\sum_{r=1}^{N}\sum_{s=1}^{N}k_{rs}.\label{eq:centering}
\end{equation}

An eigen-decomposition of $K$ is performed to produce a set of eigenvalues
$\lambda_{1}\ge\lambda_{2}\ge\ldots\ge\lambda_{N}$ and their corresponding
eigenvectors $\psi_{1},\ldots,\psi_{N}$. The embedding of $x_{i}$
into $\mathbb{R}^{q}$ is given by:
\[
\widetilde{x}_{i}\triangleq\left(\frac{1}{\sqrt{\lambda_{1}}}\left\langle \psi_{1},k\left(\cdot,x_{i}\right)\right\rangle ,\frac{1}{\sqrt{\lambda_{2}}}\left\langle \psi_{2},k\left(\cdot,x_{i}\right)\right\rangle ,\ldots,\frac{1}{\sqrt{\lambda_{q}}}\left\langle \psi_{q},k\left(\cdot,x_{i}\right)\right\rangle \right)
\]
where $k\left(\cdot,x_{i}\right)$ is the $i$-th column of the matrix
$K$.

Choosing the kernel that achieves the best dimensionality reduction
for a given dataset is an open problem and is highly dependent on
the dataset at hand (this property is shared with the diffusion map
algorithm which is described in Chapter \ref{cha:Diffusion-Maps}).
Common choices for kernels are linear kernel in which case KPCA coincides
with PCA, Gaussian kernels \cite{KPCA_book02,Kernel04}, Polynomial
kernels \cite{KPCA_book02,Kernel04} and hyperbolic tangent \cite{KPCA_book02}
which is common in neural networks.

Successful applications of KPCA include speech recognition \cite{KPCA_speech04},
novelty detection \cite{KPCA_novelty07} and face detection \cite{KPCA_face02}.
The size of the kernel matrix is $N\times N$ which renders its storage
and calculation infeasible for large datasets. This limitation can
be relieved by constructing a sparse kernel \cite{KPCA_sparse00}.

\subsection{Multidimensional Scaling (MDS)}

MDS \cite{MDS64,MDS_94} algorithms find an embedding that best preserves
the inter-point distances among the vectors in $\Gamma$. The input
to the algorithms is a matrix that contains all pair-wise distances
or dissimilarities. The output is a set of points in a low-dimensional
space - usually, the Euclidean space.

The embedding is facilitated by minimizing a loss/cost function $C\left(\widetilde{\Gamma}\right)$
which is also known as a \emph{stress function}. This function measures
the error between the pairwise distances among the vectors in $\Gamma$
and $\widetilde{\Gamma}$. A common cost function is the \emph{raw}
stress function which is given by
\[
C_{raw}\left(\widetilde{\Gamma}\right)=\sum_{i=1}^{N}\sum_{j=1}^{N}\left(d\left(x_{i},x_{j}\right)-d\left(\widetilde{x}_{i},\widetilde{x}_{j}\right)\right)^{2}
\]
where $d\left(\cdot,\cdot\right)$ is the used distance measure.

Another choice for a cost function is the \emph{Sammons} cost function
which is defined as
\[
C_{Sammons}\left(\widetilde{\Gamma}\right)=\frac{1}{\sum_{i=1}^{N}\sum_{j=1}^{N}d\left(x_{i},x_{j}\right)}\sum_{i=1}^{N}\sum_{j=1}^{N}\frac{\left(d\left(x_{i},x_{j}\right)-d\left(\widetilde{x}_{i},\widetilde{x}_{j}\right)\right)^{2}}{d\left(x_{i},x_{j}\right)}.
\]

The minimization of the stress function is obtained by applying common
optimization methods such as the conjugate gradient method, the pseudo-Newton
technique \cite{MDS_94} or by performing spectral decomposition of
the pairwise dissimilarity matrix. \emph{Metric multidimensional scaling}
algorithms generalize the optimization procedure to a variety of stress
functions and weighted input matrices - however, the result is still
given in Euclidean space. 

MDS enables the visualization of high-dimensional datasets when the
embedding space is $\mathbb{R}^{2}$ or $\mathbb{R}^{3}$. This property
was used for numerous applications such as fMRI data analysis \cite{MDSfMRI98}
and molecular modeling \cite{MDS_VB04} to name a few. Variations
of the MDS algorithm have been proposed. These include Stochastic
Proximity Embedding (SPE) \cite{MDS_SPE03}, FastMap \cite{MDS_FastMap95}
and Stochastic Neighbor Embedding (SNE) \cite{SNE02}. In fact, the
ISOMAP algorithm, which is described below, may be considered as a
special case of metric multidimensional scaling where the geodesic
distance measure constitutes the metric.

\subsubsection{\emph{Generalized multidimensional scaling }(GMDS)}

\emph{Generalized multidimensional scaling} \cite{GMDS1}\emph{ }generalizes
the classical Metric multidimensional scaling by allowing any valid
metric space to be used for the output. One of the important uses
facilitated by this extension is the representation of the intrinsic
metric structure of one surface in terms of the intrinsic geometry
of another (also known as the isometric representation problem). Furthermore,
using a distance measure%
\footnote{It is not exactly a distance measure since it does not possess all
the required properties of a distance measure.%
} that is derived from the Gromov-Hausdorff distance \cite{GH}, one
can compute the partial embedding distance between two surfaces, thus
facilitating the partial matching of surfaces. GMDS also allows to
find local differences between two shapes.

Formally, let $\Gamma^{1}=\left\{ x_{i}^{1}\right\} _{i=1}^{N_{1}}$
and $\Gamma^{2}=\left\{ x_{i}^{2}\right\} _{i=1}^{N_{2}}$ be given
two datasets representing the sampled surfaces of two continuous surfaces
$S$ and $Q$, respectively. The geodesic distances between the samples
$\left\{ x_{i}^{1}\right\} $ and $\left\{ x_{i}^{2}\right\} $ are
given by the $N_{1}\times N_{1}$ and $N_{2}\times N_{2}$ matrices
$D_{\Gamma^{1}}=\left(d_{S}\left(x_{i}^{1},x_{j}^{1}\right)\right)$
and $D_{\Gamma^{2}}=\left(d_{Q}\left(x_{i}^{2},x_{j}^{2}\right)\right)$,
respectively.

The general stress function is defined as
\[
C\left(U;D_{\Gamma^{2}},d_{S},W\right)=\left(\frac{1}{\sum_{j>i}w_{ij}}\sum_{j>i}\left(w_{ij}\left(d_{S}\left(u_{i},u_{j}\right)-d_{Q}\left(x_{i}^{2},x_{j}^{2}\right)\right)\right)^{p}\right)^{\frac{1}{p}}
\]
for $1\le p<\infty$ and
\[
C\left(U;D_{\Gamma^{2}},d_{S},W\right)=\max_{i,j=1,\ldots,N_{2}}w_{ij}\left|d_{S}\left(u_{i},u_{j}\right)-d_{Q}\left(x_{i}^{2},x_{j}^{2}\right)\right|
\]
for $p=\infty$ where the matrix $U$ represents the positions of
$N_{2}$ points on $S$ in some local or global parametric coordinates
$u_{i}$, and $W=\left(w_{ij}\right)$ is a symmetric matrix of nonnegative
weights. It should be noted that (a) the metric in the embedding space
is not given analytically and is approximated numerically and (b)
the $L_{2}$ norm of the classical MDS algorithm is generalized to
the $L_{p}$ norm.

\subsection{Summary}

Classical techniques for dimensionality reduction such as PCA and
MDS, are simple to implement and can be efficiently computed. However,
the pitfall of these methods is that they are \emph{global} i.e. they
take into account the distances between \emph{all} pairs of points.
This makes them highly sensitive to noise and outliers since the embedding
requires to take into account all the outlier and noise. By taking
into account the outliers, the embedding may deviate from the normal
data. This pitfall is amended by local methods for dimensionality
reduction which are described next. 

Beyer \emph{et al}. \cite{Beyer99} show that under a wide variety
of conditions, inspection of nearest neighbors is not meaningful.
Specifically, they show that as the dimension increases the difference
between the distance to the nearest neighbor and the farthest neighbor
diminishes and thus renders the nearest neighbor notion practically
meaningless. Consequently, using distance as a similarity measurement
must be used with caution. This phenomenon may occur even at 10 to
15 dimensions.

\section{Local methods\label{sub:Local-methods}}

Local-information-preserving dimensionality reduction methods (which
we refer to as \emph{local} methods in short) are based on the assumption
that the only relevant information lies in local distance measurements.
These measurements vary and are usually a function of the Euclidean
distance (see below). Accordingly, they look for a low-dimensional
embeddings that preserve only local properties of the high-dimensional
set. Utilizing the Newtonian paradigm, these methods derive the global
geometry of the dataset by aggregating the local information i.e.
consider for each point only the information which is inherent in
its closest neighboring points. By doing so, they can identify and
thus give less importance to outliers and thereby produce an embedding
that better fits the original dataset. In this sense, local methods
amend the pitfall of global methods. 

An interesting property of local dimensionality methods is that they
can be formulated using the KPCA framework. This is facilitated by
tailoring an appropriate kernel function to each of the methods.

\subsection{Laplacian Eigenmaps}

Laplacian Eigenmaps \cite{Laplacian03} aim at embedding low dimensional
manifolds that reside in a high-dimensional ambient space. It finds
a low-dimensional embedding that best preserves the distance from
\emph{each point only to its closest neighbors}. This is in contrast
to MDS where the embedding attempt to minimize \emph{all} pair-wise
distances. 

The algorithm consists of three steps. First, a graph is constructed
on $\Gamma$ where each graph vertex corresponds to a data point.
Edges connect each point (vertex) only to its closest neighbors. The
closest neighbors of a point $x_{i}$ can be chosen either as its
$k$ nearest neighbors or they can be chosen as the points in the
$n$-dimensional hyper-ball of radius $\varepsilon$ centered at $x_{i}$.
Depending on the neighborhood type that is chosen, either $k$ or
$\varepsilon$ is given as a parameter to the algorithm. 

Second, a weight function $W$ is chosen for the edges of the graph.
This function can be the Gaussian kernel, which is also referred to
as the heat kernel ($t\in\mathbb{R}$ is given as a parameter):
\[
W_{ij}=\left\{ \begin{array}{cc}
\exp\left(\left\Vert x_{i}-x_{j}\right\Vert ^{2}\left/t\right.\right) & if\, x_{i}\, and\, x_{j}\, are\, connected\\
0 & otherwise
\end{array}.\right.
\]
Alternatively, in order to avoid the parameter $t$, the weight function
can be a simple adjacency function, i.e.
\[
W_{ij}=\begin{cases}
\begin{array}{cc}
1 & if\, x_{i}\, and\, x_{j}\, are\, connected\\
0 & otherwise
\end{array} & .\end{cases}
\]
It was shown in \cite{Laplacian03} that when $\Gamma$ is approximately
lying on a submanifold, choosing $W$ to be the heat kernel corresponds
to an approximation of the heat kernel on the submanifold.

The third and last step of the algorithm minimizes the cost function
\[
C\left(\Gamma,\widetilde{\Gamma}\right)=\sum_{i=1}^{N}\sum_{j=1}^{N}\left\Vert \widetilde{x}_{i}-\widetilde{x}_{j}\right\Vert ^{2}W_{ij}.
\]
Note that minimizing $C\left(\Gamma,\widetilde{\Gamma}\right)$ ensures
that neighboring points $x_{i}$ and $x_{j}$ are mapped to close
points since the choice of weights $W_{ij}$ penalizes $C\left(\Gamma,\widetilde{\Gamma}\right)$
otherwise. The solution to this minimization problem is obtained by
solving the following generalized eigenvector problem:
\[
L\varphi=\lambda D\varphi
\]
where $D$ is the degree matrix of the graph which is defined as $D_{ii}=\sum_{j=1}^{N}W_{ij}$
and $L$ is the graph Laplacian which is given by $L=D-W.$ The eigenvalues
of the solution are denoted by $\lambda_{1}\le\lambda_{2}\le\ldots\le\lambda_{N}$
and their corresponding eigenvectors are given by $\varphi_{1},\ldots,\varphi_{N}$.
The embedding of a point $x_{i}$ into $\mathbb{R}^{q}$ is defined
as
\[
\widetilde{x_{i}}=\left(\varphi_{1}\left(i\right),\varphi_{2}\left(i\right),\ldots,\varphi_{q}\left(i\right)\right),\, i=1,\ldots N
\]
where $\varphi_{k}\left(i\right)$ denotes the $i$-th coordinate
of the $k$-th eigenvector. 

Examples where Laplacian eigenmaps have been successfully applied
include face recognition \cite{Laplacianfaces05}, clustering \cite{LaplacianClutering01}
and image segmentation \cite{weiss99,SM00}.

\subsection{ISOMAP}

\begin{figure}[!t]
\begin{centering}
\includegraphics[%bb=102bp 300bp 480bp 550bp,
width=3in]{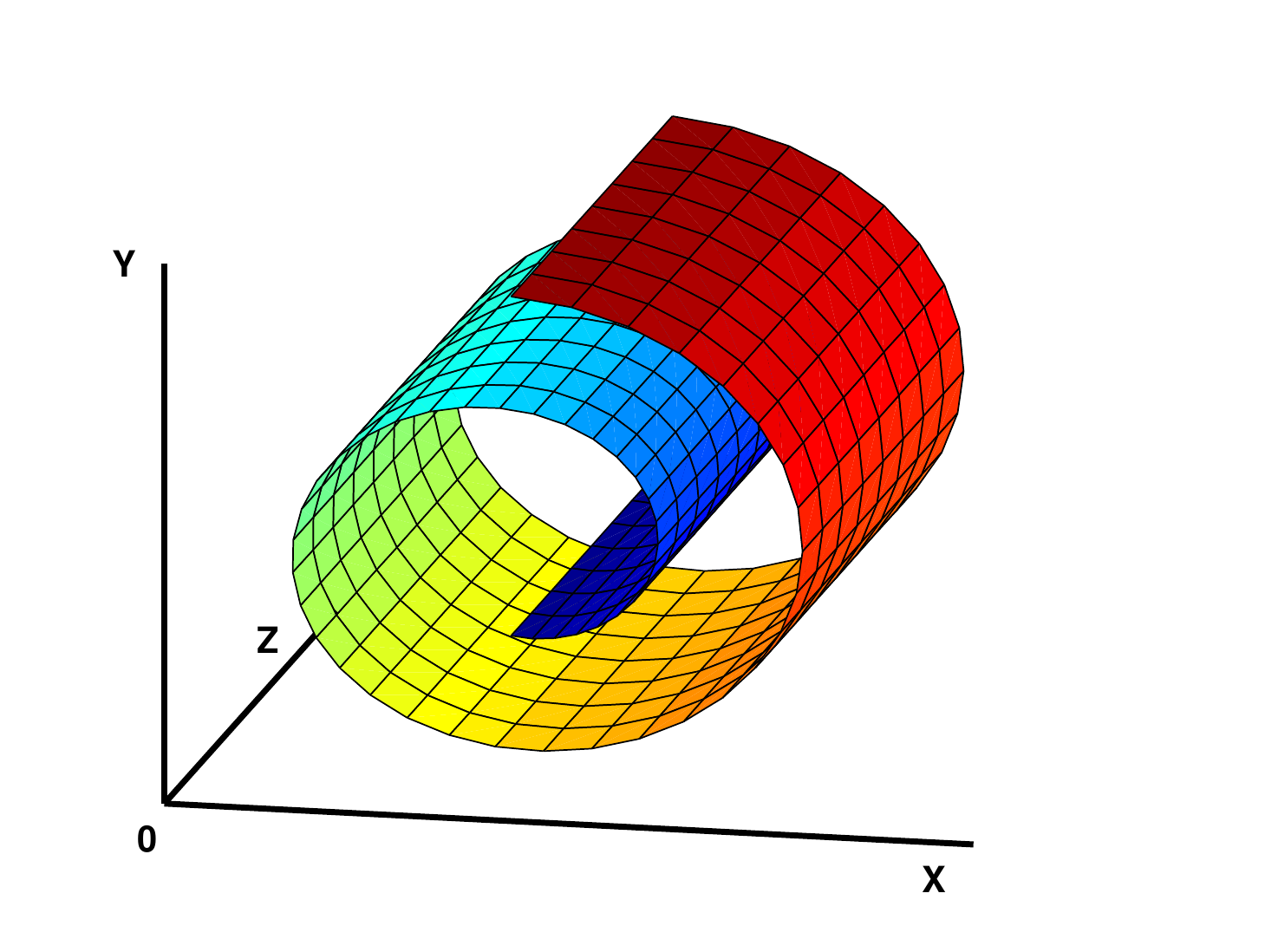}
\par\end{centering}

\caption{A 3D plot of a Swiss Roll manifold.\label{cap:SwissRoll}}
\end{figure}

The main drawback of MDS is that it produces an embedding that tries
to preserve all pair-wise \emph{Euclidean} distances as captured in
the high-dimensional space. Using the Euclidean distance may lead
to an embedding that does not reflect the geometry of the manifold
since a small Euclidean distance between a pair of points does not
imply they are close over the manifold. An example where this phenomenon
occurs is the Swiss roll manifold \cite{SwissRoll} which is displayed
in Fig. \ref{cap:SwissRoll}. Consequently, a dimensionality reduction
algorithm that preserves the pair-wise distances \emph{over the manifold}
will produce an embedding that conforms better with the geometry of
the manifold. This is exactly what ISOMAP \cite{ISO00} does. Specifically,
ISOMAP applies MDS using the \emph{geodesic distance} measure instead
of the Euclidean one. The geodesic distance between a pair of points
is defined as the length of the shortest path connecting these points
that passes only through points on the manifold. 

Initially, a sparse graph whose vertices correspond to the data points
in $\Gamma$ is constructed where edges connect each vertex to its
$k$ nearest neighbors. The pair-wise geodesic distance matrix is
then \emph{approximated} using either the Dijkstra \cite{Dijkstra}
or the Floyd-Warshall \cite{FloydWarshall} algorithms. Since the
graph is sparse, the pair-wise geodesic distances that is produced
provides only an estimate to the exact geodesic distances. The low-dimensional
embedding is obtained by applying MDS to the pair-wise geodesic distances
matrix.

Two main shortcomings of ISOMAP are its topological instability and
its requirement for the original dataset to be convex. Its performance
is degraded in the presence of graph short-circuits (also known as
loops) and holes in the manifold \cite{ISO_bad05}. Manifolds that
contain holes are dealt with by dividing the manifold into sub-manifolds
without holes \cite{ISO_bad05}. Graph loops can be overcome by removal
of nearest neighbors that do not conform with a predefined local linearity
constraint \cite{ISO_04}. Another approach for handling graph loops
is to remove data points which have high total flow in the shortest
path algorithms \cite{ISO_07}.

\subsection{Local Linear Embedding (LLE)}

LLE \cite{LLE00} assumes that the manifold is locally linear. Accordingly,
it approximates the geometry of a manifold by fitting hyperplanes
to local neighborhoods of points i.e. the manifold is expressed as
an aggregation of locally linear patches. To achieve this, every point
is formulated as a linear combination of its $k$ nearest neighbors,
where $k$ is given as a parameter to the algorithm. The output of
this formulation is a $N\times k$ weight matrix $W$. On one hand,
an embedding that maps the high-dimensional data into a low-dimensional
space can be approximated by a linear mapping. On the other hand,
the weights in $W$ describe intrinsic geometric properties of the
data that are invariant to linear transformations. Combining these
facts implies that the same weights will hold for the low-dimensional
embedding points as well. Thus, the second step of the algorithm finds
the low-dimensional embedding of the points which best preserves the
weights of the high-dimensional linear combinations. Thus, the LLE
algorithm consists of two steps where each step solves an optimization
problem.

The first step finds the linear combination weights $\left\{ W_{ij}\right\} _{i=1,\ldots,N;\, j=1,\ldots,k}$
that minimize the following cost function 
\[
C\left(W\right)=\sum_{i=1}^{N}\left|x_{i}-\sum_{j=1}^{k}W_{ij}x_{j}\right|^{2}
\]
where the constraint $\sum_{j=1}^{k}W_{ij}=1$ is imposed for every
$i=1,\ldots,N$. 

The second step finds the low-dimensional embedding points $\left\{ \widetilde{x}_{i}\right\} $
which minimize the cost function
\[
C\left(\widetilde{\Gamma}\right)=\sum_{i=1}^{N}\left|\widetilde{x}_{i}-\sum_{j=1}^{k}W_{ij}\widetilde{x}_{i}\right|^{2}
\]
with respect to the weights $\left\{ W_{ij}\right\} $ that were found
in the first step.

The LLE algorithm successfully embeds non-convex manifold as opposed
to the ISOMAP algorithm and is also more robust to short-circuits
than ISOMAP. However, LLE performance is degraded when it is applied
on manifolds that include holes \cite{LLE_bad02,LLE_bad03}. Nevertheless,
LLE has been successfully applied to image colorization \cite{LLE_colorization05},
sound source localization \cite{LLE_sound_app05} and to image super-resolution
\cite{LLE_superres04}.

\subsection{Hessian LLE (HLLE)}

The Hessian LLE or Hessian eigenmaps \cite{Hessian02} algorithm extends
the class of datasets whose dimensionality can be successfully reduced
by ISOMAP. Specifically, the convexity requirement is relaxed. This
method is based on the underlying assumption that the manifold is
locally isometric to an open, connected subset in the low-dimensional
space. In order to reduce the dimensionality of $\Gamma$ a $\mathcal{H}$-functional,
which measures the average \emph{curviness} over the manifold, is
introduced. The $\mathcal{H}$-functional is derived by aggregating
all point-wise local Hessian estimations. The Hessian is required
to be invariant to the position of the point and therefore is represented
in the local tangent space. The low-dimensional embedding is obtained
by minimization of $\mathcal{H}\left(\Gamma\right)$ with respect
to the Frobenius norm. 

The embedding algorithm consists of the following steps: first, the
basis of the local tangent space is calculated at each point by applying
PCA to its $k$ nearest neighbors. Next, a matrix $L_{i}$ whose columns
comprise of a column of ones and all cross products of the basis vectors
up to the $q$-th order, is constructed. Gram-Schmidt orthogonalization
is applied to $L_{i}$ and the representation of the local Hessian
$H_{i}^{\left(tan\right)}$ in the tangent space is estimated using
the last $\frac{q\left(q+1\right)}{2}$ columns of $L_{i}$. Then,
the $\mathcal{H}$-functional is constructed as $\mathcal{H}\left(\Gamma\right)=\sum_{i=1}^{N}\left\Vert H_{i}^{\left(tan\right)}\right\Vert _{F}^{2}$
where $\left\Vert \cdot\right\Vert _{F}$ is the Frobenius norm. This
functional is minimized via its eigen-decomposition in order to approximate
its null space. The $\left(q+1\right)$-dimensional subspace corresponding
to the $q+1$ smallest eigenvalues are extracted. There will be a
zero eigenvalue that is associated with the subspace of constant functions.
The next $q$ eigenvalues correspond to eigenvectors spanning a $q$-dimensional
space $S_{q}$ where our embedding coordinates are to be found. Finally,
a basis $\varphi_{1},\ldots,\varphi_{q}$ for $S_{q}$ is selected.
This basis must constitute an orthonormal basis when it is restricted
to a specific fixed neighborhood (which may be chosen arbitrarily
from those used in the algorithm). The embedding of a point $x_{i}$
into $\mathbb{R}^{q}$ is obtained as
\[
\widetilde{x_{i}}=\left(\varphi_{1}\left(i\right),\varphi_{2}\left(i\right),\ldots,\varphi{}_{q}\left(i\right)\right),\, i=1,\ldots,N.
\]

\subsection{Local Tangent Space Analysis (LTSA)}

LTSA \cite{LocalTanSpAnal02} shares with HLLE the property of using
the local tangent space in order to discover and represent local properties
of $\Gamma$. It assumes local linearity of the manifold. This assumption
implies that a high-dimensional point $x_{i}$ and its corresponding
low-dimensional point $\widetilde{x}_{i}$ can be mapped to the same
local tangent space via two linear transformations. LTSA looks for
both a low-dimensional embedding and a linear transformation the maps
the low dimensional embedding into the local tangent space of $\Gamma$.

Initially, in a similar manner to HLLE, the basis of the local tangent
space $\Theta_{i}$ is calculated at each point by applying PCA to
its $k$ nearest neighbors. We denote by $\widetilde{\Gamma}_{i}$
the low-dimensional embedding of the points in the neighborhood of
$x_{i}$. Using the property from the last paragraph, an embedding
$\widetilde{\Gamma}_{i}$ and a linear transformation $T_{i}$ that
minimize $\sum_{i=1}^{N}\left\Vert \widetilde{\Gamma}_{i}C_{k}-T_{i}\Theta_{i}\right\Vert _{F}^{2}$
are calculated where $C_{k}$ is the centering matrix of size $k$
and $\left\Vert \cdot\right\Vert _{F}$ is the Frobenius norm.

Successful applications of LTSA include local smoothing of manifolds
\cite{LTSA_smoothapp04} and dimensionality reduction of micro-arrays
\cite{LTSA_app_microarray05}.

\subsection{Random projections}

Random projection \cite{D06} is a technique whose theoretical foundation
stems from the works of Johnson and Lindenstrauss \cite{JL84} and
Bourgain \cite{B85} as referenced in the beginning of Section \ref{sec:Previous-work}.
In order to reduce the dimensionality of $\Gamma$ using random projections,
a random basis $\Upsilon=\left\{ \rho_{i}\right\} _{i=1}^{n}$ is
first generated where $\rho_{i}\in\mathbb{R}^{q}$. Two common choices
for generating a random basis are:
\begin{enumerate}
\item The vectors $\left\{ \rho_{i}\right\} _{i=1}^{n}$ are uniformly distributed
over the $q$ dimensional unit sphere. 
\item The elements of the vectors $\left\{ \rho_{i}\right\} _{i=1}^{n}$
are chosen from a Bernoulli +1/-1 distribution and the vectors are
normalized so that $\left\Vert \rho_{i}\right\Vert _{l_{2}}=1$ for
$i=1,\ldots,n$. 
\end{enumerate}
Next, a $q\times n$ matrix $R$ whose columns are composed of the
vectors in $\Upsilon$, is constructed. The embedding $\widetilde{x}_{i}$
of $x_{i}$ is obtained by 
\[
\widetilde{x}_{i}=R\cdot x{}_{i}
\]

\subsection{Multilayer autoencoders}

Multilayer autoencoders \cite{NeuralNets93,NeuralNets06} use feed-forward
neural networks which have an odd number of hidden layers. These networks
are tailored for dimensionality reduction by setting the input and
output layers to have $n$ nodes ( recall that $n$ is the dimension
of the original dataset $\Gamma)$ while setting the middle layer
to contain $q$ nodes ($q$ is the dimension of the embedding space).
Accordingly, $q$ must be given as input to the algorithm. The data
points in $\Gamma$ are input to the network and the network is trained
so that the mean square error between the input and the output is
minimized. The embedding of a data point is determined according to
the values of the nodes in the middle layer when it is fed to the
network. Using linear neurons produces similar results to PCA \cite{PCA_adaptive_kung94}
and classical MDS \cite{MDS_94} while non-linear dimensionality reduction
is obtained by using non-linear activation functions such as Sigmoids.

Backpropagation is usually a poor approach for training multilayer
autoencoders since it converges slowly and is sensitive to local minima
due to the high number of connections in the autoencoder. In order
to overcome this difficulty, several approaches were proposed. Restricted
Boltzmann Machines \cite{NeuralNetsBoltzmann06} were used in \cite{NeuralNets06}
for pre-training using simulated annealing \cite{NeuralNetsSimulatedAnneal83}.
Following the pre-training, backpropagation was applied to the network
in order to fine-tune its weights. Genetic algorithms can also be
used for training multilayer autoencoders \cite{NeuralNetsGenetic75,NeuralNetsGenetic00}.

\subsection{Self-Organizing Maps}

Self-Organizing Maps \cite{SOM01,SOM_App1,SOM_App2} are a neural
network algorithm which embeds the high dimensional data into a 2D
or 3D space. The low-dimensional embedding is called a \emph{map}
and one of the primary uses of this method is visualization of high-dimensional
data. The algorithm uses a deformable template to translate data similarities
into spatial relationships. The template consists of nodes in a grid
formation (hexagonal or rectangular) which discretisize the low-dimensional
space. Each node is associated with a position in the map space and
a weight vector of the same dimension as the input data vectors. 

The training consists of an iterative procedure that performs the
following. For each training vector $v$, find the node whose weight
is the closest to $v$ according to the Euclidean distance. Update
its weights and its map neighbors $u$ so they become closer to $v$:
\[
w_{t+1}\left(u\right)=w_{t}\left(u\right)+\Theta\left(u,t\right)\alpha\left(t\right)\left(v-u\right)
\]
where $\alpha\left(t\right)$ is a monotonically decreasing learning
coefficient and $\Theta\left(u,t\right)$ is a neighborhood function
that depends on the grid distances between $u$ and closest node to
$v$.

Embedding of a vector $x$ to the map is done by assigning the map
coordinates of the node whose weight vector is the closest to $x$.

\subsection{Generative Topographic Mapping }

Generative topographic map (GTM) \cite{GTM,GTM_App1,GTM_App2} is
a probabilistic method that, similarly to SOM, embeds high-dimensional
data in 2D or 3D space using a discrete grid to describe the latent
space. The algorithm assumes that the high-dimensional data was generated
by a smooth non-linear mapping of low-dimensional data into the high-dimensional
space and addition of Gaussian noise to it. The Gaussian noise assumption
renders the model a constrained mixture of Gaussians. The nonlinear
mapping using a radial basis function network (RBF). Given the training
data, the parameters of the mapping, the noise and the low-dimensional
probability distribution are learned using the expectation-maximization
(EM) algorithm.

\subsection{Other methods}

One can find additional methods for dimensionality reduction which
can be regarded as variants of the above methods. These include Kernel
Maps \cite{KernelMaps}, Conformal Eigenmaps \cite{ConformalEigenmaps},
Principal Curves \cite{PrincipalCurves}, Geodesic Nullspace Analysis
\cite{GeodesicNullAnal}, Maximum Variance Unfolding \cite{MaxVarUnfold},
Stochastic Proximity Embedding \cite{MDS_SPE03}, FastMap \cite{MDS_FastMap95},
Locality Preserving Projection \cite{LocalPreservProj} and various
methods that perform alignment of local linear models \cite{LLC_org,LLC_var1,LLC_var2,LLC_var3}.

The diffusion maps \cite{CL_DM06} algorithm is another local method
for dimensionality reduction which is of special interest to this
thesis since it is closely related to our diffusion basis \cite{mine-DB07}
scheme (Chapter \ref{cha:Diffusion-bases}). Therefore, we provide
a detailed description of the DM algorithm in the next chapter.

\chapter{The Diffusion Maps (DM) Algorithm\label{cha:Diffusion-Maps}}

Recently, Coifman and Lafon \cite{CL_DM06} introduced the \emph{Diffusion
Maps (DM)} scheme which is used for manifold learning. DM embed high
dimensional data into a Euclidean space of substantially smaller dimension
while preserving the geometry of the dataset. The global geometry
is preserved by maintaining the local neighborhood geometry of each
point in the dataset - a property that classifies it as a local method
(see Section \ref{sub:Local-methods}). However, DM use a random walk
distance that is more robust to noise since it takes into account
all the paths connecting a pair of points. Furthermore, DM can provide
parametrization of the data when only a point-wise similarity matrix
is available - a property it shares with MDS. This may occur either
when there is no access to the original data or when the original
data consists of abstract objects. 

Given a set of data points as defined in Section \ref{eq:gamma},
the DM algorithm includes the following steps:
\begin{enumerate}
\item Construction of an undirected graph $G$ on $\Gamma$ with a weight
function $w_{\varepsilon}$ that corresponds to the \emph{local} point-wise
similarity between the points in $\Gamma$%
\footnote{$G$ is sparse since only the points in the local neighborhood of
each point are considered. Wider neighborhood are explored via a diffusion
process. %
}. In case we are only given $w_{\varepsilon},$ this step is skipped. 
\item Derivation of a random walk on the graph $G$ via a Markov transition
matrix $P$ that is derived from $w_{\varepsilon}$. 
\item Eigen-decomposition of $P$. 
\end{enumerate}
By designing a local geometry that reflects quantities of interest,
it is possible to construct a diffusion operator whose eigen-decomposition
enables the embedding of $\Gamma$ into a space $Y$ of substantially
lower dimension. The Euclidean distance between a pair of points in
the reduced space defines a diffusion metric that measures the proximity
of points in terms of their connectivity in the original space. Specifically,
the Euclidean distance between a pair of points, in $Y$, is equal
to the random walk distance between the corresponding pair of points
in the original space.

The eigenvalues and eigenfunctions of $P$ define a natural embedding
of the data through the diffusion map. Furthermore, the study of the
eigenvalues allows us to use the eigenfunctions for dimensionality
reduction.

\section{Building the graph $G$ and the weight function $w_{\varepsilon}$\label{sub:Building-the-graph}}

Let $\Gamma$ be a set of points in $\mathbb{R}^{n}$ as defined in
Eq. \ref{eq:gamma}. We construct the graph $G(V,E),\,\left|V\right|=N,\,\left|E\right|\ll N^{2}$,
on $\Gamma$ in order to study the intrinsic geometry of this set.
A weight function $w_{\varepsilon}\left(x_{i},x_{j}\right)$, which
measures the pairwise similarity between the points, is introduced.
For all $x_{i},x_{j}\in\Gamma$, the weight function has the following
properties:
\begin{itemize}
\item symmetry: $w_{\varepsilon}\left(x_{i},x_{j}\right)=w_{\varepsilon}\left(x_{j},x_{i}\right)$ 
\item non-negativity: $w_{\varepsilon}\left(x_{i},x_{j}\right)\ge0$ 
\item fast decay: given a scale parameter $\varepsilon>0$, $w_{\varepsilon}\left(x_{i},x_{j}\right)\rightarrow0$
when $\left\Vert x_{i}-x_{j}\right\Vert \gg\varepsilon$ and $w_{\varepsilon}\left(x_{i},x_{j}\right)\rightarrow1$
when $\left\Vert x_{i}-x_{j}\right\Vert \ll\varepsilon$. The sparsity
of $G$ is a result of this property. 
\end{itemize}
Note that the parameter $\varepsilon$ defines a notion of neighborhood.
In this sense, $w_{\varepsilon}$ defines the local geometry of $\Gamma$
by providing a first-order pairwise similarity measure for $\varepsilon$-neighborhoods
of every point $x_{i}$. Higher order similarities are derived through
a diffusion process. A common choice for $w_{\varepsilon}$ is the
Gaussian kernel 
\begin{equation}
w_{\varepsilon}\left(x_{i},x_{j}\right)=\exp\left(-\frac{\left\Vert x_{i}-x_{j}\right\Vert ^{2}}{2\varepsilon}\right).\label{eq:gaussian}
\end{equation}
However, other weight functions can be used and the choice of the
weight function essentially depends on the application at hand. An
automatic procedure for choosing $\varepsilon$ is described in Section
\ref{sec:Choosing_epsilon}.

\section{Construction of the normalized graph Laplacian}

The non-negativity property of $w_{\varepsilon}$ allows to normalize
it into a Markov transition matrix $P$ where the states of the corresponding
Markov process are the data points. This enables to analyze $\Gamma$
via a random walk.

Formally, $P=\left(p\left(x_{i},x_{j}\right)\right)_{i,j=1,\dots,N}$
is constructed as follows:

\begin{equation}
p\left(x_{i},x_{j}\right)=\frac{w_{\varepsilon}\left(x_{i},x_{j}\right)}{d\left(x_{i}\right)}\label{eq:markov_normalization}
\end{equation}
 where 
\begin{equation}
d\left(x_{i}\right)=\sum_{j=1}^{N}w_{\varepsilon}\left(x_{i},x_{j}\right)\label{eq:degree}
\end{equation}
 is the degree of $x_{i}$. If we let $D=\left(d_{ij}\right)$ be
a $N\times N$ diagonal matrix where $d_{ii}=d\left(x_{i}\right)$,
and we let $W_{\varepsilon}$ be the weight matrix that corresponds
to the weight function $w_{\varepsilon}$, $P$ can be derived as
\begin{equation}
P=D^{-1}W_{\varepsilon}.\label{eq:P_mat}
\end{equation}
$P$ is a Markov matrix since it is row stochastic i.e. the sum of
each row in $P$ is 1 and $p\left(x_{i},x_{j}\right)\ge0$, $i,j=1,\ldots,N$.
Thus, $p\left(x_{i},x_{j}\right)$ can be viewed as the probability
to move from $x_{i}$ to $x_{j}$ in \emph{a single} time step. By
raising this quantity to a power $t$, this influence is propagated
to nodes in the neighborhood of $x_{i}$ and $x_{j}$ and the result
is the probability for this move in $t$ time steps. We denote this
probability by $p_{t}\left(x_{i},x_{j}\right)$ and it can be defined
by induction:
\[
p_{t}\left(x_{i},x_{j}\right)=\left\{ \begin{array}{cc}
p\left(x_{i},x_{j}\right) & t=1\\
\sum_{k=1}^{N}p_{t-1}\left(x_{i},x_{k}\right)p_{t-1}\left(x_{k},x_{j}\right) & t>1
\end{array}\right.
\]

These probabilities measure the connectivity of the points within
the graph. The parameter $t$ controls the scale of the neighborhood
in addition to the scale control provided by $\varepsilon$.

\section{Eigen-decomposition\label{sub:Spectral-decomposition}}

The close relation between the asymptotic behavior of $P$, i.e. the
properties of its eigen-decomposition and the clusters that are inherent
in the data, was explored in \cite{C97,fowlkes04spectral}. We denote
the left and the right bi-orthogonal eigenvectors of $P$ by $\left\{ \mu_{k}\right\} _{k=1,\dots,N}$
and $\left\{ \nu_{k}\right\} _{k=1,\dots,N}$, respectively. Let $\left\{ \lambda_{k}\right\} _{k=1,\dots,N}$
be the eigenvalues of $P$ where $\left|\lambda_{1}\right|\ge\left|\lambda_{2}\right|\ge...\ge\left|\lambda_{N}\right|$.

It is well known that $\lim_{t\rightarrow\infty}p_{t}\left(x_{i},x_{j}\right)=\mu_{1}\left(x_{j}\right).$
Coifman \emph{et al}. \cite{CLLMNWZ05} show that for a finite time
$t$ we have
\begin{equation}
p_{t}\left(x_{i},x_{j}\right)=\sum_{k=1}^{N}\lambda_{k}^{t}\nu_{k}\left(x_{i}\right)\mu_{k}\left(x_{j}\right).\label{eq:eigen_decomp}
\end{equation}
An appropriate choice of $\varepsilon$ achieves a fast decay of $\left\{ \lambda_{k}\right\} $.
Thus, to achieve a relative accuracy $\delta>0$, only a few terms
$\eta\left(\delta\right)$ are required in the sum in Eq. \ref{eq:eigen_decomp}.
By accuracy, we mean that $\eta\left(\delta\right)$ is chosen such
that 
\begin{equation}
\sqrt{\sum_{i,j}\left(p_{t}\left(x_{i},x_{j}\right)-\sum_{k=1}^{\eta\left(\delta\right)}\lambda_{k}^{t}\nu_{k}\left(x_{i}\right)\mu_{k}\left(x_{j}\right)\right)^{2}}<\delta.\label{eq:choosing-eta}
\end{equation}
Due to the fast decay of $\left\{ \lambda_{k}\right\} $, we will
typically have $\eta\left(\delta\right)\ll N.$ A simple linear search
procedure is used to calculate $\eta\left(\delta\right).$

Coifman and Lafon \cite{CL_DM06} introduced the \emph{diffusion distance}
\begin{equation}
D_{t}^{2}\left(x_{i},x_{j}\right)=\sum_{k=1}^{N}\frac{\left(p_{t}\left(x_{i},x_{k}\right)-p_{t}\left(x_{k},x_{j}\right)\right)^{2}}{\mu_{1}\left(x_{k}\right)}.\label{eq:diffusion_distance1}
\end{equation}
This formulation is derived from the known random walk distance in
Potential Theory: $D_{t}^{2}\left(x_{i},x_{j}\right)=p_{t}\left(x_{i},x_{i}\right)+p_{t}\left(x_{j},x_{j}\right)-2p_{t}\left(x_{i},x_{j}\right)$
where 2 is due to the fact that the graph $G$ is undirected.

Averaging along all the paths from $x_{i}$ to $x_{j}$ results in
a distance measure that is more robust to noise and topological short-circuits
than the geodesic distance or the shortest-path distance which are
used in the ISOMAP algorithm \cite{ISO00}. Finally, the diffusion
distance can be expressed in terms of the right eigenvectors of $P$
(see \cite{LKC06} for a proof):
\begin{equation}
D_{t}^{2}\left(x_{i},x_{j}\right)=\sum_{k=2}^{N}\lambda_{k}^{2t}\left(\nu_{k}\left(x_{i}\right)-\nu_{k}\left(x_{j}\right)\right)^{2}.\label{eq:diffusion_distance2}
\end{equation}
The sum starts from 2 since $\nu_{1}$ is constant. It follows that
in order to compute the diffusion distance, one can simply use the
right eigenvectors of $P$. Moreover, this facilitates the embedding
of the original points in a Euclidean space $\mathbb{R}^{\eta\left(\delta\right)-1}$
by:
\begin{equation}
\Xi_{t}:x_{i}\rightarrow\left(\lambda_{2}^{t}\nu_{2}\left(x_{i}\right),\lambda_{3}^{t}\nu_{3}\left(x_{i}\right),\dots,\lambda_{\eta\left(\delta\right)}^{t}\nu_{\eta\left(\delta\right)}\left(x_{i}\right)\right).\label{eq:DM_embedding_map}
\end{equation}
which also provides coordinates on the set $\Gamma$. Essentially,
$\eta\left(\delta\right)\ll n$ due to the fast decay of the eigenvalues
of $P$. Furthermore, $\eta\left(\delta\right)$ depends only on the
primary intrinsic variability of the data as captured by the random
walk and not on the original dimensionality of the data. This data-driven
method enables the parametrization of any set of points - abstract
or not - provided the similarity matrix $w_{\varepsilon}$ of the
points is available.

\section{Choosing $\varepsilon$\label{sec:Choosing_epsilon}}

The choice of $\varepsilon$ is critical to achieve the optimal performance
of the DM algorithm (and also for the DB algorithm that will be introduced
in Chapter \ref{cha:Diffusion-bases}) since it defines the size of
the local neighborhood of each point. On one hand, a large $\varepsilon$
produces a coarse analysis of the data as the neighborhood of each
point will contain a large number of points. In this case, the similarity
weight will be close to one for most pairs of points. On the other
hand, a small $\varepsilon$ might produce many neighborhoods that
contain only one point. In this case, the similarity weight will be
zero for most pairs of points. Clearly, an adequate choice of $\varepsilon$
lies between these two extreme cases and should be derived from the
data.

In the following, we describe an automatic procedure to determine
such an $\varepsilon$ for a Gaussian kernel based weight function
when the dataset $\Gamma$ approximately lies on a low dimensional
manifold \cite{mine-DB07}. That is, although $\Gamma$ is given in
the ambient Euclidean space $\mathbb{R}^{n}$, it is actually restricted
to an intrinsic low dimensional manifold $M$. We denote by $d$ the
intrinsic dimension of $M$. Let $L=I-P=I-D^{-1}W$ be the \emph{normalized
graph Laplacian} \cite{C97} where $P$ was defined in Eq. \ref{eq:P_mat}
and $I$ is the identity matrix. The matrices $L$ and $P$ share
the same eigenvectors. Furthermore, Singer \cite{Amit06} proved that
if the points in $\Gamma$ are independently uniformly distributed
over $M$ then with high probability
\begin{equation}
\frac{1}{\varepsilon}\sum_{j=1}^{N}L_{ij}f\left(x_{j}\right)=\frac{1}{2}\bigtriangleup_{M}f\left(x_{i}\right)+O\left(\frac{1}{N^{1/2}\varepsilon^{1/2+d/4}},\varepsilon\right)\label{eq:Laplace-Beltrami}
\end{equation}
where $f:M\rightarrow\mathbb{R}$ is a smooth function and $\bigtriangleup_{M}$
is the continuous Laplace-Beltrami operator of the manifold $M$.
The error term is composed of a variance term $O\left(\frac{1}{N^{1/2}\varepsilon^{1/2+d/4}}\right)$,
which is minimized by a large value of $\varepsilon$, and a bias
term $O\left(\varepsilon\right)$, which is minimized by a small value
of $\varepsilon$.

We utilize the scheme that was proposed in \cite{Hein05} and examine
the sum of the weight matrix elements
\begin{equation}
S_{\varepsilon}=\sum_{i=1}^{N}\sum_{j=1}^{N}w_{\varepsilon}\left(x_{i},x_{j}\right)=\sum_{i=1}^{N}\sum_{j=1}^{N}\exp\left(-\frac{\left\Vert x_{i}-x_{j}\right\Vert ^{2}}{2\varepsilon}\right)\label{eq:sum_vol}
\end{equation}
 as a function of $\varepsilon$. Let $Vol\left(M\right)$ be the
volume of the manifold $M$. The sum in Eq. \ref{eq:sum_vol} can
be approximated by its mean value integral
\begin{equation}
S_{\varepsilon}\approx\frac{N^{2}}{Vol^{2}\left(M\right)}\int_{M}\int_{M}\exp\left(-\frac{\left\Vert x-x'\right\Vert ^{2}}{2\varepsilon}\right)dxdx'\label{eq:mean_integral}
\end{equation}
 provided the variance term in Eq. \ref{eq:Laplace-Beltrami} is sufficiently
small.

\begin{figure}[!h]

\begin{centering}
\includegraphics[width=0.85\columnwidth]{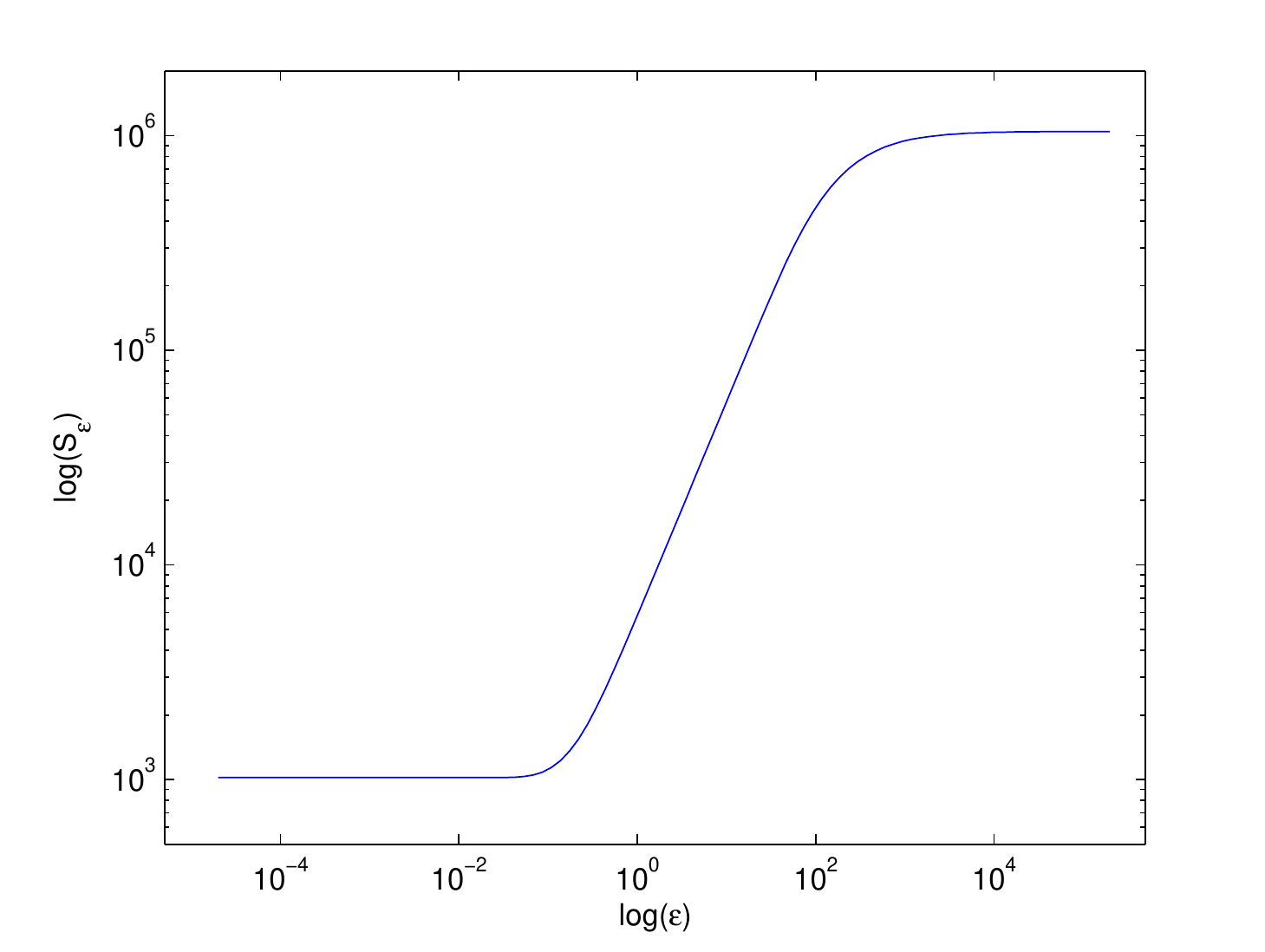}
\par\end{centering}

\caption{A plot of $S_{\varepsilon}$ as a function of $\varepsilon$ on a
log-log scale.\label{fig:epsilon_plot}}
\end{figure}

Moreover, we use the fact that for small values of $\varepsilon$,
the manifold looks locally like its tangent space $\mathbb{R}^{d}$
and thus
\begin{equation}
\int_{M}\exp\left(-\frac{\left\Vert x-x'\right\Vert ^{2}}{2\varepsilon}\right)dx\approx\int_{\mathbb{R}^{d}}\exp\left(-\frac{\left\Vert x-x'\right\Vert ^{2}}{2\varepsilon}\right)dx=\left(2\pi\varepsilon\right)^{d/2}.\label{eq:narrow}
\end{equation}
 Combining Eqs. \ref{eq:sum_vol}-\ref{eq:narrow}, we get
\[
S_{\varepsilon}\approx\frac{N^{2}}{Vol\left(M\right)}\left(2\pi\varepsilon\right)^{d/2}.
\]
 Applying logarithm on both sides yields
\[
\log\left(S_{\varepsilon}\right)\approx\frac{d}{2}\log\left(\varepsilon\right)+\log\left(\frac{N^{2}\left(2\pi\right)^{d/2}}{Vol\left(M\right)}\right).
\]
Consequently, the slope of $S_{\varepsilon}$ as a function of $\varepsilon$
on a log-log scale is $\frac{d}{2}$. However, this slope is only
linear in a limited subrange of $\varepsilon$ since $\lim_{\varepsilon\rightarrow\infty}S_{\varepsilon}=N^{2}$
and $\lim_{\varepsilon\rightarrow0}S_{\varepsilon}=N$ as illustrated
in Fig. \ref{fig:epsilon_plot}. In this subrange, the error terms
in Eq. \ref{eq:Laplace-Beltrami} are smaller than they are in the
rest of the $\varepsilon$ range. Thus, an adequate $\varepsilon$
is chosen from this linear subrange.

Given the properties of $S_{\varepsilon}$, a simple algorithm for
choosing $\varepsilon$ can be devised. First, a graph of $S_{\varepsilon}$
on a log-log space is constructed. The $x$ axis should range from
the $log$ of the minimal value to the maximal value of the similarity
matrix. If the minimal value is zero, a near zero number should be
taken. Given the graph, we look for a section where $S_{\varepsilon}$
is (nearly) linear i.e. a linear line. This can be achieved by looking
for a near-constant section of the derivative of the graph. We choose
$\varepsilon$ from this section.

\section{Conclusion}

In this chapter, the diffusion maps method \cite{CL_DM06} for dimensionality
reduction was introduced. The main contribution of this chapter is
a novel algorithm - theory and description - for the automatic derivation
of the parameter $\varepsilon$ which is crucial to the performance
of the algorithm since it defines the local neighborhood that is examined
in the algorithm.

\chapter{Diffusion Bases (DB)\label{cha:Diffusion-bases}}

Diffusion bases (DB) is a dual algorithm to the DM algorithm in the
sense that it explores the variability among the \emph{coordinates}
of the original data \cite{mine-DB07}. Both algorithms share a graph
Laplacian construction, however, the DB algorithm uses the Laplacian
eigenvectors as an orthonormal system and projects the original data
onto it. The DB algorithm also assumes that there is access to the
original data $\Gamma$.

Let $\Gamma$ be the original dataset as defined in Eq. \ref{eq:gamma}
and let $x_{i}\left(j\right)$ denote the $j$-th coordinate of $x_{i}$,
$1\le j\le n$. We define the vector 
\begin{equation}
x'_{j}\triangleq\left(x_{1}\left(j\right),\dots,x_{N}\left(j\right)\right)\label{eq:coordinate_vector}
\end{equation}
to be the $j$-th coordinate of all the points in $\Gamma$. We construct
the set $\Gamma'=\left\{ x'_{j}\right\} _{j=1}^{n}.$ The DM algorithm
is executed on the set $\Gamma'$. The right eigenvectors of $P$
constitute an orthonormal basis $\left\{ \nu_{k}\right\} _{k=1,\dots,n},\,\nu_{k}\in\mathbb{R}^{n}$.
These eigenvectors capture the \emph{non-linear} coordinate-wise variability
of the original data. This bares some similarity to PCA, however,
the diffusion process yields a more accurate estimation of the variability
than PCA due to: (a) its ability to capture non-linear manifolds within
the data by local exploration of each coordinate; (b) its robustness
to noise. Furthermore, this process is more general than PCA and it
coincides with it when the weight function $w_{\varepsilon}$ (Section
\ref{sub:Building-the-graph}) is \emph{linear}.

Next, we use the eigenvalue decay property of the eigen-decomposition
to extract only the first $\eta\left(\delta\right)$ eigenvectors
$B\triangleq\left\{ \nu_{k}\right\} _{k=1,\dots,\eta\left(\delta\right)}$
(we are not excluding the first eigenvector as mentioned in Section
\ref{sub:Spectral-decomposition}), which contain the \emph{non-linear}
directions with the highest variability of the coordinates of the
original dataset $\Gamma$.

The dimensionality of $\Gamma$ is reduced by projecting it onto the
basis $B$. Let 
\begin{equation}
\Gamma_{B}=\left\{ \widetilde{x}_{i}\right\} _{i=1}^{N}\label{eq:gamma_b}
\end{equation}
be the set of these projections, similarly to as it was defined in
Eq. \ref{eq:gamma_gal}. The embedding $\widetilde{x}_{i}$ of $x_{i}$
is defined as 
\[
\widetilde{x}_{i}=\left(\left\langle x_{i},\nu_{1}\right\rangle ,\dots,\left\langle x_{i},\nu_{\eta\left(\delta\right)}\right\rangle \right),\,\, i=1,\dots,N
\]
 and $\left\langle \cdot,\cdot\right\rangle $ denotes the inner product
operator. $\Gamma_{B}$ contains the coordinates of the original points
in the orthonormal system whose axes are given by $B$. Alternatively,
$\Gamma_{B}$ can be interpreted in the following way: the coordinates
of $\widetilde{x}_{i}$ contain the correlation between $x_{i}$ and
the directions given by the vectors in $B$. A summary of the \emph{DiffusionBasis}
procedure is given in Algorithm \ref{alg:Diffusion-Basis-Calculation}.

\section{Numerical enhancement of the eigen-decomposition\label{sub:Numerical-enhancement-of}}

The Markov matrix, which is obtained in Eq. \ref{eq:markov_normalization},
is not symmetric. In general, working with a symmetric matrix is faster
and more accurate. A symmetric matrix $A$, which is conjugate to
$P$, can be obtained in the following way:
\begin{equation}
a\left(x_{i},x_{j}\right)=\frac{w_{\varepsilon}\left(x_{i},x_{j}\right)}{\sqrt{d\left(x_{j}\right)}\sqrt{d\left(x_{i}\right)}}\label{eq:symmetric_conj_mat}
\end{equation}
where $d\left(x_{j}\right)$ and $d\left(x_{i}\right)$ are defined
in Eq. \ref{eq:degree}. Let $\left\{ \vartheta_{k}\right\} _{k=1,\dots,N}$
be the eigenvectors of $A.$ It can be shown (see \cite{C97}) that
$P$ and $A$ have the same eigenvalues and that 
\begin{equation}
\nu_{k}=\frac{\vartheta_{k}}{\vartheta_{1}};\,\,\mu_{k}=\vartheta_{k}\vartheta_{1}\label{eq:modified_eigs}
\end{equation}
where $\left\{ \mu_{k}\right\} $ and $\left\{ \nu_{k}\right\} $
are the left and right eigenvectors of $P$, respectively. This leads
to modifications of the DM algorithm and the DB algorithm (Algorithm
\ref{alg:Diffusion-Basis-Calculation}). The modified DM (abbreviated
as MDM from this point on) algorithm calculates the eigenvectors $\left\{ \nu_{k}\right\} _{k=1}^{\eta\left(\delta\right)}$
using the symmetric matrix $A$ in Eq. \ref{eq:symmetric_conj_mat}
and applying Eq. \ref{eq:modified_eigs} while the modified DB algorithm
(abbreviated as MDB from this point on) projects the data points in
$\Gamma$ onto the orthonormal basis $\left\{ \vartheta_{k}\right\} _{k=1}^{\eta\left(\delta\right)}$
instead of $\left\{ \nu_{k}\right\} _{k=1}^{\eta\left(\delta\right)}$
where the number $\eta\left(\delta\right)$ was introduced in Section
\ref{sub:Spectral-decomposition}. The MDB algorithm is given in Algorithm
\ref{alg:Modified-diffusion}.

\begin{algorithm}[!h]
\textbf{DiffusionBasis($\Gamma'$, $w_{\varepsilon}$, $\varepsilon$,
$\delta$)}
\begin{enumerate}
\item Calculate the weight function $w_{\varepsilon}\left(x'_{i},x'_{j}\right),\,\, i,j=1,\dots n$,
(Eq. \ref{eq:gaussian}). 
\item Construct a Markov transition matrix $P$ by normalizing the sum of
each row in $w_{\varepsilon}$ to be 1: 
\[
p\left(x'_{i},x'_{j}\right)=\frac{w_{\varepsilon}\left(x'_{i},x'_{j}\right)}{d\left(x'_{i}\right)}
\]
 where $d\left(x'_{i}\right)=\sum_{j=1}^{n}w_{\varepsilon}\left(x'_{i},x'_{j}\right)$. 
\item Perform eigen-decomposition of $p\left(x'_{i},x'_{j}\right)$
\[
p\left(x'_{i},x'_{j}\right)\equiv\sum_{k=1}^{n}\lambda_{k}\nu_{k}\left(x'_{i}\right)\mu_{k}\left(x'_{j}\right)
\]
 where the left and the right eigenvectors of $P$ are given by $\left\{ \mu_{k}\right\} $
and $\left\{ \nu_{k}\right\} $, respectively, and $\left\{ \lambda_{k}\right\} $
are the eigenvalues of $P$ in descending order of magnitude. 
\item Project the original data $\Gamma$ onto the orthonormal system $B\triangleq\left\{ \nu_{k}\right\} _{k=1,\dots,\eta\left(\delta\right)}$
to produce:
\[
\Gamma_{B}=\left\{ \widetilde{x}_{i}\right\} _{i=1}^{N},\,\widetilde{x}_{i}\in\mathbb{R}^{\eta\left(\delta\right)}
\]
where
\[
\widetilde{x}_{i}=\left(\left\langle x_{i},\nu_{1}\right\rangle ,\dots,\left\langle x_{i},\nu_{\eta\left(\delta\right)}\right\rangle \right),\,\, i=1,\dots,N,\,\nu_{1},\ldots,\nu_{\eta\left(\delta\right)}\in B
\]
 and $\left\langle \cdot,\cdot\right\rangle $ is the inner product. 
\item \textbf{return} $\Gamma_{B}$. 
\end{enumerate}
\caption{The Diffusion Basis algorithm. \label{alg:Diffusion-Basis-Calculation}}
\end{algorithm}

\begin{algorithm}[!h]
\textbf{ModifiedDiffusionBasis($\Gamma'$, $w_{\varepsilon}$, $\varepsilon$,
$\delta$)}
\begin{enumerate}
\item Calculate the weight function $w_{\varepsilon}\left(x'_{i},x'_{j}\right),\,\, i,j=1,\dots n$ 
\item Construct the matrix $A$ 
\[
a\left(x'_{i},x'_{j}\right)=\frac{w_{\varepsilon}\left(x'_{i},x'_{j}\right)}{\sqrt{d\left(x'_{j}\right)}\sqrt{d\left(x'_{i}\right)}}
\]
 where $d\left(x'_{i}\right)=\sum_{j=1}^{n}w_{\varepsilon}\left(x'_{i},x'_{j}\right)$. 
\item Perform eigen-decomposition of $a\left(x'_{i},x'_{j}\right)$
\[
a\left(x'_{i},x'_{j}\right)\equiv\sum_{k=1}^{n}\lambda_{k}\vartheta_{k}\left(x'_{i}\right)\vartheta_{k}\left(x'_{j}\right)
\]
 where $\left\{ \vartheta_{k}\right\} $ are the eigenvectors of $A$,
and $\left\{ \lambda_{k}\right\} $ are the eigenvalues of $A$ in
descending order of magnitude. 
\item Project the original data $\Gamma$ onto the orthonormal system $B\triangleq\left\{ \vartheta_{k}\right\} _{k=1,\dots,\eta\left(\delta\right)}$
to produce:
\[
\Gamma_{B}=\left\{ \widetilde{x}_{i}\right\} _{i=1}^{N},\,\widetilde{x}_{i}\in\mathbb{R}^{\eta\left(\delta\right)}
\]
 where
\[
\widetilde{x}_{i}=\left(\left\langle x_{i},\vartheta_{1}\right\rangle ,\dots,\left\langle x_{i},\vartheta_{\eta\left(\delta\right)}\right\rangle \right),\,\, i=1,\dots,N,\,\vartheta_{1},\ldots,\vartheta_{\eta\left(\delta\right)}\in B
\]
 and $\left\langle \cdot,\cdot\right\rangle $ denotes the inner product. 
\item \textbf{return} $\Gamma_{B}$. 
\end{enumerate}
\caption{The modified Diffusion Basis algorithm. \label{alg:Modified-diffusion}}
\end{algorithm}

\section{The algebraic connection between the DM and DB algorithms}

Let $\Gamma$ be a set of vectors as defined in Eq. \ref{eq:gamma}.
We define weight functions 
\[
w_{\varepsilon_{DM}}\left(x_{i},x_{j}\right)=\exp\left(-\frac{\left\Vert x_{i}-x_{j}\right\Vert ^{2}}{\varepsilon_{DM}}\right),\,\, i,j=1,\ldots,N
\]
 and 
\[
w_{\varepsilon_{DB}}\left(x'{}_{i},x'{}_{j}\right)=\exp\left(-\frac{\left\Vert x'{}_{i}-x'{}_{j}\right\Vert ^{2}}{\varepsilon_{DB}}\right),\,\, i,j=1,\ldots,n
\]
to be used by the modified diffusion maps (MDM) and the modified diffusion
bases (MDB) algorithms (Section \ref{sub:Numerical-enhancement-of}),
respectively. We denote by $B^{DM}=\left\{ \vartheta_{k}^{DM}\right\} _{k=1,\dots,N}$
and $B^{DB}=\left\{ \vartheta_{k}^{DB}\right\} {}_{k=1,\dots,n}$
the eigenvectors that are the result of the eigen-decomposition in
the MDM and MDB algorithms, respectively. $B^{DM}$ and $B^{DB}$
are orthonormal systems in $\mathbb{R}^{N}$ and $\mathbb{R}^{n}$,
respectively. Accordingly, every coordinate vector $x'_{i}$ can be
expressed in $B^{DM}$ as 
\[
x'_{i}=\sum_{k=1}^{N}\left\langle x'_{i},\vartheta_{k}^{DM}\right\rangle \vartheta_{k}^{DM}.
\]
 In a similar manner, every vector $x_{i}$ can be expressed in $B^{DB}$
as 
\[
x_{i}=\sum_{k=1}^{n}\left\langle x_{i},\vartheta_{k}^{DB}\right\rangle \vartheta_{k}^{DB}.
\]
 Let $M_{\Gamma}$ be a matrix whose rows are composed of the vectors
in $\Gamma$. In this setting, the columns of $M_{\Gamma}$ are given
by the coordinate vectors $\left\{ x'_{i}\right\} _{i=1}^{n}$ (Eq.
\ref{eq:coordinate_vector}). We define the set $\left\{ \vartheta_{kl}^{DMDB}\right\} {}_{k=1,\dots,N;l=1,\dots,n}$
where $\vartheta_{kl}^{DMDB}=\vartheta_{k}^{DM}\times\vartheta_{l}^{DB}$
and $\times$ denotes the cross product. It is easy to verify that
$B^{DMDB}$ is an orthonormal system in $\mathbb{R}^{N\times n}$.
We also define $M_{\Gamma}$ as a row vector $\overrightarrow{M}_{\Gamma}$
by horizontally concatenating its rows 
\[
\overrightarrow{M}_{\Gamma}\triangleq\left(x_{1},x_{2},\ldots,x_{N}\right).
\]
 Thus, $\overrightarrow{M}_{\Gamma}$ can be expressed in $B^{DMDB}$
as 
\[
\overrightarrow{M}_{\Gamma}=\sum_{k=1}^{N}\sum_{l=1}^{n}\left\langle \overrightarrow{M}_{\Gamma},\vartheta_{kl}^{DMDB}\right\rangle \vartheta_{kl}^{DMDB}.
\]

An attempt to directly calculate $B^{DMDB}$ will involve a weight
function of size $N\times n$ by $N\times n$ which is not feasible
even for relatively small values of $N$ and $n$. However, we see
that $B^{DMDB}$ can indirectly be calculated as the Cartesian product
of the orthonormal bases that result from the eigen-decomposition
step in the DM and DB algorithms.

\section{A graph theoretic link between DB and DM}

In this section, we describe a graph theoretic framework that links
between the DM and DB algorithms by constructing a special graph.
This construction demonstrates the duality between these methods.

\begin{figure}[!h]
\begin{centering}
\includegraphics[bb=81bp 336bp 515bp 780bp,clip,width=3in]{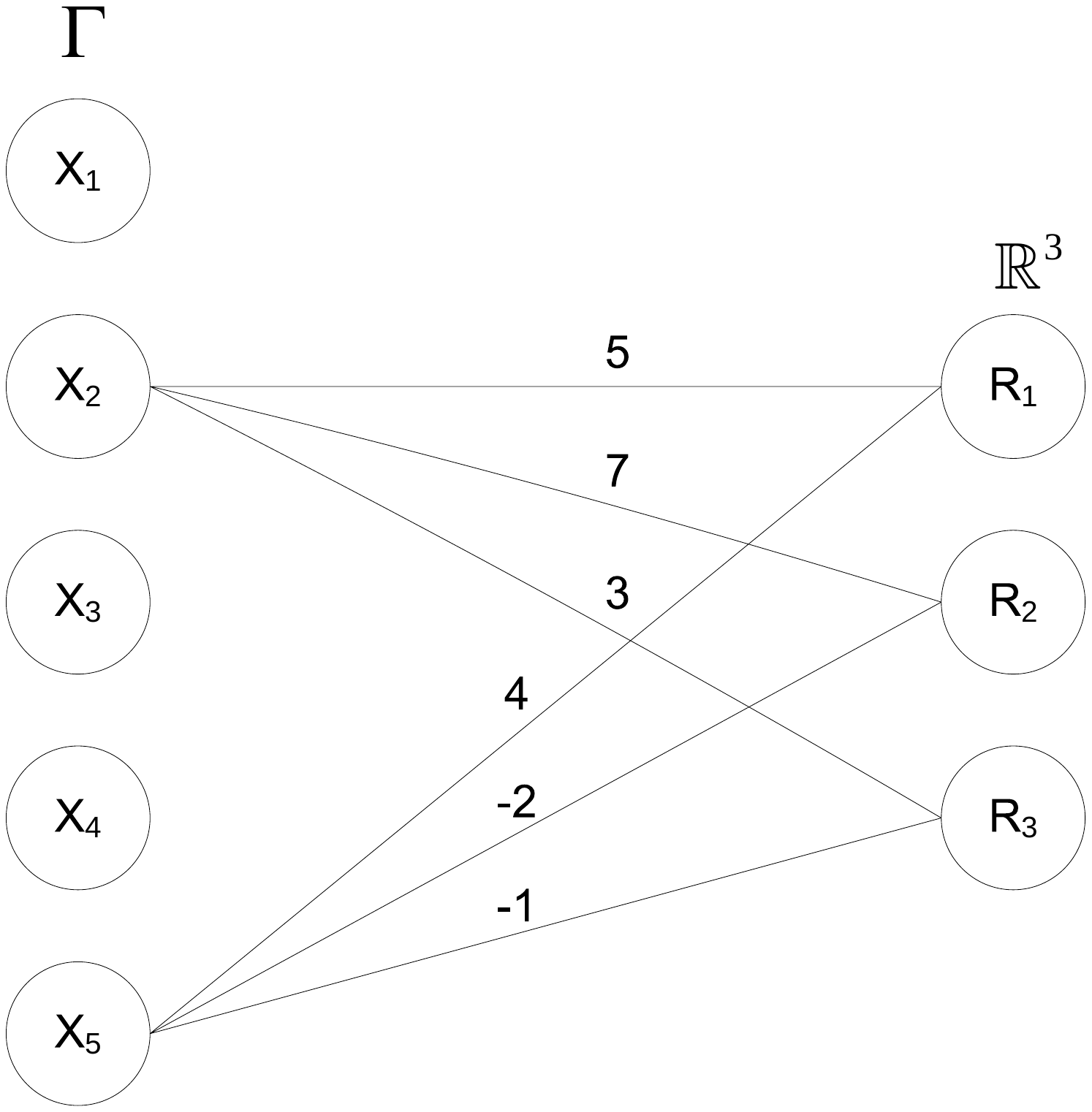}
\par\end{centering}

\caption{An example for a bi-partite graph construction for the set $\Gamma=\left\{ x_{1},\ldots,x_{5}\right\} \subseteq\mathbb{R}^{3}$
where the edge weights are specified for $x_{2}=\left(5,7,3\right)$
and $x_{5}=\left(4,-2,-1\right)$. \label{fig:bi-partite} }
\end{figure}

Let $\Gamma$ be a set of vectors as defined Eq. \ref{eq:gamma}.
This set of vectors can be represented by a full weighted bi-partite
graph $G=\left(U\cup V,W\right),\,\left|U\right|=N,\,\left|V\right|=n$
where $U$ corresponds to the vectors $\left\{ x_{i}\right\} $ and
$V$ corresponds to the dimensions of $\mathbb{R}^{n}$. The weight
of each edge $\left(u_{i},v_{j}\right)\in W$ is given by $x_{i}\left(j\right)\equiv x'{}_{j}\left(i\right)$
- the value in the $j$-th coordinate of the vector $x_{i}$ which
is equal to the $i$-th coordinate of the vector $x_{j}$. The size
of the weight matrix $W$ is $N\times n$. We demonstrate this construction
in Fig. \ref{fig:bi-partite} for the set $\Gamma=\left\{ x_{1},\ldots,x_{5}\right\} \subseteq\mathbb{R}^{3}$
where the edge weights are specified for $x_{2}=\left(5,7,3\right)$
and $x_{5}=\left(4,-2,-1\right)$.

For the case of a hyper-spectral cube $I$, the graph construction
is as follows. Let $I\left(x,y,z\right)$ be the intensity of wavelength
$z$ at spatial position $\left(x,y\right)$ where $z=0,\ldots,L-1$
and $\,\, x=0,\ldots,C-1;\, y=0,\ldots,R-1$. We construct a bi-partite
graph $G=\left(U\cup V,W\right)$ where $U$ correspond to the spatial
positions and $V$ corresponds to the wavelengths. Their sizes are
given by $\left|U\right|=RC$ and $\left|V\right|=L$. A vertex $u_{i}\in U$
corresponds to the spatial position $\left(\left\lfloor \frac{i}{C}\right\rfloor ,i-\left\lfloor \frac{i}{C}\right\rfloor \cdot C\right)$
where $\left\lfloor \cdot\right\rfloor $ denotes the floor operator.
The weight of an edge $\left(u_{i},v_{j}\right)\in W$ is equal to
the $j$-th wavelength intensity at spatial position $\left(\left\lfloor \frac{i}{C}\right\rfloor ,i-\left\lfloor \frac{i}{C}\right\rfloor \cdot C\right)$.

In the following, we describe a construction that corresponds to the
kernel that is used in Section \ref{sub:Building-the-graph}. First,
we construct a function $f:M\mapsto Q$ which takes a matrix $M$
and produces a matrix $Y$ that depends on $M$ and $M^{T}$. An example
for $f$ is the row pairwise correlation where $Y=MM^{T}$and
\begin{equation}
Y_{corr}\left(i,j\right)=\sum_{k}M\left(i,k\right)M\left(j,k\right)=\sum_{k}M\left(i,k\right)M^{T}\left(k,j\right).\label{eq:Y_corr}
\end{equation}
 Another example is the row pairwise squared $l_{2}$ distances 
\begin{equation}
Y_{dist}\left(i,j\right)=\sum_{k}\left(M\left(i,k\right)-M\left(j,k\right)\right)^{2}=\sum_{k}\left(M\left(i,k\right)-M^{T}\left(k,j\right)\right)^{2}.\label{eq:Y_dist}
\end{equation}
An alternative definition for $f$ is a function $\widetilde{f}:\Gamma\times\Gamma\rightarrow\mathbb{R}$.
In this context, the function $Y_{corr}\left(i,j\right)\equiv\left\langle x_{i},x_{j}\right\rangle $
(Eq. \ref{eq:Y_corr}) can be defined using $\widetilde{f}$ when
$\widetilde{f}$ is the inner product. Similarly, $Y_{dist}\left(i,j\right)\equiv\left\Vert x_{i}-x_{j}\right\Vert ^{2}$
(Eq. \ref{eq:Y_dist}) can be defined by $\widetilde{f}$ when $\widetilde{f}$
is the squared $l_{2}$ distance.

Next, we define an \emph{element-wise} function $g$ on a matrix $M$
where $g\left(M\right)$ denotes the result of applying $g$ on every
element of $M$. We denote by $g\left(M\left(i,j\right)\right)$ the
result of the application of $g$ to the element at row $i$ and column
$j$. With the above definitions, we can define any kernel on $I$
for both the DM and the DB algorithms. A DM kernel is constructed
from $W$ by $K\left(W\right)\triangleq g\left(f\left(W\right)\right)$,
while a DB kernel is constructed from $W$ by $K\left(W\right)\triangleq g\left(f\left(W^{T}\right)\right)$.
For example, the Gaussian kernel in Eq. \ref{eq:gaussian} can be
defined using $f\equiv Y_{dist}\left(i,j\right)$ and $g\left(x\right)=e^{-x/\varepsilon}$.

Given a kernel $K$ for which $f$ is $Y_{mul}$ and $g$ is the identity
function, DM and DB are connected through the eigenvectors of $WW^{T}$and
$W^{T}W$\emph{.} In order to verify this, we look at the singular
value decomposition of $W=BSR^{T}$ (assuming there is one). Since
$WW^{T}=BSR^{T}RSB^{T}=BS^{2}B^{T}$ and $W^{T}W=RSB^{T}BSR^{T}=RD^{2}R^{T}$,
the result of the eigen-decomposition in the DM and DB algorithms
are given by $B$ and $R$, respectively.

\section{Conclusion}

In this chapter, a novel method for dimensionality reduction was introduced
- diffusion bases. The method is closely related to the diffusion
maps algorithm \cite{CL_DM06}. The algebraic and graph-theoretic
connection between the methods was presented. Both methods preserve
the geometrical structure of the dataset whose dimension is being
reduced. 

Table \ref{tab:DMDB_Comparison} summarizes the differences between
the DM and DB algorithms. When the number of items in the dataset
is higher than the dimensionality of the items (this is the common
case), the diffusion bases algorithm has a lower time and space complexity
than the diffusion maps algorithm.

\begin{table}[!h]

\centering{}%
\begin{longtable}{|l|c|c|}
\hline 
{\scriptsize Criterion} & {\scriptsize DM} & {\scriptsize DB}\tabularnewline
\hline 
\hline 
{\scriptsize Similarity matrix is built on} & {\scriptsize data items} & {\scriptsize coordinates of the data items}\tabularnewline
\hline 
{\scriptsize Size of similarity matrix} & {\scriptsize $N\times N$} & {\scriptsize $n\times n$}\tabularnewline
\hline 
{\scriptsize Time complexity of } &  & \tabularnewline
{\scriptsize similarity matrix construction} & {\scriptsize $O\left(nN^{2}\right)$} & {\scriptsize $O\left(Nn^{2}\right)$}\tabularnewline
\hline 
{\scriptsize Embedding mapping} & {\scriptsize $x_{i}\rightarrow\left(\lambda_{2}^{t}\nu_{2}\left(x_{i}\right),\lambda_{3}^{t}\nu_{3}\left(x_{i}\right),\dots,\lambda_{\eta\left(\delta\right)}^{t}\nu_{\eta\left(\delta\right)}\left(x_{i}\right)\right)$} & {\scriptsize $\widetilde{x}_{i}=\left(\left\langle x_{i},\nu_{1}\right\rangle ,\dots,\left\langle x_{i},\nu_{\eta\left(\delta\right)}\right\rangle \right)$}\tabularnewline
\hline 
\end{longtable}\caption{Comparison between the DM and DB algorithms.}
\label{tab:DMDB_Comparison}
\end{table}

\chapter{Introduction to Hyper-spectral Imagery\label{cha:Introduction-to-Hyper-spectral}}

A significant recent breakthrough in imagery has been the development
of hyper-spectral sensors and software to analyze the resulting image
data. Fifteen years ago only spectral remote sensing experts had access
to hyper-spectral images or software tools to take advantage of such
images. Over the past decade hyper-spectral image analysis has matured
into one of the most powerful and fastest growing technologies in
fields such as remote sensing and biology to name a few. It is now
gradually entering to the area of medical diagnostics.

The ``hyper'' in hyper-spectral means ``over'' as in ``too many''
and refers to the large number of measured wavelength bands. Hyper-spectral
images are spectrally overdetermined, which means that they provide
ample spectral information to identify and distinguish spectrally
unique materials. Hyper-spectral imagery provides the potential for
more accurate and detailed information extraction than it is possible
with any other type of remotely sensed data. A hyper-spectral cube
is a good example for a high dimensional dataset. Therefore, the methods
in previous chapters will be applied to these datasets in order to
reduce their dimensionality.

In this chapter we review definitions of some basic hyper-spectral
concepts and present some recent hyper-spectral image analysis research.

\section{Spectral imaging: basics}

To understand the advantages of hyper-spectral imagery, it may help
to first review some basic spectral remote sensing concepts. We recall
that each photon of light has a wavelength determined by its energy
level. Light and other forms of electromagnetic radiation are commonly
described in terms of their wavelengths. For example, visible light
has wavelengths between 0.4 and 0.7 microns, while radio waves have
wavelengths greater than 30 cm. Reflectance is the percentage of the
light hitting a material that is then reflected by that material (as
opposed to being absorbed or transmitted). A reflectance spectrum
shows the reflectance of a material measured across a range of wavelengths.
Some materials will reflect certain wavelengths of light, while other
materials will absorb the same wavelengths. These patterns of reflectance
and absorption across wavelengths can be used to uniquely identify
certain materials. We refer to a unique reflectance pattern of a material
as a \emph{spectral signature}. In Chapter \ref{cha:Automatic-identification-of}
we extract features from spectral signatures in order to uniquely
identify materials. These signatures are obtained in laboratory conditions
in the wavelength range of $400-2400$ nanometer. It should be mentioned
that given a sufficient amount of noise and a low spectral resolution
(a small number of wavelengths) two signatures might be similar. Nevertheless,
this is not the case in the data that is used in this thesis. Our
data has high spectral resolution - which is referred to as \emph{hyper}-spectral
(see the next section). This guarantees the uniqueness of the signatures
even in the presence of noise.

Field and laboratory spectrometers usually measure reflectance at
many narrow, closely spaced spectral bands, so that the resulting
spectra appear to be continuous curves. When a spectrometer is used
in an imaging sensor, the resulting images record a reflectance spectrum
for each pixel in the image. A tutorial on this topic can be found
in \cite{Monteiro04}.

\section{Hyper-spectral data}

Most multi-spectral imaging devices measure radiation reflected from
a surface at a few wide, separated spectral bands. Most hyper-spectral
imagers, on the other hand, measure reflected radiation at a series
of narrow and contiguous spectral bands. When we look at a spectrum
for one pixel in a hyper-spectral image, it resembles a spectrum that
would be measured in a spectroscopy laboratory. This type of detailed
pixel spectrum can provide much more information about the surface
than a multi-spectral pixel spectrum where \emph{multi} indicates
a lower resolution of captured spectral bands.

Although most hyper-spectral sensors measure hundreds of wavelengths,
it is not the number of measured wavelengths that defines a sensor
as hyper-spectral but rather it is the narrowness and contiguous nature
of the measurements. For example, a sensor that measures only 20 bands
could be considered hyper-spectral if those bands were contiguous
and, say, 10 nanometer (\emph{nm}) wide. If a sensor measured 20 spectral
bands that were, say, 100 \emph{nm} wide, or that were separated by
non-measured wavelength ranges, the sensor would no longer be considered
hyper-spectral.

Standard multi-spectral image classification techniques were generally
developed to classify multi-spectral images into broad categories.
Hyper-spectral imagery provides an opportunity for more detailed image
analysis. For example, using hyper-spectral data, spectrally similar
materials can be distinguished, and sub-pixel scale information can
be extracted. To materialize this potential, new image processing
techniques have been developed. In Chapter \ref{cha:Segmentation-and-Anomalies}
we introduce algorithms for segmentation of hyper-spectral images
and detection of anomalies (manifested as sub-pixel segments).

\section{Applications of hyper-spectral image analysis}

Hyper-spectral imagery has been used to detect and map a wide variety
of materials using their characteristic reflectance spectra. For example,
hyper-spectral images have been used by geologists for mineral mapping
(\cite{Salisbury91,Clark92}) and for detection of soil properties
including moisture, organic content, and salinity (\cite{Jensen86}).
Vegetation scientists have successfully used hyper-spectral imagery
to identify vegetation species (\cite{Vieira88}), study plant canopy
chemistry (\cite{BenDor01}), and detect vegetation stress (\cite{Aber95}).
Military personnel have used hyper-spectral imagery to detect military
vehicles under partial vegetation canopy, and also for many other
military target detection objectives.

\subsection{Medical applications}

With new advances in sensor technology and the recent affordability
of high-performance spectral imagers, hyper-spectral instruments have
enabled a host of new applications - with key efforts in the area
of medical imaging. Rather than flying over battlefields, these hyper-spectral
sensors can be deployed to scan a patient's body in search of pre-cancerous
regions or to provide much needed spectral information through endoscopy
procedures. These hyper-spectral medical instruments hold great potential
for non-invasive diagnosis of cancer, assessment of wound conditions,
etc. For the patient, tremendous advantage is obtained by being able
to not only diagnose the condition in a non-invasive manner but also
to potentially treat the condition at the time of diagnosis. Great
interest has been generated on behalf of health care providers to
investigate the promise of reducing health care costs and timeliness
of treatment for many types of disease conditions through the use
of hyper-spectral scanning procedures.

Physicians have successfully used hyper-spectral imagery to detect
skin tumors. Skin cancers may not be visually obvious since the visual
signature may appear as a shape distortion rather than discoloration.
Hyper-spectral imaging offers an instant, non-invasive diagnostic
procedure (comparing to skin biopsy) based on the analysis of the
spectral signatures of skin tissue (\cite{Kong05}). Hyper-spectral
imaging was found efficient in providing information regarding the
spatial tissue oxygen saturation in patients with peripheral vascular
disease (\cite{Patent}). In addition hyper-spectral imaging systems
have been shown to be useful in the monitoring of the spatial distribution
of skin oxygenation (\cite{Criqui85,Potter01}), general surgery,
plastic surgery, hemorrhagic shock, and burns (\cite{Patent,Mansfield98,Sowa01,Afromowitz88,Cancio}).

\section{Spectral libraries}

Spectral libraries are collections of reflectance spectra measured
from materials of known composition, usually in the field or laboratory.
Many investigators construct spectral libraries of materials from
their field sites as part of every project, to facilitate analysis
of multi-spectral or hyper-spectral imagery from those sites. Several
high quality spectral libraries are also publicly available (e.g.
(\cite{Merton99,Grove92,Elvidge90,Korb96,Salisbury91a,Salisbury94})).

\section{Classification and target identification}

There are many unique image analysis algorithms that have been developed
to take advantage of the extensive information contained in hyper-spectral
imagery. Most of these algorithms also provide accurate, although
more limited, analysis of multi-spectral data. Spectral analysis methods
usually compare pixel spectra with a reference spectrum (often called
a target). Target spectra can be derived from a variety of sources,
including spectral libraries, regions of interest within a spectral
image, or individual pixels within a spectral image. The most commonly
used hyper-spectral/multi-spectral image analysis methods will be
described in Section \ref{related}.

\section{Hyper-spectral cameras\label{sec:Hyper-spectral-Cameras}}

Hyper-spectral cameras produce hyper-spectral images of a viewed scene
and they gain growing popularity in recent years. These cameras can
be installed, for example, on a satellite, on a microscope or on other
devices. There are many uses for this type of imagery: biologists
use hyper-spectral images of human tissues in order to detect malignant
cells, while agronomists use remote sensed images of the earth to
estimate amounts of crops.

A regular CCD camera provides the geometry of a scene, however, with
very limited spectral information as it is equipped with sensors that
only capture details that are visible to the naked eye. In contrast,
a hyper-spectral image acquisition device is equipped with multiple
sensors - each sensor is sensitive to a particular range of the light
spectrum. The output of the device contains the reflectance values
of a scene at \emph{all} the wavelengths that the sensors are designed
to capture. Thus, a hyper-spectral image is composed of a set of images
- one for each wavelength%
\footnote{A hyper-spectral image is sometimes referred to as a hyper-spectral
\emph{cube} of layered images. %
}. We refer to a set of wavelength values at a coordinate $\left(x,y\right)$
as a \emph{hyper-pixel.} Each hyper-pixel can be represented by a
vector in $\mathbb{R}^{n}$ where $n$ is the number of wavelengths.
This data can be used to achieve inferences that can not be derived
from a limited number of wavelengths which are obtained by regular
cameras. 

An example for a hyper-spectral camera is \emph{AVIRIS }which stands
for Airborne Visible InfraRed Imaging Spectrometer. It captures images
in 224 contiguous spectral bands with wavelengths ranging from 400
to 2500 nanometers. Typically, the camera is mounted aboard a NASA
ER-2 airplane which flies at approximately 20 km above sea level.
In Chapter \ref{cha:Automatic-identification-of}, the spectrum of
the signatures is acquired in the range of $400-2400$ nanometers.
For comparison, a regular camera, on the other hand, captures a scene
in the wavelength range of $400-700$ nanometer which is the range
visible to the human eye.

\chapter{Segmentation and Detection of Anomalies in Hyper-Spectral Images
via Diffusion Bases\label{cha:Segmentation-and-Anomalies}}

%Hyper-spectral cameras capture images at hundreds and even thousands different wavelengths. The images they produce are referred to as hyper-spectral images. Biology, agriculture, medicine and remote sensing are some of the fields where hyper-spectral images are used. While intensity images provide only the geometry of a scene, hyper-spectral images offer spectral information as well. This additional information enables the development of novel efficient techniques for solving classical image processing problems such as segmentation, change detection etc.
While intensity images provide only the geometry of a scene, hyper-spectral
images offer spectral information as well. This additional information
enables the development of novel and efficient techniques for solving
classical image processing problems such as segmentation, change detection,
etc. A hyper-spectral image can be modeled as a cube of layered images
- one layer per wavelength. This cube is processed in order to extract
additional information which lies in the spectral domain. Applying
classical image processing techniques separately on each layer does
not utilize the inter-wavelength spectral properties of the objects
in the image. These methods also fail when sub-pixel segmentation
is needed, for example in various remote sensing applications.

In this chapter, we present a novel approach for the segmentation
of hyper-spectral images. Our approach reduces the dimensionality
of the data by embedding it in a low dimensional space while maintaining
the coherency of the information that is crucial for the derivation
of the segmentation. Furthermore, our algorithm is able to detect
anomalies which come in the form of sub-pixel segments.

\section{Introduction\label{sec:Introduction6}}

Every substance in nature has a unique \emph{spectral signature} (see
Section \ref{sec:Hyper-spectral-Cameras}) which characterizes its
physical properties. One can analyze hyper-spectral images using two
approaches: (a) understanding the physical nature; (b) Employing image
processing and computer vision techniques. Inter-wavelength connections,
which are inherent in spectral signatures, can not be utilized when
classical image processing techniques are separately applied on each
wavelength image. Thus, the entire hyper-spectral cube needs to be
processed in order to analyze the physical nature of the scene. Naturally,
this has to be done efficiently due to the large volume of the data.

We propose a general framework for automatic segmentation of hyper-spectral
volumes by encapsulating both approaches (physical and image processing)
via manifold learning. It is based on the \emph{diffusion bases} scheme
(see Chapter \ref{cha:Diffusion-bases}). Our approach provides a
coherent methodology that projects the hyper-spectral cube onto a
space of substantially smaller dimension while preserving its geometrical
structure. We present a novel segmentation approach that is tailored
for the data in the dimension-reduced space. We distinguish between
segmentation of areas that contain more than one pixel and sub-pixel
segmentation (anomalies detection) that is common in remote sensing.
The algorithms for these segmentation types consist of two steps.
Both segmentation types share the first step that performs dimensionality
reduction. However, the second step is different: sub-pixel segmentation
is accomplished by applying a local variance detection scheme while
for the other case, we use a histogram-based method to cluster hyper-pixels
in the reduced-dimensional space. The proposed algorithm is fast,
automatic (no supervision is required) and it is robust to noise,
so there is no need for a pre-processing denoising stage.

This chapter is organized as follows: in Section \ref{sec:Related-Works4}
we present a survey of related work on segmentation. In Section \ref{sec:WWG}
we introduce the two phase \emph{Wavelength-wise Global} (WWG) hierarchical
segmentation algorithm. Section \ref{sec:Experimental-Results5} contains
experimental results from the application of the algorithm on several
hyper-spectral volumes. Concluding remarks are given in Section \ref{sec:Future-Research5}.

\section{Related work\label{sec:Related-Works4}}

We review in this section some current methods for segmentation of
still images. For an extensive survey of current state-of-the-art
methods for dimensionality reduction the reader is referred to Chapter
\ref{cha:Dimensionality-reduction}.

Automatic segmentation of still images has been investigated by many
researchers from various fields of science. Segmentation methods can
be divided into the following groups: Graph-based techniques, boundary-based
methods, region-based methods and hybrid methods.

Shi and Malik \cite{SM00} propose a technique that solves image segmentation
as a graph partitioning problem. They define a global criterion, the
\emph{normalized cut}, which is optimized in order to partition the
graph. The graph partition induces a binary segmentation of the image.
Perona and Freeman \cite{perona98factorization} propose a fast affinity
approximation algorithm,%that is based on the normalized cuts algorithm
which takes into account only two groups, \emph{foreground} and \emph{background}.
The foreground group is computed as a factorized approximation of
the pairwise affinity of the elements in the scene.

Weiss \cite{weiss99} suggests a unified treatment of three affinity
matrix-based algorithms - Shi and Malik \cite{SM00}, Perona and Freeman
\cite{perona98factorization} and Scott and Longuet-Higgins \cite{ScottHiggins90}.
He shows close connections between them while pointing out their distinguishing
features. Their similarities lie in the fact that they all use the
eigenvectors of the pairwise similarity matrix. They differ by the
eigenvectors they look at and by the ways they normalize the affinity
matrix.

Fowlkes \emph{et al.} \cite{fowlkes-efficient} propose an approach
that is based on a numerical solution of eigenfunction problems, known
as the Nystr\"{o}m extension in order to reduce the high computational
cost of grouping algorithms that are based on spectral partitioning
like the normalized cuts. Belongie \emph{et al.} \cite{649340,fowlkes04spectral},
present a modification to the previous algorithm that does not require
the weighted adjacency matrix to be positive definite.

Boundary-based techniques find objects in the image via their contours.
The simplest form of such segmentation is to apply edge-detection
\cite{Gonzalez02}. More advanced techniques apply physics-based deformable
models that change under the laws of Newton mechanics, in particular,
by the theory of elasticity expressed in the Lagrange dynamics. In
general\emph{, }these models start from an initial shape and gradually
evolve to the contour of the image objects\emph{. }One type of these
shapes is \emph{snakes }(also known as \emph{active contours})\emph{,
}were first introduced by Kass \emph{et al}. \cite{Snake87g}. In
a later paper, Zhu \emph{et al}. \cite{SnakeUni} propose an algorithm
in which the segmentation is derived by minimizing a generalized Bayes/MDL
criterion using the variational principle and combining aspects of
snakes/balloons and region growing (region-based methods are described
below). In a more recent paper \cite{ChanVese06}, Chan and Vese propose
a model for active contours to detect objects, based on curve evolution,
the Mumford\textendash{}Shah functional \cite{MumfordShah} for segmentation
and level sets \cite{LevelSets}.%
\begin{comment}
(by using level sets, this method can also by classified as region-based
method, examples of which are described below)
\end{comment}
{} Their model can detect objects whose boundaries are not necessarily
defined by a gradient\emph{.}

Region-based methods group together similar pixels according to some
homogeneity criteria. They assume that pixels, which belong to the
same homogeneous region, are more alike than pixels from different
homogeneous regions. Examples for these methods are the split-and-merge
and the region-growing techniques \cite{CL94,Horowitz74}. In many
cases, these methods fail to produce satisfactory segmentation results
in the presence of noise.

Another approach to image segmentation uses histograms. However, most
of the histogram-based algorithms deal only with gray level images.
Dealing with 3D color histograms is a difficult task. The technique
in \cite{Underwood:1977:ICA} projects 3D color spaces onto 2D or
even 1-D spaces and segment the image according to the obtained surfaces.
Other techniques \cite{Sarabi81,1045267} transform the 3D histogram
into a binary tree where each node corresponds to a different range
of RGB values. 

Hybrid methods improve the segmentation results by combining the advantages
of different techniques. Hybrid techniques gain their advantage by
incorporating both global (histogram) and local (regions and boundaries)
information. The Watershed algorithm \cite{Vincent91} is an example
of a hybrid technique. It begins with a boundary based method and
continues with a region growing technique. Navon \emph{et al.} \cite{bb20529}
propose an iterative algorithm for segmentation of still color image,
which is based on adaptive automatic derivation of local thresholds.
The algorithm uses the segmentation result of the watershed algorithm
as an initial solution and continues with a region-growing technique.

Some algorithms, which are used for specific image processing tasks
such as segmentation, can be extended to handle hyper-spectral data
volumes. The normalized cuts algorithm \cite{SM00} can be extended
in a straight forward manner to handle hyper-spectral data. However,
this method uses a pixel similarity matrix without attempting to reduce
the dimensionality of the data, which renders it to be computational
expensive. Furthermore, the choice of pixel similarity metrics can
be further investigated to yield better results.

\section{The wavelength-wise global (WWG) algorithm for\protect \\
above pixel segmentation\label{sec:WWG}}

We introduce a two-phase approach for the segmentation of hyper-spectral
images. The first stage reduces the dimensionality of the data using
the DB algorithm and the second stage applies a histogram-based method
to cluster the low-dimensional data.

We model a hyper-spectral image as a three dimensional cube where
the first two coordinates correspond to the position $(x,y)$ and
the third coordinate corresponds to the wavelength $\lambda_{k}$.
Let 
\begin{equation}
I=\left\{ p_{ij}^{\lambda_{k}}\right\} _{i,j=1,\ldots,m;k=1,\ldots,n}\in\mathbb{R}^{m\times m\times n}\label{eq_hyper_image}
\end{equation}
be a hyper-spectral image cube, where the size of the image is $m\times m$,
the number of wavelengths is $n$ and $p_{ij}^{\lambda_{k}}$ is the
reflectance value at position $\left(i,j\right)$ at wavelength $\lambda_{k}$.
An example for a hyper-spectral image cube is illustrated in Fig.
\ref{cap:Hyper-spectral-cube}. For notation simplicity, we assume
that the images are square. It is important to note that almost always
$n\ll m^{2}$.

$I$ can be viewed in two ways:
\begin{enumerate}
\item \textbf{Wavelength-wise}: $I=\left\{ I^{\lambda_{l}}\right\} $ is
a collection of $n$ images of size $m\times m$ where 
\begin{equation}
I^{\lambda_{l}}\triangleq\left(p_{ij}^{\lambda_{l}}\right)\in\mathbb{R}^{m\times m},1\leq l\leq n\label{eq_wavelength_image}
\end{equation}
 is the image that corresponds to wavelength $\lambda_{l}$. 
\item \textbf{Point-wise}: $I=\left\{ \overrightarrow{I}_{ij}\right\} _{i,j=1}^{m}$
is a collection of $n$-dimensional vectors where 
\begin{equation}
\overrightarrow{I}_{ij}\triangleq\left(p_{ij}^{\lambda_{1}},\dots,p_{ij}^{\lambda_{n}}\right)\in\mathbb{R}^{n}\mbox{ , }1\leq i,j\leq m\label{eq_wavelength_vector}
\end{equation}
 is the hyper-pixel (see Section \ref{sec:Introduction6}) at position
$(i,j)$. 
\end{enumerate}
The proposed WWG algorithm assumes the wavelength-wise setting of
a hyper-spectral image. Note that the DB and DM algorithms receive
vectors as their inputs. Thus, we regard each image as a $m^{2}$-
dimensional vector. Formally, let
\begin{equation}
\breve{I}\triangleq\left(\pi_{i,\lambda_{l}}\right)_{i=1,\ldots,m^{2};l=1,\ldots,n}\in\mathbb{R}^{m^{2}\times n}\label{eq_hyper_image_matrix}
\end{equation}
 be a 2D matrix corresponding to $I$ where 
\[
\pi_{i+(j-1)\cdot m,\lambda_{k}}\triangleq p_{ij}^{\lambda_{k}}\mbox{ , }1\leq k\leq n\mbox{ , }1\leq i,j\leq m,
\]
($p_{ij}^{\lambda_{k}}$ is defined in Eq. \ref{eq_hyper_image}).
Each column in $\breve{I}$ corresponds to the image at wavelength
$\lambda_{k}$. We denote by
\begin{equation}
\breve{I}^{\lambda_{k}}\triangleq\left(\begin{array}{c}
\pi_{1,\lambda_{k}}\\
\vdots\\
\pi_{m^{2},\lambda_{k}}
\end{array}\right)\in\mathbb{R}^{m^{2}}\mbox{, }1\leq k\leq n\label{eq_column_vector}
\end{equation}
 the column vector that corresponds to $I^{\lambda_{k}}$ (see Eq.
\ref{eq_wavelength_image}).

\subsection{\label{sec:phase1-Dimensionality-Reduction}Phase 1: Reduction of
dimensionality via DB}

Different sensors can produce values at different scales. Thus, in
order to have a uniform scale for all the sensors, each column vector
$\breve{I}^{\lambda_{k}}\mbox{, }1\leq k\leq n$, is normalized to
be in the range {[}0,1{]}. Then, we form the set of vectors $\Gamma=\left\{ \breve{I}^{\lambda_{1}},\ldots,\breve{I}^{\lambda_{n}}\right\} $
from the columns of $\breve{I}$ and we execute Algorithm \ref{alg:Diffusion-Basis-Calculation}
on $\Gamma$. The dimensionality can also be reduced using Algorithm
\ref{alg:Modified-diffusion}. In this case, we will refer to the
segmentation algorithm as the \emph{modified}-WWG algorithm. In either
case, we denote the low-dimensional embedding by $\Gamma_{B}$ as
defined in \ref{eq:gamma_b}.

\subsection{\label{sec:phase2-Segmentation-by-colors}Phase 2: Segmentation by
colors }

We introduce a histogram-based segmentation algorithm that extracts
objects from $\Gamma$ using $\Gamma_{B}$. For notational convenience,
we denote $\eta\left(\delta\right)-1$ by $\eta$ hereinafter. We
denote by $G$ the cube representation of the set $\Gamma_{B}$ in
accordance with Eq. \ref{eq_hyper_image}:
\[
G\triangleq\left(g_{ij}^{k}\right)_{i,j=1,\ldots,m;k=1,\ldots,\eta}\mbox{ , }G\in\mathbb{R}^{m\times m\times\eta}.
\]
 We assume a wavelength-wise setting for $G.$ Let $\breve{G}$ be
a 2D matrix in the setting defined in Eq. \ref{eq_hyper_image_matrix}
that corresponds to $G$. Thus, the matrix $G^{l}\triangleq\left(g_{ij}^{l}\right)_{i,j=1,\ldots,m}\in\mathbb{R}^{m\times m}$,
$1\leq l\leq\eta$ corresponds to a column in $\breve{G}$ and $\overrightarrow{g}_{ij}\triangleq\left(g_{ij}^{1},\dots,g_{ij}^{\eta}\right)\in\mathbb{R}^{\eta}$,
$1\leq i,j\leq m$ corresponds to a row in $\breve{G}$. The coordinates
of $\overrightarrow{g}_{ij}$ will be referred to as \emph{colors}
from this point on.

The segmentation is achieved by clustering hyper-pixels with similar
colors. This is based on the assumption that objects in the image
will be composed of hyper-pixels that have similar \emph{color} vectors
in $\Gamma_{B}$. These colors contain the correlations between the
original hyper-pixels and the global inter-wavelength changes of the
image. Thus, homogeneous regions in the image have similar correlations
with the changes i.e. \emph{close} colors where closeness between
colors is measured by the Euclidean distance.

The segmentation-by-colors algorithm consists of the following steps:
\begin{enumerate}
\item \textbf{\label{alg_step:Normalization_and_uniform_quantization}Normalization
of the input image cube} $G$: \\
 First, we normalize each color layer of the image cube to be in {[}0,1{]}.
Let $G^{k}$ be the $k$-th ($k$ is the color index) color layer
of the image cube $G$. We denote by $\widehat{G}^{k}=\left(\widehat{g}_{ij}^{k}\right)_{i,j=1,\ldots,m}$
the normalization of $G^{k}$ and define it to be
\begin{equation}
\widehat{g}_{ij}^{k}\triangleq\frac{g_{ij}^{k}-\min\left\{ G^{k}\right\} \textrm{ }}{\max\left\{ G^{k}\right\} -\min\left\{ G^{k}\right\} },\,1\leq k\leq\eta.\label{eq:V_k}
\end{equation}

\item \textbf{Uniform quantization of the normalized image cube} $\widehat{G}$:
\\
 Let $l\in\mathbb{N}$ be the given number of quantization levels.
We uniformly quantize every value in $G^{k}$ to be one of $l$ possible
values. The quantized matrix is given by $Q$: 
\begin{equation}
Q\triangleq\left(q_{ij}^{k}\right)_{i,j=1,\ldots,m;k=1,\ldots,\eta}\mbox{ , }q_{ij}^{k}\in\left\{ 1,\ldots,l\right\} \label{eq_quantized_matrix}
\end{equation}
 where $q_{ij}^{k}=\left\lfloor l\cdot\widehat{g}_{ij}^{k}\right\rfloor $.
We denote the quantized \emph{color} vector at coordinate $(i,j)$
by
\begin{equation}
\overrightarrow{c}_{ij}\triangleq\left(q_{ij}^{1},\dots,q_{ij}^{\eta}\right)\in\mathbb{R}^{\eta}\mbox{ , }1\leq i,j\leq m.\label{eq_quantized_color_vector}
\end{equation}

\item \textbf{Construction of the frequency} \textbf{\emph{color}} \textbf{histogram:}
\\
 We construct the frequency function $f\colon\left\{ 1,\ldots,l\right\} ^{\eta}\rightarrow\mathbb{N}$
where for every $\kappa\in\left\{ 1,\ldots,l\right\} ^{\eta}$, $f\left(\kappa\right)$
is the number of quantized color vectors $\overrightarrow{c}_{ij}\mbox{ , }1\leq i,j\leq\eta$,
that are equal to $\kappa$. 
\item \textbf{Finding peaks in the histogram:} \\
We detect local maximum points (called \emph{peaks}) of the frequency
function $f$. We assume that each peak corresponds to a different
object in the image cube $G$. Here we use the classical notion of
segmentation - separating object from the background. Indeed, the
highest peak corresponds to the largest homogeneous area which is
mostly the background. The histogram may have many peaks. Therefore,
we apply an iterative procedure in order to find the $\theta$ highest
peaks where the number $\theta$ of the sought after peaks is given
as a parameter to the algorithm. This parameter corresponds to the
number of objects we are looking for. We assume that a peak and its
neighborhood in the histogram correspond to a single object in the
scene. Accordingly, the next peaks are sought after \emph{outside}
this neighborhood. In order to accomplish this, the algorithm is given
an integer parameter $\xi$, which specifies the size of the neighborhood.
This size is defined as the $l_{1}$ cube with radius $\xi$ around
a peak. Formally, we define the $\xi$-neighborhood of a coordinate
$\left(x_{1},\cdots,x_{\eta}\right)$ to be
\begin{equation}
N_{\xi}\left(x_{1},\cdots,x_{\eta}\right)=\left\{ \left(y_{1},\cdots,y_{\eta}\right)\left|\max_{k}\left\{ \left|y_{k}-x_{k}\right|\right\} \le\xi,\, x_{k},y_{k}\in\mathbb{N}\right.\right\} .\label{eq:neighborhood}
\end{equation}
The coordinates outside the neighborhood $N_{\xi}$ are the candidates
for the locations of new peaks. Thus, the next iterations find the
rest of the peaks. The peaks are labeled $1,\dots,\theta$. The output
of this step is a set of vectors 
\[
\Psi=\left\{ \overrightarrow{\rho}_{i}\right\} _{i=1,\dots,\theta},\,\overrightarrow{\rho}_{i}=\left(\rho_{i}^{1},\dots,\rho_{i}^{\eta}\right)\in\mathbb{N^{\eta}}
\]
that contains the colors that are associated with the highest peaks.
A summary of this step is given in algorithm \ref{alg:PeakFinder-Algorithm.}. 
\item \textbf{Finding the nearest peak to each} \textbf{\emph{color:}} \\
Once the highest peaks are found, each quantized \emph{color} vector
is associated with a single peak. The underlying assumption is that
the quantized \emph{color} vectors, which are associated with the
same peak, belong to the same object in the \emph{color} image-cube
$I$. Each quantized color is associated with the peak that is the
closest to it with respect to the Euclidean distance and it is labeled
by the number of its associated peak. We denote by 
\[
\gamma\colon\overrightarrow{c}_{ij}\mapsto d\in\left\{ 1,\ldots,\theta\right\} 
\]
this mapping function, where 
\[
\gamma\left(\overrightarrow{c}_{ij}\right)\triangleq\arg\min_{1\le k\le\theta}\left\{ \left\Vert \rho_{k}-\overrightarrow{c}_{ij}\right\Vert _{l_{\eta}}\right\} .
\]

\item \textbf{Construction of the output image:} \\
 The final step assigns a unique color $k{}_{i},\,1\le i\le\theta$
to each coordinate in the image according to its label $\gamma\left(\overrightarrow{c}_{ij}\right)$.
We denote the output image of this step by $\Omega=\left(\Omega_{ij}\right)$. 
\end{enumerate}
\begin{algorithm}[!h]
\textbf{PeaksFinder($f$, $\theta$, $\xi$)}
\begin{enumerate}
\item $\Psi\leftarrow\phi$ 
\item \textbf{while} $\left|\Psi\right|\le\theta$ 
\item ~~~~Find the next global maximum $c$ of $f$. 
\item ~~~~Add the coordinates of $c$ to $\Psi$. 
\item ~~~~Zero all the values of $f$ in the $\xi$-neighborhood of
$c$ (Eq. \ref{eq:neighborhood}). 
\item \textbf{end while} 
\item \textbf{return} $\Psi$. 
\end{enumerate}
\caption{The \emph{PeaksFinder} Algorithm. \label{alg:PeakFinder-Algorithm.}}
\end{algorithm}

\subsection{Hierarchical extension of the WWG algorithm}

We construct a hierarchical extension to the WWG algorithm in the
following way: given the output $\Omega$ of the WWG algorithm, the
user can choose one of the objects and apply the WWG algorithm on
the \emph{original} hyper-pixels which belong to this object. Let
$\chi$ be the color of the chosen object. We define $\Gamma\left(\chi\right)$
to be the set of the \emph{original} hyper-pixels which belong to
this object: 
\[
\Gamma\left(\chi\right)=\left\{ \left(p_{ij}^{1},\dots,p_{ij}^{n}\right)\left|\Omega_{ij}=\chi\right.,\,\, i,j=1,\dots,m\right\} .
\]
This facilitates a \emph{drill-down} function that enables a finer
segmentation of a \emph{specific} object in the image. We form the
set \textbf{$\Gamma\left(\chi\right)'$} from $\Gamma\left(\chi\right)$
as described in the beginning of Chapter \ref{cha:Diffusion-bases}
and run the \emph{DiffusionBasis} algorithm (Algorithm \ref{alg:Diffusion-Basis-Calculation}
or \ref{alg:Modified-diffusion}) on \textbf{$\Gamma\left(\chi\right)'$}.
Obviously, the size of the input is smaller than that of the original
image, thus allowing a finer segmentation of the chosen object. We
denote the result of this stage by $\Gamma_{B}\left(\chi\right)$.
Next, the WWG is applied on $\Gamma_{B}\left(\chi\right)$ and the
result is given by $\Omega\left(\chi\right)$. The drill-down algorithm
is outlined in Algorithm \ref{cap:Iterative-segmentation-algorithm}.
This step can be applied on other objects in the image as well as
on the drill-down result.

\begin{algorithm}[!h]
\textbf{DrillDown($\Gamma'$, $w_{\varepsilon}$, $\varepsilon$,
$\delta$ ,$\chi$)}
\begin{enumerate}
\item $\Gamma_{B}$= \textbf{DiffusionBasis($\Gamma'$, $w_{\varepsilon}$,
$\varepsilon$, $\delta$)~~~~~~~~~~~~~~ //} Algorithm
\ref{alg:Diffusion-Basis-Calculation} 
\item $\Omega$= \textbf{WWG}($\Gamma_{B}$)~~~~~~~~~~~~~~~~~~~~~~~~~~~~~~~~~~~//
Described in Sec. \ref{sec:WWG} 
\item $\Gamma_{B}\left(\chi\right)$= \textbf{DiffusionBasis($\Gamma\left(\chi\right)'$,
$w_{\varepsilon}$, $\varepsilon$, $\delta$)~~~~~~~~~//}
Algorithm \ref{alg:Diffusion-Basis-Calculation} 
\item $\Omega\left(\chi\right)$= \textbf{WWG}($\Gamma_{B}\left(\chi\right)$)~~~~~~~~~~~~~~~~~~~~~~~~~~~//
Described in Sec. \ref{sec:WWG} 
\end{enumerate}
\caption{A drill-down segmentation algorithm \label{cap:Iterative-segmentation-algorithm}}
\end{algorithm}

\subsection{Sub-pixel segmentation}

The first steps of the sub-pixel segmentation procedure are similar
to those that are used by the above-pixel segmentation: the dimensionality
reduction is used as described in Section \ref{sec:phase1-Dimensionality-Reduction}
and $\Gamma_{B}$ is normalized according to Eq. \ref{eq:V_k}.

Let $\widehat{G}=\left\{ \widehat{G}^{k}\right\} _{k=1}^{\eta}$ be
the normalized 3D image volume in the dimension-reduced space as described
in Eq. \ref{eq:V_k} and let $\widehat{g}_{ij}$ be a hyper-pixel
in $\widehat{G}$. Sub-pixel segments have high contrast with hyper-pixels
in their local neighborhood. This is manifested by differences in
several of their colors (see Section \ref{sec:phase2-Segmentation-by-colors}).
This is due to the difference in their correlations with the vectors
in the diffusion basis. We refer to such color points as \emph{isolated
points}. The following steps are applied in order to detect isolated
points.

We define the $\alpha$-neighborhood of $\widehat{g}_{ij}^{k}$ to
be 
\[
\alpha\left(\widehat{g}_{ij}^{k}\right)\triangleq\left\{ \widehat{g}_{mn}^{k}\right\} _{m=i-\alpha,\dots,i+\alpha,n=j-\alpha,\dots,j+\alpha}.
\]
 We compute the number of pixels in $\alpha\left(\widehat{g}_{ij}^{k}\right)$
whose differences from $\widehat{g}_{ij}^{k}$ are above a given threshold
$\tau_{1}$. We denote this number by
\[
\Delta_{\alpha}\left(\widehat{g}_{ij}^{k}\right)\triangleq\left|\left\{ \widehat{g}_{mn}^{k}:\left|\widehat{g}_{ij}^{k}-\widehat{g}_{mn}^{k}\right|>\tau_{1},\,\,\widehat{g}_{mn}^{k}\in\alpha\left(\widehat{g}_{ij}^{k}\right)\right\} \right|.
\]
 If the size of $\Delta_{\alpha}\left(\widehat{g}_{ij}^{k}\right)$
is larger than a given threshold $\tau_{2}$ then $\widehat{g}_{ij}^{k}$
is classified as an isolated point. $\tau_{2}$ determines the number
of pixels that are different from $\widehat{g}_{ij}^{k}$ in the neighborhood
of $\widehat{g}_{ij}^{k}$ in order for $\widehat{g}_{ij}^{k}$ to
be classified as an isolated point.

A coordinate \emph{$\left(i,j\right)$} is classified as a \emph{sub-pixel
segment} if there are $k_{1}$ and $k_{2}$ such that $\widehat{g}_{ij}^{k_{1}}$
and $\widehat{g}_{ij}^{k_{2}}$ are isolated points. The requirement
that a coordinate will contain an isolated point in at least \emph{two}
images prevents misclassification of noisy isolated points as sub-pixel
segments.

\section{Experimental results\label{sec:Experimental-Results5}}

The results are divided into three parts: (a) segmentation of hyper-spectral
microscopy images; (b) segmentation of remote-sensed hyper-spectral
images; (c) sub-pixel segmentation of remote-sensed images. We provide
the results using the two dimensionality reduction schemes that were
described in Algorithms \ref{alg:Diffusion-Basis-Calculation} and
\ref{alg:Modified-diffusion}.

We denote the size of the hyper-spectral images by $m\times m\times n$
where the size of every wavelength image is $m\times m$ and $n$
is the number of wavelengths. The geometry (objects, background, etc.)
of each hyper-spectral image is displayed using a gray image $\Upsilon$.
This image is obtained by averaging the hyper-spectral image along
the wavelengths. Given a hyper-spectral image $I$ of size $m\times m\times n$,
$\Upsilon=\left\{ \upsilon_{ij}\right\} _{i,j=1,\ldots,m}$ is obtained
by 
\[
\upsilon_{ij}=\frac{1}{n}\sum_{k=1}^{n}I_{ij}^{k}\,\,\,1\le i,j\le m.
\]
 We refer to $\Upsilon$ as the \emph{wavelength-averaged-version}
(WAV) of the image. All the results were obtained using the automatic
procedure for choosing $\varepsilon$ which is described in Section
\ref{sec:Choosing_epsilon}.

\paragraph{Segmentation~of~hyper-spectral~microscopy~images }

\begin{figure*}[!t]
\begin{centering}
\begin{tabular}{>{\centering}m{0.48\textwidth}||>{\centering}m{0.48\textwidth}}
\multicolumn{1}{c}{
\includegraphics[%bb=141bp 297bp 467bp 554bp,clip,
width=0.4\textwidth,keepaspectratio]{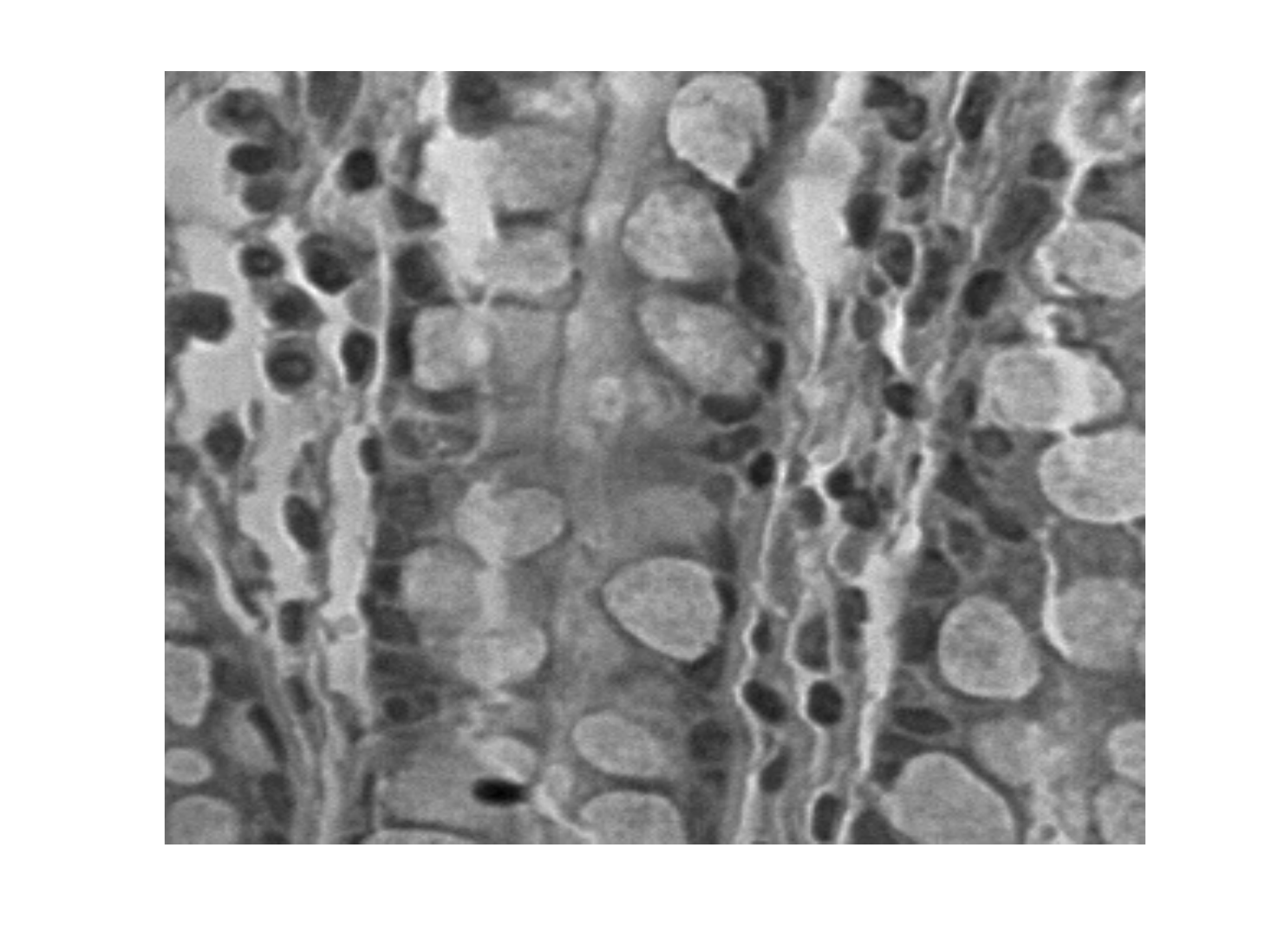}} &\multicolumn{1}{c}{}\tabularnewline
\multicolumn{1}{c}{
\includegraphics[%bb=141bp 297bp 467bp 554bp,clip,
width=0.4\textwidth,keepaspectratio]{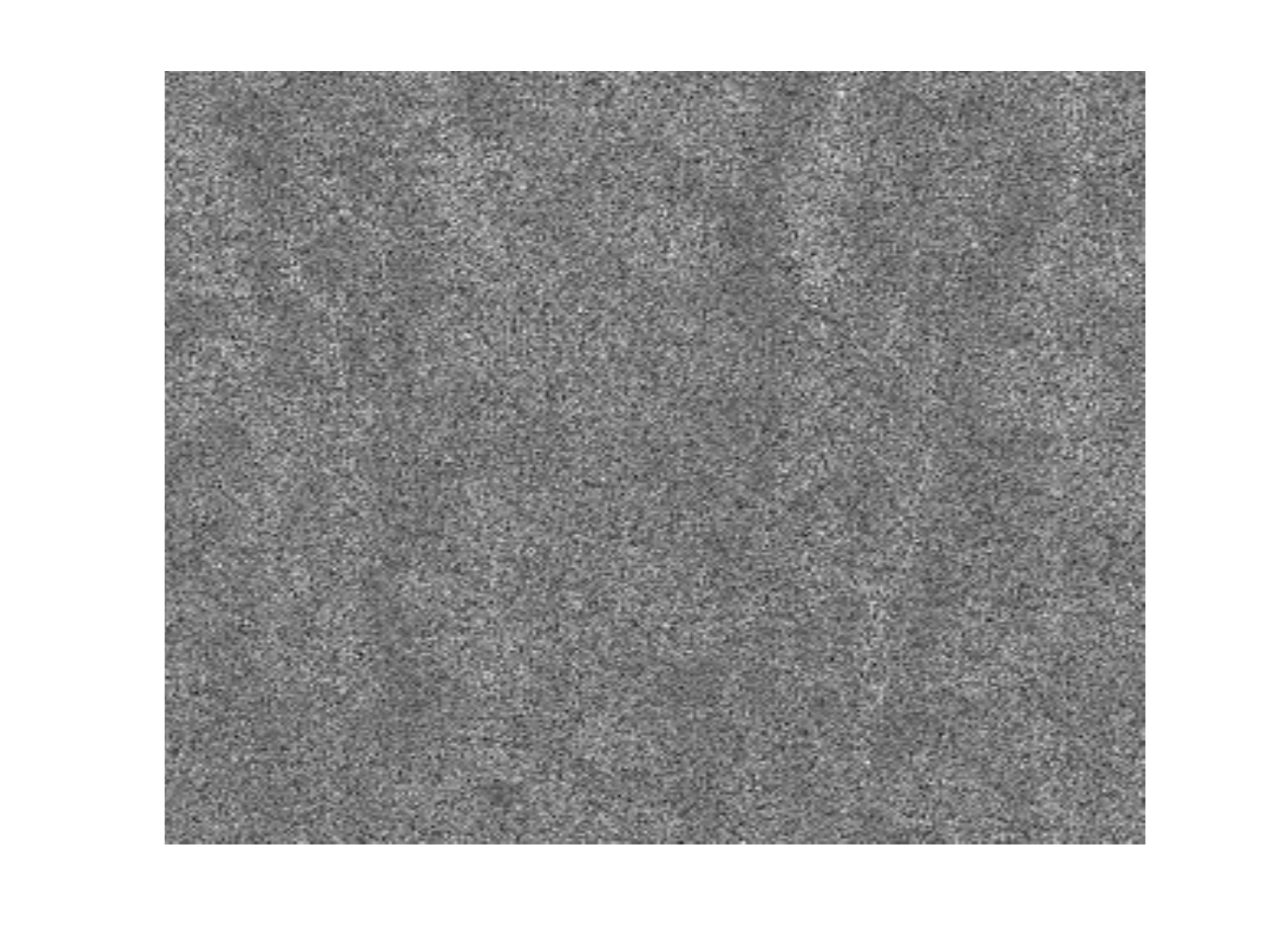}} &\multicolumn{1}{c}{}\tabularnewline
\multicolumn{1}{c}{
\includegraphics[%bb=143bp 298bp 466bp 553bp,clip,
width=0.4\textwidth,keepaspectratio]{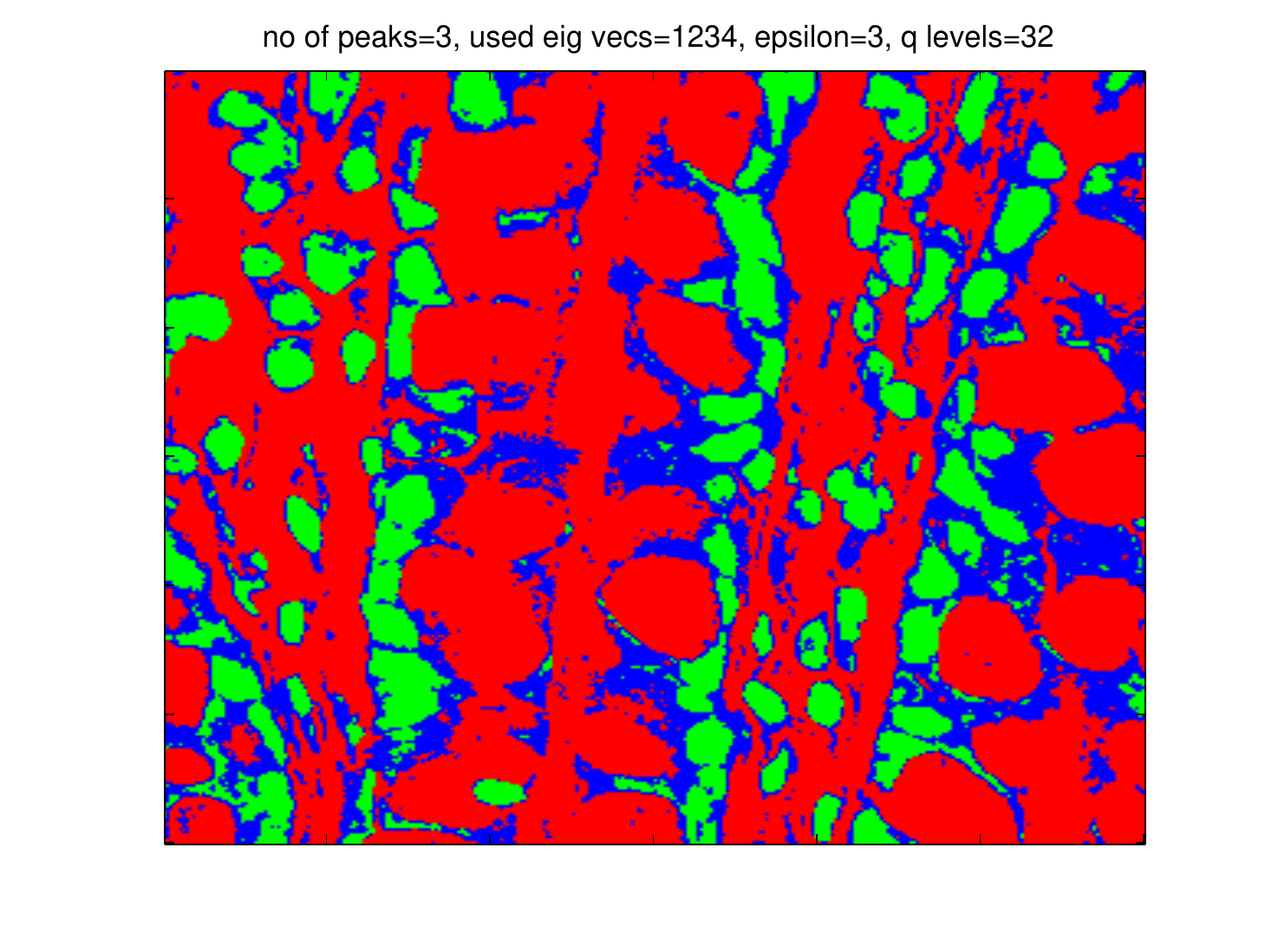}} &\multicolumn{1}{c}{}\tabularnewline
\end{tabular}
\par\end{centering}

\caption{\label{fig:BIG-micro}A hyper-spectral microscopy image of a healthy
human tissue. (a) The WAV of the original image. (b) The $50^{th}$
wavelength. (c) The $95^{th}$ wavelength. (d) The results of applying
the WWG algorithm with $\eta\left(\delta\right)=4,\,\theta=3,\,\xi=3,\, l=32$.
(e) The results of applying the modified-WWG algorithm with $\eta\left(\delta\right)=4,\,\theta=3,\,\xi=1,\, l=16$. }
\end{figure*}

\begin{figure*}[!t]
\begin{centering}
\begin{tabular}{>{\centering}m{0.48\textwidth}||c}
\multicolumn{1}{c}{
\includegraphics[%bb=142bp 298bp 467bp 554bp,clip,
width=0.3\textwidth,keepaspectratio]{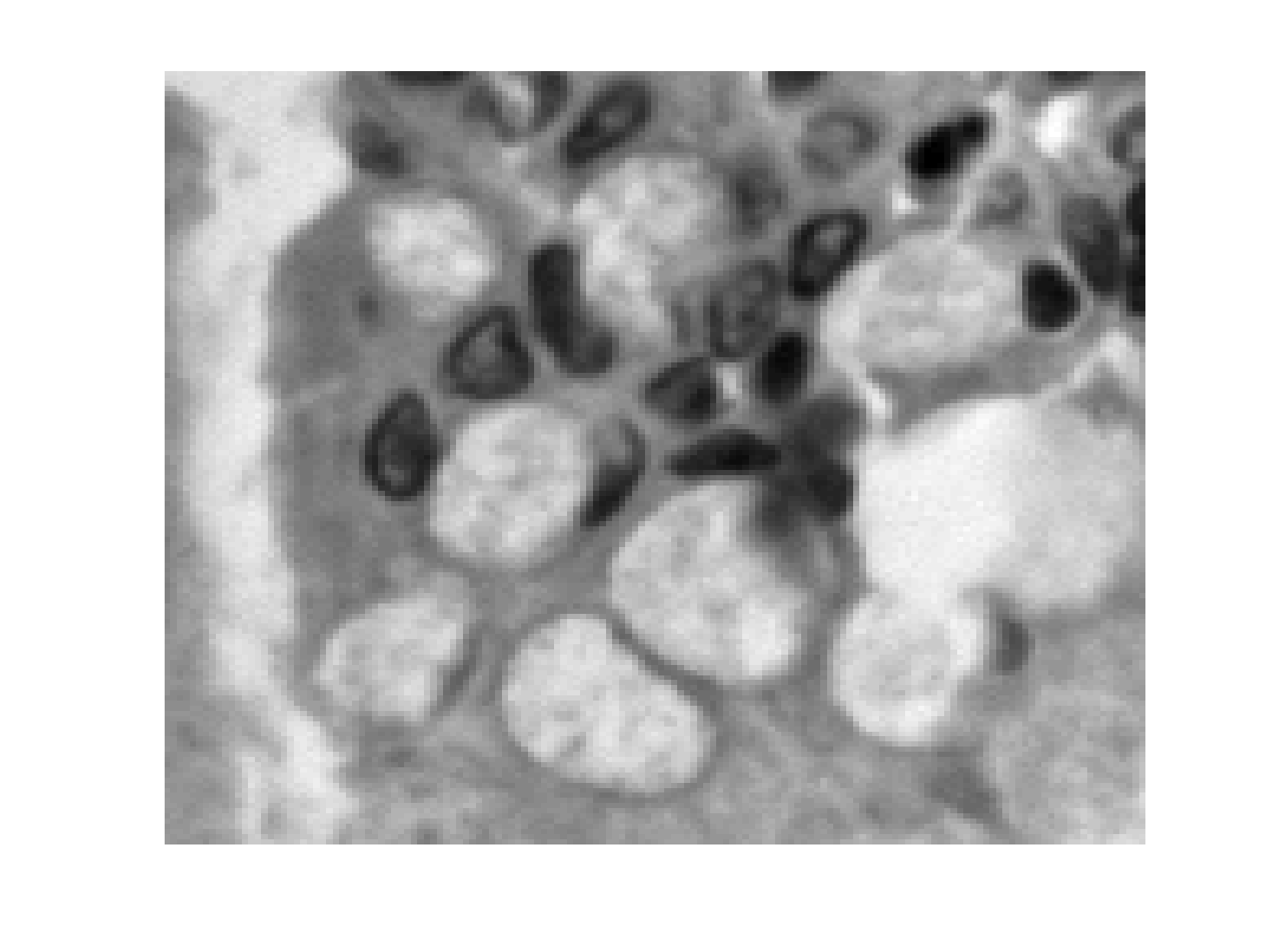}} 
& \multicolumn{1}{c}{}\tabularnewline
\multicolumn{1}{c}{
\includegraphics[%bb=142bp 298bp 467bp 554bp,clip,
width=0.3\textwidth,keepaspectratio]{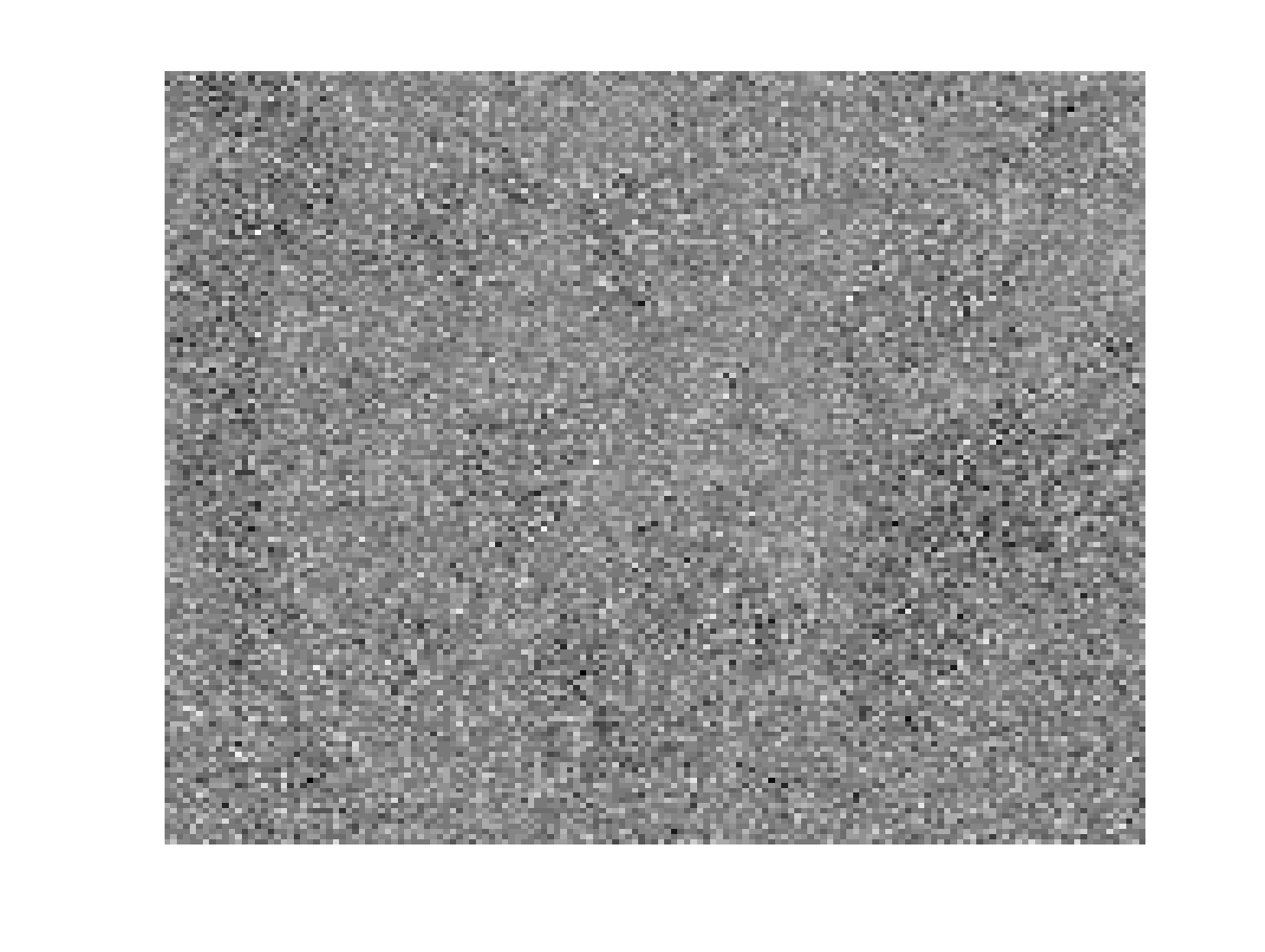}}& \multicolumn{1}{c}{}\tabularnewline
\multicolumn{1}{c}{
\includegraphics[%bb=91bp 225bp 541bp 581bp,clip,
width=0.3\textwidth,keepaspectratio]{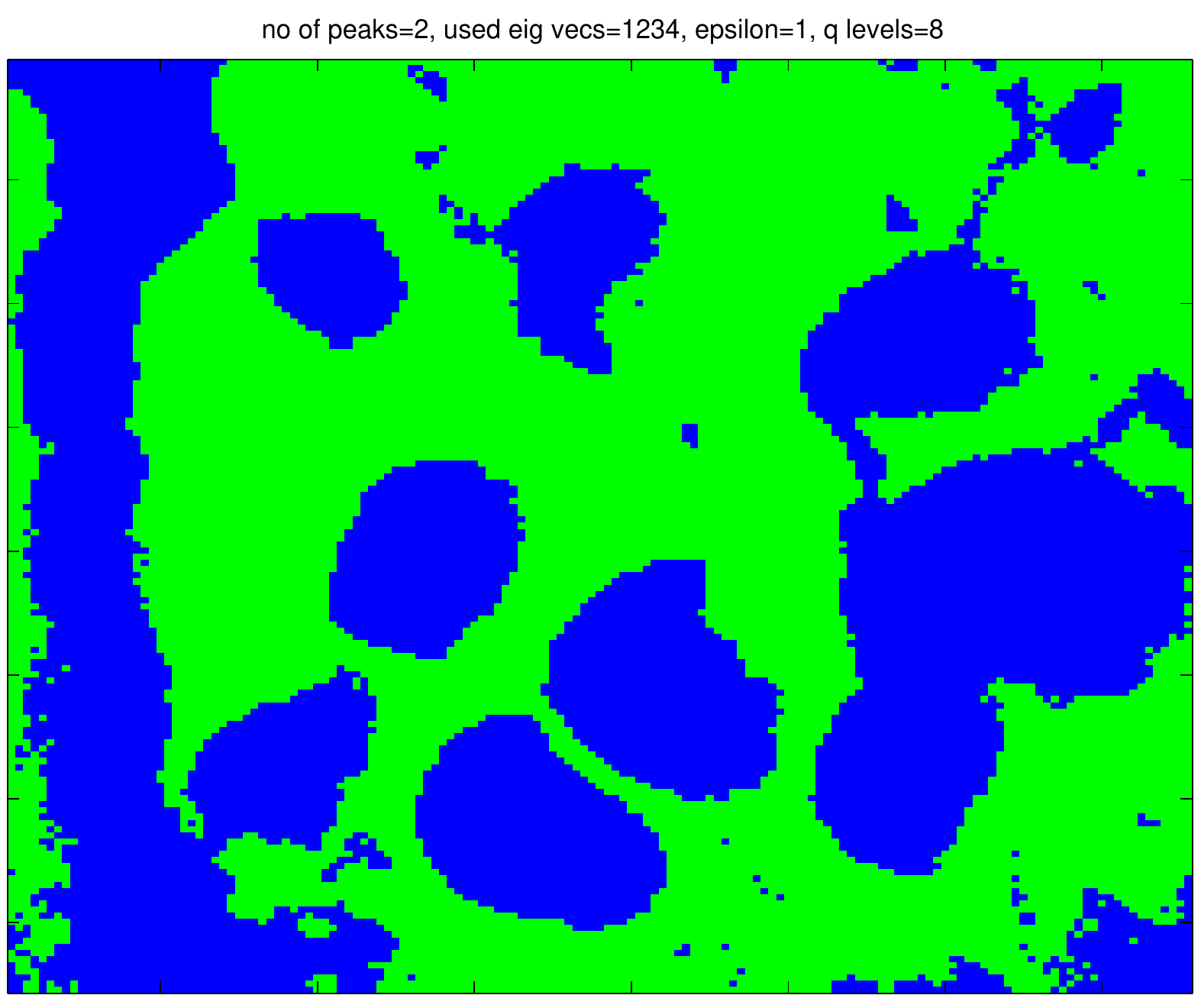}}& \multicolumn{1}{c}{}\tabularnewline
\multicolumn{1}{c}{
\includegraphics[%bb=140bp 295bp 468bp 555bp,clip,
width=0.3\textwidth,keepaspectratio]{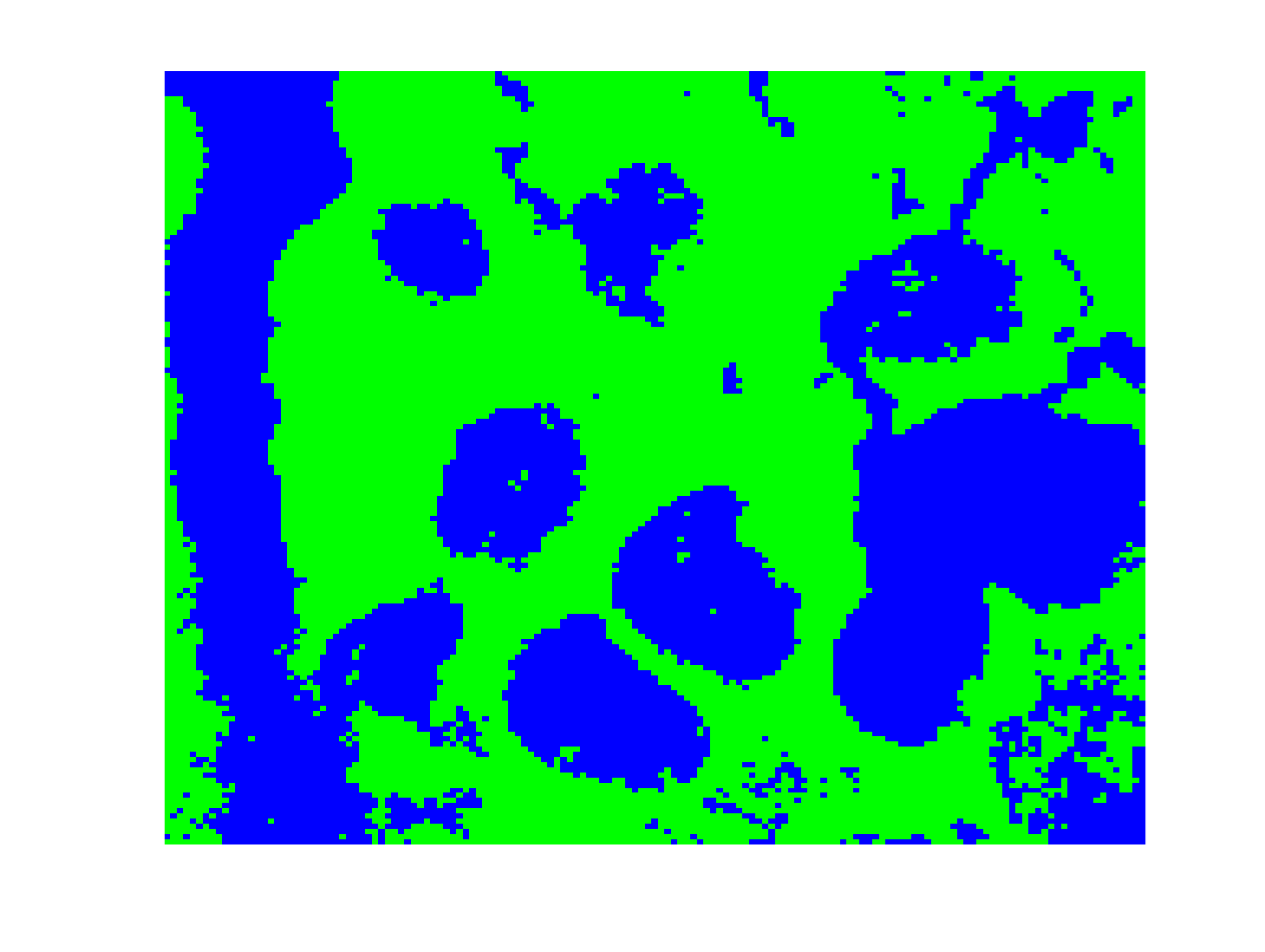}}& \multicolumn{1}{c}{}\tabularnewline
\end{tabular}
\par\end{centering}

\caption{A hyper-spectral microscopy image of a healthy human tissue. (a) The
WAV of the original image. (b) The $40^{th}$ wavelength. (c) The
$107^{th}$ wavelength. (d) Results of applying the k-mean algorithm
to (a). (e) The results after the first iteration of the WWG algorithm
using $\eta\left(\delta\right)=2,\,\theta=4,\,\xi=1,\, l=8$. (f)
The results after the second iteration of the WWG algorithm on the
hyper-pixels in the green area of (e) using $\eta\left(\delta\right)=2,\,\theta=4,\,\xi=6,\, l=32$.
(g) The results after the first iteration of the modified-WWG with
$\eta\left(\delta\right)=3,\,\theta=2,\,\xi=3,\, l=16$. (h) The results
after the second iteration of the modified-WWG on the hyper-pixels
in the green area of (g) using $\eta\left(\delta\right)=3,\,\theta=2,\,\xi=1,\, l=8$.
Both (e) and (g) exhibit better inter-nuclei separation than (d).
\label{fig:Example1x}}
\end{figure*}

Figures \ref{fig:BIG-micro} and \ref{fig:Example1x} contain samples
of healthy human tissues and the results after applying the WWG algorithm
to them. The images are of sizes $300\times300\times128$ and $151\times151\times128$,
respectively. The images contain three types of substances: cell nuclei,
glandular cell cytoplasm and lamina propria cell cytoplasm.

Figure \ref{fig:BIG-micro}(a) shows the WAV of the image. Figures
\ref{fig:BIG-micro}(b) and \ref{fig:BIG-micro}(c) show the $50^{th}$
and $95^{th}$ wavelengths, respectively. The images in the $50^{th}$
through the $70^{th}$ wavelengths are less noisy than the rest. Figures
\ref{fig:BIG-micro}(d) and \ref{fig:BIG-micro}(e) display the results
after the application of the WWG and the modified-WWG algorithms,
respectively. The algorithm clearly segments this image into three
parts: the lamina propria cell cytoplasm is colored in blue, the glandular
cell cytoplasm is colored in red and the cell nuclei is colored in
green.

Figure \ref{fig:Example1x}(a) shows the WAV of the image. Figures
\ref{fig:Example1x}(b) and \ref{fig:Example1x}(c) show the $40^{th}$
and $107^{th}$ wavelengths, respectively. The images in the $40^{th}$
through the $55^{th}$ wavelengths are less noisy than the rest. Figure
\ref{fig:Example1x}(d) shows results after applying the k-mean algorithm
to (a). Figures \ref{fig:Example1x}(e) and \ref{fig:Example1x}(f)
display the results of the application of the \emph{hierarchical}
extension of the WWG algorithm while Figs. \ref{fig:Example1x}(g)
and \ref{fig:Example1x}(h) display the results of the application
of the \emph{hierarchical} extension of the modified-WWG algorithm.
Figures \ref{fig:Example1x}(e) and \ref{fig:Example1x}(g) depict
the first iteration of the WWG algorithm and the modified-WWG algorithm,
respectively. The second iteration receives as input the hyper-pixels
that are in the green region of Figs. \ref{fig:Example1x}(f) and
\ref{fig:Example1x}(h). The results of the second iteration of the
WWG algorithm and the modified-WWG algorithm are shown in Figs. \ref{fig:Example1x}(f)
and \ref{fig:Example1x}(h), respectively. The image is clearly segmented
into three parts. The lamina propria cell cytoplasm is colored in
green, the glandular cell cytoplasm is colored in blue and the cell
nuclei is colored in red. Both (e) and (g) exhibit better inter-nuclei
separation than (d).
%\multicolumn{1}{c}{\includegraphics[bb=141bp 297bp 467bp 554bp,clip,width=0.45\textwidth,keepaspectratio]{DB_paper/Big300_band50}} & \multicolumn{1}{c}{}\tabularnewline
%\multicolumn{2}{c}{}\tabularnewline
%\multicolumn{1}{c}{} & \multicolumn{1}{c}{}\tabularnewline
%\end{tabular}
%\par\end{centering}
%
%\caption{Two wavelengths of a hyper-spectral microscopy image of a healthy
%human colon tissue.\label{fig:Two-wavelengths-of}}
%\end{figure*}

\paragraph{Segmentation~of~remote-sensed~images}

\begin{figure*}[!t]
\begin{centering}
\begin{tabular}{>{\centering}m{0.48\columnwidth}||>{\centering}m{0.48\columnwidth}}
\multicolumn{1}{c}{

\includegraphics[%bb=141bp 297bp 467bp 554bp,clip,
width=0.4\textwidth,keepaspectratio]{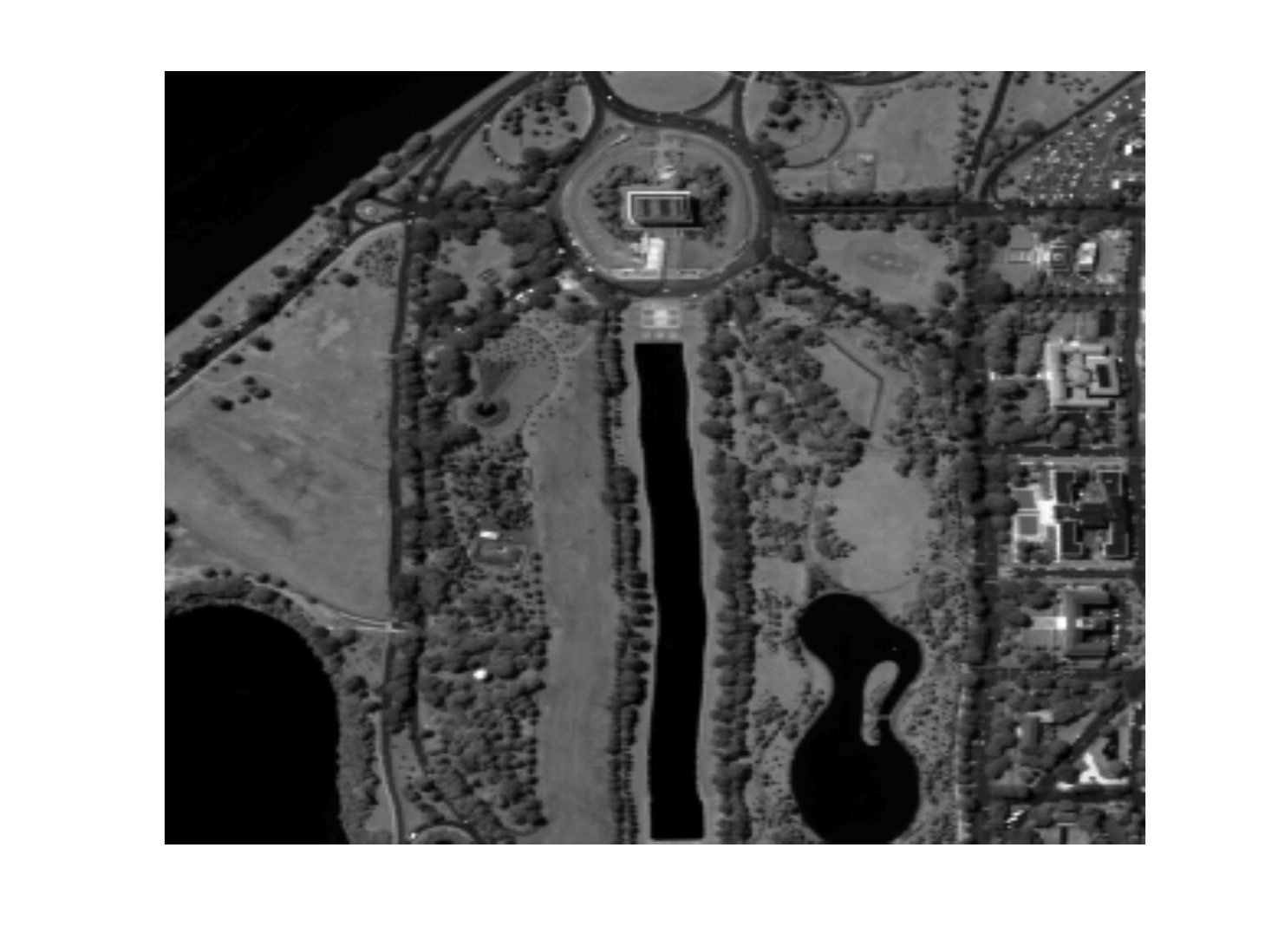}}& \multicolumn{1}{c}{}\tabularnewline
\multicolumn{1}{c}{

\includegraphics[%bb=141bp 297bp 467bp 554bp,clip,
width=0.4\textwidth,keepaspectratio]{dcmall_band80}}& \multicolumn{1}{c}{}\tabularnewline
\multicolumn{1}{c}{

\includegraphics[%bb=141bp 297bp 467bp 554bp,clip,
width=0.4\textwidth,keepaspectratio]{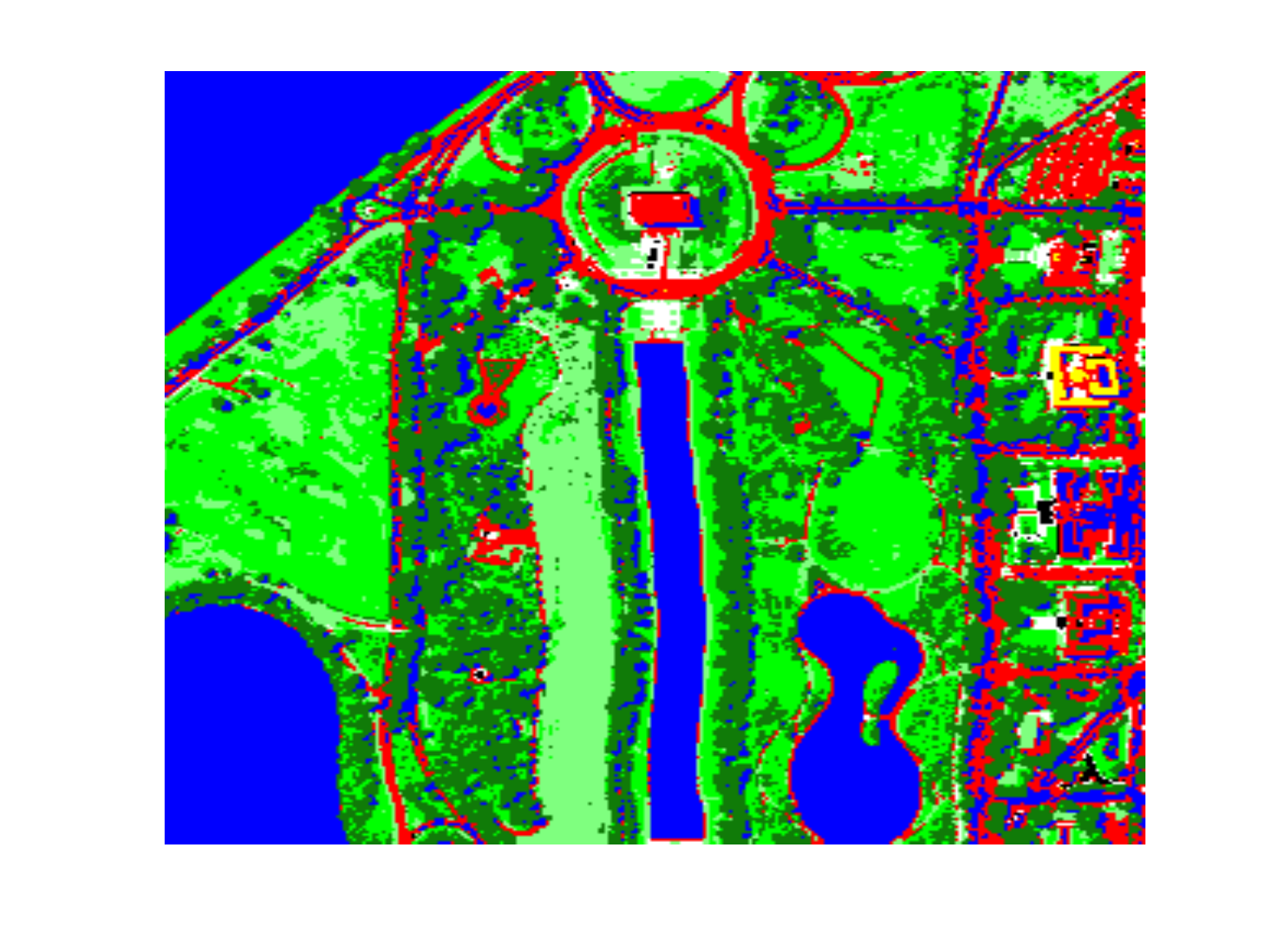}}& \multicolumn{1}{c}{}\tabularnewline
\end{tabular}
\par\end{centering}

\caption{A hyper-spectral satellite image of Washington DC mall. (a) The WAV of the image. The image contains water,
two types of grass, trees, two types of roofs, roads, trails and shadow.
(b) The $10^{th}$ wavelength. (c) The $80^{th}$ wavelength. (d)
The result after the application of the WWG algorithm using $\eta\left(\delta\right)=4,\,\theta=8,\,\xi=7,\, l=32$.
(e) The result after the application of the modified-WWG algorithm
using $\eta\left(\delta\right)=4,\,\theta=8,\,\xi=7,\, l=32$. The
water is colored in blue, the grass is colored in two shades of light
green, the trees are colored in dark green, the roads are colored
in red, the roofs are colored in pink and yellow, the trails are colored
in white and the shadow is colored in black.\label{fig:DC_mall}}
\end{figure*}

Figure \ref{fig:DC_mall} contains a hyper-spectral satellite image
of Washington DC mall and the result
after applying the WWG algorithm to it. The image is of size $300\times300\times100$.
Figure \ref{fig:DC_mall}(a) shows the WAV of the image. The image
contains water, two types of grass, trees, two types of roofs, roads,
trails and shadow. Figures \ref{fig:DC_mall}(b) and \ref{fig:DC_mall}(c)
show the $10^{th}$ and $80^{th}$ wavelengths, respectively. Figures
\ref{fig:DC_mall}(d) and \ref{fig:DC_mall}(e) are the results of
the WWG algorithm and modified-WWG algorithm, respectively, where
the water is colored in blue, the grass is colored in two shades of
light green, the trees are colored in dark green, the roads are colored
in red, the roofs are colored in pink and yellow, the trails are colored
in white and the shadow is colored in black.

\paragraph{Sub-pixel~segmentation: }

The results of the application of the sub-pixel segmentation algorithm
on a $300\times300\times121$ hyper-spectral image of a mountain terrain
are given in Fig. \ref{fig:sub_pixel}. The image was taken from a
high altitude airplane and it contains twenty four sub-pixel segments.
Figure \ref{fig:sub_pixel}(a) shows the WAV of the image. Figures
\ref{fig:sub_pixel}(b) and \ref{fig:sub_pixel}(c) show the $35^{th}$
and $50^{th}$ wavelengths, respectively. The noise in the image is
caused by atmosphere conditions and calibration errors of the hyper-spectral
camera. These problems can be found in most of the wavelengths. Figure
\ref{fig:sub_pixel}(d) displays $\widehat{G}^{2}$ with squares around
the sub-pixel segments that were found. The dimensionality was reduced
using Algorithm \ref{alg:Modified-diffusion} with $\eta\left(\delta\right)=6$.
The sub-pixel segmentation was obtained using $\tau_{1}=0.04,\,\tau_{2}=3$.
The algorithm detects \emph{all} twenty four segments.

Our algorithm is \emph{highly robust} to noise as it is demonstrated
by the results at the presence of noisy wavelengths which are depicted
in Figs. \ref{fig:Example1x}(b)-\ref{fig:sub_pixel}(b) and \ref{fig:Example1x}(c)-\ref{fig:sub_pixel}(c).

\begin{center}
\begin{figure*}[!t]
\begin{centering}
\begin{tabular}{>{\centering}m{0.48\columnwidth}||>{\centering}m{0.48\columnwidth}}
\multicolumn{1}{c}{

\includegraphics[%bb=141bp 297bp 467bp 554bp,clip,
width=0.45\textwidth,keepaspectratio]{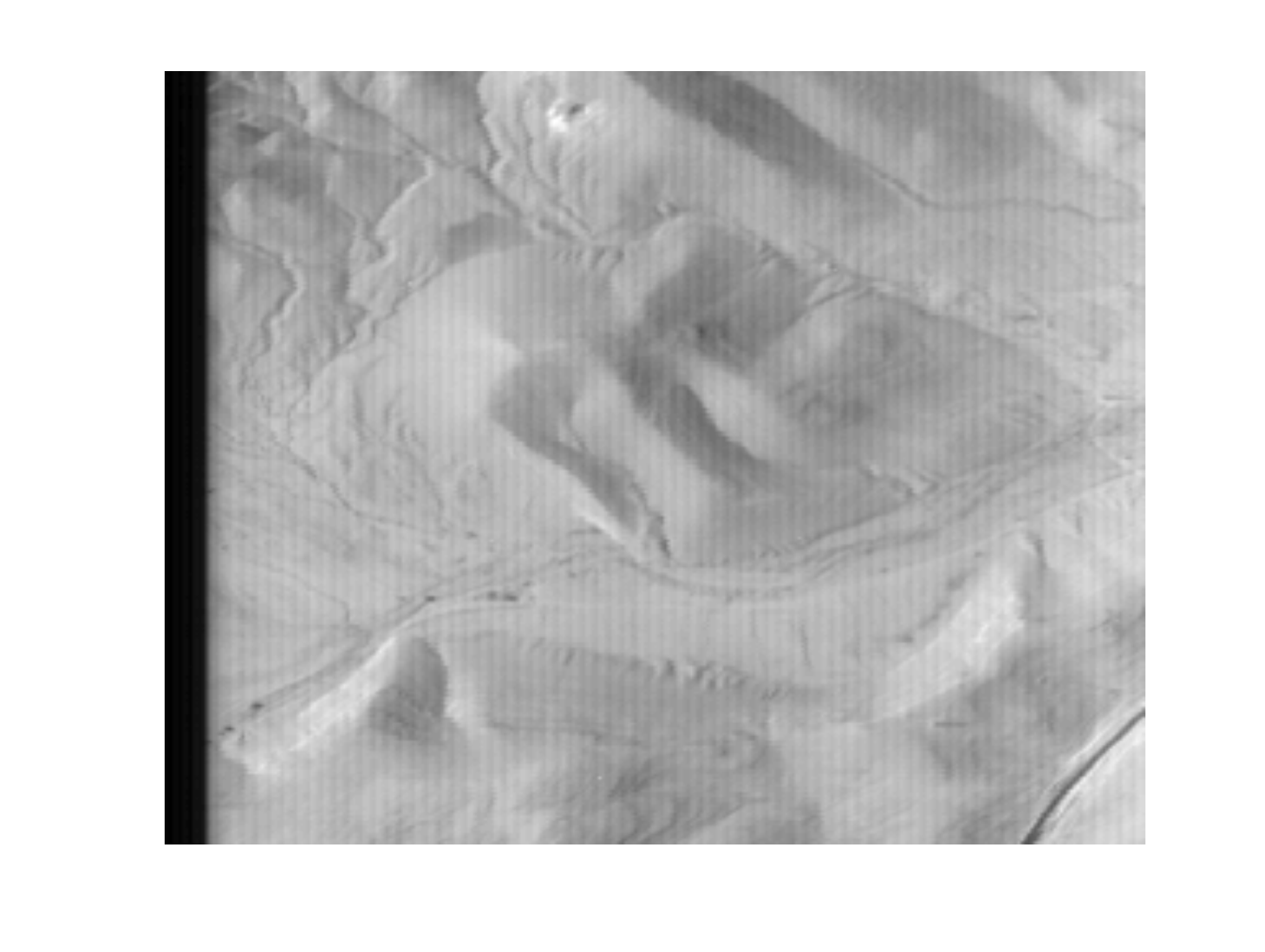}}& \multicolumn{1}{c}{}\tabularnewline
\multicolumn{1}{c}{

\includegraphics[%bb=141bp 297bp 467bp 554bp,clip,
width=0.45\textwidth,keepaspectratio]{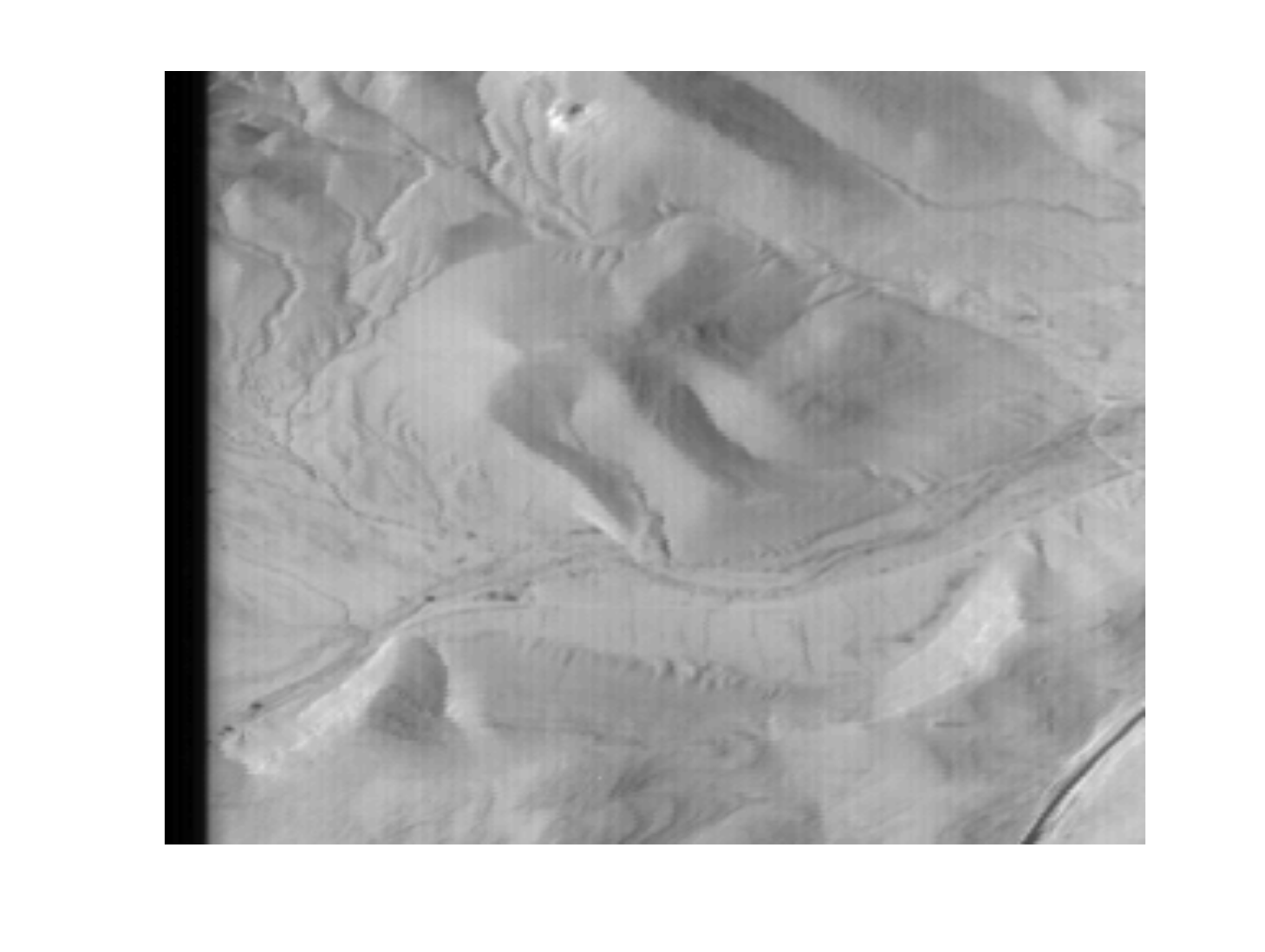}}& \multicolumn{1}{c}{}\tabularnewline
\end{tabular}
\par\end{centering}

\caption{A hyper-spectral image of a mountain terrain which contains 24 sub-pixel
segments. (a) The WAV of the image. (b) The $35^{th}$ wavelength.
(c) The $50^{th}$ wavelength. (d) $\widehat{G}^{2}$ with squares
around the detected sub-pixel segments. The parameters that were used
are: $\eta\left(\delta\right)=6$, $\tau_{1}=0.04$, $\tau_{2}=3$.\label{fig:sub_pixel}}
\end{figure*}

\par\end{center}

\section{Conclusion}

In this chapter, a novel algorithm for the segmentation of hyper-spectral
images was introduced. The algorithm utilizes the diffusion bases
algorithm which is described in Chapter \ref{cha:Diffusion-bases}
in order to reduce the dimensionality of the data. Two segmentation
algorithms were proposed: one for the segmentation of the dimension-reduced
data and the other for the detection of sub-pixel segments i.e. anomalies.
The algorithm was successfully applied to images produced by a hyper-spectral
microscope and also to remote-sensed images. This demonstrates the
broad range of domains the proposed algorithms can tackle.

\section{Discussion\label{sec:Future-Research5}}

The results in Section \ref{sec:Experimental-Results5} were obtained
using a Gaussian kernel. It was shown in \cite{CL_DM06} that any
positive semi-definite kernel may be used for the dimensionality reduction.
Rigorous analysis of families of kernels to facilitate the derivation
of an optimal kernel for a given set $\Gamma$ is an open problem
and is currently being investigated by the authors.

Every hyper-spectral image cube that we tested in Section \ref{sec:Experimental-Results5}
produced a different set of eigenvalues. We assume that the eigenvalues
play a vital role in achieving the dimensionality reduction while
having unique physical properties. Analysis of the behavior of the
eigenvalues is crucial. As a result of this analysis, an automatic
algorithm may be constructed for choosing the eigenvectors which will
produce the best segmentation results.

The parameter $\eta\left(\delta\right)$ determines the dimensionality
of the diffusion space. Automatic choice of this threshold is vital
in order to detect the objects in the image cube. A rigorous way for
choosing $\eta\left(\delta\right)$ is currently being studied by
the authors. Naturally, $\eta\left(\delta\right)$ is \emph{data driven}
(similarly to $\varepsilon$) i.e. it depends on the set $\Gamma$
at hand.

Detection of sub-pixel objects in remote-sensed images is a preliminary
step to their identification. These sub-pixel objects are composed
of several materials that are mixed. Their identification requires
an unmixing algorithm that separates the different substances of the
object.

\chapter{Neuronal Tissues Sub-Nuclei Segmentation Using Multi-Contrast MRI\label{cha:Neuronal-Tissues-Sub-Nuclei}}

Many brain surgeries, forebrain functionality mapping, research of
brain-related diseases and additional various pathological procedures
rely on a sub-nuclei mapping of the examined neuronal tissue. Using
contemporary acquisition tools, to construct a reliable \emph{in-vivo}
sub-nuclei mapping is a highly complex task. 

In this chapter we proposes a novel algorithm, called \emph{Sub-Nuclei
Finder} (SNF), which segments neuronal tissues into their sub-nuclei
arrangement. Using multi-contrast Magnetic Resonance Imaging (MRI),
we are able to separate a neuronal tissue into its sub-nuclei in 3D
space. Our approach includes image enhancement, non-linear dimensionality
reduction and clustering in 3D space. For the image enhancement, we
use a \emph{wavelet}-based method, which reduces the undesired noise
and simultaneously emphasizes and amplifies features of interest.
As for the dimensionality reduction procedure, we use the frameworks
of \emph{Diffusion Maps} (Chapter \ref{cha:Diffusion-Maps}) and \emph{Diffusion
Bases} (Chapter \ref{cha:Diffusion-bases}) to achieve a non-linear
local-geometry-driven reduction of dimensionality. In order to segment
the enhanced low-dimensional data, we use a \emph{Silhouette-Driven
K-Means} (SDK) algorithm, which automatically identifies the number
of clusters for the desired separation. 

We demonstrate our algorithm on the \emph{Thalamus} sub-part of the
human brain. Our algorithm successfully segments the human thalamus
into its sub-nuclei in strong agreement with a known histological
atlas. Moreover, the results on a number of healthy subjects are highly
correlated.

\section{Introduction and related work}

We introduce a novel framework for performing a sub-nuclei segmentation
of brain tissue. The proposed framework is comprised of a special
MRI-based acquisition method and a new algorithm for the segmentation
task. In this section, we survey a number of issues that are relevant
to this framework. We explore each issue including its significant
related work. We start by describing in Sections \ref{sub:The-Problem-of}
and \ref{sub:Sub-Nuclei-Segmentation-Using} the problem of sub-nuclei
segmentation that motivates this chapter. Then, we explore the issues
that are relevant to the proposed \emph{Sub-Nuclei Finder} (SNF) algorithm,
which are: image enhancement in Section \ref{sub:Image-Enhancement}
and cluster analysis in Section \ref{sub:Cluster-Analysis}. Dimensionality
reduction is also a key part of the proposed scheme and it was explored
in Chapter \ref{cha:Dimensionality-reduction}.

The rest of this chapter is organized as follows: in Section \ref{sec:Building-Blocks},
we describe two of the building blocks of the SNF algorithm. Namely,
a wavelet-based image-enhancement technique and a silhouette-driven
K-Means clustering technique. These techniques are used by the SNF
algorithm which is described in Section \ref{sec:The-Algorithm-(Sub-Nuclei}.
Experimental results are given in Section \ref{sec:Experimental-Results6}.
We conclude in Section \ref{sec:Conclusion-and-Future6}.

\subsection{\label{sub:The-Problem-of}The problem of sub-nuclei segmentation}

The ability to map a neuronal tissue to its sub-parts, including its
sub-nuclei arrangement, is needed for certain brain surgeries, forebrain
functionality mapping and for pathologically tracking and researching
brain related diseases. 

In the field of functional brain studies (e.g. fMRI), changes in the
functional-activation have been detected relating certain diseases
\cite{Blinkenberg00,Heckers00,Rauch01,Rubia,Volz99,Vuilleumier01}.
An accurate mapping of the brain is necessary for accurately locating
the specific areas of these changes.

\begin{figure}[!h]
\begin{centering}
\includegraphics[bb=80bp 210bp 530bp 570bp,clip,width=1\columnwidth]{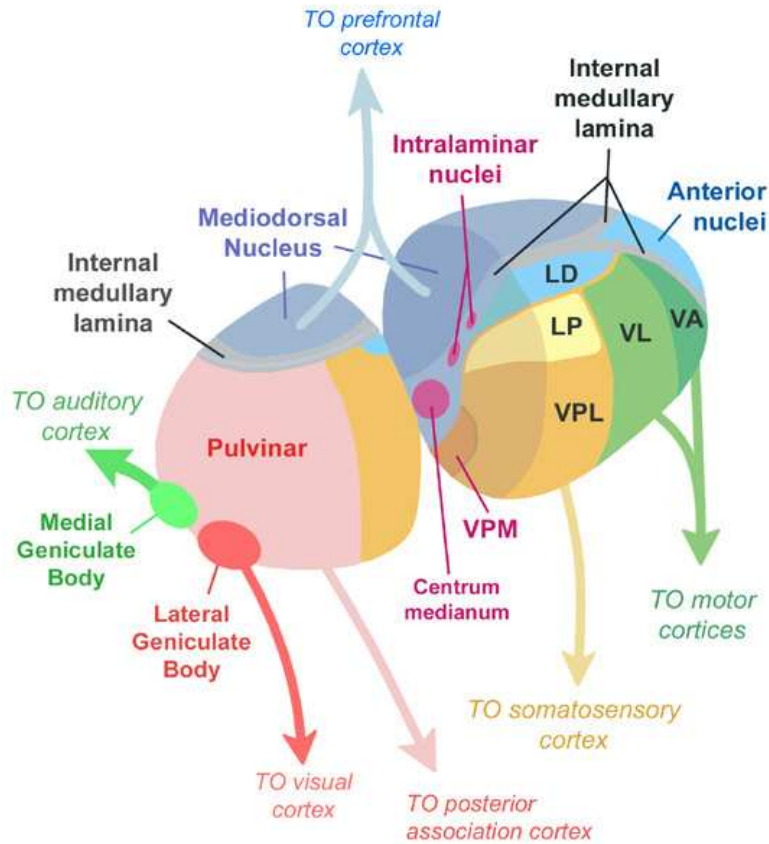}
\par\end{centering}

\caption{\label{cap:The-major-nuclei}The major nuclei in the human thalamus
and their main connections - taken from \cite{A_Brief_Review_of_Neuroanatomy}}
\end{figure}

In order to further comprehend the necessity of a sub-nuclei mapping,
we take the \emph{thalamus} as a case-study. The \emph{thalamus} is
a large ovoid mass of gray matter situated in the posterior part of
the forebrain. It functions as a central relay station conveying sensory
impulses from the spinal cord to the cerebrum. A number of sub-parts
which are known as \emph{nuclei} constitute the thalamus. Each nuclei
has its own function and cyto-architecture (cellular composition).
Most of the histological researches \cite{Morel97,Scannell99,Smeets99,Van_Buren72}
have detected 9-16 major nuclei. Figure \ref{cap:The-major-nuclei}
shows a 3D scheme of the major nuclei in the human thalamus. 

Changes in the thalamus nuclei have been observed in a large number
of diseases, e.g. Parkinson's disease \cite{Parkinsons}, Wallerian
degeneration \cite{wallerian_degeneration}, chronic pain syndrome
\cite{chronic_pain_syndrom}, multiple sclerosis \cite{multiple_sclerosis}
and schizophrenia \cite{schizophrenia}. In some of these diseases
the treatment can include a removal by surgery or an electric stimulation
of the involved nucleus/nuclei \cite{Tornqvist}. 

Moreover, we indicate that using a nuclei map, based on one brain,
for another different brain, is problematic due to the wide variability
that exists in the nuclei sizes and locations in each brain of each
individual \cite{Van_Buren72}.

Therefore, there is a crucial need for a non-invasive \emph{in-vivo}
method for mapping a neuronal tissue into its sub-nuclei. The \emph{in-vivo}
demand causes the histological methods, such as those in \cite{Duvernoy,Kikinis,Morel97,Van_Buren72},
to be irrelevant for the aforementioned tasks.

\subsection{\label{sub:Sub-Nuclei-Segmentation-Using}Sub-nuclei segmentation
using MRI}

Standard usage of existing acquisition methods such as Computer Tomography
(CT) and Magnetic Resonance Imaging (MRI) does not provide adequate
contrast for distinguishing between nuclei. Recently, the medial-dorsal
nucleus (MD) of the thalamus has been detected by a standard usage
of MRI \cite{Magnotta00}, but without the ability to detect the rest
of the thalamus nuclei. Hence, it seems that a promising breakthrough
would be a non-standard usage of an existing acquisition method.

MRI has a potential to be a basic sensing method for this task. It
satisfies the non-invasive and \emph{in-vivo} demands and it outperform
other sensors in the aspect of resolution and separability capability.
Another major advantage of MRI is its multi-contrast methodology (see
details below). 

The attempts made thus far using standard MRI \cite{Ashburner97,Atkins98,Lemieux2003,Stokking2000,Wu05}
have obtained a gross segmentation of the brain into its three main
compartments: white matter, gray matter and cerebrospinal fluid (CSF).
Some of these attempts did a more accurate separation into a small
number of sub-cortical parts. As we mentioned above, only one nucleus
(MD) of the thalamus has been successfully detected in the standard
procedure. In addition, a MRI high-resolution acquisition was done
\emph{in-vitro} delivering an improved separation between the cortical
layers \cite{Bendersky03,Fatterpekar02,Kruggel03,Barbier02}, though
its long acquisition time is not realistic for \emph{in-vivo} procedures. 

Regarding the thalamus, a recent hypothesis argues that its nuclei
are distinguished in their characteristic fiber orientation \cite{Niemann00,Wiegell99,Wiegell2000}.
Indeed, based on this hypothesis, division of the thalamus into several
nuclei was made using Diffusion Tensor Imaging (DTI) \cite{Jonasson05,Wiegell03}.
DTI is a novel MRI-based technique, which enables the visualization
of the location, orientation and anisotropy of the white matter tracts
of the brain.

As aforementioned, one of the major advantages of MRI is that it is
a multi-contrast method. Namely, controlling several parameters of
the acquisition process (such as \emph{time to repeat}, \emph{time
to echo,} etc.), one can affect the produced images contrast. The
contrast variability, in accordance with the acquisition parameters,
is usually distinctive for each type of neuronal tissue. This is the
main motivation for the usage of different contrast acquisition-methods
for neuronal tissues separation. In \cite{Zavaljevski00}, five contrast
acquisition-methods were used, which enabled a segmentation of a brain
into its white matter, gray matter, CSF and few sub-cortical structures.
However, they did not deliver a segmentation of the nuclei level.
Yovel and Assaf \cite{Yovel06} used ten contrast acquisition-methods
and managed to segment a human thalamus into seven histologically-distinct
areas. 

In this chapter we achieve the desired sub-nuclei segmentation by
using a novel algorithm (Section \ref{sec:The-Algorithm-(Sub-Nuclei}).
The inputs to the algorithm are ten different MRI-based contrast mechanisms
(Section \ref{sub:The-Input-Data}) and the output of the algorithm
is a division of the examined tissue into sub-nuclei.

\subsection{\label{sub:Image-Enhancement}Image enhancement}

Generally speaking, the goal of image enhancement is to process an
image such that the achieved result is more suitable than the original
image for a defined task. Specifically, it can be said that the goal
of image enhancement is to supply techniques that can enhance the
obscured details of the features, that are most relevant to a specific
application, and simultaneously reduce the noise. For instance, in
our work we aim to enhance features, that separate between neuronal
tissue nuclei, and also to reduce the noise caused during the acquisition
and the quantization processes. Due to the ambiguity in the definition
of features and noise, the field of image enhancement is objective
and it is application-depended. Hence, there is no general standard
of image enhancement quality that can serve as a design criterion
for these techniques.

Many solutions have been suggested for image enhancement and we will
briefly survey several relevant ones. Histogram modification techniques
\cite{17-Frei77,16-Hummel75} are attractive due to their simplicity
and relative low complexity. These techniques are performed by fixing
a transformation (usually a non-linear one) that is applied on the
image histogram. This transformation is applied to every pixel in
the image and it maps each gray level into a new gray level. However,
these techniques suffer from the fact that the local relations between
the pixels are not used. Therefore, adaptive histogram equalization
methods were developed in order to overcome this restriction. Unfortunately,
these methods are usually based on a constant-sized window which is
not suitable for emphasizing features of various sizes.

Gorden and Rangayyan \cite{5-Gorden84} presented a method, that is
based on adaptive neighborhoods. It emphasizes the interesting features,
but at the same time amplified the noise causing digitization artifacts.
Another adaptive neighborhoods method was presented in \cite{6-Dhawan86,7-Dhawan87,8-Dhawan88},
however, it combines knowledge regarding the specific feature that
needs to be enhanced. This knowledge dependence is not applicable
in our case since the data we work with contains different types of
features for which this preliminary knowledge is unavailable. A non-linear
stretching of a linear combination of the original image and smoothed
versions of it was proposed in \cite{9-Tahoces91}. This technique
is able to emphasize edges while providing a minimal noise amplification.

Unsharp masking methods \cite{Gonzalez02,19-Rosenfeld82} sharpen
the edges by subtracting a smoothed image (e.g. using a Laplacian
filter) from the original image. However, these methods are less effective
when they are applied to images that contain features of various sizes
(such as in our case where the nuclei sizes vary), due to their linearity
and their single scale analysis. An attempt to overcome this limitation
is presented in \cite{22-Beghdadi89,5-Gorden84} where a local contrast
and non-linear transformations are used. 

A promising category for image enhancement uses \emph{wavelet analysis}.
The wavelet analysis (transform) consists of a decomposition of a
signal (2D signal in case of an image) using a family of functions
referred to as the \emph{wavelet family.} See \cite{Daub92,Mallat98}
for a fundamental discussion on wavelets. Contrary to the Fourier
transform, which reveals only the frequency attributes of an image,
the wavelet transform delivers a powerful insight into both spatial
and frequency characteristics of the image and also represents the
image in various scales of resolution, referred to as \emph{multi-resolution
analysis}. The variation of resolution enables the transform coefficients
to locally characterize the irregularities of the signal (image).
Hence, wavelet analysis can provide a basis for sophisticated image
enhancement tasks (such as in our case) that outperform other existing
techniques.

In our work, we use a wavelet-based image enhancement method, originally
implemented for mammograms in \cite{Laine95}. Section \ref{sec:Wavelet-for-Contrast}
describes this method in more details.

\subsection{\label{sub:Cluster-Analysis}Cluster analysis}

The final step of the SNF algorithm, is a clustering procedure. This
step divides the processed data, which is received from the previous
steps of the SNF algorithm, into groups. Each group represents a nucleus
of the examined neuronal tissue. Next, we present a brief survey of
cluster analysis.

\emph{Cluster analysis}, also known as clustering, is a sub-domain
of \emph{unsupervised learning}. In the field of cluster analysis,
a set of objects is divided to an unknown number of clusters. Ideally,
the number of clusters should be discovered during the procedure.
However, some clustering techniques, e.g. K-Means (see below), require
the number of sought after clusters as input. The main guideline for
the clustering process is that objects of the same class should be
``similar'' while objects that belong to different classes should
be ``dissimilar''.

Data representation affects the results of the clustering procedure.
In some cases, balancing this phenomenon can be achieved by a careful
selection of a distance metric. For instance, in our case, we use
the diffusion metric (Section \ref{sub:Spectral-decomposition}).
Thus, although commonly successful, clustering analysis is an experimental
procedure in its nature.

There exists a large number of clustering techniques. Most of them
can be classified into two main categories: \emph{hierarchical clustering}
and \emph{partitioning clustering.}

In hierarchical clustering techniques, the clusters created during
the process are repeatedly combined into larger clusters, resulting
in a hierarchical structure in a shape of a tree. For further details
regarding this category, the reader is referred to \cite{Kotsiantis04}.

Partitioning clustering techniques divide the set of objects into
homogeneous clusters, which afterwords are examined separately. In
these techniques, an object dissimilarity metric is chosen. Moreover,
the number of the requested clusters and their initialization must
be chosen. Hardy \cite{Hardy96} introduces a comparison between different
methods for choosing the number of clusters.

In this chapter, we use a technique (Section \ref{sec:Silhouette-Driven-K-Means}),
which is based of the K-Means clustering algorithm, for the clustering
phase of the SNF algorithm. In the following, we briefly describe
the \emph{K-Means} \cite{kmeans_org} technique, which belongs to
the partitioning clustering category. In K-Means, the \emph{k} signifies
the number of clusters that is set a-priori, while the \emph{means}
signifies the way that clusters centroids are calculated. For simplicity,
we refer to a data object as a ``point''. The K-Means technique
uses a two-step iterative algorithm, which minimizes the sum of point-to-centroid
squared distances, summed over all \emph{k} clusters. Initially, the
points are assigned at random to the \emph{k} clusters. In the first
step, the centroid of every cluster is calculated by taking the mean
across all points in the cluster. In the second step, every point
is reassigned to the cluster whose centroid is the closest one to
it. These two steps are repeated until there is no further change
in the assignment of the data points (the stopping criterion). An
undesirable situation is when this process converges to a local optimum,
i.e. a partition of points in which is not the optimal (global minimum).
This problem can be solved by using a different initialization. 

Most of the clustering algorithms are not intended to find the optimal
separation, but rather find a separation that gives a local optimum
according to a certain criterion used by the algorithm. Nevertheless,
there are techniques that find the global optimum. Several of them
are described in \cite{Hosel01}.

The robustness of a clustering algorithm is a critical issue. The
term ``robustness'' used here refers to the following question:
if some input data points deviate slightly from their current values,
will we get the same clustering result? This deviation is referred
to as a \emph{perturbation} of the original data. A good clustering
technique should be insensitive to noise and perturbations of the
data and capture the ``real'' structure of the data. In our case,
the robustness to noise and the emphasis of the ``interesting''
features of the data are achieved using an image enhancement phase
(Section \ref{sec:Wavelet-for-Contrast}) prior to the clustering
phase. An extensive source for in-depth discussion on robustness and
robust statistics can be found in \cite{Huber81}.

For more detailed reviews on cluster analysis see \cite{Lior05,Han01,Hosel01,Kogan05,Kotsiantis04,Finland06,Cluster_Analysis01}.

\section{\label{sec:Building-Blocks} Building blocks of the SNF algorithm}

In this section, we describe the methods that are used in the SNF
algorithm (Section \ref{sec:The-Algorithm-(Sub-Nuclei}) for image
enhancement and clustering. First, we describe the method we use for
image enhancement and we continue with a description of the method
we use for clustering. The DB and DM methods that are used for the
dimensionality reduction were described in Chapters \ref{cha:Diffusion-Maps}
and \ref{cha:Diffusion-bases}.

\subsection{\label{sec:Wavelet-for-Contrast}Wavelet-based image enhancement}

This section contains a description of the wavelet-based image enhancement
technique which is used in the SNF algorithm. This technique, was
originally developed for mammography enhancement (\cite{Laine95})
and it consists of the following steps:
\begin{enumerate}
\item Wavelet decomposition
\item Denoising
\item Non-linear mapping (amplification) of wavelet coefficients
\item Wavelet reconstruction
\end{enumerate}
In the following we describe in detail the steps.

\paragraph{\label{sub:Phase-1--WaveletDecomposition}Step 1: Wavelet decomposition}

Application of a wavelet transform to the original image which decomposes
the image into wavelet coefficients. We use $C_{l}\triangleq\left\{ c_{l,i}\right\} _{i=1}^{M_{l}}$
to denote the set of wavelet coefficients in level (scale) $l$, where
$M_{l}$ is the number of wavelet coefficients in level $l$.

\paragraph{\label{sub:Pase-2--Denoising}Step 2: Denoising}

An image enhancement technique must take into account the noise in
the image, which is caused by the acquisition and the quantization
processes. Ignoring the existence of noise might result in its amplification.
The ability to separate between the features and the noise (this is
known as \emph{denoising}) is a challenge since there is no clear
distinction between them. Thus, denoising techniques focus on suppressing
noise in the signal (the image) while simultaneously minimizing undesired
impacts on features of interest.

In the wavelet framework, simple denoising is achieved by discarding
small wavelet coefficients. We use the \emph{non-linear wavelet shrinkage}
method, which was introduced in \cite{Donoho93}. In this method,
a threshold $N_{l}$ is given for each resolution level. The coefficients
in level $l$ that are smaller than $N_{l}$ are zeroed and the rest
of the coefficients are \emph{shrunk}. Formally, for each level $l$:
\begin{equation}
\Psi_{l}(x)\triangleq sign(x)\cdot\left\{ \begin{array}{cc}
|x|-N_{l} & \textrm{if }|x|>N_{l}\\
0 & \textrm{otherwise}
\end{array}\right.\label{eq:Denoising}
\end{equation}
for all $x\in C_{l}$ (step 1), where $N_{l}$ denotes the noise threshold
for level $l$, and its derivation is described in \cite{Donoho95}.

\paragraph{\label{sub:Phase-3--mapping}Step 3: Non-linear mapping of wavelet
coefficients}

This phase performs the actual enhancement. It is preformed automatically
using a non-linear mapping function on the modified coefficients from
step 2. Designing the aforementioned mapping function is based on
the following guidelines:
\begin{itemize}
\item A low-contrast area should be enhanced more than a high-contrast area,
i.e. in the wavelet coefficients domain, lower values in $\left\{ \left.\Psi_{l}(x)\right|x\in C_{l}\right\} $
(Eq. \ref{eq:Denoising}) should be amplified more than higher values
of $\left\{ \left.\Psi_{l}(x)\right|x\in C_{l}\right\} $.
\item A sharp edge must not get blurred. This implies that the coefficients
should be divided into sub-ranges, where each sub-range should be
handled differently (a piecewise function). 
\item Avoiding the creation of new artifacts in the form of changing the
locations of existing local extrema and/or creating new local extrema.
This restricts our amplification function to be monotone.
\end{itemize}
Therefore, based on \cite{Laine95}, we use the following non-linear
(piecewise linear) mapping function, for a specific scale $l$:
\begin{equation}
\Phi_{l}(y)\triangleq\left\{ \begin{array}{cc}
y-(G_{l}-1)\cdot T_{l} & \textrm{if }y<-T_{l}\\
G_{l}\cdot y & \textrm{if }|y|\leq T_{l}\\
y+(G_{l}-1)\cdot T_{l} & \textrm{if }y>T_{l}
\end{array}\right.\label{eq:Amplification}
\end{equation}
for all $y\in\left\{ \left.\Psi_{l}(x)\right|x\in C_{l}\right\} $,
where $G_{l}>1$ is a \emph{gain factor} and $T_{l}$ is the mapping
threshold for level $l$.

The gain factor $G_{l}$ and the threshold $T_{l}$ in Eq. \ref{eq:Amplification},
are chosen for each resolution level $l$. One way for defining the
threshold $T_{l}$ is by $T_{l}\triangleq t\cdot\max\left\{ \left.\Psi_{l}(x)\right|x\in C_{l}\right\} $,
where $0<t<1$ is application dependent. Using a small $t$ leads
to the enhancement of weak features in different scales. $G_{l}$
and $t$ are chosen empirically.

One can limit the introduction of artifacts, which may be due to the
non-linearity nature of the mapping, by carefully choosing the wavelet
filters and designing the mapping function according to the above
guidelines.

\paragraph{Step 4: Wavelet reconstruction}

The last phase of the enhancement procedure is the wavelet reconstruction.
The output image is reconstructed from the new enhanced wavelet coefficients,
after the application of denoising (step 2) and the non-linear mapping
(step 3).

\begin{figure}[!h]
\begin{centering}
\includegraphics[bb=80bp 250bp 530bp 554bp,clip,width=0.9\columnwidth]{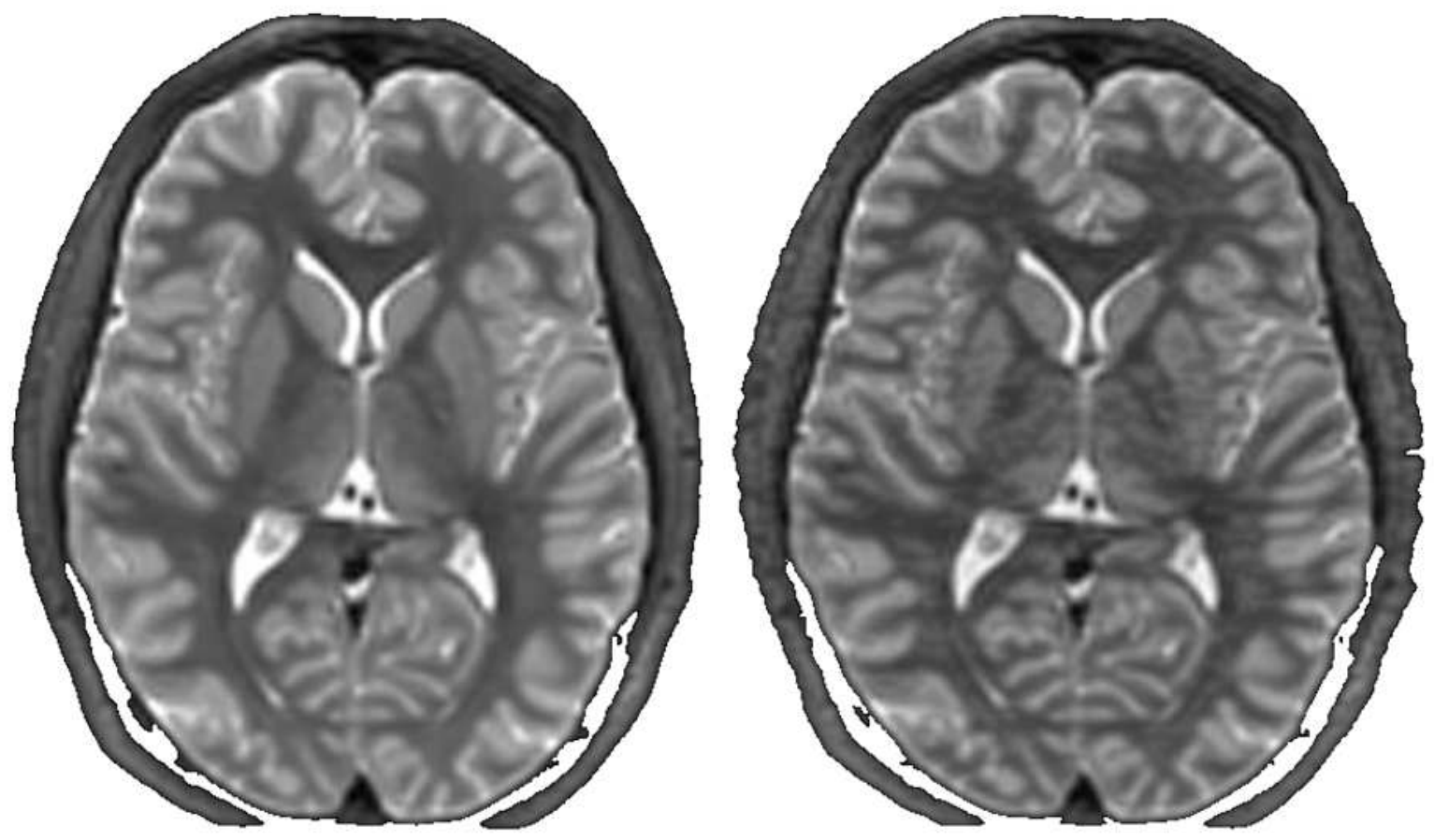}
\par\end{centering}

\begin{centering}
\includegraphics[bb=80bp 280bp 530bp 554bp,clip,width=0.9\columnwidth]{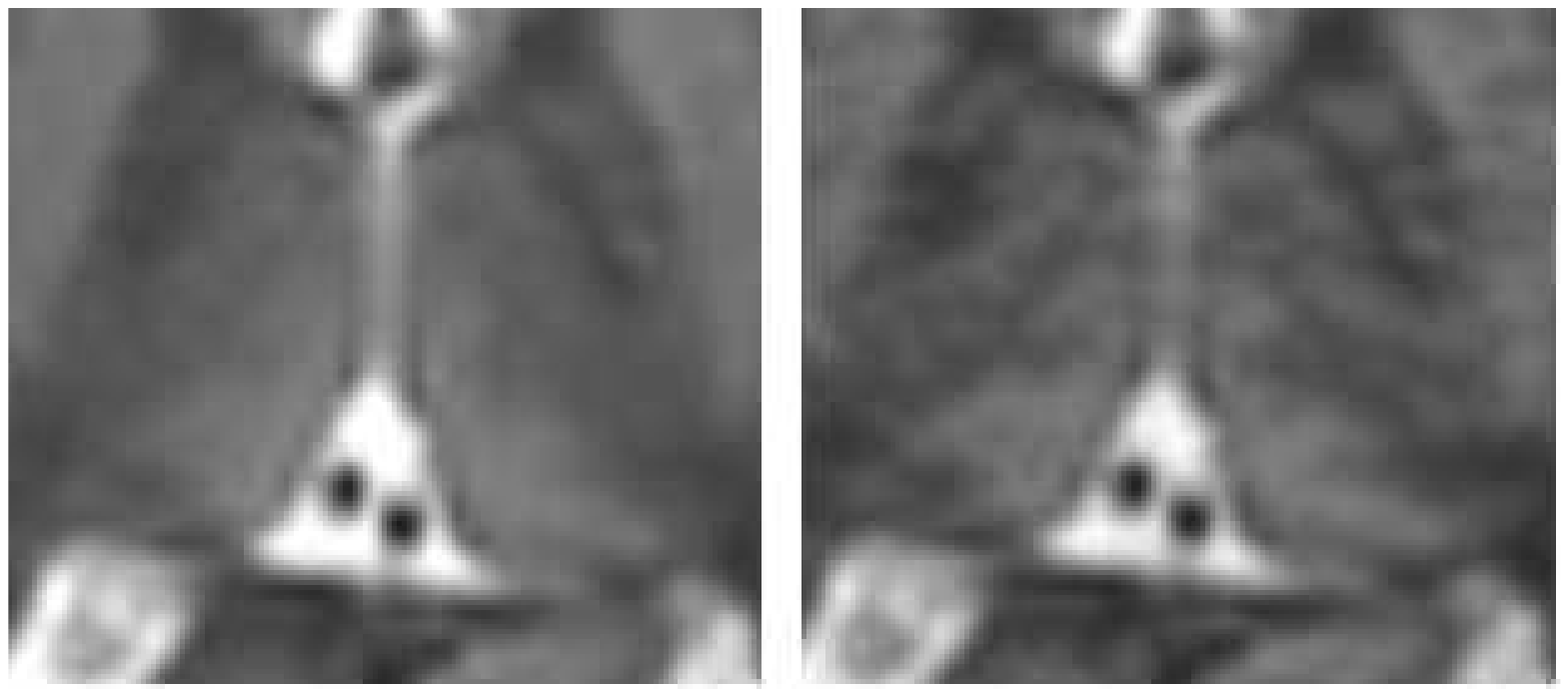}
\par\end{centering}

\caption{\label{cap:slice-28-of}Slice 28 of the STIR contrast acquisition-method
(Table \ref{cap:The-parameters-used}). Left column: before enhancement.
Right column: after enhancement with the parameters $G_{l}=80$, $l=1,...,7$,
$t=0.001$. Top row: the complete slice. Bottom row: magnification
of the thalamus region. Features that were hidden in the original
image appear in the enhanced image while features that appear in the
original image are not damaged in the enhanced image.}
\end{figure}

The presented image enhancement technique outperforms traditional
techniques (Section \ref{sub:Image-Enhancement}) due to its simultaneous
enhancement of features of various sizes. This stems from its multi-scale
feature detection ability and its non-linear enhancement of small
features. Furthermore, edges of large features are not blurred. Figure
\ref{cap:slice-28-of} demonstrates the enhancement of a single slice
of an MRI image acquired with one of the contrast acquisition-methods
that are described in Section \ref{sub:The-Input-Data}. In this figure,
features that were hidden in the original image appear in the enhanced
image while features that appear in the original image are not damaged
in the enhanced image.

\subsection{\label{sec:Silhouette-Driven-K-Means}Silhouette-driven \emph{k}-means
(SDK)}

This section contains a description of the clustering technique we
use in the SNF algorithm. Our clustering technique is based on the
K-Means algorithm (Section \ref{sub:Cluster-Analysis}) and it includes
an extension that automatically finds the number of clusters $k$.

\begin{figure}[!h]
\begin{centering}
\begin{tabular}{cc}
\includegraphics[bb=80bp 210bp 530bp 600bp,clip,width=0.48\columnwidth]{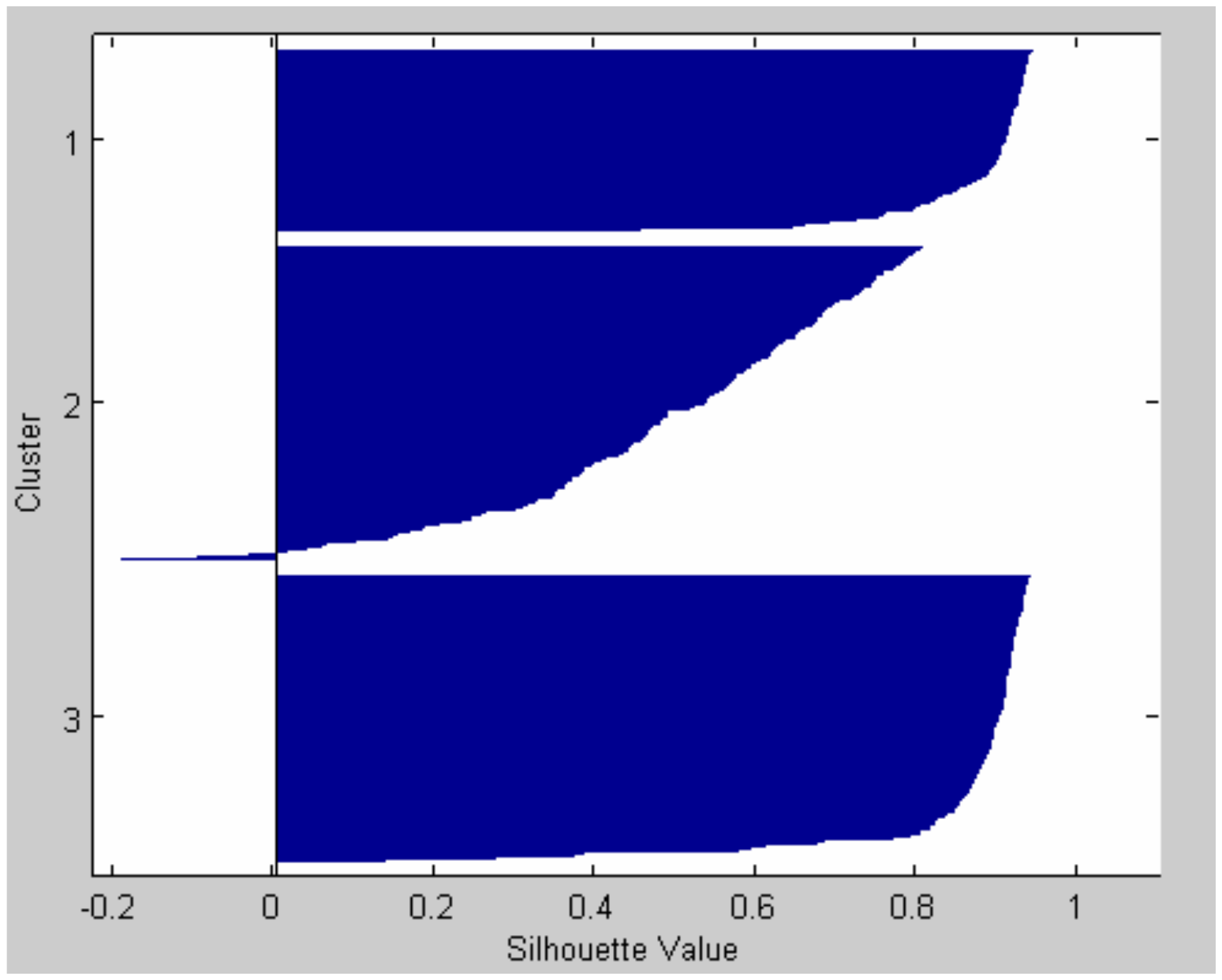} & \includegraphics[bb=80bp 210bp 530bp 600bp,clip,width=0.48\columnwidth]{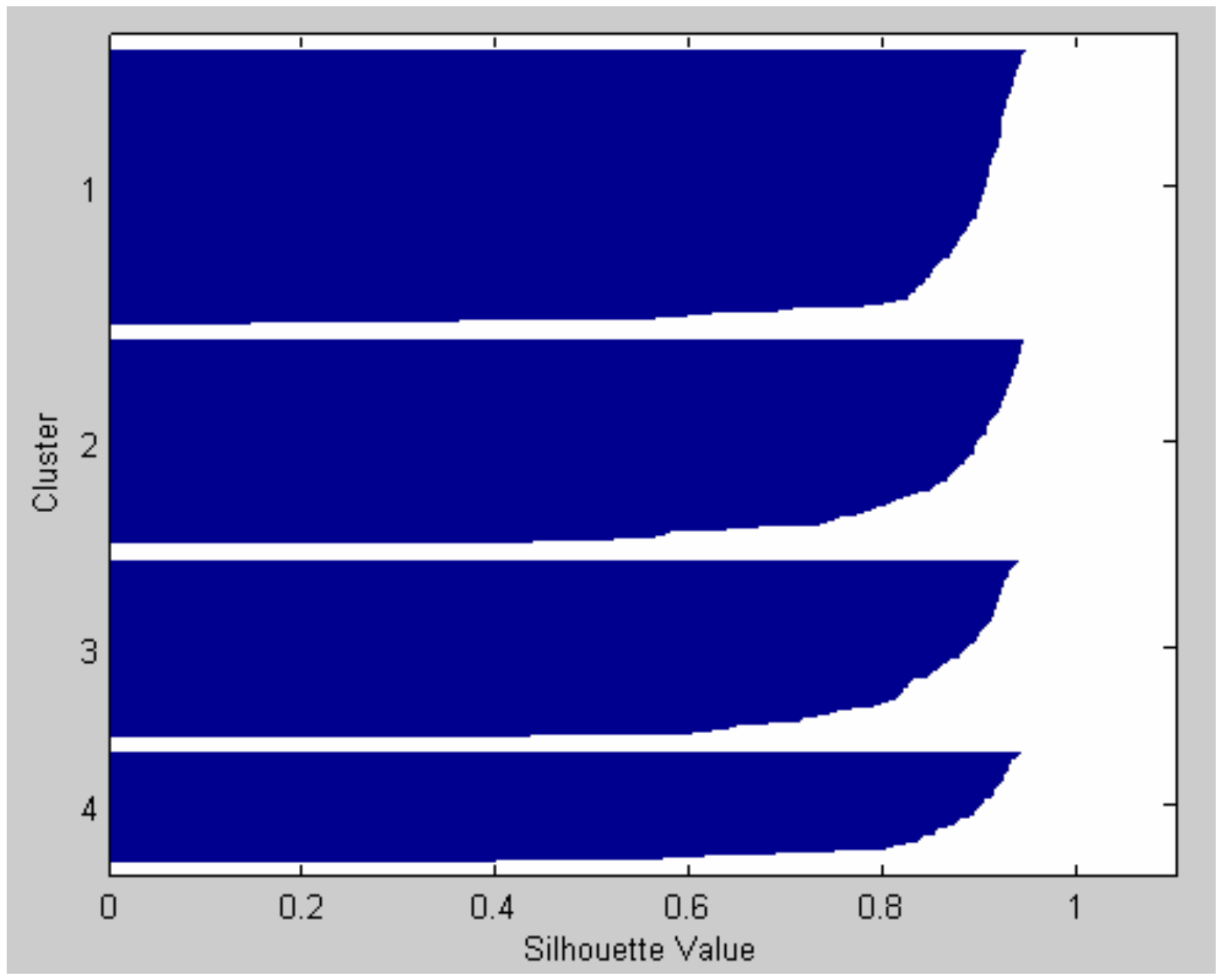}\tabularnewline
\end{tabular}
\par\end{centering}

\caption{\label{cap:Cluster-silhouttes.}Using silhouette values to compare
between clustering results. The values of the \emph{y}-axis are the
clusters' indices and the values of the \emph{x}-axis are the silhouette
values of the clusters elements. The silhouette values of each cluster
are in ascending order from bottom to top. }

Left: a silhouette plot of the result from clustering a data into
$k=3$ clusters. The first and the third clusters have large silhouette
values (most of them are grater than 0.8), indicating these clusters
are well separated from neighboring clusters. The second cluster has
relatively low silhouette values (some of them are negative), indicating
this cluster is not well separated. The mean across the silhouette
values of this clustering result is 0.72.

Right: a silhouette plot of the result from clustering the same data
into $k=4$ clusters. All four clusters have large silhouette values
(most of them are grater than 0.8), indicating these clusters are
well separated. The mean across the silhouette values of this clustering
result is 0.86.

The two silhouette plots (left and right) and the mean silhouette
values indicate that the clustering result for $k=4$ (right) is better
than the one for $k=3$ (left).
\end{figure}

We use the clustering criterion that is known as \emph{silhouette}
\cite{Silhouettes87,Kaufman90}. Let $\Upsilon$ be a given clustering
result. A silhouette in $\Upsilon$ measures how similar a point is
to points in its own cluster compared to points in other clusters.
Silhouette values range from $-1$ to $+1$ and are computed, for
a given clustering result $\Upsilon$, as follows: for each data point
$u$, we denote by $Cluster_{\Upsilon}(u)$ the set of points in the
cluster in $\Upsilon$ that $u$ belongs to and by $\left|Cluster_{\Upsilon}(u)\right|$
the number of points in the cluster. We define 
\begin{equation}
a_{\Upsilon}(u)\triangleq\frac{\sum_{v\in Cluster_{\Upsilon}(u),\, v\neq u}\left\Vert u-v\right\Vert }{\left|Cluster_{\Upsilon}(u)\right|-1}\label{eq:Au}
\end{equation}
 to be the average distance from point $u$ to all other points in
its cluster in $\Upsilon$. Let 
\begin{equation}
b_{\Upsilon}(u)\triangleq\min_{C\neq Cluster_{\Upsilon}(u)}\left\{ \frac{\sum_{v\in C}\left\Vert u-v\right\Vert }{\left|C\right|-1}\right\} \label{eq:Bu}
\end{equation}
 be the minimal average distance from a point $u$ to all points in
other clusters in $\Upsilon$. Using Eqs. \ref{eq:Au} and \ref{eq:Bu},
the silhouette value of data point $u$ is defined as
\begin{equation}
S_{\Upsilon}(u)\triangleq\frac{b_{\Upsilon}(u)-a_{\Upsilon}(u)}{\max\left\{ b_{\Upsilon}(u),\, a_{\Upsilon}(u)\right\} }\,.\label{eq:Su}
\end{equation}
The silhouette values (derived in Eq. \ref{eq:Su}) of all of the
data points can be used as a criterion for assessing the clustering
result $\Upsilon$. Figure \ref{cap:Cluster-silhouttes.} shows a
comparison between two plots of the silhouette values of two clustering
results of the same data for two different values of $k$. These plots
indicate which clustering result is better.

In the following, we describe our Silhouette-Driven K-means (SDK)
algorithm. In the SDK algorithm, we search for a $k$ as large as
possible, which produces a clustering with satisfying silhouette values
compared to other $k$'s. The inputs to the algorithm are $D$, $\gamma$
and $\tilde{k}$, where $D$ is the data points, $0<\gamma\leq1$
is a threshold and $\tilde{k}$ is a maximal expected number of clusters
$k$ that depends on the examined data and is usually known a-priori.
For instance, for the thalamus, we use a maximum value of $\tilde{k}=16$,
which is based on known results regarding the number of its sub-nuclei
(Section \ref{sub:Sub-Nuclei-Segmentation-Using}). For each $k=2,...,\tilde{k}$,
we run K-Means($D$, $k$) on the data and compute the mean across
the silhouette values of the clustering result. Next, we find $\hat{k}$
that produces the maximal mean silhouette value. Then, we search for
the largest $k\in[\hat{k},\tilde{k}]$ that gives a mean silhouette
value, that when multiplied by $\gamma$, is larger than the mean
silhouette value for $\hat{k}$. Algorithm \ref{cap:Algorithm-The-silhouette-driven-k-means}
formulates the SDK algorithm. 

\begin{algorithm}[!h]
\textbf{SDK($D$, $\gamma$, $\tilde{k}$)}
\begin{enumerate}
\item For $k=2,...,\tilde{k}$ do:\\
~~calculate $\Upsilon_{k}=$K-Means($D$, $k$) (Section \ref{sub:Cluster-Analysis}),\\
~~calculate $m_{k}=\textrm{mea}\textrm{n}_{u\in D}\left(S_{\Upsilon_{k}}(u)\right)$
(using Eq. \ref{eq:Su})
\item Find $\hat{k}=\arg\max_{k=2,...,\tilde{k}}\left(m_{k}\right)$
\item For $k=\tilde{k}$ down to $\hat{k}$ do:\\
~~if $m_{k}\geq\gamma\cdot m_{\hat{k}}$ break 
\item Return $\Upsilon_{k}$
\end{enumerate}
\caption{\label{cap:Algorithm-The-silhouette-driven-k-means}The Silhouette-Driven
K-means (SDK) algorithm }
\end{algorithm}

\section{\label{sec:The-Algorithm-(Sub-Nuclei}The \emph{Sub-Nuclei Finder}
(SNF) algorithm}

This section describes our algorithm for sub-nuclei segmentation.
The input to the algorithm is a multi-contrast MRI data. This type
of data is described in Section \ref{sub:Sub-Nuclei-Segmentation-Using}
and the specific experimental data that is being used in our work
is given in Section \ref{sub:The-Input-Data}. In addition to the
multi-contrast MRI values, we also use spatial information. This means
that three additional coordinates $(x,y,z)$ are added to each features
vector, representing the location of the \emph{voxel} (a voxel is
a pixel in a 3D image cube) in the 3D brain space. The spatial information
is added since each sub-nuclei is continuous by definition. It implies
that neighboring voxels are more likely to belong to the same sub-nucleus
than distant ones. The contribution of the spatial coordinates to
the feature vector is regulated by a weight coefficient. The output
of the algorithm is a mapping of each voxel to its sub-nuclei cluster.
In the following we provide a list with the notation that are used
for the description of the SNF algorithm.

\textbf{\uline{Notation}}:

\begin{longtable}{ll>{\raggedright}p{0.7\linewidth}}
$S$ & - & number of slices in the multi-contrast MRI data\tabularnewline
$z$ & - & a slice in the multi-contrast MRI data, $z=1,...,S$ ($z$ can also
be considered as the \emph{z}-axis in the 3D brain space)\tabularnewline
$M$ & - & number of contrast acquisition-methods\tabularnewline
$m$ & - & a contrast acquisition-method, $m=1,...,M$\tabularnewline
$I_{z,m}$ & - & 2D image of slice $z$ acquired using the contrast acquisition-method
$m$\tabularnewline
$I_{m}$ & - & 3D image consists of all the 2D images $I_{z,m}$, $z=1,...,S,$ piled
one over the other\tabularnewline
$\Omega[I]$ & $\triangleq$ & $\left\{ \left.I_{m}\right|m=1,...,M\right\} $ - the input data which
is received from the multi-contrast MRI acquisition\tabularnewline
$E_{z,m}$ & - &  image $I_{z,m}$ after enhancement\tabularnewline
$E_{m}$ & - & 3D image that contains all the 2D images $E_{z,m}$, $z=1,...,S$,
piled one over the other\tabularnewline
$N_{x}$ and $N_{y}$ & - & width and height of a slice, respectively\tabularnewline
$x$ and $y$ & - & $x$ and $y$ coordinates of a pixel $(x,y)$ in the image $I_{z,m}$,
respectively, $x=1,...,N_{x}$, $y=1,...,N_{y}$\tabularnewline
$v(x,y,z,m)$ & $\triangleq$ & $I_{m}(x,y,z)$ - the value of contrast acquisition-method (feature)
$m$ in position $(x,y,z)$\tabularnewline
$\overrightarrow{f_{v,d}(x,y,z)}$ & $\triangleq$ & $\left(v(x,y,z,1),\, v(x,y,z,2),...,\, v(x,y,z,d)\right)$ - a $d$-feature
vector of a 3D brain%
\footnote{The feature space is not a 3D space but a space of $d$ dimensions,
where each point in this $d$ dimensional space represents the features
acquired by the multi-contrast MRI for each point of the 3D space
of the brain. %
} point $(x,y,z)$ defined by $v$\tabularnewline
$\Gamma_{v,d}$ & $\triangleq$ & $\left\{ \overrightarrow{f_{v,d}(x,y,z)}\right\} $ - the set of points
in the feature space of dimension $d$, defined by $v$\tabularnewline
$R$ & - & the set of 3D brain points which belong to the Region of Interest
(ROI)\tabularnewline
$\Gamma_{v,d,R}$ & $\triangleq$ & $\left\{ \left.\overrightarrow{f_{v,d}(x,y,z)}\right|(x,y,z)\in R\right\} $
- the set of points from $\Gamma_{v,d}$, whose related 3D brain points
belong to the ROI $R$\tabularnewline
$\delta$ & - & a desired accuracy of the dimensionality reduction process, $\delta>0$
(see Section \ref{eq:eigen_decomp})\tabularnewline
$\beta$ & - & a factor that controls the influence of the spatial information in
respect to the contrast information, $0\leq\beta\leq1$\tabularnewline
$\gamma$ & - & a threshold for the SDK algorithm (Algorithm \ref{cap:Algorithm-The-silhouette-driven-k-means}),
$0<\gamma\leq1$ \tabularnewline
$\tilde{k}$ & - & maximal expected number of clusters (Section \ref{sec:Silhouette-Driven-K-Means})\tabularnewline
$\Upsilon$ & - & clustering result\tabularnewline
\end{longtable}

\bigskip{}
The SNF algorithm consists of two steps. In the first step, we locate
our \emph{ROI} - region of interest - in the 3D brain space (Section
\ref{sub:ROI-Extraction}). In the second step, we segment the data
that belongs to the ROI into its sub-nuclei in the 3D brain space
(Section \ref{sub:Sub-Nuclei-Segmentation}), using the input data
and the ROI information.

A summary of the SNF algorithm is given in Algorithm \ref{cap:The-Sub-Nuclei-Finder-algorithm}.
\textbf{$\Omega[I]$} is the input data received from the multi-contrast
MRI acquisition\textbf{. $\beta$}, \textbf{$\delta$} and $\gamma$
are parameters that are empirically chosen. \textbf{$\tilde{k}$}
is the maximal expected number of clusters, which is usually known
a-priori (Section \ref{sec:Silhouette-Driven-K-Means}).

\begin{algorithm}[!h]
\textbf{SNF($\Omega[I]$, $\beta$, $\delta$, $\gamma$, $\tilde{k}$)}
\begin{enumerate}
\item Calculate \textbf{$R$ = ExtractROI($\Omega[I]$, $\beta$, $\delta$)}
(Algorithm \ref{cap:Algorithm-ROI-extraction})
\item Calculate \textbf{$\Upsilon$ = SubNucleiSegmentation($\Omega[I]$,
$R$, $\beta$, $\delta$, $\gamma$, $\tilde{k}$)} (Algorithm \ref{cap:Algorithm-Sub-Nuclei-Segmentation})
\item Return \textbf{$\Upsilon$}
\end{enumerate}
\caption{The SNF algorithm \label{cap:The-Sub-Nuclei-Finder-algorithm}}
\end{algorithm}

\subsection{\label{sub:ROI-Extraction}ROI extraction}

This section describes the procedure of locating the ROI in the 3D
brain space. It is a semi-automatic procedure, which involves basic
expert knowledge. 

\begin{figure}[!h]
\begin{centering}
\includegraphics[bb=80bp 210bp 530bp 600bp,clip,width=0.7\columnwidth]{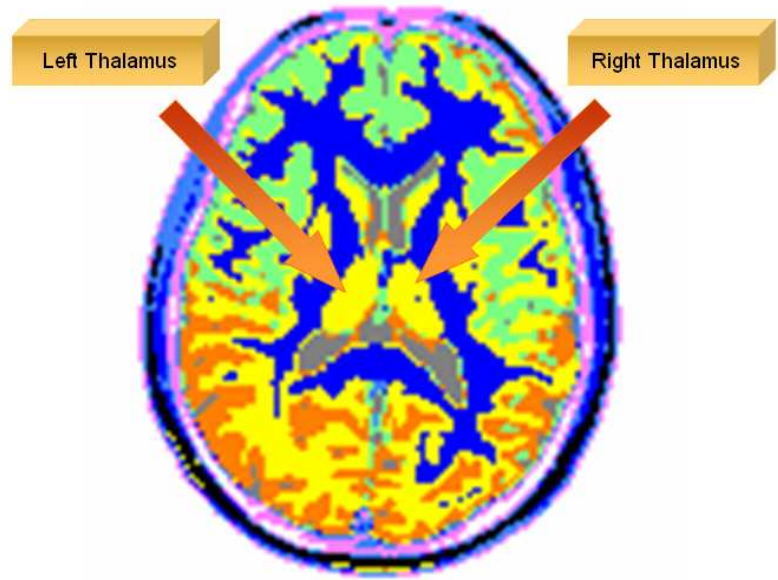}
\par\end{centering}

\caption{\label{cap:Detecting-the-Thalamus}Detecting the thalamus on slice
30 after the application of the \emph{K-Means} algorithm in the ROI
extraction procedure. The presented image is one slice of the output
from the application of step 5 in Algorithm \ref{cap:Algorithm-ROI-extraction}
on the multi-contrast MRI data.}
\end{figure}

First, we apply the DB algorithm (Section \ref{cha:Diffusion-bases})
on the full input data (cube) to reduce its dimensionality. Next,
we add the spatial information to the dimension-reduced data, i.e.
we attach to each voxel its 3D position in the brain. We multiply
the \emph{x}, \emph{y} and \emph{z} coordinates by a scalar to control
their influence and append the result as extra coordinates to the
feature vector of the voxel. Then, we apply the \emph{K-Means} algorithm
(Section \ref{sub:Cluster-Analysis}) to the new set of feature vectors
to produce a coarse segmentation of the brain into $k$ clusters.
The number $k$ is empirically chosen to separate the ROI from the
rest of the data. Large $k$ will cause an undesirable phenomenon
in which the ROI is divided into separate clusters. Figure \ref{cap:Detecting-the-Thalamus}
demonstrates a good separation of the thalamus ROI. Finally, an expert
locates the ROI using the produced segmentation to construct the ROI
mask. The ROI does not have to appear in all of the slices, e.g. in
the data that is used in Section \ref{sec:Experimental-Results6},
the thalamus ROI appears only in slices 22-34.

\begin{figure}[!h]
\begin{centering}
\includegraphics[width=0.8\columnwidth,keepaspectratio]{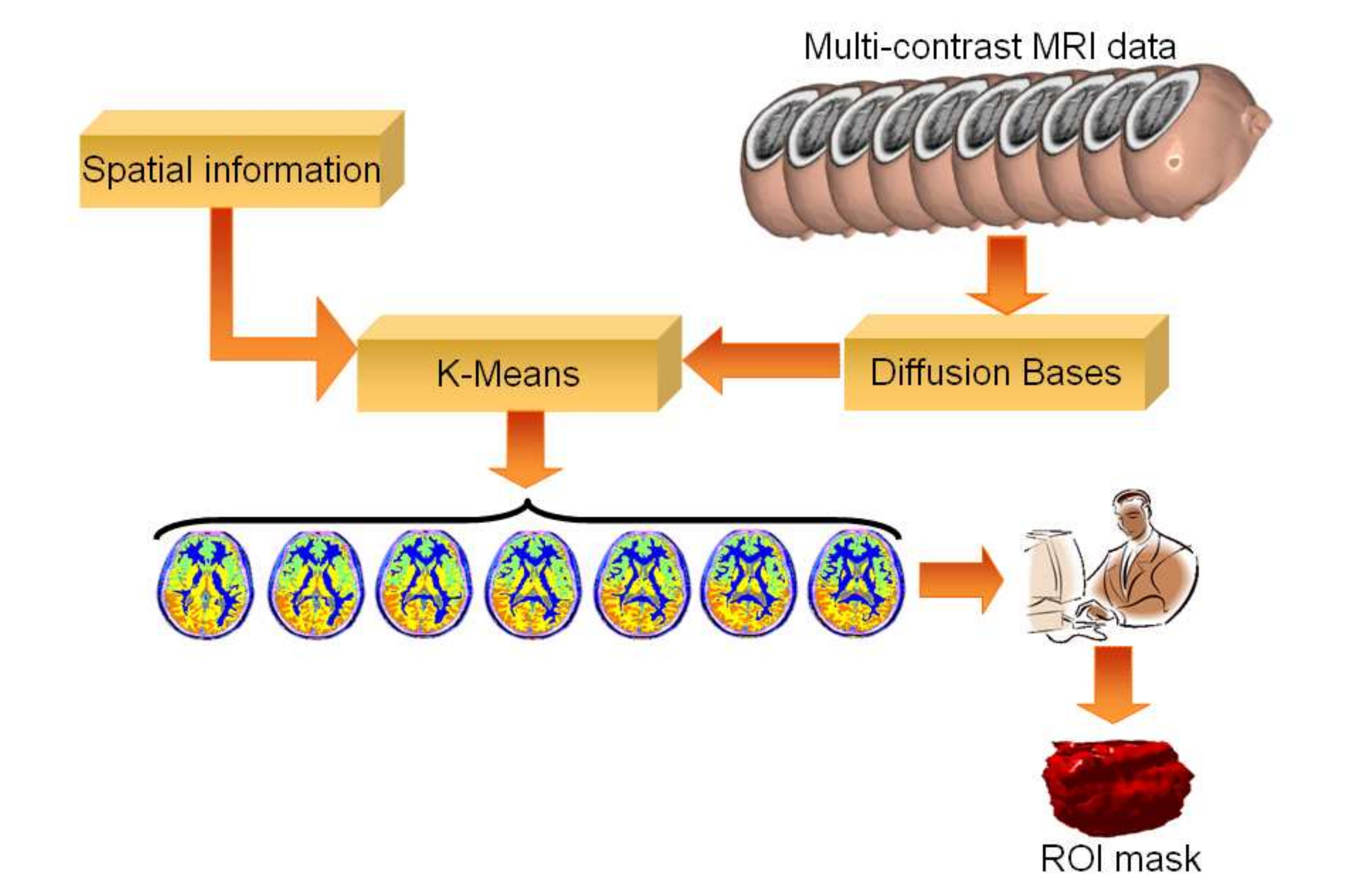}
\par\end{centering}

\caption{\label{cap:Extracting-the-ROI}Functional-flow for extraction of a
ROI mask.}
\end{figure}

The functional-flow of the ROI extraction procedure is illustrated
in Figure \ref{cap:Extracting-the-ROI}. The implementation of the
ROI extraction procedure is described in Algorithm \ref{cap:Algorithm-ROI-extraction}.

\newpage{}
\begin{algorithm}[t]
\textbf{ExtractROI($\Omega[I]$, $\beta$, $\delta$)}
\begin{enumerate}
\item Construct $\Gamma_{v,d}$ from $\Omega[I]$, $d=M$ (see notation
above)
\item Calculate $\varepsilon=\max_{w\in\Gamma_{v,d}}\left\{ \min_{q\in\Gamma_{v,d},\, q\neq w}\left\{ ||w-q||^{2}\right\} \right\} $
\item Calculate $\Gamma_{v',d'}$ ($d'<d$) by applying DB on $\Gamma_{v,d}$
with accuracy \textbf{$\delta$}, using the Gaussian kernel defined
in Eq. \ref{eq:gaussian} with $\varepsilon$ from step 2, where $v'(x,y,z,m)$,
$m=1,...,d'$, is the value of the $m^{\textrm{th}}$ coordinate of
the feature vector related to the 3D brain point $(x,y,z)$ in the
embedding space
\item Add to each vector in $\Gamma_{v',d'}$ the spatial information of
its related 3D point in the brain to produce $\Gamma_{v'',d''}$,
where $d''=d'+3$, $v''(x,y,z,m)\triangleq\beta\cdot v'(x,y,z,m)$,
$m=1,...,d'$, $v''(x,y,z,d'+1)\triangleq(1-\beta)\cdot x$, $v''(x,y,z,d'+2)\triangleq(1-\beta)\cdot y$
and $v''(x,y,z,d'+3)\triangleq(1-\beta)\cdot z$
\item Calculate \emph{$\Upsilon$=}K-Means($\Gamma_{v'',d''}$, $k$) (Section
\ref{sub:Cluster-Analysis}), where the number of clusters $k$ is
empirically chosen
\item For each slice $z=1,...,S$, the ROI is manually marked on the visualization
of $\Upsilon$ and its voxels are added to $R$
\item Return $R$
\end{enumerate}
\caption{ROI extraction procedure \label{cap:Algorithm-ROI-extraction}}
\end{algorithm}

\subsection{\label{sub:Sub-Nuclei-Segmentation}Sub-nuclei segmentation}

This section describes the procedure that segments the data, which
belongs to the ROI, into sub-nuclei, using the set of points in the
feature space and the ROI information. It is a fully automatic procedure
and no human intervention is needed. An optional procedure that can
follow may allow an expert to manually merge clusters. %After running this procedure, the expert can manually unite clusters according to the desired level of sub-nuclei separation.

First, we apply the wavelet-based image enhancement (Section \ref{sec:Wavelet-for-Contrast})
on each slice of each contrast acquisition-method. This procedure
produces denoised and contrast-improved slices. Since our distance
metric is not invariant to scaling, we need to normalize the data.
We normalize each enhanced contrast acquisition-method (feature) separately
based on its range of values \emph{inside the ROI}. The normalization
procedure is described in Algorithm \ref{cap:Algorithm-normalization-process}.
Then, we use the normalized enhanced data, the ROI mask (Section \ref{sub:ROI-Extraction})
and the spatial information as the input to the DM procedure (Section
\ref{cha:Diffusion-Maps}). The ROI mask is used to select the desired
region from the normalized enhanced data. This selected data is combined
with the spatial information to be used as an input to the DM procedure.
The complexity of the DM procedure is reduced since points which are
outside the ROI are excluded. Then, we apply the SDK algorithm (Section
\ref{sec:Silhouette-Driven-K-Means}) on the dimension-reduced output
of the DM algorithm in order to segment the ROI, into its sub-nuclei. 

\begin{figure}[!h]
\begin{centering}
\includegraphics[bb=80bp 210bp 530bp 600bp,clip,width=0.8\columnwidth]{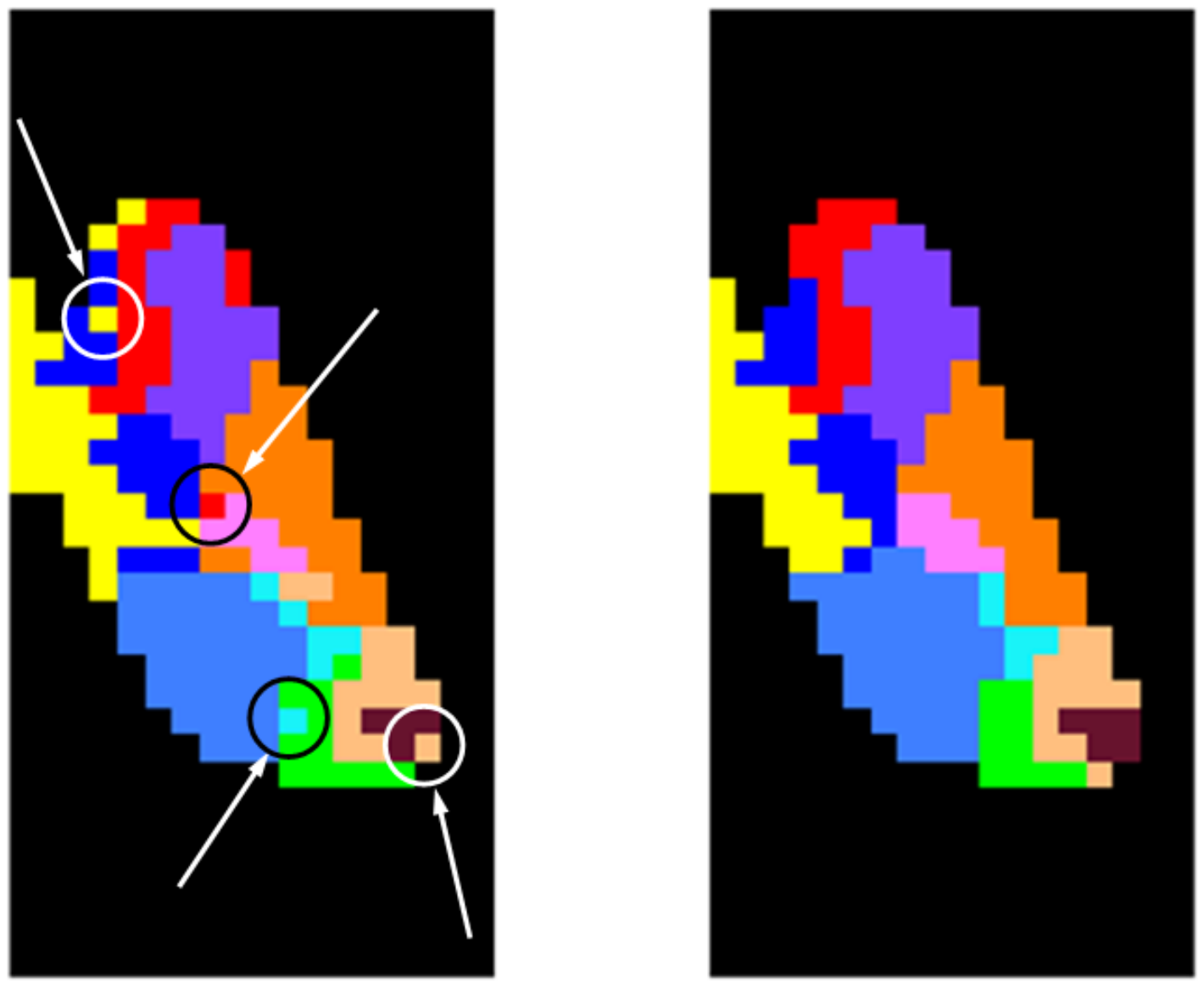}
\par\end{centering}

\begin{centering}
\includegraphics[bb=80bp 210bp 530bp 600bp,clip,width=0.8\columnwidth]{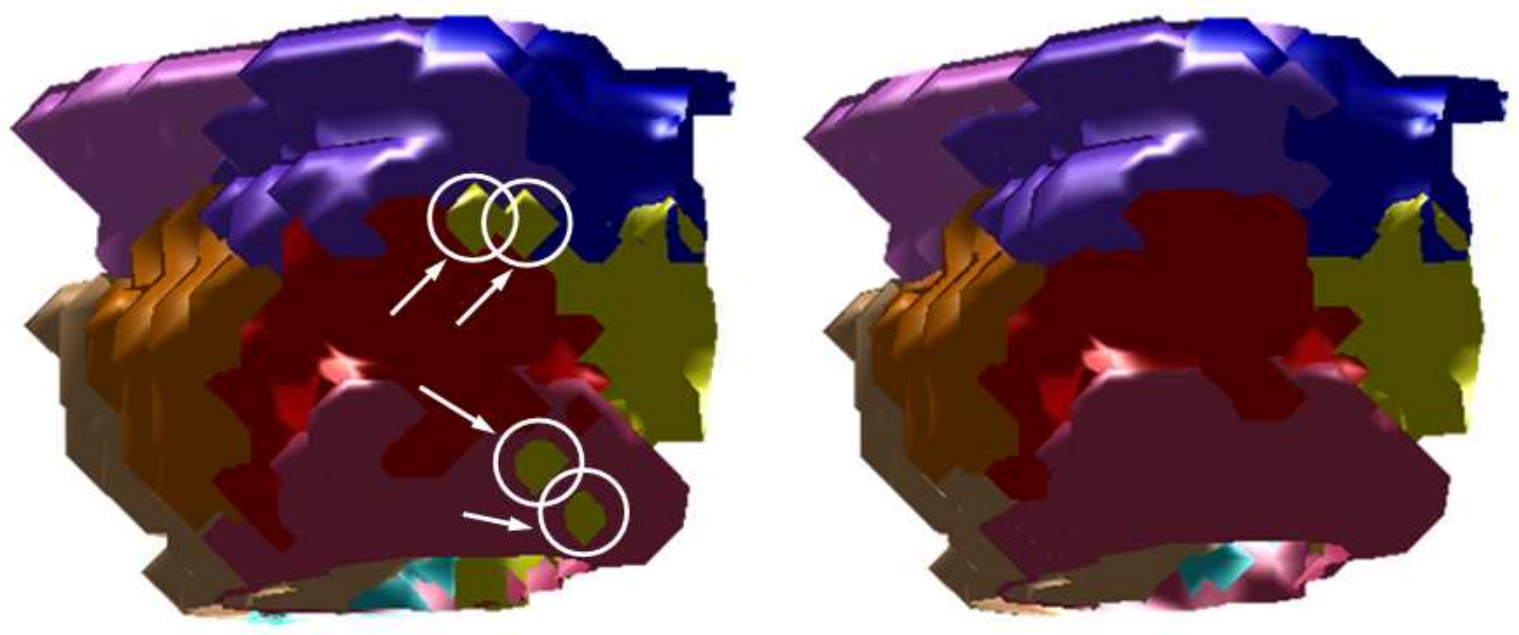}
\par\end{centering}

\caption{\label{cap:Discarding-Isolated-Points.}Merging isolated points. Left
column: Before merging of isolated points. Right column: after merging
of isolated points. Top row: slice 27 of the right thalamus. Bottom
row: all slices (3D space) of the right thalamus\textbf{.} The arrows
and circles mark some of the isolated points.}
\end{figure}

The clustering result contains isolated points, i.e. single-voxel
segments that are located inside larger segments. We merge the isolated
points using a procedure that is described in Algorithm \ref{cap:Algorithm-clean},
in order to get more homogeneous segments. This is based on the assumption
that the nuclei are continuous. Figure \ref{cap:Discarding-Isolated-Points.}
shows a comparison between the segmentation results before and after
the merging of isolated points.

\begin{figure}[!h]
\begin{centering}
\includegraphics[width=0.8\columnwidth]{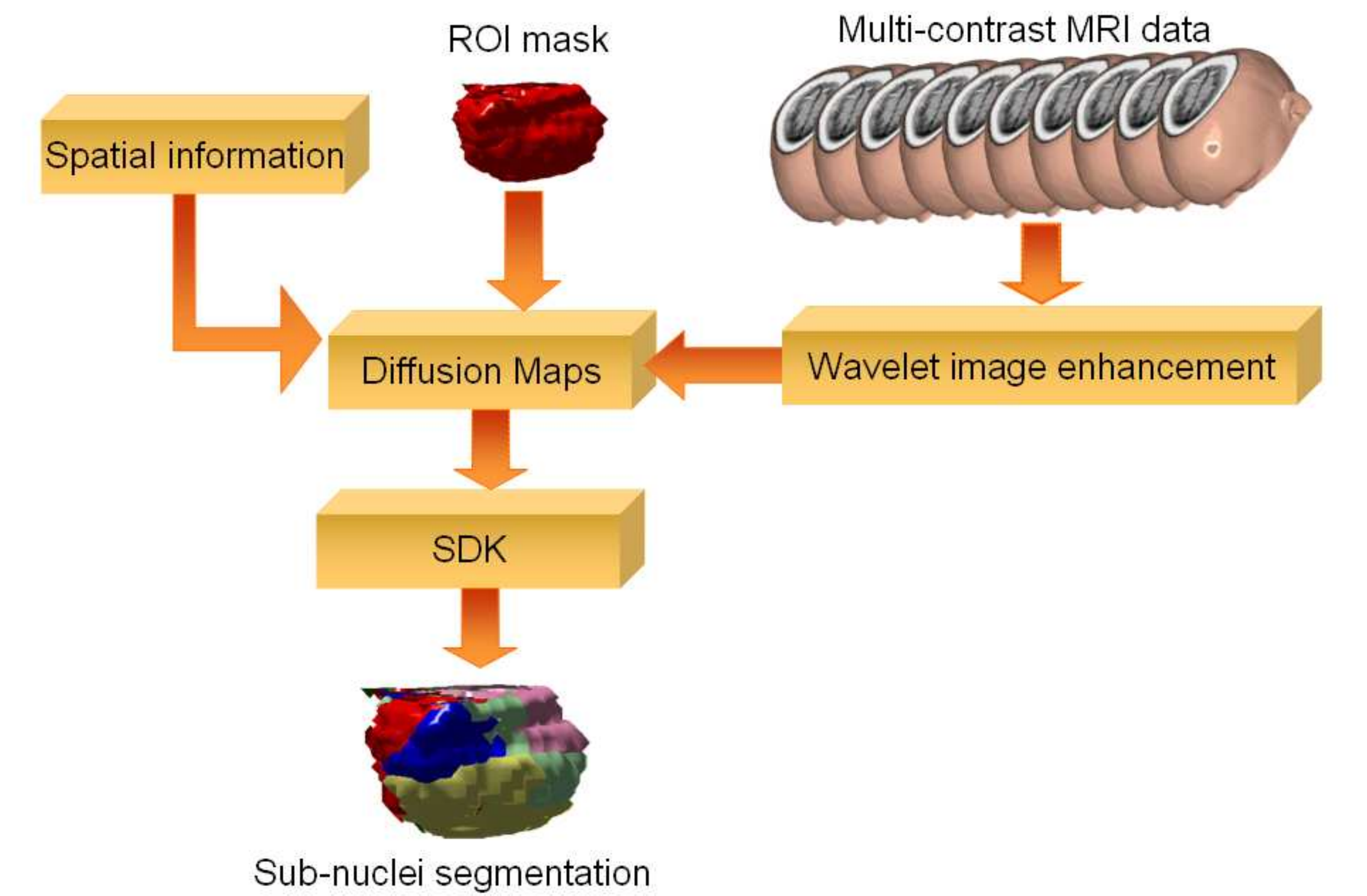}
\par\end{centering}

\caption{\label{cap:Segmenting-to-sub-nuclei}Functional-flow for segmentation
of a given ROI.}
\end{figure}

The functional-flow of the sub-nuclei segmentation procedure is illustrated
in Figure \ref{cap:Segmenting-to-sub-nuclei}. The implementation
of the sub-nuclei segmentation procedure is described in Algorithm
\ref{cap:Algorithm-Sub-Nuclei-Segmentation}.

\begin{algorithm}[t]
\textbf{NormalizeContrastMethod($E_{m}$, $R$)}
\begin{enumerate}
\item Calculate
\[
E_{m}^{normalized}\triangleq\frac{E_{m}-mean_{R}\left(E_{m}\right)}{std_{R}\left(E_{m}\right)}
\]
 where $mean_{R}\left(E_{m}\right)$ and $std_{R}\left(E_{m}\right)$
are the mean and the standard deviation values of $E_{m}$ voxels
that belong to the ROI $R$, respectively
\item Transform $E_{m}^{normalized}$ to be in {[}0,1{]} by applying the
following mapping
\[
E_{m}^{[0,1]}\triangleq\frac{E_{m}^{normalized}-\min_{R}\left(E_{m}^{normalized}\right)}{\max_{R}\left(E_{m}^{normalized}\right)-\min_{R}\left(E_{m}^{normalized}\right)}
\]
 where $\min_{R}\left(E_{m}^{normalized}\right)$ and $\max_{R}\left(E_{m}^{normalized}\right)$
are the minimum and the maximum values of $E_{m}^{normalized}$ voxels
that belong to the ROI $R$, respectively
\item Return $E_{m}^{[0,1]}$
\end{enumerate}
\caption{Normalization procedure \label{cap:Algorithm-normalization-process}}
\end{algorithm}

\begin{algorithm}[!h]
\textbf{MergeIsolatedPoints($\Upsilon$)}

\bigskip{}
For each slice in $\Upsilon$ do:
\begin{enumerate}
\item Search for all the ``isolated points'' that have less than or exactly
one neighbor of the same cluster in its 8-connected neighborhood
\item For every isolated point, search for its closest or containing cluster
by searching in its 8-connected neighborhood
\item For every point: if it is an isolated point, transfer it into its
most closer cluster, otherwise leave it in its current cluster
\end{enumerate}
Return $\Upsilon$\bigskip{}

\caption{Merging isolated points procedure \label{cap:Algorithm-clean}}
\end{algorithm}

\begin{algorithm}[!h]
\textbf{SubNucleiSegmentation($\Omega[I]$, $R$, $\beta$, $\delta$,
$\gamma$, $\tilde{k}$)}
\begin{enumerate}
\item Construct $I_{z,m}$, $z=1,...,S$, $m=1,...,M$, from $I_{m}\in\Omega[I]$
\item Apply the wavelet-based image enhancement algorithm (Section \ref{sec:Wavelet-for-Contrast})
on each $I_{z,m}$ to produce enhanced images $E_{z,m}$, $z=1,...,S$,
$m=1,...,M$ 
\item Calculate $E_{m}^{[0,1]}$ \emph{=} \textbf{NormalizeContrastMethod($E_{m}$,
$R$)}, \\
$m=1,...,M$ (Algorithm \ref{cap:Algorithm-normalization-process})
\item Construct $\Gamma_{w,d}$ from $\Omega[E^{[0,1]}]$, $d=M$ (see notation),
where $w(x,y,z,m)\triangleq E_{m}^{[0,1]}(x,y,z)$, $m=1,...,d$,
is the value of coordinate $m$ in the enhanced space of a 3D brain
point $(x,y,z)$. 
\item Construct $\Gamma_{w,d,R}$ from $\Gamma_{w,d}$ and $R$
\item Add to each vector in $\Gamma_{w,d,R}$ the spatial information of
its related 3D brain point to produce $\Gamma_{w',d',R}$, where $d'=d+3$,
$w'(x,y,z,m)\triangleq\beta\cdot w(x,y,z,m)$, $m=1,...,d$, $w'(x,y,z,d+1)\triangleq(1-\beta)\cdot x$,
$w'(x,y,z,d+2)\triangleq(1-\beta)\cdot y$ and $w'(x,y,z,d+3)\triangleq(1-\beta)\cdot z$
\item Calculate $\varepsilon=\max_{u\in\Gamma_{w',d',R}}\left\{ \min_{q\in\Gamma_{w',d',R},\, q\neq u}\left\{ ||u-q||^{2}\right\} \right\} $
\item Calculate $\Gamma_{w'',d'',R}$ ($d''<d$') by applying DM on $\Gamma_{w',d',R}$
with accuracy \textbf{$\delta$}, using the Gaussian kernel defined
in Eq. \ref{eq:gaussian} with $\varepsilon$ from step 7, where $w''(x,y,z,m)$,
$m=1,...,d''$, is the value of coordinate $m$ of the feature vector
in the embedded space, related to the 3D brain point $(x,y,z)$
\item Calculate $\Upsilon$= \textbf{SDK(}$\Gamma_{w'',d'',R}$\textbf{,
$\gamma$, $\tilde{k}$)} (Algorithm \ref{cap:Algorithm-The-silhouette-driven-k-means})
\item Calculate $\Upsilon^{cleaned}$ = \textbf{MergeIsolatedPoints(}$\Upsilon$\textbf{)}
(Algorithm \ref{cap:Algorithm-clean}) 
\item Return $\Upsilon^{cleaned}$ 
\end{enumerate}
\caption{The sub-nuclei segmentation procedure \label{cap:Algorithm-Sub-Nuclei-Segmentation}}
\end{algorithm}

\section{\label{sec:Experimental-Results6}Experimental results}

In this section we present the results after the application of the
SNF algorithm (Algorithm \ref{cap:The-Sub-Nuclei-Finder-algorithm})
on a multi-contrast MRI data of a human brain. The goal was to segment
the thalamus into its sub-nuclei. Details about the acquisition process
of the experimental data are given in Section \ref{sub:The-Input-Data}.
Using the ROI extraction procedure of the SNF algorithm (Section \ref{sub:ROI-Extraction})
on this data, we were able to detect the thalamus (Figure \ref{cap:Detecting-the-Thalamus}).
The volume of each detected thalamic part (right and left) was found
to be approximately 7050 $mm^{3}$. Our algorithm detected up to 13
clusters of the right/left thalamus. 

The \emph{spectrum} of a cluster is a vector, whose coordinates represent
different contrast-acquisition-method information that is related
to the points in this cluster. Specifically, each coordinate contains
a value associated with an image from a different (Algorithm \ref{cap:Algorithm-normalization-process})
contrast acquisition-method after normalization. A coordinate value
is the median across the values of the pixels that are associated
with the points of the cluster which were obtained from the relevant
contrast acquisition-method after normalization. The values of the
spectrum are ordered according to their appearance in Table \ref{cap:The-parameters-used}.
In the following, we formulate the calculation of a spectrum for a
given cluster.

Let $C$ be a cluster. For each point $c_{i}\in C$, $i=1,...,\left|C\right|$,
let $(x_{i},y_{i},z_{i})$ be the coordinates of $c_{i}$ in the 3D
space of the brain and let $I_{m}^{[0,1]}(x_{i},y_{i},z_{i})$ be
the value of the pixel obtained by contrast acquisition-method $m$
after normalization at point $(x_{i},y_{i},z_{i})$, $m=1,...,M$
(see Section \ref{sec:The-Algorithm-(Sub-Nuclei} for notation and
Algorithm \ref{cap:Algorithm-normalization-process} for the normalization
process). The spectrum of $C$ is defined as 
\begin{equation}
\sigma(C)\triangleq\left(\textrm{media}\textrm{n}_{i=1,...,\left|C\right|}\left\{ I_{1}^{[0,1]}(x_{i},y_{i},z_{i})\right\} ,...,\,\textrm{media}\textrm{n}_{i=1,...,\left|C\right|}\left\{ I_{M}^{[0,1]}(x_{i},y_{i},z_{i})\right\} \right)\,.\label{eq:Spectrum}
\end{equation}
The values of the coordinates of $\sigma$ are in the range of {[}0,1{]}.

\begin{figure}[!t]
\begin{centering}
\includegraphics[bb=80bp 210bp 530bp 600bp,clip,width=0.9\linewidth]{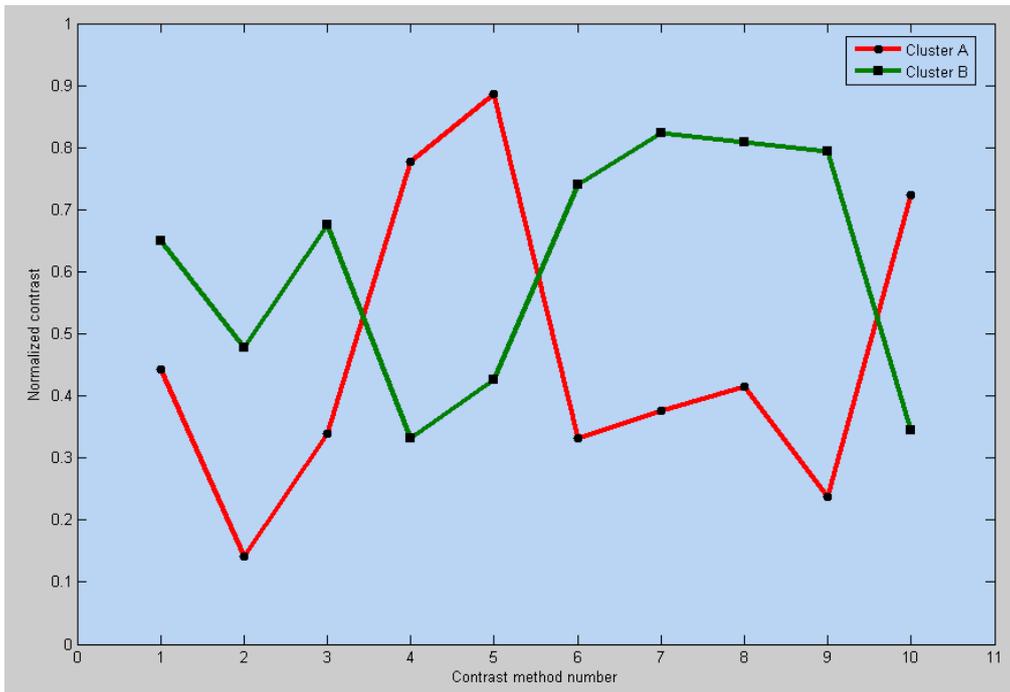}
\par\end{centering}

\caption{\label{cap:ClustersGraphs}Two dissimilar spectra (colored red and
green) of two different clusters. The \emph{x}-axis depicts the contrast
acquisition-method number from Table \ref{cap:The-parameters-used}
and the \emph{y}-axis depicts the values of the spectra for each contrast
acquisition-method.}
\end{figure}

Some of the clusters significantly differ in their spectrum, supporting
the hypothesis that nuclei can be separated by their multi-contrast
data. Figure \ref{cap:ClustersGraphs} presents two dissimilar spectra
of two different clusters.

\begin{figure}[!h]
\begin{centering}
\includegraphics[bb=80bp 210bp 530bp 600bp,clip,width=0.8\columnwidth]{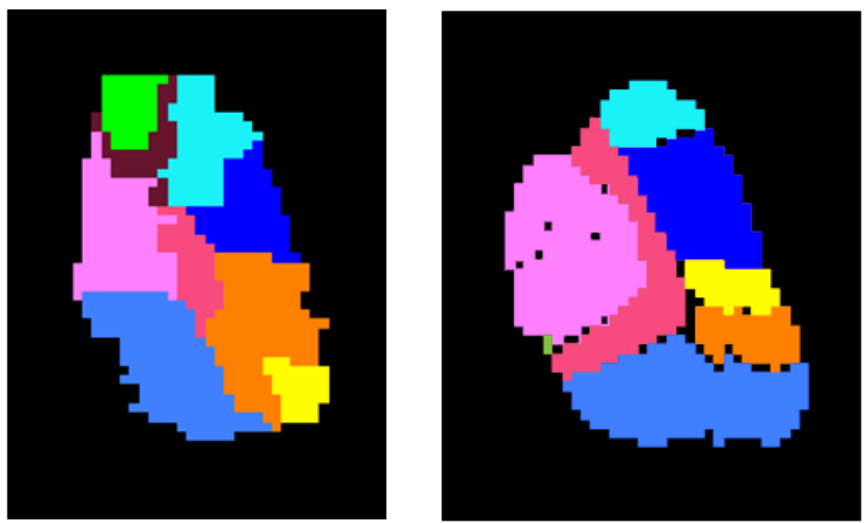}
\par\end{centering}

\caption{\label{cap:The-differences-in-atlases}The variability in the geometric
characteristics (shape, size and location) of the sub-nuclei among
histological atlases of the right thalamus as shown in two different
histological atlases. Left: Kikinis atlas \cite{Kikinis}. Right:
Morel atlas \cite{Morel97}. Each color represents a different nucleus.}
\end{figure}

Due to the variability that exists in the nuclei shapes, sizes and
locations in the brain of each individual (Section \ref{sub:The-Problem-of}),
there is no agreeable measurement that we can use to evaluate the
performance of the SNF algorithm. A combination of inter-subject variability
and utilization of different types of histological methods is the
cause for variability among different histological atlases. Figure
\ref{cap:The-differences-in-atlases} gives a visual demonstration
of this phenomenon. It shows the same slice from two different atlases,
colored according to the sub-nuclei arrangement. The sub-nuclei in
one atlas differ from their counterparts in the second atlas in their
geometric characteristics (shape, size and location).

Therefore, we measure the quality of the outputs from the SNF algorithm
by:
\begin{itemize}
\item comparing the detected nuclei locations to a histological atlas in
Section \ref{sub:Matching-To-Histological-Atlas}
\item matching between subjects in Section \ref{sub:Matching-Between-Subjects} 
\item matching between the left and the right parts of the thalamus of the
same subject in Section \ref{sub:Matching-the-Left}.
\end{itemize}
In Section \ref{sub:Assessing-the-Qauility} we demonstrate the contribution
of the image enhancement and the dimensionality reduction procedures
in the SNF algorithm. We do this by comparing the results produced
using versions of the SNF algorithm that exclude each of these procedures,
to the results produced using the full SNF algorithm, as defined in
Algorithm \ref{cap:The-Sub-Nuclei-Finder-algorithm}. Finally, in
Section \ref{sub:Summary-of-All}, we summarize the experimental results.

\subsection{\label{sub:The-Input-Data}The experimental data}

We tested our algorithm on a number of multi-contrast MRI data. The
data was acquired using a 3T GE 8-channel head coil MRI scanner. The
scans are of nine healthy males aged 25 to 30. Each subject was scanned
with ten different contrasts ($M=10$, see notation in Section \ref{sec:The-Algorithm-(Sub-Nuclei}),
which were achieved by variation of acquisition parameters such as
\emph{time to repeat} and \emph{time to echo}. Table \ref{cap:The-parameters-used}
describes the values of the parameters that were used for each contrast
acquisition-method. 

\begin{table}[!h]
\begin{centering}
{\footnotesize }%
\begin{tabular}{|l|>{\raggedright}p{2.4cm}|>{\centering}p{1.35cm}|>{\centering}p{1cm}|>{\centering}p{2cm}|l|}
\hline 
{\footnotesize \#} & \textbf{\footnotesize Contrast Acquisition-Method} & \textbf{\footnotesize Time to repeat ($ms$)} & \textbf{\footnotesize Time to echo ($ms$)} & \textbf{\footnotesize Experiment time ($minutes$)} & \textbf{\footnotesize Extra information}\tabularnewline
\hline 
\hline 
{\footnotesize 1} & {\footnotesize FLAIR} & {\footnotesize 9000} & {\footnotesize 140} & {\footnotesize 4:50} & {\footnotesize Time to invert = 2100 }\emph{\footnotesize ms}\tabularnewline
\hline 
{\footnotesize 2} & {\footnotesize T2 weighted} & {\footnotesize 7000} & {\footnotesize 150} & {\footnotesize 3:00} & {\footnotesize Echo train length = 32}\tabularnewline
\hline 
{\footnotesize 3} & {\footnotesize Proton Density} & {\footnotesize 7000} & {\footnotesize 6} & {\footnotesize 3:00} & {\footnotesize Echo train length = 32}\tabularnewline
\hline 
{\footnotesize 4} & {\footnotesize T1 weighted} & {\footnotesize 550} & {\footnotesize 8} & {\footnotesize 5:00} & \tabularnewline
\hline 
{\footnotesize 5} & {\footnotesize T1 + MgT} & {\footnotesize 550} & {\footnotesize 8} & {\footnotesize 6:20} & \multicolumn{1}{c|}{{\footnotesize Irradiation frequency = 1.2 }\emph{\footnotesize kHz}}\tabularnewline
\hline 
{\footnotesize 6} & {\footnotesize T2 short TE} & {\footnotesize 600} & {\footnotesize 2} & {\footnotesize 5:00} & {\footnotesize Flip Angle = $20^{o}$}\tabularnewline
\hline 
{\footnotesize 7} & {\footnotesize T2 medium TE} & {\footnotesize 600} & {\footnotesize 15} & {\footnotesize 5:00} & {\footnotesize Flip Angle = $20^{o}$}\tabularnewline
\hline 
{\footnotesize 8} & {\footnotesize T2 long TE} & {\footnotesize 600} & {\footnotesize 32} & {\footnotesize 5:00} & {\footnotesize Flip Angle = $20^{o}$}\tabularnewline
\hline 
{\footnotesize 9} & {\footnotesize STIR} & {\footnotesize 5000} & {\footnotesize 25} & {\footnotesize 3:00} & {\footnotesize Time to invert = 130 $ms$}\tabularnewline
\hline 
{\footnotesize 10} & {\footnotesize SPGR} & {\footnotesize 400} & {\footnotesize 2} & {\footnotesize 2:30} & {\footnotesize Flip Angle = $12^{o}$}\tabularnewline
\hline 
\end{tabular}
\par\end{centering}{\footnotesize \par}

\caption{\label{cap:The-parameters-used}The parameters used in the acquisition
of each contrast acquisition-method. \protect \\
\textbf{FLAIR: F}luid \textbf{L}evel \textbf{A}ttenuated \textbf{I}nversion
\textbf{R}ecovery\textbf{, M}g\textbf{T: M}agnetization \textbf{T}ransfer,
\textbf{STIR:} \textbf{S}hort \textbf{T}au \textbf{I}nversion \textbf{R}ecovery,
\textbf{SPGR:} \textbf{S}poiled \textbf{G}radient \textbf{R}ecalled
\textbf{E}cho.}
\end{table}

\begin{figure}[!h]
\begin{centering}
\includegraphics[bb=80bp 210bp 530bp 600bp,clip,width=0.9\columnwidth]{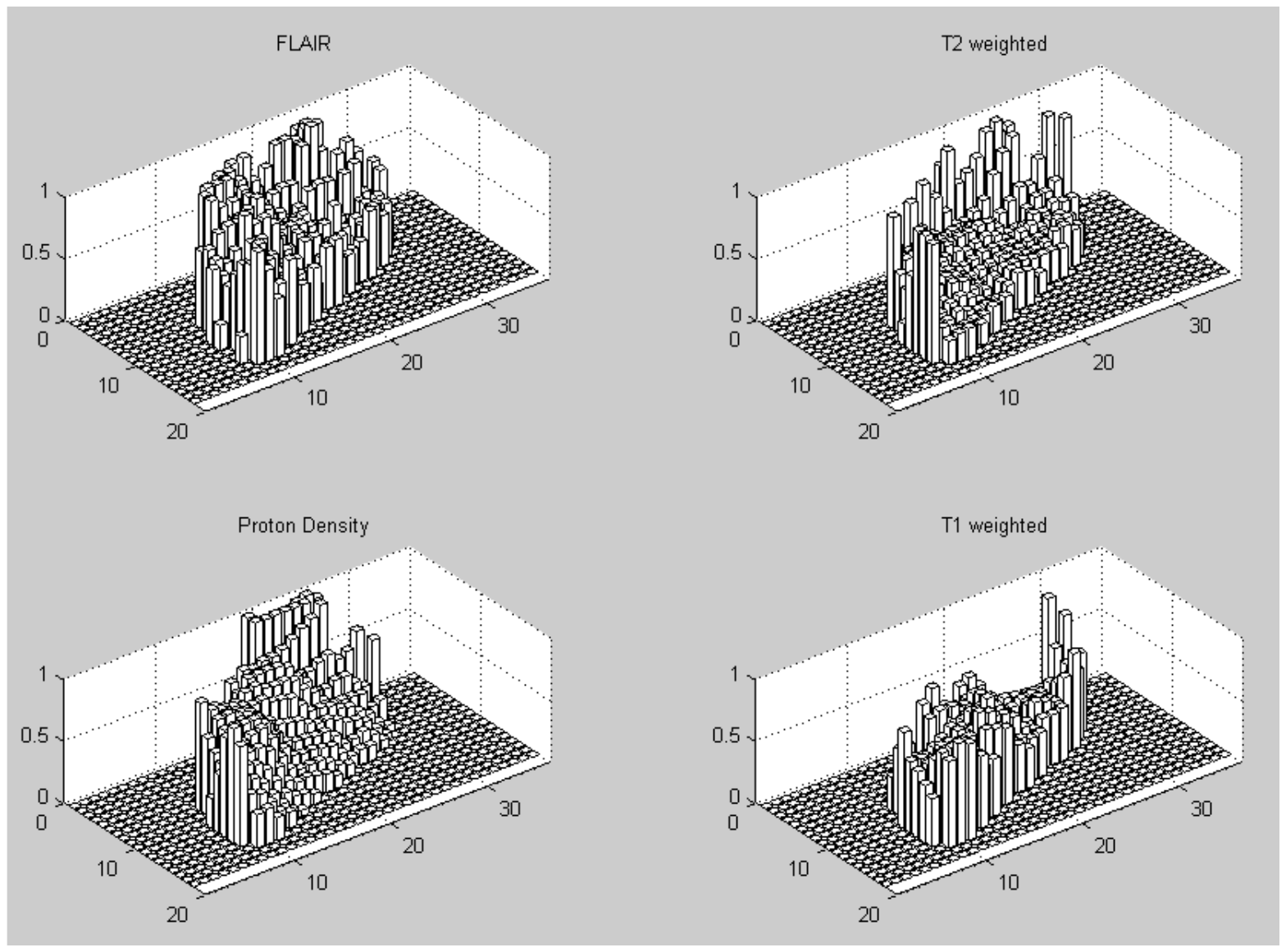}
\par\end{centering}

\caption{\label{cap:1-4}The variability among the normalized values of contrast
acquisition-methods 1-4 in slice 27 of the right thalamus of one of
the subjects.}
\end{figure}

\begin{figure}[!h]
\begin{centering}
\includegraphics[bb=80bp 210bp 530bp 600bp,clip,width=0.9\columnwidth]{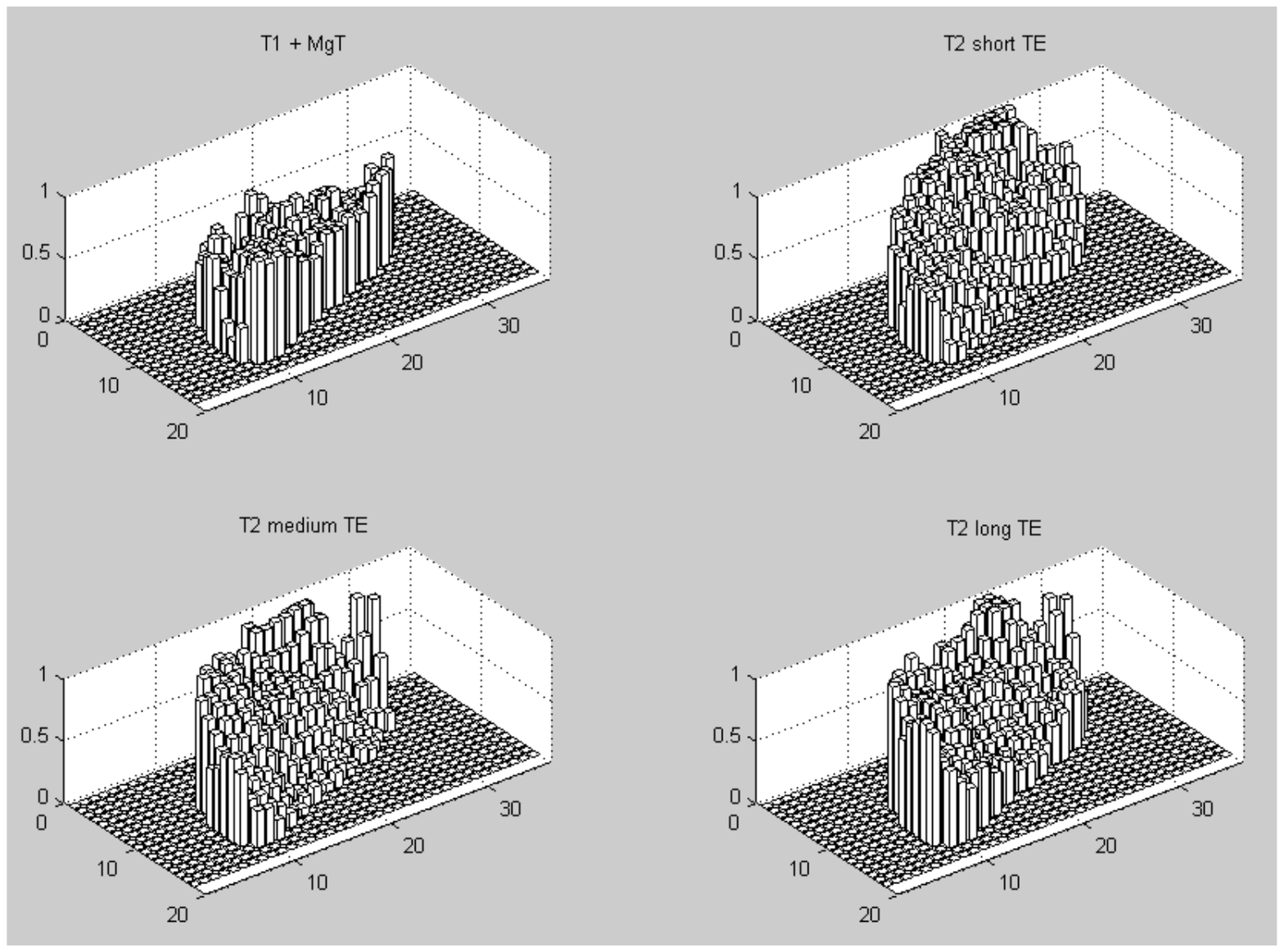}
\par\end{centering}

\caption{\label{5-8}The variability among the normalized values of contrast
acquisition-methods 5-8 in slice 27 of the right thalamus of one of
the subjects.}
\end{figure}

\begin{figure}[!h]
\begin{centering}
\includegraphics[bb=80bp 300bp 530bp 500bp,clip,width=0.9\columnwidth]{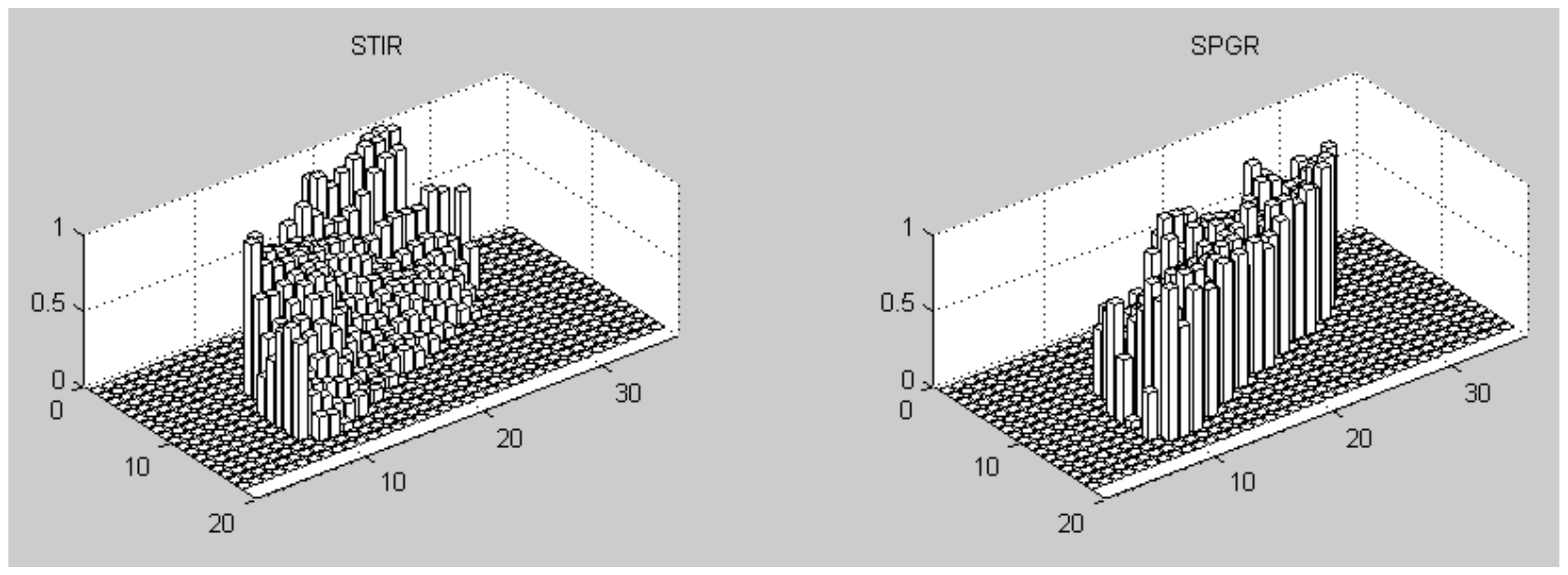}
\par\end{centering}

\caption{\label{9-10}The variability among the normalized values of contrast
acquisition-methods 9-10 in slice 27 of the right thalamus of one
of the subjects.}
\end{figure}

\begin{figure}[!h]
\begin{centering}
\includegraphics[bb=80bp 300bp 530bp 500bp,clip,width=0.98\linewidth]{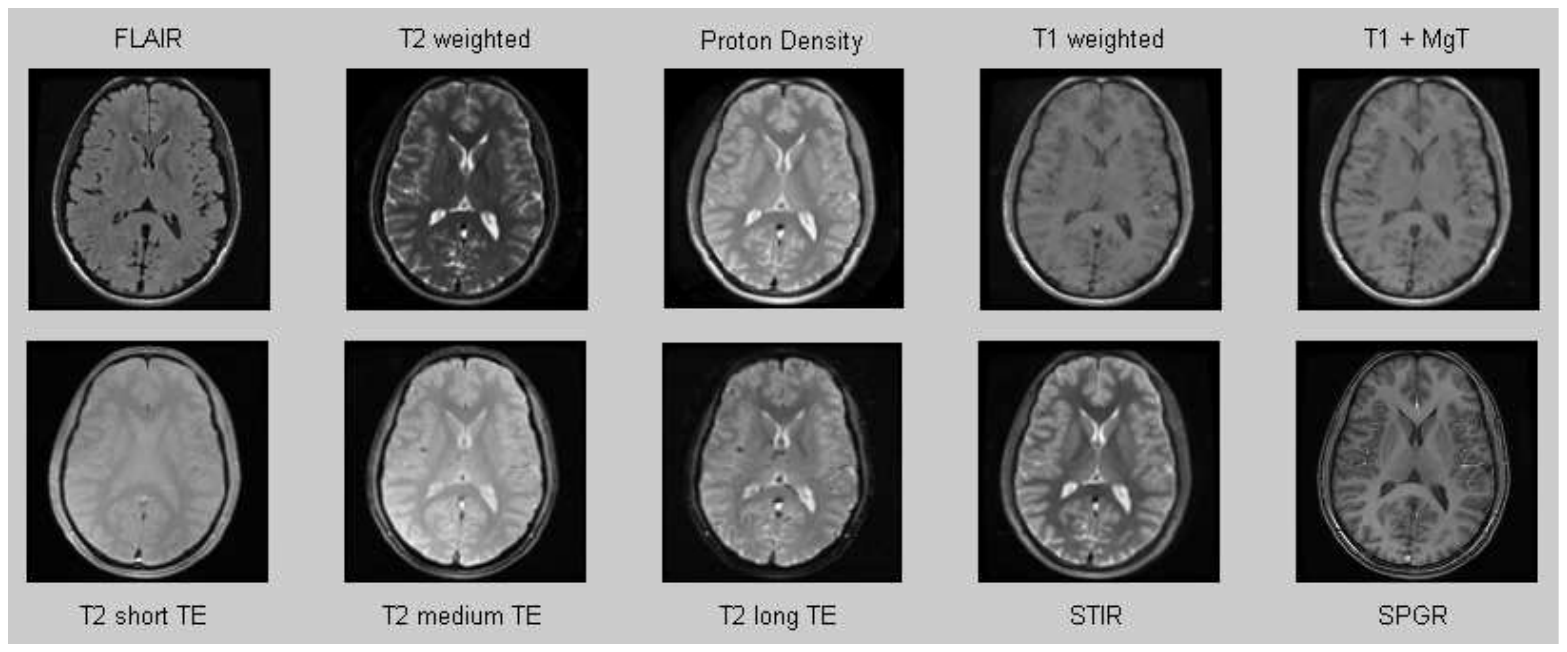}
\par\end{centering}

\caption{\label{cap:The-resulted-images}The images of all ten contrast acquisition-method
of slice 28 of one of the subjects.}
\end{figure}

Each specific contrast scan produced 48 images made in the axial plane
with a Field of View (FOV) of $20\times20$ $cm^{2}$. Each image
represents a slice, 1.5 $mm$ thick, with no gaps and contains $128\times128$
pixels with spatial resolution of 1.5 $mm$ in both axes. Hence, the
scanning provides a 3D resolution of $1.5\times1.5\times1.5$ $mm^{3}$.
After the scanning phase, the various contrast images, which belong
to the same subject and to the same slice, were realigned in order
to correct head motion corrections. In addition, for comparison reasons,
all subjects volumes were co-registered and re-sliced to a single
chosen subject. Figures \ref{cap:1-4}-\ref{cap:The-resulted-images}
give a visual illustration of the variability between the different
contrast acquisition-methods. Figures \ref{cap:1-4}, \ref{5-8} and
\ref{9-10} show the values of the ten resulted contrasts of the thalamus
ROI for a specific slice of one of the subjects. Figure \ref{cap:The-resulted-images}
shows the ten contrast images (one for each contrast acquisition-method)
for a specific slice of one of the subjects.

\subsection{\label{sub:Matching-To-Histological-Atlas}Matching to histological
atlas}

In order to evaluate the results of the SNF algorithm (Algorithm \ref{cap:The-Sub-Nuclei-Finder-algorithm})
applied on the multi-contrast MRI data, we compare them to a histological
atlas. We apply the SNF algorithm on the data that belongs to one
of the nine subjects from Section \ref{sub:The-Input-Data}. We use
the atlas of Kikinis \cite{Kikinis} for the comparison. 

The atlas is first digitized and registered to the subject. Each voxel
of the atlas is labeled according to the nucleus it belongs to. After
labeling the atlas voxels, we compute the centroids of each labeled
cluster in the 3D space. In addition, we compute the centroids of
the clusters produced by the SNF algorithm on the subject. Then, we
calculate in the common 3D space the distances between the centroids
of the subject and the centroids of the atlas. We define the distance
between two clusters as the distance between their centroids. Every
cluster of the subject is assigned the label of the cluster in the
atlas that is closest to it. This process may assign the same label
to two or more different clusters. This phenomenon is acceptable for
the sake of comparison and it indicates that the segmentation separates
the data into more sub-nuclei than the ones in this specific atlas. 

\begin{figure}[!h]
\begin{centering}
\includegraphics[bb=86bp 359bp 508bp 482bp,clip,width=1\linewidth]{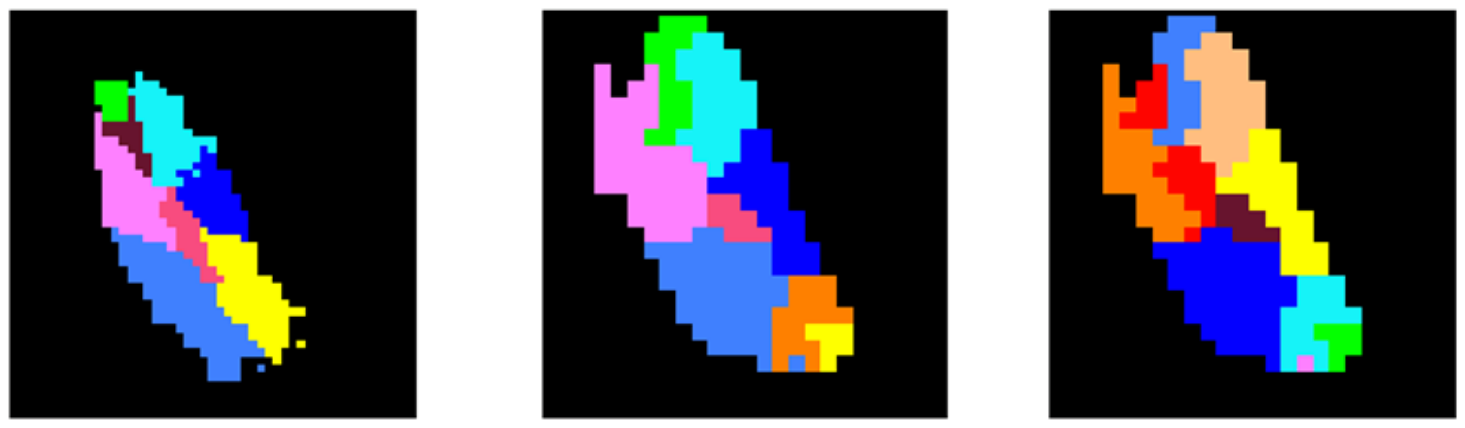}
\par\end{centering}

\caption{\label{cap:Slice-27-Matched}A comparison between the Kikinis histological
atlas \cite{Kikinis} and the output of the SNF algorithm applied
on the right thalamus of a subject. Left: slice 27 from the atlas.
Middle: slice 27 of the result from matching between the atlas and
the output of the SNF algorithm. Right: slice 27 of the same output
before the match (the colors palette is random). Each color represents
a different nucleus. There is a strong resemblance between the geometric
characteristics (shape, size and location) of the matched nuclei and
those of their counterparts in the atlas.}
\end{figure}

\begin{figure}[!h]
\begin{centering}
\includegraphics[bb=80bp 250bp 530bp 550bp,clip,width=1\linewidth]{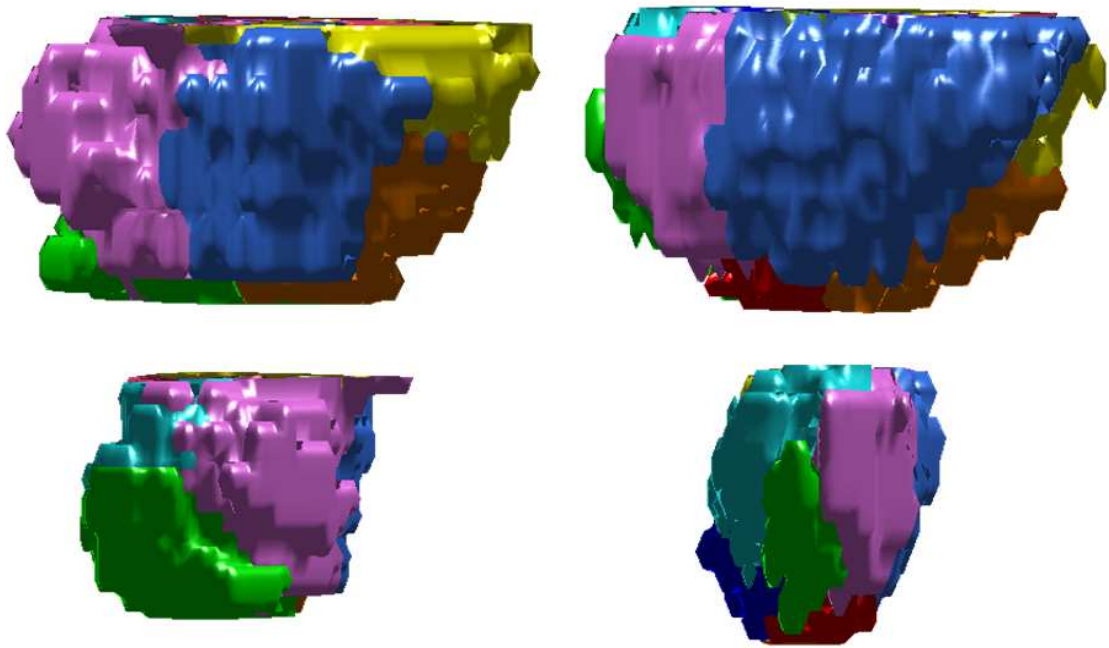}
\par\end{centering}

\caption{\label{cap:Match3D}3D view corresponding to Fig. \ref{cap:Slice-27-Matched}.
Left column: the result from matching between the atlas and the output
of the SNF algorithm. Right column: similar angles from the atlas.
Each color represents a different nucleus. There is a strong resemblance
between the geometric characteristics (shape, size and location) of
the matched nuclei and those of their counterparts in the atlas.}
\end{figure}

The geometric characteristics (shape, size and location) of the matched
segmented nuclei were consistent with the registered atlas. The mean
distance error from the segmented nuclei centroids to their matched
atlas nuclei centroids was 2.8 $mm$ (while the total volume of the
right thalamus was found to be 7,043 $mm^{3}$). A perfect match is
not feasible due to the inter-subject variability (Section \ref{sub:The-Problem-of})
and the imperfect registration. Figure \ref{cap:Slice-27-Matched}
shows a single slice from the result of a match between the atlas
and the output of the SNF algorithm applied on the data of the right
thalamus of one of the subjects. Figure \ref{cap:Match3D} shows some
3D views of this match. There is high resemblance between the geometric
characteristics of the matched nuclei and the geometric characteristics
of their counterparts in the atlas.

\subsection{\label{sub:Matching-Between-Subjects}Match between subjects}

\begin{figure}[!h]
\begin{centering}
\includegraphics[bb=80bp 250bp 530bp 500bp,clip,width=0.65\columnwidth]{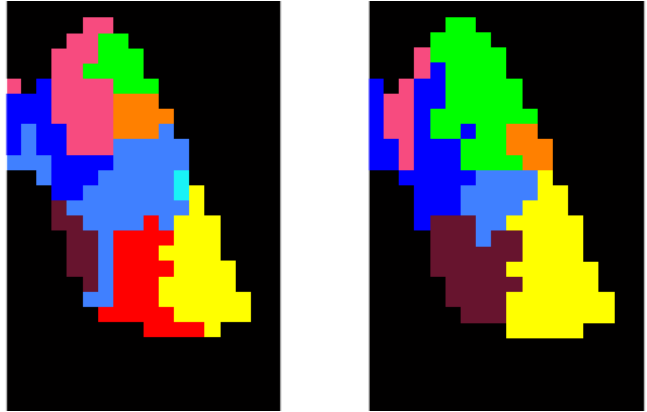}
\par\end{centering}

\caption{\label{Matching-Subjects}A comparison between the outputs of the
SNF algorithm applied on the data of the right thalami of two different
subjects. Left: slice 26 of the output of the SNF algorithm applied
on the right thalamus of the first subject. Right: slice 26 of the
matching result between the output of the SNF algorithm applied on
the right thalamus of the second subject and the output of the SNF
algorithm applied on the right thalamus of the first subject. Each
color represents a different nucleus. There is a resemblance between
the geometric characteristics (shape, size and location) of the matched
nuclei, despite the inter-subject variability (Section \ref{sub:The-Problem-of}).}
\end{figure}

\begin{figure}[!h]
\begin{centering}
\includegraphics[bb=80bp 330bp 530bp 470bp,clip,width=1\linewidth]{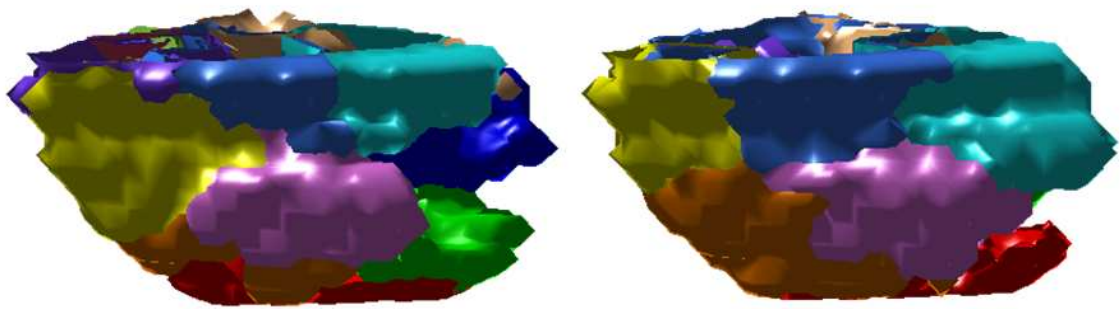}
\par\end{centering}

\caption{\label{cap:Match3DSubjects}3D view corresponding to  Fig. \ref{Matching-Subjects}.
Left: the output of the SNF algorithm applied on the right thalamus
of the first subject. Right: a similar angle of the matching result
of the SNF algorithm applied on the right thalamus of the first subject.
Each color represents a different nucleus. There is a resemblance
between the geometric characteristics (shape, size and location) of
the matched nuclei, despite the inter-subject variability (Section
\ref{sub:The-Problem-of}).}
\end{figure}

\begin{figure}[!h]
\begin{centering}
\includegraphics[bb=80bp 220bp 530bp 600bp,clip,width=0.85\linewidth]{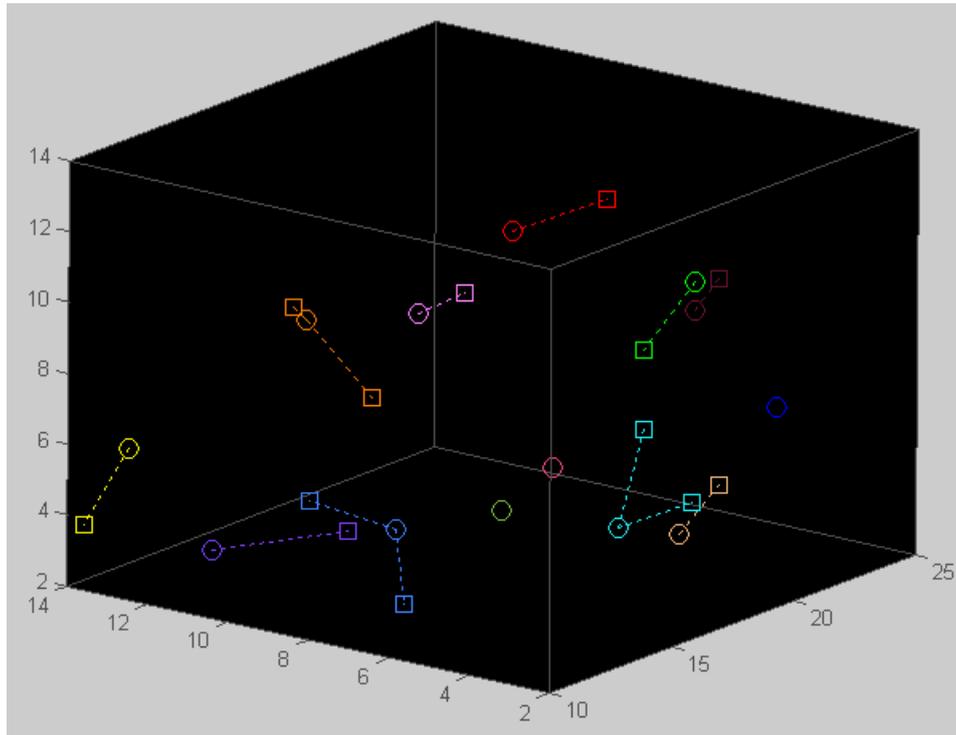}
\par\end{centering}

\caption{\label{Matching-Subjects-Graphs}A 3D view corresponding to  a match
of the centroids of the nuclei that were detected by the SNF algorithm
applied on the data of the right thalami of two different subjects.
The centroids of the nuclei that belong to the first subject are marked
with circles and the centroids of the nuclei that belong to the second
subject are marked with squares. The dotted lines indicate the matching
errors, which are due to the inter-subject variability (Section \ref{sub:The-Problem-of})}
\end{figure}

We further examine the performance of the SNF algorithm (Algorithm
\ref{cap:The-Sub-Nuclei-Finder-algorithm}) by comparing between the
results it produced for different subjects. We used the same matching
technique that was presented in Section \ref{sub:Matching-To-Histological-Atlas}.
Figure \ref{Matching-Subjects} shows a single slice from the result
of a match between the outputs of the SNF algorithm applied on the
data of the right thalami of two different subjects. Figure \ref{cap:Match3DSubjects}
shows a 3D view of this match. There is a certain resemblance between
the geometric characteristics (shape, size and location) of the matched
nuclei, despite the inter-subject variability (Section \ref{sub:The-Problem-of}).
The mean distance error between the matched centroids was 3.55 $mm$.
Figure \ref{Matching-Subjects-Graphs} shows the matched centroids
in the 3D brain space. It shows the matching errors that are due to
the imperfect match.

\begin{figure}[!h]
\begin{centering}
\includegraphics[bb=80bp 250bp 530bp 550bp,clip,width=0.85\linewidth]{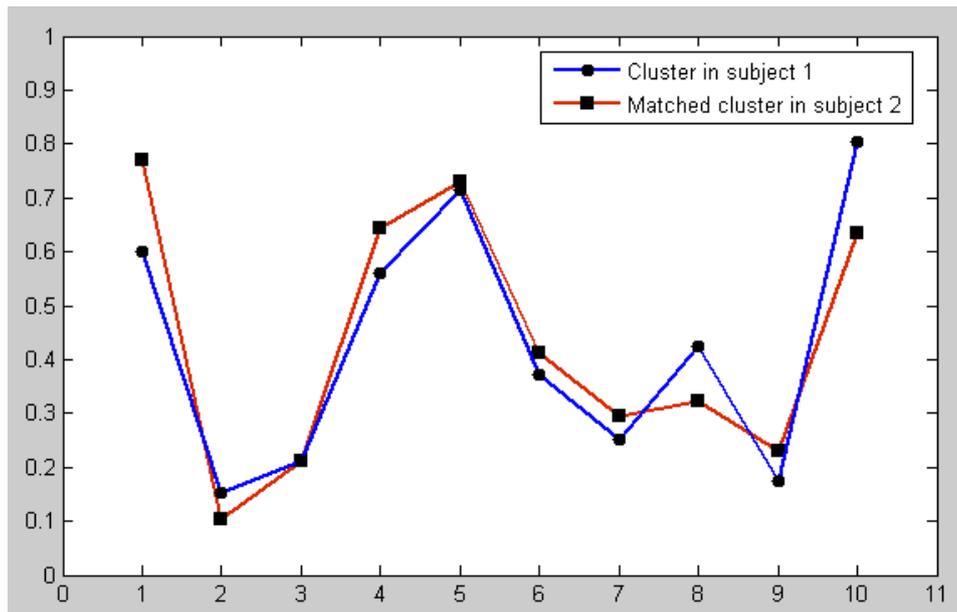}
\par\end{centering}

\caption{\label{Matching-Subjects-Spectrums} A comparison between spectra
of two matched clusters that were detected by the SNF algorithm applied
on the data of the right thalami of two different subjects. The \emph{x}-axis
depicts the contrast acquisition-method number from Table \ref{cap:The-parameters-used}
and the \emph{y}-axis depicts the values of the spectra for each contrast
acquisition-method. There is a strong resemblance between the two
spectra in most of the contrast acquisition-methods.}
\end{figure}

We also examined the spectra (Eq. \ref{eq:Spectrum}) of the matched
clusters. Figure \ref{Matching-Subjects-Spectrums} illustrates the
high resemblance between the spectra of two matched clusters from
two different subjects. % To evaluate the extent of the spectra resemblance, we calculate the vector whose coordinates are the absolute differences between the spectra of a pair of matched clusters: (0.172, 0.075, 0.131, 0.101, 0.102, 0.148, 0.139, 0.199, 0.100, 0.088). One of the clusters belongs to the first subject and the other cluster belongs to the second subject. Later, these error measurements are used for comparison with additional matching results. 

\subsection{\label{sub:Matching-the-Left}Matching between the two parts of the
thalamus}

\begin{figure}[!h]
\begin{centering}
\includegraphics[bb=80bp 250bp 530bp 500bp,clip,width=0.57\linewidth]{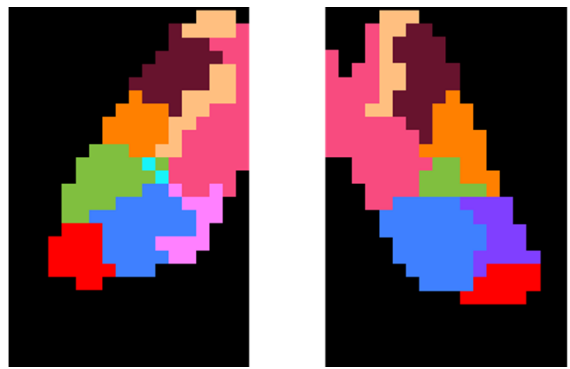}
\par\end{centering}

\caption{\label{Matching-Right-To-Left}A comparison between the outputs of
the SNF algorithm applied on the two parts of the thalamus of a subject.
Left: slice 27 of the output of the SNF algorithm applied on the left
part of the thalamus of the subject. Right: slice 27 of the result
from matching between the output of the SNF algorithm applied on the
right part of the thalamus of the subject and the output of the SNF
algorithm applied on the left part of the thalamus of the subject.
Each color represents a different nucleus. There is a resemblance
between the geometric characteristics (shape, size and location) of
the matched nuclei. However, it is not a perfect match since the thalamus
right and left parts are not identical.}
\end{figure}

\begin{figure}[!h]
\begin{centering}
\includegraphics[bb=80bp 330bp 530bp 460bp,clip,width=0.9\linewidth]{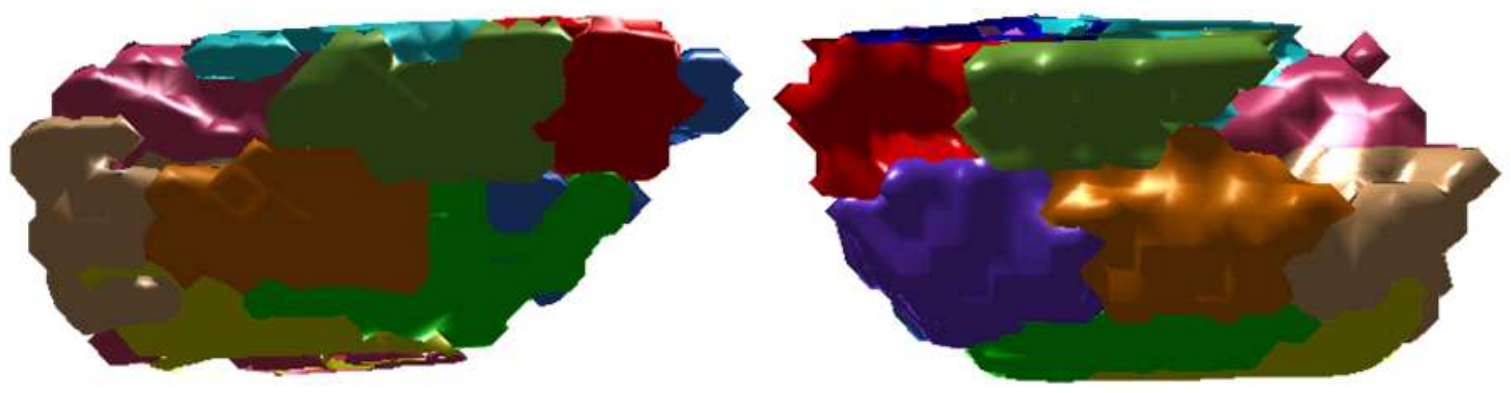}
\par\end{centering}

\caption{\label{cap:Match3DRightToLeft}A 3D view corresponding to  Fig. \ref{Matching-Right-To-Left}.
Left: the output of the SNF algorithm applied on the left part of
the thalamus of the subject. Right: a similar angle of the result
of the SNF algorithm applied on the right part of the thalamus of
the subject. Each color represents a different nucleus. There is a
resemblance between the geometric characteristics (shape, size and
location) of the matched nuclei. However, it is not a perfect match
since the thalamus right an left parts are not identical.}
\end{figure}

In order to further evaluate and validate the output of the SNF algorithm
(Algorithm \ref{cap:The-Sub-Nuclei-Finder-algorithm}), we take advantage
of the fact that the left part and the right part of the thalamus
are similar (Figure \ref{cap:Detecting-the-Thalamus}). We applied
the SNF algorithm separately on each of these thalamus parts of a
subject. The output from the right part to resembles the output from
the left part. We use the matching technique that was presented in
Section \ref{sub:Matching-To-Histological-Atlas} to evaluate the
similarity. The matching technique used a mirror image of the left
part of the thalamus. Figure \ref{Matching-Right-To-Left} demonstrates
the matching between the outputs of the SNF algorithm applied on the
data of the two thalamus parts of the same subject. Figure \ref{cap:Match3DRightToLeft}
shows a 3D view of this match. There is a resemblance between the
geometric characteristics (shape, size and location) of the matched
nuclei. However, it is not a perfect match since the thalamus right
and left parts are not identical.

\begin{figure}[!h]
\begin{centering}
\includegraphics[bb=80bp 250bp 550bp 550bp,clip,width=0.9\linewidth]{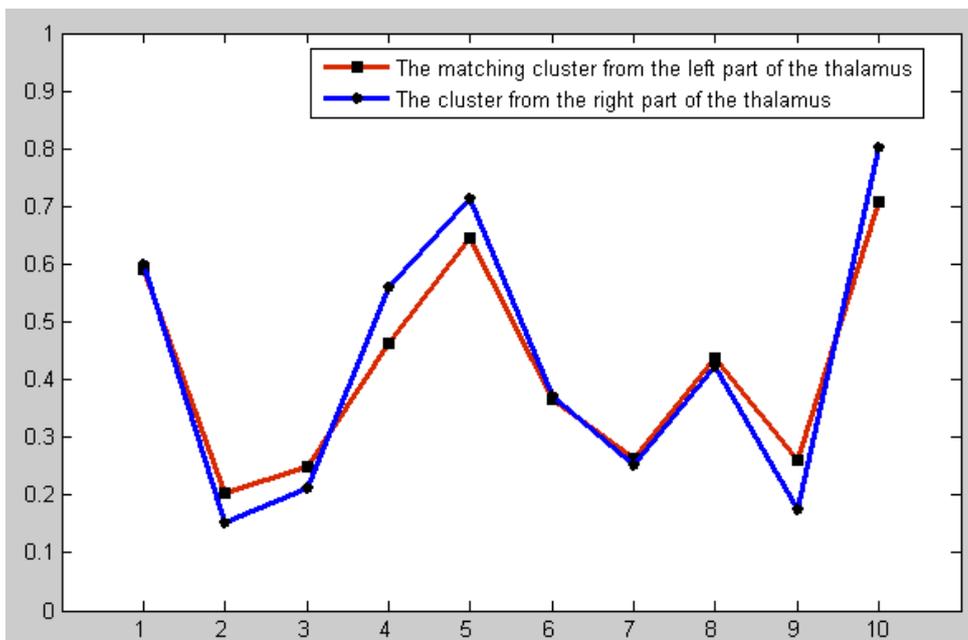}
\par\end{centering}

\caption{\label{Matching-Right-Left-Spectrums} A comparison between spectra
of two matched clusters: one cluster is part of the result of the
SNF algorithm applied on the left side of the thalamus of a subject
(in red, marked with squares) and the other cluster is part of the
result of the SNF algorithm applied on the right side of the thalamus
of the same subjects (in blue, marked with circles). The \emph{x}-axis
depicts the contrast acquisition-method number from Table \ref{cap:The-parameters-used}
and the \emph{y}-axis depicts the mean of the normalized values of
the contrast acquisition-methods in the cluster. There is a high resemblance
between the two spectra in most of the contrast-acquisition-methods.}
\end{figure}

To achieve better accuracy, a more advanced registration technique
is needed to register the left part and the right part since there
is not a perfect match between their shapes. The mean distance error
between matched centroids was 4.14 $mm$. Figure \ref{Matching-Right-Left-Spectrums}
illustrates high resemblance between two spectra that belong to two
matched clusters: one cluster is part of the output of the SNF algorithm
applied on the left side of the thalamus of a subject and the other
cluster is part of the output of the SNF algorithm applied on the
right side of the thalamus of the same subject. 

%To evaluate the extent of the spectra (Eq. ) resemblance, we calculate the vector whose coordinates are the absolute differences between the spectra of a pair of matched clusters: (0.040, 0.044, 0.087, 0.063, 0.050, 0.087, 0.087, 0.052, 0.109, 0.0754). One of the clusters belongs to the right part and the other cluster belongs to the left part. A comparison between this vector and the similar vector from section  (calculated for the same cluster from the right part) indicates that the spectra of two parts of the same subject are closer than the spectra of the same part in two different subjects. 

\begin{figure}[!h]
\begin{centering}
\includegraphics[bb=129bp 261bp 486bp 530bp,clip,width=0.57\linewidth]{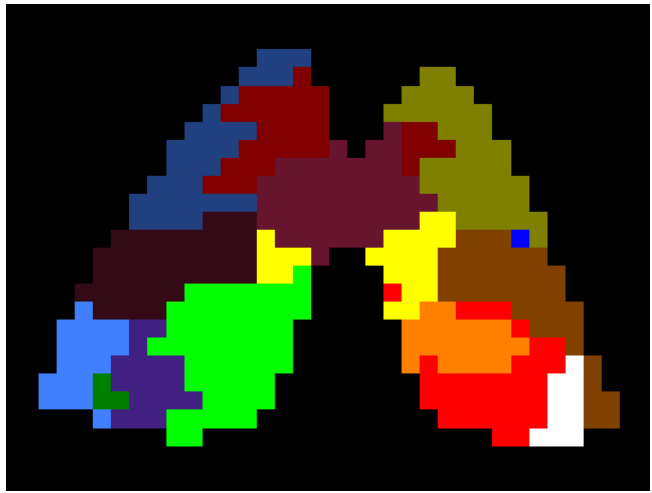}
\par\end{centering}

\caption{\label{Right+Left} Slice 26 of the output of the SNF algorithm applied
on the whole thalamus (both the left part and the right part are included
in the ROI). There is a geometric (shape, size and location) resemblance
between the segments from the two parts. Matched clusters from the
left part and the right part are not necessarily colored in the same
color since they were discovered as different clusters due to the
use of the spatial information.}
\end{figure}

We also apply the SNF algorithm on the whole thalamus area, i.e. on
the ROI that includes both the left and the right parts of the thalamus.
The result is depicted in Fig. \ref{Right+Left}. Matched clusters
from the left part and the right part are not necessarily colored
in the same color since they were discovered as different clusters
due to the use of the spatial information.

\subsection{\label{sub:Assessing-the-Qauility}Contribution of the main steps
of the SNF algorithm}

In this section we show the benefits of using the image enhancement
and the dimensionality reduction steps of the SNF algorithm. Following
are some comparisons between results of the SNF algorithm using all
its steps (we refer to this version as the \emph{full} SNF algorithm)
and results of the SNF algorithm where each of these steps is excluded
(we refer to these versions as the \emph{partial} SNF algorithms).

\paragraph{\label{sub:Image-Enhancement-reduction}Image enhancement}

\begin{figure}[!h]
\begin{centering}
\includegraphics[bb=80bp 250bp 530bp 500bp,clip,width=1\linewidth]{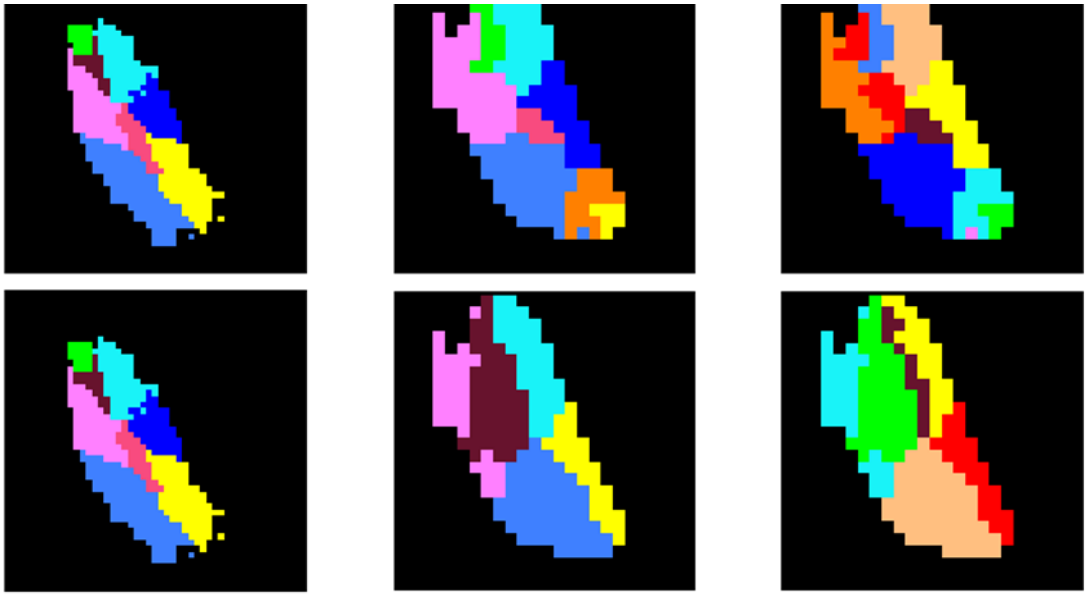}
\par\end{centering}

\caption{\label{cap:Slice-27-Matched-No_Enhancement}The contribution of the
image enhancement step to the performance of the SNF algorithm. Top
row: slice 27 from the result of a match between the Kikinis histological
atlas \cite{Kikinis} and the output of the full SNF algorithm (including
the image enhancement step) applied on the right thalamus of a subject.
Bottom row: slice 27 from the result of a match between the histological
atlas and the output of the partial SNF algorithm (without the image
enhancement step) on the same data. Left column: slice 27 from the
atlas. Middle column: slice 27 of the matching result between the
output of the relevant SNF algorithm version (full/partial) applied
on the data and the atlas. Right column: slice 27 of the same output
of the relevant SNF algorithm version (full/partial) before the match
(the color palette is random). Each color represents a different nucleus.
A comparison between the geometric characteristics (shape, size and
location) of the detected nuclei (by both algorithm versions) and
their counterparts in the atlas, reveals a better similarity to the
output of the full SNF algorithm (top row).}
\end{figure}

We exclude the image enhancement procedure from the SNF algorithm
and apply it on the same subject that was examined in Section \ref{sub:Matching-To-Histological-Atlas}.
The results were poor in comparison with the results of the full SNF
algorithm. The partial SNF algorithm detected only 6 clusters in the
right thalamus that were matched to 5 nuclei of the atlas, while the
full SNF algorithm detected 13 clusters, that were matched to 10 nuclei
of the atlas. The superiority of the full SNF algorithm stems from
the noise reduction and the non-linear feature amplification mechanism
of the wavelet-based image enhancement (Section \ref{sec:Wavelet-for-Contrast}).
Figure \ref{cap:Slice-27-Matched-No_Enhancement} illustrates this
comparison. The number of sub-nuclei, which were detected by the application
of the partial SNF algorithm (excluding the image enhancement step),
is less than the number of sub-nuclei, that were detected by the application
of the full SNF algorithm (including the image enhancement step).
Moreover, a comparison between the geometric characteristics (shape,
size and location) of the detected nuclei (by both algorithm versions)
and their counterparts in the atlas, reveals that the result of the
full SNF algorithm bares a stronger similarity to the atlas. 

\paragraph{\label{sub:Dimensionality-Reduction-reduction}Dimensionality reduction}

\begin{figure}[!h]
\begin{centering}
\includegraphics[bb=80bp 250bp 530bp 500bp,clip,width=1\linewidth]{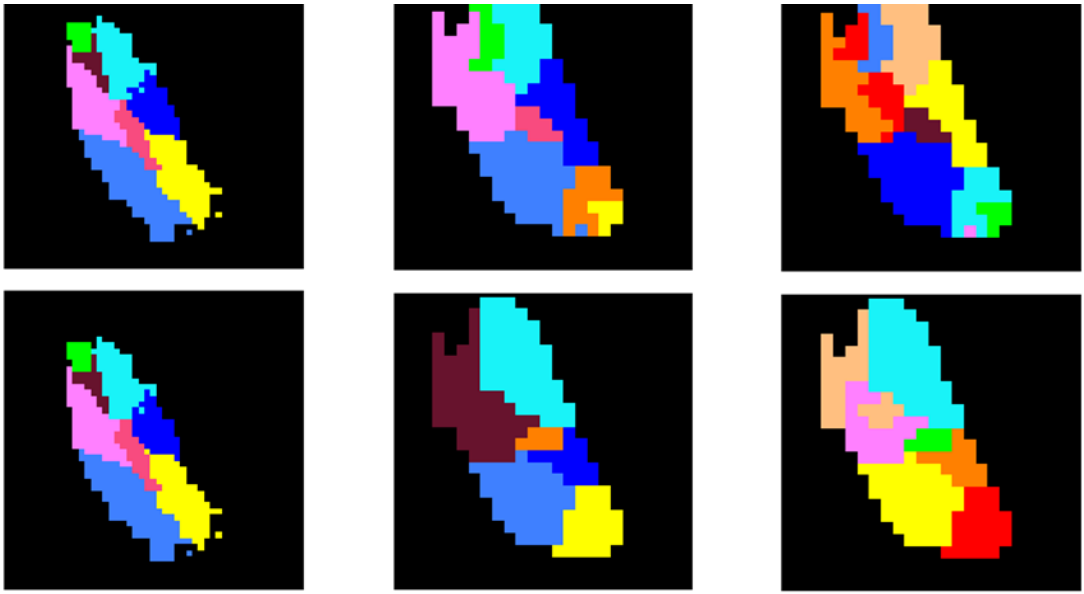}
\par\end{centering}

\caption{\label{cap:Slice-27-Matched-No_Dimensionality}The contribution of
the dimensionality reduction step to the performance of the SNF algorithm.
Top row: slice 27 from the result of a match between the histological
atlas \cite{Kikinis} and the output of the full SNF algorithm (including
the dimensionality reduction step) applied on the right thalamus of
a subject. Bottom row: slice 27 from the result of a match between
the histological atlas and the output of the partial SNF algorithm
(without the dimensionality reduction step) on the same data. Left
column: slice 27 from the atlas. Middle column: slice 27 of the matching
result between the output of the relevant SNF algorithm version (full/partial)
applied on the data and the atlas. Right column: slice 27 of the same
output of the relevant SNF algorithm version (full/partial) before
the match (the colors palette is random). Each color represents a
different nucleus. A comparison between the geometric characteristics
(shape, size and location) of the detected nuclei (by both algorithm
versions) and their counterparts in the atlas, reveals a better similarity
to the output of the full SNF algorithm (top row).}
\end{figure}

We exclude the dimensionality reduction phase from the SNF algorithm
and apply it on the data of the same subject that we examined in Section
\ref{sub:Matching-To-Histological-Atlas}. There was substantial degradation
in the quality of the results: the partial SNF algorithm detected
only 10 clusters in the right thalamus, that were matched to 8 nuclei
in the atlas, while the full SNF algorithm detected 13 clusters, that
were matched to 10 nuclei in the atlas. Figure \ref{cap:Slice-27-Matched-No_Dimensionality}
illustrates this comparison. The number of sub-nuclei, which were
detected by the partial SNF algorithm (excluding the dimensionality
reduction step), is less than the number of sub-nuclei, which were
detected by the full SNF algorithm (including the dimensionality reduction
step). Moreover, a comparison between the geometric characteristics
(shape, size and location) of the detected nuclei (by both algorithm
versions) and their counterparts in the atlas, reveals a better similarity
to the output of the full SNF algorithm.

Furthermore, the time complexity of the partial SNF algorithm was
much higher in comparison to that of the full SNF algorithm: 170 $seconds$
versus 106 $seconds$ on a Pentium4 2.4$GHz$ with 768$MB$ RAM. This
is due to the clustering procedure, whose time complexity significantly
grows with increase of the dimensionality of the data.

\begin{figure}[!h]
\begin{centering}
\includegraphics[bb=80bp 250bp 530bp 500bp,clip,width=1\linewidth]{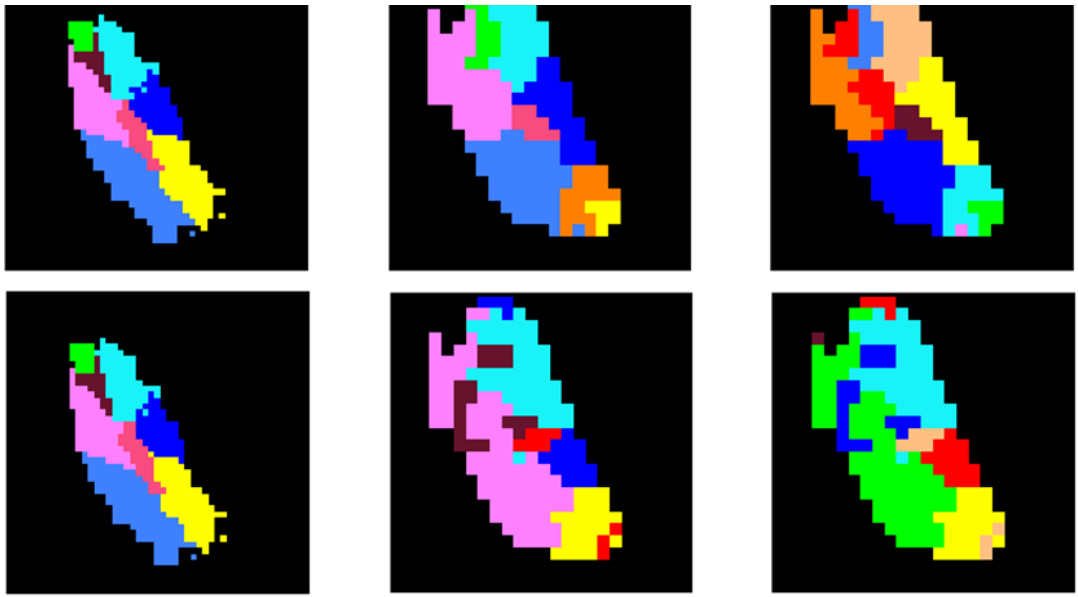}
\par\end{centering}

\caption{\label{cap:Slice-27-Matched-PCA}A comparison between the output of
the DM-based SNF algorithm (the original SNF algorithm) and the output
of the PCA-based SNF algorithm. Top row: slice 27 from the result
of a match between the Kikinis histological atlas \cite{Kikinis}
and the output of the DM-based SNF algorithm applied on the right
thalamus of a subject. Bottom row: slice 27 of the result from matching
between the atlas and the output of the PCA-based SNF algorithm applied
on the same data. Left column: slice 27 from the atlas. Middle column:
slice 27 of the result from matching between the atlas and the output
of the relevant SNF algorithm version (DM-based/PCA-based) applied
on the same data. Right column: slice 27 of the same output of the
relevant SNF algorithm version (DM-based/PCA-based) before the match
(the colors palette is random). Each color represents a different
nucleus. A comparison between the geometric characteristics (shape,
size and location) of the detected nuclei (by both algorithm versions)
and their counterparts in the atlas, reveals a better similarity to
the output of the DM-based SNF algorithm (top row). Moreover, the
DM-based SNF algorithm detected sub-nuclei that are geometrically
smoother than those that were detected by the PCA-based SNF algorithm.}
\end{figure}

\begin{figure}[!h]
\begin{centering}
\includegraphics[bb=80bp 320bp 530bp 460bp,clip,width=1\linewidth]{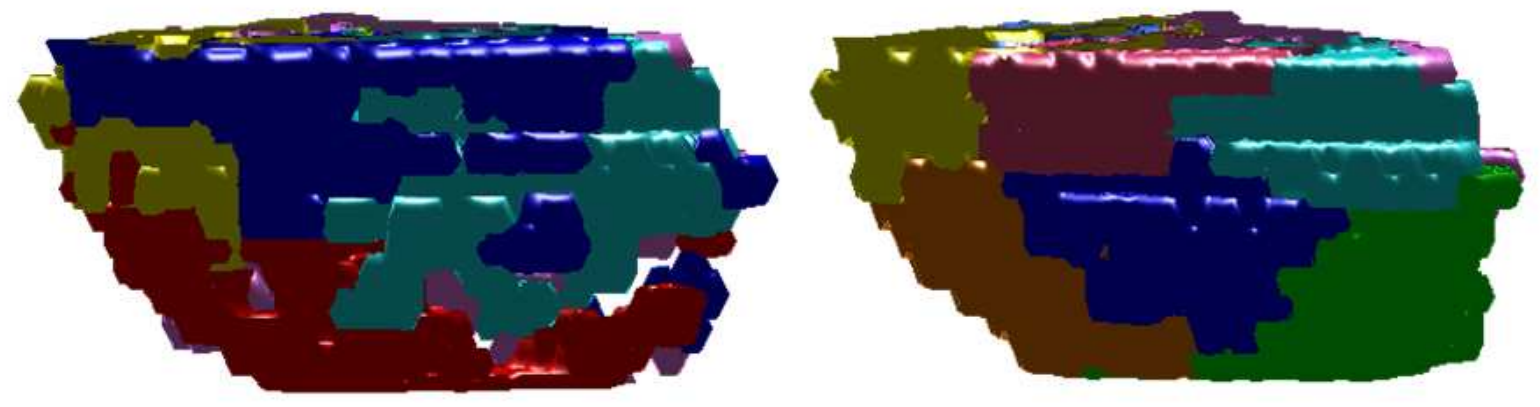}
\par\end{centering}

\caption{\label{cap:Match3DPCA}A 3D view corresponding to  Fig. \ref{cap:Slice-27-Matched-PCA}.
Left: the result of a match between the Kikinis histological atlas
\cite{Kikinis} and the output of the PCA-based SNF algorithm applied
on the right thalamus of a subject. Right: a similar angle of the
result of a match between the atlas and the output of the DM-based
SNF algorithm applied on the same data. The DM-based SNF algorithm
detected sub-nuclei that are geometrically smoother than those that
were detected by the PCA-based SNF algorithm.}
\end{figure}

We also investigated the performance of the dimensionality reduction
process we use in Algorithm \ref{cap:Algorithm-Sub-Nuclei-Segmentation},
which is based on the DM framework (Chapter \ref{cha:Diffusion-Maps}),
to the classical dimensionality reduction technique - PCA (Section
\ref{sub:Global-methods}). We used PCA instead of the DM for the
dimensionality reduction step and kept the same number of eigenvectors
in both the PCA process and the DM process. As expected, the time
complexity was reduced from 106 $seconds$ (using DM) to 75 $seconds$
(using PCA) since PCA has a lower complexity than local non-linear
methods (Section \ref{sub:Local-methods}). However, the results of
the PCA-based SNF algorithm were inferior to those of the DM-based
SNF algorithm (the original SNF algorithm). The PCA-based SNF algorithm
managed to detect only 7 clusters in the right thalamus that were
matched to 6 nuclei of the atlas, while the DM-based SNF algorithm
detected 13 clusters that were matched to 10 nuclei of the atlas.
Moreover, the mean distance error between the segmented nuclei centroids
to their matched atlas nuclei centroids was 3.53 $mm$ in the PCA-based
SNF algorithm outputs, while in the DM-based SNF algorithm outputs
this value was 2.8 $mm$ (Section \ref{sub:Matching-To-Histological-Atlas}).
Figure \ref{cap:Slice-27-Matched-PCA} illustrates this comparison.
Figure \ref{cap:Match3DPCA} shows a 3D view of the compared results.
The number of sub-nuclei, that were detected by the application of
the PCA-based SNF algorithm, is less than the number of sub-nuclei,
which were detected by the application of the DM-based SNF algorithm.
Moreover, a comparison between the geometric characteristics (shape,
size and location) of the detected nuclei (by both algorithm versions)
and their counterparts in the atlas, reveals a better similarity to
the output of the DM-based SNF algorithm. Furthermore, the DM-based
SNF algorithm detected sub-nuclei that are geometrically smoother
than those that were detected by the PCA-based SNF algorithm.

\subsection{\label{sub:Summary-of-All}Summary of results}

Tables \ref{cap:Table-Results-for-all-data-types} and \ref{cap:Table-Results-for-all-algorithm-types}
summarize the experimental results. Table \ref{cap:Table-Results-for-all-data-types}
summarizes the results of evaluating the SNF algorithm applied on
different types of data. In all cases, there was a good match between
the number of nuclei of the two examined objects: subject and atlas,
two subjects and the two parts of the thalamus. Specifically, 10 out
of the 13 nuclei were shown to match. The visual match between the
results is very high for the subject and atlas. This is corroborated
by a low matched centroids mean distance error ($mm$). The results
for the two subjects and the two part of thalamus are also good however,
they are inferior to those of the subject and atlas. Table \ref{cap:Table-Results-for-all-algorithm-types}
summarizes the results of evaluating  different versions of the SNF
algorithm applied on the same data.

\begin{table}[!h]
{\footnotesize }%
\begin{tabular}{|>{\centering}m{0.22\linewidth}|>{\centering}m{0.22\linewidth}||>{\centering}m{0.22\linewidth}|>{\centering}m{0.22\linewidth}|}
\cline{1-2} 
\textbf{\footnotesize Criterion} & \multicolumn{1}{>{\centering}m{0.22\linewidth}|}{\textbf{\footnotesize }%
\begin{tabular}{|>{\centering}m{0.9\linewidth}|}
\hline 
\multicolumn{1}{|>{\centering}m{0.9\linewidth}}{\textbf{\footnotesize subject and atlas}}\tabularnewline
\hline 
\hline 
{\footnotesize (Section \ref{sub:Matching-To-Histological-Atlas})}\tabularnewline
\hline 
\end{tabular}\textbf{\footnotesize{} }} & \textbf{\footnotesize }%
\begin{tabular}{|>{\centering}m{0.9\linewidth}}
\textbf{\footnotesize two subjects}\tabularnewline
{\footnotesize (Section \ref{sub:Matching-Between-Subjects})}\tabularnewline
\end{tabular}\textbf{\footnotesize{} } & \textbf{\footnotesize }%
\begin{tabular}{|>{\centering}p{0.9\linewidth}}
\textbf{\footnotesize two parts of the thalamus}\tabularnewline
{\footnotesize (Section \ref{sub:Matching-the-Left})}\tabularnewline
\end{tabular}\textbf{\footnotesize{} }\tabularnewline
\cline{1-2} 
{\footnotesize visual resemblance} & {\footnotesize very high} & {\footnotesize high} & \multicolumn{1}{>{\centering}m{0.22\linewidth}}{{\footnotesize medium}}\tabularnewline
{\footnotesize number of detected clusters} & {\footnotesize 13} & {\footnotesize 13} & \multicolumn{1}{>{\centering}m{0.22\linewidth}}{{\footnotesize 13}}\tabularnewline
\cline{3-4} 
{\footnotesize number of matched clusters} & {\footnotesize 10} & {\footnotesize 10} & {\footnotesize 10}\tabularnewline
\hline 
{\footnotesize matched centroids mean distance error ($mm$)} & \multicolumn{1}{>{\centering}m{0.22\linewidth}|}{{\footnotesize 2.8}} & {\footnotesize 3.55} & {\footnotesize 4.14}\tabularnewline
\end{tabular}{\footnotesize \par}

\caption{\label{cap:Table-Results-for-all-data-types}Evaluation summary of
the SNF algorithm applied on different types of data.}
\end{table}

\begin{table}[!h]
{\footnotesize }%
\begin{tabular}{|>{\centering}m{0.17\linewidth}|>{\centering}m{0.17\linewidth}||>{\centering}m{0.17\linewidth}|>{\centering}m{0.17\linewidth}|>{\centering}m{0.17\linewidth}}
\cline{1-3} 
\textbf{\footnotesize Criterion} & \multicolumn{1}{>{\centering}m{0.17\linewidth}|}{\textbf{\footnotesize }%
\begin{tabular}{|>{\centering}m{0.95\linewidth}}
\hline 
\textbf{\footnotesize SNF}\tabularnewline
\hline 
\hline 
{\footnotesize (Section \ref{sub:Matching-To-Histological-Atlas})}\tabularnewline
\hline 
\end{tabular}\textbf{\footnotesize{} }} & \multicolumn{1}{>{\centering}m{0.17\linewidth}||}{\textbf{\footnotesize }%
\begin{tabular}{|>{\centering}m{0.9\linewidth}}
\textbf{\footnotesize SNF without image enhancement}\tabularnewline
{\footnotesize (Section \ref{sub:Image-Enhancement-reduction})}\tabularnewline
\end{tabular}\textbf{\footnotesize{} }} & \textbf{\footnotesize }%
\begin{tabular}{|>{\centering}p{0.95\linewidth}|}
\multicolumn{1}{|>{\centering}p{0.95\linewidth}}{\textbf{\footnotesize SNF without dimensionality reduction}}\tabularnewline
{\footnotesize (Section \ref{sub:Dimensionality-Reduction-reduction})}\tabularnewline
\end{tabular}\textbf{\footnotesize{} } & \textbf{\footnotesize }%
\begin{tabular}{|>{\centering}m{0.95\linewidth}}
\textbf{\footnotesize PCA-based SNF}\tabularnewline
{\footnotesize (Section \ref{sub:Dimensionality-Reduction-reduction})}\tabularnewline
\end{tabular}\textbf{\footnotesize{} }\tabularnewline
\cline{1-3} 
{\footnotesize visual resemblance to atlas} & {\footnotesize very high} & {\footnotesize medium} & {\footnotesize medium} & {\footnotesize medium}\tabularnewline
{\footnotesize number of detected clusters} & {\footnotesize 13} & {\footnotesize 6} & {\footnotesize 10} & {\footnotesize 7}\tabularnewline
\cline{3-5} 
{\footnotesize number of matched clusters} & {\footnotesize 10} & {\footnotesize 5} & {\footnotesize 8} & \multicolumn{1}{>{\centering}m{0.17\linewidth}|}{{\footnotesize 6}}\tabularnewline
\hline 
{\footnotesize running time ($seconds$)} & \multicolumn{1}{>{\centering}m{0.17\linewidth}|}{{\footnotesize 106}} & {\footnotesize 76} & {\footnotesize 170} & \multicolumn{1}{>{\centering}m{0.17\linewidth}|}{{\footnotesize 75}}\tabularnewline
\end{tabular}{\footnotesize \par}

\caption{\label{cap:Table-Results-for-all-algorithm-types}Evaluation summary
of the results of different versions of the SNF algorithm applied
on the same data. All the results are according to the match between
the outputs and the histological atlas \cite{Kikinis}.}
\end{table}

\section{Conclusion and future work\label{sec:Conclusion-and-Future6}}

In this chapter, we introduced a novel algorithm (the SNF algorithm)
for automatic segmentation of neuronal tissue into its sub-nuclei.
By using an \emph{in-vivo} multi-contrast MRI data of a human brain
as an input, we achieved a good segmentation of the thalamus into
its sub-nuclei. Namely, the SNF algorithm automatically identified
the major nuclei of the thalamus. The results of the SNF algorithm
were found to strongly agree with a known histological atlas \cite{Kikinis}.
Moreover, the outputs were found to be consistent for different subjects
and for the two parts of the thalamus of the same subject%
\footnote{in both cases up to inner-subject variability.%
}. % The results of the SNF algorithm can be improved by upgrading the 2D wavelet-based image enhancement into a 3D wavelet-based image enhancement that would emphasize features that exist in the 3D space. In addition, a kernel that improves the results in the DM and DB procedures should be investigated. 

The results of the algorithm can be improved by a better selection
of the input data. For instance, one should perform an \emph{in-vitro}%
\footnote{Since the desired acquisition time is too long to perform in-\emph{vivo}
and to allow a histological examination.%
} experiment, that includes acquisition of a large number of different
contrast acquisition-methods in higher resolution. Then, after the
application of the SNF algorithm on the contrast acquisition-methods,
the spectra of the clusters should be examined to identify a reasonable
number of contrast acquisition-methods, that give the best separation,
when combined together. Furthermore, an \emph{in-vitro} histological-based
sub-nuclei segmentation of the examined sample should be performed,
in order to validate the results. Then, the identified contrast acquisition-methods
should be used \emph{in-vivo}. The described technique should produce
a more accurate segmentation (separation into smaller nuclei) than
the segmentation by histological methods.

%The SNF algorithm, that was introduced in this work, is not restricted only to MRI-based data and it can be used on a variety of multi-dimensional inputs (after several adjustments), e.g. hyper-spectral data.

\chapter{Video Segmentation via Diffusion Bases\label{cha:Video-segmentation-via}}

Identifying moving objects in a video sequence, which is produced
by a static camera, is a fundamental and critical task in many computer-vision
applications. A common approach performs background subtraction and
thus identifies moving objects as the portion of a video frame that
differs significantly from a background model. A good background subtraction
algorithm has to be robust to changes in the illumination and it should
avoid detecting non-stationary background objects such as moving leaves,
rain, snow, and shadows. In addition, the internal background model
should quickly respond to changes in background such as objects that
start to move or stop.

In this chapter, we present a new algorithm for video segmentation
that processes the input video sequence as a 3D matrix where the third
axis is the time domain. Our approach identifies the background by
reducing the input dimension using the \emph{diffusion bases} methodology.
Furthermore, we describe an iterative method for extracting and deleting
the background. The algorithm has two versions and thus covers the
complete range of backgrounds: one for scenes with static backgrounds
and the other for scenes with dynamic (moving) backgrounds.

\section{Introduction \label{sub:introduction}}

Video surveillance systems, tracking systems, imaging-based statistical
packages that count people, games, etc. seek to automatically identify
people, objects, or events of interest in different environment types.
Typically, these systems consist of stationary cameras, that are directed
at offices, parking lots, playgrounds, fences and so on, together
with computer systems that process the video frames. Human operators
or other processing elements are notified about salient events. There
are many needs for automated surveillance systems in commercial, law
enforcement, and military applications. One possible application is
a continuous 24-hour monitoring of surveillance video to alert security
officers of a burglary in progress or a suspicious individual loitering
in a parking lot. 

In addition to the obvious security applications, video surveillance
technology has been proposed to measure traffic flow, detect accidents
on highways, monitor pedestrian congestion in public spaces, compile
consumer demographics in shopping malls and amusement parks, log routine
maintenance tasks at nuclear facilities, and count endangered species.
The numerous military applications include patrolling national borders,
measuring the flow of refugees in troubled areas, monitoring peace
treaties, and providing secure perimeters around bases.

A common element in surveillance systems is a module that performs
background subtraction to distinguish between background pixels, which
should be ignored, and foreground pixels, which should be processed
for identification or tracking. The difficulty in background subtraction
is not to differentiate, but to maintain the background model, its
representation and its associated statistics. In particular, capturing
the background in frames where the background can change over time.
These changes can be moving trees, leaves, water flowing, sprinklers,
fountains, video screens (billboards) just to name a few typical examples.
Other forms of changes are weather changes like rain and snow, illumination
changes like turning on and off the light in a room and changes in
daylight. We refer to this background type as \emph{dynamic background}
(DBG) while a background that slightly changes or does not change
at all is referred to as \emph{static background} (SBG).

Subtraction of backgrounds, which are captured by static cameras,
is also useful to achieve low-bit rate video compression for transmission
of rich multimedia content. The subtracted background is transmitted
once, followed by the segmented objects which are detected.

In this chapter, we present a new method for capturing the background.
It is based on the application of the \emph{diffusion bases} (DB)
algorithm (Chapter \ref{cha:Diffusion-bases}). Moreover, we introduce
a real time iterative method for background subtraction in order to
separate between background and foreground pixels while overcoming
the presence of changes in the background. The main steps of the algorithm
are:
\begin{itemize}
\item Extract the background frame by dimensionality reduction via the application
of the DB algorithm. 
\item Subtract the background from the input sequence. 
\item Threshold the subtracted sequence. 
\item Detect the foreground objects by applying %for example
\emph{depth first search} (DFS) on a graph model of the background-subtracted
sequence. 
\end{itemize}
We propose two versions of the algorithm - one for static background
and the other for dynamic background. To handle dynamic background,
a learning process is applied to data that contains only the background
objects in order to construct a frame that will contain the DBG to
be extracted. The proposed algorithm outperform current state-of-the-art
algorithms.

The rest of this chapter is organized as follows: in Section \ref{sub:relatedWork},
related algorithms for background subtraction are presented. The main
algorithm, that is called the \emph{background subtraction algorithm
using diffusion bases} (BSDB), is presented in Section \ref{BSDB}.
In Section \ref{EXPERIMENTAL}, we present experimental results, a
performance analysis of the BSDB algorithm and we compare it to other
background subtraction algorithms.

\section{Related work \label{sub:relatedWork}}

Background subtraction is a widely used approach for detection of
moving objects in video sequences that are captured by static cameras.
This approach detects moving objects by differentiating between the
current frame and a reference frame, often called the background frame,
or background model. In order to extract the objects of interest,
a threshold can be applied to the subtracted frame. The background
frame should faithfully represent the scene. It should not contain
moving objects. In addition, it must be regularly updated in order
to adapt to varying conditions such as illumination and geometry changes.
This section provides a review of the current state-of-the-art background
subtraction techniques. These techniques range from simple approaches,
aiming to maximize speed and minimizing the memory requirements, to
more sophisticated approaches, aiming to achieve the highest possible
accuracy under any possible circumstances. The goal of these approaches
is to run in real-time. Additional references can be found in \cite{CLK00,P04,M00}.
\begin{itemize}
\item \textbf{Temporal median filter:} In \cite{LV01}, is was proposed
to use the median value of the last $n$ frames as the background
model. This provides an adequate background model even if the $n$
frames are subsampled with respect to the original frame rate by a
factor of 10 \cite{CGPP03}. The median filter is computed on a special
set of values that contains the last $n$ subsampled frames and the
last computed median value. This combination increases the stability
of the background model \cite{CGPP03}. \\
The main disadvantage of a median-based approach is that its computation
requires a buffer with the recent pixel values. Furthermore, no deviation
measure was provided with which the subtraction threshold can be adapted.
\item \textbf{Gaussian average:} This approach models the background independently
at each pixel location $(i,j)$ \cite{WADP97}. The model is based
on ideally fitting a Gaussian probability density function (pdf) to
the last $n$ pixels. At each new frame at time $t$, a running average
is computed by $\psi_{t}=\alpha I_{t}+(1-\alpha)\psi_{t-1}$ where
$I_{t}$ is the current frame, $\psi_{t-1}$ is the previous average
and $\alpha$ is an empirical weight that is often chosen as a trade-off
between stability and quick update.\\
In addition to speed, the advantage of the running average is given
by a low memory requirement. Instead of a buffer with the last $n$
pixel values, each pixel is classified using two parameters $(\psi_{t},\sigma_{t})$,
where $\sigma_{t}$ is the standard deviation. Let $p_{i,j}^{t}$
be the $(i,j)$ pixel at time $t$. $p_{i,j}^{t}$ is classified as
a foreground pixel if $|p_{i,j}^{t}-\psi_{t-1}|>k\sigma_{t}$. Otherwise
$p_{i,j}^{t}$ is classified as background pixel.
\item \textbf{Mixture of Gaussians}: In order to cope with rapid changes
in the background, a multi-valued background mode was suggested in
\cite{SG99}. In this model, the probability of observing a certain
pixel $x$ at time $t$ is represented by a mixture of $k$ Gaussians
distributions: $P(x_{t})=\Sigma_{i=1}^{k}w_{i,t}\eta(x_{t},\psi_{i,t},\Sigma_{i,t})$
where for each $i$-th Gaussian in the mixture at time $t$, $w$
estimates what portion of the data is accounted for by this Gaussian,
$\psi$ is the mean value, $\Sigma$ is the covariance matrix and
$\eta$ is a Gaussian probability density function. In practice, $k$
is set to be between 3 and 5. %Figure  illustrated this model.\\
Each of the $k$ Gaussian distributions describes only one of the
observable background or foreground objects. The distributions are
ranked according to the ratio between their peak amplitude $w_{i}$
and their standard deviation $\sigma_{i}$. Let $Th$ be the threshold
value. The first $B$ distributions that satisfy $\Sigma_{i=1}^{B}w_{i}>Th$
are accepted as background. All the other distributions are considered
as foreground.\\
Let $I_{t}$ be a frame at time $t$. At each frame $I_{t}$, two
events take place simultaneously: assigning the new observed value
$x_{t}$ to the best matching distribution and estimating the updated
model parameters. The distributions are ranked and the first that
satisfies $(x_{t}-\psi_{i,t})/\sigma_{i,t}>2.5$ is a match for $x_{t}$.
\item \textbf{Kernel density estimation (KDE)}: This approach models the
background distribution by a non-parametric model that is based on
a Kernel Density Estimation (KDE) of the buffer of the last \emph{n}
background values (\cite{EHD00}). KDE guarantees a smooth, continuous
version of the histogram of the most recent values that are classified
as background values. This histogram is used to approximate the background
pdf.\\
The background pdf is given as a sum of Gaussian kernels centered
at the most recent $n$ background values, $x_{t}$: $P(x_{t})=\frac{1}{n}\Sigma_{i=1}^{n}\eta(x_{t}-x_{i},\Sigma_{t})$
where $\eta$ is the kernel estimator function and $\Sigma_{t}$ represents
the kernel function bandwidth. $\Sigma$ is estimated by computing
the median absolute deviation over the sample for consecutive intensity
values of the pixel. Each Gaussian describes just one sample data.
The buffer of the background values is selectively updated in a FIFO
order for each new frame $I_{t}$.\\
In this application two similar models are concurrently used, one
for long-term memory and the other for short-term memory. The long-term
model is updated using a \emph{blind} update mechanism that prevents
incorrect classification of background pixels.
\item \textbf{Sequential kernel density approximation}: Mean-shift vector
techniques have been employed for various pattern recognition problems
such as image segmentation and tracking (\cite{C03,CM02}). The mean-shift
vector is an effective technique capable of directly detecting the
main modes of the pdf from the sample data using a minimum set of
assumptions. However, it has a very high computational cost since
it is an iterative technique and it requires a study of the convergence
over the whole data space. As such, it is not immediately applicable
to modeling background pdfs at the pixel level.\\
To solve this problem, computational optimizations are used to mitigate
the computational high cost (\cite{PJ04}). Moreover, the mean-shift
vector can be used only for an off-line model initialization \cite{JCD04},
i.e. the initial set of Gaussian modes of the background pdf is detected
from an initial sample set. The real-time model is updated by simple
heuristics that handle mode adaptation, creations, and merging.
\item \textbf{Co-occurrence of image variations}: This method exploits spatial
co-occurrences of image variations (\cite{SWFS03}). It assumes that
neighboring blocks of pixels that belong to the background should
have similar variations over time. The disadvantage of this method
is that it does not handle blocks at the borders of distinct background
objects.\\
This method divides each frame to distinct blocks of $N\times N$
pixels where each block is regarded as an $N^{2}$-component vector.
This trades-off resolution with high speed and better stability. During
the learning phase, a certain number of samples is acquired at a set
of points, for each block. The temporal average is computed and the
differences between the samples and the average, called the \emph{image
variations}, is calculated. Then the $N^{2}\times N^{2}$ covariance
matrix is computed with respect to the average. An eigenvector transformation
is applied to reduce the dimensions of the image variations.\\
For each block $b$, a classification phase is performed: the corresponding
current eigen-\emph{image-variations} are computed on a neighboring
block of $b$. Then the image variation is expressed as a linear interpolation
of its L-nearest neighbors in the eigenspace. The same interpolation
coefficients are applied on the values of $b$, to provide an estimate
for its current eigen-\emph{image-variations}.
\item \textbf{Eigen-backgrounds}: This approach is based on an eigen-decomposition
of the whole image \cite{ORP00}. During a learning phase, samples
of $n$ images are acquired. The average image is computed and subtracted
from all the images. The covariance matrix is computed and the best
eigenvectors are stored in an eigenvector matrix. For each frame $I$,
a classification phase is executed: $I$ is projected onto the eigenspace
and then projected back onto the image space. The output is the background
frame, which does not contain any small moving objects. A threshold
is applied on the difference between $I$ and the background frame. 
\end{itemize}

\section{The Background Subtraction Algorithm using Diffusion Bases (BSDB)}

\label{BSDB}

In this section we present the BSDB algorithm. The algorithm has two
versions:

\paragraph*{Static background subtraction using DB (SBSDB) }

We assume that the background is static (SBG) -- see Section \ref{SBG}.
The video sequence is captured on-line. Gray level images are sufficient
for the processing.

\paragraph*{Dynamic background subtraction using DB (DBSDB) }

We assume that the background is moving (DBG) -- see Section \ref{DBG}.
This algorithm uses off-line (training) and on-line (detection) procedures.
As opposed to the SBSDB, this algorithm requires color (RGB) frames. 

Both algorithms assume that the camera is static.

\subsection{Static background subtraction algorithm using DB (SBSDB)\label{SBG}}

In this section we describe the on-line algorithm that is applied
on a video sequence that is captured by a static camera. We assume
that the background is static. The SBSDB algorithm captures the static
background, subtracts it from the video sequence and segments the
subtracted output.

The input to the algorithm is a sequence of video frames in gray-level
format. The algorithm produces a binary mask for each video frame.
The pixels in the binary mask that belong to the background are assigned
0 values while the other pixels are assigned to be 1.

%\paragraph*{Off-line algorithm for capturing static background\label{sub:findSteadyBg}}

\subsubsection{Off-line algorithm for capturing static background\label{sub:findSteadyBg}}

In order to capture the static background of a scene, we reduce the
dimensionality of the input sequence by applying the DB algorithm
(see Chapter \ref{cha:Diffusion-bases}). The input to the algorithm
consists of $n$ frames that form a datacube.

Formally, let 
\begin{equation}
D_{n}=\left\{ s_{i,j}^{t}\right\} _{i,j=1,\ldots,N;\, t=1,\ldots,n}\label{eq:D_n}
\end{equation}
be the input datacube of $n$ frames each of size $N\times N$ where
$s_{i,j}^{t}$ is the pixel at position $(i,j)$ in the video frame
at time $t$. We define the vector $P_{i,j}\triangleq\left(s_{i,j}^{1},\ldots,s_{i,j}^{n}\right)$
to be the values of the $(i,j)^{th}$ coordinate at all the $n$ frames
in $D_{n}$. This vector resembles a \emph{hyper}-\emph{pixel} (see
Section \ref{sec:Introduction6}) and will be named so from this point
on. Let $\Omega_{n}=\left\{ P_{i,j}\right\} _{i,j=1,\ldots,N}$ be
the set of all hyper-pixels. We define $F_{t}\triangleq(s_{1,1}^{t},\ldots,s_{N,N}^{t})$
to be a 1-D vector representing the video frame at time $t$. We refer
to $F_{t}$ as a frame-vector. Let $\Omega'_{n}\triangleq\left\{ F_{t}\right\} _{t=1}^{n}$
be the set of all frame-vectors.

We apply the DB algorithm to $\Omega_{n}$ by $\Omega_{BS}$=\textbf{DiffusionBasis}($\Omega'_{n}$,
$w_{\varepsilon}$, $\varepsilon$, $\eta$) where $w_{\varepsilon}$
is defined by Eq. \ref{eq:gaussian}, the \textbf{DiffusionBasis}
procedure was defined in algorithm \ref{alg:Diffusion-Basis-Calculation}
and $\varepsilon$, $\eta$ are defined in Section \ref{sub:Building-the-graph}
and \ref{sub:Spectral-decomposition}, respectively (see also Section
\ref{sec:Choosing_epsilon} for an algorithm to calculate $\varepsilon$).
The output is the projection of every hyper-pixel on the diffusion
basis which embeds the original data $D_{n}$ into a reduced space.
The first vector of $\Omega_{BS}$ represents the background of the
input frames. Let $bg_{V}=\left(x_{i}\right)$, $i=1,\ldots,N^{2}$
be this vector. We reshape $bg_{V}$ into the $N\times N$ matrix
$bg_{M}=\left(x_{ij}\right)$. Then, $bg_{M}$ is normalized to be
between 0 to 255. The normalized background is denoted by $\widehat{bg}_{M}$.

%\paragraph*{On-line algorithm for capturing a static background\label{sub:findBgRT} }

\subsubsection{On-line algorithm for capturing a static background\label{sub:findBgRT} }

In order to make the algorithm suitable for on-line applications,
the incoming video sequence is processed by using a \emph{sliding
window} (SW) of size $m$. Thus, the number of frames that are input
to the algorithm is $m$. Naturally, we seek to minimize $m$ in order
to obtain a faster result from the algorithm. We found empirically
that the algorithm produces good results for values of $m$ as low
as $m=5,6$ and $7$. The delay of 5 to 7 frames is negligible and
renders the algorithm suitable for on-line applications.

\begin{figure}[!h]
\begin{centering}
\includegraphics[bb=0bp 0bp 512bp 108bp,width=1\columnwidth]{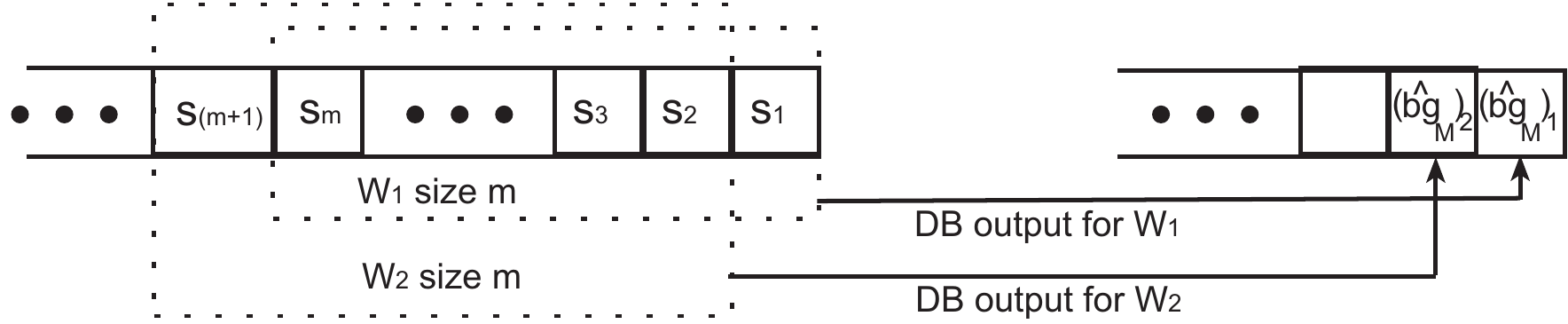} 
\par\end{centering}

\caption{Illustration of how the SW is shifted right to left. $W_{1}=\left(s_{1},\ldots,s_{m}\right)$
is the SW for $s_{1}$. $W_{2}=\left(s_{2},\ldots,s_{m+1}\right)$
is the SW for $s_{2}$, etc. The backgrounds of $s_{i}$ and $s_{i+1}$
are denoted by $(\widehat{bg}_{M})_{i}$ and $(\widehat{bg}_{M})_{i+1}$,
$i=1,\ldots,n-m+1$, respectively.}

\label{fig:slidingWindow}
\end{figure}

Let $S=\left(s_{1},\ldots,s_{i},\ldots,s_{m},s_{m+1},\ldots,s_{n}\right)$
be the input video sequence. we apply the algorithm that is described
in Section \ref{sub:findSteadyBg} to every SW. The output is a sequence
of background frames 
\begin{equation}
\widehat{BG}=\left((\widehat{bg}_{M})_{1},\ldots,(\widehat{bg}_{M})_{i},\ldots,(\widehat{bg}_{M})_{m},(\widehat{bg}_{M})_{m+1},\ldots,(\widehat{bg}_{M})_{n}\right)\label{eq:BG_hat}
\end{equation}
where $(\widehat{bg}_{M})_{i}$ is the background that corresponds
to frame $s_{i}$ and $(\widehat{bg}_{M})_{n-m+2}$ till $(\widehat{bg}_{M})_{n}$
are set to $(\widehat{bg}_{M})_{n-m+1}$. Figure \ref{fig:slidingWindow}
describes how the SW is shifted.

The SW results in a faster execution time of the DB algorithm. The
weight function $w_{\varepsilon}$ (Eq. \ref{eq:gaussian}) is not
recalculated for all the frame in the SW. Instead, $w_{\varepsilon}$
is only updated according to the new frame that enters the SW and
the one that exits the SW. Specifically, let $W_{t}=\left(s_{t},\ldots,s_{t+m-1}\right)$
be the SW at time $t$ and let $W_{t+1}=\left(s_{t+1},\ldots,s_{t+m}\right)$
be the SW at time $t+1$. At time $t+1$, $w_{\varepsilon}$ is calculated
only for $s_{t+m}$ and the entries that correspond to $s_{t}$ are
removed from $w_{\varepsilon}$.\\

%\paragraph*{The SBSDB algorithm\label{sub:algSteadyBg}}

\subsubsection{The SBSDB algorithm\label{sub:algSteadyBg}}

The SBSDB on-line algorithm captures the background of each SW according
to Section \ref{sub:findBgRT}. Then it subtracts the background from
the input sequence and thresholds the output to get the background
binary mask.

Let $S=\left(s_{1},\ldots,s_{n}\right)$ be the input sequence. For
each frame $s_{i}\in S$, $i=1,\ldots,n$, we do the following: 
\begin{itemize}
\item Let $W_{i}=\left(s_{i},\ldots,s_{i+m-1}\right)$ be the SW of $s_{i}$.
The on-line algorithm for capturing the background (Section \ref{sub:findBgRT})
is applied to $W_{i}$. The output is the background frame $(\widehat{bg}_{M})_{i}$. 
\item The SBSDB algorithm subtracts $(\widehat{bg}_{M})_{i}$ from the original
input frame to produce $\bar{s}_{i}=s_{i}-(\widehat{bg}_{M})_{i}$.
Each pixel in $\bar{s}_{i}$ that has a negative value is set to 0. 
\item A threshold, which is computed in Section \ref{sub:GrayThreshold},
is applied to $\bar{s}_{i}$. For $k,l=1,\ldots,N$ the output is
defined as follows: 
\[
\tilde{s}_{i}(k,l)=\left\{ \begin{array}{ll}
0, & if\, it\, is\, a\, background\, pixel;\\
1, & otherwise.
\end{array}\right.
\]
 
\end{itemize}
%\paragraph*{Threshold computation for a grayscale input\label{sub:GrayThreshold}}

\subsubsection{Threshold computation for a grayscale input\label{sub:GrayThreshold}}

The threshold $Th$, which separates between background and foreground
pixels, is calculated in the last step of the SBSDB algorithm. The
SBSDB algorithm subtracts the background from the input frame and
sets pixels with negative values to zero. Furthermore, a linear low-pass
filter is applied to the histogram in order to smooth it so that the
threshold value could be accurately computed and not influenced by
local variations that are due to noise.

Usually, the number of background pixels is larger than the the number
of the foreground ones. After the subtraction, the values of the background
pixels are close to zero. Thus, the histogram of a frame after subtraction
will be high at small values (background) and low at high values (foreground).
In order to separate between the background and foreground, a point
between the peak and the low point of the histogram is sought after.
If the slope is too high, that means that we are \emph{leaving} in
the background area of the histogram. If the slope is too low, that
mean we are in the foreground area of the histogram. Consequently,
a good point of separation is where the slope of the histogram becomes
moderate. 

Let $h$ be the histogram of a frame and let $\mu$ be a given parameter
which provides a threshold for the slope of $h$. $\mu$ is chosen
to be the magnitude of the slope where $h$ becomes moderate. This
point separates between We scan $h$ from its global maximum to the
right. We set the threshold $Th$ to be the smallest value of $x$
that satisfies $\left|h'\left(x\right)\right|<\mu$ where $h'_{x}$
is the first derivative of $h$ at point $x$, i.e. the slope of $h$
at point $x$. The background/foreground classification of the pixels
in the input frame $\bar{s}_{i}$ is determined according to $Th$.
Specifically, for $k,l=1,\ldots,N$ 
\[
\tilde{s}_{i}(k,l)=\left\{ \begin{array}{cc}
0, & if\,\bar{s}_{i}(k,l)<Th;\\
1, & otherwise.
\end{array}\right.
\]

\begin{figure}[!h]
\begin{centering}
\includegraphics[bb=0bp 0bp 575bp 500bp,clip,width=0.4\columnwidth]{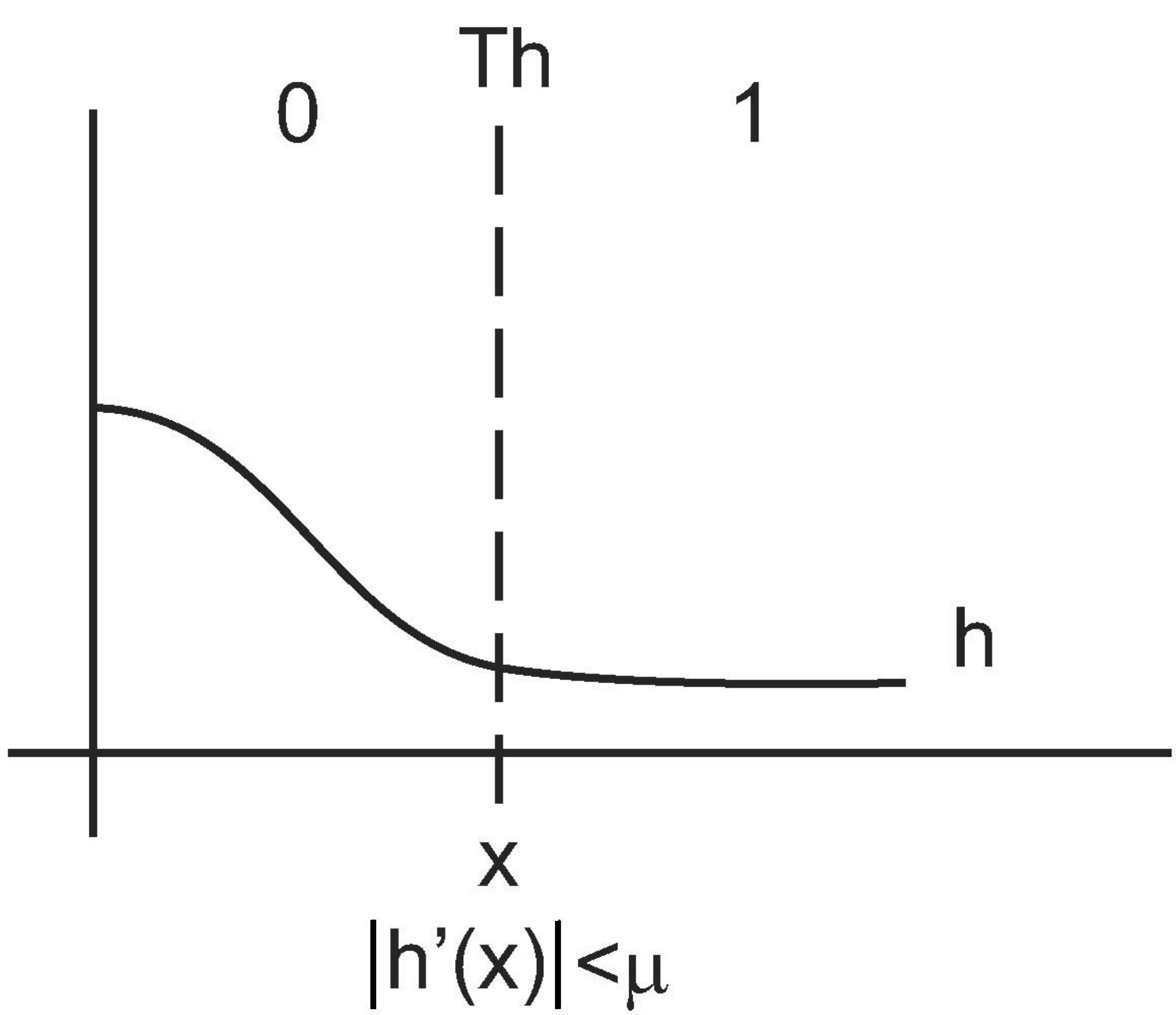} 
\par\end{centering}

\caption{An example how to use the histogram $h$ for finding the threshold
value. $Th$ is set to to the smallest $x$ for which $\left|h'(x)\right|<\mu$.}

\label{fig:hist_gray} 
\end{figure}

Fig. \ref{fig:hist_gray} illustrates how to find the threshold.

%-----------------------------------

\subsection{Dynamic background subtraction algorithm using DB (DBSDB)\label{DBG}}

%The DBSDB is an two-step on-line algorithm that is used for video
%sequences in which the background is dynamic (moving). The first
%step consists of an offline training procedure that captures the
%dynamic background of the scene from a video sequence that does not
%contain foreground objects. The second step captures the dynamic
%background, subtracts it from the and the background that was
%captured during the first step from applies an on-line background
%subtraction
In this section, we describe an on-line algorithm that handles video
sequences that are captured by a static camera. We assume that the
background is dynamic (moving). The DBSDB applies an off-line procedure
that captures the dynamic background and an on-line background subtraction
algorithm. In addition, the DBSDB algorithm segments the video sequence
after the background subtraction is completed.

\begin{figure}[!h]
\begin{centering}
\includegraphics[bb=0bp 270bp 595bp 580bp,clip,width=0.7\columnwidth]{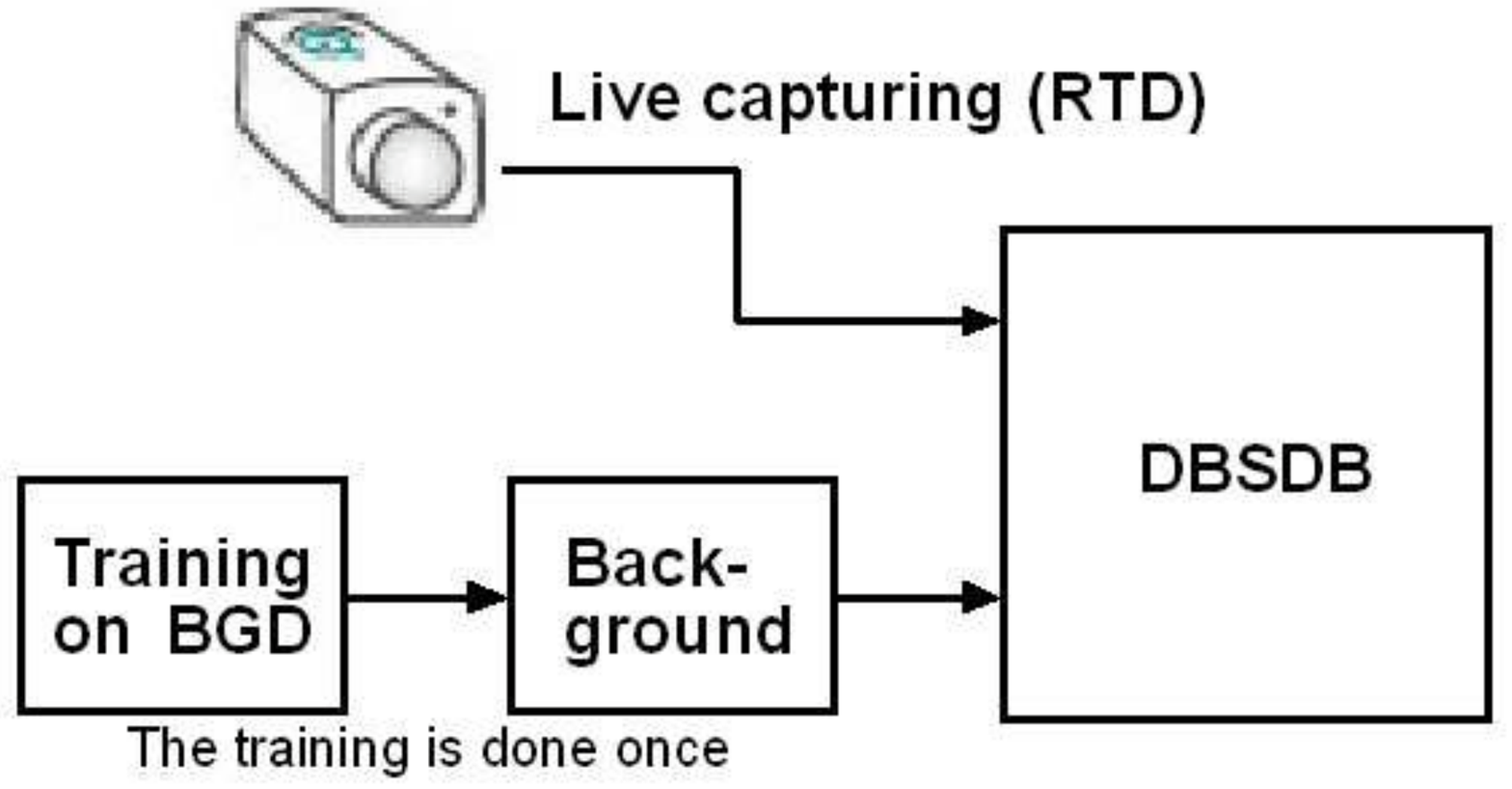} 
\par\end{centering}

\caption{The inputs to the DBSDB algorithm. The training is done once on the
BGD. It produces the background which is input to the DBSDB. The RTD
is the on-line input to the DBSDB.}

\label{fig:input} 
\end{figure}

The input to the algorithm consists of two components: 
\begin{itemize}
\item \textbf{Background training data}: A video sequence of the scene without
foreground objects. This training data can be obtained for instance
from the frames in the beginning of the video sequence. This sequence
is referred to as the \emph{background data} (BGD). %A small
%number of frames is sufficient.

\item \textbf{Data for classification}: A video sequence that contains background
and foreground objects. The classification of the objects is performed
on-line. We refer to this sequence as the \emph{real-time data} (RTD). 
\end{itemize}
For both input components, the video frames are assumed to be in RGB
- see Fig. \ref{fig:input}.

The algorithm is applied to every video frame and a binary mask is
constructed in which the pixels that belong to the background are
set to 0 while the foreground pixels are set to 1.

%\paragraph*{Iterative method for capturing a dynamic background: training \label{sub:captureDynamicBg}}

\subsubsection{Iterative method for capturing a dynamic background - training \label{sub:captureDynamicBg}}

The algorithm that is described in Section \ref{sub:algSteadyBg},
does not handle well on-going changes in the background, such as illumination
differences between frames, moving leaves, water flowing, etc. In
the following, we present a method that is not affected by background
changes. 

An iterative procedure is applied on the BGD in order to capture the
movements in the scene. This procedure constitutes the training step
of the algorithm. Let $B=\left(b_{1},\ldots,b_{m}\right)$ be the
BGD input sequence and let $bg_{M}^{final}$ be the output background
frame. $bg_{M}^{final}$ is initialized to zeros. Each iteration contains
the following steps: 
\begin{itemize}
\item Application of the off-line algorithm (Section \ref{sub:findSteadyBg})
in order to capture the static background of $B$. The BGD is treated
as a single sliding window of length $m$. The output consists of
the background frames $bg_{M}$ and $\widehat{bg}_{M}$ where $\widehat{bg}_{M}$
is the normalization of $bg_{M}$. 
\item $\widehat{bg}_{M}$ is subtracted from each frame in $B$ by $\bar{b}_{j}=b_{j}-\widehat{bg}_{M}$,
$j=1,\ldots,m$. In case the input is in grayscale format, we set
to zero each pixel in $\bar{b}_{j}$ that has a negative value. The
output is the sequence $\bar{B}=\left(\bar{b}_{1},\ldots,\bar{b}_{m}\right)$. 
\item $bg_{M}$ is added to $bg_{M}^{final}$ by $bg_{M}^{final}=bg_{M}^{final}+bg_{M}$. 
\item $\bar{B}$ is the input for the next iteration, $B=\bar{B}$. 
\end{itemize}
The iterative process stops when a given number of pixels in $B$
are equal to or smaller than zero. Finally, $bg_{M}^{final}$ is normalized
to be between 0 to 255. The normalized background is denoted by $\widehat{bg}_{M}^{final}$.
The output of this process is composed of $bg_{M}^{final}$ and $\widehat{bg}_{M}^{final}$.

%\paragraph*{The DBSDB algorithm \label{sub:algDynamicBg}}

\subsubsection{The DBSDB algorithm \label{sub:algDynamicBg}}

In this section, we describe the DBSDB algorithm which handles video
sequences that contain a dynamic background. The DBSDB algorithm consists
of five steps. The first and second steps are training steps which
capture the background of the BGD by processing the RGB sequence and
its corresponding gray-scale sequence, respectively. The third and
fourth steps classify to foreground and background the RTD using the
result of the first and second steps. The final step combines the
output from the third and fourth steps.

Formally, let $S^{rgb}=\left(s_{1}^{rgb},\ldots,s_{n}^{rgb}\right)$
and $B^{rgb}=\left(b_{1}^{rgb},\ldots,b_{m}^{rgb}\right)$ be the
RTD, which is the on-line captured video sequence, and the BGD, which
is the off-line video sequence for the training step (Section \ref{sub:captureDynamicBg}),
respectively.\\

The DBSDB algorithm consists of the following:

%\begin{itemize}

\paragraph{Step 1: The grayscale training }
\begin{itemize}
\item Convert $B^{rgb}$ into grayscale format. The grayscale sequence is
denoted by $B^{g}$. 
\item Apply the SBSDB algorithm to $B^{g}$ as was done in Section \ref{sub:algSteadyBg},
\emph{excluding} the threshold computation (Section \ref{sub:GrayThreshold}).
The output is a sequence of background frames $\bar{B}^{g}$. 
\item Capture the \emph{dynamic background} (DBG) in $\bar{B}^{g}$ (Section
\ref{sub:captureDynamicBg}). The output is the background frame given
by $\left(\widehat{bg}_{M}^{final}\right)^{g}$. 
\end{itemize}

\paragraph{\noindent Step 2: The RGB training }

\noindent Capture the color DBG by applying the algorithm from Section
\ref{sub:captureDynamicBg} to \emph{each} of the RGB channels of
$B^{rgb}$ (Section \ref{sub:captureDynamicBg}). The result forms
the color background frame which is denoted by $\left(\widehat{bg}_{M}^{final}\right)^{rgb}$.

\paragraph{Step 3: The grayscale classification }

\noindent $S^{rgb}$ is converted into grayscale format. The grayscale
sequence is denoted by $S^{g}$. The SBSDB algorithm is applied to
$S^{g}$ as it is described in Section \ref{sub:algSteadyBg}\emph{,
excluding} the threshold computation (Section \ref{sub:GrayThreshold}).
% This process is performed once or, in some cases, iteratively twice.
The output is denoted by $\bar{S}^{g}$.

For each frame $\bar{s}_{i}^{g}\in\bar{S}^{g}$, $i=1,...,n$, we
do the following: 
\begin{itemize}
\item $\widehat{(bg}_{M}^{final})^{g}$ is subtracted from $\bar{s}_{i}^{g}$
by $\tilde{s}_{i}^{g}=\bar{s}_{i}^{g}-\left(\widehat{bg}_{M}^{final}\right)^{g}$.
Then, each pixel in $\tilde{s}_{i}^{g}$ that has a negative value
is set to 0. 
\item A threshold is applied to $\tilde{s}_{i}^{g}$. The threshold is computed
as in Section \ref{sub:GrayThreshold}. The output is set to: 
\[
\breve{s}_{i}^{g}(k,l)=\left\{ \begin{array}{ll}
0, & if\, it\, is\, a\, background\, pixel;\\
1, & otherwise
\end{array}\right.
\]
 for $k,l=1,...,N$. 
\end{itemize}

\paragraph{Step 4: The RGB classification }

\noindent For each frame $s_{i}^{rgb}\in S^{rgb}$, $i=1,...,n$,
we do the following: 
\begin{itemize}
\item $\left(\widehat{bg}_{M}^{final}\right)^{rgb}$ is subtracted from
$s_{i}^{rgb}$ by $\bar{s}_{i}^{rgb}=s_{i}^{rgb}-\left(\widehat{bg}_{M}^{final}\right)^{rgb}$. 
\item $\bar{s}_{i}^{rgb}$ is normalized to be between 0 to 255. The normalized
frame is denoted by $\tilde{s}_{i}^{rgb}$. 
\item A threshold is applied to every channel in $\tilde{s}_{i}^{rgb}$.
The derivation of the threshold is described in Section \ref{sub:RGBThreshold}.
The output is set to: 
\[
\breve{s}_{i}^{rgb}(k,l)=\left\{ \begin{array}{ll}
0, & if\, it\, is\, a\, background\, pixel;\\
1, & otherwise
\end{array}\right.
\]
 for $k,l=1,\ldots,N$. 
\end{itemize}

\paragraph{Step 5: The DFS step}

\noindent This step combines the $\breve{s}_{i}^{g}$ and $\breve{s}_{i}^{rgb}$
from the grayscale and RGB classification steps, respectively. Since
$\breve{s}_{i}^{g}$ contains false negative detections (not all the
foreground objects are found) and $\breve{s}_{i}^{rgb}$ contains
false positive detections (background pixels are classified as foreground
pixels). Therefore, we use each foreground pixel in $\breve{s}_{i}^{g}$
as a reference point from which we begin the application of a DFS
on $\breve{s}_{i}^{rgb}$. This step is described in details in Section
\ref{sub:dfs}. 

%\end{itemize}

\subsubsection{Threshold computation for RGB input\label{sub:RGBThreshold}}

In the following, we describe how the threshold for the RGB classification
step (step 4) is derived. The last step of the DBSDB algorithm derives
thresholds that are applied in order to separate between background
pixels and foreground pixels in each of the RGB components. The DBSDB
algorithm subtracts the background from the input frame. Consequently,
the histogram of a frame after the subtraction is high in the center
and low at the right and left ends, where the center area corresponds
to the background pixels. The DBSDB algorithm smooths the histogram
in order to compute the threshold values accurately.

Let $h$ be the histogram and let $\mu$ be a given parameter which
provides a threshold for the slope of $h$. $\mu$ should be chosen
to be the absolute value of the slope where $h$ becomes moderate.
We denote the thresholds to be $Th^{r}$ and $Th^{l}$. We scan $h$
from its global maximum to the left. $Th^{l}=y$ if $y$ is the first
coordinate that satisfies $h'(y)<\mu$ where $h'(y)$ denotes the
first derivative of $h$ at point $y$, i.e. the slope of $h$ at
point $y$. We also scan $h$ from its global maximum to the right.
$Th^{r}=x$ if $x$ is the first coordinate that satisfies $h'(x)>-\mu$.

\begin{figure}[!h]
\begin{centering}
\includegraphics[bb=104bp 288bp 488bp 550bp,clip,width=0.5\columnwidth]{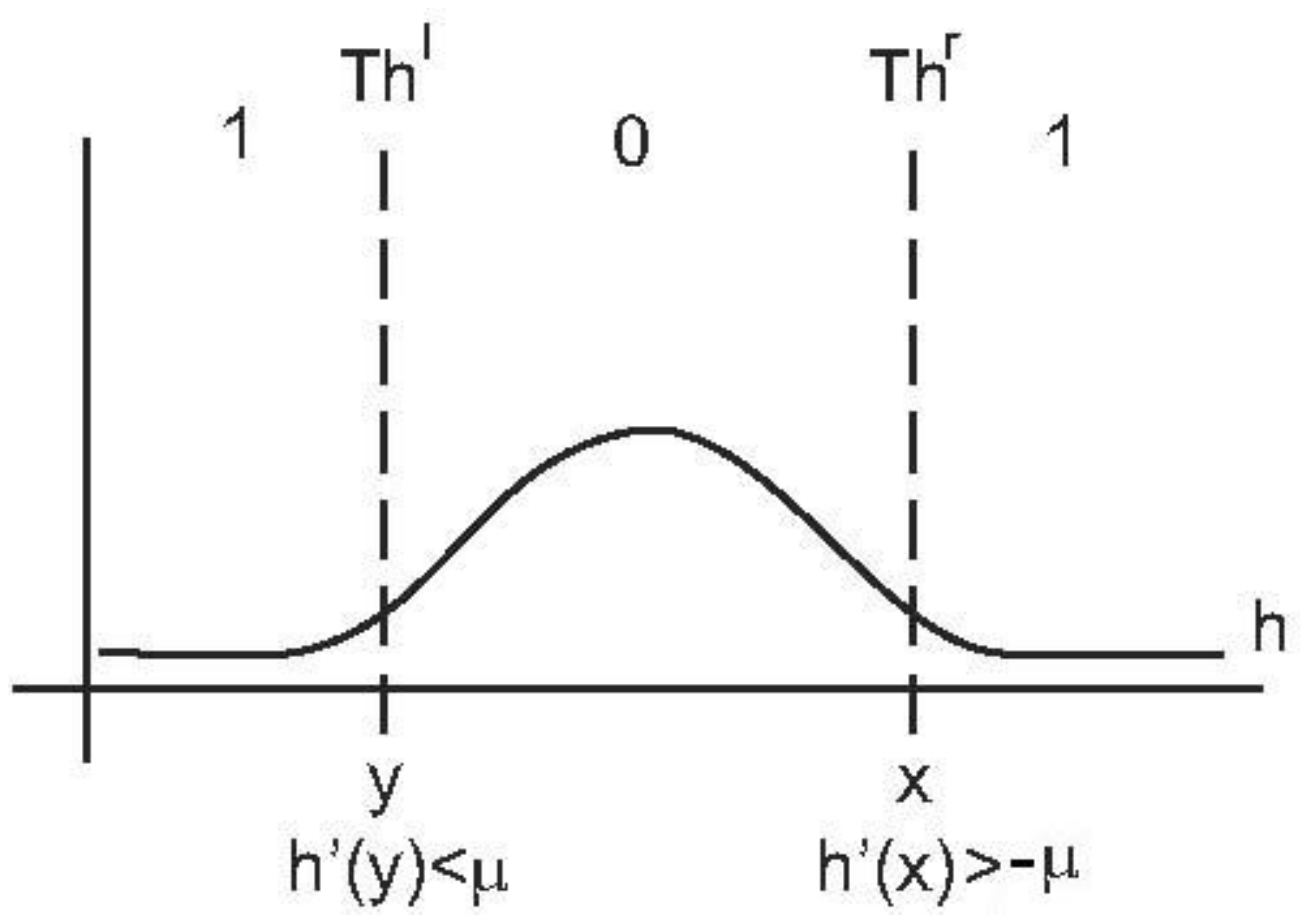} 
\par\end{centering}

\caption{An example that uses the histogram $h$ for finding the threshold
values. %$Th^{r}$
%and $Th^{l}$ are set to $x$ and $y$,
%respectively, where the slope of $h$ becomes smaller than $\mu$.
}

\label{fig:hist_rgb}
\end{figure}

The classification of the pixels in the input frame $\tilde{s}_{i}^{rgb}$
is determined according to $Th^{r}$ and $Th^{l}$. For each color
component and for each $k,l=1,\ldots,N$ 
\[
\breve{s}_{i}^{rgb}(k,l)=\left\{ \begin{array}{cc}
0, & if\,\, Th^{l}<\tilde{s}_{i}^{rgb}(k,l)<Th^{r};\\
1, & otherwise
\end{array}\right.
\]
 See Fig.\ref{fig:hist_rgb} for an example how the thresholds are
derived.

The process is executed three times, one for each of the RGB channels.
The outputs are combined by a pixel-wise OR operation.

\subsubsection{Scan by depth-first search (DFS) \label{sub:dfs}}

The last step of the DBSDB algorithm is the application of a DFS.
Let $\breve{s}_{i}^{g}\in\breve{S}^{g}$ and $\breve{s}_{i}^{rgb}\in\breve{S}^{rgb}$
be the $i^{th}$ output frames of the grayscale and the RGB classification
steps (steps 3 and 4), respectively. Each frame is a binary mask represented
by a matrix. The DFS step combines both outputs. In $\breve{s}_{i}^{g}$
there are false negative detections and in $\breve{s}_{i}^{rgb}$
there are false positive detections. We use each foreground pixel
in $\breve{s}_{i}^{g}$ as a reference point from which we begin a
DFS in $\breve{s}_{i}^{rgb}$. The goal is to find the connected components
of the graph whose vertices are constructed from the pixels in $\breve{s}_{i}^{rgb}$
and whose edges are constructed according to the 8-neighborhood of
each pixel.

\begin{figure}[!h]

\begin{centering}
\includegraphics[bb=0bp 0bp 292bp 170bp,width=0.8\columnwidth]{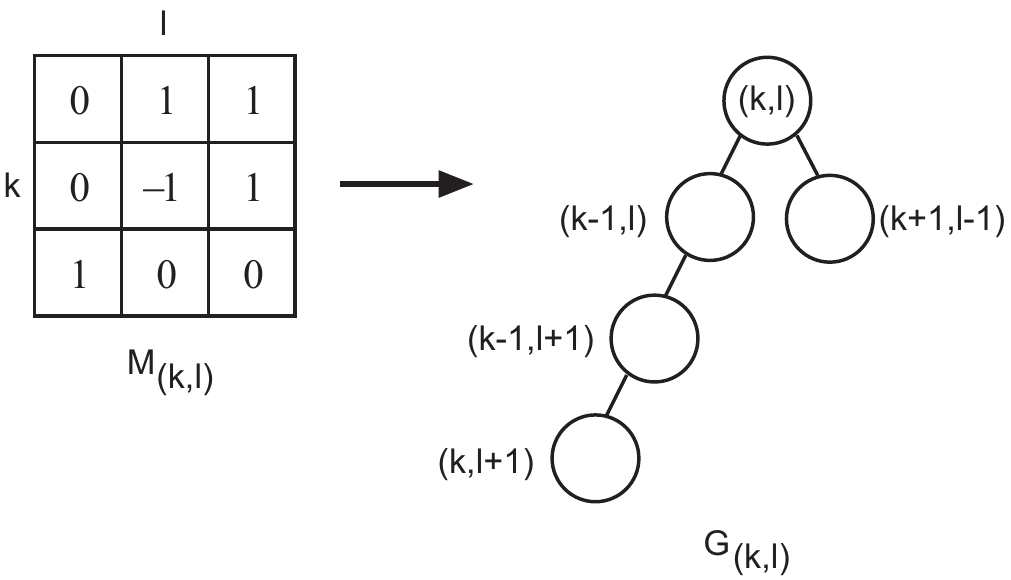} 
\par\end{centering}

\caption{$G_{(k,l)}$ is a graph representation of the 8-neighborhood matrix
$M_{(k,l)}$ whose root is the pixel $\breve{s}_{i}^{rgb}(k,l)$.
A root pixel is set to -1, a foreground pixel is set to 1 and a background
pixel is set to 0.}

\label{fig:dfs} 
\end{figure}

The graph on which the DFS is applied, is constructed as follows: 
\begin{itemize}
\item A pixel $\breve{s}_{i}^{rgb}(k,l)$ is a root if $\breve{s}_{i}^{g}(k,l)$
is a foreground pixel and it has not been classified yet as a foreground
pixel by the algorithm. 
\item A pixel $\breve{s}_{i}^{rgb}(k,l)$ is a node if it is a foreground
pixel and was not marked yet as a root. 
\item Let $\breve{s}_{i}^{rgb}(k,l)$ be a node or a root and let $M_{(k,l)}$
be a $3\times3$ matrix that represents its 8-neighborhood. A pixel
$\breve{s}_{i}^{rgb}(q,r)\in M_{(k,l)}$ is a child of $\breve{s}_{i}^{rgb}(k,l)$
if $\breve{s}_{i}^{rgb}(q,r)$ is a node (see Fig.\ref{fig:dfs}). 
\end{itemize}
The DFS is applied from each root in the graph. Each node, that is
scanned by the DFS, represents a pixel that belongs to the foreground
objects that we wish to find. The scanned pixels are marked as the
new foreground pixels and the others as the new background pixels.

\subsection{A parallel extension of the SBSDB and the DBSDB algorithms}

We propose parallel extensions to the SBSDB and the DBSDB algorithms.
We describe this scheme for the SBSDB algorithm. Nevertheless, the
same scheme can be used for the DBSDB algorithm.

First, the data cube $D_{n}=\left\{ s_{i,j}^{t}\right\} _{i,j=1,...,N;\, t=1,...,n}$
(Eq. \ref{eq:D_n}) is decomposed into overlapping blocks $\left\{ \beta_{k,l}\right\} $.
Next, the SBSDB algorithm is independently applied on each block.
This step can run in \emph{parallel}. The final result of the algorithm
is constructed using the results from each block. Specifically, the
result from each block is placed at its original location in $D_{n}$.
The result for pixels that lie in overlapping areas between adjacent
blocks is obtained by applying a logical $OR$ operation on the corresponding
blocks results.

%----------------------------

\section{Experimental results }

\label{EXPERIMENTAL}

In this section, we present the results from the application of the
SBSDB and DBSDB algorithms. The section is divided into three parts:
%(a) A comparison between the outputs obtained by the application of
%the DB based algorithm and the outputs that are obtained when the DB
%algorithm is replaced by \emph{principal components analysis} (PCA).
The first part is composed from the results of the SBSDB algorithm
when applied to a SBG video. The second part contains the results
from the application of the DBSDB algorithm to a DBG video. In the
third part we compare between the results obtained by our algorithm
and those obtained by five other background-subtraction algorithms.

\subsection{Performance analysis of the SBSDB algorithm}

\begin{figure}[!h]
\begin{centering}
\includegraphics[bb=0bp 267bp 595bp 575bp,clip,width=1\columnwidth]{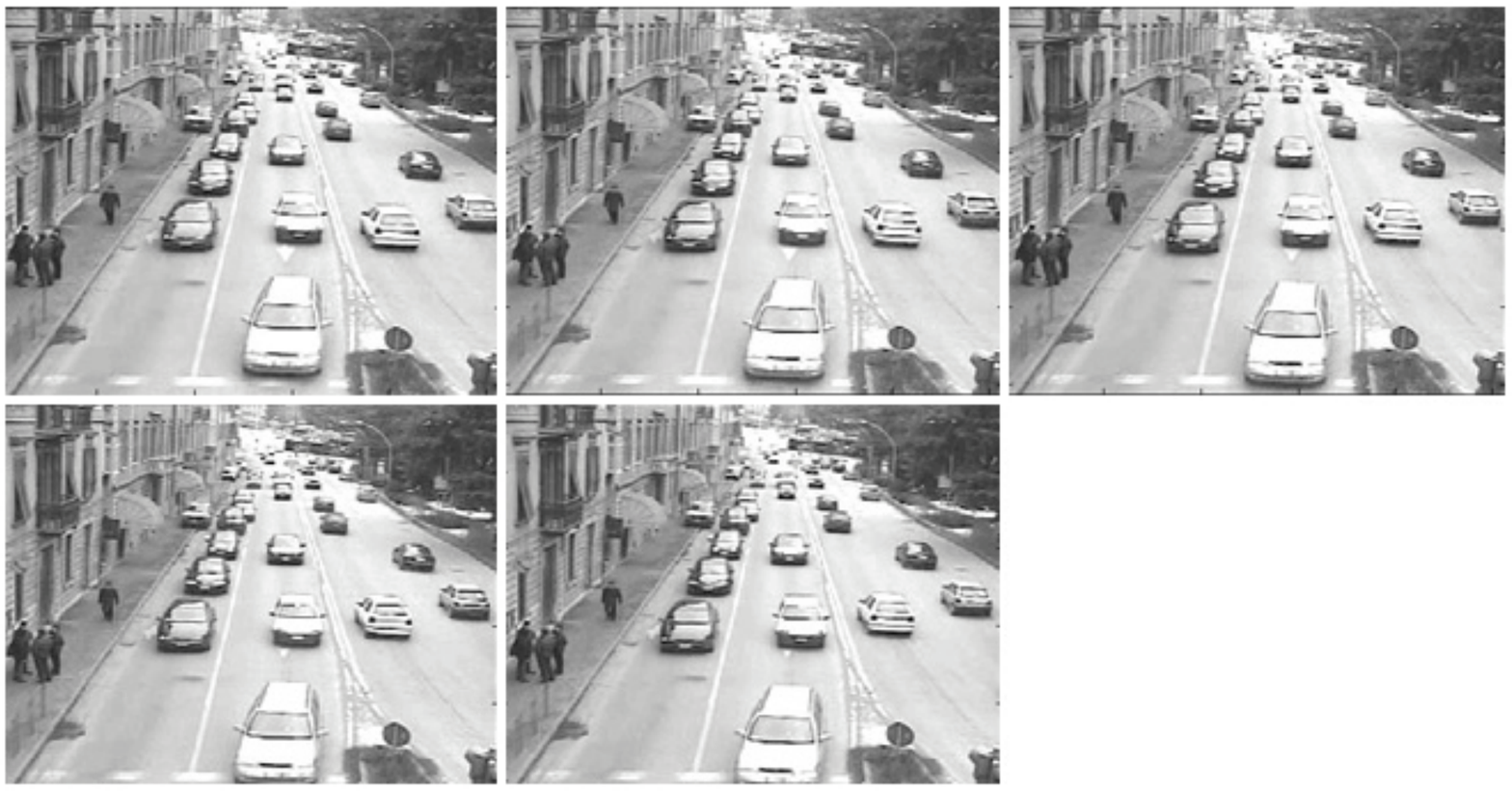} 
\par\end{centering}

\caption{The frames that $W_{s}$ contains. The test frame $s$ is the top-left
frame. The frames are ordered from top-left to bottom-right. }

\label{fig:steady_bg_org}
\end{figure}

\begin{figure}[!h]
\begin{centering}
\includegraphics[bb=0bp 358bp 595bp 492bp,clip,width=1\columnwidth]{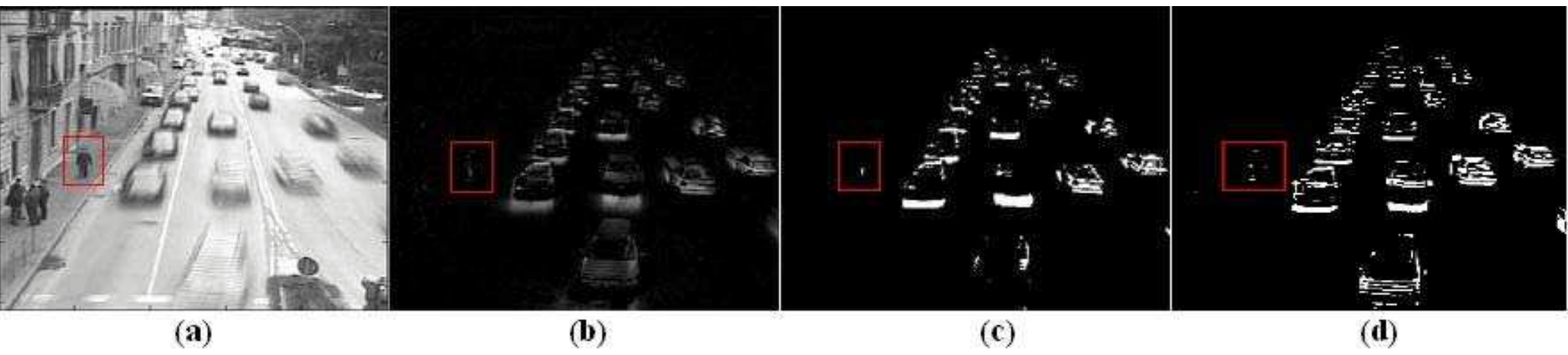} 
\par\end{centering}

\caption{(a) The background for the test frame $s$. (b) The test frame $s$
after the subtraction of the background. (c) The output for the test
frame $s$. (d) The output for the test frame $s$ from the parallel
version of the algorithm. %The person that walks on the
%sidewalk (marked by a square around him), on the left side of the
%frame, is recognized by the SBSDB algorithm.
}

\label{fig:steady_bg}
\end{figure}

We apply the SBSDB algorithm to a video sequence that consists of
190 grayscale frames of size $256\times256$. The video sequence was
captured by a static camera with a frame rate of 15 fps. The video
sequence shows moving cars over a static background. We apply the
sequential version of the algorithm where the size of the SW is set
to 5. We also apply the parallel version of the algorithm where the
video sequence is divided to four blocks in a $2\times2$ formation.
The overlapping size between two (either horizontally or vertically)
adjacent blocks is set to 20 pixels and the size of the SW is set
to 10. Let $s$ be the test frame and let $W_{s}$ be the SW starting
at $s$. In Fig. \ref{fig:steady_bg_org} we show the frames that
$W_{s}$ contains. The output of the SBSDB algorithm for $s$ is shown
in Fig. \ref{fig:steady_bg}.

\subsection{Performance analysis of the DBSDB algorithm}

We apply the DBSDB algorithm to five RGB video sequences. The first
four video sequences have a frame rate of 30 fps. The last video sequence
has a frame rate of 24 fps. All the video sequences, except the first
video sequence, are of size $320\times240$. The first video sequence
is of size $210\times240$. The video sequences were produced by a
static camera and contain dynamic backgrounds.

The input video sequences are: 
\begin{enumerate}
\item People walking in front of a fountain. It contains moving objects
in the background such as water flowing, waving trees and a video
screen whose content changes over time. The DBSDB input is a RTD that
contains 170 frames and a BGD that contains 100 frames. The output
of the DBSDB is presented in Fig. \ref{fig:res_jap}(g).

\begin{figure}[!h]
\begin{centering}
\includegraphics[bb=0bp 284bp 595bp 558bp,clip,width=1\columnwidth]{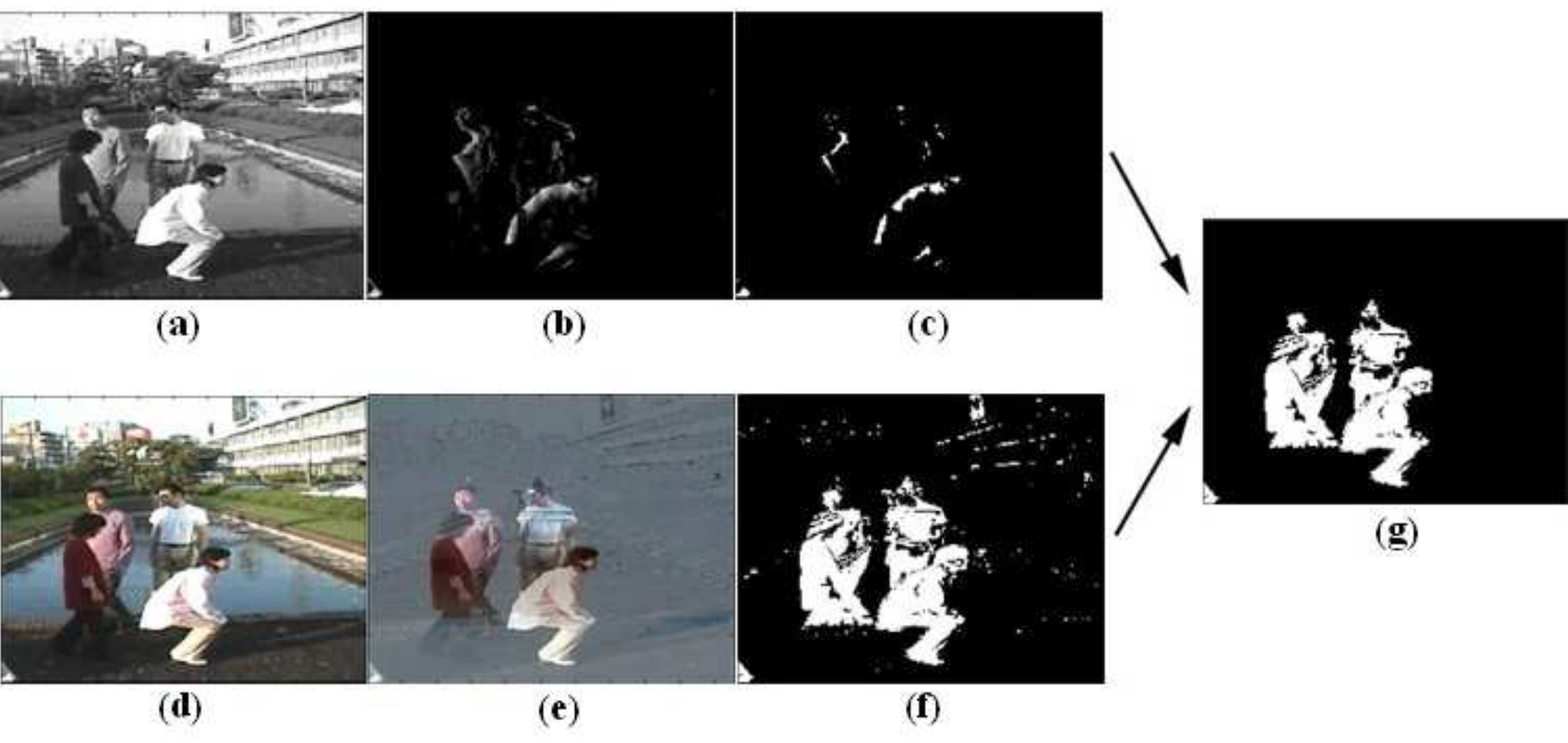} 
\par\end{centering}

\caption{(a), (d) The original test frames in grayscale and RGB, respectively.
(b), (e) The grayscale and RGB test frames after the background subtraction
in the classification steps (steps 3 and 4) of the DBSDB algorithm,
respectively. (c), (f) Results after the thresholding of (b) and (e),
respectively. (g) The final output of the DBSDB algorithm after the
application of the DFS.}

\label{fig:res_jap}
\end{figure}

\item A person walking in front of bushes with waving leaves. The DBSDB
input is a RTD that contains 88 frames and a BGD that contains 160
frames. The output of the DBSDB algorithm is presented in Fig. \ref{fig:res_bushes}(g).

\begin{figure}[!h]
\begin{centering}
\includegraphics[bb=0bp 284bp 595bp 560bp,clip,width=1\columnwidth]{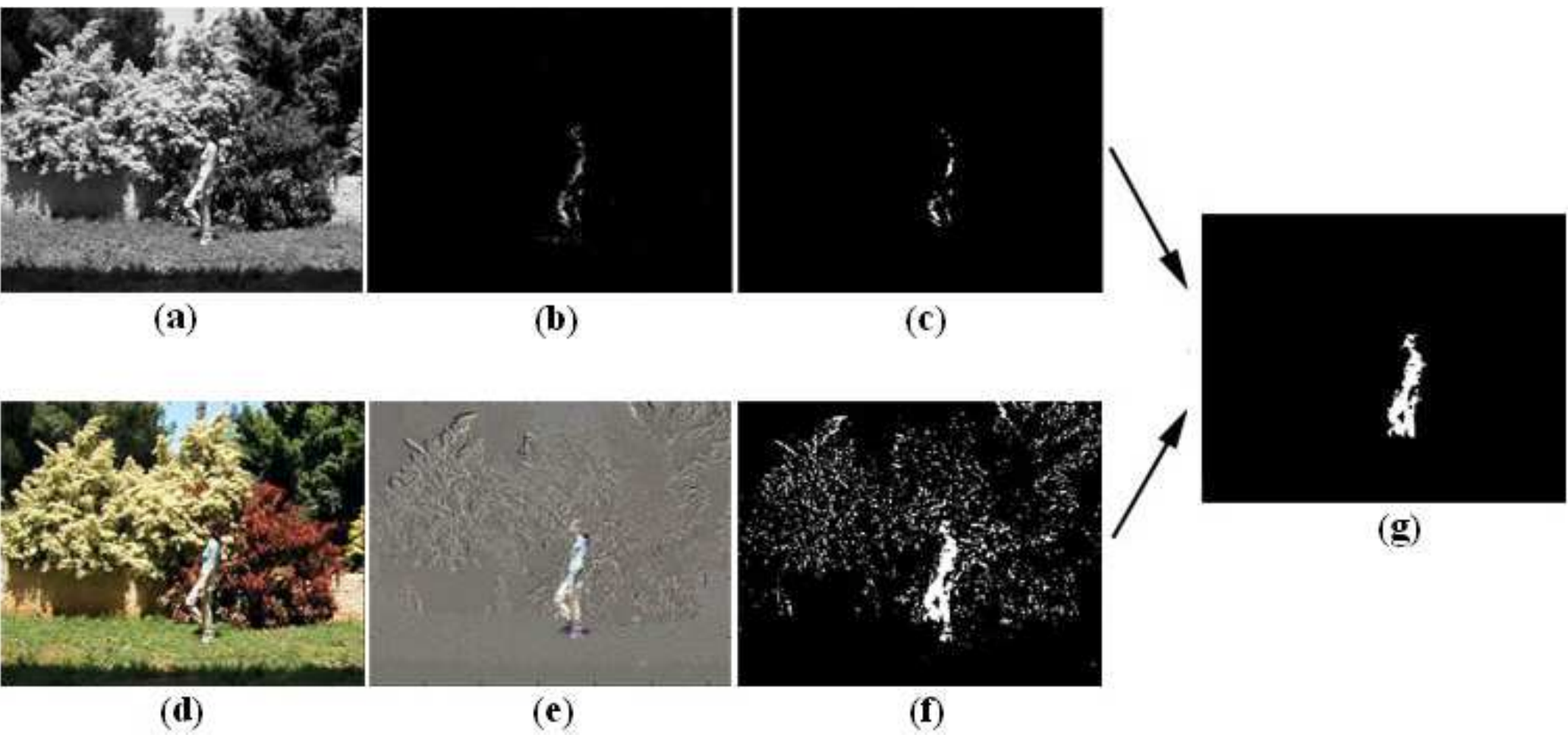} 
\par\end{centering}

\caption{(a), (d) The original test frames in grayscale and RGB, respectively.
(b), (e) The grayscale and RGB test frames after the background subtraction
in the classification steps (steps 3 and 4) of the DBSDB algorithm,
respectively. (c), (f) Results after the thresholding of (b) and (e),
respectively. (g) The final output of the DBSDB algorithm after the
application of the DFS.}

\label{fig:res_bushes}
\end{figure}

\item A moving ball in front of waving trees. The DBSDB input is a RTD that
contains 88 frames and a BGD that contains 160 frames. A frame from
the video sequence is shown in Fig. \ref{fig:more_res}(a). The output
of the DBSDB algorithm is presented in Fig. \ref{fig:more_res}. Figure
\ref{fig:more_res}(d) contains the result of the sequential version
of the algorithm and Fig. \ref{fig:more_res}(g) contains the results
of the parallel version. In results of the parallel version, the video
sequence was divided to four blocks in a $2\times2$ formation. The
overlapping size between two (either horizontally or vertically) adjacent
blocks was set to 20 pixels and the size of SW was set to 30. 
\item A ball jumping in front of trees and a car passing behind the trees.
The DBSDB input is a RTD that contains 106 frames and a BGD that contains
160 frames. A frame from the video sequence is shown in Fig. \ref{fig:more_res}(b).
The output of the DBSDB algorithm is presented in Fig. \ref{fig:more_res}(e). 
\item A person walking in front of a sprinkler. The DBSDB input is a RTD
that contains 121 frames and a BGD that contains 100 frames. A frame
from the video sequence is shown in Fig. \ref{fig:more_res}(c). The
output of the DBSDB algorithm is presented in Fig. \ref{fig:more_res}(f).

\begin{figure}[!h]
\begin{centering}
\includegraphics[bb=0bp 138bp 595bp 704bp,clip,width=1\columnwidth]{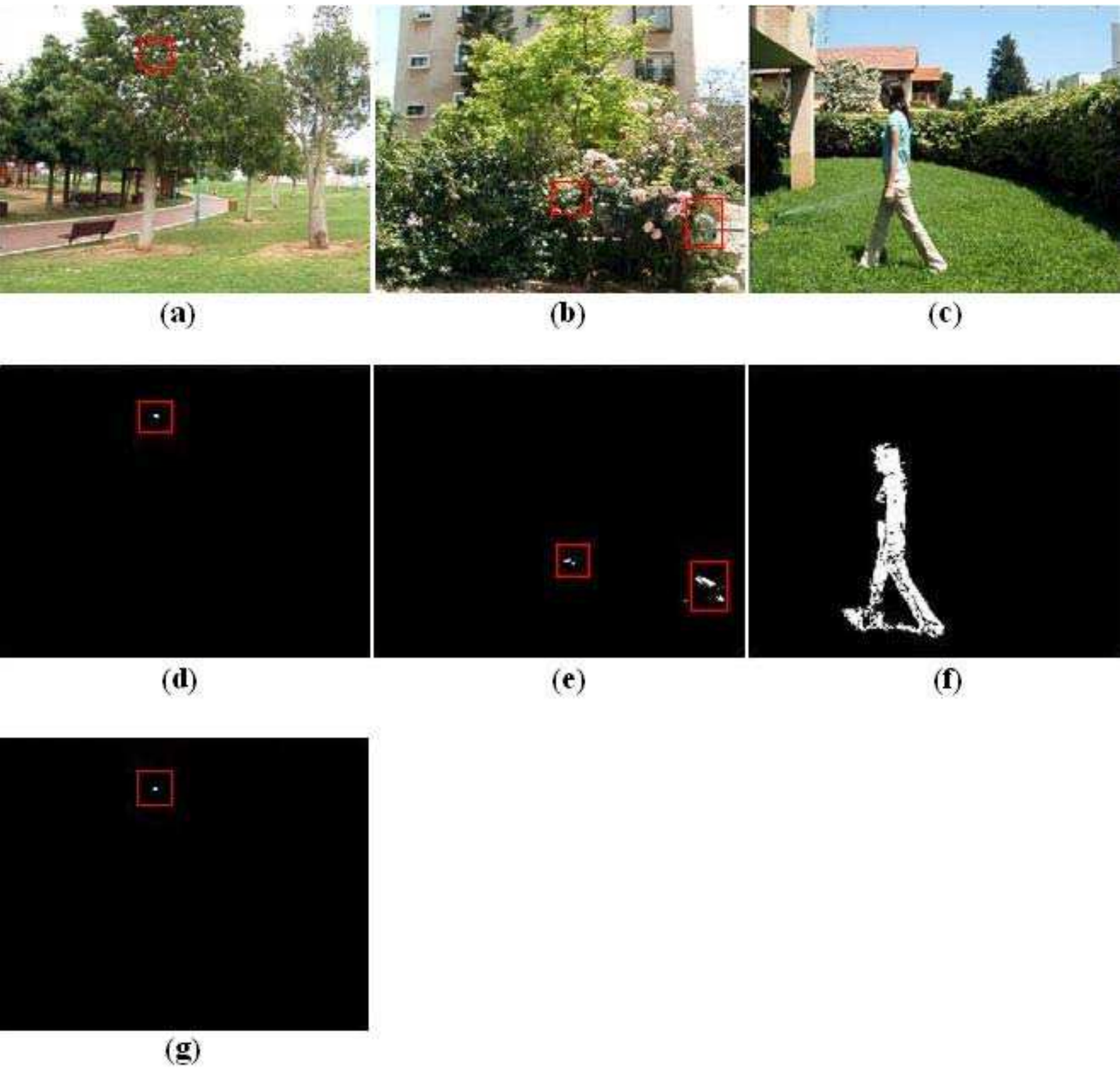} 
\par\end{centering}

\caption{ (a)-(c) The original test frames. (d)-(g) The segmented outputs from
the application of the DBSDB algorithm. (a) A ball in front of waving
trees. (d), (g) The result of the sequential and parallel versions
of the algorithm applied on (a), respectively. (b), (e) A ball jumping
in front of a tree and a car passing behind the trees. (c), (f) A
person walking in front of a sprinkler.}

\label{fig:more_res} 
\end{figure}

\end{enumerate}

\subsection{Performance comparison between the BSDB algorithm and other algorithms
\label{sub:comparison}}

We compared between the BSDB algorithm and five different background
subtraction algorithms. The input data and the results were taken
from \cite{TKBM99}. All the test sequences were captured by a camera
that has three CCD arrays. The frames are of size 160x120 in RGB format
and are sampled at 4Hz. The test frame that was segmented, in video
sequences where the background changes, is taken to be the frame that
appears 50 frames after the frame where the background changes. On
every output frame (besides the output of the BSDB algorithm), a speckle
removal \cite{TKBM99} was applied to eliminate islands of 4-connected
foreground pixels that contain less than 8 pixels. All other parameters
were adjusted for each algorithm in order to obtain visually optimal
results over the entire dataset. The parameters were used for all
sequences. Each test sequence begins with at least 200 background
frames that were used for training the algorithms, except for the
bootstrap sequence. Objects such as cars, which might be considered
foreground in some applications, were deliberately excluded from the
sequences.

\begin{figure}[!h]
\begin{centering}
\includegraphics[clip,width=1\columnwidth]{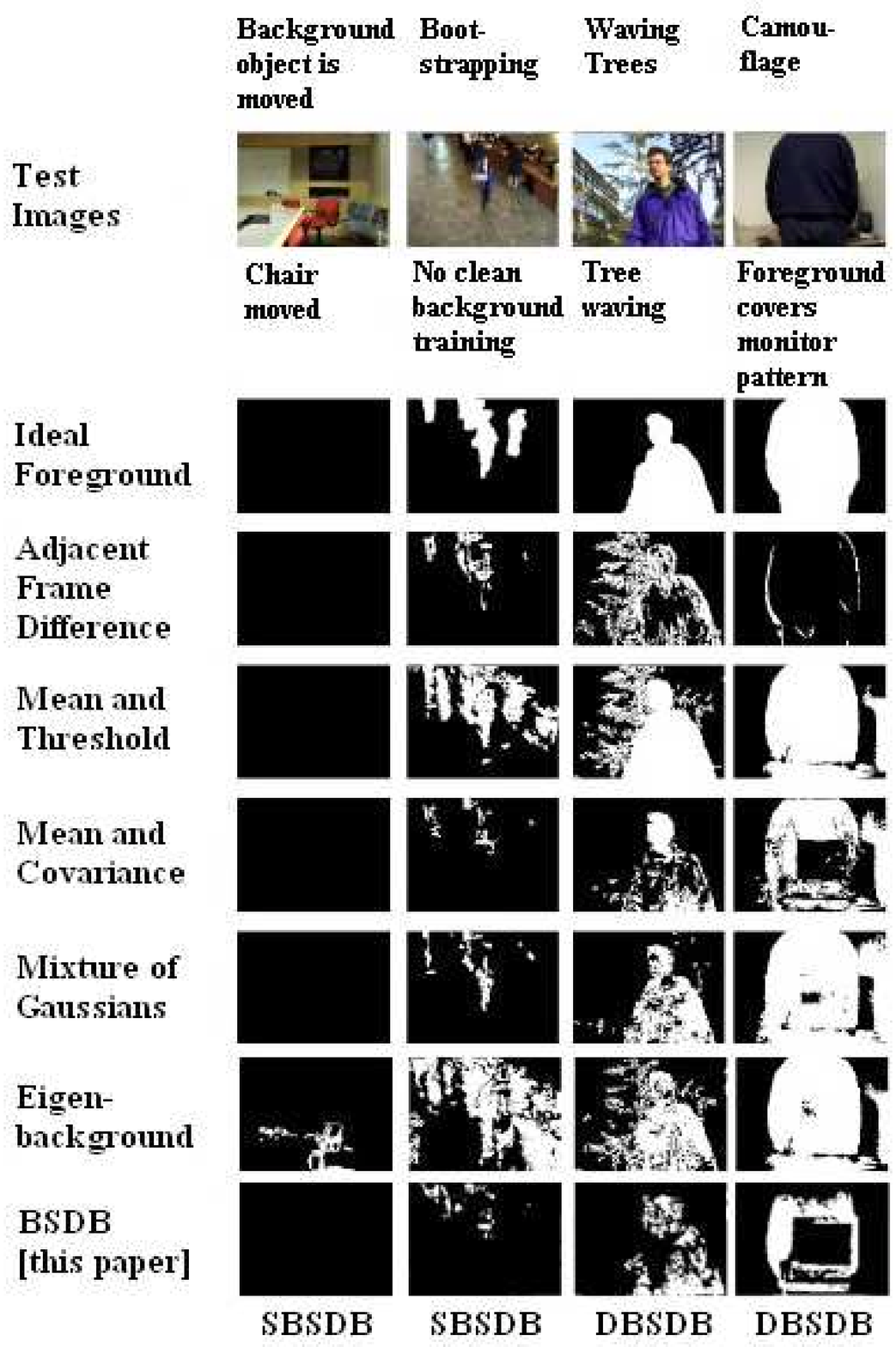} 
\par\end{centering}

\caption{The outputs from the applications of the BSDB and five other algorithms.
Each row shows the results of one algorithm, and each column represents
one problem in background maintenance. The top row shows the test
frames. The second row shows the optimal background outputs.}

\label{fig:comp}
\end{figure}

Each of the sequences poses a different problem in background maintenance.
The chosen sequences and their corresponding problems are:
\begin{enumerate}
\item \textbf{Background object is moved} - Problem: A background object
that changes its position. These objects should not be considered
as part of the foreground. The sequence contains a person that walks
into a conference room, makes a telephone call, and leaves with the
phone and a chair in a different position. The test frame is the one
that appears 50 frames after the person has left the scene. 
\item \textbf{Bootstrapping} - Problem: A training period without foreground
objects is not available. The sequence contains an overhead view of
a cafeteria. There is constant motion and every frame contains people. 
\item \textbf{Waving Trees} - Problem: Backgrounds can contain moving objects.
The sequence contains a person walking in front of a swaying tree. 
\item \textbf{Camouflage} - Problem: Pixels of foreground objects may be
falsely recognized as background pixels. The sequence contains a monitor
on a desk with rolling interference bars. A person walks into the
scene and stands in front of the monitor. 
\end{enumerate}
We apply six background subtraction algorithms to the five video sequences,
including the algorithm that is presented in this chapter. The background
subtraction algorithms are: 
\begin{enumerate}
\item \textbf{Adjacent Frame Difference} - Each frame is subtracted from
the previous frame in the sequence. Absolute differences greater than
a threshold are marked as foreground. 
\item \textbf{Mean and Threshold} - Pixel-wise mean values are computed
during a training phase, and pixels within a fixed threshold of the
mean are considered background. 
\item \textbf{Mean and Covariance} - The mean and covariance of pixel values
are updated continuously \cite{KWM94}. Foreground pixels are determined
by applying a threshold to the Mahalanobis distance. 
\item \textbf{Mixture of Gaussians} - This algorithm is reviewed in Section
\ref{sub:relatedWork}. 
\item \textbf{Eigen-background} - This algorithm is reviewed in Section
\ref{sub:relatedWork}. 
\item \textbf{BSDB} - The algorithm presented in this chapter (Section \ref{BSDB}).
\end{enumerate}
The outputs of these algorithms are shown in Fig. \ref{fig:comp}.

We applied the SBSDB algorithm to the first two video sequences: the
moved chair and the bootstrapping. In both cases, the background in
the video sequence is static. In the first video sequence, the SBSDB
algorithm handles the changes in the position of the chair that is
a part of the background. The SBSDB algorithm does not require a training
process so it can handle the second video sequence where there is
no clear background for training. Algorithms that require a training
process can not handle this case.

We applied the DBSDB algorithm to the waving trees and the camouflage
video sequences. In both cases, the background in the video sequences
is dynamic. In the moving trees video sequence, the DBSDB algorithm
captures the movement of the waving trees and successfully eliminates
it from the video sequence where as the other algorithms produce false
positive detections. The DBSDB algorithm does not handle well the
last video sequence where the foreground object covers the background
moving object (the monitor). In this case the number of false negative
detections is significant.

\section{Conclusion and future work}

We introduced in this chapter the BSDB algorithm for automatic segmentation
of video sequences. The algorithm contains two versions: the SBSDB
algorithm for video sequences with static background and the DBSDB
algorithm for video sequences that contain dynamic background. The
BSDB algorithm captures the background by reducing the dimensionality
of the input via the DB algorithm. The SBSDB algorithm uses an on-line
procedure while the DBSDB algorithm uses both an off-line training
procedure and an on-line procedure. During the training phase, the
DBSDB algorithm captures the dynamic background by iteratively applying
the DB algorithm on the background training data. 

The BSDB algorithm presents a high quality segmentation of the input
video sequences. Moreover, it was shown that the BSDB algorithm produces
results that are highly competitive with current state-of-the-art
algorithms by coping with difficult situations of background maintenance.

The performance of the BSDB algorithm can be enhanced by improving
the accuracy of the threshold values. Furthermore, it is necessary
to develop a method for automatic computation of $\mu$, which is
used in the threshold computation (Sections \ref{sub:GrayThreshold}
and \ref{sub:RGBThreshold}).

Additionally, the output of the BSDB algorithm contains a fair amount
of false negative detections when a foreground object obscures a brighter
background object. This will be improved in future versions of the
algorithm.

The BSDB algorithm can be useful to achieve low-bit rate video compression
for transmission of rich multimedia content. The captured background
is transmitted once followed by the detected segmented objects.

\chapter{Automatic Identification of Features in Hyper-Spectral Data \label{cha:Automatic-identification-of}}

In this chapter we show how to uniquely construct a spectral signature
that enables to identify and distinguish spectrally unique materials
in a data base of spectral signatures. 

This Chapter uses the basic concepts of hyper-spectral imagery which
were introduced Chapter \ref{cha:Introduction-to-Hyper-spectral}.
This is followed by a detailed overview of the image-processing method
that we introduce as well as two algorithms that implement this approach.%Finally we describe the execution of these algorithms in IDL+ENVI. 

\section{Introduction}

In this chapter, we introduce a new approach that chooses only certain
band values from the inspected substance. These values reflect the
physical properties of the material. In this sense, we achieve dimensionality
reduction, and make it easy to link between the resultant match and
the material's properties.

After extracting the ``interesting'' features two methods for the
analysis of the signal are presented: 
\begin{description}
\item [{1.~Exact~search:}] Building a tree structure classifier, using
only a part of the extracted data (depends on the spectral library
size). 
\item [{2.~Approximate~search:}] Constructing a classifier, which enables
to find spectra that have some common features with the target. 
\end{description}
The rest of this chapter is organized as follows: related works are
presented in Section \ref{related}. Section \ref{Features} describes
the algorithm for extraction of spectral features. In Section \ref{sec:exact-search}
we introduce the exact search, including: tree construction, target
identification and illustration. The approximate search, including:
the feature distribution table, the target identification and an example
are described in Section \ref{Approximate_identification}. %The IDL+ENVI implementation of these algorithms is given in section .

\section{Related work\label{related} }

Several algorithms for the analysis of hyper-spectral imagery can
be found in the literature. Most of them suffer from the following
disadvantage: physical properties of the materials under inspection
are not used during the data processing. Consequently, when the output
results are analyzed, it is hard to understand how the the material
characteristics are connected to reasons that led to the results/identification.
Another common disadvantage of existing methods is that they tend
to use all the data in the hyper-spectral image (e.g. Spectral Angle
Mapper \cite{Carvalho00}). We assume that a large portion of this
data is unnecessary, and adds complexity. Moreover, in our proposed
algorithm, the comparison between spectral signatures is based on
the location of the spectral features in the spectrum. This means
that the reflectance value of the spectral feature at this location
does not affect the comparison.

In the following, we describe current state-of-the-art methods for
spectral identification.
\begin{description}
\item [{Spectral~Angle~Mapper~(SAM)}] (\cite{Kruse92,Kruse93,Kruse93a}).
The reflectance of a hyper spectral pixel can be described as an $N$-dimensional
vector, where $N$ is the number of bands. The length of the vector
is the brightness, and the direction of this vector can be regarded
as the spectral features that are contained in the pixel. Changes
in the illumination affect only the length of the vector but not its
angle. The classification of a target is done by calculating the angle
between the vector of the analyzed pixel and the vectors (representing
materials) from the spectral libraries/database. The pixel that classifies
the material is the one with the lowest spectral angel value.

The drawback of the SAM method is that it does not distinguish between
positive and negative correlation values since it takes into account
only the absolute value of the correlation. The SCM method was designed
to overcomes this limitation.

\item [{Spectral~Correlation~Mapper~(SCM)}] (\cite{Carvalho00}). This
method performs a similar computation to the one that is performed
by the SAM method. Namely, the angles between the spectra DB and the
analyzed pixel. However, the SCM method standardizes the vectors of
the reflected spectra. Thus, the Pearsonian Correlation Coefficients
(PCC) are used.%The difference between SAM and SCM is that the vectors of the reflected spectrum are being standardized before calculating the angle. Namely, if in the SAM, the angle can be calculated in SCM the Pearsonian Correlation Coefficient (PCC) is used but the computation is very similar. The difference is that the PCC standardizes the data.
\item [{Spectral~Identification~Method~(SIM)}] (\cite{Carvalho01,Carvalho05}).
This method uses the statistical procedure ANOVA (\cite{Davis73,Souza98,Steel80,Vieira88})
that is based on linear regression. It is not affected by negative
correlation values. By combining it with the SCM coefficients, it
uses the negative/positive information. This technique exhibits three
estimates according to the levels of significance of the materials.
\item [{Spectral~Feature~Fitting\ (SFF)}] (\cite{Clark95}). This approach
examines the specific absorption features in the spectra. The U.S.
Geological Survey (\cite{Clark01}) has developed an advanced method
based on this approach called \emph{Tetracorder}. %In the ENVI program XXX, a simpler form of this method can be found. It is called Spectral Feature Fitting.
SFF is a simpler method that is based on this approach.%An implementation of the SFF method can be found in the ENVI software.
In this technique, the user defines a range of wavelengths which form
a unique absorption feature that exists in the chosen target (the
isolation of the feature is done using a continuum mathematical function).
Often the highpoint enclosing the feature is identified and a line
fit between points is used in order to normalize the absorption feature
by embedding the original spectra into the continuum. The comparison
is based on the following two characteristics: The depth of the features
in the target versus the depth in the analyzed pixel and the shape
of the features in the target versus the shape in the analyzed pixel
(using a least-square technique).
\item [{Optimum~Index~Factor~(OIF)}] (\cite{Chavez84}). In this method,
an index, which is called the OIF, is used to select the optimal combination
of any number of bands in the satellite image in order to create a
color composite. The bands that contain the highest amount of information
are chosen. The OIF is based on the total variance within bands and
correlation coefficients between bands. Every combination of bands
is assigned an OIF. The optimal combination of bands, which is chosen,
is defined as the combination of bands with the lowest correlation
coefficient between bands and with the highest total variance within
bands.\\
This approach is quite different from the previous approaches. While
other approaches try to deal with the \emph{whole} spectrum, this
approach reduces the high-dimensionality of the data (manifested as
superfluous data with high inter-band correlation). \\
The approach that is introduced in this chapter follows the logic
of OIF (selection of specific band for classification) and deals with
the high-dimensionality of the data, unlike all the methods that were
introduced above. 
\end{description}

\section{Feature extraction\label{Features} }

The spectra data that is used as input to the proposed algorithm represent
absorption values. Nevertheless, it can operate in the same way if
the spectra represent radiation, reflection or any other type of hyper-spectral
data. The description of the general algorithm is accompanied by a
detailed example of a database that contains 173 entries (spectra)
each of which contains 2001 wavelengths. The length of each spectrum
(vector) in the database is predefined - in our case it is 2001 -
which is the number of wavelengths. In order to construct a robust
and efficient classifier, the dimensionality of the vectors has to
be reduced. This is achieved by extraction of characteristic features
of the spectra. If a spectrum is treated as a continuous curve, then
its geometrical characteristics are determined by the physical properties
of the material. We use four types of geometrical characteristics
to represent the physical properties that can facilitate the unique
identification of a spectral signature of the material in the database.
We demonstrate it on the following four features%
\footnote{A different number and type of physical features can be chosen.%
}: 
\begin{enumerate}
\item Location of deep minima. 
\item Location of shallow and one-side minima. 
\item Location of near-horizontal flat intervals. 
\item Location of inflection points. 
\end{enumerate}
These features describe a spectral signature based upon the \emph{absorption}
features and back-scattering parameters of the material. Deep minima
correspond to wavelengths where the light absorption is high while
shallow and one-side minima correspond to wavelengths where the light
absorption is small to moderate. Near-horizontal flat intervals correspond
the subranges of the spectrum where the absorption amount is approximately
constant. Finally, inflection points are locations in the spectrum
where there is a change in the manner the light is absorbed. Specifically,
looking at the spectrum subranges on both sides of the inflection
point, the rate and direction of absorption will change from one subrange
to the other. We demonstrate the strength of this spectral signature
using the clay minerals \emph{Montmorillonite} and \emph{Kaolinite}.
Both minerals have an absorption band at 2205 \emph{nm} where as only
Kaolinite has an additional small absorption band at 2165 \emph{nm}
(shallow minima). The proposed method enables to distinguish between
these two clay minerals and points out their different features. Both
the inflection point position of an absorption band and flat intervals
in a spectral signature, are features that help to distinguish between
spectral features. Currently, our experiments show that the order
in which the selected features are chosen does not affect the final
result. However, the order of the selected features might affect the
final output when different parameters, which correspond to different
physical assumptions, are chosen.

The four features that we use were chosen to demonstrate the performance
of the proposed algorithm. Nevertheless, one can exclude some features
or add new features. For example, the area below two adjacent minima
can be another feature. There is no limitations on the number of features
and their meaning. Any feature can be added, provided its effectiveness
was established.

% The algorithm uses some parameters. All the hardwired" parameters (numbers) that appear throughout the algorithm are parameters that were chosen arbitrarily but they can be modified and changed as needed.

The problem is to correctly locate essential physical events even
in the presence of strong noise. Let $\textbf{y}=\{y(k)\}$ be a set
of original raw spectra of length $N$ (number of wavelengths). We
denote by 
\[
Dy(k)=\frac{y(k+1)-y(k-1)}{2},\quad D^{2}y(k)=\frac{Dy(k+1)-Dy(k-1)}{2},~~k=2,\ldots,N
\]
the first and the second centered differences of the array $\textbf{y}$,
respectively. The resulting sets $\{y(k)\}$, $\{Dy(k)\}$ and $\{D^{2}y(k)\}$
are assumed to be corrupted by noise. Therefore, we filter their elements
by applying a low-pass shape-preserving B-spline filter. The smoothed
arrays are denoted by $\{Y(k)\}$, $\{DY(k)\}$ and $\{D^{2}Y(k)\}$.

\subsection{Finding the locations of the features\label{sub:Finding-the-locations}}

In the following, we describe how the locations of the four features
are found. Five thresholds $T_{1},\ldots,T_{5}$ are used during the
calculation. 
\begin{description}
\item [{Location~of~deep/shallow~minima:}] First, we find all the points
$\{k_{c}\}$, where the sequence $\{DY(k)\}$ changes its sign, i.e.
where $DY(k_{c}-1)<0,\; DY(k_{c}+1)>0$ or $DY(k_{c}-1)>0,\; DY(k_{c}+1)<0$.
Without loss of generality, we describe the procedure for points where
$DY(k_{c}-1)<0,\; DY(k_{c}+1)>0$. We classify the deep/shallow minima
points in the following way: let $k_{c}$ be a marked point and let
$k_{cl}\leq k_{c}$ be the point such that for all $k_{cl}\leq k\leq k_{c},\; DY(k)<0$.
Let $k_{cr}\ge k_{c}$ be the point such that for all $k_{c}<k\leq k_{cr}\; DY(k)>0$.
We denote by $V_{l}\stackrel{\Delta}{=}\{k:k_{cl}\leq k<k_{c}\}$
and $V_{r}\stackrel{\Delta}{=}\{k:k_{c}<k\leq k_{cr}\}$ the intervals
to the left and to the right of the point $k_{c}$, respectively.
We calculate $M_{l}\stackrel{\Delta}{=}\max_{k\in V_{l}}Y(k)$, $M_{r}\stackrel{\Delta}{=}\max_{k\in V_{r}}Y(k)$,
$M\stackrel{\Delta}{=}\max(M_{r},M_{l})-Y(k_{c})$ and $m\stackrel{\Delta}{=}\min(M_{r},M_{l})-Y(k_{c})$
and apply the following conditions:

\begin{itemize}
\item If $M<T_{2}$ then we discard the point $k_{c}$. 
\item If $M>T_{2}$ and $m>T_{1}$ then we mark the point $k_{c}=k_{dm}$
as a deep minima. 
\item If $M>T_{2}$ but $m<T_{1}$ then we mark the point $k_{c}=k_{sm}$
as a shallow or one-side minima. 
\end{itemize}
\item [{Location~of~near-horizontal~flat~intervals:}] We find intervals
of length of a given length $\mu$ in which $0<|DY(k)|<T_{5}$. The
central points of these intervals, which we denote by $\left\{ k_{f}\right\} $,
mark the locations of near-horizontal flat intervals. 
\item [{Location~of~inflection~points:}] First, we find all the points
$\{k_{q}\}$, where the second difference $\{D^{2}Y(k)\}$ changes
its sign i.e. $D^{2}Y(k_{q}-1)<0$ and $D^{2}Y(k_{q}+1)>0$ or $D^{2}Y(k_{q}-1)>0$
and $D^{2}Y(k_{q}+1)<0$. Without loss of generality, we describe
the inflection points classification procedure for points $\left\{ k_{q}\right\} $
where $D^{2}Y(k_{q}-1)<0$ and $D^{2}Y(k_{q}+1)>0$. We denote by
$k_{ql}\leq k_{q}$ the point such that for all $k_{ql}\leq k\leq k_{q},\; D^{2}Y(k)<0$
and we denote by $k_{qr}\ge k_{q}$ the point such that for all $k_{q}<k\leq k_{qr}\; D^{2}Y(k)>0$.
Let $V_{w}\stackrel{\Delta}{=}\{k:k_{ql}\leq k\leq k_{qr}\}$ and
$V_{n}\stackrel{\Delta}{=}\{k:k_{q}-\mu\leq k\leq k_{q}+\mu\}$ be
the wide and narrow neighborhoods of the point $k_{q}$, respectively.
We calculate $M_{w}\stackrel{\Delta}{=}\max_{k\in V_{w}}|(DY(k)|$
and $m_{n}\stackrel{\Delta}{=}\min_{k\in V_{n}}|(DY(k)|$. We discard
the point $k_{q}$ if $M_{w}<T_{3}$ (near flat interval) or if $m_{n}>T_{4}$
(near vertical interval). The set of points $\left\{ k_{q}\right\} $
that were not discarded are marked as inflection points.
\end{description}

\paragraph{Thinning the features arrays: }

In order to make the feature extraction robust to noise, the feature
vector is thinned in the following way: 
\begin{itemize}
\item If an interval of a given length $\xi$ , contains several deep minima
points, we retain for that interval only the deepest minimum point
i.e. the point for which the value of $Y$ is the smallest. Other
points inside the interval are discarded.
\item A similar procedure is applied to shallow minima points, indicators
of flat intervals and inflection points. However, in the latter case,
points with the smallest $|DY|$ values are retained. 
\end{itemize}

\paragraph{\noindent Merging~the~features~arrays }

\noindent After finding the feature points, they are merged into a
single feature vector. The merging procedure considers the features
in the following order: deep minima points, shallow minima points,
flat intervals and inflection points. This order does not affect the
output of the algorithm. It only affects the structure of the classifier
tree. Consequently, we carry out the following procedures:
\begin{itemize}
\item If within distance of $d$ ($d$ is chosen empirically) samples from
an inflection point lies a minimum point or an indicator of a flat
interval then this inflection point is discarded.
\item If within distance of $d$ samples from an indicator of a of flat
interval lies a minimum point then this indicator is discarded.
\item If within distance of $d$ samples from a shallow minimum lies a deep
minimum point then this shallow minimum point is discarded. 
\end{itemize}
Finally, for a given spectrum $\#n$, we get a reduced set of features,
which we arrange as a $4\times K$ matrix ${\bf X}_{n}$ where $K$
denotes a maximal number of sought after features per feature type.% We assume that each spectrum has at most K locations for each selected feature.

\begin{figure}[!h]
\begin{centering}
\includegraphics[bb=60bp 425bp 570bp 780bp,clip,width=0.9\columnwidth]{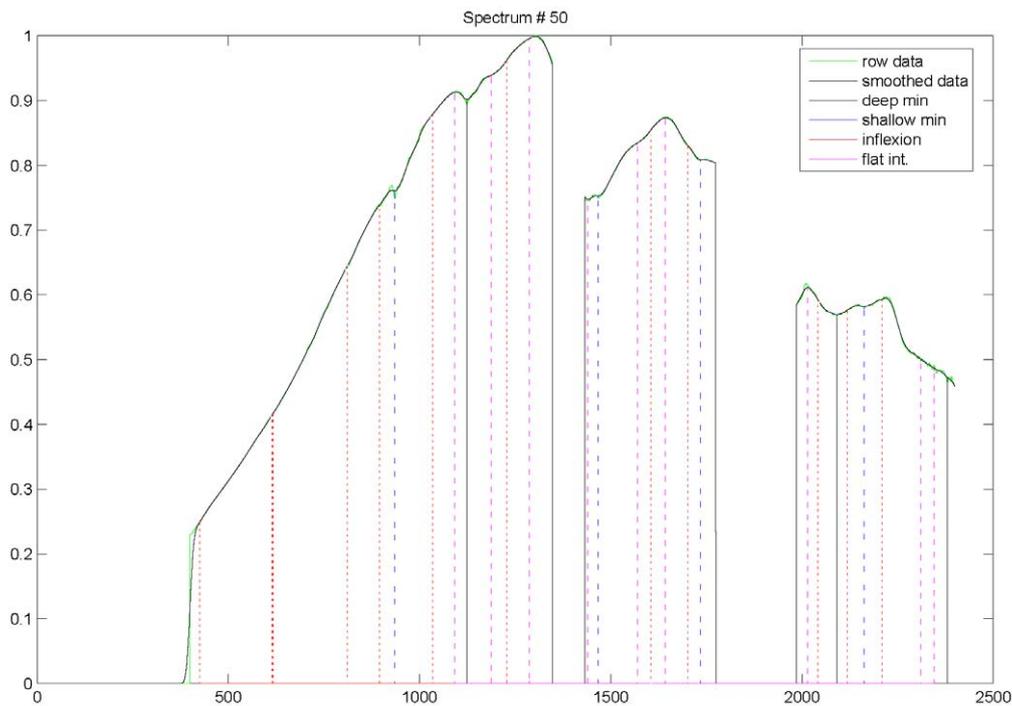} 
\par\end{centering}

\caption{Spectrum $\#50$. Green curve -- original raw data, black curve --
smoothed data after the application of 4$^{th}$-order B-spline to
the raw data, black vertical lines -- deep minima, blue vertical lines
-- shallow minima, magenta vertical lines -- flat intervals, red vertical
lines -- inflection points. Obtained using $T_{1}=0.003$, $T_{2}=0.002$,
$T_{3}=0.001$, $T_{4}=0.001$, $T_{5}=0.0005$.}

\label{s50} 
\end{figure}

\textbf{Example:} The construction is illustrated by a specific example
in which $K=10$. 
\begin{equation}
{\bf X}_{50}=\left(\begin{array}{cccccccccc}
1124 & 2091 & 2380 & 0 & 0 & 0 & 0 & 0 & 0 & 0\\
935 & 1467 & 1735 & 2162 & 0 & 0 & 0 & 0 & 0 & 0\\
1092 & 1187 & 1287 & 1440 & 1570 & 1642 & 2014 & 2311 & 2345 & 0\\
426 & 617 & 811 & 895 & 1034 & 1228 & 1604 & 1701 & 2042 & 2119
\end{array}\right).\label{eq1}
\end{equation}
The structure of the matrix in Eq. \ref{eq1} is as follows: The first
row contains the wavelengths of deep minima points. There are only
three such points and thus the rest of the row is filled with zeros.
The second row contains the wavelengths of shallow minima points.
The third row contains the wavelengths of indicators of flat intervals
and the last row contains the wavelengths of the first ten inflection
points. We display the spectrum and its features in Figure~\ref{s50}.

\section{Exact search\label{sec:exact-search} }

In this section, we introduce an algorithm that looks for a material
whose spectral signature matches \emph{exactly} the spectral signature
of a given material.

\subsection{Construction of the classification tree\label{sub:Construction-of-the}}

We design a tree structured classifier that, using characteristic
features of a spectrum, assigns it to a certain material from the
given database. Here is the outline of the construction of this classifier. 
\begin{description}
\item [{Initialization:}] As an input data, we use the extracted selected
features for all the $N$ spectra in the database. The data is organized
in four matrices $S^{sh}$, $sh=1,2,3,4$, of sizes $N\times K$.
The rows correspond to different spectra. The $K$ columns of $S^{1},\ldots,S^{4}$
contain the wavelength locations of the extracted features from all
the spectra. Specifically, the $K$ columns of $S^{1}$ contain the
locations of deep minima points of all spectra in ascending order,
the columns of $S^{2}$ contain the shallow minima points, the columns
of $S^{3}$ contain indicators of flat intervals and the columns of
$S^{4}$ contain the inflection points. We can visualize this as four
matrices as is illustrated in Fig. \ref{sheets}. 
\begin{figure}[!h]
\begin{centering}
\includegraphics[width=0.5\columnwidth]{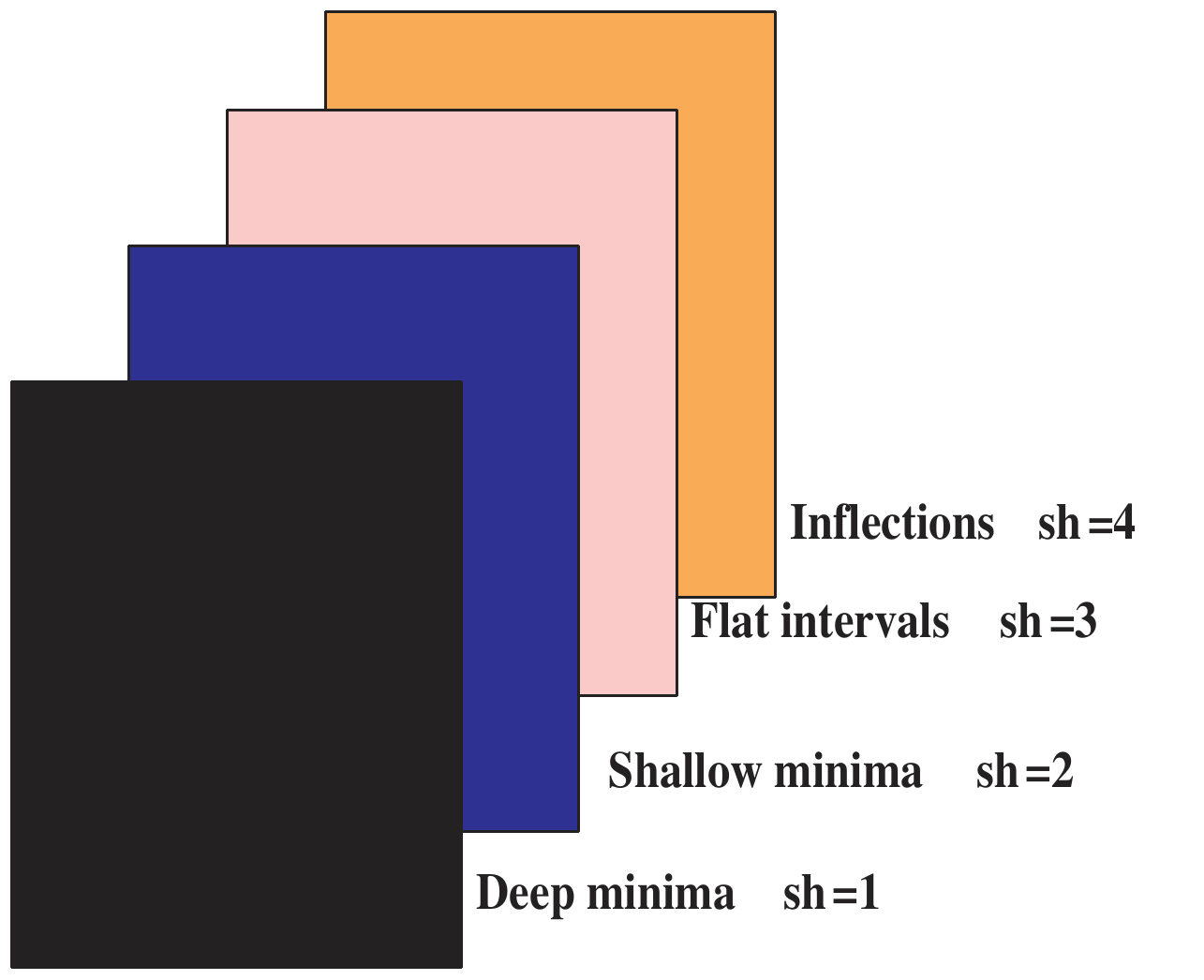} 
\par\end{centering}

\caption{The arrangement of four matrices $sh=1,2,3,4$}

\label{sheets} 
\end{figure}

\end{description}
We now construct the tree. The first node of the tree includes all
the spectra in the database. The tree construction uses a rule to
split each node to a \emph{son} node and a \emph{daughter} node. At
each split, one feature (one of the deep minima, shallow minima etc.)
is examined to see whether it distinguishes between the different
spectra. The son node is assigned with all the spectra that \emph{cannot}
be distinguished by this feature while the daughter node is assigned
with those that can. This procedure is repeated recursively for every
node until each node contains a single spectrum or all the features
have been exhausted.
\begin{description}
\item [{Split~of~the~first~node:}] We start with the matrix $\textbf{S}=S^{sh}$,
$sh=1$. We introduce a shift parameter $sf$ and initialize its value
to zero. The column $\left\{ \textbf{S}\left(k,1+sf\right)\right\} _{k=1}^{N}$
in the matrix contains the locations of the first deep minimum of
all the spectra. Let $m=\min_{k}\textbf{S}\left(k,1\right)$.

The first node in the tree is split according to the following \texttt{Split
rule:} \emph{All the spectra, such that $S^{1}(k,1)-m\leq\alpha$
($\alpha$ is a given tolerance parameter for distance comparison
whose value is determined empirically) are assigned to the son node.
The rest of the spectra (if it is not empty) is assigned to the daughter
node}. The split value $m$ is saved as $m_{1}$ as well as the parameters
$sh_{1}=sh=1$ and $sf_{1}=sf=0$.

\item [{Handling~of~the\emph{~daughter}~node:}] If this node contains
only one object then it is marked as a terminal node. Otherwise, the
node is stored for subsequent processing. The shift parameter $sf$
is left unchanged. 
\item [{Handling~of~the\emph{~son}~node:}] If this node contains only
one object, then it is marked as a \emph{terminal} node. Otherwise,
the following steps are performed. Recall that the entries in $\{\textbf{S}(k,1+sf)\}_{k\in son}$
in column number $1+sf$ are all equal to the same value $m$ (up
to $K$ samples). Therefore, if $m\neq0$, then we continue to the
next column by $sf=sf+1$. If $m=0$ or if $sf>K-1$ then we proceed
to the next matrix $S^{sh}$ by increasing $sh$ by 1 and setting
$sf=0$. This means that the deep minima features are not sufficient
for separation of the objects which belong to the \emph{son} node
and we go to the next set of features. Then, the node is subjected
to a split. However, if the \emph{son} node is a terminal node, then
the \emph{daughter} node, which shares the same parent node with the
\emph{son} node, is subjected to a split.
\item [{Split~of~the~node~number\emph{~nod}:}] We take the matrix
$\textbf{S}\stackrel{\Delta}{=}\{S^{sh}(k,1:K)\}_{k\in nod}$. Let
$m=\min_{k\in nod}\textbf{S}(k,sf+1)$. The \texttt{Split rule} is:
\emph{All the spectra, such that $\textbf{S}(k,sf+1)-m\leq\alpha$,
are assigned to the son node. The rest of the spectra (if not empty)
is assigned to the daughter node}. The split value $m$ is saved as
$m_{nod}$ as well as the parameters $sh_{nod}$ and $sf_{nod}$.
\item [{Handling~of~a~daughter~node:}] As above.
\item [{Handling~of~a~son~node:}] As above. A possible situation that
may happen for some son node, is when the parameters become $sh=4$
and $sf=K+1$. This means that the objects within this node are identical
with respect to the available set of features. In this case, the node
is marked as a terminal node. 
\end{description}
The above steps are repeated until all the nodes become terminal nodes.
Then, the tree is saved and the terminal nodes that contain more than
one material are saved separately.

We emphasize that the order in which the matrices are processed is
unimportant. Using a different processing order will produce a different
tree structure, however, the outcome of the identification will be
\emph{the same}. This is due to the fact that every pair of different
spectra can be distinguished by a set of features. Going over this
set in any order will always distinguish between the spectra once
all the distinguishing features have been visited.

\subsection{Identification of a spectrum}
\begin{description}
\item [{Input:}] A raw spectrum $\#n$ from the given database and the
constructed tree. 
\item [{Extraction~of~features:}] According to the procedure described
in Section \ref{Features}. The features are gathered into the matrix
${\bf X}_{n}$ of size $4\times K$ as displayed in Eq. \ref{eq1}. 
\end{description}
The identification procedure traverses the tree starting from the
root.
\begin{description}
\item [{Traversal~decision~at~the~first~node:}] Initially, the object
lies in node 1, whose split value is $m=m_{1}$. In the root node
we calculate the difference $d=X(1,1)-m$. If $d\leq\alpha$ then
the traversal continues to the \emph{son}  node, otherwise -- it continues
to the \emph{daughter}  node. If the destination node is a terminal
then we already have the answer.
\item [{Traversal~decision~at~an~arbitrary~node\emph{~nod}:}] Assume
that the traversal arrived at a non-terminal node \emph{nod}. Assume
the parameters at this node are $sh=sh_{nod}$, $sf=sf_{nod}$ and
the split value is $m=m_{nod}$. We calculate the difference $d=X(sh,sf+1)-m$.
If $d\leq\alpha$ then the traversal continues to the \emph{son}  node,
otherwise -- it continues to the \emph{daughter} node. 
\end{description}
The traversal of the tree continues until the object reaches a terminal
node. The output is an index of the identified spectrum and the features
that were actually involved in its identification. If the spectrum
belongs to a group of identical spectra in the database, then the
output of the classifier consists of the indices of the whole group.

\subsection{Illustration by example of the tree construction}

In the following a step-by-step of the tree construction in Section
\ref{sub:Construction-of-the} is illustrated using an example. From
the given database (173 entries of spectra where each consists of
2001 wavelengths), we select the eight spectra which are given in
Table \ref{tab:Example_spectra}. We let $K=10$ and $\alpha=10$.

\begin{table}[!h]
\caption{A set of spectra used for a step-by-step illustration of the tree
construction in Section \ref{sub:Construction-of-the}.\label{tab:Example_spectra}}

\centering{}%
\begin{tabular}{|c|c|c|c|c|c|c|c|c|}
\hline 
Serial number & 1 & 2 & 3 & 4 & 5 & 6 & 7 & 8\tabularnewline
\hline 
\hline 
Row number in the database  & 173 & 4 & 27 & 87 & 43 & 163 & 150 & 156\tabularnewline
\hline 
\end{tabular}
\end{table}

\begin{figure}[!h]
\includegraphics[width=1\columnwidth]{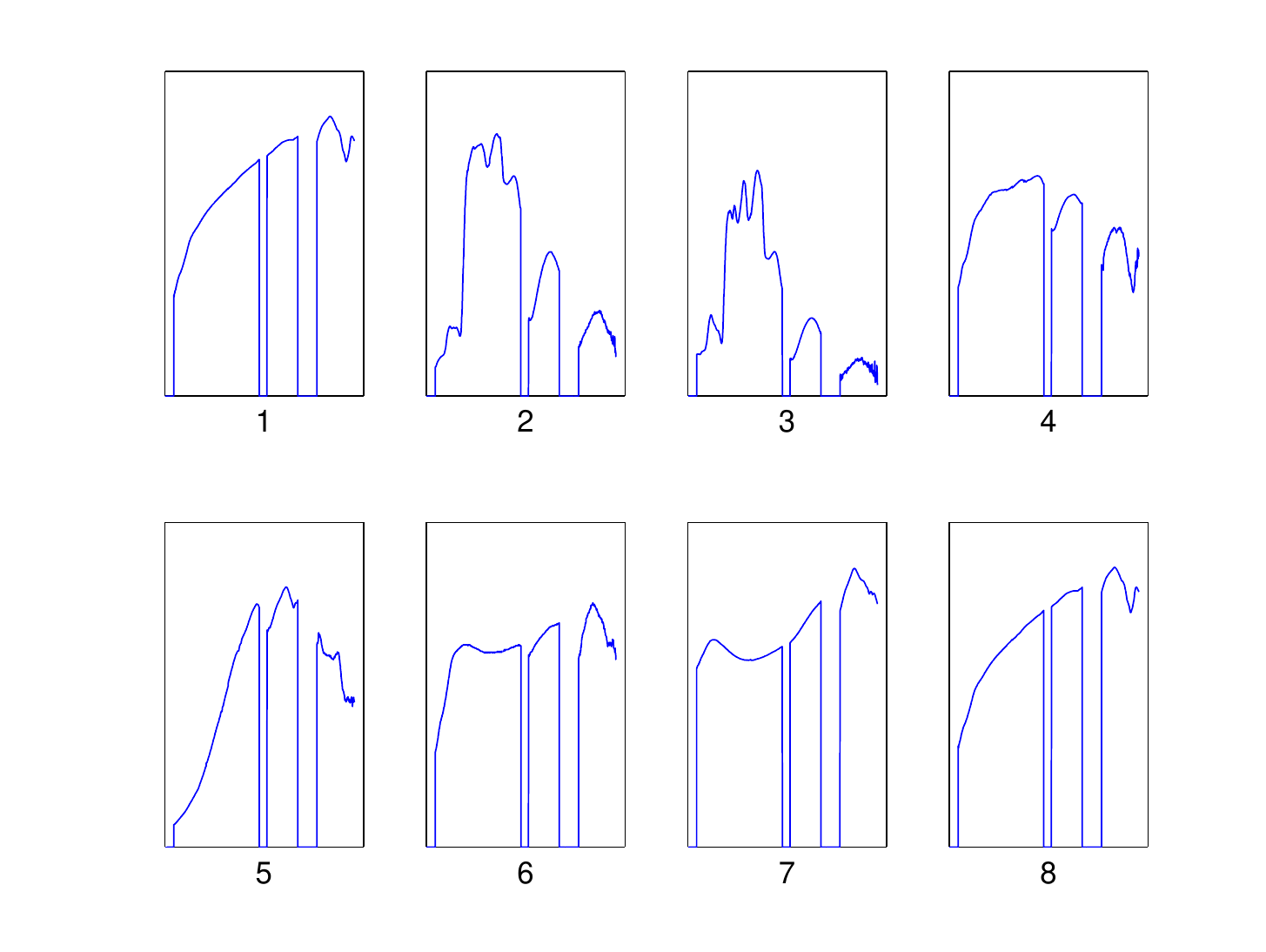}

\caption{Spectra $\#173$,$\#4$, $\#27$, $\#87$, $\#43$, $\#163$, $\#163$,
$\#150$, $\#156$ and their corresponding serial number from table
\ref{tab:Example_spectra}.}

\label{s50i} 
\end{figure}

Their corresponding graphs are given in Fig. \ref{s50i}. 

The first node of tree consists of the eight spectra. The matrix in
Eq. \ref{ex1} is the ten deep minima of each spectra. Column 1 is
the serial number of the spectra in the database as specified in Table
\ref{tab:Example_spectra}.

\begin{equation}
S^{1}=\left(\begin{array}{rrrrrrrrrrr}
1 & 2311 & 0 & 0 & 0 & 0 & 0 & 0 & 0 & 0 & 0\\
2 & 672 & 976 & 1191 & 1450 & 0 & 0 & 0 & 0 & 0 & 0\\
3 & 673 & 794 & 855 & 976 & 1192 & 1452 & 0 & 0 & 0 & 0\\
4 & 1446 & 2153 & 0 & 0 & 0 & 0 & 0 & 0 & 0 & 0\\
5 & 1728 & 2163 & 2308 & 2353 & 0 & 0 & 0 & 0 & 0 & 0\\
6 & 2380 & 0 & 0 & 0 & 0 & 0 & 0 & 0 & 0 & 0\\
7 & 2380 & 0 & 0 & 0 & 0 & 0 & 0 & 0 & 0 & 0\\
8 & 2311 & 0 & 0 & 0 & 0 & 0 & 0 & 0 & 0 & 0
\end{array}\right).\label{ex1}
\end{equation}

\begin{figure}[!h]
\begin{centering}
\includegraphics[width=0.7\columnwidth]{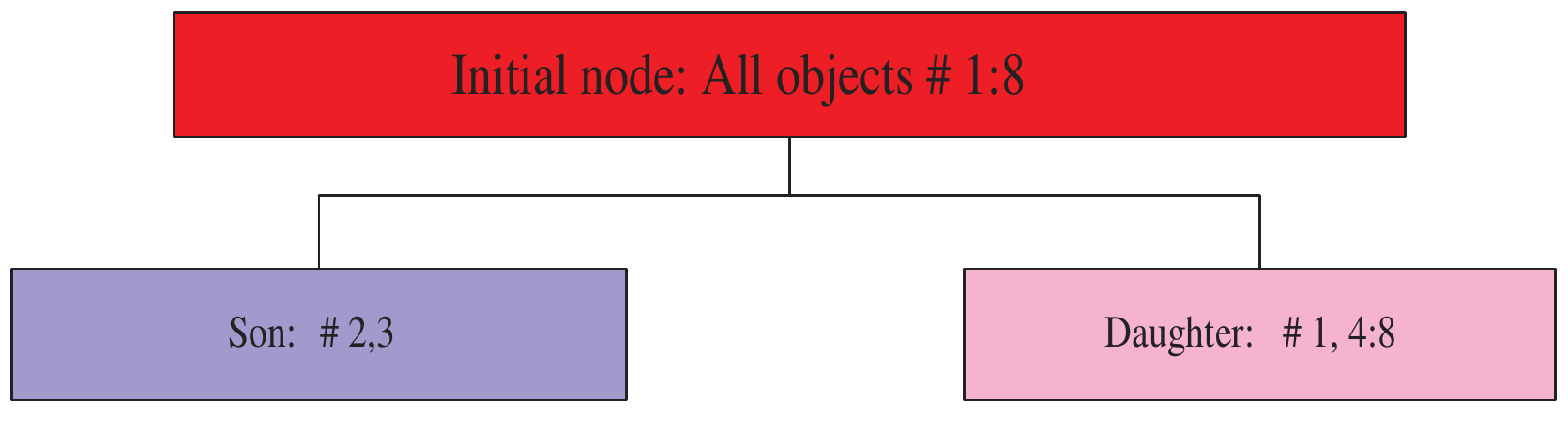} 
\par\end{centering}

\caption{The tree after the first split.}

\label{f1} 
\end{figure}

We look at the second column of matrix $S^{1}$ (Eq. \ref{ex1}).
We look for the minimum value in this column and we get that it is
$m=672$. The distances from feature (deep-minima) locations 672,
673 from $m$ are less or equal to $\alpha=10$. Therefore, we assign
rows 2 and 3 to the son node while the others are assigned to the
daughter node (see Fig. \ref{f1}).

The matrix $S^{1}$ for the son is set to: 
\begin{equation}
S_{son}^{1}=\left(\begin{array}{rrrrrrrrrrrl}
2 & 672 & 976 & 1191 & 1450 & 0 & 0 & 0 & 0 & 0 & 0\\
3 & 673 & 794 & 855 & 976 & 1192 & 1452 & 0 & 0 & 0 & 0
\end{array}\right).\label{ex2}
\end{equation}

The matrix $S^{1}$ for the daughter is set to: 
\begin{equation}
S_{daug}^{1}=\left(\begin{array}{rrrrrrrrrrr}
1 & 2311 & 0 & 0 & 0 & 0 & 0 & 0 & 0 & 0 & 0\\
4 & 1446 & 2153 & 0 & 0 & 0 & 0 & 0 & 0 & 0 & 0\\
5 & 1728 & 2163 & 2308 & 2353 & 0 & 0 & 0 & 0 & 0 & 0\\
6 & 2380 & 0 & 0 & 0 & 0 & 0 & 0 & 0 & 0 & 0\\
7 & 2380 & 0 & 0 & 0 & 0 & 0 & 0 & 0 & 0 & 0\\
8 & 2311 & 0 & 0 & 0 & 0 & 0 & 0 & 0 & 0 & 0
\end{array}\right).\label{ex3}
\end{equation}

We continue with the third column of $S_{son}^{1}$ (Eq. \ref{ex2}). 

\begin{figure}[!h]
\begin{centering}
\includegraphics[width=0.7\columnwidth]{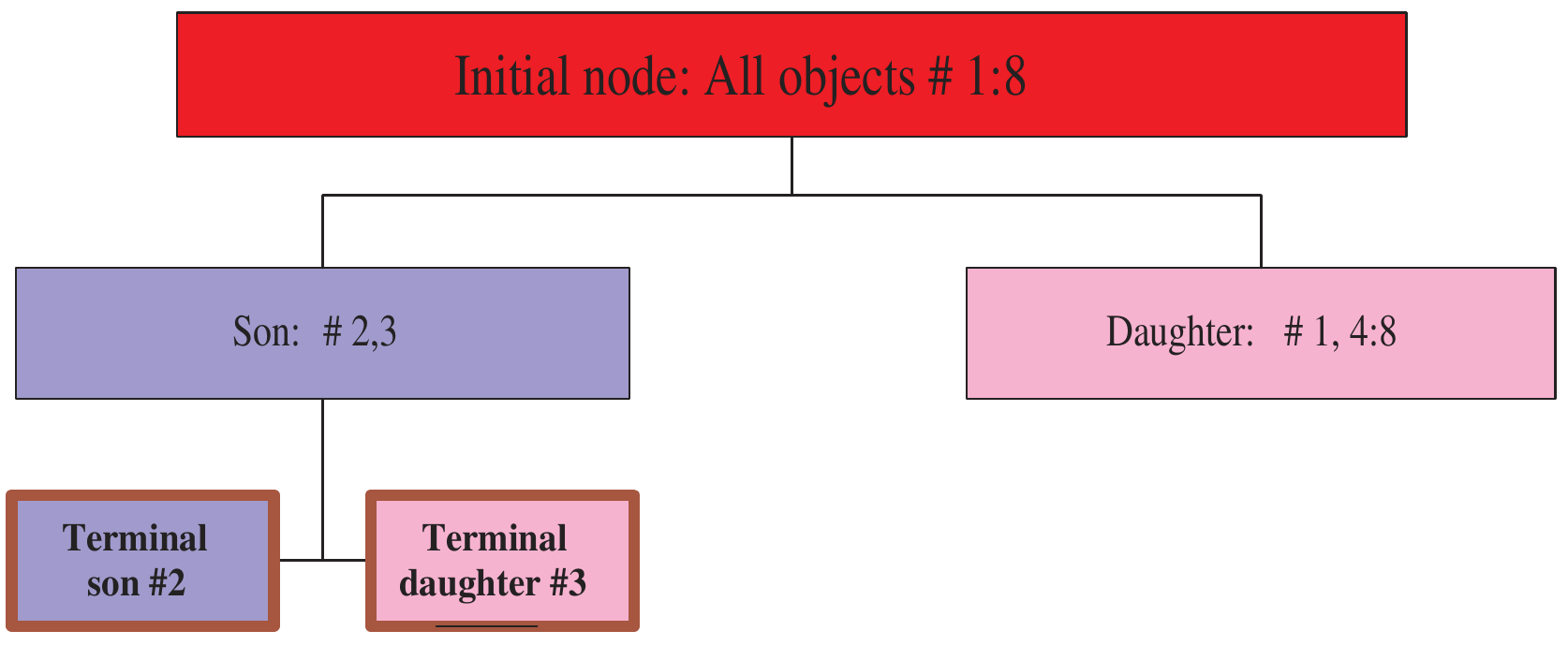} 
\par\end{centering}

\caption{The tree after the second split.}

\label{f2} 
\end{figure}

We get that the $m=794$. Entry 3 in $S_{son}^{1}$ is a terminal
son, and entry 2 in $S_{son}^{1}$ is a terminal daughter. The current
structure of the tree is given in Fig. \ref{f2}.

\begin{figure}[!h]
\begin{centering}
\includegraphics[width=0.7\columnwidth]{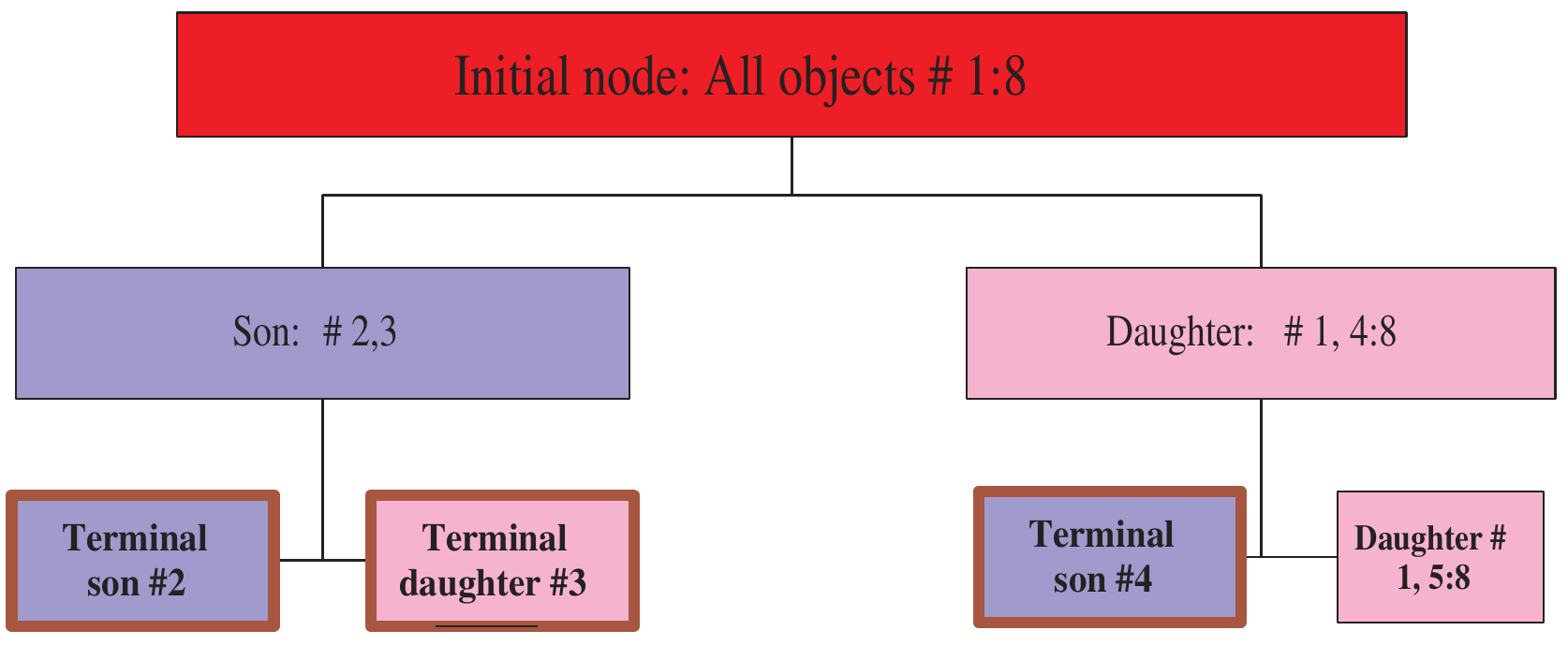} 
\par\end{centering}

\caption{The tree after the third split.}

\label{f3} 
\end{figure}

We now proceed to $S_{daug}^{1}$ (Eq. \ref{ex3}). Calculating $m$
for the second column yields $m=1446$. Therefore, entry 4 is a son
terminal node, and the rest is daughter node. The resulting structure
of the tree is given in Fig. \ref{f3}.

The matrix $S^{1}$ (Eq. \ref{ex3}) for the daughter becomes: 
\begin{equation}
S_{daug}^{1}=\left(\begin{array}{rrrrrrrrrrr}
1 & 2311 & 0 & 0 & 0 & 0 & 0 & 0 & 0 & 0 & 0\\
5 & 1728 & 2163 & 2308 & 2353 & 0 & 0 & 0 & 0 & 0 & 0\\
6 & 2380 & 0 & 0 & 0 & 0 & 0 & 0 & 0 & 0 & 0\\
7 & 2380 & 0 & 0 & 0 & 0 & 0 & 0 & 0 & 0 & 0\\
8 & 2311 & 0 & 0 & 0 & 0 & 0 & 0 & 0 & 0 & 0
\end{array}\right).\label{ex4}
\end{equation}

Calculating $m$ for the first column of $S_{daug}^{1}$ produces
$m=1728$. Therefore, the current node is split into a son terminal
node, which contains material 5, and a daughter node which contains
the rest. The tree structure that has been constructed so far is given
in Fig. \ref{f4}:

\begin{figure}[!h]
\begin{centering}
\includegraphics[width=0.7\columnwidth]{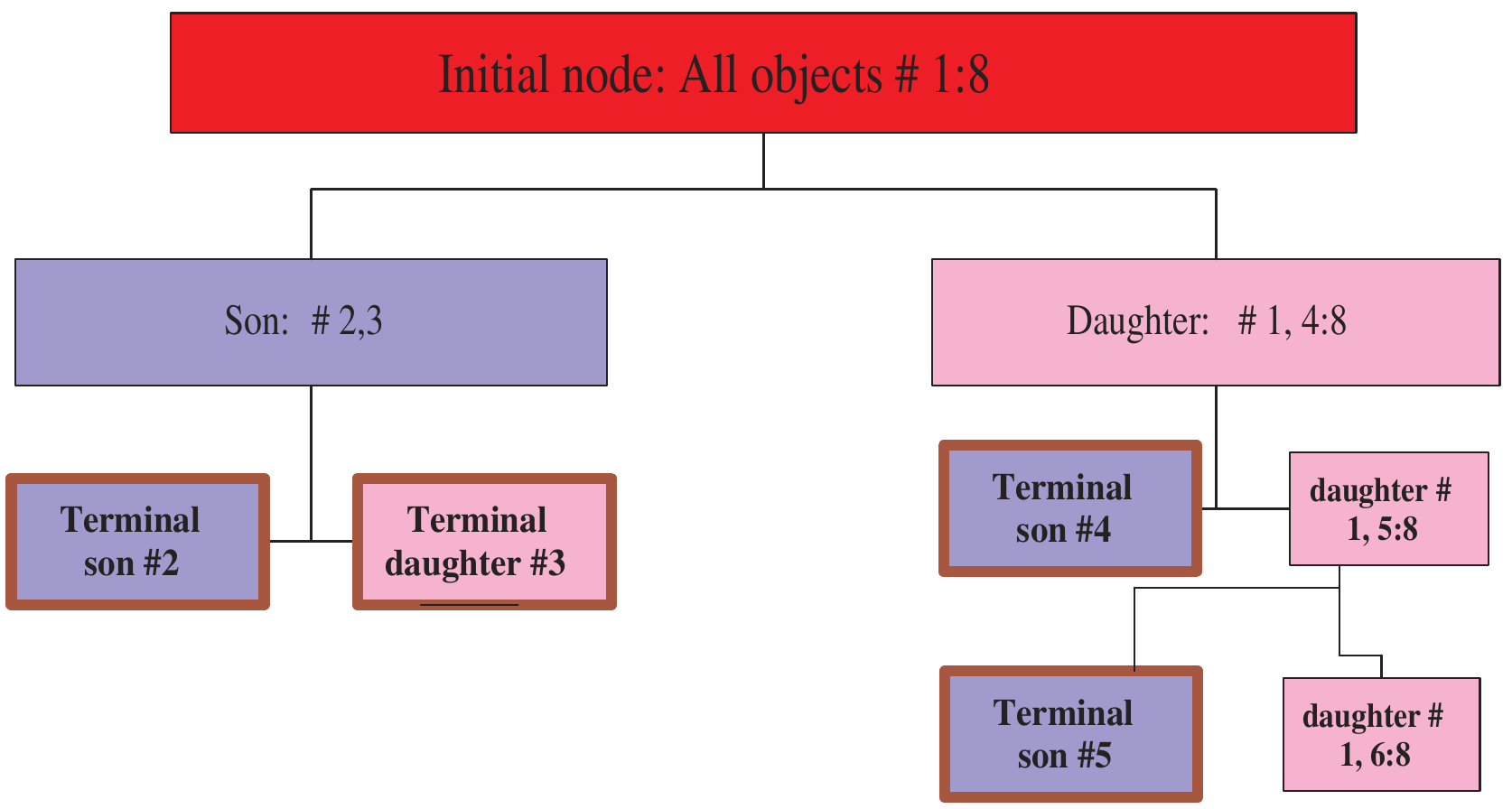} 
\par\end{centering}

\caption{The tree after the fourth split.}

\label{f4} 
\end{figure}

The matrix $S^{1}$ for the daughter (Eq. \ref{ex4}) is: 
\begin{equation}
S_{daug}^{1}=\left(\begin{array}{rrrrrrrrrrr}
1 & 2311 & 0 & 0 & 0 & 0 & 0 & 0 & 0 & 0 & 0\\
6 & 2380 & 0 & 0 & 0 & 0 & 0 & 0 & 0 & 0 & 0\\
7 & 2380 & 0 & 0 & 0 & 0 & 0 & 0 & 0 & 0 & 0\\
8 & 2311 & 0 & 0 & 0 & 0 & 0 & 0 & 0 & 0 & 0
\end{array}\right).\label{ex5}
\end{equation}

Te value of $m$ for the first column is $m=2311$. Therefore, the
node is split into a son node which contains materials 1 and 8 and
a daughter node which contains materials 6 and 7. The resulting structure
of the tree is given in Fig. \ref{f5}:

\begin{figure}[!h]
\begin{centering}
\includegraphics[width=0.7\columnwidth]{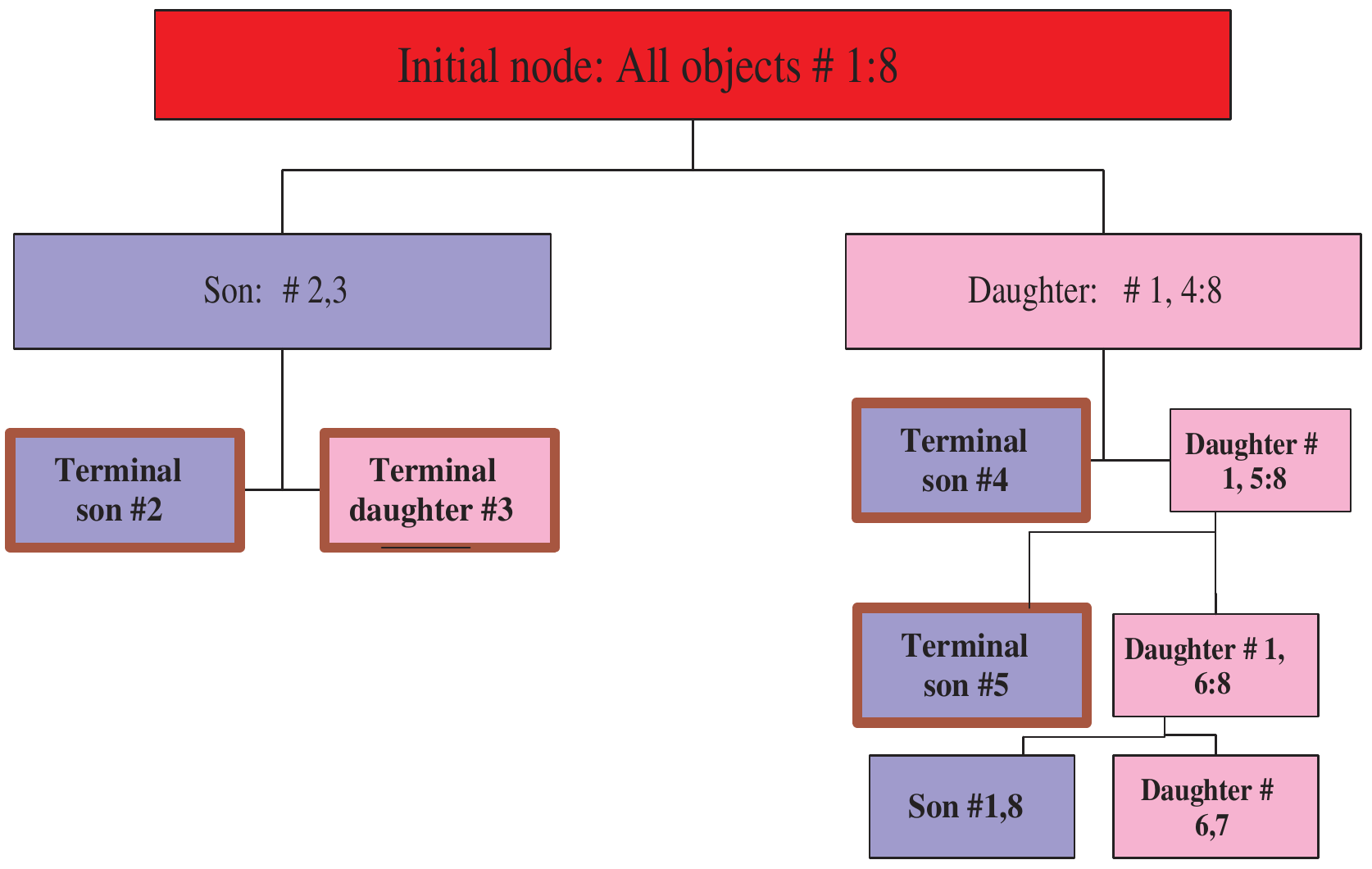} 
\par\end{centering}

\caption{The tree after the fifth split.}

\label{f5} 
\end{figure}

The matrices $S^{1}$ for son and daughter are: 
\begin{equation}
S_{son}^{1}=\left(\begin{array}{rrrrrrrrrrr}
1 & 2311 & 0 & 0 & 0 & 0 & 0 & 0 & 0 & 0 & 0\\
8 & 2311 & 0 & 0 & 0 & 0 & 0 & 0 & 0 & 0 & 0
\end{array}\right)\label{ex6}
\end{equation}

and 
\begin{equation}
S_{daug}^{1}=\left(\begin{array}{rrrrrrrrrrr}
6 & 2380 & 0 & 0 & 0 & 0 & 0 & 0 & 0 & 0 & 0\\
7 & 2380 & 0 & 0 & 0 & 0 & 0 & 0 & 0 & 0 & 0
\end{array}\right).\label{ex7}
\end{equation}

We check the third column of the son in Eq. \ref{ex6}. Since it contains
zero values, we go to matrix $S_{son}^{2}$.

\begin{equation}
S_{son}^{2}=\left(\begin{array}{rrrrrrrrrrr}
1 & 1698 & 0 & 0 & 0 & 0 & 0 & 0 & 0 & 0 & 0\\
8 & 1698 & 0 & 0 & 0 & 0 & 0 & 0 & 0 & 0 & 0
\end{array}\right).\label{ex8}
\end{equation}
 Both entries are assigned to son and we go to matrix $S_{son}^{3}$.

\begin{equation}
S_{son}^{3}=\left(\begin{array}{rrrrrrrrrrr}
1 & 622 & 1040 & 1604 & 2104 & 0 & 0 & 0 & 0 & 0 & 0\\
8 & 622 & 1040 & 1604 & 2104 & 0 & 0 & 0 & 0 & 0 & 0
\end{array}\right).\label{ex9}
\end{equation}

The algorithm continues to check the columns in an attempt to separate
between entries. Finally, it reaches matrix $S^{4}$ which is given
by

\begin{equation}
S_{son}^{4}=\left(\begin{array}{rrrrrrrrrrr}
1 & 470 & 1750 & 2281 & 0 & 0 & 0 & 0 & 0 & 0 & 0\\
8 & 470 & 1750 & 2281 & 0 & 0 & 0 & 0 & 0 & 0 & 0
\end{array}\right).\label{ex10}
\end{equation}
 Since we arrived at the last feature, this node appears to be unsplittable.
Consequently, the node is marked as terminal and we continue to the
remaining daughter. The current structure of the tree is given in
Fig. \ref{f6}:

\begin{figure}[!h]
\begin{centering}
\includegraphics[width=0.7\columnwidth]{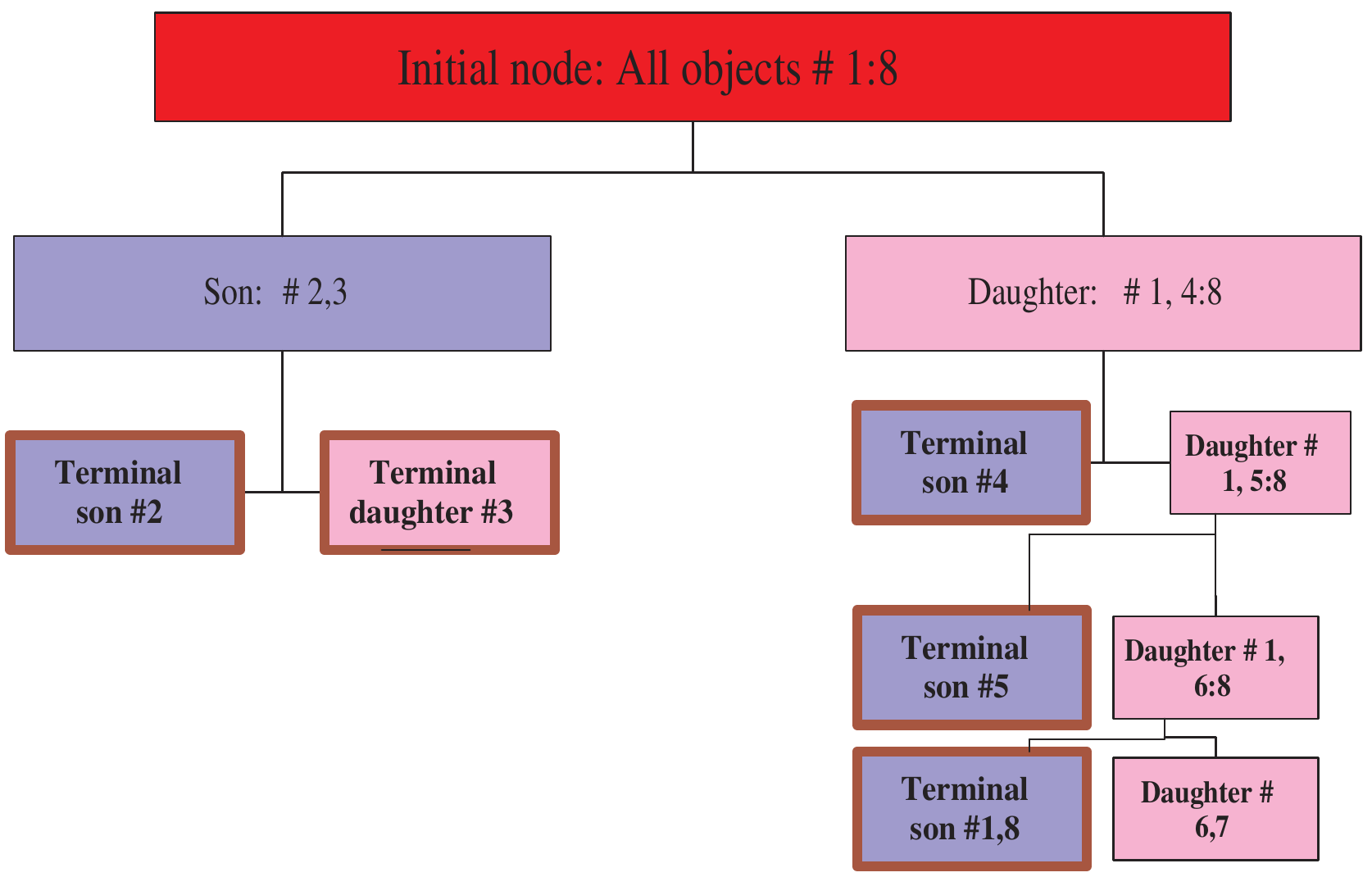} 
\par\end{centering}

\caption{The tree after the sixth split.}

\label{f6} 
\end{figure}

We process the daughter in Eq. \ref{ex7}. Here it is again

\begin{equation}
S_{daug}^{1}=\left(\begin{array}{rrrrrrrrrrr}
6 & 2380 & 0 & 0 & 0 & 0 & 0 & 0 & 0 & 0 & 0\\
7 & 2380 & 0 & 0 & 0 & 0 & 0 & 0 & 0 & 0 & 0
\end{array}\right).\label{ex11}
\end{equation}
 The second column contains identical values. Therefore, both materials
are assigned to a son node and we go to $S_{son}^{2}$, which yields
\begin{equation}
S_{son}^{2}=\left(\begin{array}{rrrrrrrrrrr}
6 & 958 & 1113 & 2313 & 0 & 0 & 0 & 0 & 0 & 0 & 0\\
7 & 968 & 1011 & 2311 & 0 & 0 & 0 & 0 & 0 & 0 & 0
\end{array}\right).\label{ex12}
\end{equation}
 The entries in the second column are identical up to the feature
distance tolerance $\alpha$, therefore, they are both assigned to
a son. Next, we check the third column. The distance between the entries
is $\alpha=10$, therefore, entry 7 is a son terminal-node and entry
6 is a daughter terminal-node. However, because many features of these
two spectra are similar, it is most probable that these spectra are
associated with the same material. The final tree is given in Fig.
\ref{fig:f_final}:

\begin{figure}[!h]
\begin{centering}
\includegraphics[width=0.8\columnwidth]{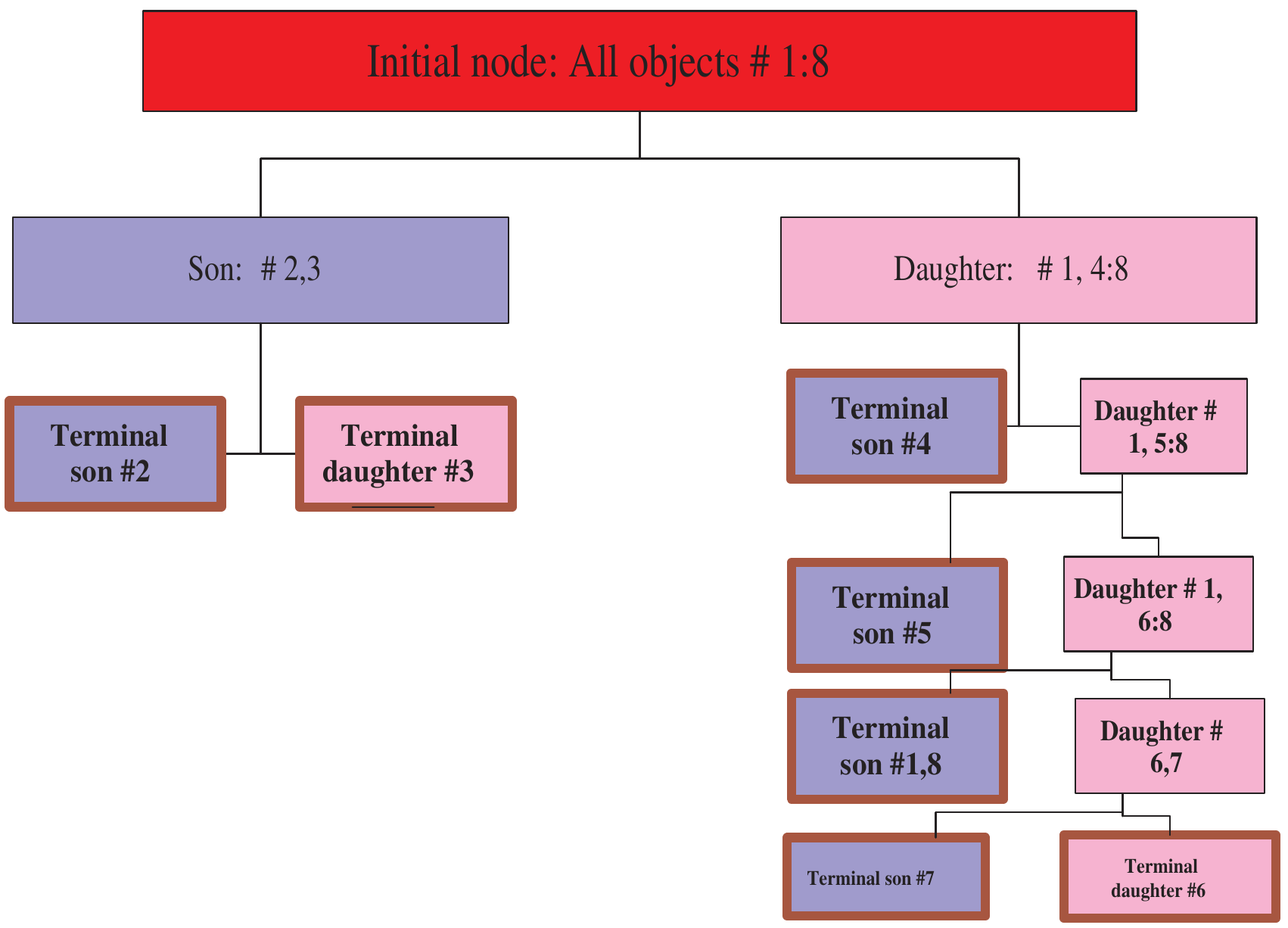} 
\par\end{centering}

\caption{The constructed tree of the step-by-step example }

\label{fig:f_final}
\end{figure}

\subsection{Experimental results}

As mentioned above, the spectra database that we use contains the
absorption values of materials with 2001 wavelengths. We have 173
vectors which are composed of 2001 wavelengths. There are seven pairs
of identical spectra in the given database: {[}160 161{]}, {[}154
172{]}, {[}152 170{]}, {[}111 165{]}, {[}112 164{]}, {[}153 171{]},
{[}156 173{]}. 

The thresholds that we use are: $T_{1}=0.003$, $T_{2}=0.002$, $T_{3}=0.001$,
$T_{4}=0.001$, $T_{5}=0.0005$. $\mu=5$ - length of flat interval,
$\xi-$ length of thinning interval, $d=25$ - merging threshold,
$\alpha=10$ - distance tolerance parameter for exact search comparisons
(for the definition of all threshold parameters, the reader is referred
to Section \ref{sub:Finding-the-locations}. The maximal number of
sought after features per feature-type was set to $K=10$.

\begin{figure}[!h]
\begin{centering}
\includegraphics[width=1\columnwidth]{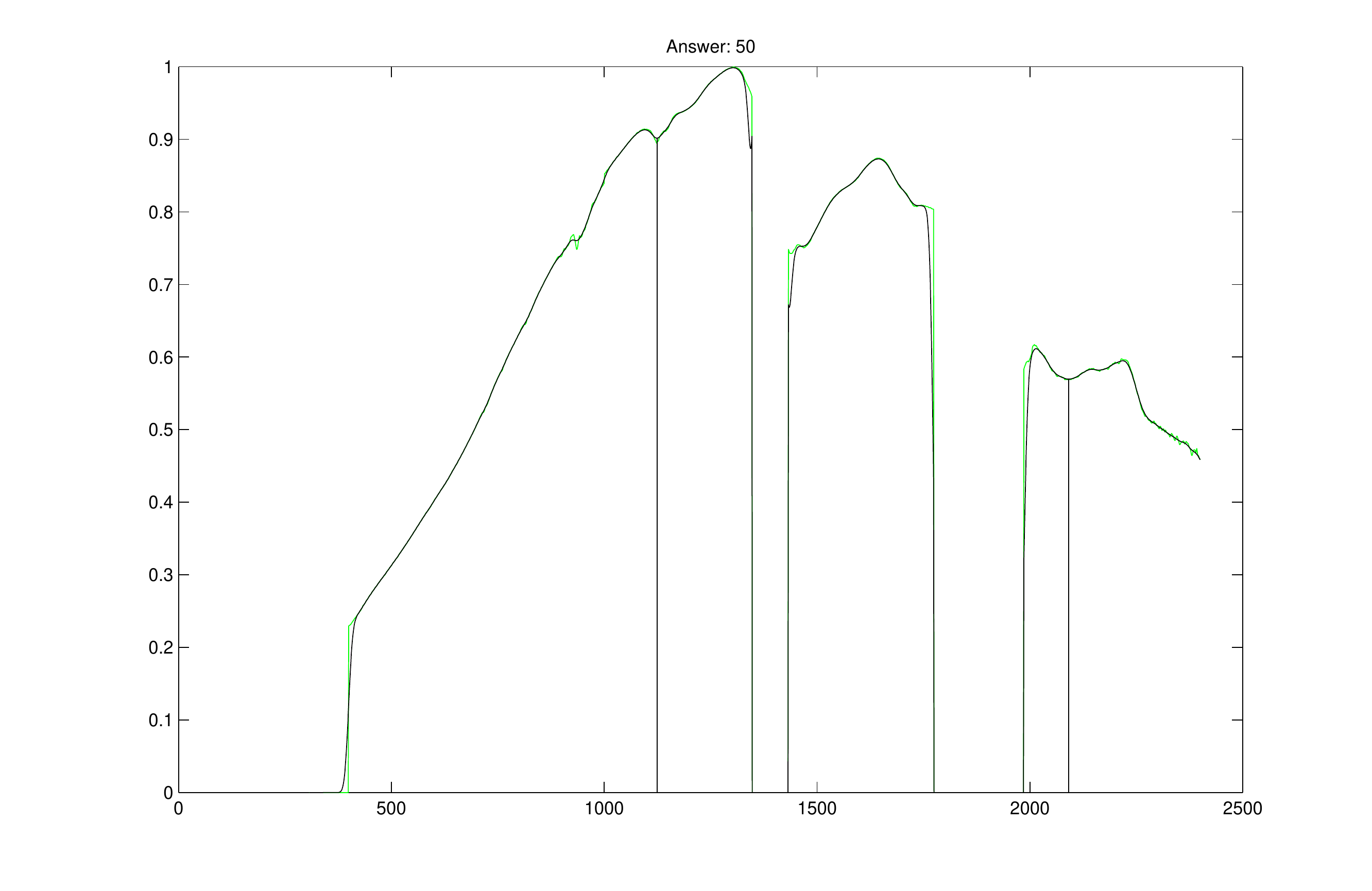}
\par\end{centering}

\caption{Identified spectrum $\#50$. Only two deep minima that are located
at wavelengths $1124nm$ and $2091nm$ (see Eq. \ref{eq1}) were used
for its identification.}

\label{s50ii} 
\end{figure}

We display a few outputs from the identification process. Figure \ref{s50ii}
demonstrates the identification of material $\#50$. The identification
was achieved by merely using the two deep minima that are located
at wavelengths $1124nm$ and $2091nm$.

\begin{figure}[!h]
\begin{centering}
\includegraphics[width=1\columnwidth]{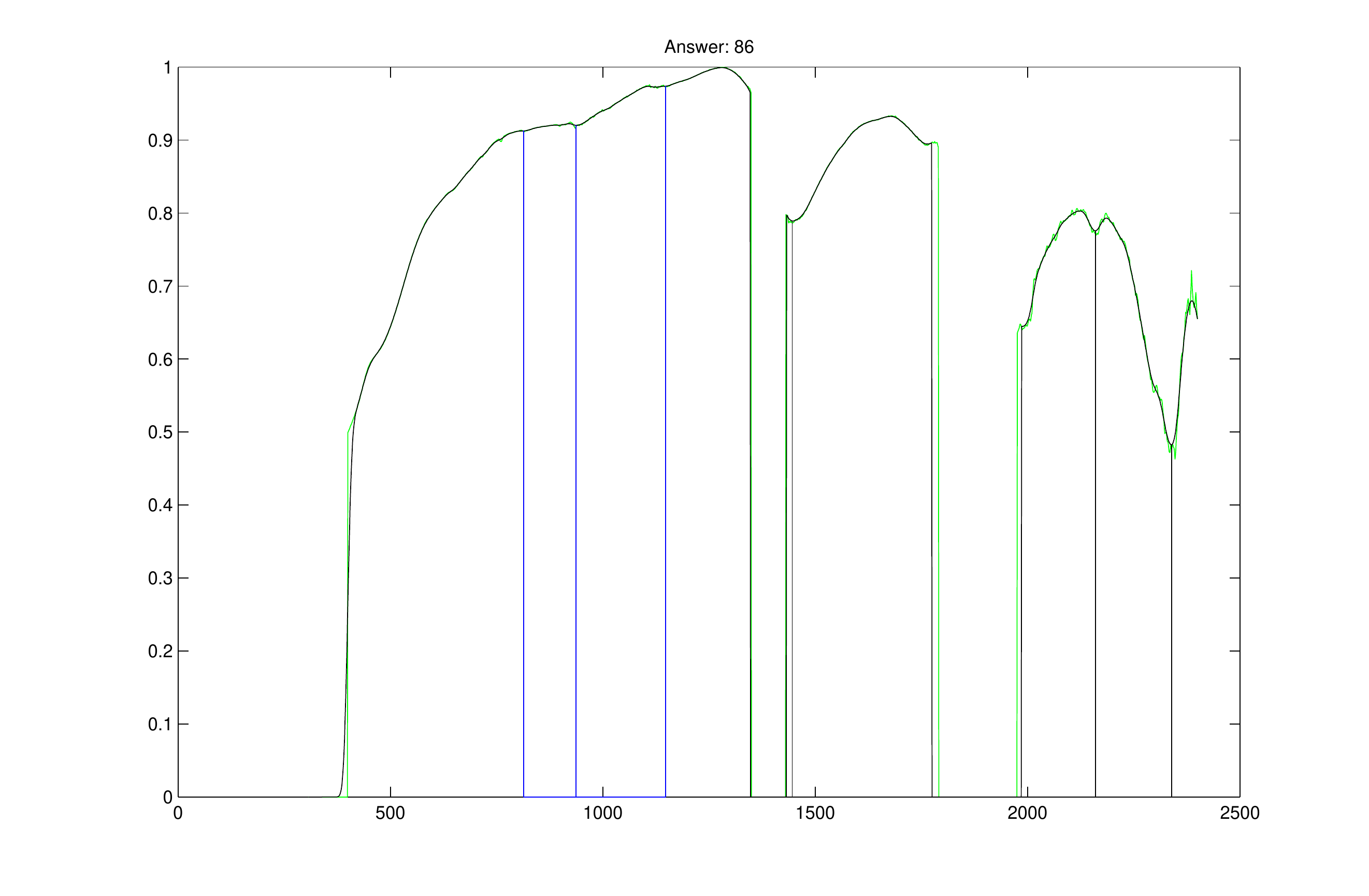} 
\par\end{centering}

\caption{Identified spectrum $\#86$. Three deep and three shallow minima form
its signature.}

\label{s86i} 
\end{figure}

\begin{figure}[!h]
\begin{centering}
\includegraphics[width=1\columnwidth]{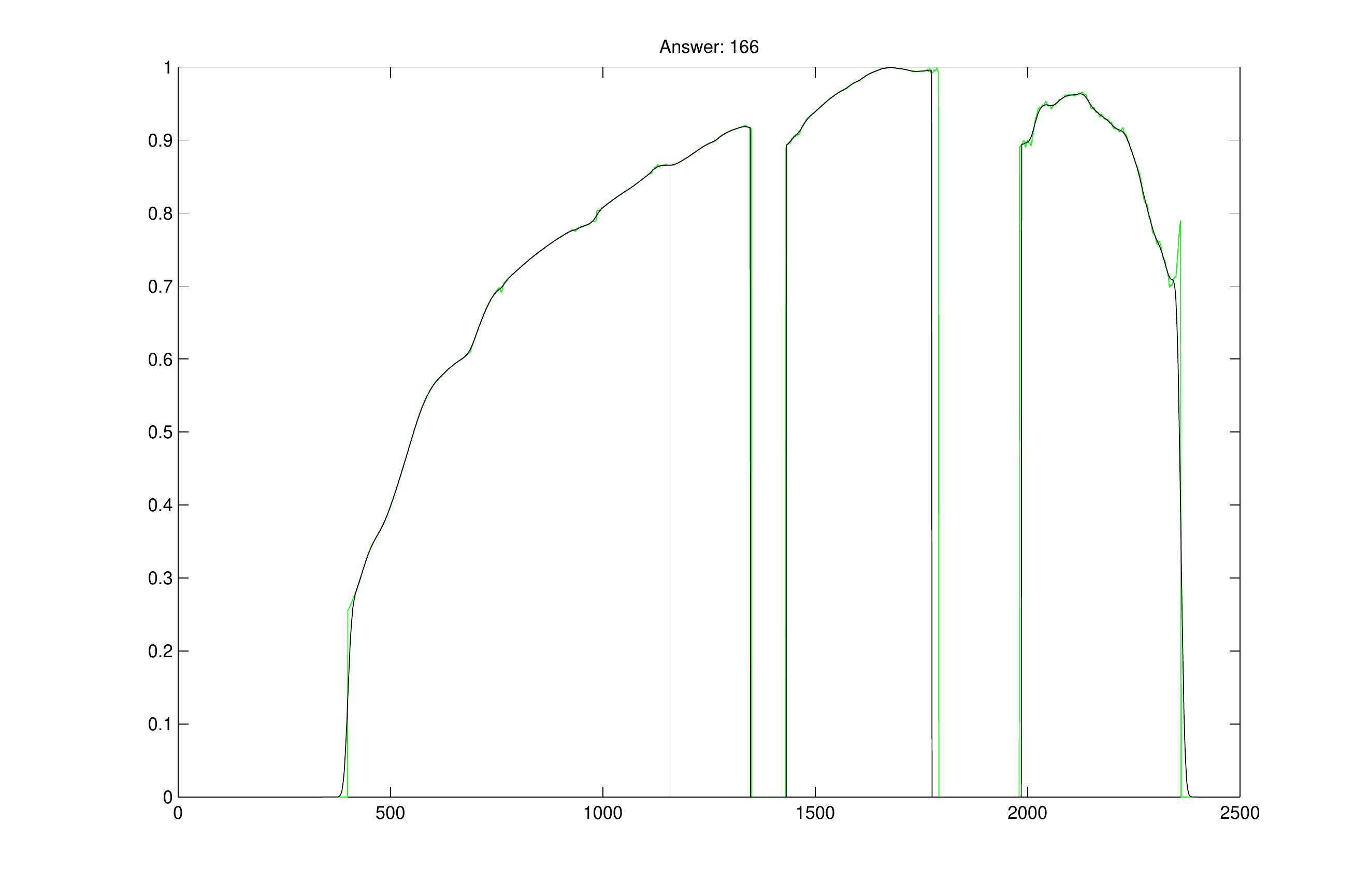} 
\par\end{centering}

\caption{Identified spectrum $\#166$. Only one shallow minimum characterizes
this spectrum.}

\label{s166i} 
\end{figure}

\begin{figure}[!h]
\begin{centering}
\includegraphics[width=1\columnwidth]{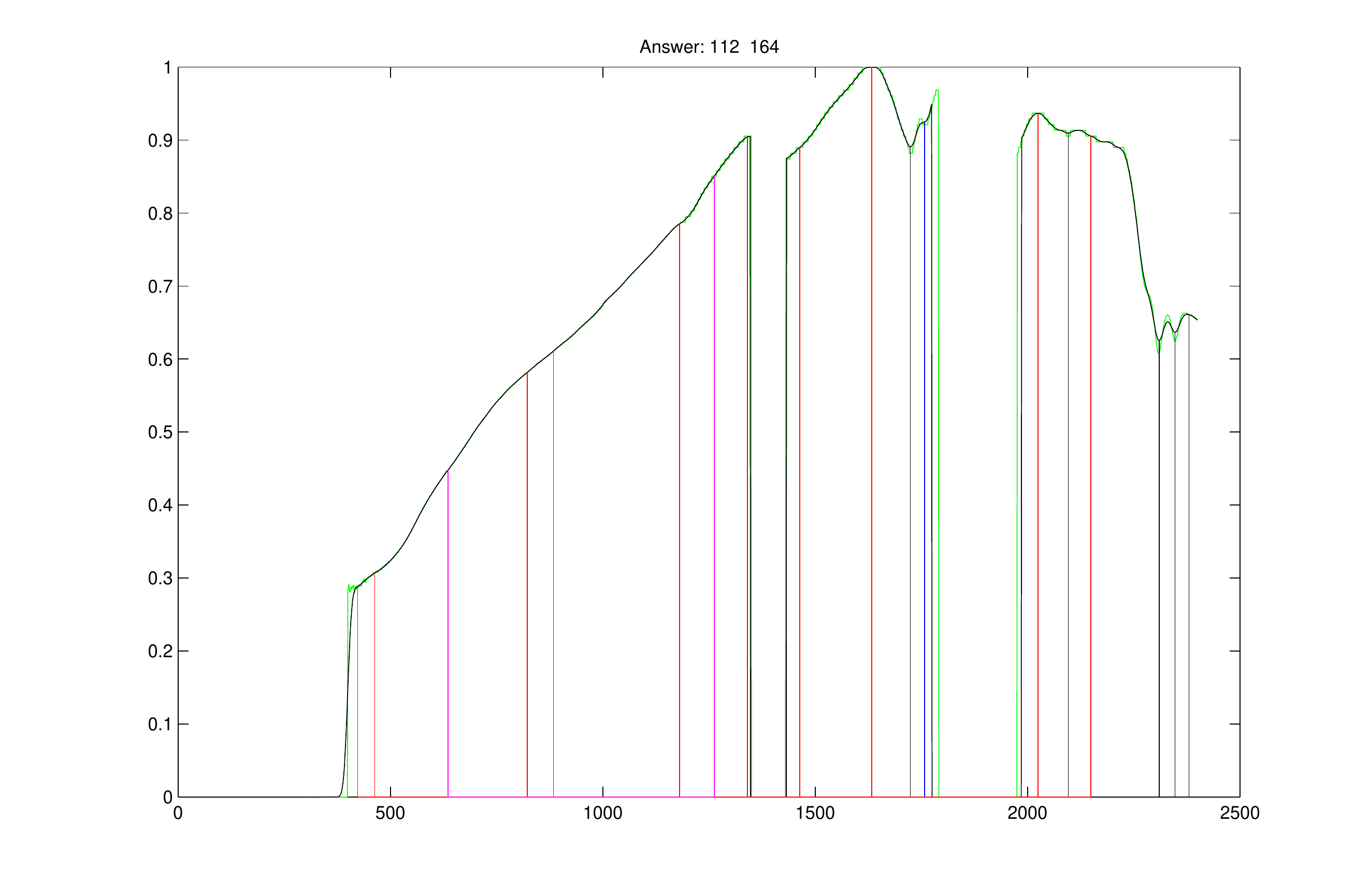} 
\par\end{centering}

\caption{A pair of identical spectra $\#112$ and $\#164$. The classifier
used all available features in an attempt to distinguish between these
two spectra.}

\label{s112_164i} 
\end{figure}

Figure \ref{s86i} depicts the spectrum graph of material $\#86$.
Its signature consists of three deep minima and three shallow minima.
The spectral signature of material $\#166$ is includes only one shallow
minima as it is illustrated in Fig. \ref{s166i}. The spectra graphs
of materials $\#112$ and $\#164$ are depicted in Fig. \ref{s112_164i}.
This materials are identical and the identification process succeeded
not to distinguish between them after using \emph{all} their spectral
features during the identification process. 

\section{Approximate search}

\label{Approximate_identification}

\subsection{Introduction}

In this section we describe a search scheme for a specific spectrum
in a given database that produces spectra that are \emph{similar}
to the presented spectrum. We call this scheme \emph{approximate search}.
It uses the same characteristic features that were described in Section
\ref{sec:exact-search} - the exact search scheme. The \textbf{new}
classifier is tailored to find all spectra in the database that have
some common features with the presented spectrum (up to a predefined
tolerance interval). There are no limitations on the number of spectra
in the database that the classifier can handle. This classifier is
especially useful when applied to noisy measurements. It can also
be utilized to perform unmixing where the measurements of endmembers
are corrupted with noise.

\subsection{The approximate search scheme}

The scheme consists of three operational blocks: 
\begin{description}
\item [{1.}] \textbf{Feature extraction block} - produces a diverse set
of geometric features from the spectra in the given reference database.
These features are related to the physical properties of their corresponding
materials. 
\item [{2.}] \textbf{Feature distribution table construction block} - provides
a table that describes the distribution of the selected features among
the spectra in the reference database. It provides a fast search for
the spectra, which possess the prescribed features. 
\item [{3.}] \textbf{Identification block} - extracts selected features
from the presented spectrum by using the feature distribution table
(block 2). Then, it finds all the spectra in the given database that
have some features \emph{close} to the features of the presented spectrum.
The closeness is measured using a given parameter which will be defined
later. This block produces as output a set of spectra that are similar
to the presented spectrum. The algorithm can handle spectra that do
not necessarily belong to the reference database. 
\end{description}
We demonstrate the capabilities and the performance of the algorithm
using two databases: The first, which is denoted by \emph{DB1}, is
the same database that was used in Section \ref{sec:exact-search}.
The second database, which is denoted by \emph{DB2}, consists of 90
spectra whose structure is similar to the structure of spectra from
\emph{DB1}. To validate the performance of the approximate search
algorithm, three types of experiments were conducted: 
\begin{enumerate}
\item The reference database is \emph{DB1} and the tested spectra were taken
from \emph{DB1}. 
\item The reference database is \emph{DB2} and the tested spectra were taken
from \emph{DB2}. 
\item The reference database is \emph{DB1} and the tested spectra were taken
from \emph{DB2}. 
\end{enumerate}
Selection of features (block 1) is done as in Section \ref{Features}.

\subsection{Construction of the feature distribution table (block 2)\label{ss42} }

In the following, we describe the construction of the table that describes
the distribution of the selected characteristic features among the
spectra in the reference database. 
\begin{description}
\item [{Organization~of~the~input~data:}] As an input data, we use
the extracted characteristic features of all $N$ spectra in the reference
database (see Section \ref{sec:exact-search}). The data is organized
in a matrix $\textbf{S}=\left(s_{ij}\right)$ of size $N\times4K$,
$1\le i\le N,\,1\le j\le4K$. The rows correspond to different spectra
($N$) in the reference database. The initial $K$ columns in $\textbf{S}$
contain the locations of deep minima points in ascending order of
all spectra. If, for example, spectrum $\#\, i_{0}$ has only $m<K$
deep minima points, then, the positions $s_{i_{0},j},m<j\le K$ are
filled with zeros. Recall that the locations of all the features do
not exceed 2500 in \emph{DB1} and \emph{DB2}. This number corresponds
to the highest wavelength at which the spectra was captured. We make
use of this number to distinguish between different features in $S$.
The following $K$ columns are assigned to locations of shallow minima
and are organized similarly to the previous $K$ columns. To distinguish
between the shallow minima and deep minima, each sample is increased
by adding $3000$. Column $2K+1$ till $3K$ contain indicators of
flat intervals that are increased by adding $6000$, and the last
$K$ columns, i.e. columns $3K+1$ till $4K$, contain the locations
of inflection points that are increased by adding 9000.

In the all the coming figures, the $x$ axis starts at 400. This number
corresponds to the minimal wavelength at which the spectra were captured.

Figure \ref{spe6} displays the spectra $\#36-\#41$ from \emph{DB1}.
Tables \ref{T1}-\ref{T4} show the wavelength locations of the deep
minima, shallow minima, flat interval indicators and inflection points,
respectively. In these tables, $K$ was taken to be $10$.

\begin{figure}[!h]
\begin{centering}
\includegraphics[width=1\columnwidth]{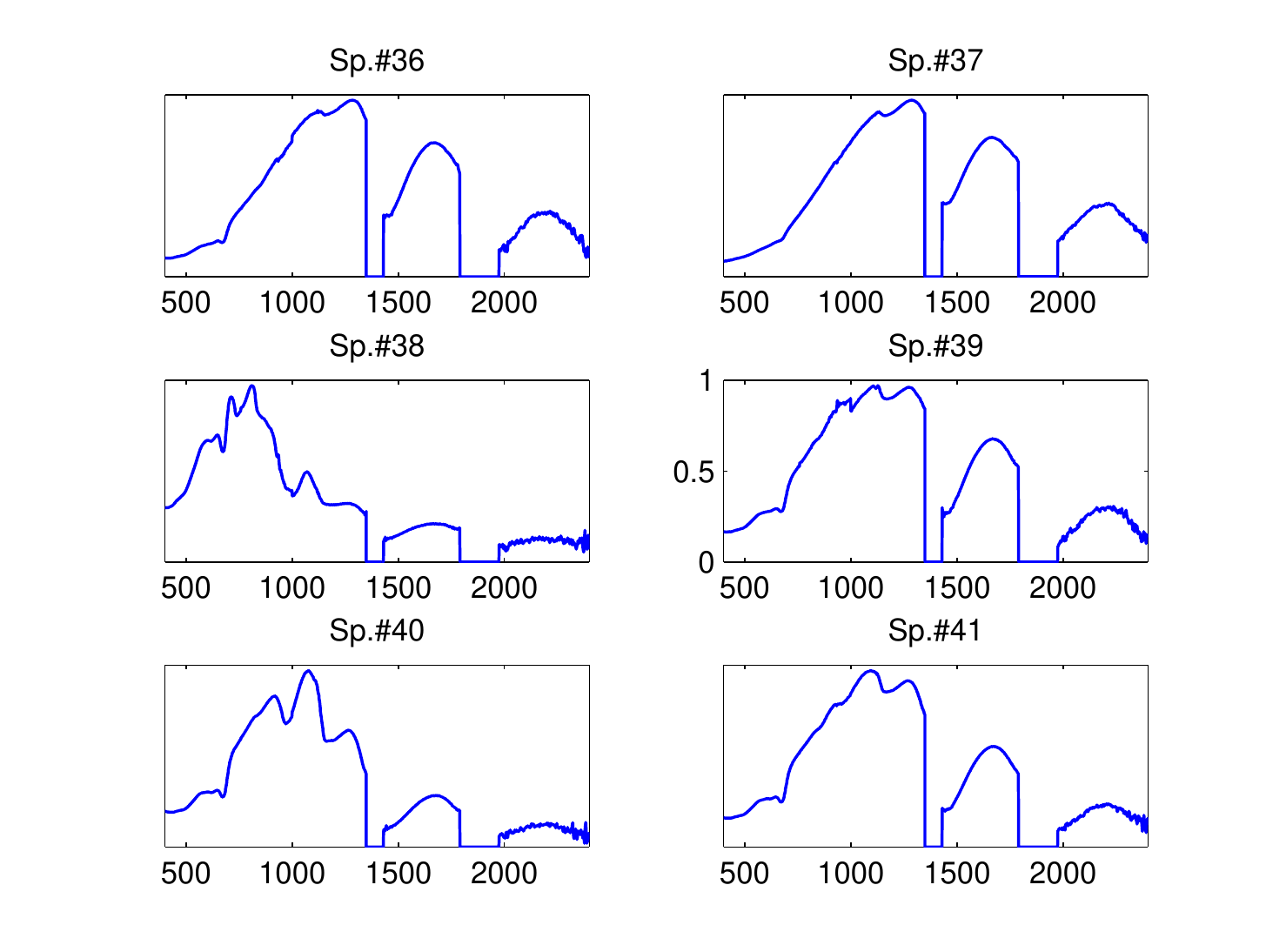} 
\par\end{centering}

\caption{Spectra $\#36-\#41$ from \emph{DB1}.}

\label{spe6} 
\end{figure}

\begin{table}[!h]
\begin{centering}
\begin{tabular}{|l|llllllllll|}
\hline 
36  & 669  & 1157  & 0  & 0  & 0  & 0 & 0  & 0  & 0  & 0\tabularnewline
37 & 1162 & 1446  & 0  & 0  & 0  & 0  & 0  & 0  & 0  & 0\tabularnewline
38  & 673  & 743  & 1008  & 1185  & 2011  & 2236  & 2337  & 0  & 0  & 0\tabularnewline
39  & 670  & 1009  & 1171  & 1444  & 0  & 0  & 0  & 0  & 0  & 0\tabularnewline
40  & 432 & 672 & 973  & 1167  & 2001  & 2148 & 2343 & 0 & 0 & 0\tabularnewline
41  & 671 & 1168  & 1445  & 0  & 0 & 0  & 0 & 0 & 0 & 0\tabularnewline
\hline 
\end{tabular}
\par\end{centering}

\caption{Deep minima locations in ascending order of spectra $\#36-\#41$ are
placed in the initial $K=10$ columns of the matrix $\textbf{S}$.}

\label{T1} 
\end{table}

\begin{table}[!h]
\begin{centering}
\begin{tabular}{|l|llllllllll|}
\hline 
36  & 4454 & 5008  & 5143 & 5166 & 5196 & 5299 & 5322  & 3000 & 3000 & 3000\tabularnewline
37 & 5155 & 3000 & 3000 & 3000 & 3000 & 3000 & 3000 & 3000 & 3000 & 3000\tabularnewline
38  & 3616 & 5046 & 5067  & 5088  & 5117 & 5188 & 3000 & 3000 & 3000 & 3000\tabularnewline
39  & 3970 & 4119 & 5008 & 5043 & 5171 & 5197  & 5232 & 5261 & 5290 & 3000\tabularnewline
40  & 3619 & 5039 & 5062  & 5088  & 5117  & 5181  & 5215  & 5260 & 3000  & 3000\tabularnewline
41  & 3420 & 3617 & 5056 & 5169 & 5258  & 5324  & 3000 & 3000 & 3000 & 3000\tabularnewline
\hline 
\end{tabular}
\par\end{centering}

\caption{Shallow minima locations added by 3000 in ascending order of spectra
$\#36-\#41$ are placed in columns 11-20 of the matrix $\textbf{S}$.}

\label{T2} 
\end{table}

\begin{table}[!h]
\begin{centering}
\begin{tabular}{|l|llllllllll|}
\hline 
36  & 6458 & 6598 & 6931 & 7121 & 7283 & 7669  & 8084 & 6000 & 6000  & 6000\tabularnewline
37 & 6470 & 6592 & 6658 & 7131 & 7286 & 7449 & 7665 & 6000 & 6000 & 6000\tabularnewline
38  & 6420 & 7224 & 7455 & 7556  & 7646  & 6000 & 6000 & 6000 & 6000 & 6000\tabularnewline
39  & 6453 & 6600 & 7274 & 7451 & 7673 & 6000 & 6000 & 6000 & 6000 & 6000\tabularnewline
40  & 6645 & 6916 & 7077 & 7265 & 7450  & 6000 & 6000 & 6000 & 6000 & 6000\tabularnewline
41  & 6454  & 6644  & 6932 & 7094  & 7269 & 7453 & 7673  & 7996 & 8087 & 6000\tabularnewline
\hline 
\end{tabular}
\par\end{centering}

\caption{Flat intervals locations added by 6000 in ascending order of spectra
$\#36-\#41$ are placed in columns 21-30 of the matrix $\textbf{S}$.}

\label{T3} 
\end{table}

\begin{table}[!h]
\begin{centering}
\begin{tabular}{|l|llllllllll|}
\hline 
36  & 9424  & 10238  & 9000  & 9000 & 9000 & 9000 & 9000 & 9000 & 9000 & 9000 \tabularnewline
37 & 9425 & 9929 & 11049 & 11094 & 9000 & 9000 & 9000 & 9000 & 9000 & 9000 \tabularnewline
38  & 9861 & 9000 & 9000 & 9000 & 9000 & 9000 & 9000 & 9000 & 9000 & 9000 \tabularnewline
39  & 9428 & 9841 & 10063 & 10022 & 9000 & 9000 & 9000 & 9000 & 9000 & 9000 \tabularnewline
40  & 9472 & 9832 & 9000 & 9000 & 9000 & 9000 & 9000 & 9000 & 9000 & 9000 \tabularnewline
41  & 9843 & 9000 & 9000 & 9000 & 9000 & 9000 & 9000 & 9000 & 9000 & 9000 \tabularnewline
\hline 
\end{tabular}
\par\end{centering}

\caption{Inflection points locations added by 9000 in ascending order of spectra
$\#36-\#41$ are placed in columns 31-40 of the matrix $\textbf{S}$.}

\label{T4} 
\end{table}

\item [{Initialization~of~the~feature~distribution~table~$\textbf{T}$:}] In
the beginning, the size of $\textbf{T}$ is unknown. The largest size
of $\textbf{T}$ is $N\times I$ where $I$ is the total number of
different features found in the entire database. Let $t$ be a binary
row vector of $N+1$ zeros, where $N$ is the number of spectra in
the reference database. The column $\{\textbf{S}(k,1)\}_{k=1}^{N}$
in the matrix $\textbf{S}$ contains the locations of the first deep
minimum of all the spectra. Let $m=\min_{1\le k\le N}\textbf{S}(k,1)$.
We set $t(1)=m$.

\texttt{Selection Criterion:} \emph{We find for $1\le k\le N$ all
the spectra in the reference database such that $\left|S(k,1)-m\right|\leq\alpha$
where the parameter $\alpha$ determines the tolerance interval.}

Let the indices of these spectra that satisfy the \texttt{Selection
Criterion} be $\left\{ k_{r}\right\} _{r=1}^{R}$. We set $t(k_{r}+1)=1$,
$r=1,\ldots,R$. %Thus, the first term in row  is the closest to the left location of the deep minimum in the whole database. The next  terms of row  may be either 0 or 1.
Locations that contain 1 mark indices of the spectra that satisfy
the \texttt{Selection Criterion} i.e. indices of spectra that have
a deep minimum whose distance from $m$ is not greater than $\alpha$.
We define the first row of table $\textbf{T}$ to be $t$. 

\item [{First~update~of~the~feature~matrix~$S$:}] We update the
matrix $\textbf{S}$ in the following way. For each $1\le j\le4K$,
rows $\left\{ s(k_{r},j)\right\} ,\,1\le r\le R$ correspond to the
indices marked in the previous step. All these rows are subjected
to one-sample left side shift. Formally, 
\begin{equation}
\tilde{s}(k_{r},j)\stackrel{\Delta}{=}s(k_{r},j+1),\,1\le r\le R,\,1\le j\le4K-1,\;\tilde{s}(k_{r},4K)=0.\label{ap1}
\end{equation}
It may happen that after the shift by Eq. \ref{ap1}, the first $K-1$
columns in some row $\kappa$ become $\tilde{s}(\kappa,j)=0$. It
means that spectrum $\#\kappa$ has only one deep minima which is
within distance $\alpha$from $m$. In this case, we apply additional
shifts of the type in Eq. \ref{ap1} to row $\tilde{s}(\kappa,j)$
until a non-zero term appears at the column 1. These rows are discarded
since all their deep minima were considered. These spectra are reconsidered
when other feature types are being processed e.g. shallow minima.
Note that terms that are equal to 3000, 6000 or 9000 are also treated
as 0. Thus, a set of features in spectrum $\#\kappa$ other than deep
minima (most probably, shallow minima) appear at the first ten positions.
These features will be considered no sooner than when all the deep
minima features of all the spectra in the reference database are used.
The output is the updated feature matrix ${\bf \tilde{S}}$.
\item [{Complement~of~table~$\textbf{T}$:}] We repeat the above operations
on the updated feature matrix ${\bf \tilde{S}}$. Thus, we produce
the second row in table $\textbf{T}$ and the updated feature matrix
${\bf \tilde{S}}$. Subsequent iterations produce more rows in table
$\textbf{T}$ that correspond to the ascending sequence of features.
The values of the features appear at the first column of table $\textbf{T}$.
During the updating process of the feature matrix ${\bf \tilde{S}}$,
some rows become 0 (or 3000,6000, 9000). We remove these rows from
the matrix ${\bf \tilde{S}}$. Thus, its size is reduced. The iterations
continue until the matrix ${\bf \tilde{S}}$ becomes a one-row matrix
consisting of zeros (or 3000,6000, 9000).

Table $\textbf{T}=\left\{ t(i,j)\right\} ,\,1\le i\le I,\,\,1\le j\le N+1,$
is completed. Its first column comprises all the features present
in the matrix ${\bf \left\{ S\right\} }$ in ascending order. This
number of features is $I$. The remaining columns consist of zeros
and ones. Locations of ones indicate spectra that have a feature.
For example, if $t(k,1)=1520$ and $t(k,m)=1$ then it means that
spectrum $\#m-1$ has a feature at the location 1520 (up to a tolerance
interval $\alpha$). Since $1520<3000$, this feature is a deep minimum.
If $t(w,1)=4715$ and $t(w,v)=1$ then it means that spectrum $\#v-1$
has a feature at location 4715-3000=1715 (up to a tolerance interval
$\alpha$). Since $3000<4715<6000$, this feature is a shallow minimum.

\end{description}
\begin{table}[!h]
\begin{centering}
\begin{tabular}{|l|llllll|}
\hline 
Features & 36  & 37  & 38  & 39  & 40 & 41 \tabularnewline
\hline 
973 & 0 & 0 & 0 & 0 & 1 & 0 \tabularnewline
1008  & 0 & 0 & 1 & 1 & 0 & 0 \tabularnewline
1157  & 1 & 1 & 0 & 0 & 1 & 0 \tabularnewline
1168  & 0 & 0 & 0 & 1 & 0 & 1 \tabularnewline
1185  & 0 & 0 & 1 & 0 & 0 & 0 \tabularnewline
1444  & 0 & 1 & 0 & 1 & 0 & 1 \tabularnewline
2001  & 0 & 0 & 1 & 0 & 1 & 0 \tabularnewline
\hline 
\end{tabular}
\par\end{centering}

\caption{Deep minima of spectra $\#36-\#41$ in a portion of the feature distribution
table $\textbf{T}$.}

\label{T5} 
\end{table}

From table \ref{T5} we see that spectra $\#38$ and $\#39$ have
a deep minimum at location 1008, spectra $\#37$ and $\#39$ have
a deep minimum at location 1444, etc.

\begin{table}[!h]
\begin{centering}
\begin{tabular}{|l|llllll|}
\hline 
Features & 36  & 37  & 38  & 39  & 40 & 41 \tabularnewline
\hline 
3616 & 0 & 0 & 1 & 0 & 1 & 1 \tabularnewline
3970  & 0 & 0 & 0 & 1 & 0 & 0 \tabularnewline
4119  & 0 & 0 & 0 & 1 & 0 & 0 \tabularnewline
5008  & 1 & 0 & 0 & 1 & 0 & 0 \tabularnewline
5039  & 0 & 0 & 1 & 1 & 1 & 0 \tabularnewline
5056  & 0 & 0 & 0 & 1 & 0 & 0 \tabularnewline
4623  & 0 & 0 & 0 & 0 & 1 & 1\tabularnewline
\hline 
\end{tabular}
\par\end{centering}

\caption{Shallow minima of spectra $\#36-\#41$ in another portion of the feature
distribution table $\textbf{T}$.}

\label{T6} 
\end{table}

From table \ref{T6} we see that spectra $\#38,\#40$ and $\#41$
have shallow minimum at location 616=3616-3000, spectra $\#36$ and
$\#39$ have shallow minimum at location 2008=5008-3000, etc.

\begin{table}[!h]
\begin{centering}
\begin{tabular}{|l|llllll|}
\hline 
Features & 36  & 37  & 38  & 39  & 40 & 41 \tabularnewline
\hline 
6453 & 1 & 0 & 0 & 1 & 0 & 1\tabularnewline
6592  & 1 & 1 & 0 & 1 & 0 & 0 \tabularnewline
6644  & 0 & 0 & 0 & 0 & 1 & 1 \tabularnewline
6931  & 1 & 0 & 1 & 0 & 0 & 1 \tabularnewline
7121  & 1 & 1 & 0 & 0 & 0 & 0 \tabularnewline
7265  & 0 & 0 & 0 & 1 & 1 & 1 \tabularnewline
7283  & 1 & 1 & 0 & 0 & 0 & 0 \tabularnewline
\hline 
\end{tabular}
\par\end{centering}

\caption{Flat intervals of spectra $\#65-\#70$ in a portion of the feature
distribution table $\textbf{T}$.}

\label{T7} 
\end{table}

From table \ref{T7} we see that spectra $\#36,\#39$ and $\#41$
have flat interval at location 453=6453-6000, spectra $\#36$ and
$\#37$ have flat interval at location 1121=7121-6000, etc.

\begin{table}[!h]
\begin{centering}
\begin{tabular}{|l|llllll|}
\hline 
Features & 36  & 37  & 38  & 39  & 40 & 41 \tabularnewline
\hline 
9424 & 1 & 1 & 0 & 1 & 0 & 0\tabularnewline
9472  & 0 & 0 & 0 & 0 & 1 & 0 \tabularnewline
9832  & 0 & 0 & 0 & 1 & 1 & 0 \tabularnewline
9861  & 0 & 0 & 1 & 0 & 0 & 0 \tabularnewline
\hline 
\end{tabular}
\par\end{centering}

\caption{inflection points of spectra $\#36-\#41$ in a portion of the feature
distribution table $\textbf{T}$.}

\label{T8} 
\end{table}

We see from table \ref{T8} that spectra $\#36,\#37$ and $\#39$
have an inflection point at location 424=9424-9000, spectra $\#39$
and $\#40$ have an inflection point at location 832= 9832-9000, etc.

\subsection{Identification of a spectrum}
\begin{description}
\item [{Input:}] A raw \emph{test} spectrum $s_{n}$ and the constructed
feature distribution table $\textbf{T}$ (as was described in Section
\ref{ss42}) are loaded. The \emph{test} spectrum may or may not belong
to the reference database. 
\item [{Extraction~of~features:}] The features of spectrum $s_{n}$ are
extracted according to the procedure described in Section \ref{Features}.
The features are gathered into the row vector ${\bf X}_{n}=\left\{ x_{n}(k)\right\} _{k=1}^{4K}$,
whose structure is similar to the structure of a row in the matrix
$\textbf{S}$ with the difference that we discard the values 0, 3000,
6000 and 9000. For example, 
\begin{eqnarray}
 &  & {\bf X}_{39}=(670\;1009\;1171\;1444\;3970\;4119\;5008\;5043\;5171\;5197\;5232\;5261...\nonumber \\
 &  & 5290\;3000\;6453\;6600\;7274\;7673\;9428\;9841\;10063\;10222)\label{re2}
\end{eqnarray}
 is the features of spectrum 39 without 0, 3000, 6000 and 9000 (see
tables \ref{T1}, \ref{T2}, \ref{T3} and \ref{T4}). 
\item [{Reduction~of~the~feature~distribution~table~$\textbf{T}$:}] Recall
that for $1\le i\le I$, the first column of $t(i,1)$ in table $\textbf{T}$
contains all the features from the reference database. Also, $N$
is the number of spectra in the reference database and $I$ is the
number of features. We define an \emph{indicator} column vector $\textbf{r}_{n}\stackrel{\Delta}{=}\left(r_{n}(1),\ldots,r_{n}(I)\right)$
which is initialized with zeros. Next, we test all the features $\left\{ x_{n}(k)\right\} $
in the following way: if the feature $x_{n}(k)$ is close to a feature
$\left\{ t(\gamma,1)\right\} $, $1\le\gamma\le I$, such that 
\begin{equation}
|x_{n}(k)-t(\gamma,1)|<\alpha\label{pe3}
\end{equation}
 then, we set $r_{n}(\gamma)=1$, where $\alpha$ is a tolerance parameter.
It may happen that inequality \ref{pe3} is valid for two features
$t(\gamma_{1},1)$ and $t(\gamma_{2},1)$. In this case, we choose
$t(\gamma_{1},1)$. Let $\Gamma=\left\{ \gamma_{k}\right\} _{k=1}^{L}$
be the set of indices such that $r_{n}(\gamma_{k})=1$.

Finally, we prepare $\tilde{\textbf{T}}_{n}$, which is the reduced
version of table $\textbf{T}$ that contains only rows whose indices
belong to $\Gamma$ i.e. 
\[
\tilde{\textbf{T}}=\left\{ t(i,j)\right\} ,\; i\in\Gamma,\;1\le j\le N+1.
\]

Table \ref{T9} presents a portion of the reduced feature distribution
table $\textbf{T}_{39}$ for the spectra $\#36-\#41$, which corresponds
to deep minima (see table \ref{T1}).

\begin{table}[!h]
\begin{centering}
\begin{tabular}{|l|l|l|l|l|l|l|}
\hline 
$t(i,1)$ & S.36 $t(i,2)$  & S.37 $t(i,3)$ & S.38 $t(i,4)$  & S.39 $t(i,5)$  & S.40 $t(i,6)$ & S.41 $t(i,7)$ \tabularnewline
\hline 
669  & 1 & 0 & 1 & 1 & 1 & 1 \tabularnewline
1008  & 0 & 0 & 1 & 1 & 0 & 0 \tabularnewline
1168  & 0 & 0 & 0 & 1 & 0 & 1 \tabularnewline
1444  & 0 & 1 & 0 & 1 & 0 & 1 \tabularnewline
\hline 
$C(39,j)$ & 1 & 1 & 2 & 4 & 1 & 3\tabularnewline
\hline 
\end{tabular}
\par\end{centering}

\caption{Portion of the reduced feature distribution table $\textbf{T}_{39}$
for the spectra $\#36-\#41$.}

\label{T9} 
\end{table}

\item [{Finding~spectra~that~have~common~features~with~the~presented~spectrum:}] For
this purpose, we calculate for $i\in\Gamma$ the sums of the columns
$t(i,j),\;2\le j\le N+1,$ of the reduced table $\tilde{\textbf{T}}$
that correspond to the spectra in the reference database: 
\[
C_{n,j}=\sum_{i\in\Gamma}t(i,j),\;2\le j\le N+1.
\]

The value of $C_{n,j}$ determines the number of features in spectrum
$\#j$ from the database that coincide (up to the tolerance parameter
$\alpha$) with the features of the presented spectrum $\#n$. The
values of $C_{39,j}$ for Table \ref{T9} are given in its bottom
row.

Note that the user can use all the available sets of features (deep
and shallow minima, flat intervals an inflection points) or any combination
of these sets. In the above example (table \ref{T9}), only deep minima
are considered. 

\item [{Histograms:}] We elaborate the above process using histograms of
the total number of spectra that have common features against the
number of common features. In other words, we calculate the numbers
of common features $C_{n,j}$.

The histogram in Fig. \ref{histr} displays the distribution of spectra
from the reference database (spectra $\#36$ to $\#41$) according
to the number of common features (deep minima) with the presented
spectrum ($\#39$). We can see that three spectra have only one common
feature with spectrum $\#39$, where as one spectrum has two common
features and one spectrum has three common features with spectrum
$\#39$.

\begin{figure}[!h]

\begin{centering}
\includegraphics[width=0.35\paperheight]{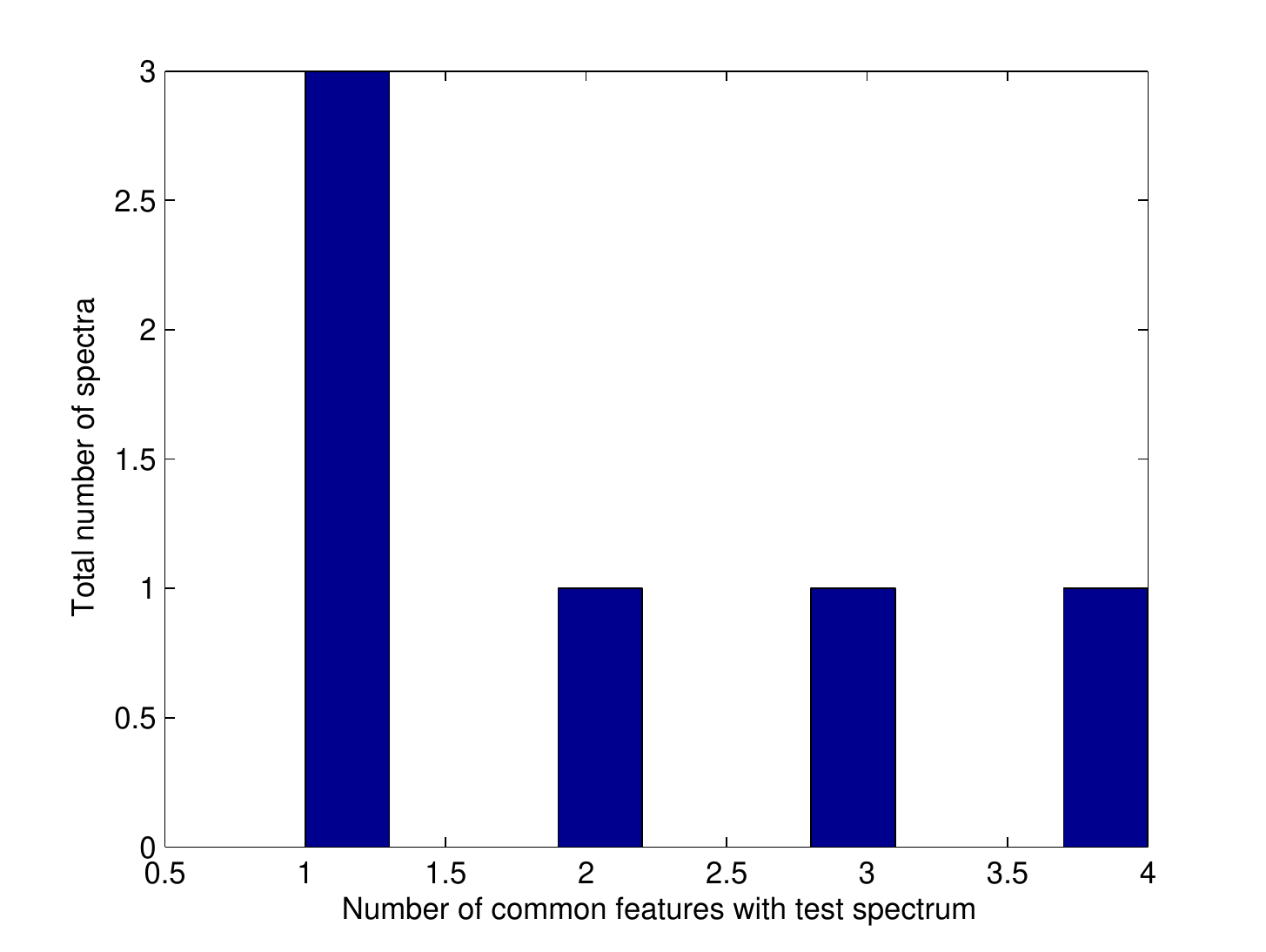} 
\par\end{centering}

\caption{Distribution of spectra $\#36$ to $\#41$ from \emph{DB1} according
to the number of common features (deep minima) with the presented
spectrum $\#39$ from \emph{DB1}.}

\label{histr} 
\end{figure}

The histogram in Fig. \ref{histf} displays the distribution of spectra
from the full reference database (spectra $\#1$ to $\#173$) according
to the number of common features (deep minima) with the presented
spectrum $\#33$.

\begin{figure}[!h]

\begin{centering}
\includegraphics[width=0.35\paperheight]{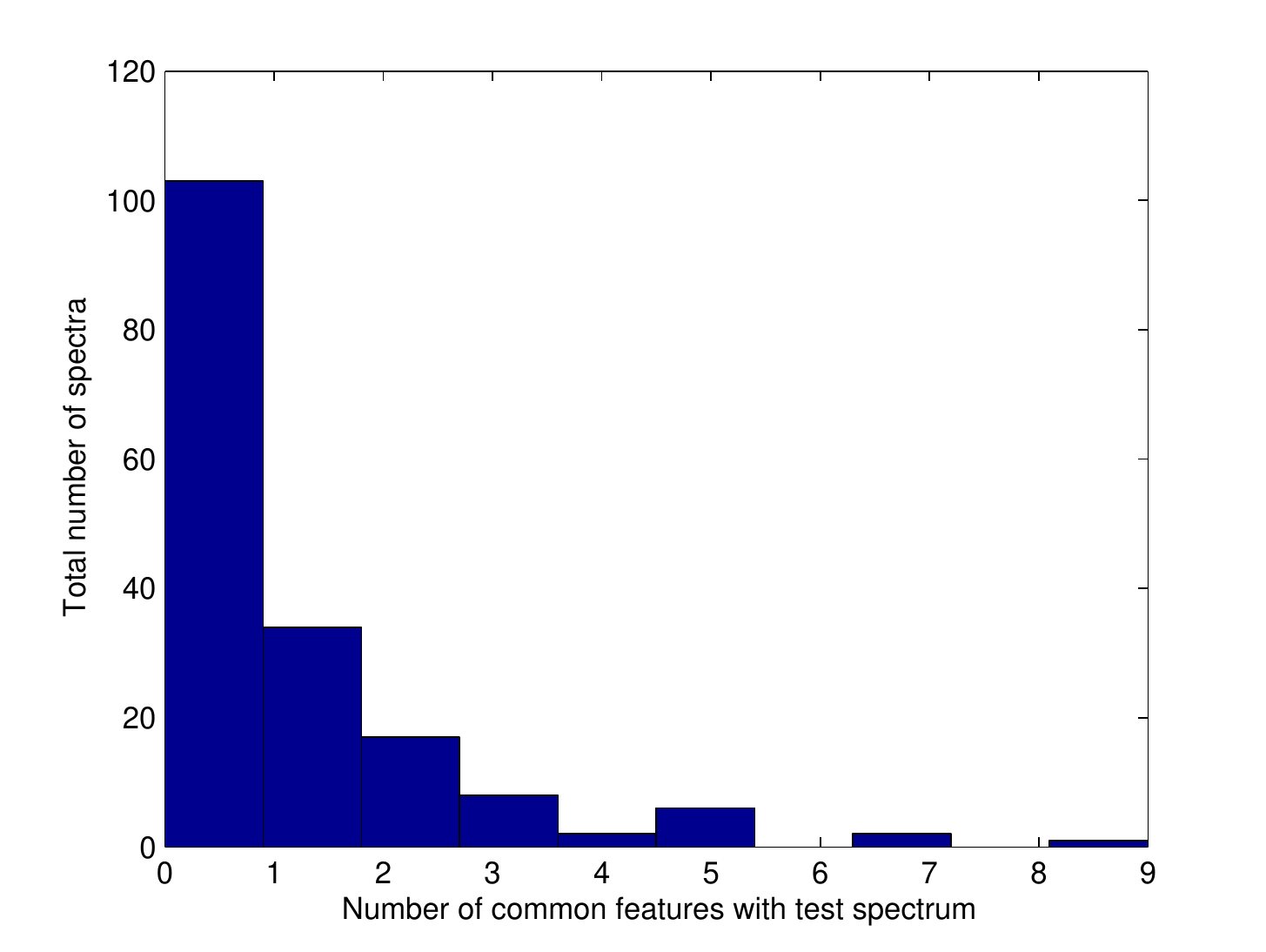} 
\par\end{centering}

\caption{Distribution of spectra $\#1$ to $\#173$ from \emph{DB1} according
to the number of common features (deep minima) with the presented
spectrum $\#33$ from \emph{DB1}.}

\label{histf} 
\end{figure}

Two spectra have seven (deep minima) features with spectrum $\#33$,
six spectra have five common deep minima with spectrum $\#33$.

The histogram in Fig. \ref{hista} displays the distribution of spectra
from the full reference database (spectra $\#1$ to $\#173$) according
to the number of common features (from all the available sets of features)
with the presented spectrum $\#33$.

\begin{figure}[!h]

\begin{centering}
\includegraphics[width=0.35\paperheight]{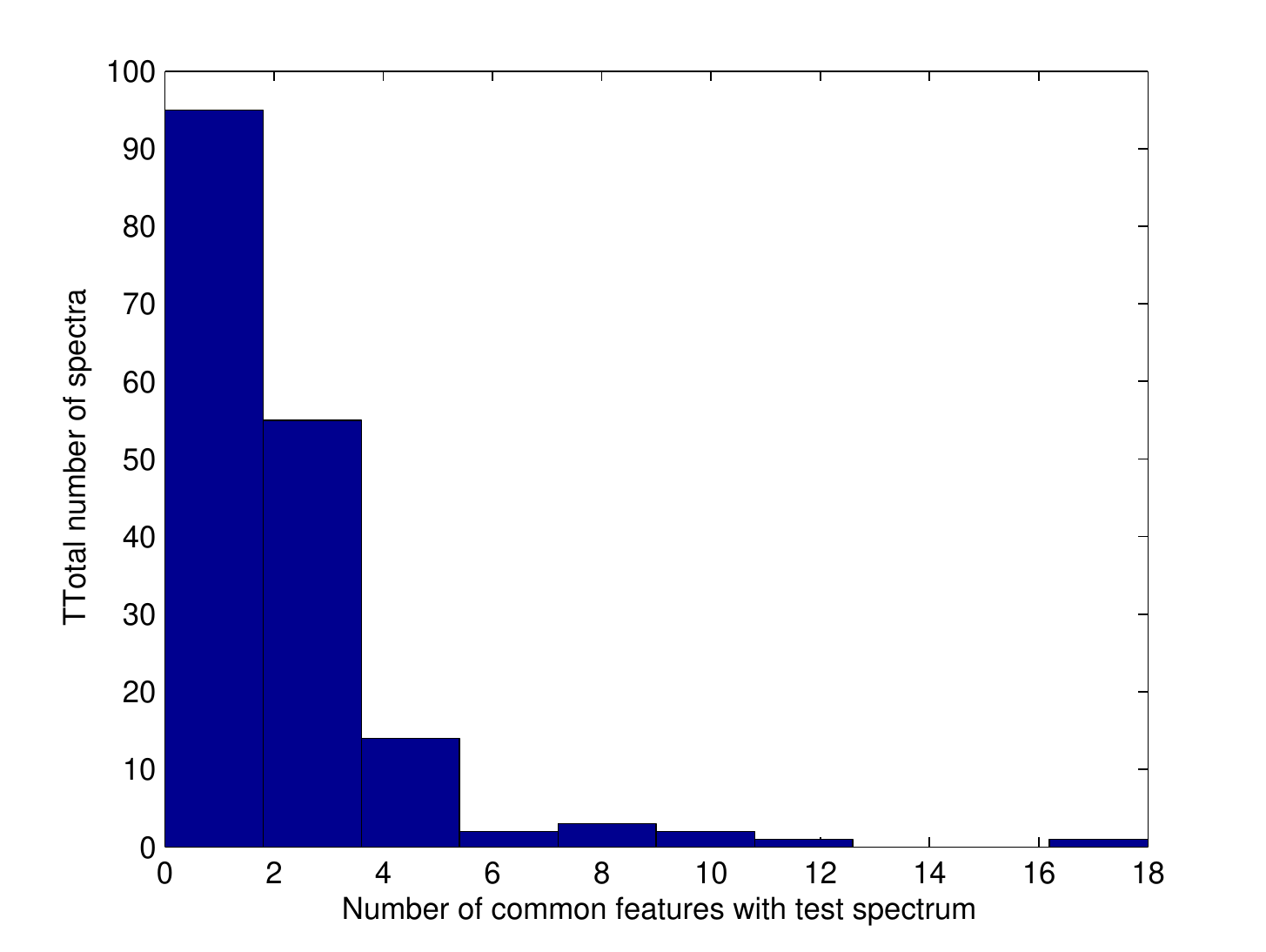} 
\par\end{centering}

\caption{Distribution of spectra $\#1$ to $\#173$ from \emph{DB1} according
to the number of common features (from all available features) with
the presented spectrum $\#33$ from \emph{DB1}.}

\label{hista} 
\end{figure}

\item [{The~output~from~the~identification~process:}] Once the histogram
related to the presented spectrum $\#n$ is displayed, the user indicates
the number $M$ of common features to be used. Then, the algorithm
produces a list of spectra that have $\geq M$ (parameter) common
features (up to the size of the tolerance parameter) with the presented
spectrum. %Optionally, the plots of these spectra are displayed versus the plot of presented spectrum. Common features are also displayed in these images. 
\end{description}

\subsection{Examples}

This section included the results of the proposed feature extraction
algorithm applied to spectra from DB1. For comparison, hierarchical
clustering was applied on the investigated spectra using the shortest
Euclidean distance measure for similarity. The Euclidean distance
compares the \emph{entire} spectrum regardless of the features.

\subsubsection{Examples where both reference and test spectra are taken from \emph{DB1}}

\begin{figure}[!h]
\begin{centering}
\includegraphics[%bb=147bp 257bp 480bp 549bp,clip,
width=0.9\columnwidth,height=0.2\paperheight]{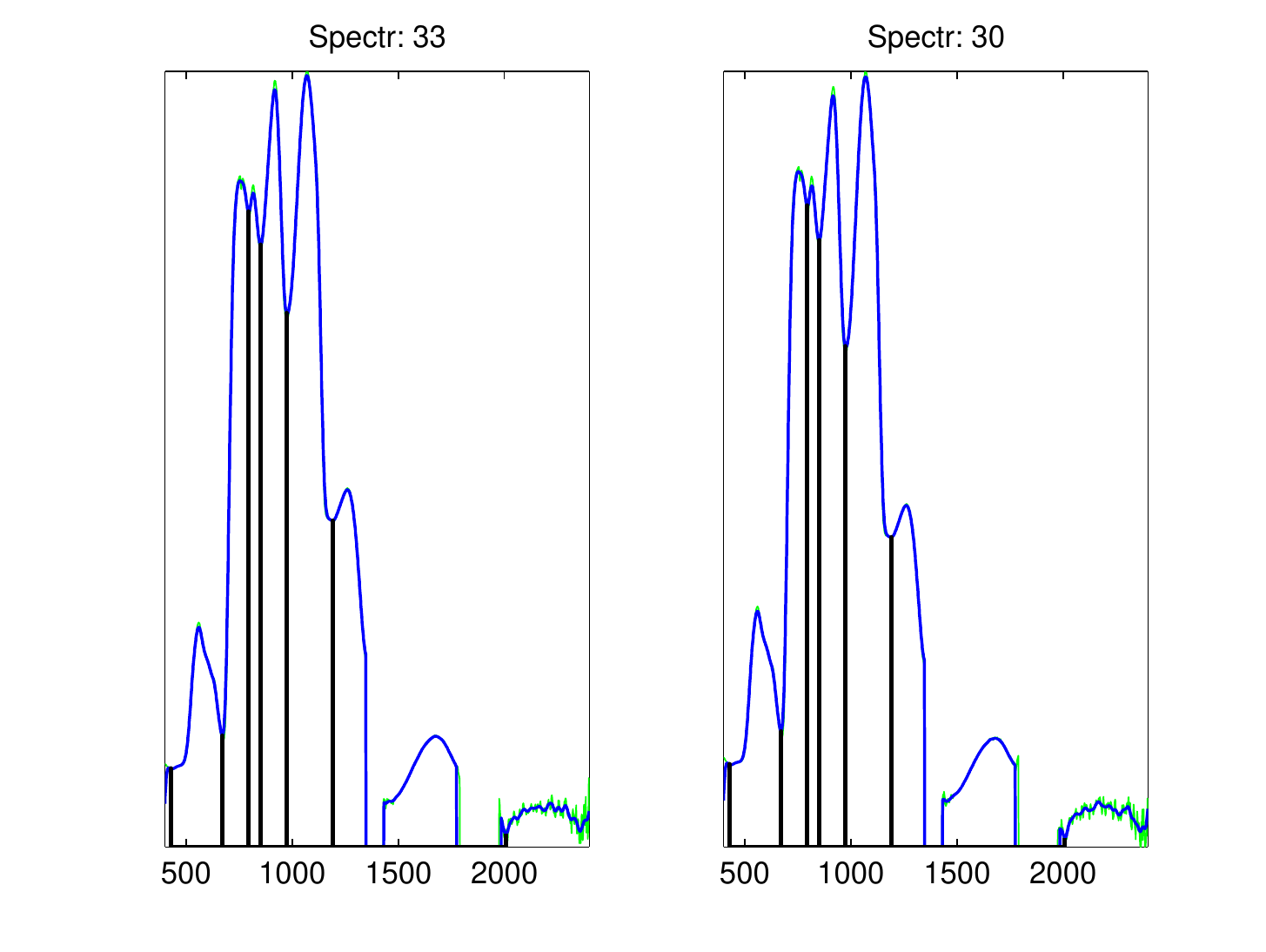} 
\par\end{centering}

\caption{Spectrum $\#30$ (right) in \emph{DB1} has seven common deep minima
with the tested spectrum $\#33$ (left) in \emph{DB1}.}

\label{com33_30} 
\end{figure}

\begin{figure}[!h]
\begin{centering}
\includegraphics[%bb=147bp 257bp 479bp 549bp,clip,
width=0.9\columnwidth,height=0.2\paperheight]{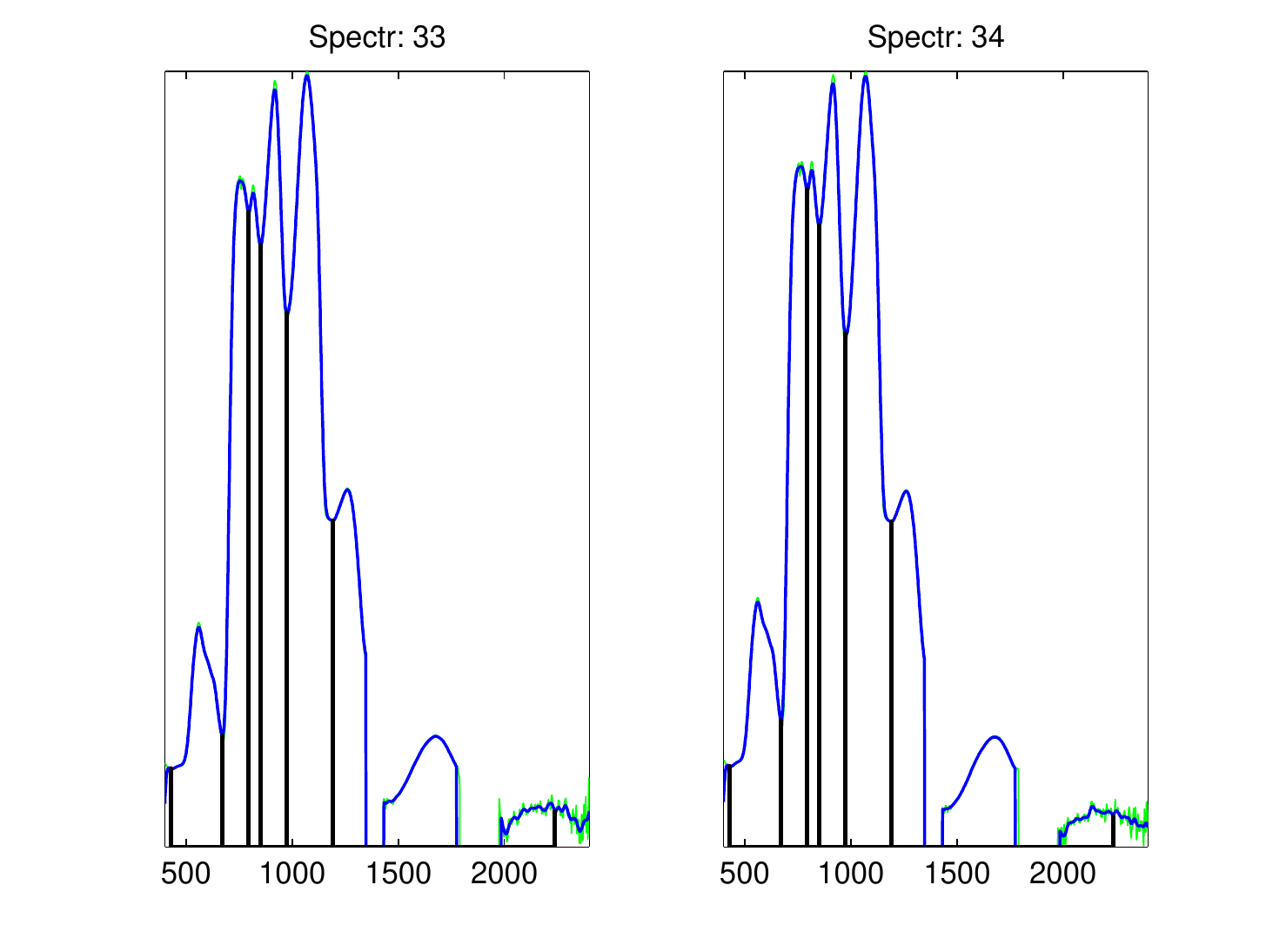}
\par\end{centering}

\caption{Spectrum $\#34$ (right) in \emph{DB1} has seven common deep minima
with the tested spectrum $\#33$ (left) in \emph{DB1}.}

\label{com33_34} 
\end{figure}

\begin{figure}[!h]
\begin{centering}
\includegraphics[%bb=147bp 257bp 480bp 549bp,clip,
width=0.9\columnwidth,height=0.2\paperheight]{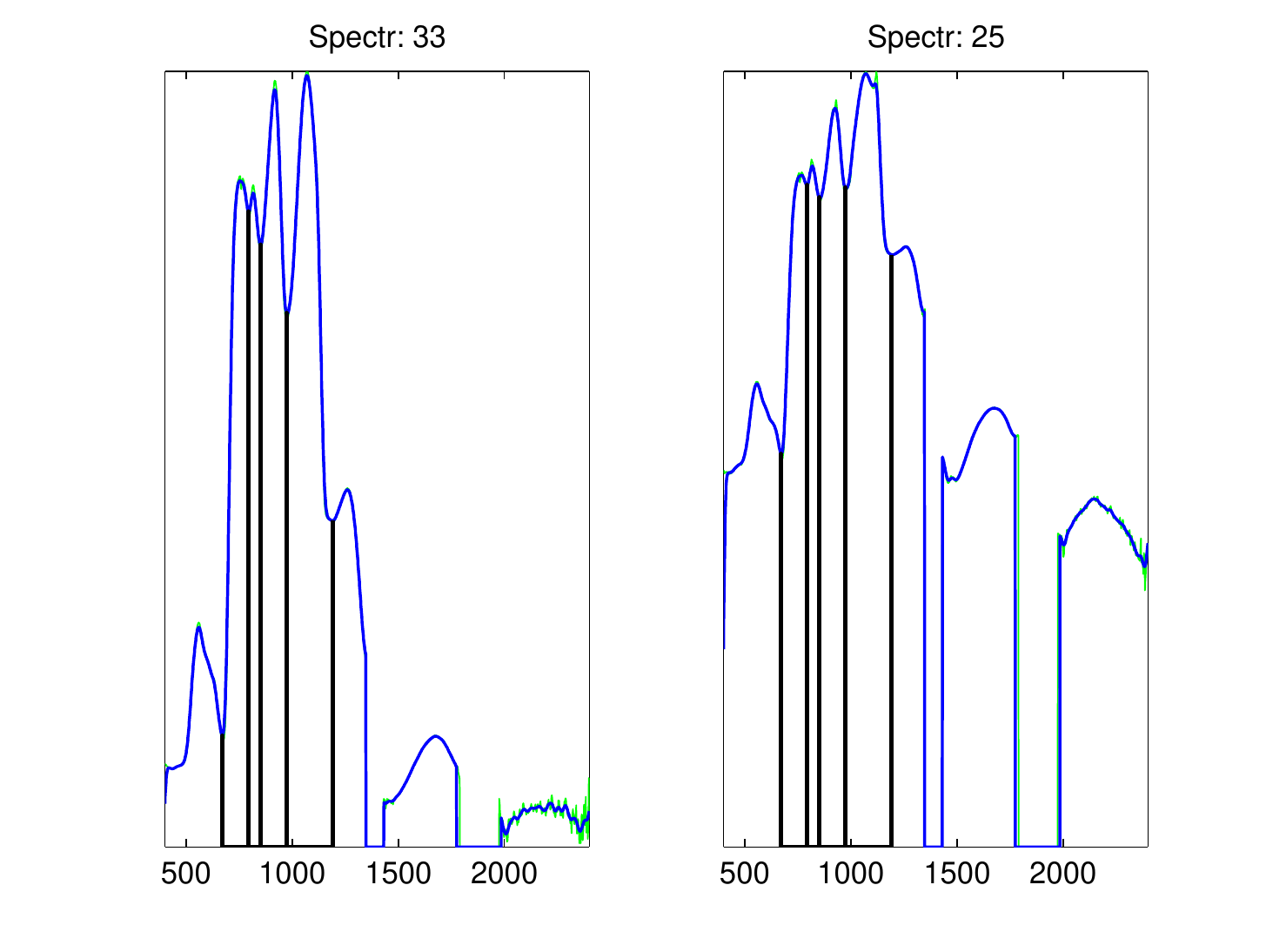}
\par\end{centering}

\caption{Spectrum $\#25$ (right) in \emph{DB1} has five common deep minima
with the tested spectrum $\#33$ (left) in \emph{DB1}.}

\label{com33_25} 
\end{figure}

\begin{figure}[!h]
\begin{centering}
\includegraphics[%bb=147bp 257bp 480bp 549bp,clip,
width=0.9\columnwidth,height=0.2\paperheight]{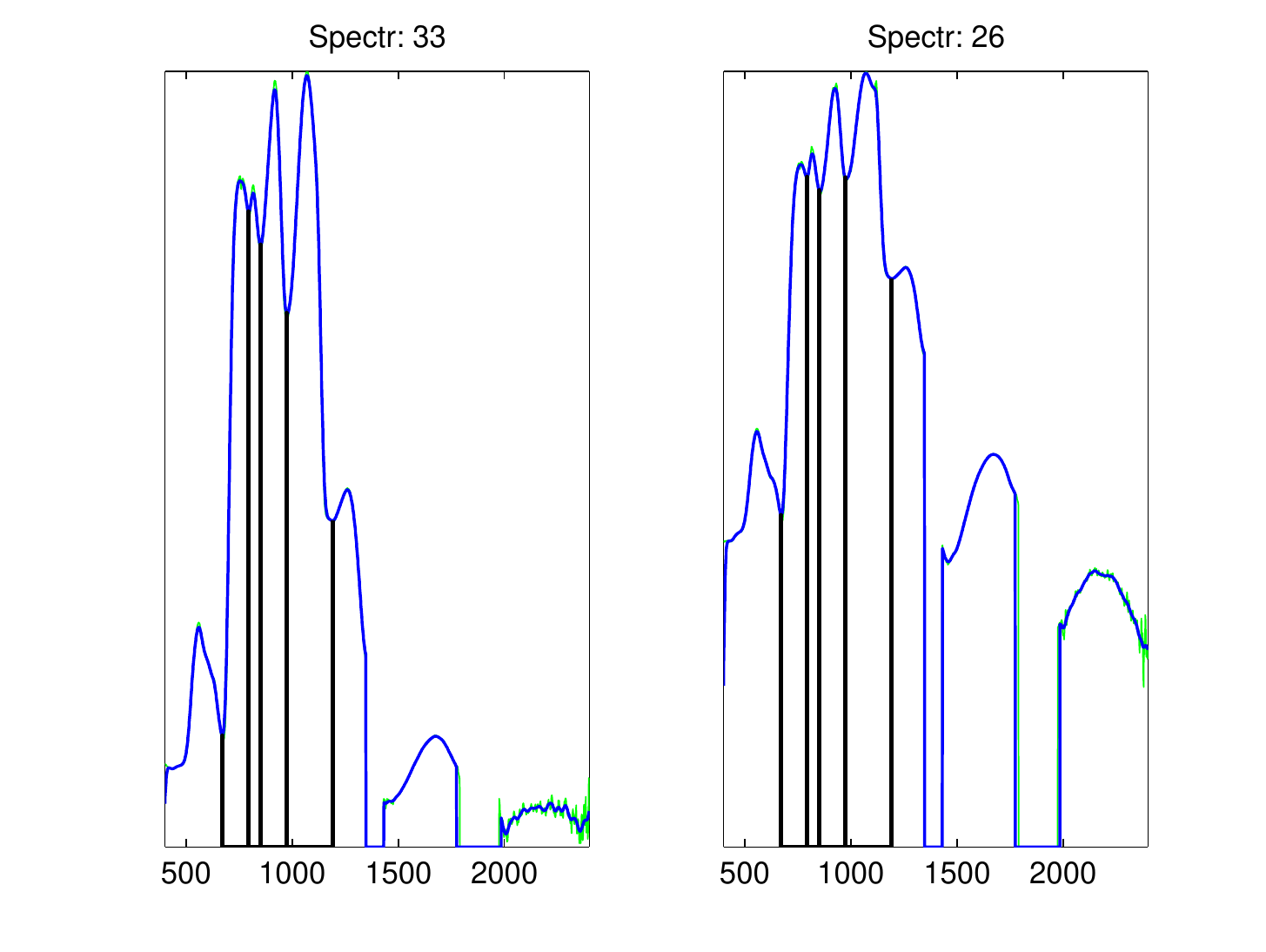} 
\par\end{centering}

\caption{Spectrum $\#26$ (right) in \emph{DB1} has five common deep minima
with the tested spectrum $\#33$ (left) in \emph{DB1}.}

\label{com33_26} 
\end{figure}

\begin{figure}[!h]

\begin{centering}
\includegraphics[%bb=147bp 257bp 480bp 549bp,clip,
width=0.9\columnwidth,height=0.2\paperheight]{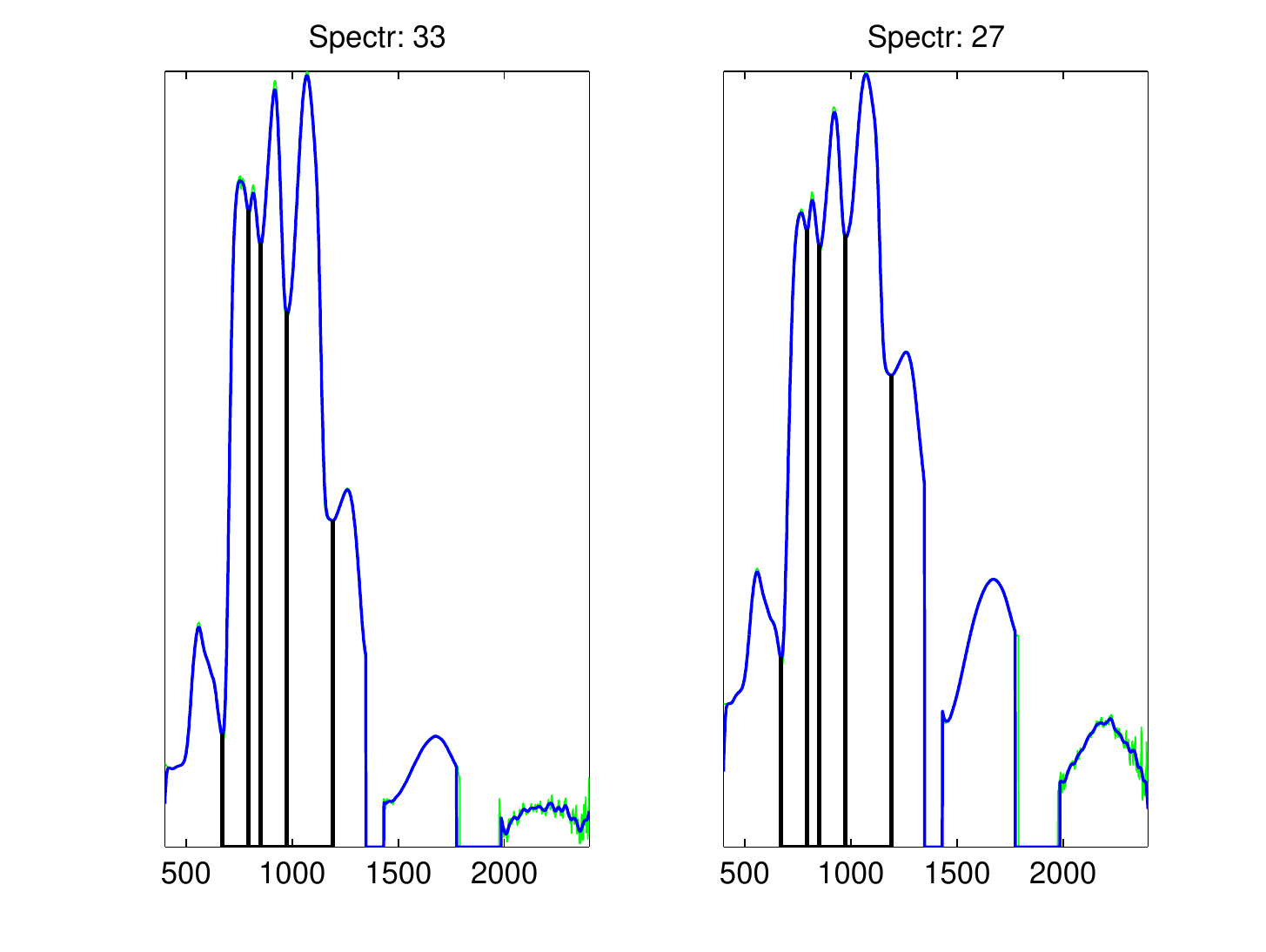}
\par\end{centering}

\caption{Spectrum $\#27$ (right) in \emph{DB1} has five common deep minima
with the tested spectrum $\#33$ (left) in \emph{DB1}.}

\label{com33_27} 
\end{figure}

\begin{figure}[!h]

\begin{centering}
\includegraphics[%bb=147bp 257bp 480bp 549bp,clip,
width=0.9\columnwidth,height=0.2\paperheight]{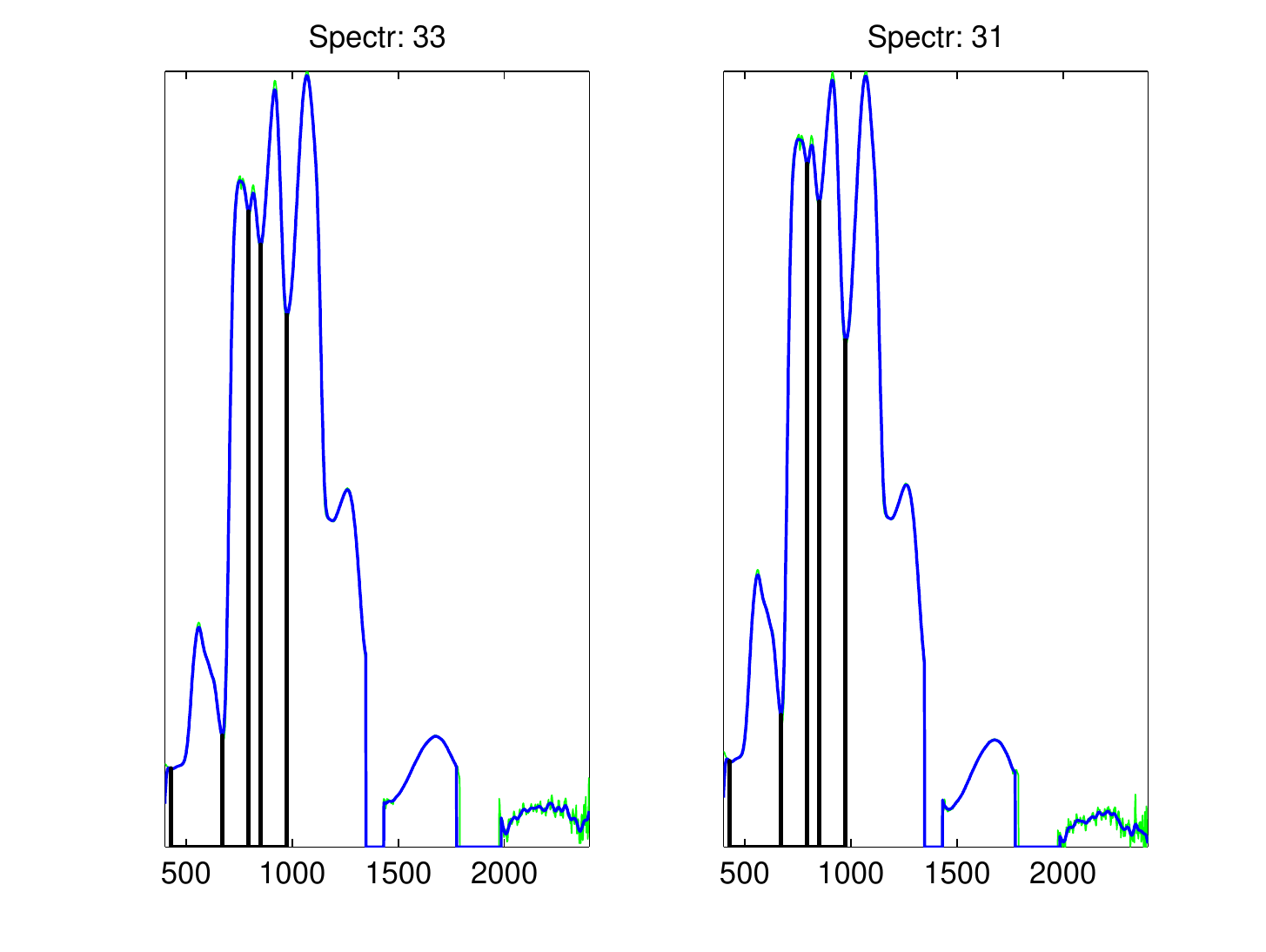}
\par\end{centering}

\caption{Spectrum $\#31$ (right) in \emph{DB1} has five common deep minima
with the tested spectrum $\#33$ (left) in \emph{DB1}.}

\label{com33_31} 
\end{figure}

\begin{figure}[!h]

\begin{centering}
\includegraphics[%bb=147bp 257bp 479bp 549bp,clip,
width=0.9\columnwidth,height=0.2\paperheight]{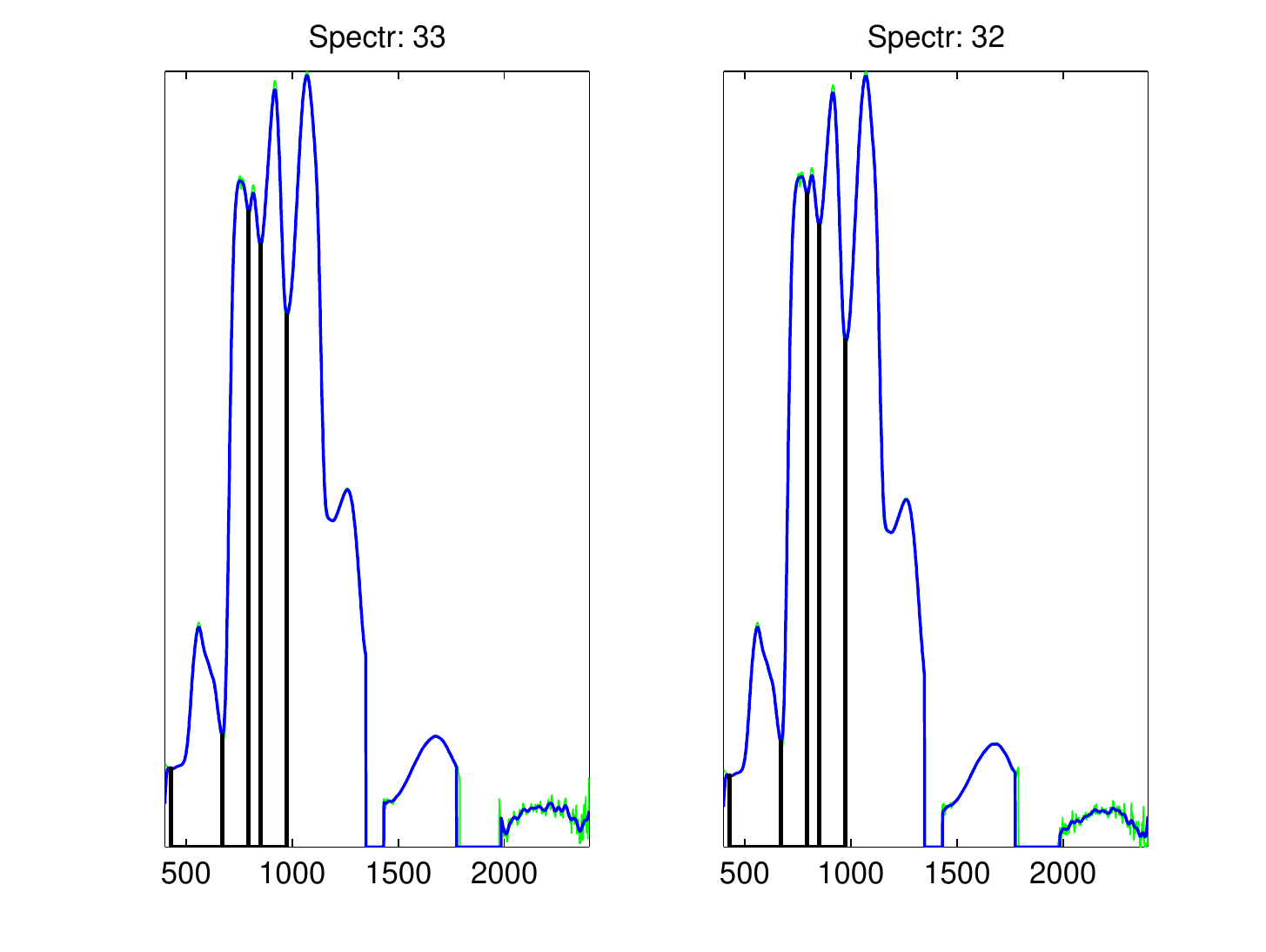} 
\par\end{centering}

\caption{Spectrum $\#32$ (right) in \emph{DB1} has five common deep minima
with the tested spectrum $\#33$ (left) in \emph{DB1}.}

\label{com33_32} 
\end{figure}

\begin{figure}[!h]

\begin{centering}
\includegraphics[%bb=147bp 257bp 479bp 549bp,clip,
width=0.9\columnwidth,height=0.2\paperheight]{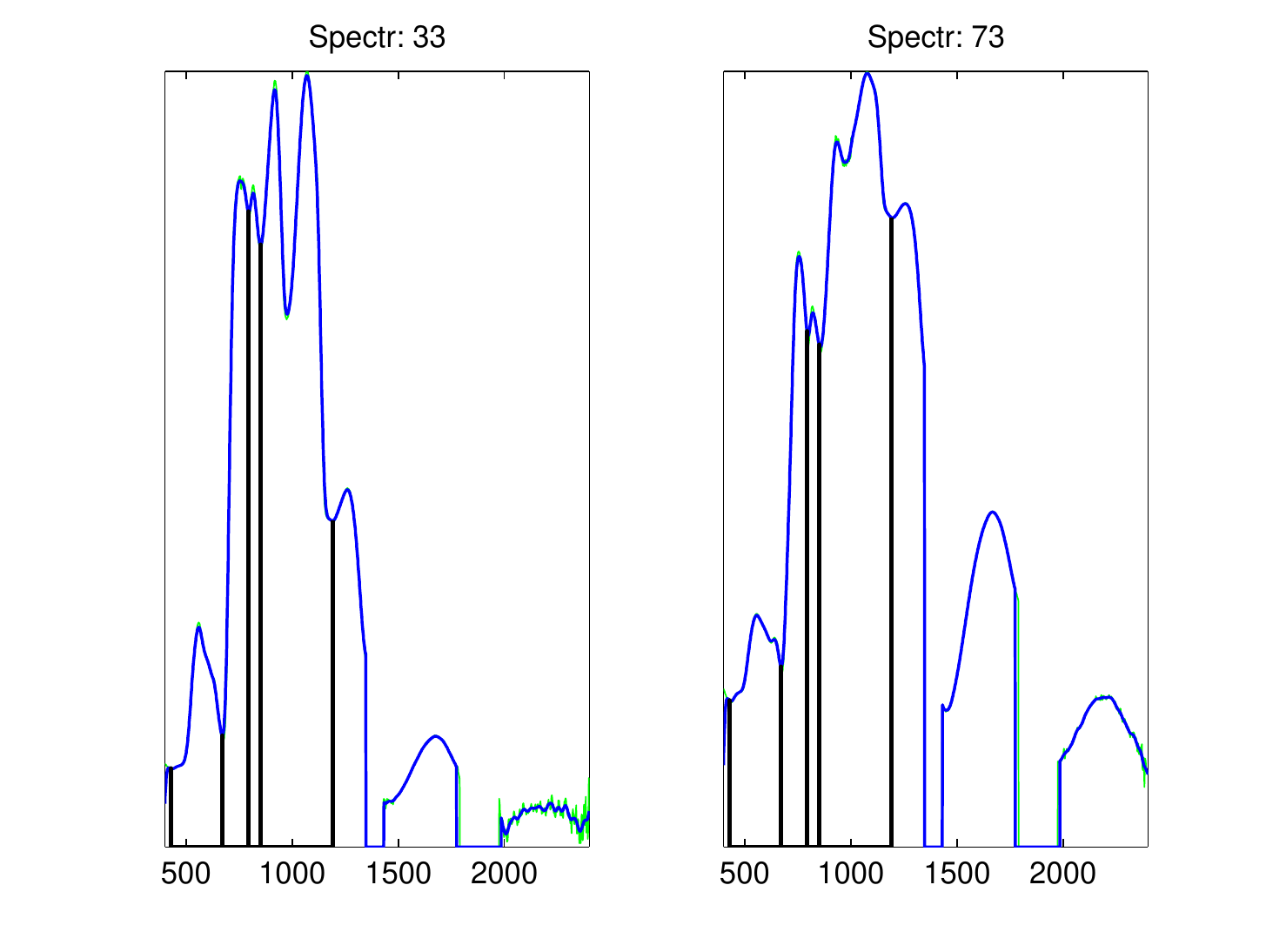}
\par\end{centering}

\caption{Spectrum $\#73$ (right) in \emph{DB1} has five common deep minima
with the tested spectrum $\#33$ (left) in \emph{DB1}.}

\label{com33_73} 
\end{figure}

In this section, the reference database is \emph{DB1} and the tested
spectra belong to \emph{DB1} as well. 
\begin{description}
\item [{Illustrations~for~the~reduced~database:}] We searched the reduced
database (spectra $\#36$ to $\#41$) for spectra that have common
deep minima with spectra $\#39$. We found that spectra $\#39$ has
two and three common deep minima with spectrum $\#38$ and $\#41$,
respectively. 
\begin{figure}[!h]

\begin{centering}
\includegraphics[%bb=147bp 257bp 476bp 549bp,clip,
width=0.9\columnwidth,height=0.2\paperheight]{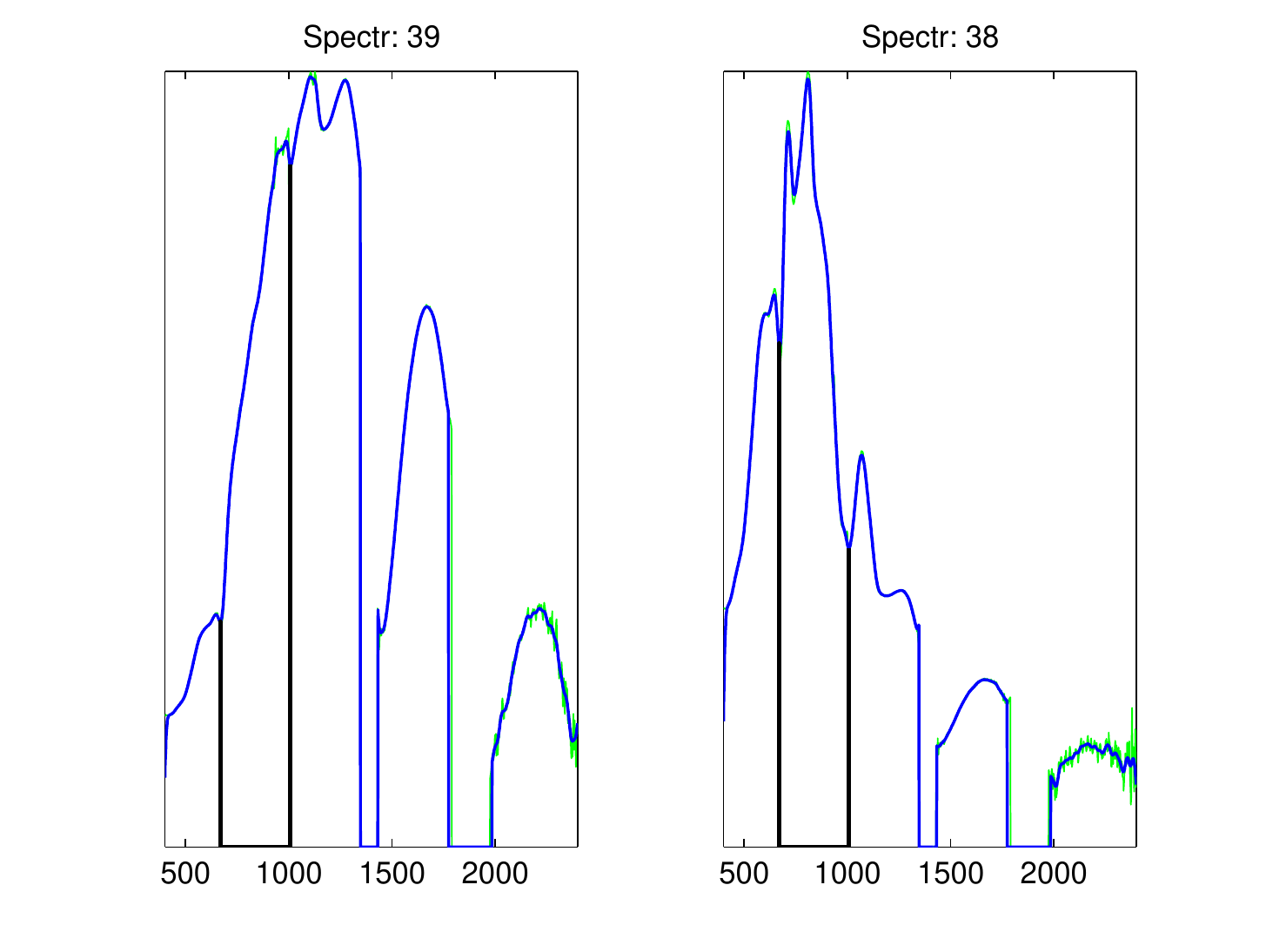} 
\par\end{centering}

\caption{Spectrum $\#38$ from \emph{DB1} has two common deep minima with the
tested spectrum $\#39$, which is also in \emph{DB1}.}

\label{com39_38} 
\end{figure}

\begin{figure}[!h]

\begin{centering}
\includegraphics[%bb=147bp 257bp 476bp 549bp,clip,
width=0.9\columnwidth,height=0.2\paperheight]{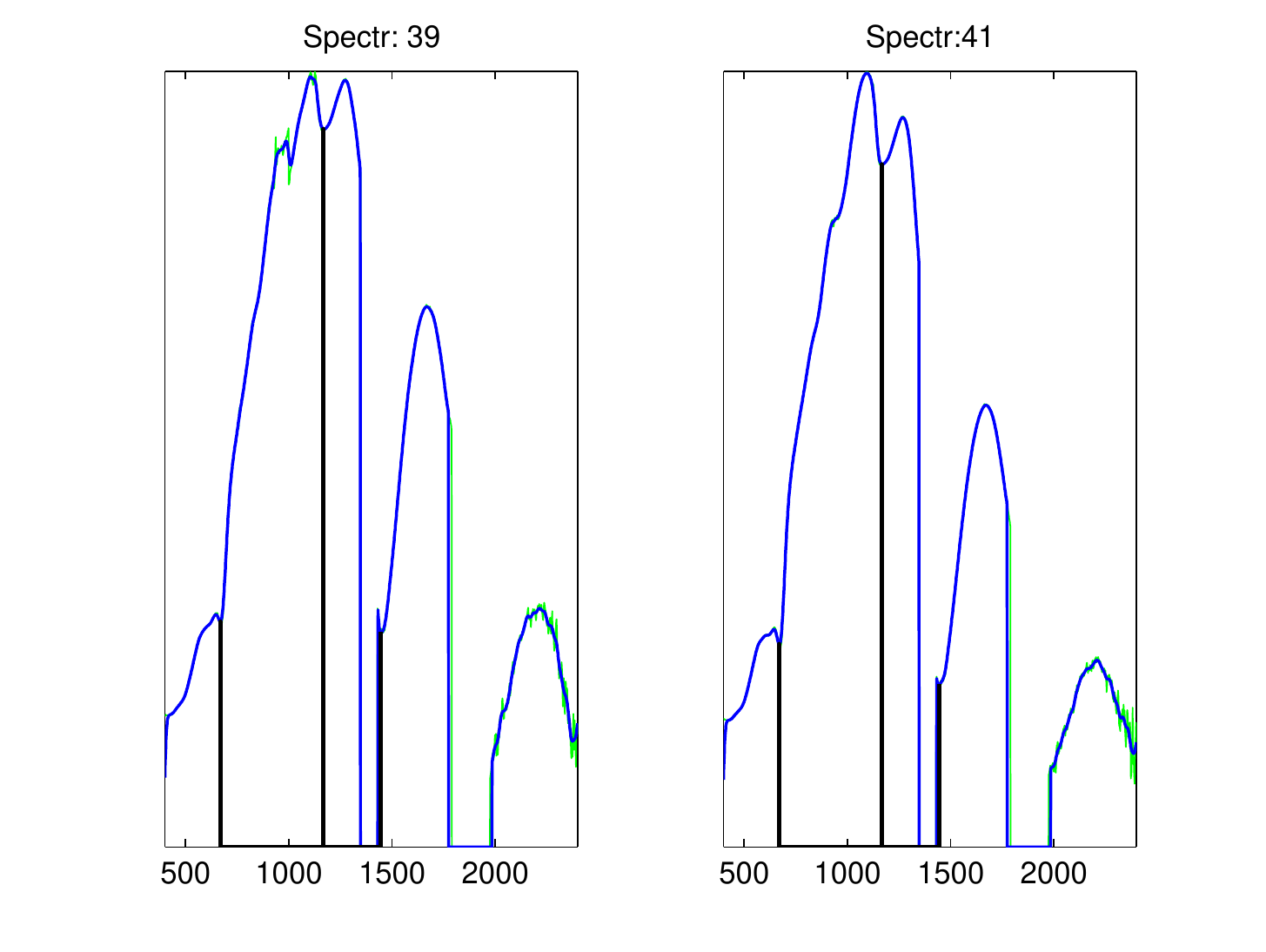}
\par\end{centering}

\caption{Spectrum $\#41$ from \emph{DB1} has three common deep minima with
the tested spectrum $\#39$, which also in \emph{DB1}.}

\label{com39_41} 
\end{figure}

We see from Fig. \ref{com39_38} that the materials that correspond
to spectra $\#38$ and $\#39$ are weakly connected one to the other.
A material that is more closely related to spectrum $\#39$ is spectrum
$\#41$, as can be seen from Fig. \ref{com39_41} where it is clearly
demonstrated that spectra $\#41$ and $\#39$ originate from the same
material. The hierarchical clustering algorithm found spectra $\#38$
and $\#39$ to have a weak similarity. However, it also found that
spectrum $\#38$ is the closest to $\#41$ and not $\#39$. 

\item [{Illustrations~for~the~full~database~(deep~minima):}] We looked
for similar spectra in \emph{DB1} (spectra $\#1$ to $\#173$). We
define two spectra to be similar if most of their features are similar.
We found that spectra $\#30$ and $\#34$ have seven common features
(deep minima) with spectrum $\#33$ as it is illustrated in Figs.
\ref{com33_30} and \ref{com33_34}. It is clear that spectra $\#30$
and $\#34$ belong to the same material as spectrum $\#33$. We also
searched \emph{DB1} for materials the strongly resemble spectrum $\#33$.
Our results showed that spectra $\#25$, $\#26$, $\#27$, $\#31$,
$\#32$ and $\#73$ have five common features with the tested spectrum
$\#33$. These findings are illustrated in Figs. \ref{com33_25}-\ref{com33_73}.
\end{description}

\begin{figure}[!h]

\begin{centering}
\includegraphics[width=0.35\paperheight]{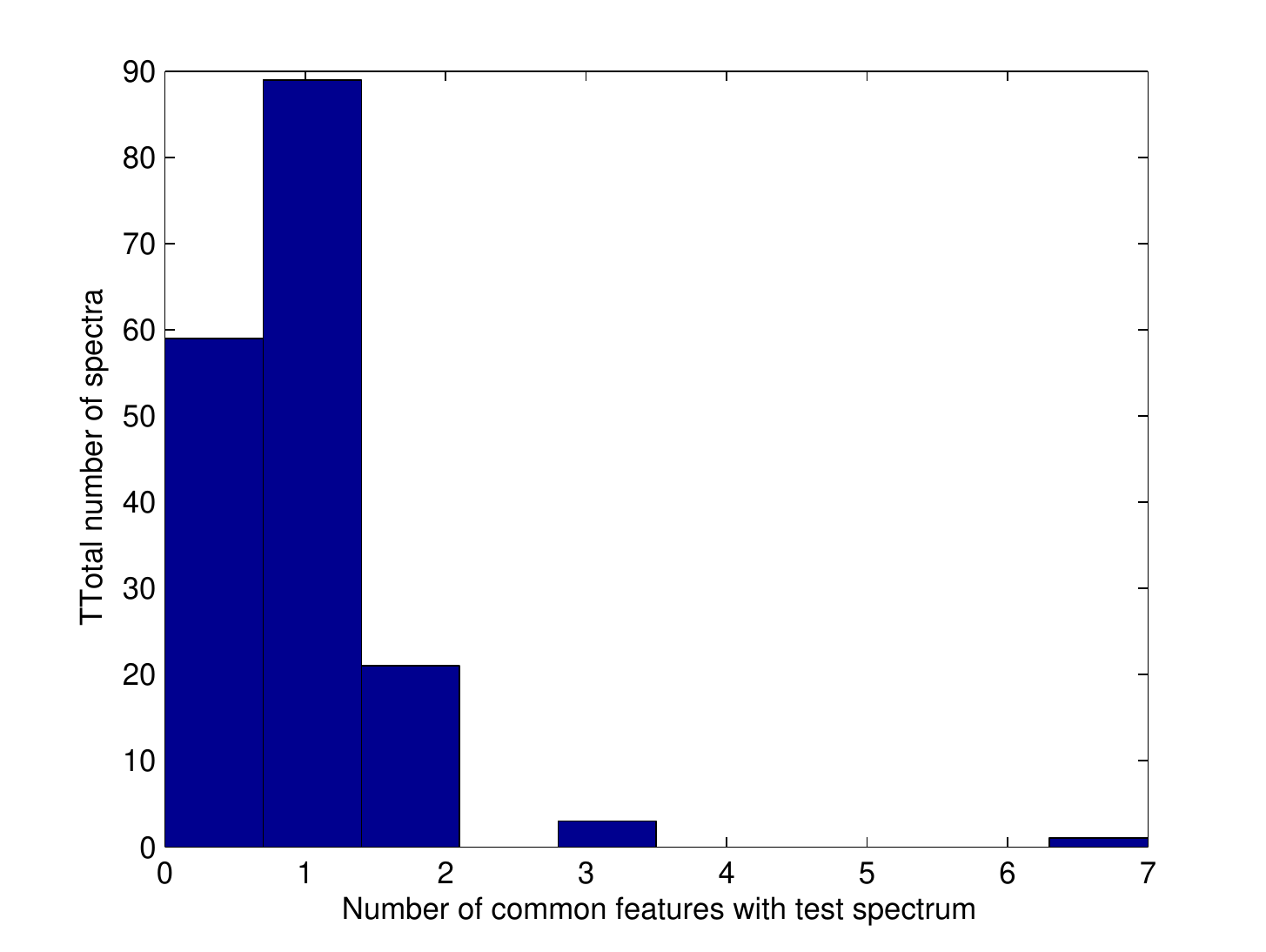} 
\par\end{centering}

\caption{Distribution of spectra $\#1$ to $\#173$ according to the number
of common inflection points with the tested spectrum $\#45$ in \emph{DB1}.}

\label{histf45i} 
\end{figure}

Obviously, spectra $\#30$ and $\#34$ are identical to spectrum $\#33$.
Moreover, it can be said that spectra $\#25$, $\#26$, $\#27$, $\#31$
and $\#32$ \emph{originate} from a similar material as spectrum $\#33$.
Most probably, this is also true for spectrum $\#73$. 
\begin{description}
\item [{Illustrations~for~the~full~database~(inflection~points):}] We
searched \emph{DB1} (spectra $\#1$ to $\#173$) for spectra that
have common inflection points with spectrum $\#45$. The histogram
in Fig. \ref{histf45i} displays the distribution of spectra from
\emph{DB1} according to the number of common inflection points with
the tested spectrum $\#45$.
\end{description}

\begin{figure}[!h]
\begin{centering}
\includegraphics[%bb=147bp 257bp 479bp 549bp,clip,
width=0.9\columnwidth,height=0.2\paperheight]{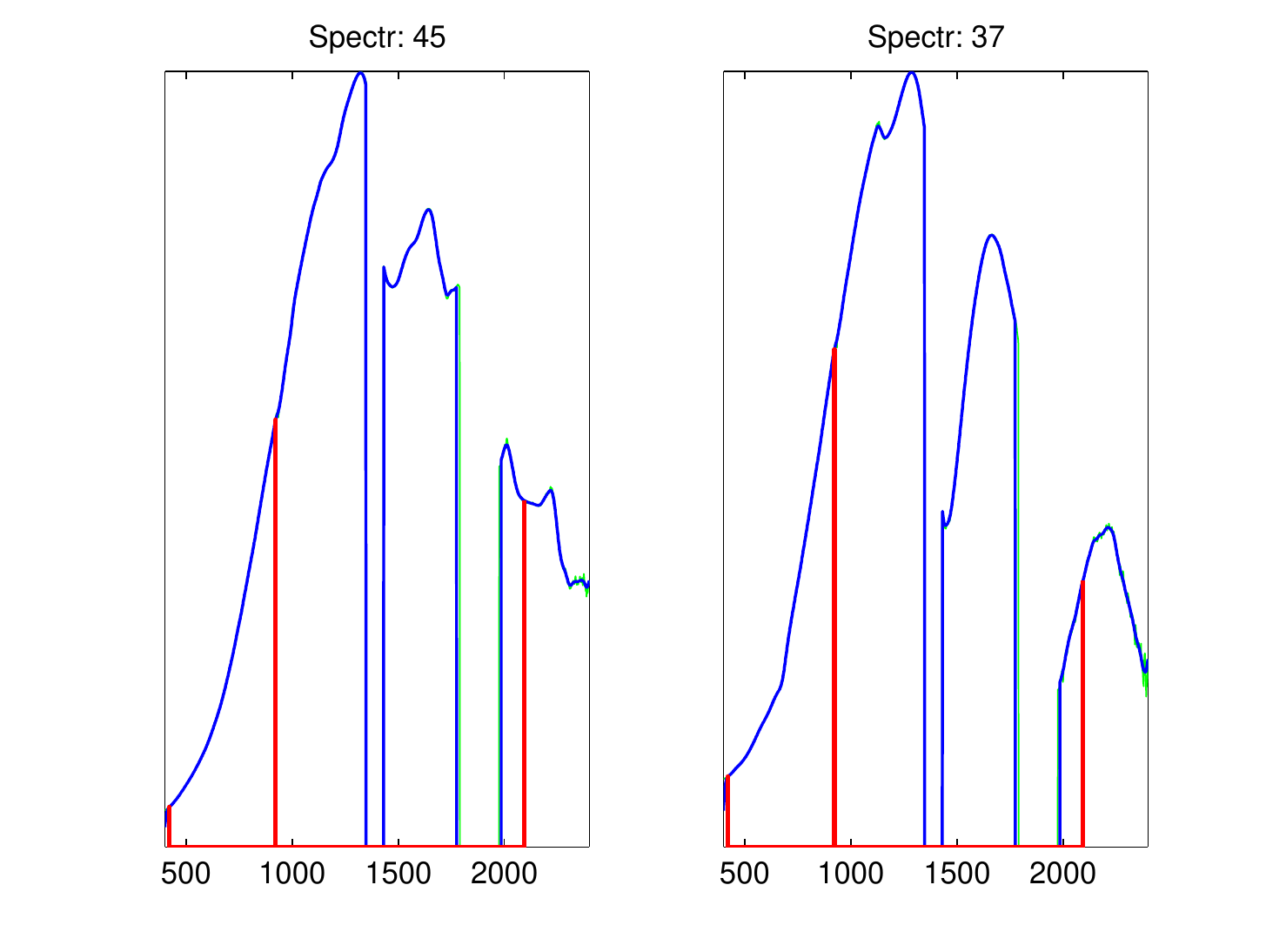} 
\par\end{centering}

\caption{Spectrum $\#37$ (right) in \emph{DB1} has three common inflection
points with the tested spectrum $\#45$ (left) in \emph{DB1}.}

\label{com45_37i} 
\end{figure}

\begin{figure}[!h]

\begin{centering}
\includegraphics[%bb=147bp 257bp 479bp 549bp,clip,
width=0.9\columnwidth,height=0.2\paperheight]{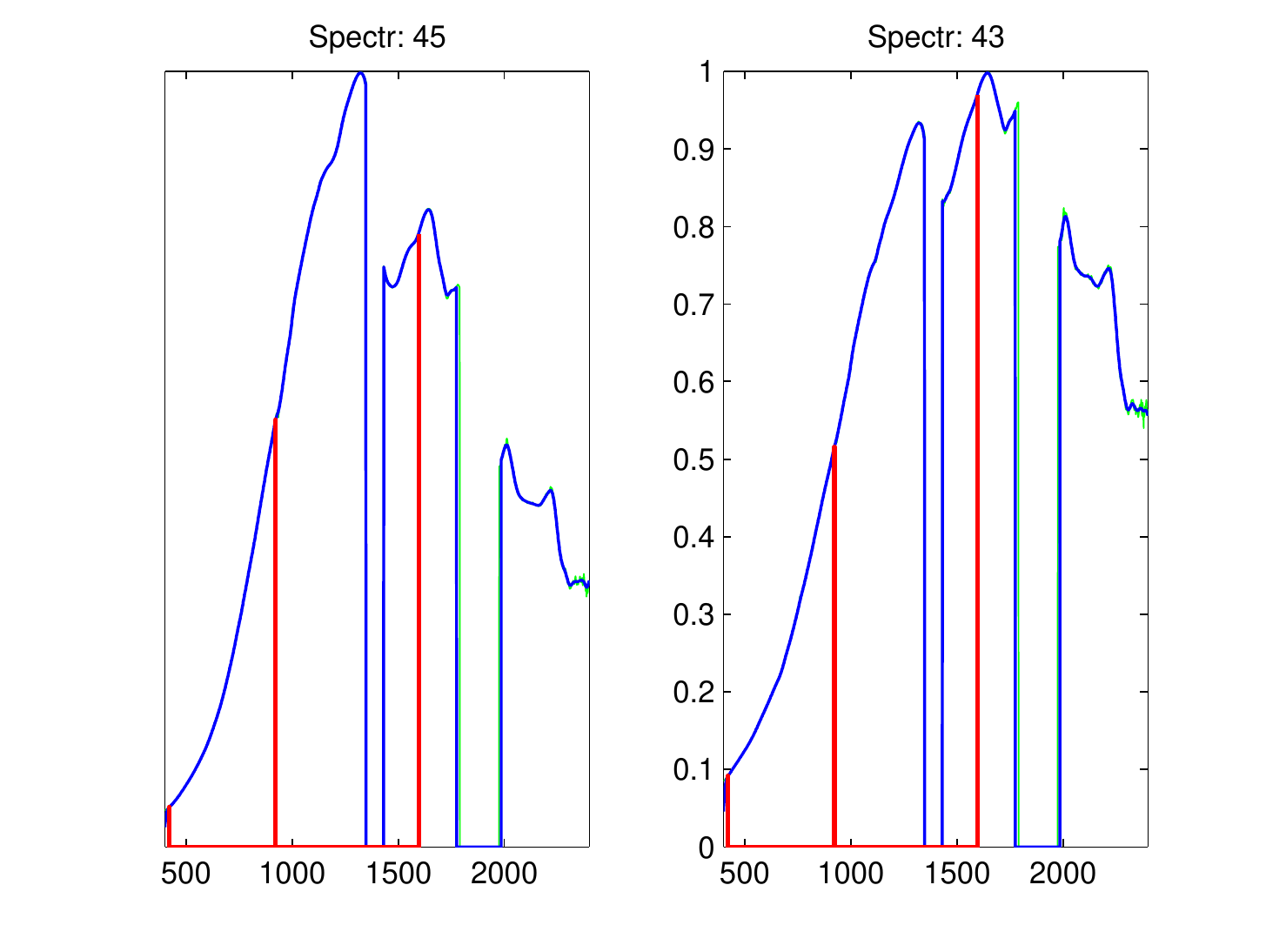}
\par\end{centering}

\caption{Spectrum $\#43$ (right) in \emph{DB1} has three common inflection
points with the tested spectrum $\#45$ (left) in \emph{DB1}.}

\label{com45_43i} 
\end{figure}

\begin{figure}[!h]

\begin{centering}
\includegraphics[%bb=147bp 257bp 479bp 549bp,clip,
width=0.9\columnwidth,height=0.2\paperheight]{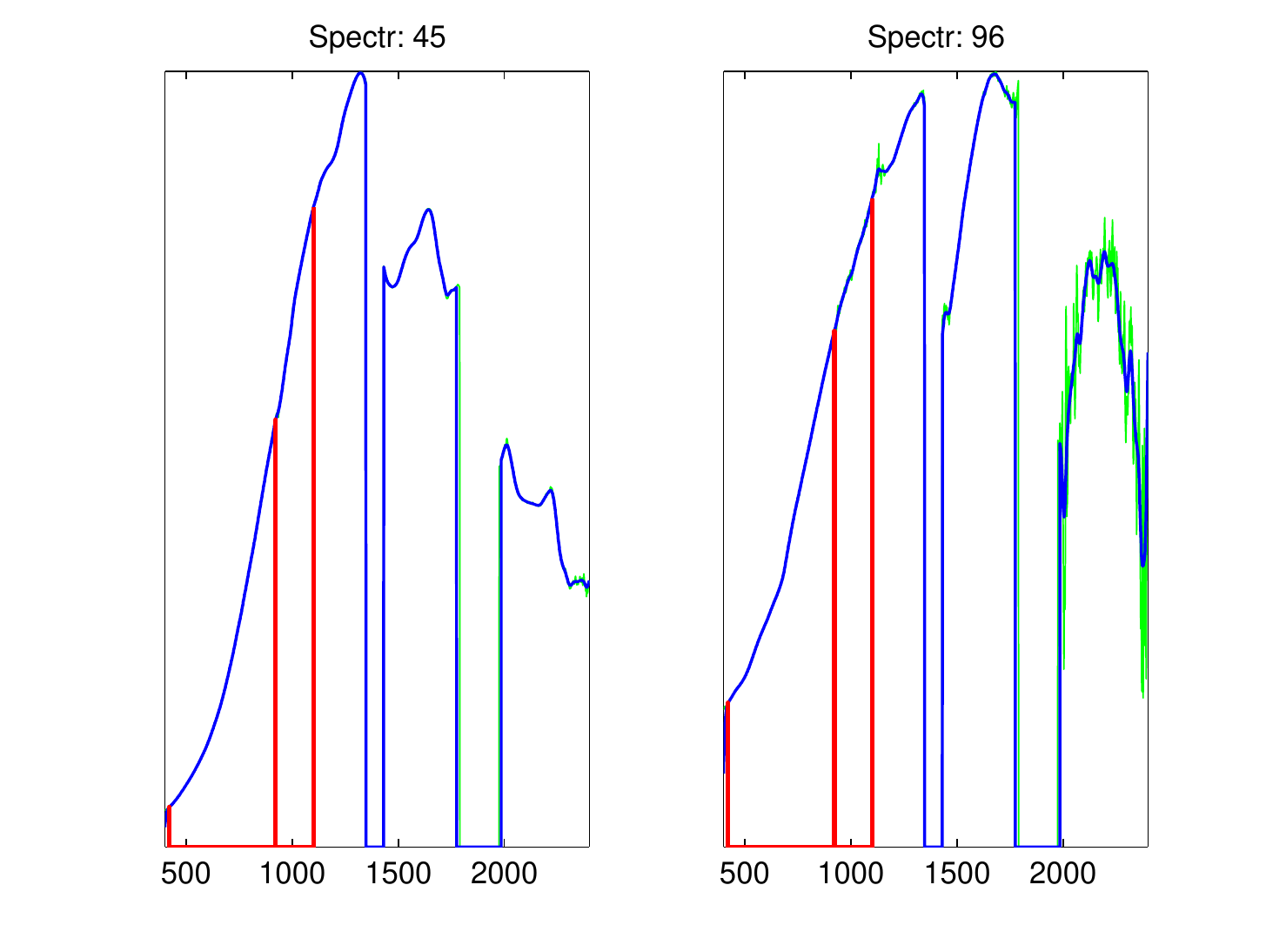}
\par\end{centering}

\caption{Spectrum $\#96$ (right) in \emph{DB1} has three common inflection
points with the tested spectrum $\#45$ (left) in \emph{DB1}.}

\label{com45_96i} 
\end{figure}

Spectra $\#37$, $\#43$ and $\#96$ have three common features with
spectrum $\#45$ as can be noticed in Figs \ref{com45_37i}-\ref{com45_96i}.

We can see from Figs. \ref{com45_37i}--\ref{com45_96i} that, most
probably, spectrum $\#43$ is derived from the same material as $\#45$,
however, spectra $\#37$ and $\#96$ are very weakly related to spectrum
$\#45$. In general, common \emph{deep minima} features indicate a
closer relation between materials than \emph{inflection points} do. 

The hierarchical clustering algorithm failed to produce the deep-minima
driven results of the proposed algorithm. Namely, spectra $\#30$,
$\#34$ and $\#33$ were put in different clusters. Moreover, spectra
$\#25$, $\#26$, $\#27$, $\#31$ and $\#32$ were also put in different
clusters. Specifically, the clusters that include the above spectra
were: {[}\textbf{30}{]} (singleton), {[}28 \textbf{31}{]}, {[}21 32
\textbf{33}{]}, {[}23 \textbf{34}{]}, {[}\textbf{25}{]}, {[}\textbf{26}{]},
{[}\textbf{27}{]}, {[}\textbf{73}{]}. The hierarchical clustering
succeeded in finding the weak inflection point correlations. However
it failed in finding the connection between spectra $\#43$and $\#45$.
Spectra $\#37$, $\#43$, $\#45$ and $\#96$ were put in the following
clusters: {[}35 \textbf{37} 42 \textbf{43}{]}, {[}\textbf{45} 47 49{]}
and {[}91 \textbf{96} 97 98{]}.

\subsubsection{Examples from \emph{DB2}}

In this section, the reference database is \emph{DB2} and the tested
spectra also belong to the same reference database. Spectra from this
database were smoothed. Typically, they have less characteristic features
than spectra from \emph{DB1}. Therefore, the identification process
of these spectra uses all four sets of features (deep and shallow
minima, flat intervals and inflection points).

\begin{figure}[!h]

\begin{centering}
\includegraphics[width=0.35\paperheight]{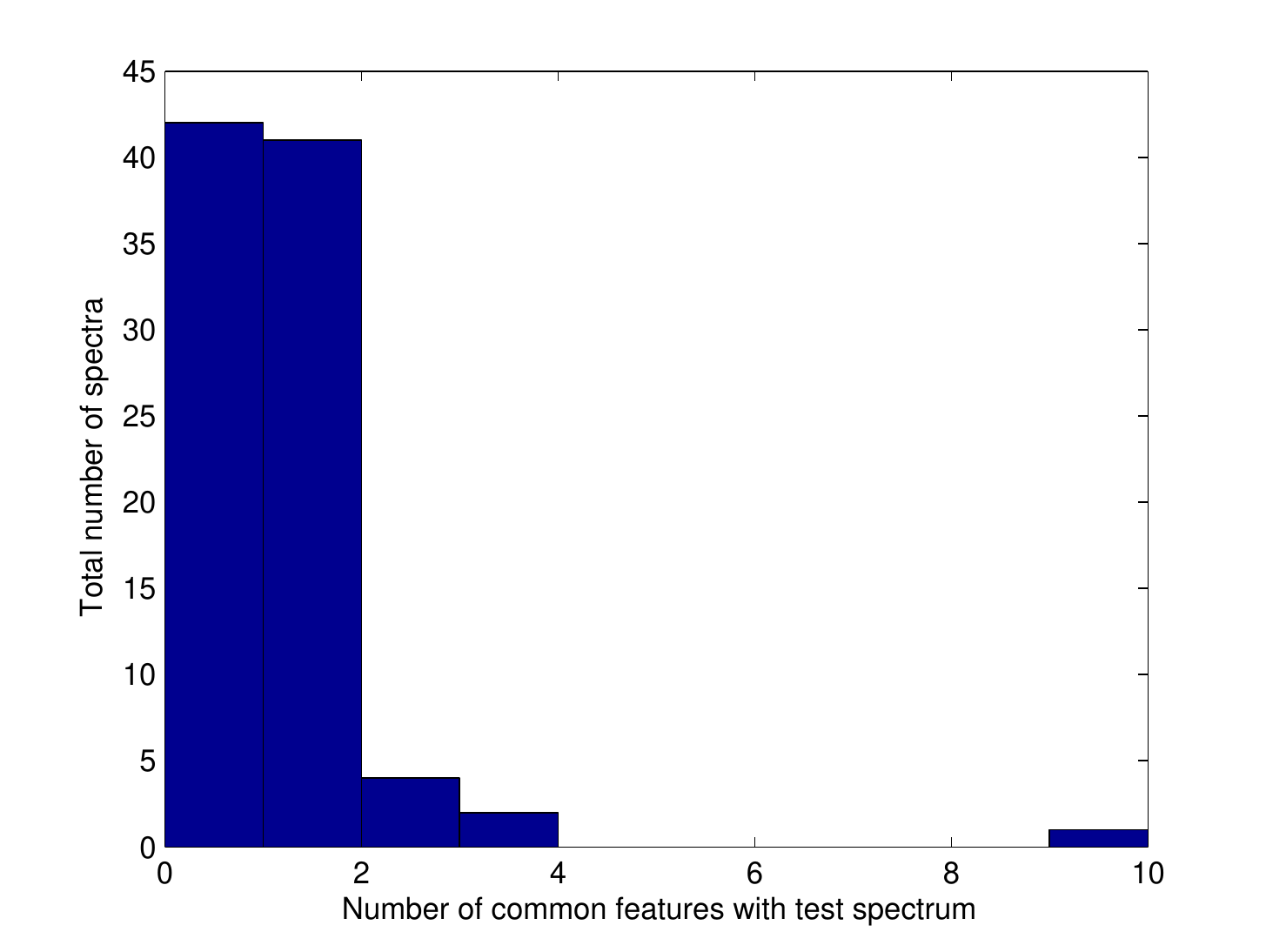}
\par\end{centering}

\caption{Distribution of spectra $\#1$ to $\#90$ from \emph{DB2} according
to the number of all common features with the tested spectrum $\#61$
from \emph{DB2}.}

\label{hist61_2} 
\end{figure}

We looked for materials in \emph{DB2} that have common features with
spectrum $\#61$ from the same database (\emph{DB2}). The histogram
in Fig. \ref{hist61_2} displays the distribution of spectra from
the reference database \emph{DB2} (spectra $\#1$ to $\#90$) according
to the number of \emph{all} common features with the tested spectrum
$\left(\#61\right)$.

The identification algorithm found that six spectra -- $\#19$, $\#23$,
$\#32$, $\#47$, $\#56$ and $\#69$ -- have at least three common
features with spectrum $\#61$. Two of them -- $\#47$ and $\#69$
-- have four common features with spectrum $\#61$.

\begin{figure}[!h]

\begin{centering}
\includegraphics[%bb=147bp 257bp 479bp 549bp,clip,
width=0.9\columnwidth,height=0.2\paperheight]{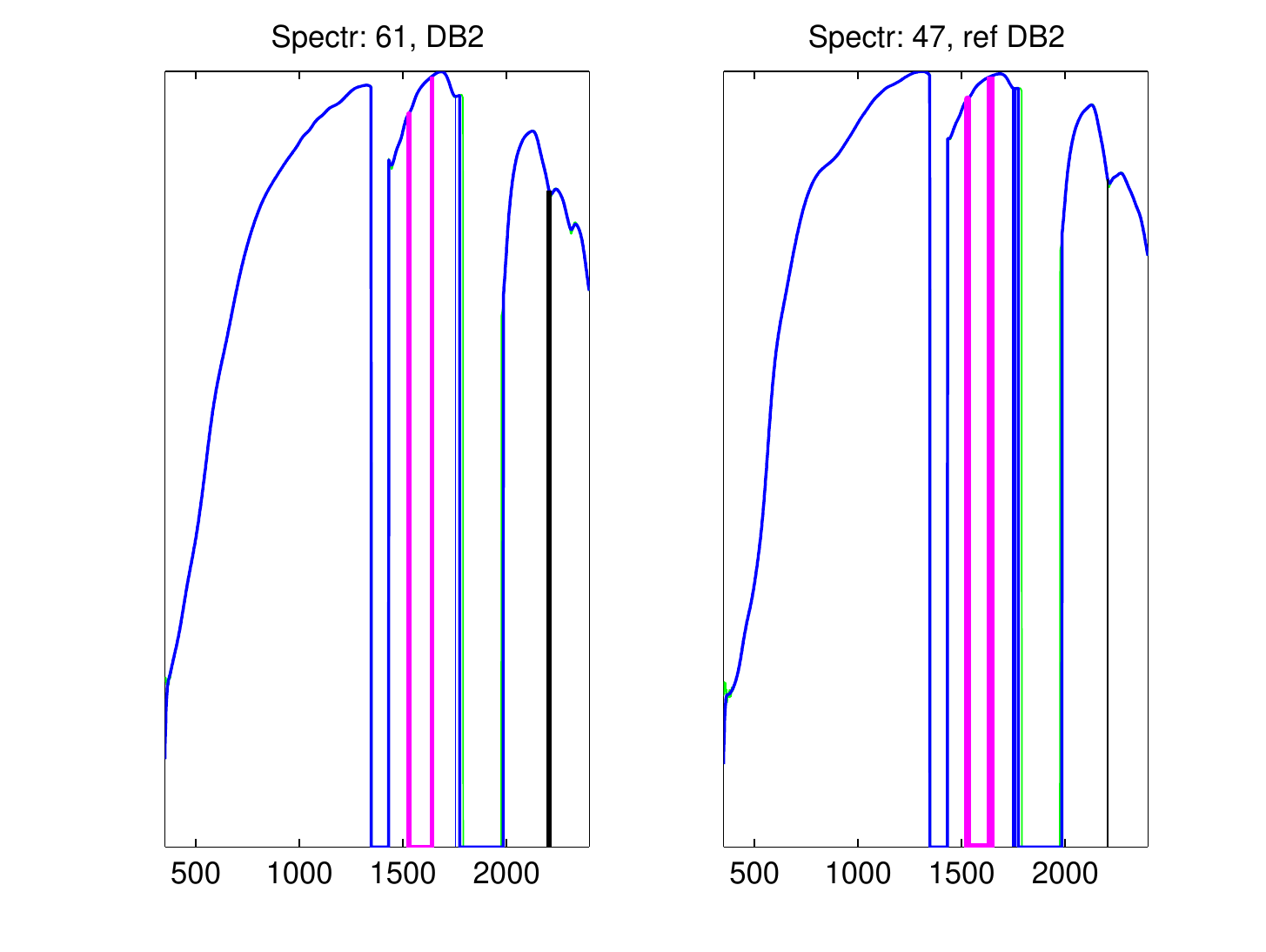}
\par\end{centering}

\caption{Spectrum $\#47$ (right) from \emph{DB2} has four common features
(one deep minimum, one shallow minimum and two flat intervals) with
the tested spectrum $\#61$ (left) from \emph{DB2}.}

\label{com61_47_2} 
\end{figure}

\begin{figure}[!h]

\begin{centering}
\includegraphics[%bb=147bp 257bp 479bp 549bp,clip,
width=0.9\columnwidth,height=0.2\paperheight]{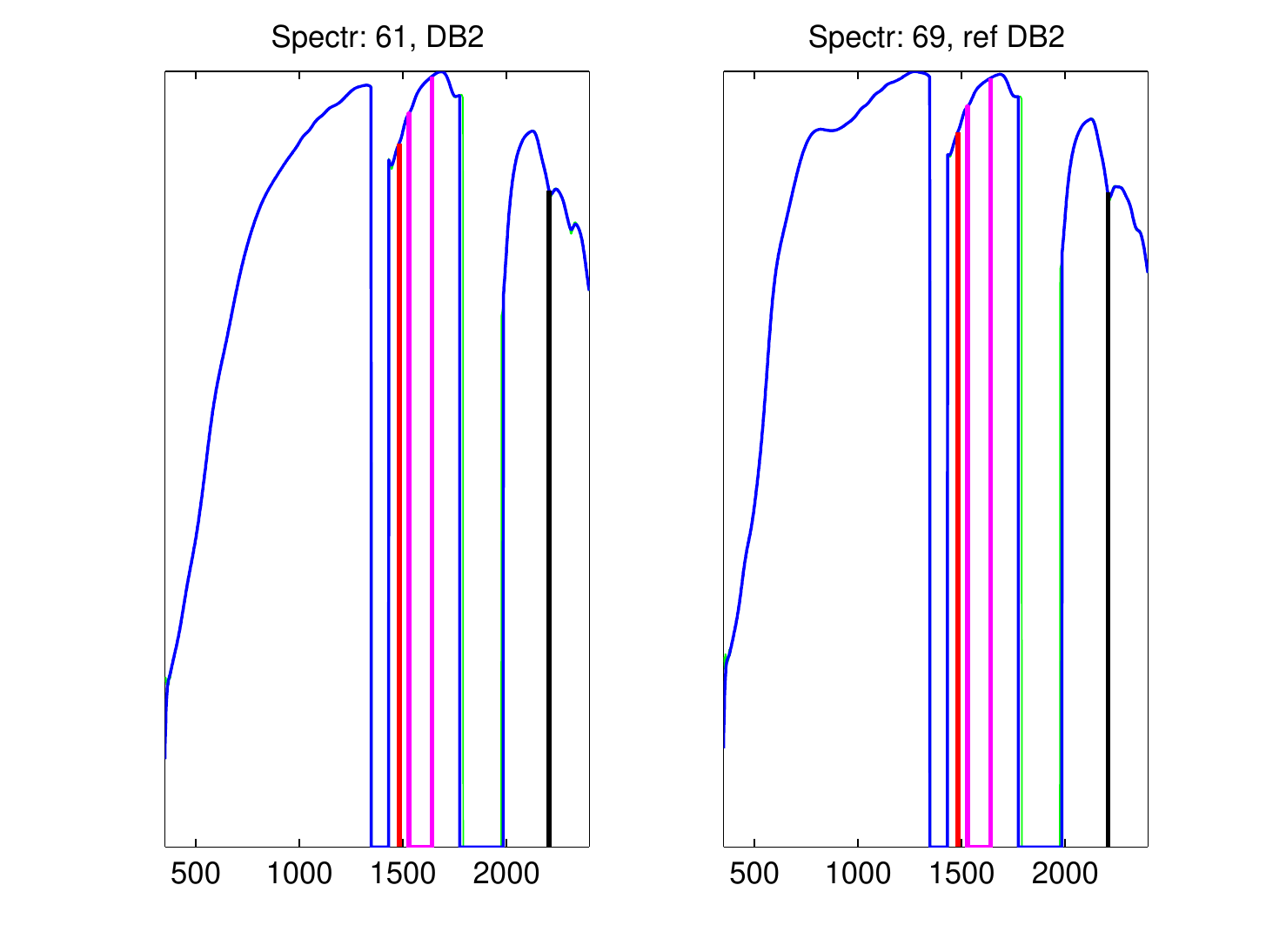} 
\par\end{centering}

\caption{Spectrum $\#69$ (right) from \emph{DB2} has four common features
(one deep minimum, one shallow minimum and two flat intervals) with
the tested spectrum $\#61$ (left) from \emph{DB2}.}

\label{com61_69_2} 
\end{figure}

Our findings suggest that, spectra $\#47$ and $\#69$ are related
to the same material that spectrum $\#61$ represents.

\begin{figure}[!h]

\begin{centering}
\includegraphics[%bb=147bp 257bp 479bp 549bp,clip,
width=0.9\columnwidth,height=0.2\paperheight]{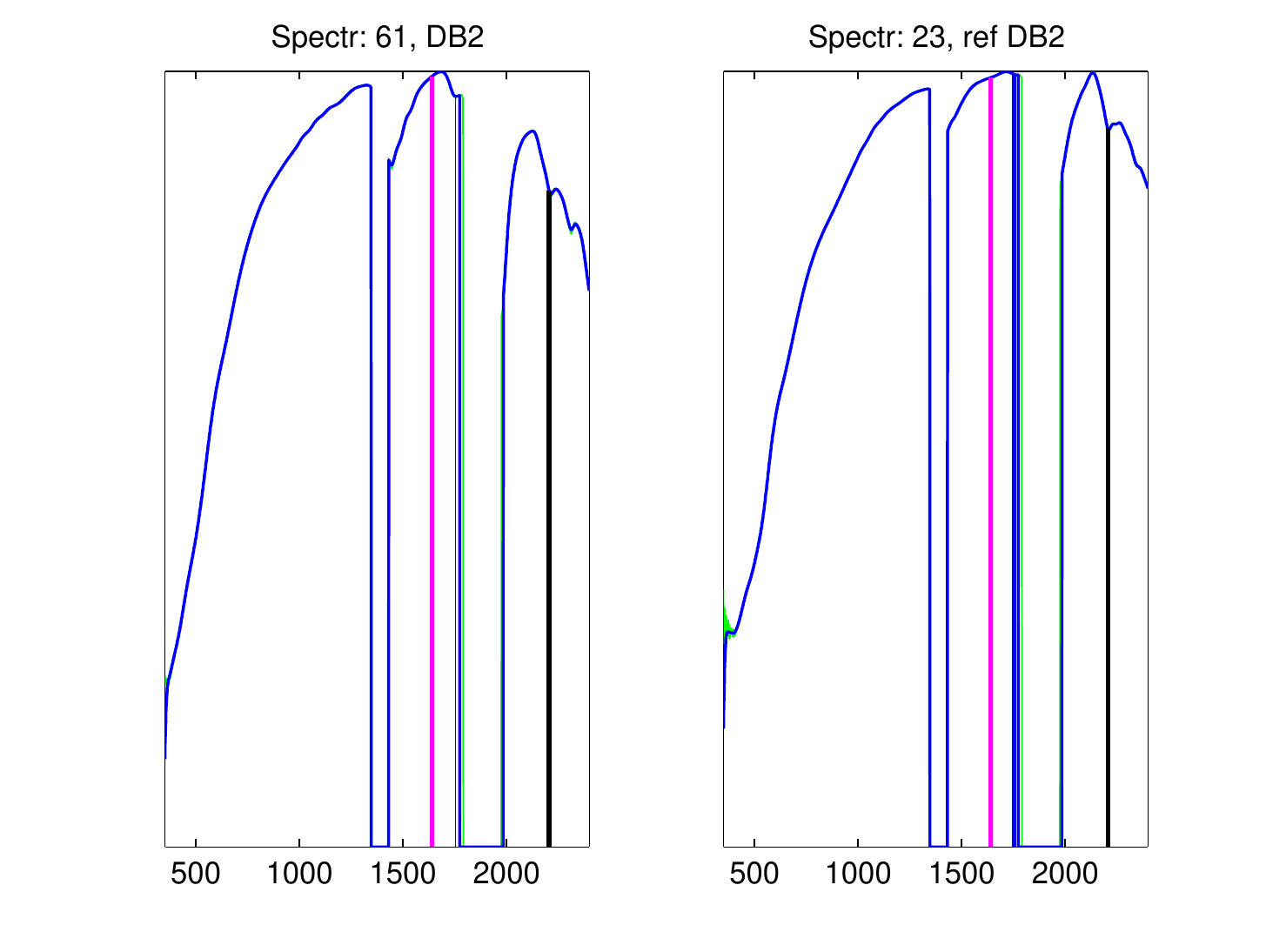}
\par\end{centering}

\caption{Spectrum $\#23$ (right) from \emph{DB2} has three common features
(one deep minimum, one shallow minimum and one flat intervals) with
the tested spectrum $\#61$ (left) from \emph{DB2}.}

\label{com61_23_2} 
\end{figure}

\begin{figure}[!h]

\begin{centering}
\includegraphics[%bb=147bp 257bp 479bp 549bp,clip,
width=0.9\columnwidth,height=0.2\paperheight]{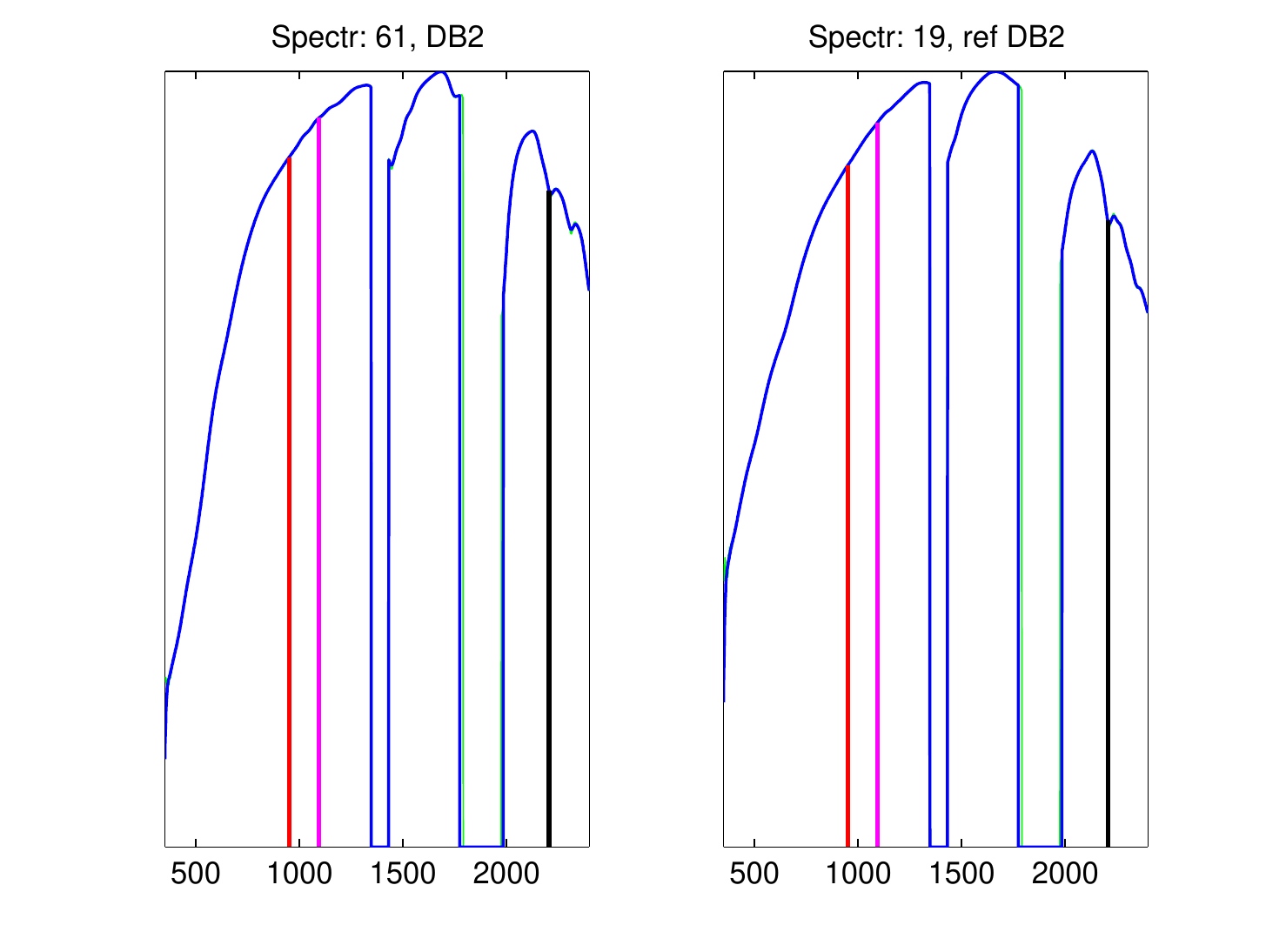}
\par\end{centering}

\caption{Spectrum $\#19$ (right) from \emph{DB2} has three common features
(one deep minimum, one shallow minimum and one flat intervals) with
the tested spectrum $\#61$ (left) from \emph{DB2}.}

\label{com61_19_2} 
\end{figure}

\begin{figure}[!h]

\begin{centering}
\includegraphics[%bb=147bp 257bp 479bp 549bp,clip,
width=0.9\columnwidth,height=0.2\paperheight]{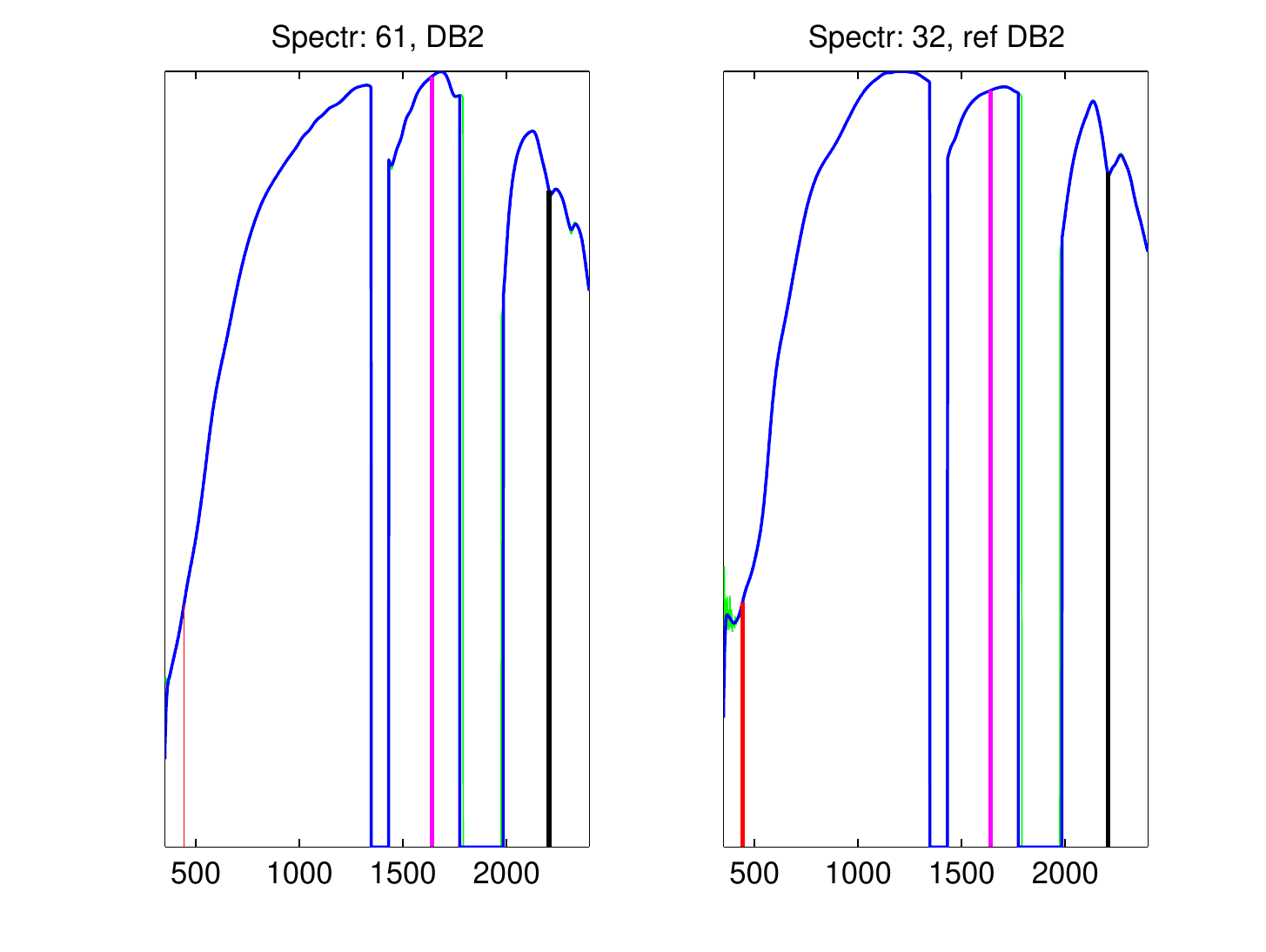}
\par\end{centering}

\caption{Spectrum $\#32$ (right) from \emph{DB2} has three common features
(one deep minimum, one shallow minimum and one flat intervals) with
the tested spectrum $\#61$ (left) from \emph{DB2}.}

\label{com61_32_2} 
\end{figure}

\begin{figure}[!h]
\begin{centering}
\includegraphics[%bb=147bp 257bp 479bp 549bp,clip,
width=0.9\columnwidth,height=0.2\paperheight]{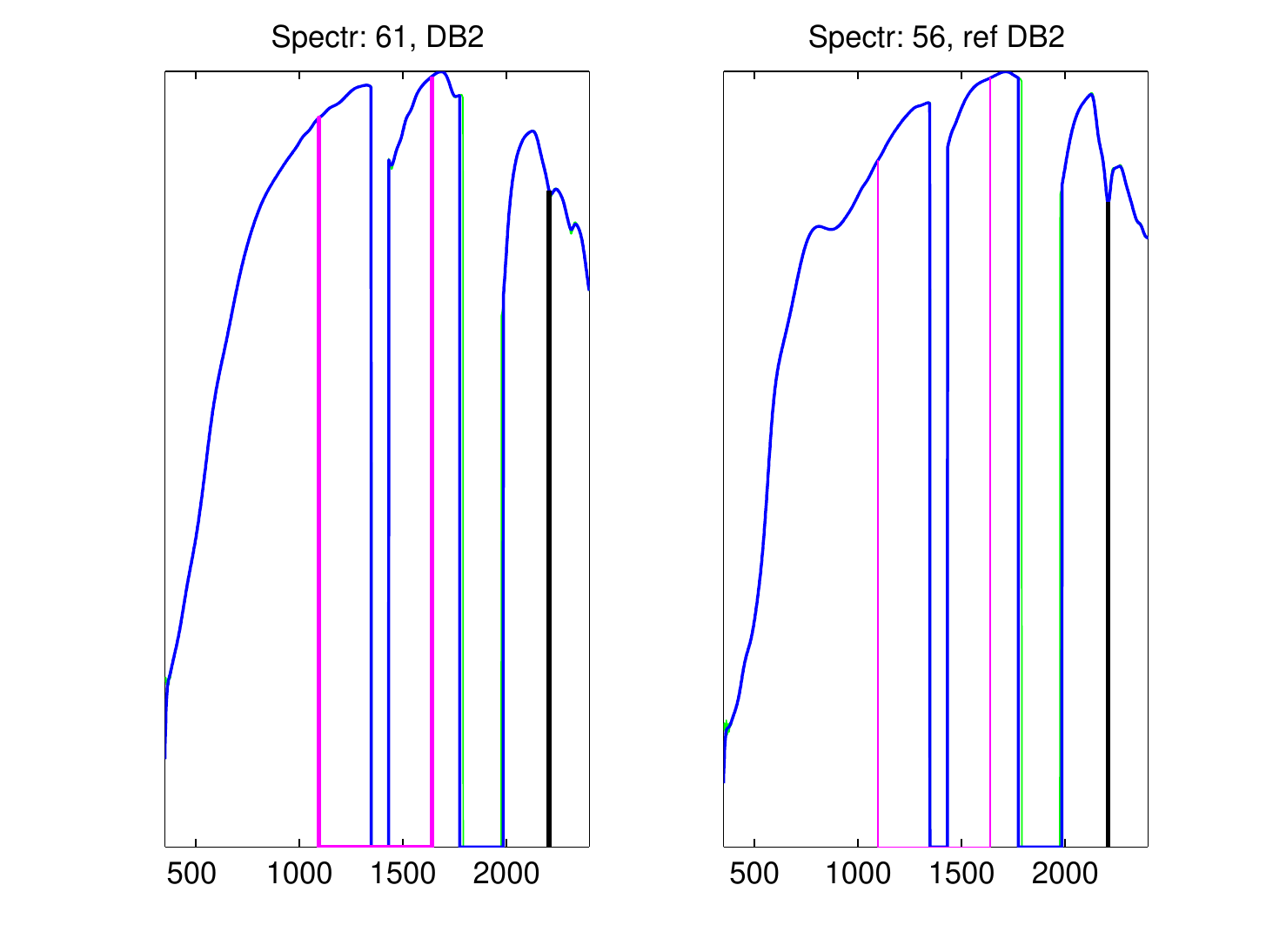}
\par\end{centering}

\caption{Spectrum $\#56$ (right) from \emph{DB2} has three common features
(one deep minimum, one shallow minimum and one flat intervals) with
the tested spectrum $\#61$ (left) from \emph{DB2}.}

\label{com61_56_2} 
\end{figure}

Figures \ref{com61_23_2}-\ref{com61_56_2} suggest that most probably,
spectra $\#19$, $\#23$, $\#32$ and $\#56$ are related to materials
whose physical properties are similar to the physical properties of
the material that corresponds to spectrum $\#61$.

The hierarchical clustering put spectrum $\#69$ in the same cluster
as $\#61$ (their cluster also included spectra $\#58$, $\#62$ and
$\#90.$ However, spectrum $\#47$ was put in a different cluster
than $\#61$'s. Specifically, spectrum $\#47$ was put in a singleton
cluster. Furthermore, spectra $\#19$, $\#23$, $\#32$ and $\#56$
were put in different clusters than $\#61$'s. Specifically, they
were put in the following clusters: {[}\textbf{56}{]} (singleton),
{[}18 \textbf{19} 59{]}, {[}\textbf{32} 52{]}, {[}6 22 \textbf{23}
26 53 68 72{]}.

\subsection{Example from different databases }

In this section, the reference database is \emph{DB1} and the tested
spectrum belongs to \emph{DB2}. All four types of features (deep and
shallow minima, flat intervals and inflection points) were used for
the identification of the tested spectra.

\begin{figure}[!h]

\begin{centering}
\includegraphics[width=0.37\paperheight]{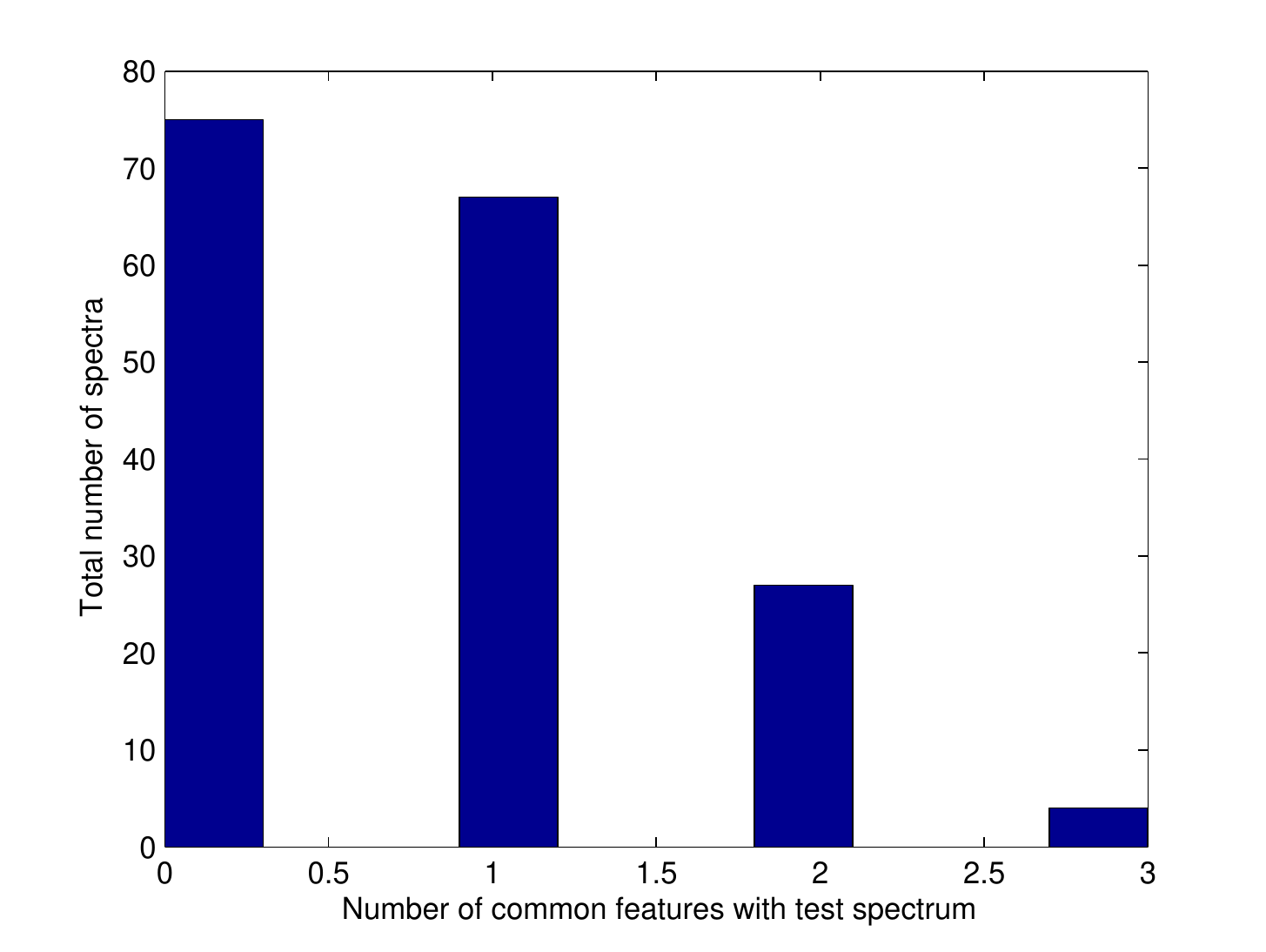} 
\par\end{centering}

\caption{Distribution of spectra $\#1$ to $\#173$ from \emph{DB1} according
to the number of common features (all types) with the tested spectrum
$\#61$ from \emph{DB2}.}

\label{hist61_2_1} 
\end{figure}

First we looked for spectra from \emph{DB1} that are similar to spectrum
$\#61$ from \emph{DB2}. The histogram in Fig. \ref{hist61_2_1} displays
the distribution of spectra from the reference database \emph{DB1}
(spectra $\#1$ to $\#173$) according to the number of common features
with the tested spectrum $\#61$ from \emph{DB2}.

We found that four spectra $\#35$, $\#47$, $\#70$ and $\#75$ have
three common features with spectrum $\#61$.

\begin{figure}[!h]

\begin{centering}
\includegraphics[%bb=147bp 257bp 479bp 549bp,clip,
width=0.9\columnwidth,height=0.2\paperheight]{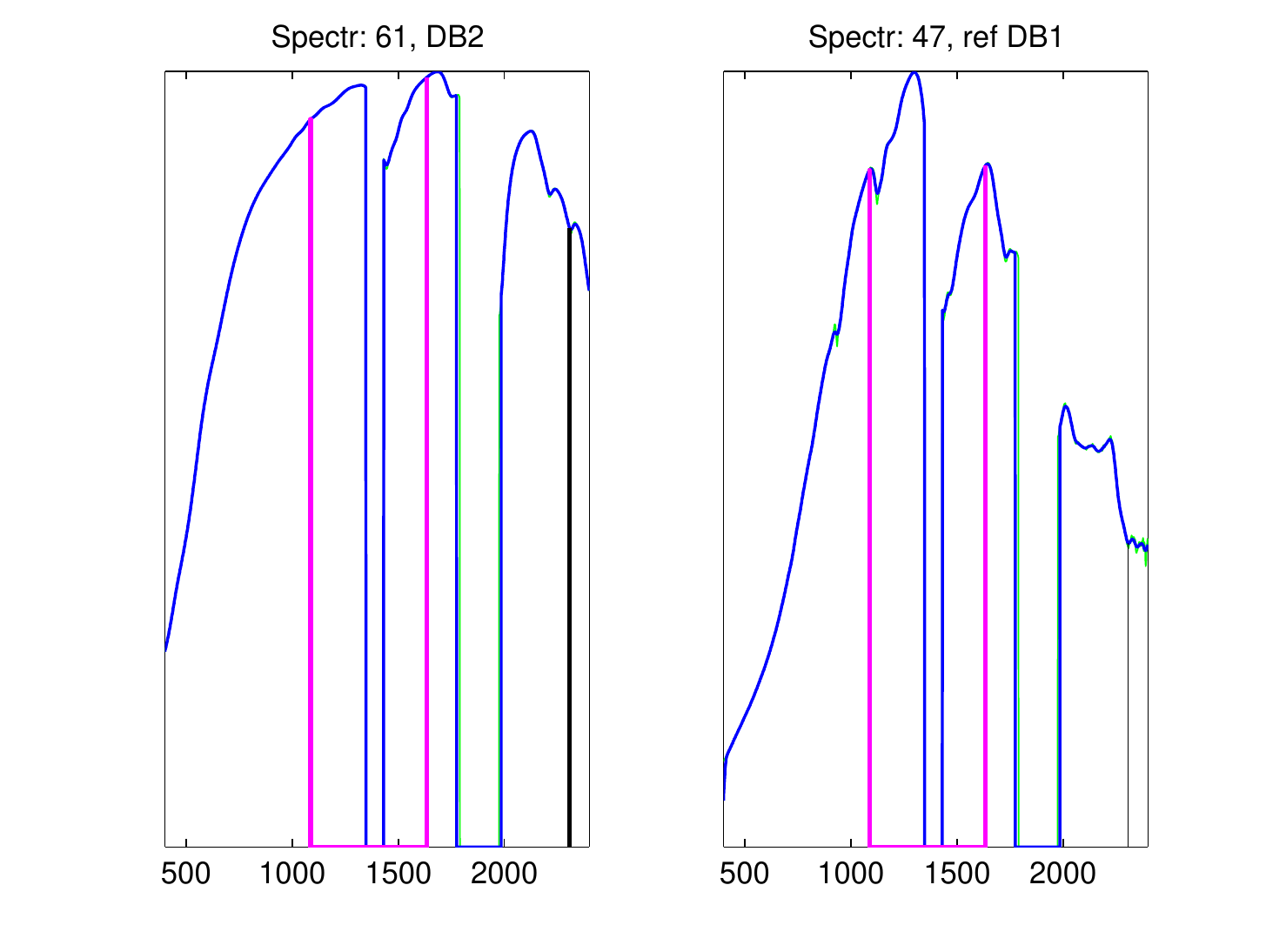} 
\par\end{centering}

\caption{Spectrum $\#47$ (right) from \emph{DB1} has three common features
(one deep minimum and two flat intervals) with the tested spectrum
$\#61$ (left) from \emph{DB2}.}

\label{com61_47_2_1} 
\end{figure}

\begin{figure}[!h]

\begin{centering}
\includegraphics[%bb=147bp 257bp 479bp 549bp,clip,
width=0.9\columnwidth,height=0.2\paperheight]{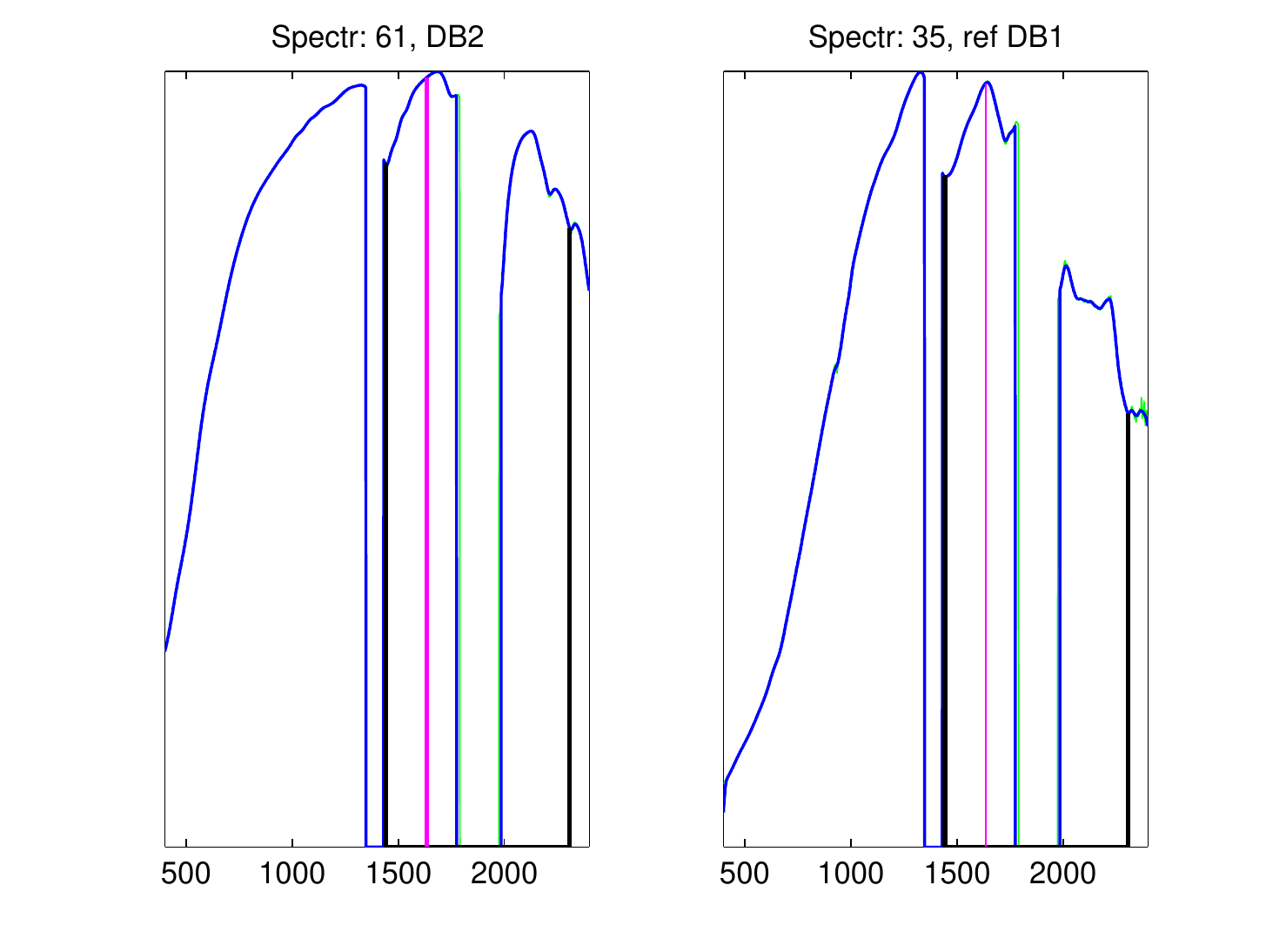}
\par\end{centering}

\caption{Spectrum $\#35$ (right) from \emph{DB1} has three common features
(two deep minima and one flat intervals) with the tested spectrum
$\#61$ (left) from \emph{DB2}.}

\label{com61_35_2_1} 
\end{figure}

\begin{figure}[!h]

\begin{centering}
\includegraphics[%bb=147bp 257bp 479bp 549bp,clip,
width=0.9\columnwidth,height=0.2\paperheight]{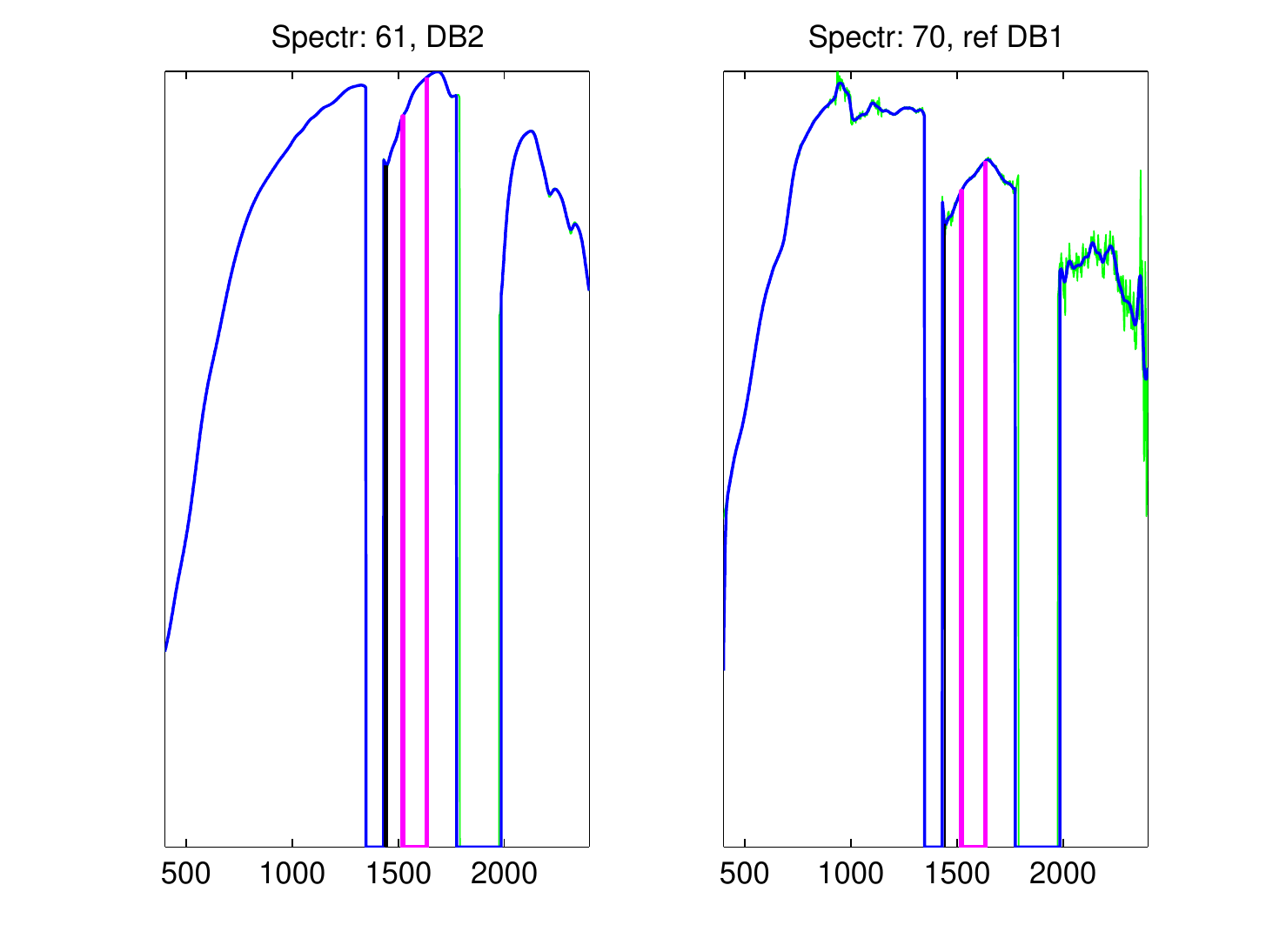} 
\par\end{centering}

\caption{Spectrum $\#70$ (right) from \emph{DB1} has three common features
(one deep minimum and two flat intervals) with the tested spectrum
$\#61$ (left) from \emph{DB2}.}

\label{com61_70_2_1} 
\end{figure}

\begin{figure}[!h]

\begin{centering}
\includegraphics[%bb=147bp 257bp 479bp 549bp,clip,
width=0.9\columnwidth,height=0.2\paperheight]{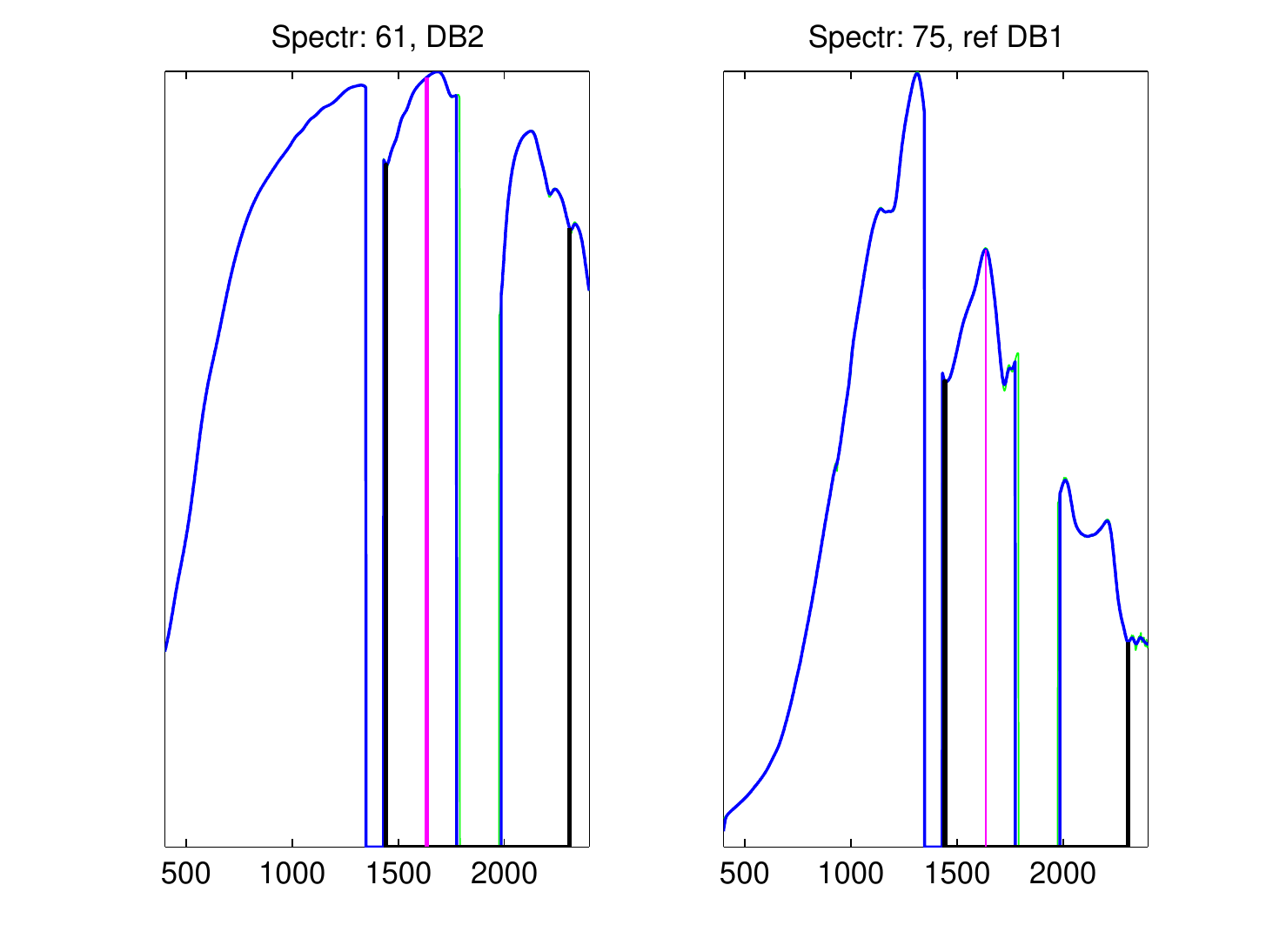} 
\par\end{centering}

\caption{Spectrum $\#75$ (right) from \emph{DB1} has three common features
(two deep minima and one flat intervals) with the tested spectrum
$\#61$ (left) from \emph{DB2}.}

\label{com61_75_2_1} 
\end{figure}

Although the spectra graphs look different, the findings show that
the materials associated with spectra $\#35$, $\#47$, $\#70$ and
$\#75$ from \emph{DB1} share some common physical properties with
spectrum $\#61$ from $DB2$. The hierarchical clustering algorithm
clustered spectrum $\#61$ from DB2 only with spectrum $\#103$ from
DB1.

\section{Conclusion and discussion\label{conclusion} }

In this chapter, dimensionality reduction was utilized to uniquely
characterize substances via their spectral signatures. Specifically,
characterizing features are extracted from the spectrum. Two complementary
methods were presented for the computation and analysis of spectral
signatures.
\begin{description}
\item [{Exact~search:}] A given spectrum is compared to all the spectra
in a spectra database and only exact matches are sought after. Seven
pairs of identical spectra were found. In most cases, only a small
number of the available four feature types are utilized. Thus, we
argue that the designed scheme is generic and can be applied to other
databases. %Properly modified, the scheme may serve as a base for an unmixing algorithm.\\
The algorithm allows different spectra in a population of spectra
to have different resolutions (sampling rates). 
\item [{Approximate~search:}] The algorithm, which was described in this
chapter, efficiently reveals the spectra in the reference database
that posses similar physical properties as a spectrum tested for identification.
The extent of the relation is interactively defined by the user. If
the tested spectrum belongs to the reference database, then the algorithm
correctly identifies it and finds all the spectra originating from
the same material as well as the spectra that belong to a similar
group of materials. %The scheme can be utilized for solving the unmixing problem.%The proposed algorithms was also tested on a database consisting of 1000 spectral signatures, in reflectance, of natural targets that were measured by a field spec (FR) spectro radiometer manufactured by ASD (Boulder, Colorado). The measurements range is between 400 to 2500 nm with a spectral sampling resolution of 1nm. All the spectral signatures are of natural targets with reoccurrence of a few targets measured in different spectral conditions and instruments. Some of the measurements were made in sun light conditions and other were made with a contact probe. No noise was added to the measured data. The analysis of the method described in this chapter was done on authentic reflectance measurements of real field measurements and was crowned with success.

\end{description}
Other physical features can be chosen to either replace or enhance
the current features that were described in this thesis such as the
area between two minima, etc. Therefore, the algorithms can be classified
as generic classifiers.

The results of the proposed algorithm were compared to those obtained
by hierarchical clustering. It was found that in most cases hierarchical
clustering fails to find connections between materials unless they
are very similar. This renders hierarchical clustering to be suitable
for exact search but not for inexact search.

\chapter{Wavelet-Based Acoustic Detection of Moving Vehicles\label{cha:Wavelet-based-acoustic}}

We propose a robust algorithm to detect the arrival of a vehicle of
arbitrary type in the presence of various noise types. This is achieved
by analysis of the acoustic signatures of vehicles and their comparison
with an existing database of recorded and processed acoustic signals.
We combine a construction of a training database of acoustic signatures
of signals emitted by vehicles using the distribution of the energies
among blocks of wavelet packet coefficients with a procedure of random
search for a near-optimal footprint (RSNOFP). This construction achieves
a minimum number of false alarms even under severe conditions such
as the presence of signals emitted by acoustic sources that differ
from those that appear in the training database. Such sources include
helicopters, airplanes, wind, speech, footsteps etc. The proposed
algorithm is robust even when the surrounding conditions of the test
sites are completely different from the conditions where the training
signals were recorded. The proposed technique has many algorithmic
variations. For example, it can be used to classify different types
of vehicles. The proposed algorithm is a generic solution for process
control that is based on a learning phase (training) followed by an
automatic real time detection. This is an example where dimensionality
reduction is utilized for classification.

\section{Introduction}

The goal of the algorithm in this chapter is to detect the arrival
of vehicles of arbitrary types such as various cars, vans, jeeps and
trucks via the analysis of their acoustic signatures with a minimal
number of false alarms. The algorithm uses an existing database of
recorded acoustics signals. The problem is complex due to the high
variability in vehicles types and the diversity of the surrounding
conditions that may contain sounds emitted by planes, helicopters,
speech, wind, steps, to name a few. These sounds also appear in many
recordings in the training database. In addition, the velocities of
the vehicles, their distances from the receiver, the roads where the
vehicles traveled on are highly diverse as well. Consequently, this
affected the recorded acoustics.

A successful detection depends on the constructed acoustic signatures
that were built from characteristic features. These signatures enable
to discriminate between vehicle (V) and non-vehicle (N) classes. Acoustic
signals emitted by vehicles have a quasi-periodic structure. It stems
from the fact that each part of the vehicle emits a distinct acoustic
signal which contains only a few dominating bands in the frequency
domain. As the vehicle moves, the conditions change and the configuration
of these bands may vary, however, the general disposition remains.
Therefore, we assume that the acoustic signature for the class of
signals emitted by a certain vehicle is obtained as a combination
of the inherent energies in blocks of wavelet packet coefficients
of the signals, each of which is related to a certain frequency band.
This assumption has been corroborated in the detection and identification
of a certain type of vehicles (\cite{AHZK01,AKZ01}). The experiments
in this section demonstrate that a choice of distinctive characteristic
features that discriminate between vehicles and non-vehicle classes
can be derived from blocks of wavelet packet coefficients. Extraction
of characteristic features (parameters) is a critical task in the
training phase of the process.

In order to identify the acoustic signatures, we combine in the final
phase of the process the outputs of two classifiers. One is the well
known Classification and Regression Trees (CART) classifier \cite{CRT93}.
The other classifier is based on the distances between the test signal\ and
sets of pattern signals from the V and N classes.

The chapter has the following structure: In Section \ref{secrw},
we briefly review related works. The structure of the available data
is described in Section \ref{sec:structure}. In Section \ref{sec:s4},
we outline the scheme of the algorithm and in Section \ref{sec:s5}
we describe it in full details. Section \ref{sec:s6} is devoted to
presentation of the experimental results. Section \ref{sec:s7} provides
some discussion. Appendix I outlines the notions of the wavelet and
wavelet packets and Appendix II provides a detailed description of
the RSNOFP method.

%\section{Formulation of the approach}
%\label{sec:s2}

\section{Related work}

\label{secrw} Several papers that handle the separation between vehicle
and non-vehicle sounds can be found in the literature. Choe \emph{et
al.} \cite{WBGV}, extracted the acoustic features by using a discrete
wavelet transform. The feature vectors were compared to the reference
vectors in the database using statistical pattern matching to determine
the type of vehicle from which the signal originated. In \cite{AAS},
discrete cosine transform was applied to signals and a time-varying
auto-regressive modeling approach was used for their analysis. Averbuch
\emph{et al.} \cite{AHZK01}, designed a system that is based on wavelet
packets coefficients in order to discriminate between different types
of vehicles. Classification and Regression Trees (CARTs) were used
for classification of new unknown signals. In a later paper \cite{AKZ01},
Averbuch \emph{et al.} used similar methods with multiscale local
cosine transform applied to the frequency domain of the acoustic signal.
The classifier that was used was based on the \emph{Parallel Coordinates}
methodology \cite{ParallelCoords85}. Wu \emph{et al}. \cite{VSS}
used the \emph{eigenfaces method} \cite{CHF}, which was originally
used for human face recognition, to distinguish between different
vehicle sound signatures. The data was sliced into frames - short
series of time slices. Each frame was then transformed into the frequency
domain. Classification was done by projecting new frames on principal
components that were calculated for a known training set. Munich \cite{BSM}
compared between several speech recognition techniques for classification
of vehicle types. These methods were applied to short time Fourier
transform of the vehicles' acoustic signatures.

\section{The structure of the acoustics signals}

\label{sec:structure} \label{sec:ss12} The recordings that we use
in this section were taken under very different conditions on different
time points. The recordings sampling rate (SR) was 48000 samples per
second (SPS). It was downsampled to SR of 1000 SPS.

We extracted from the set of recordings, which were used for training
the algorithm, fragments that contain sounds emitted by vehicles and
stored them as the V-class signals. Recorded fragments that did not
contain vehicles sounds were stored as the N-class signals. Both classes
were highly diverse. Recordings in the V-class were taken from different
types of vehicles during different field experiments under various
surrounding conditions. In particular, the velocities of the vehicles
and their distances from the recording device were different from
one recording to the other. Moreover, the vehicles traveled on either
various paved (asphalt) or unpaved roads, or on a mixture of paved
and unpaved roads. Recordings in the N-class comprised of sounds emitted
by planes, helicopters, wind gusts and speech nearby the receiver,
to name a few.

\begin{figure}[!h]
\begin{centering}
\includegraphics[bb=31bp 245bp 576bp 611bp,
width=0.495\columnwidth,height=5cm]{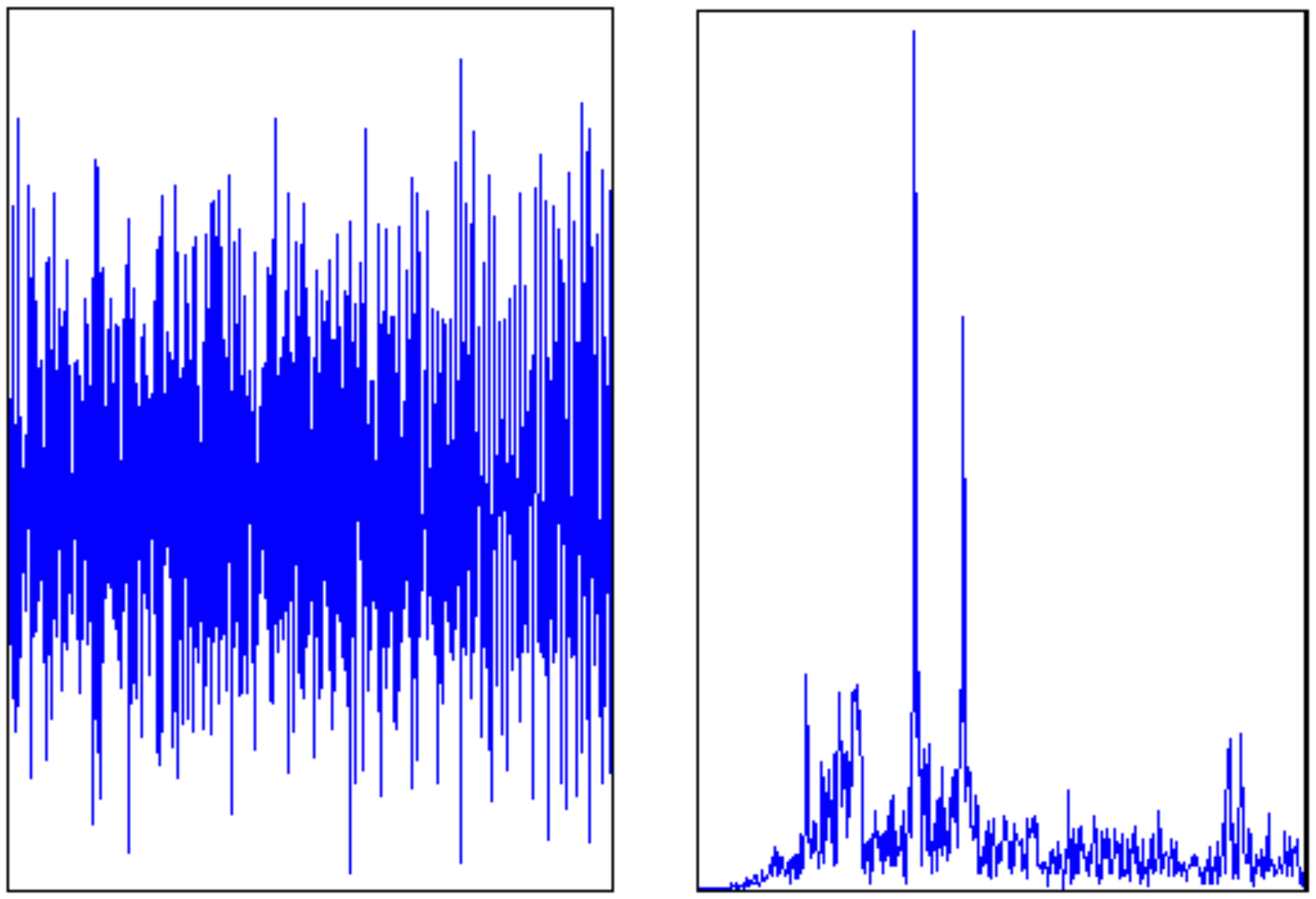}
\includegraphics[bb=41bp 245bp 576bp 611bp,
width=0.495\columnwidth,height=5cm]{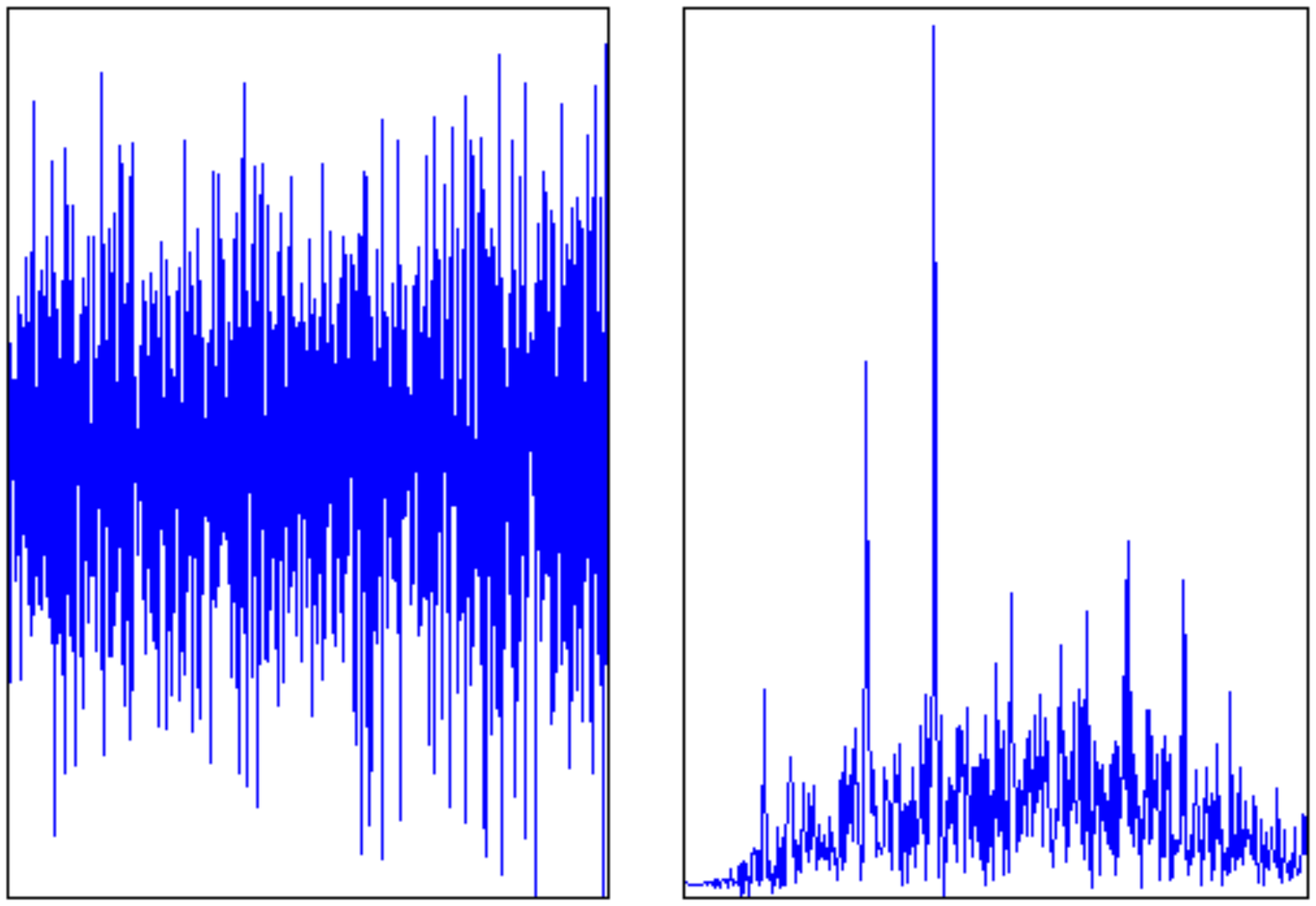} 
\par\end{centering}

\caption{Recording fragments of two cars and their spectra. Frames from left
to right: recording fragment of the first car and its spectrum, recording
fragment of the second car and its spectrum.}

\label{car} 
\end{figure}

Figure \ref{car} displays portions of acoustic signals emitted by
two cars with their Fourier transforms (spectra). 

\begin{figure}[!h]
\begin{centering}
\includegraphics[bb=20bp 244bp 583bp 611bp,
width=0.495\columnwidth,height=5cm]{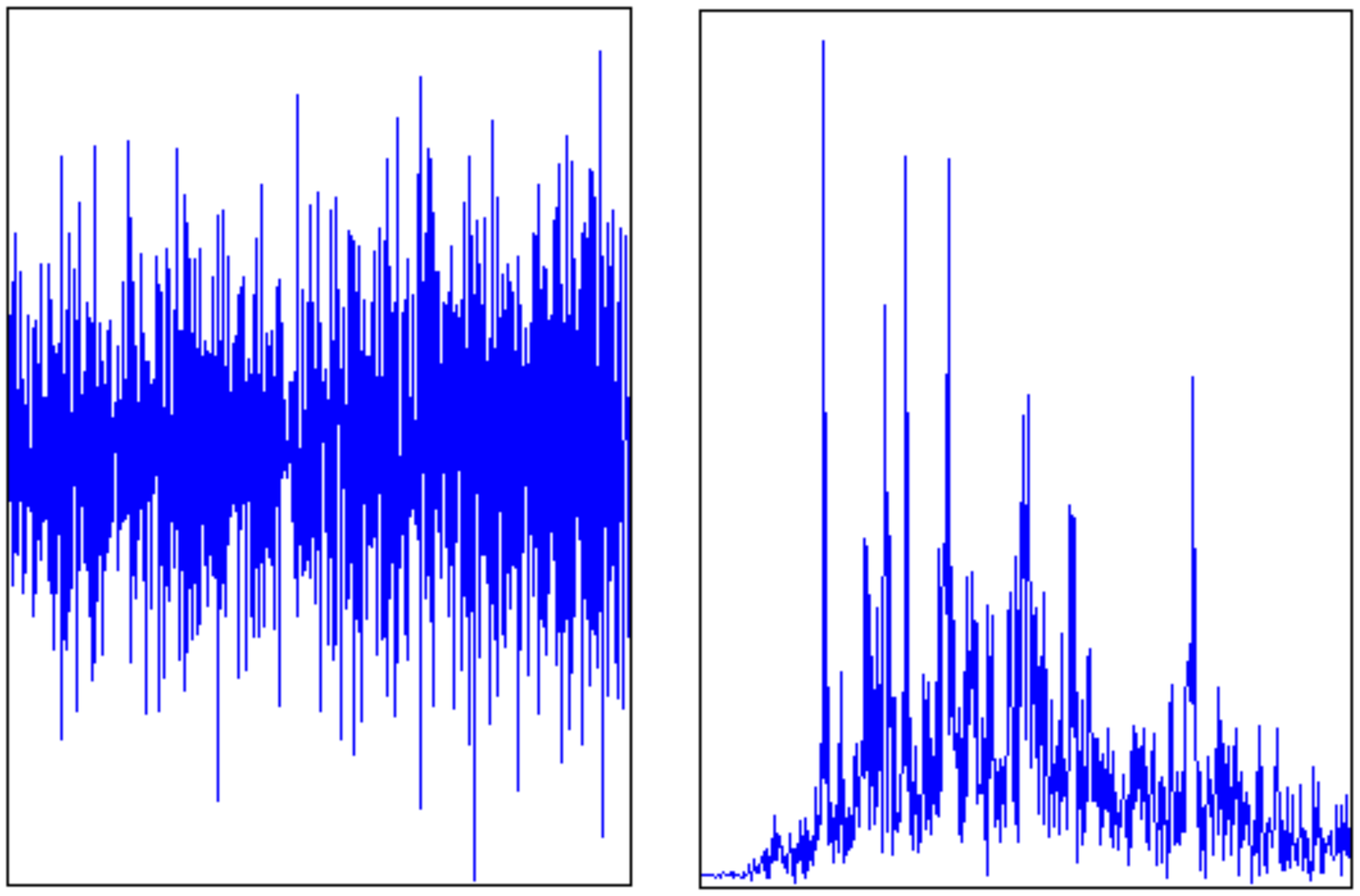}
\includegraphics[bb=30bp 244bp 583bp 611bp,
width=0.495\columnwidth,height=5cm]{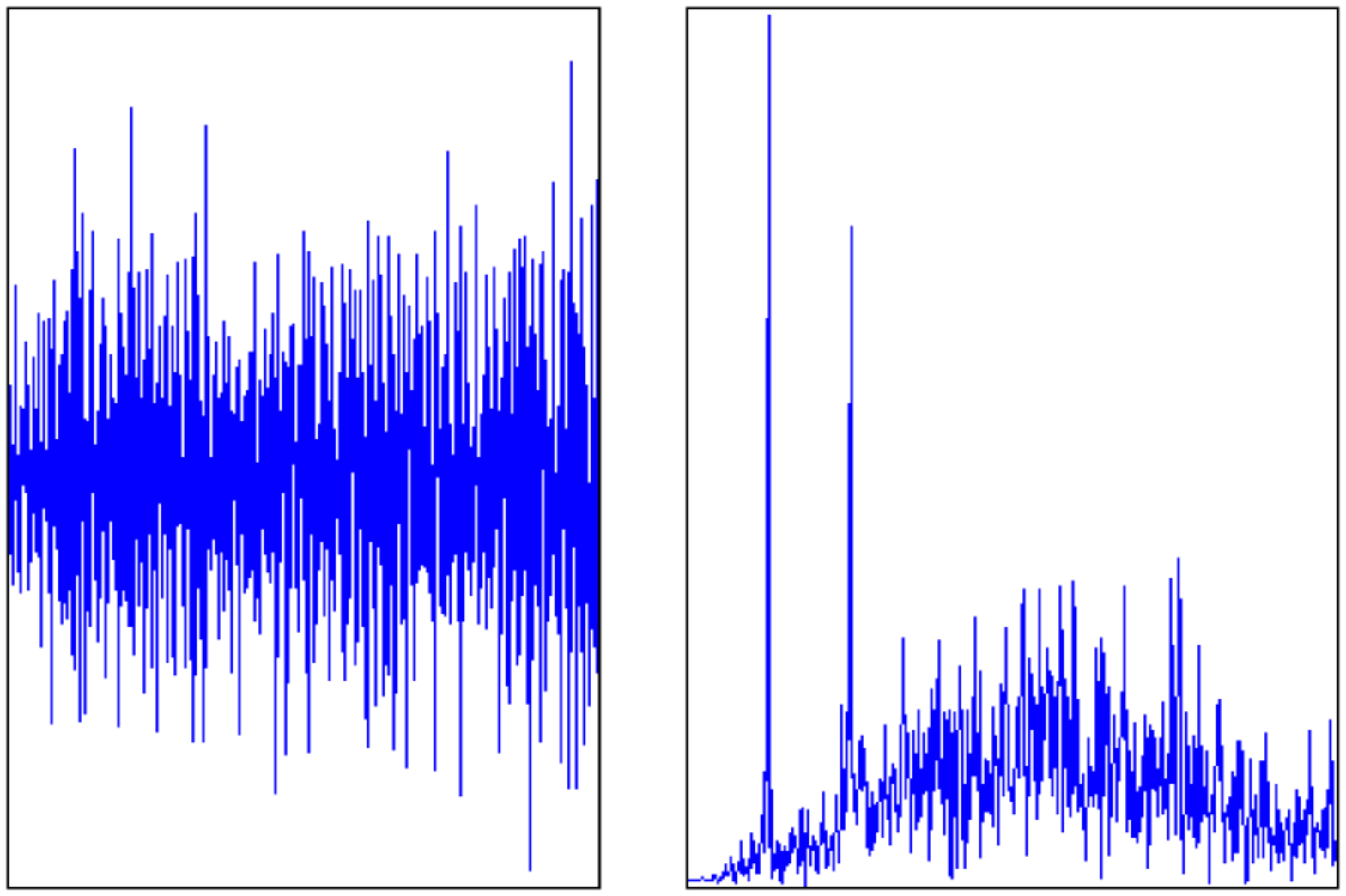} 
\par\end{centering}

\caption{Recording fragments of a truck and a van and their spectra. Frames
from left to right: recording fragment of a truck and its spectrum,
recording fragment of a van and its spectrum.}

\label{trvan} 
\end{figure}

Figure \ref{trvan} displays portions of acoustic signals emitted
by a truck and a van with their Fourier transforms.

\begin{figure}[!h]
\begin{centering}
\includegraphics[bb=14bp 244bp 583bp 611bp,
width=0.495\columnwidth,height=5cm]{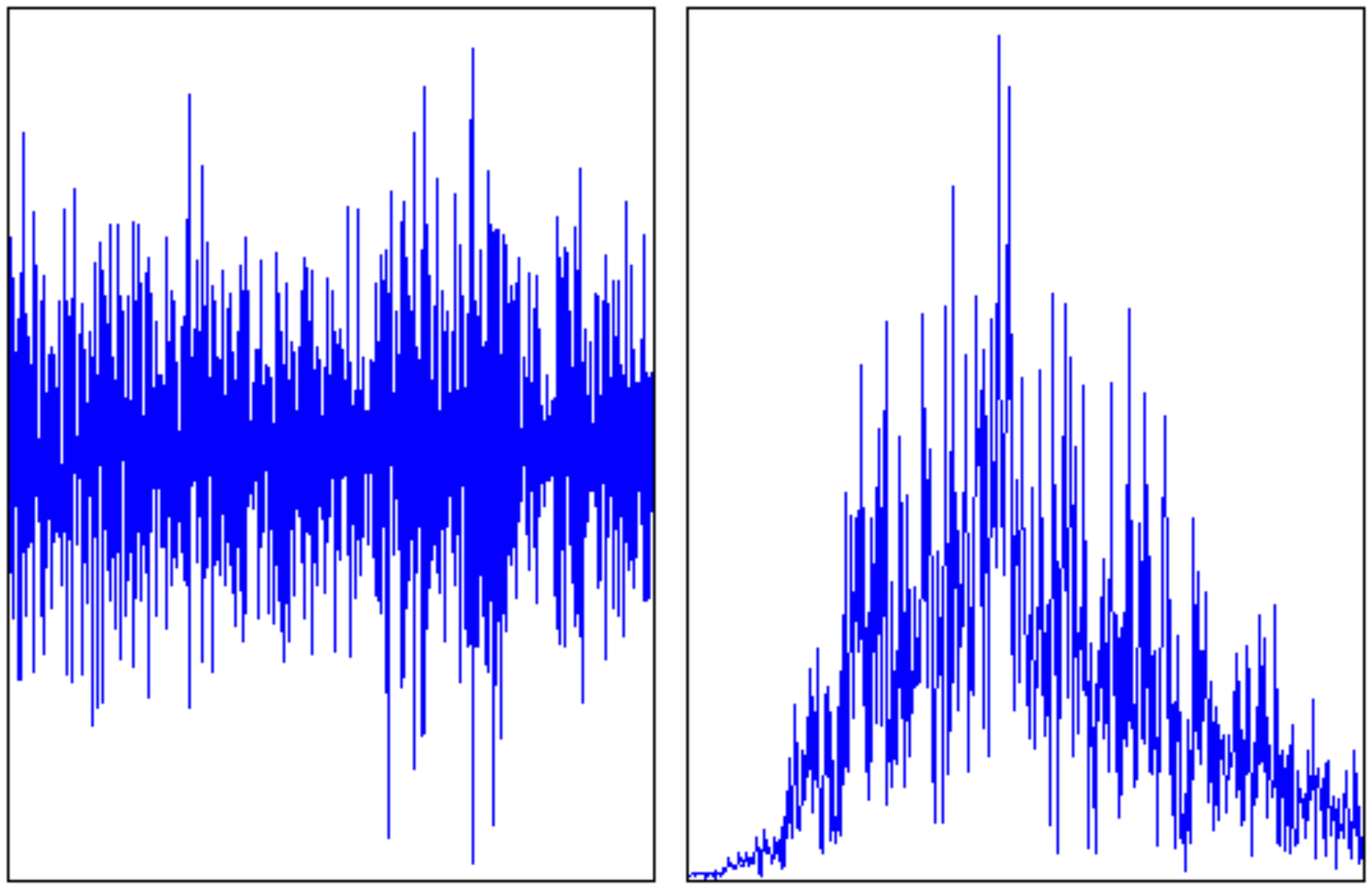}
\includegraphics[bb=14bp 244bp 583bp 611bp,
width=0.495\columnwidth,height=5cm]{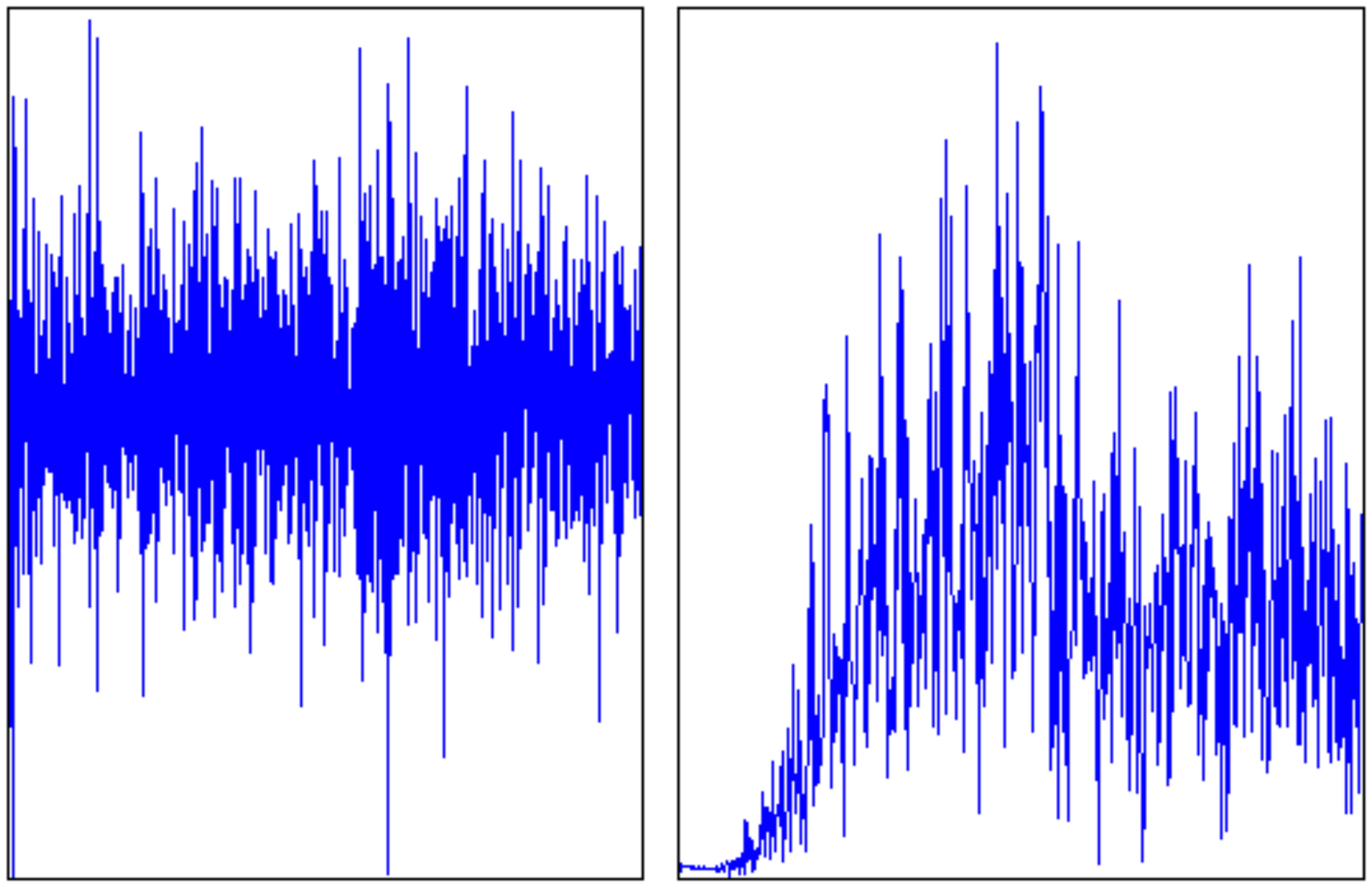} 
\par\end{centering}

\caption{Recording fragments of an airplane and a helicopter and their spectra.
Frames from left to right: recording fragment of an airplane and its
spectrum; recording fragment of a helicopter and its spectrum.}

\label{plahel} 
\end{figure}

\begin{figure}[!h]
\begin{centering}
\includegraphics[bb=6bp 244bp 583bp 611bp,
width=0.495\columnwidth,height=5cm]{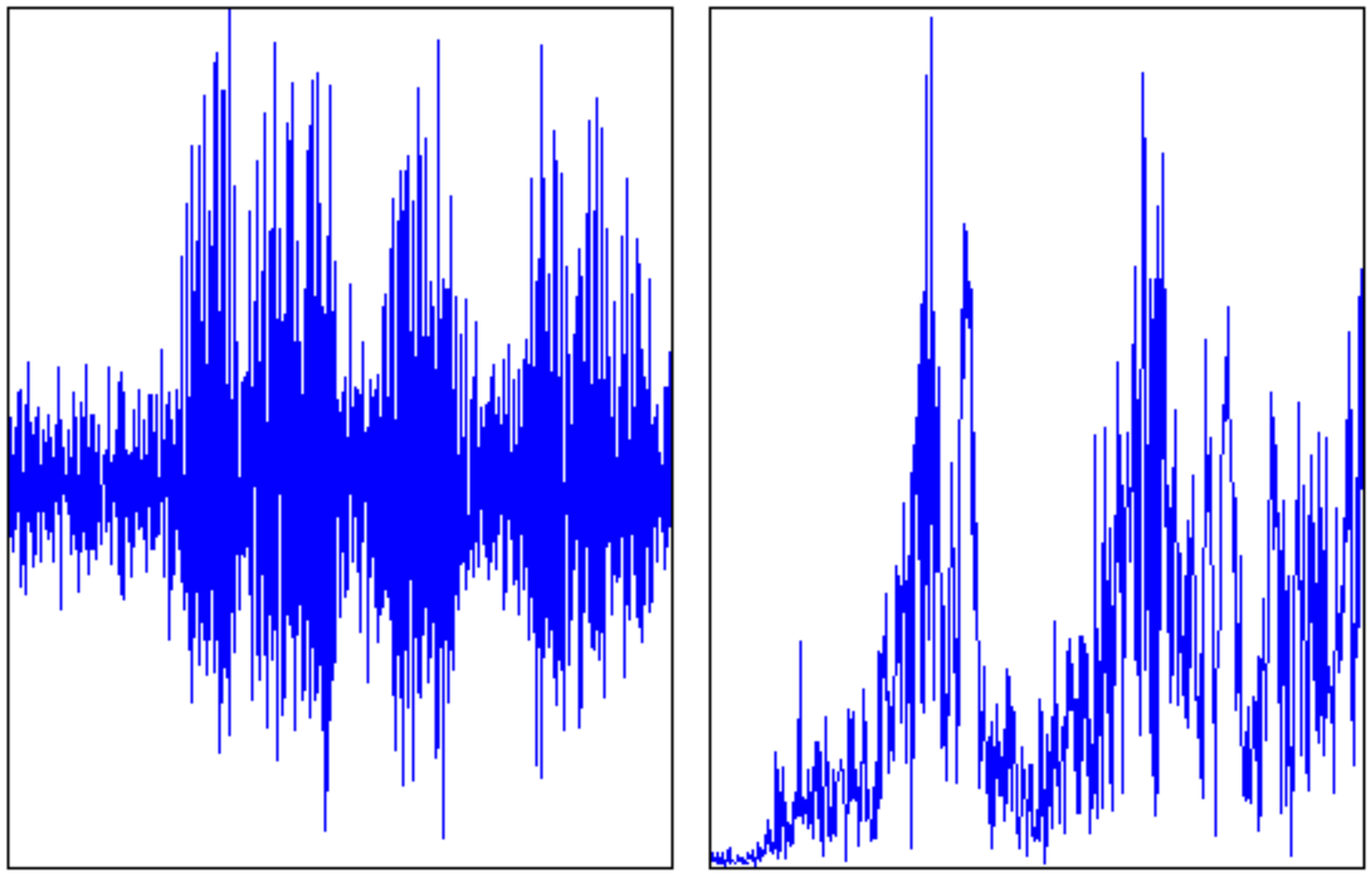}
\includegraphics[bb=17bp 244bp 588bp 611bp,
width=0.495\columnwidth,height=5cm]{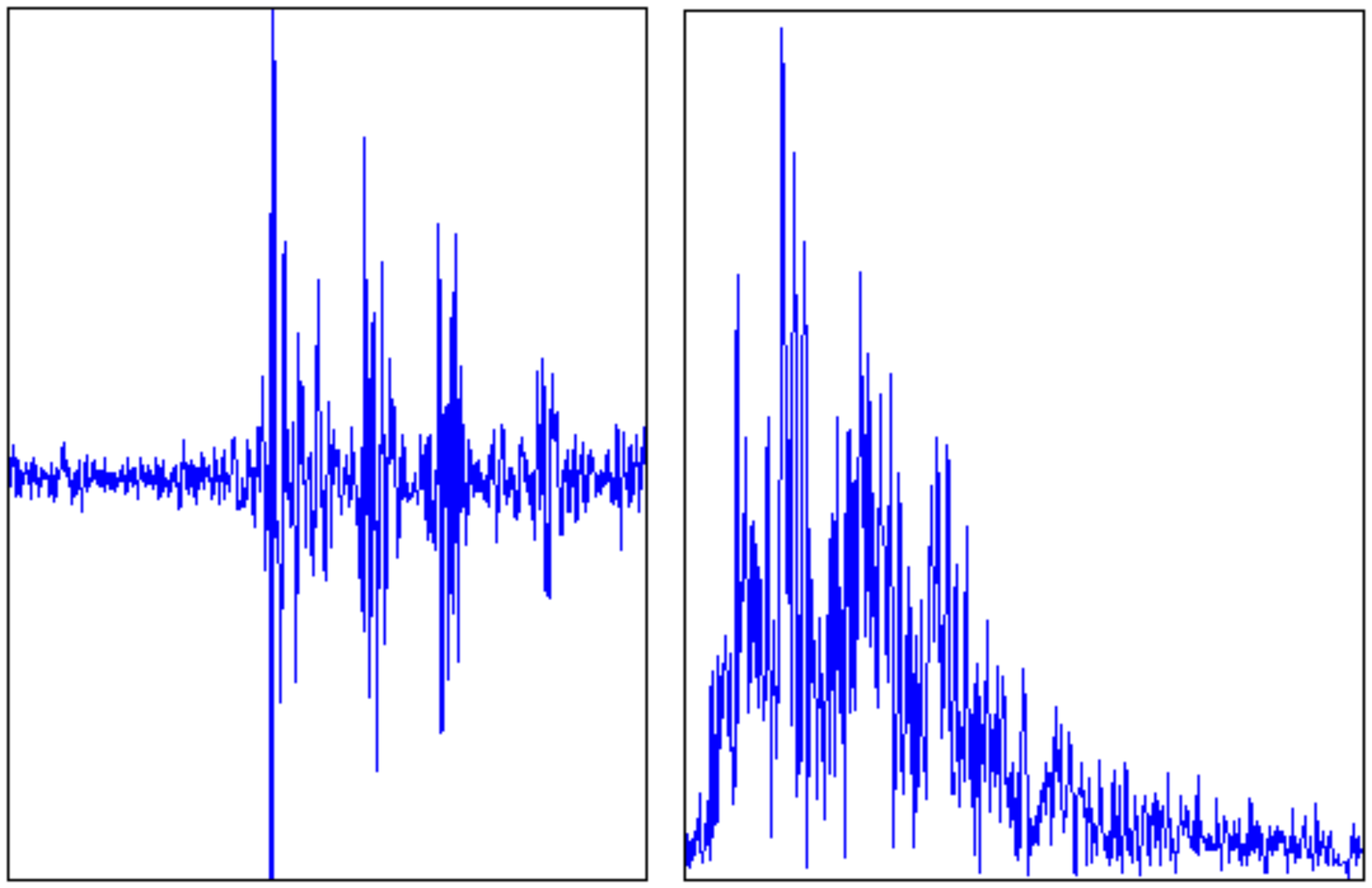} 
\par\end{centering}

\caption{Recording fragment fragments of speech and wind and their spectra.
Frames from left to right: recording fragment wind and its spectrum;
recording fragment of speech and its spectrum.}

\label{spwin} 
\end{figure}

We see that the spectra of different cars differ from one another.
It is even more apparent in the spectra of other vehicles. Figure
\ref{plahel} displays portions of acoustic signals emitted by a plane
and a helicopter with their Fourier transforms, whereas Fig. \ref{spwin}
does the same for speech and wind patterns.

Analysis of the signals revealed that even within the same class (V
or N), the signals differ significantly from one another. The same
is true for their corresponding Fourier transforms. However, there
are some properties that are common to the acoustic signals of all
moving vehicles. First, these signals are quasi-periodic in the sense
that there exist some dominating frequencies in each signal. These
frequencies may vary as motion conditions change. However, for the
same vehicle, these variations are limited to narrow frequency bands.
Moreover, the relative locations of the frequency bands are stable
(invariant) to some extent for signals that belong to the same vehicle.

Therefore, we conjectured that the distribution of the energy (or
some energy-like parameters) of acoustics signals that belong to some
class over different areas in the frequency domain, may provide a
reliable characteristic signature for this class.

\section{Formulation of the approach}

\label{sec:s4}

\begin{figure}[!h]
\begin{centering}
\includegraphics[bb=72bp 226bp 554bp 615bp,clip,
width=0.7\columnwidth,height=7cm]{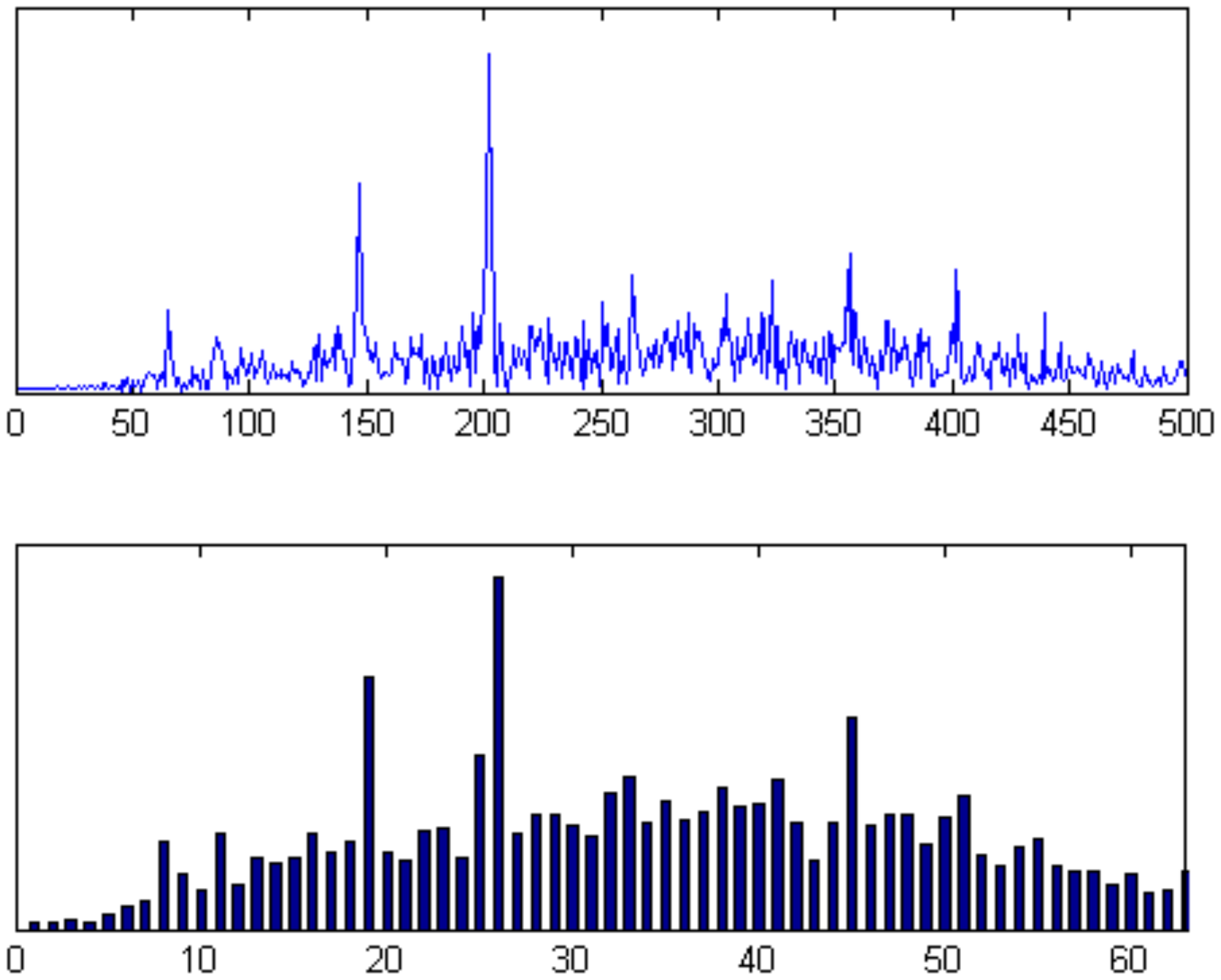} 
\par\end{centering}

\caption{Top: Fourier spectrum of the car signal in Fig. \ref{car}. Bottom:
Energies in the blocks of wavelet packets of the sixth level of the
wavelet packet transform that uses the orthogonal spline wavelets
of sixth order.}

\label{spebar} 
\end{figure}

Wavelet packet analysis (see Appendix I) is a highly relevant tool
for adaptive search for significant frequency bands in a signal or
a class of signals. Once implemented, a wavelet packet transform of
a signal, although computationally efficient, yields a huge (redundant)
variety of different partitions of the frequency domain. Due to the
lack of time invariance in the multiscale wavelet packet decomposition,
we use all the coefficients in the blocks of wavelet packet rather
than individual coefficients and waveforms. The collection of energies
in blocks of wavelet-packet-coefficients can be regarded as an averaged
version of the Fourier spectrum of the signal. However, wavelet packets
provide a more sparse and more robust representation of signals compared
to the Fourier spectrum. We can see it, for example, in Fig. \ref{spebar},
where the displayed energies in their blocks of wavelet packet coefficients
of the orthogonal spline wavelet of the sixth order in the sixth level
of the wavelet packet transform of a car acoustic signal.

The algorithms in \cite{AHZK01,AKZ01} use a variation of the Best
Basis algorithm \cite{CMW91,wic}. Specifically, it searches for a
few blocks that best discriminate a certain vehicle from other vehicles
and from the background. This approach did not prove to be robust
and efficient for solving the problem of this chapter. This is due
to the variability in vehicle types sought for in Class V and the
different types of background in Class N. Therefore, another way to
utilize the wavelet packet coefficients blocks was chosen. This method
can be characterized as a random search for the near-optimal footprint
(RSNOFP) of a class of signals. The proposed method utilizes ideas
from compressed sensing \cite{D06,DT06,CRT06}.

In order to enhance the robustness of the algorithm, we implement
three different versions of RSNOFP to corroborate one another.

The sample signals for the training phase and the on-line signals
in the detection phase are formed by applying a short-time window
on each input signal followed by a shift of this window along the
signal so that adjacent windows are overlapped.

\subsection{Outline of the approach}

The proposed scheme consists of three steps:
\begin{description}
\item [{Training~step:}] We use a set of signals with known V/N classification.
These signals are sliced into overlapped fragments of length $L$
(typically, $L=1024$). The fragments are chosen according to the
wavelet packet transform. The energies of the blocks are calculated
and three versions of RSNOFP are applied. As a result, we represent
each fragment by three different vectors of length $l\ll L$ (typically,
$l=12$ or $l=8$). The components of these vectors constitute the
characteristic features of the fragments. These vectors are used as
pattern datasets and are also utilized to produce three versions of
CART trees.
\item [{Identification~--~features~extraction~step:}] We slice the
new acquired signal\ to overlapped fragments of length $L$. Then,
the wavelet packet transform is applied to each fragment. This is
followed by calculation of the energies of the coefficient blocks.
Then, we apply three different transforms that are determined according
to the three versions of RSNOFP. As a result, we represent each fragment
by three different vectors of length $l$.
\item [{Identification~--~decision~making~phase:}] The three CART classifiers
that were constructed in the training step are applied to the vectors
that were produced by the previous step. In addition, the vectors
are tested by a second classifier that calculates the distances of
the vectors from the pattern datasets associated with the V and N
classes. The final classification to V and N of the fragment is derived
by combining the answers for the three vectors from the two classifiers.
\end{description}

\section{Description of the algorithm and its implementation}

\label{sec:s5}

\subsection{The algorithm}

\label{sec:ss31}

The algorithm is composed of three basic phases:
\begin{description}
\item [{I.}]  Extraction of characteristic features from the V and N classes.
It contains the following steps:

\begin{enumerate}
\item The analyzing wavelet filters are chosen. 
\item The training sets of the signals are constructed by slicing the training
signals into overlapped segments. 
\item The wavelet packet transform is applied to these segments. 
\item The energies in the blocks of the wavelet packet coefficients are
calculated.
\item The RSNOFP algorithm is executed. This embeds the training sets of
signals into lower-dimensional reference sets that contain their characteristic
features. 
\end{enumerate}
\item [{II.}] Building the CART classification trees. 
\item [{III.}]  Classification of the new signal to either the V or the
N class: 

\begin{enumerate}
\item The new signal is sliced into overlapped segments. 
\item The wavelet packet transform is applied to these segments. 
\item The energies in the blocks of the wavelet packet coefficients are
calculated. 
\item The set of blocks energies of each segment is embedded into a lower-dimensional
vectors that contains its characteristic features. 
\item The distances of the vector, which contains characteristic features,
from the reference sets of V and N classes are calculated. 
\item The vector is tested by the CART classifier. 
\item The vector is classified to either the V or the N class. 
\end{enumerate}
\end{description}
Next, we present a detailed description of the implementation of this
algorithm.

\subsection{Implementation }

\label{sec:ss32} \label{sec:sss321} %/subsubsection Extraction of characteristic features
\begin{figure}[!h]
\begin{centering}
\includegraphics[bb=76bp 244bp 544bp 642bp,clip,
width=0.495\columnwidth,height=5cm]{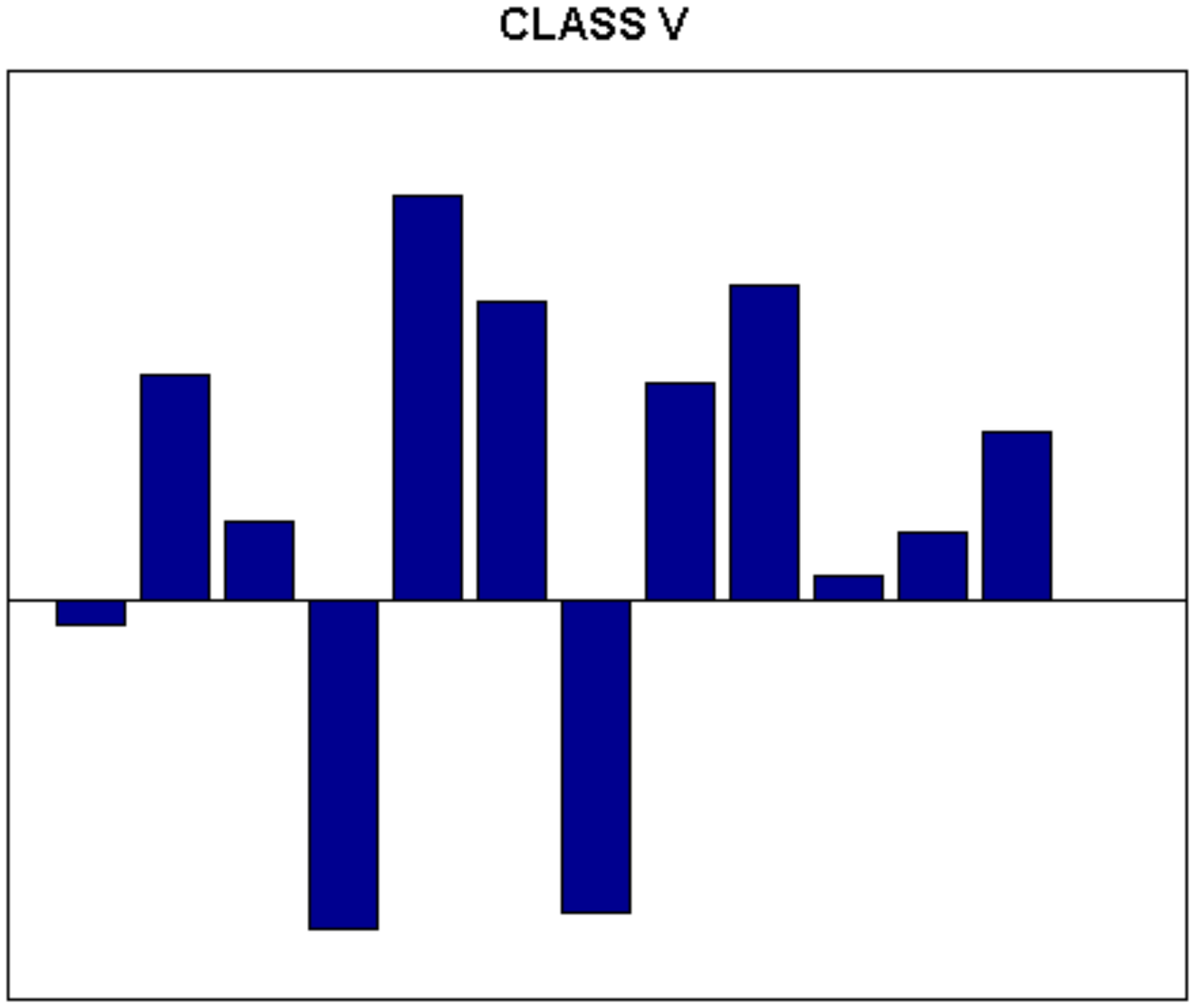}
\includegraphics[bb=76bp 244bp 544bp 642bp,clip,
width=0.495\columnwidth,height=5cm]{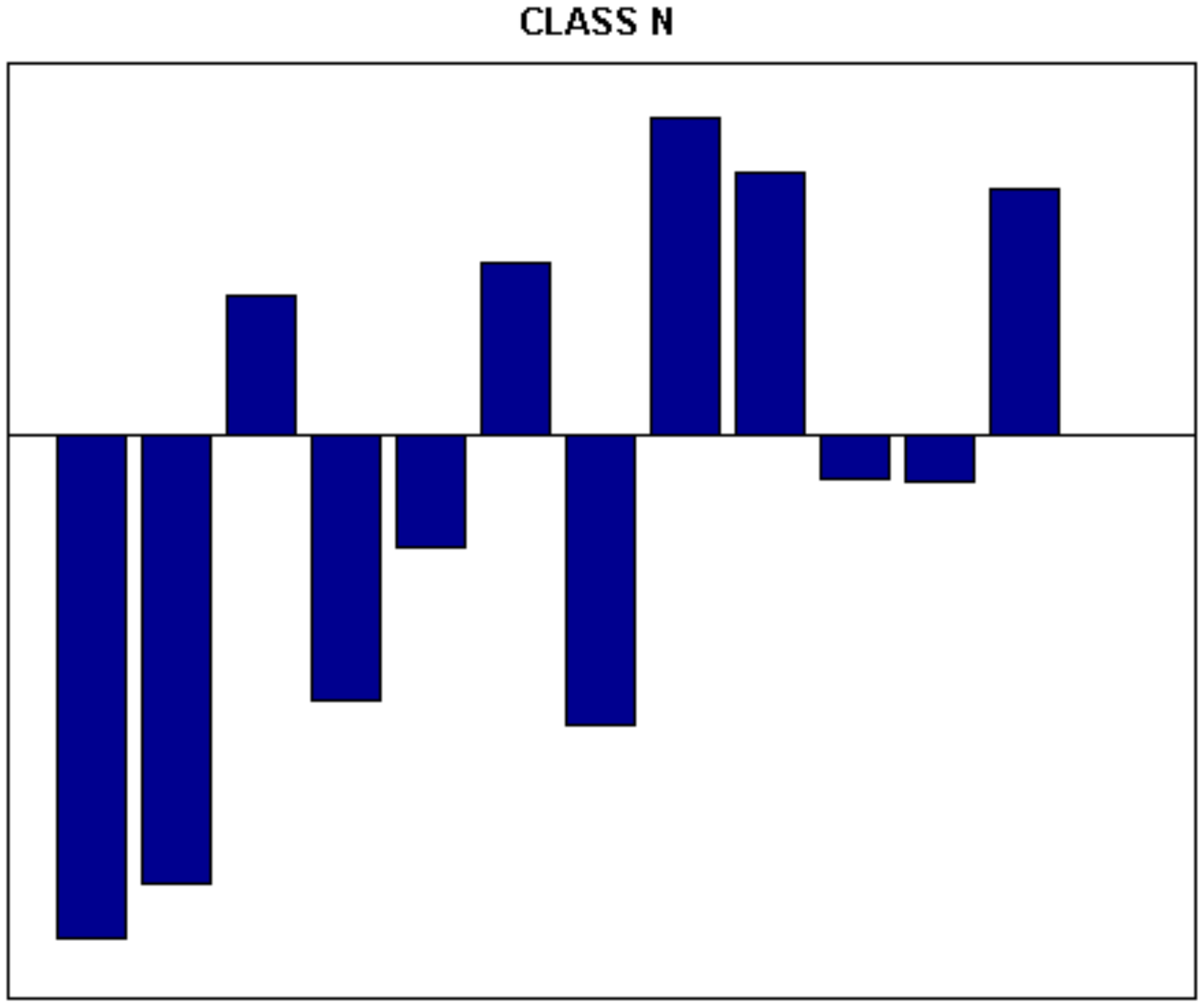} 
\par\end{centering}

\caption{Left: one row from the matrix $D_{perm}^{v}$ (features of a segment
from the V-class). Right: one row from the matrix $D_{perm}^{n}$
(features of a segment from the N-class).}

\label{smap} 
\end{figure}

\begin{description}
\item [{Choice~of~the~analyzing~waveforms:}]  A broad variety of orthogonal
and bi-orthogonal filters, which generate wavelet packets coefficients,
are available (\cite{Daub92,Mallat98,AZ1,AZ06}). We use the 6-th
order spline wavelet. This filter reduces the overlap between frequency
bands associated with different decomposition blocks. At the same
time, the transform with this filter provides a variety of waveforms
that have a fair time-domain localization. For details see Appendix
I (Section \ref{apendix1}).
\item [{Signal~preparation~for~training~the~algorithm:}]  Initially,
we gather as many recordings as possible for the V and the N classes,
which have to be separated. Then, we prepare from each selected recording,
which belongs to a certain class, a number of overlapped slices --
each of length $L=2^{J}$ samples, shifted by $S\ll L$ samples with
respect to one another. In total, we prepare $M^{v}$ and $M^{n}$
slices for the V and N classes, respectively. The slices are arranged
into two matrices, $A^{v}=\left(A_{i,j}^{v}\right)_{i=1,\ldots,M^{v};j=1,...,L}$
and $A^{n}=\left(A_{i,j}^{n}\right)_{i=1,\ldots,M^{n};j=1,...,L}$.
\item [{Embedding~the~sets~of~slices~into~sets~of~energies:}]  We
use the normalized $l_{1}$ norms of the blocks as the energy measure.
Then, the following operations are carried out:

\begin{enumerate}
\item The wavelet packet transform up to scale $m$ (typically $m=6$ if
$L=1024$) is applied to each slice of length $L$ from the V and
N classes. We take the coefficients from the sparsest (coarsest) scale
$m$. This scale contains $L=2^{J}$ coefficients that are arranged
into $2^{m}$ blocks of length $l=2^{J-m}$, each of which is associated
with a certain frequency band. These bands form a near-uniform partition
of size $2^{m}$ of the Nyquist frequency domain.
\item The ``energy'' of each block is calculated using the chosen measure.
We obtain, to some extent, the distribution of the ``energies''
of the $i$-th slice $A^{v}\left(i,:\right)$%
\footnote{We explain this for the $V$ class, however, this also applies to
the $N$ class. In the later case, the corresponding notation is $A^{v}\left(i,:\right)$.%
} over various frequency bands of widths $N_{F}/m$, where $N_{F}$
is the Nyquist frequency. It is stored in the energy vector $\vec{E}_{i}^{v}$
of length $\lambda=2^{m}=L/l$ (typically, $\lambda=64$). The energy
vectors are arranged into two matrices, $B^{v}=\left(B_{i,j}^{v}\right)_{i=1,\ldots,M^{v};j=1,...,\lambda}$
and $B^{n}=\left(B_{i,j}^{n}\right)_{i=1,\ldots,M^{n};j=1,...,\lambda}$.
The $i$-th row of the matrix $B^{v}$ is the vector $\vec{E}_{i}^{v}$.
This vector is considered to be the averaged Fourier spectrum of the
slice $A^{v}\left(i,:\right)$, as it is seen in Fig. \ref{spebar}.
We consider this vector to be a proxy of the slice. By the above operations
we reduced the dimension of the database by a factor of $l=2^{J-m}$.
\end{enumerate}
\item [{Embedding~of~sets~of~energies~into~the~sets~of~features:}]  The
subsequent operations yield a further reduction of the dimensionality
in the compressed sensing \cite{D06,DT06} spirit. It is achieved
by the application of three versions of the RSNOFP scheme to the energy
matrices $B^{v}$ and $B^{n}$. The RSNOFP scheme is described in
Appendix II. As a result, we get three pairs of the reference matrices:
$D_{rand}^{v}$$\;\&\;$$D_{rand}^{n}$, $D_{pca}^{v}$$\;\&\;$$D_{pca}^{n}$
and $D_{perm}^{v}\;\&\; D_{perm}^{n}$ and the corresponding random
matrices $\rho_{rand}$, $\rho_{pca}$ and $\rho_{perm}$. These random
matrices will be utilized in the identification phase. 
\item [{Compaction~of~the~feature~matrices~in~the~V-class:}]  In
order to refine the feature matrices of V-class, we test their rows.
Recall that each row is associated with a segment of a signal that
belongs to V-class. We calculate the Mahalanobis distances $d^{v}$
and $d^{n}$ of each row in the matrix $D_{rand}^{v}$ from the sets
$D_{rand}^{v}$ and $D_{rand}^{n}$, respectively. We remove row from
the matrix $D_{rand}^{v}$ rows for which $d^{v}>d^{n}$. The same
is done for the matrices $D_{pca}^{v}$ and $D_{perm}^{v}$.
\item [{Conclusion:}]  As a result of the above operations, the dimensionality
of the training set was substantially reduced. Typically, a segment
of length 1024 is embedded into a 12-component vector. This part of
the process might look computationally expensive, especially if, for
better robustness, large training sets are involved. However, this
procedure is called once and it is done off-line before the detection
phase that is done on-line. 
\end{description}
Figure \ref{smap} displays one row from matrix $D_{perm}^{v}$ and
one row from matrix $D_{perm}^{n}$. These are feature vectors that
correspond to segments from the V-class and the N-class. 

\subsection{Building the Classification and Regression Trees (CARTs)}

\label{classification} Once we have $D_{rand}^{v}\;\&\; D_{rand}^{n}$
, $D_{pca}^{v}\;\&\; D_{pca}^{n}$ and $D_{perm}^{v}\;\&\; D_{perm}^{n}$,
which are three pairs of the reference matrices, we proceed to build
the classifiers. For this purpose we use the vectors, which form rows
in the reference matrices. The construction of the tree is done by
a binary split of the space of input patterns $X\longrightarrow\{X_{1}\bigcup X_{2}\bigcup\ldots\bigcup X_{r}\},$
so that, if a vector appears in the subspace $X_{k}$, its classification
can be predicted with sufficient reliability. 

Given a test vector, the output of a CART classifier is the class
to which the vector belongs and the probability of this classification.
The basic idea behind the split is that the data in each descendant
subset is \emph{more pure} than the data in the parent subset. The
scheme is described in full details in the monograph \cite{CRT93}.
A brief outline of this method that is tailored to acoustic processing
is given in \cite{AHZK01}.

After the construction of the three classification trees $T_{rand}$
, $T_{pca}$ and $T_{perm}$ and the three pairs of reference matrices,
we are in a position to classify new signals that were not used in
the training phase.

\subsection{Identification of an acoustic signal}

\label{identification}

An acoustic signal to be identified is preprocessed by the same operations
that were used on the training signals. 
\begin{description}
\item [{Preprocessing~of~a~new~acoustics~signal:}]  Preprocessing
is done similarly to the feature extraction phase (Section \ref{sec:ss32}). 

\begin{enumerate}
\item The signal is sliced to $M$ overlapped segments of length $L=2^{J}$
samples each, shifted with respect to each other by $S$ samples.
The wavelet packet transform up to scale $m$ is applied to each slice.
We take the coefficients from the sparsest (coarsest) scale $m$ that
are arranged into $2^{m}$ blocks of length $l=2^{J-m}$. The ``energy''
of each block is calculated with the chosen measure. Thus, each $i$-th
slice is embedded into an energy vector $\vec{E}_{i}$ of length $\lambda=2^{m}=L/l$.
The energy vectors are arranged in the matrix $B=\left(B_{i,j}\right)_{i=1,...,M;\; j=1,...,\lambda}$
where the $i$-th row of the matrix $B$ is the vector$\vec{E}_{i}$. 
\item In order to embed the energy matrix $B$ into the feature spaces,
we multiply it by the random matrices $\rho_{rand}$, $\rho_{pca}$
and $\rho_{perm}$. These multiplications produce three feature matrices
$D_{rand}$, $D_{pca}$ and $D_{perm}$, where the $i$-th row in
each matrix is associated with the $i$-th segment of the processed
signal. 
\end{enumerate}
\item [{Identification~of~a~single~segment:}] To identify the $i$-th
segment of a signal , we take three vectors $\vec{v}_{rand}^{i}$,
$\vec{v}_{pca}^{i}$ and $\vec{v}_{perm}^{i}$, which form the $i$-th
rows of the matrices $D_{rand}$, $D_{pca}$ and $D_{perm}$, respectively. \end{description}
\begin{enumerate}
\item These vectors are submitted to their respective versions $T_{rand}$,
$T_{pca}$ and $T_{perm}$ of the classification tree. Once a vector
is submitted to the tree, it is assigned to one of the subsets $X_{k}$
of the input space $X$. These trees produce three decisions $\tau_{rand}$,
$\tau_{pca}$ and $\tau_{perm}$ together with the corresponding probabilities
$p_{rand}$, $p_{pca}$ and $p_{perm}$. The decision $\tau_{({\bf \cdot})}$
determines the most probable membership of the segment. Here $(\cdot)$
stands for \emph{rand}, or \emph{pca} or \emph{perm}. The value $\tau_{({\bf \cdot})}=1$
if the CART assigns the segment to V-class and $\tau_{({\bf \cdot})}=0$
otherwise. 
\item The distances (for example, Mahalanobis or Euclidean) between the
vectors $\vec{v}_{rand}^{i}$, $\vec{v}_{pca}^{i}$ and $\vec{v}_{perm}^{i}$
and the respective pairs of the reference sets $D_{rand}^{v}$ $\&$
$D_{rand}^{n}$, $D_{pca}^{v}$ $\&$ $D_{pca}^{n}$ and $D_{perm}^{v}$
$\&$ $D_{perm}^{n}$ are calculated. This calculation produces three
decisions $\tilde{\tau}_{rand}$, $\tilde{\tau}_{pca}$ and $\tilde{\tau}_{perm}$
together with the corresponding probabilities $\tilde{p}_{rand}$,
$\tilde{p}_{pca}$ and $\tilde{p}_{perm}$ in the following way. Let
$d^{v}$ and $d^{n}$ be the distances of a vector $\vec{v}_{({\bf \cdot})}^{i}$
from the respective pair of the reference sets $D_{({\bf \cdot})}^{v}$
and $D_{({\bf \cdot})}^{n}$. If $d^{v}<d^{n}$ then the decision
is $\tilde{\tau}_{({\bf \cdot})}=1$ (the segment is assigned to V-class),
otherwise $\tilde{\tau}_{({\bf \cdot})}=0$ (the segment is assigned
to N-class). If $d^{v}<d^{n}$ then the membership probability in
the V-class is defined as $\tilde{p}_{({\bf \cdot})}=1-d^{v}/d^{n}$,
otherwise $\tilde{p}_{({\bf \cdot})}=0$. This classification scheme
is similar to the well known Linear Discriminant Analysis (LDA) classifier
\cite{Fisher36}. If the Mahalanobis distance is used then it is identical
to LDA. We call this scheme the Minimal Distance (MinDist) classifier.
\item Two threshold values $t$ and $\tilde{t}$ are set and the results
for the $i$-th segment are combined into three 3-component column
vectors $\vec{y}_{rand}^{i}$, $\vec{y}_{pca}^{i}$ and $\vec{y}_{perm}^{i}$,
where: 
\begin{equation}
\begin{array}{lll}
y{}_{({\bf \cdot})}^{i}(1) & = & \left\{ \begin{array}{cc}
1, & if\, p_{({\bf \cdot})}>t\\
0, & otherwise
\end{array}\right.\\
y{}_{({\bf \cdot})}^{i}(2) & = & \left\{ \begin{array}{cc}
1, & if\,\widetilde{p}_{({\bf \cdot})}>\widetilde{t}\\
0, & otherwise
\end{array}\right.\\
y{}_{({\bf \cdot})}^{i}(3) & = & \tau_{({\bf \cdot})}\cdot\tilde{\tau}_{({\bf \cdot})}
\end{array}
\end{equation}
\end{enumerate}
\begin{description}
\item [{Identification~of~a~recording:}]  \end{description}
\begin{enumerate}
\item The vectors $\vec{y}_{rand}^{i}$, $\vec{y}_{pca}^{i}$ and $\vec{y}_{perm}^{i}$
are gathered into three matrices $Y_{rand}$, $Y_{pca}$ and $Y_{perm}$
of size $3\times M$, where $M$ is the number of overlapping segments
produced from the analyzed signal. The vectors $\vec{y}_{{\bf \left(\cdot\right)}}^{i}$
serve as the $i$-th columns in the respective matrices $Y_{{\bf \left(\cdot\right)}}$.
\item The rows of the matrices are processed by a moving average. 
\item The matrices $Y_{rand}$, $Y_{pca}$ and $Y_{perm}$ are combined
into the matrix $Y$ in the following way. Each entry in $Y$ is defined
as the median value of the respective entries of the three matrices:
\begin{equation}
Y(i,j)={\rm median}\left(Y_{rand}\left(i,j\right),\; Y_{pca}\left(i,j\right),\; Y_{perm}\left(i,j\right)\right).\label{e2}
\end{equation}

\end{enumerate}

\paragraph*{\noindent Conclusions: }

\noindent The matrix $Y$ contains the results for the analyzed signal.
Its first row contains the averaged answers (which have significant
probabilities) from the CART classifier. The value at each point is
the number of positive (class V) answers in the vicinity of this point,
which is divided by the length of the vicinity. It represents the
``density\textquotedbl{} of the positive answers around the corresponding
segment. The structure of the second row is similar to the structure
of the first row with the difference that these answers come from
the MinDist classifier instead of the CART classifier. The third row
of the matrix $Y$ combines the answers from both classifiers. First,
these answers are multiplied. The combined answer is equal to one
for segments where both classifiers produce the answer one (V-class)
and zero otherwise. Thus, the classifiers cross-validate one another.
Then, the results are processed by the application of the moving average
providing the ``density\textquotedbl{} for the positive answers.
The third row in the matrix $Y$ yields the most robust result from
the detection process with minimal false alarm. For a binary detection
scheme, any detection probability above 0.5 is considered as a detection.

The above scheme is described for the detection of the arrival of
any moving vehicle. Obviously, the scheme is also applicable for the
detection of the arrival of the sought after vehicles.

\section{Experimental results}

\label{sec:s6}

We conducted a series of experiments to detect the arrival of vehicles
of arbitrary type in the presence of various types of noise.

Two hundred recordings were available. They were taken in five different
areas. Many recordings contained sounds emitted by different vehicles
combined with the sounds of wind, speech, aircrafts etc. The original
sampling rate (SR) was 48000 samples per second (SPS). The signals
were downsampled to SR of 1000 SPS. The motion dynamics, the distances
of vehicles from the receiver and the surrounding conditions were
highly diverse.

\begin{comment}
The calculation of the three pairs of reference matrices $D_{rand}^{v}\;\&\; D_{rand}^{n}$,
$D_{pca}^{v}\;\&\; D_{pca}^{n}$ and $D_{perm}^{v}\;\&\; D_{perm}^{n}$
(Section \ref{sec:sss321}) requires 2-3 minutes of CPU time on a
standard PC using a 3GHz Pentium4 processor and 2GB RAM.
\end{comment}

\subsection{Detection experiments}

\label{sec:ss41}

The first task was to form the reference database of signals with
known classification (training) for the construction of the classifiers.
This database was derived from the recordings by clipping the corresponding
fragments. The CAR fragments were extracted from 10 recordings whereas
5 recordings were used for the TRUCK fragments and the for the VAN
fragments. Diverse non-vehicle fragments were extracted from 35 recordings.
Thirty eight recordings, in total, were involved in the training process
(most of them contained sounds from different sources). We tested
various families of wavelet packets, various norms for the feature
extraction and various combinations of features for the MinDist and
CART classifiers. The best results were achieved with the wavelet
packet transform that uses the sixth order spline filters and the
$l_{1}$ norm.

We separated the reference signal s into two groups. One group (V
class) contains all signal s associated with vehicles and the other
group (N class) contains all the non-vehicles signals. The signals
were sliced into overlapped segments of length $L=1024$ that were
shifted with respect to one another by $S=256$, thus, the overlap
was 3/4 of a fragment. We extracted the characteristic features from
the segments as explained in Section~\ref{sec:sss321}. Each segment
was expanded by the wavelet packet transform up to 6th level (scale)
and the $l_{1}$ norm was used as an ``energy\textquotedbl{} measure
for all the 64 blocks of the 6th level. As a result of the procedures
that were described in Section \ref{sec:sss321}, we selected various
sets of discriminant blocks. These procedures produced three pairs
of reference matrices: $D_{rand}^{v}\;\&\; D_{rand}^{n}$, $D_{pca}^{v}\;\&\; D_{pca}^{n}$
and $D_{perm}^{v}\;\&\; D_{perm}^{n}$ with the corresponding random
matrices $\rho_{rand}$, $\rho_{pca}$ and $\rho_{perm}$. Each matrix
has 12 columns according to the number of characteristic features.
These matrices were used for the MinDist classification and were also
utilized for building three CART trees $T_{rand}$, $T_{pca}$ and
$T_{perm}$. For the MinDist classification, we used all the available
features (12), unlike building the CART trees, where better results
were achieved with sets containing only 8 features.

All the available recordings were used in the detection phase. A recording
number $k$ was embedded into the three feature matrices $D_{rand}^{k}$,
$D_{pca}^{k}$ and $D_{perm}^{k}$ to associate the $i$-th row of
each matrix with the $i$-th segment of the recording (see Section
\ref{identification}). Each row was tested with the MinDist and CART
classifiers. The results were gathered into the $Y^{k}$ matrix. The
Euclidean distance was used for the MinDist classification.

In Figs. \ref{sig7}--\ref{sig438}, we present a few results from
the experiments on detection of vehicles of arbitrary types. All the
figures have the same structure. Each is composed of four parts. The
top part depicts the original recording $\#k$. The three parts below
the original recording present five rows from the $Y^{k}$ matrix
with respect to the time scale. The second part from the top presents
the combined answers (the median from three answers) from the CART
classifiers after being processed by the moving average. The third
part from the top displays, in a similar manner, the results from
the MinDist classifiers. The bottom figure illustrates the combined
results from both classifiers i.e. the product of the answers from
both classifiers after it was processed by the moving average. Any
detection probability above 0.5 is considered as a detection.

\subsubsection{Examples}
\begin{description}
\item [{Recording~$\#1$:}] We display in Fig. \ref{sig7} the results
for test recording $\#1$. This recording was part of the training
set. It is apparent that the arrivals of a car and a track at around
the $40^{th}$ and $55^{th}$ seconds of the recording, respectively,
are correctly detected by the CART and the MinDist classifiers. 
\begin{figure}[!h]
\begin{centering}
\includegraphics{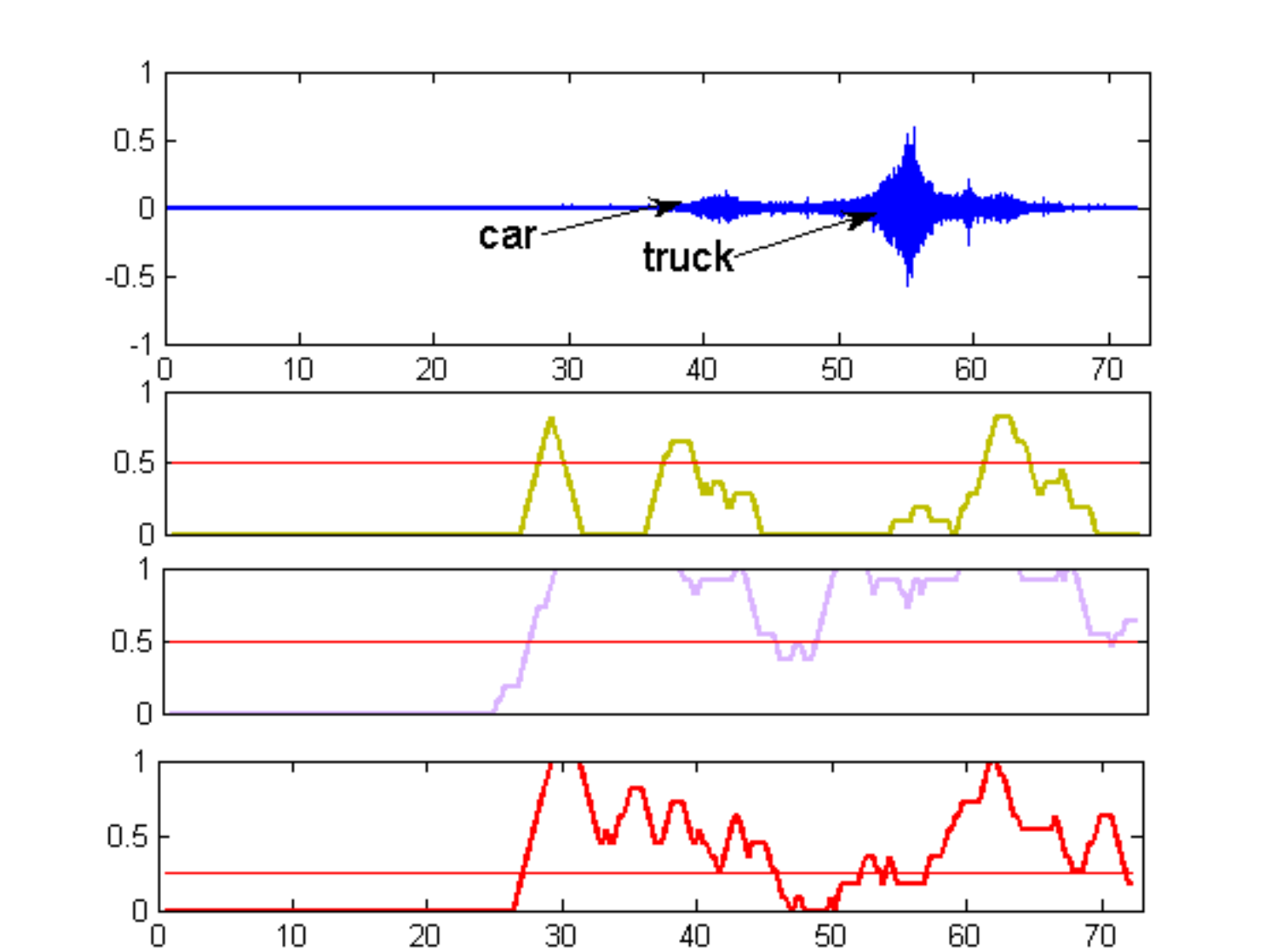} 
\par\end{centering}

\caption{Results for test recording $\#1$. The recording contains sounds emitted
by a car (at around the $40^{th}$ second) and a truck (at around
$55^{th}$ second). This recording was part of the training set.}

\label{sig7} 
\end{figure}

\item [{Recording~$\#2$:}] We display in Fig. \ref{sig16} the results
for test recording $\#2$. This recording was part of the training
set. Arrivals of two cars: one at the $11^{th}$ second and another
at the $27^{th}$ second of the recording, respectively, are correctly
detected by the CART and the MinDist classifiers. The sound of a helicopter
became audible starting from the $27^{th}$ second of the recording
until the end of the recording. It caused some false alarms, which
were eliminated by combining the results of the classifiers (bottom
figure). 
\begin{figure}[!h]
\begin{centering}
\includegraphics{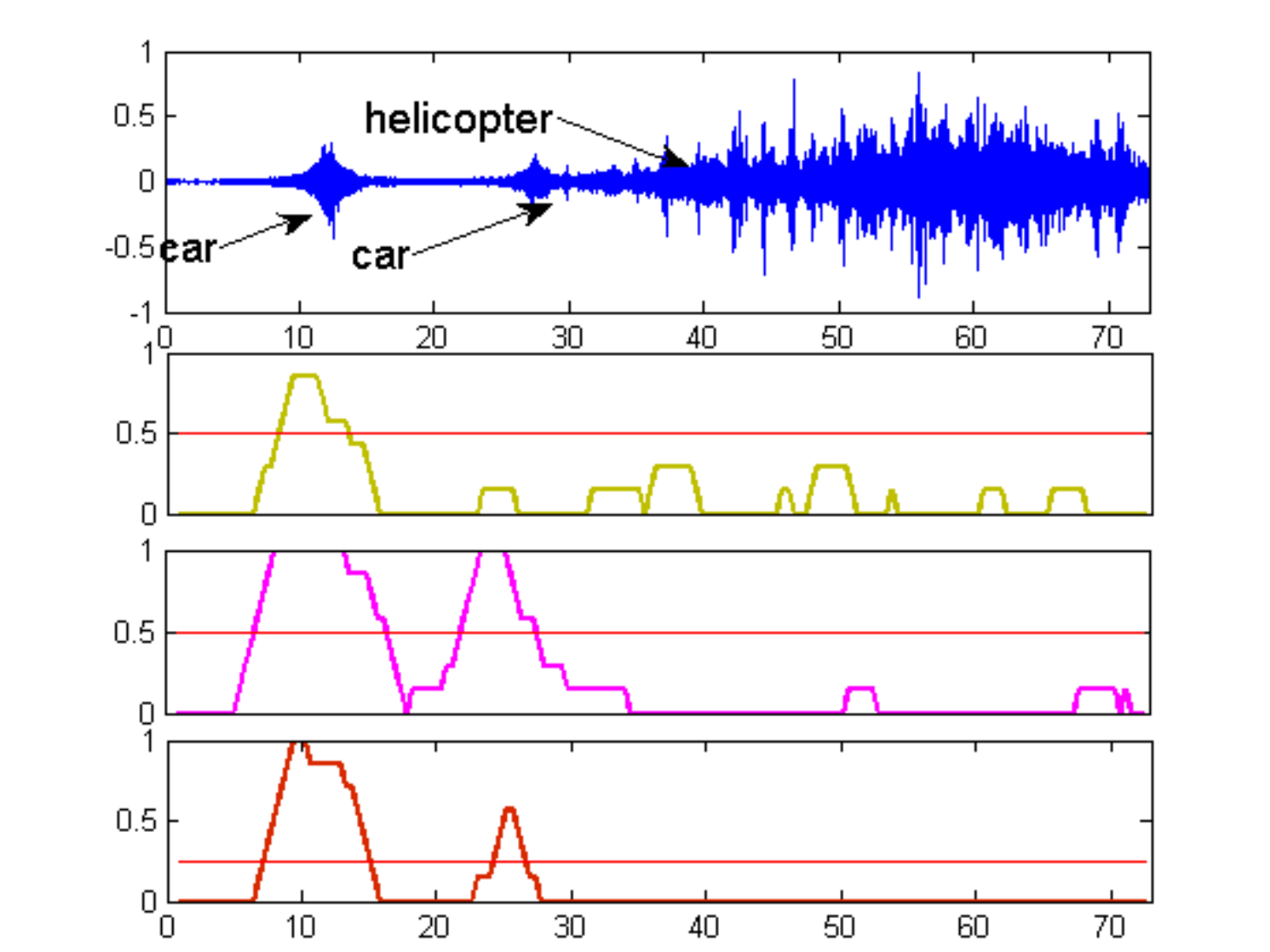} 
\par\end{centering}

\caption{Results for test recording $\#2$. The recording contains sounds emitted
by a car (at the $11^{th}$ second) and by another car (at the $27^{th}$
second). This recording was part of the training set.}

\label{sig16} 
\end{figure}

\item [{Recording~$\#3$:}] We display in Fig. \ref{sig21} the results
for test recording $\#3$. A fragment of 60 seconds from the beginning
of the recording was part of the training set of the N class. A loud
speech is present in the background until vehicles pass by. It lasted
during the first 100 seconds of the recording. In addition, there
was a plane sound from the $107^{th}$ second until the end of the
recording. A van briefly passed by at the $105^{th}$ second of the
recording. It was correctly detected by the CART and the MinDist classifiers.
The number of false alarms was reduced by combining the results of
the classifiers (bottom figure). 
\begin{figure}[!h]
\begin{centering}
\includegraphics{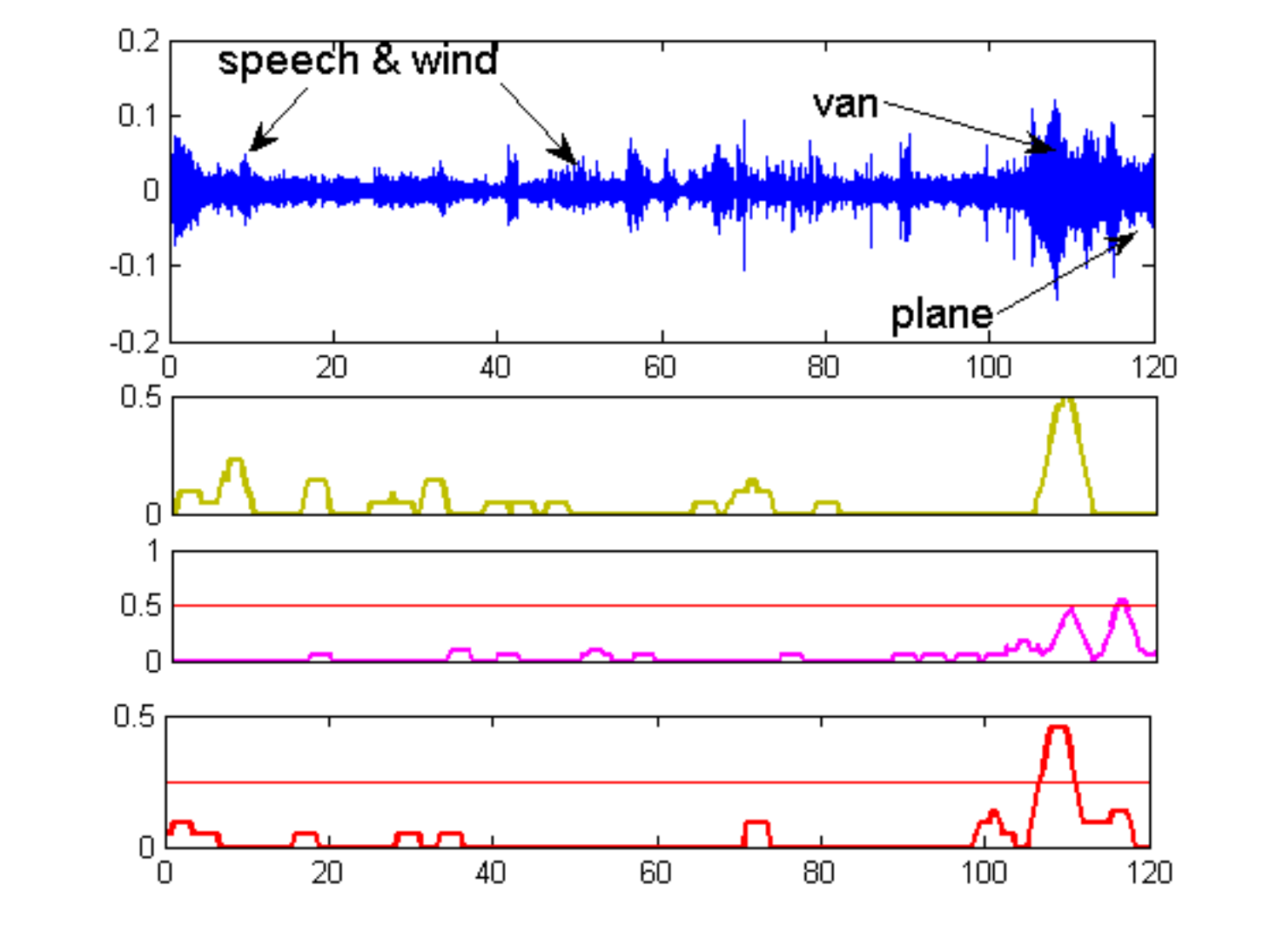} 
\par\end{centering}

\caption{Results for test recording $\#3$. The recording contains loud speech
during the first 100 seconds, sounds emitted by a car (at around the
$105^{th}$ second) and sound of a plane (from the $107^{th}$ second
until the end of the recording). The fragment of the first 60 seconds
of the recording was part of the training set of the N class.}

\label{sig21} 
\end{figure}

\item [{Recording~$\#4$:}] We display in Fig. \ref{sig110} the results
for test recording $\#4$. This recording was not part of the training
set. In the beginning of the recording, sound from a remote vehicle
is heard. Then, a jumping vehicle passed by the receiver at around
the $70^{th}$, $134^{th}$ and $200^{th}$ seconds of the recording.
In its last occurrence, it was followed by another car. All the events
were correctly detected by the CART and the MinDist classifiers. The
MinDist classifier produced some false alarms, which were eliminated
by combining the results of the classifiers (bottom figure). 
\begin{figure}[!h]
\begin{centering}
\includegraphics{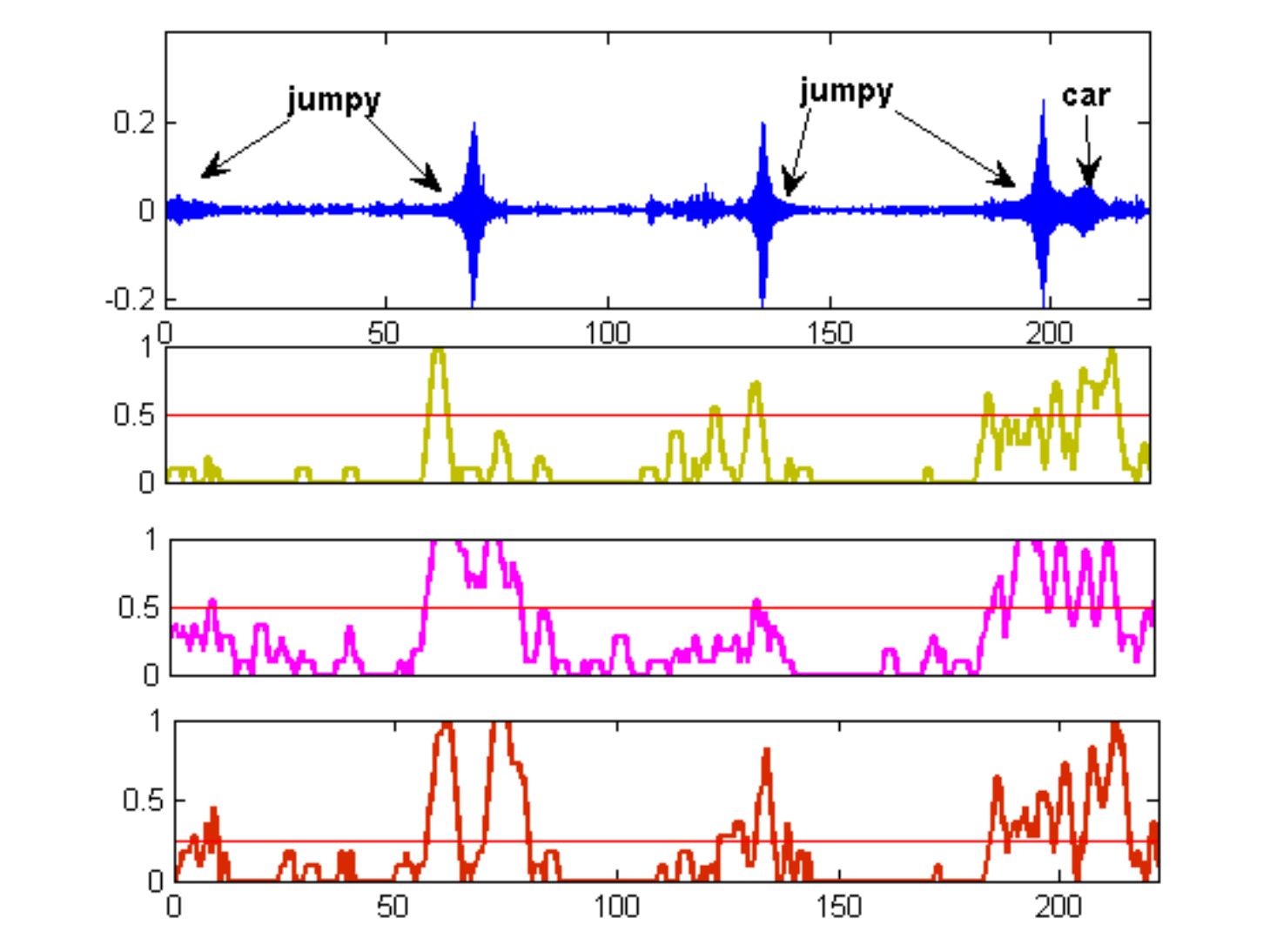}
\par\end{centering}

\caption{Results for test recording $\#4$. In the beginning, the sound from
a remote vehicle is heard. A jumping vehicle passed by the receiver
at around the $70^{th}$, $134^{th}$ and $200^{th}$ seconds of the
recording. In its last occurrence, it was followed by another car.
The recording was not part of the training set.}

\label{sig110} 
\end{figure}

\item [{Recording~$\#5$:}] We display in Fig. \ref{sig129} the results
for test recording $\#5$. This recording was not part of the training
set. In the beginning of the recording, a truck passed by the receiver
followed by a pickup truck. A car followed by a truck passed by the
receiver at around the $70^{th}$ second of the recording. A minibus
and a car passed by the receiver at the end of the recording. All
the events were correctly detected by CART and the MinDist classifiers.
The MinDist classifier produced some false alarms, which were reduced
by combining the results of the classifiers (bottom figure). 
\begin{figure}[!h]
\begin{centering}
\includegraphics{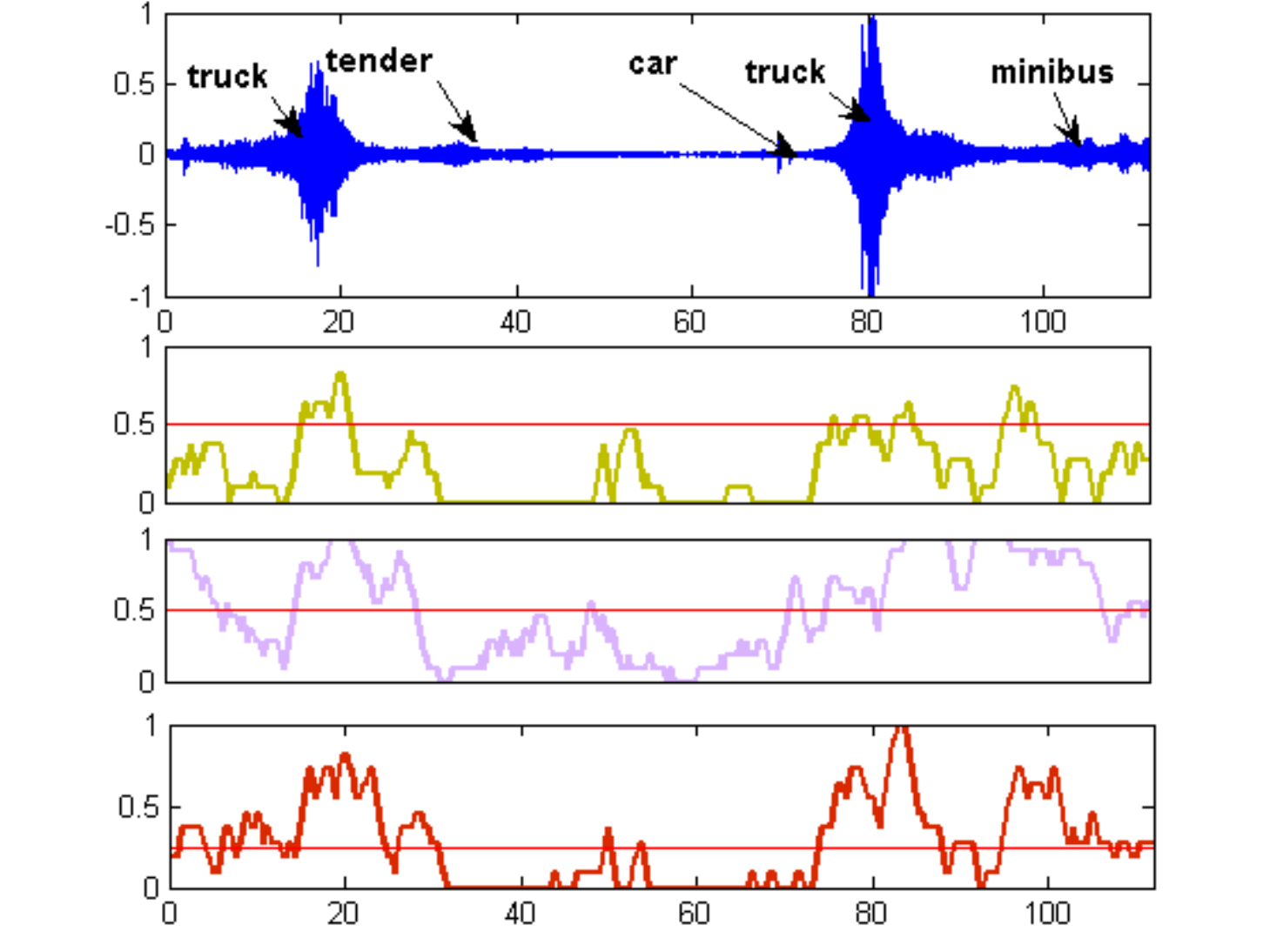}
\par\end{centering}

\caption{Results for test recording $\#5$. In the beginning of the recording,
a truck passed by the receiver followed by a pickup truck. A car followed
by a truck passed by the receiver at around the $70^{th}$ second
of the recording. A minibus and a car passed by the receiver at the
end of the recording. This recording was not part of the training
set.}

\label{sig129} 
\end{figure}

\item [{Recording~$\#6$:}] We display in Fig. \ref{sig206} the results
for test recording $\#6$. The recording was not part of the training
set. Two trucks passed by the receiver in opposing directions at around
the $50^{th}$ second of the recording. Strong wind was present. The
trucks were correctly detected by the CART and the MinDist classifiers.
The MinDist classifier produced some false alarms, which were reduced
by combining the results of the classifiers (bottom figure). 
\begin{figure}[!h]
\begin{centering}
\includegraphics{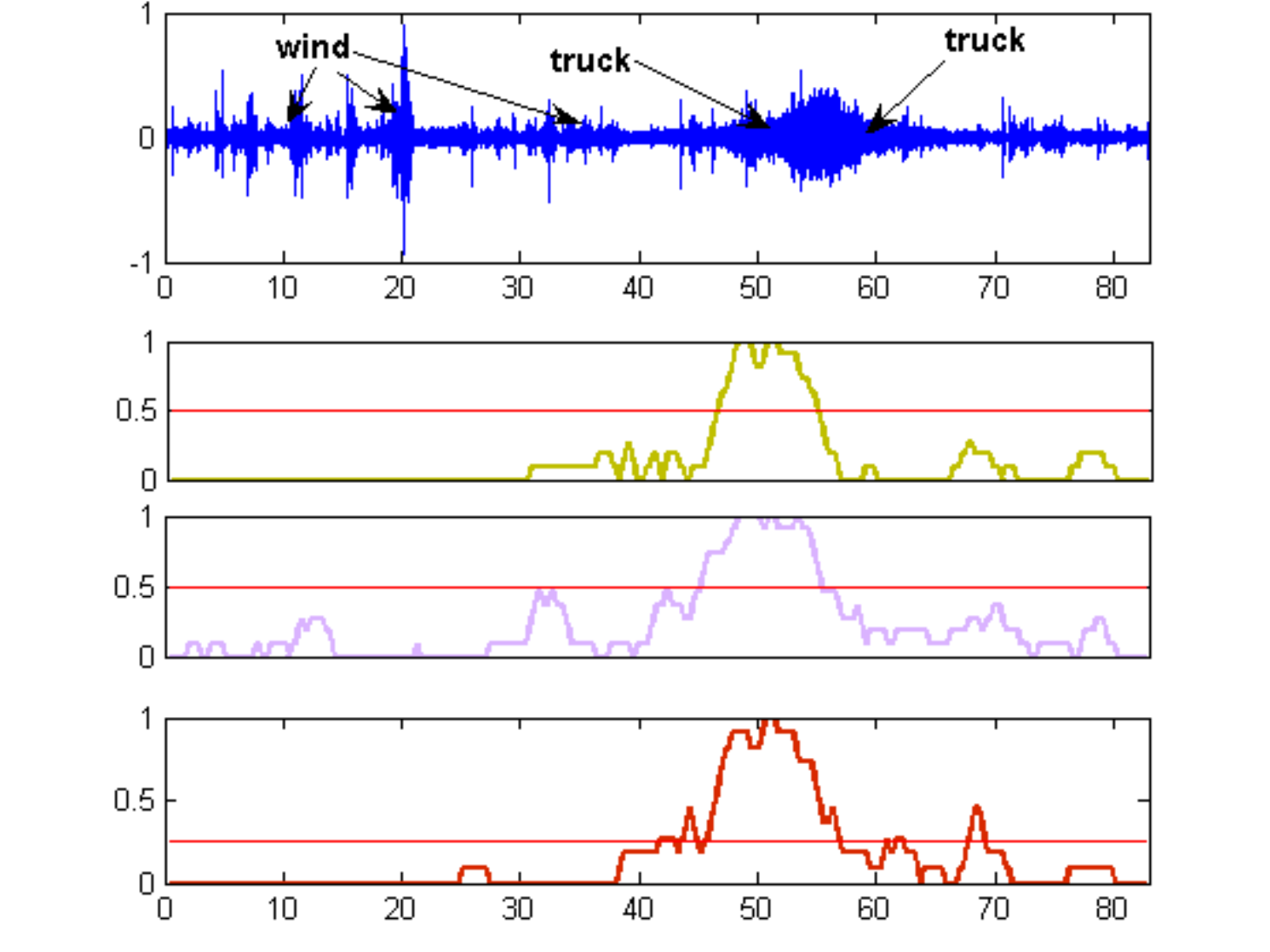}
\par\end{centering}

\caption{Results for test recording $\#6$. Two trucks passed by the receiver
in opposite directions at around the $50^{th}$ second of the recording.
Strong wind was present at the site. The recording was not part of
the training set.}

\label{sig206} 
\end{figure}

\item [{Recording~$\#7$:}] We display in Figs. \ref{sig441} the results
for test recording $\#7$. The recording was not part of the training
set. A truck passed by the receiver from the $30^{th}$ second to
the $190^{th}$ second of the recording. Then, a strong sound of an
airplane dominated the recording until the end of the recording. Strong
wind sound appeared throughout the recording. The truck was correctly
detected by the CART and the MinDist classifiers. The MinDist classifier
produced some false alarms, which were reduced by combining the results
of the classifiers (bottom figure). The plane was correctly not assigned
to the V class. 
\begin{figure}[!h]
\begin{centering}
\includegraphics{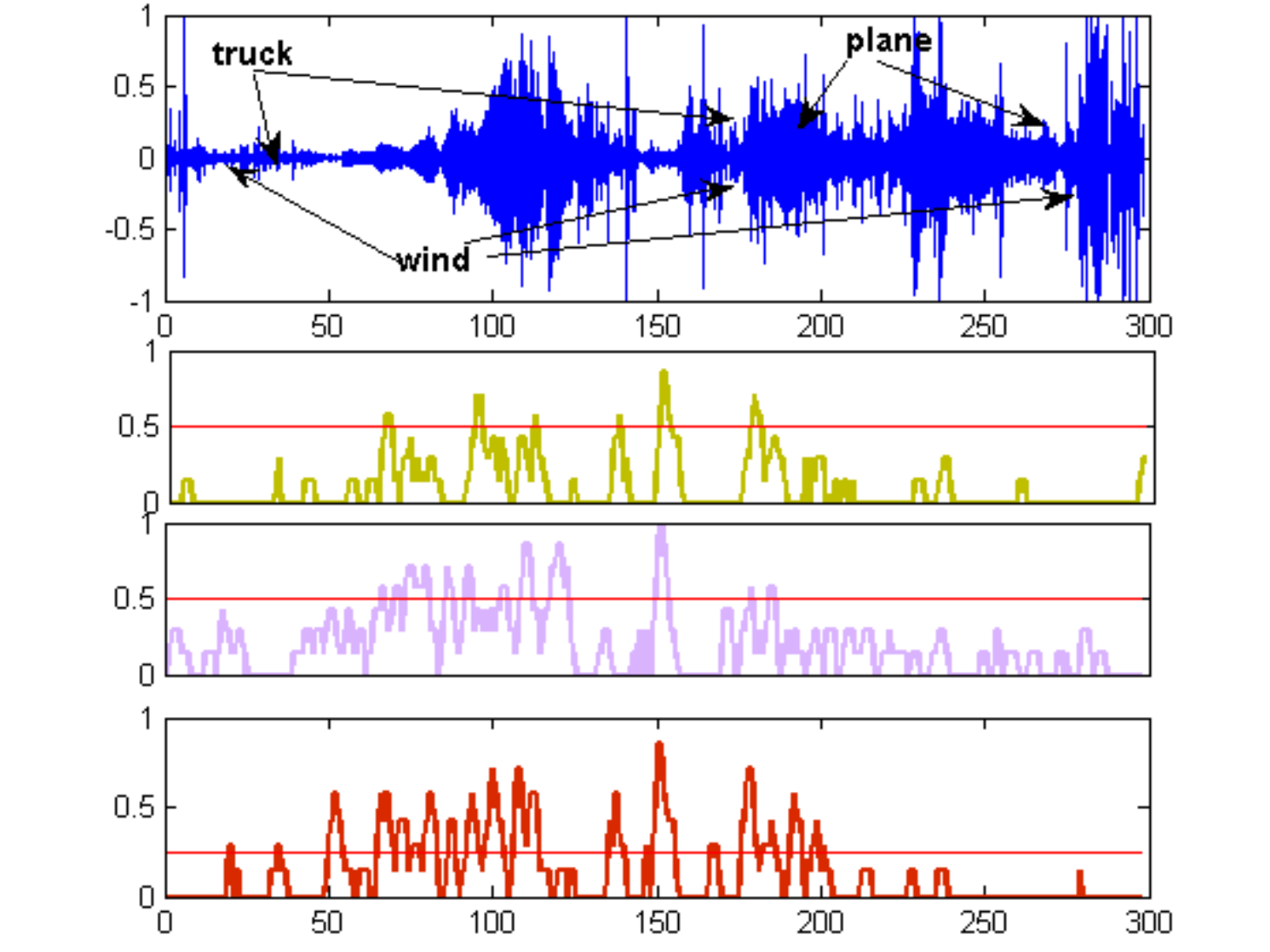}
\par\end{centering}

\caption{Results for test recording $\#7$. A truck passed by the receiver
from the $30^{th}$ second to the $190^{th}$ second of the recording.
Then, a strong sound of an airplane dominated the recording until
the end of the recording. Strong wind sound appeared throughout the
recording. This recording was not part of the training set.}

\label{sig441} 
\end{figure}

\item [{Recording~$\#8$:}] We display in Figs. \ref{sig251} the results
for test recording $\#8$. The recording was not part of the training
phase. A truck followed by a minibus passed by the receiver at around
the $40^{th}$ second of the recording. Another truck passed by at
around the $65^{th}$ second. Strong wind was present. The vehicles
were correctly detected by the CART and the MinDist classifiers. The
MinDist classifier produced some false alarms, which were eliminated
by combining the results of the classifiers (bottom figure).

\begin{figure}[!h]
\begin{centering}
\includegraphics{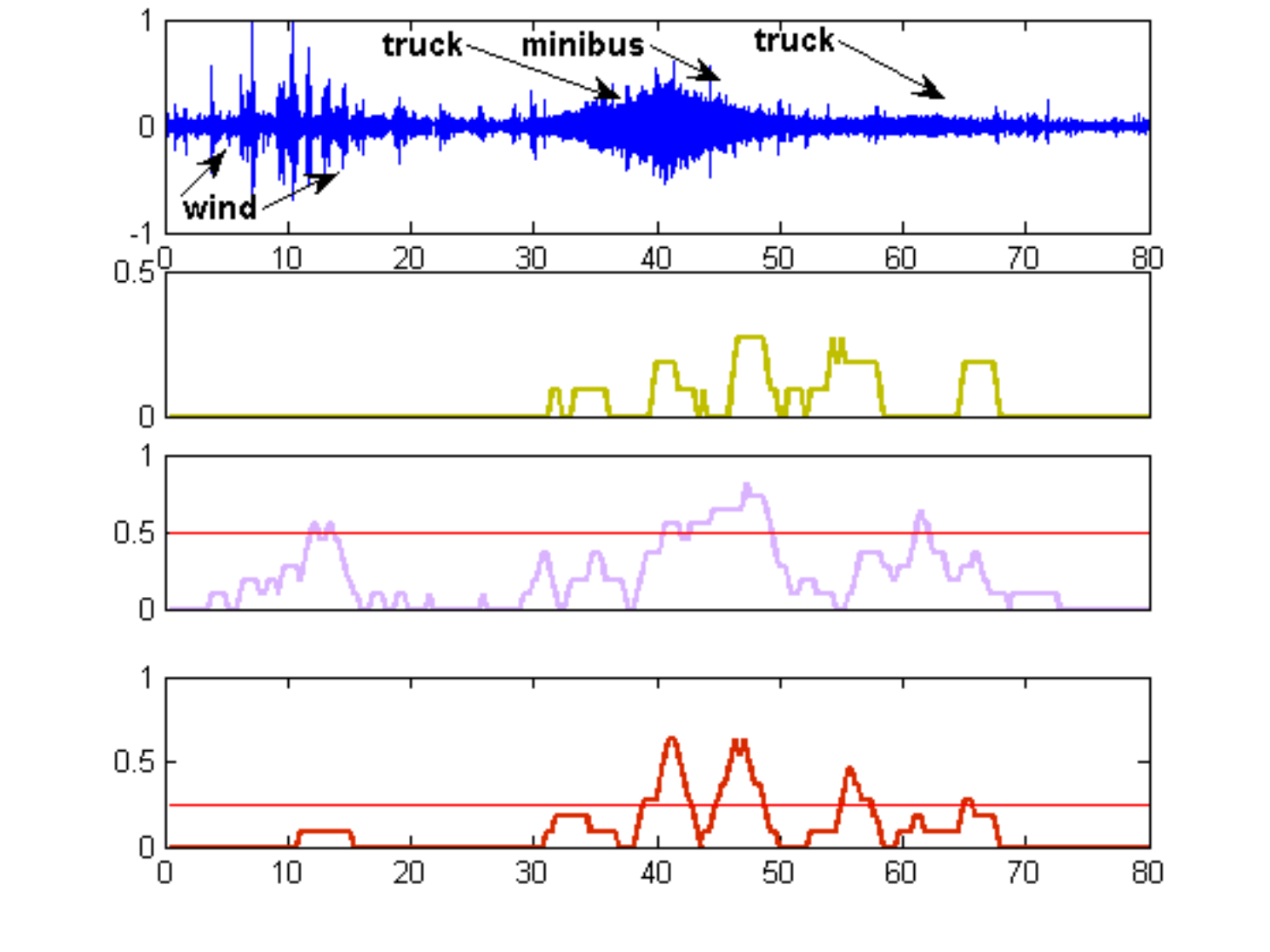}
\par\end{centering}

\caption{Results for test recording $\#8$. A truck followed by a minibus passed
by the receiver at around the $40^{th}$ second of the recording.
Another truck passed by at around the $65^{th}$second. Strong wind
was present. This recording was not part of the training set.}

\label{sig251} 
\end{figure}

\item [{Recording~$\#9$:}] We display in Fig. \ref{sig438} the results
for test recording $\#9$. The recording was not part of the training
set. A sound of a truck was heard between the $15^{th}$ and the $50^{th}$
seconds and also between the $80^{th}$ and the $110^{th}$ seconds
of the recording. An airplane sound appeared next. It lasted until
the end of the recording. The truck was correctly detected by the
CART and the MinDist classifiers. The plane was correctly not assigned
to the V class. The MinDist classifier performed poorly. 
\begin{figure}[!h]
\begin{centering}
\includegraphics{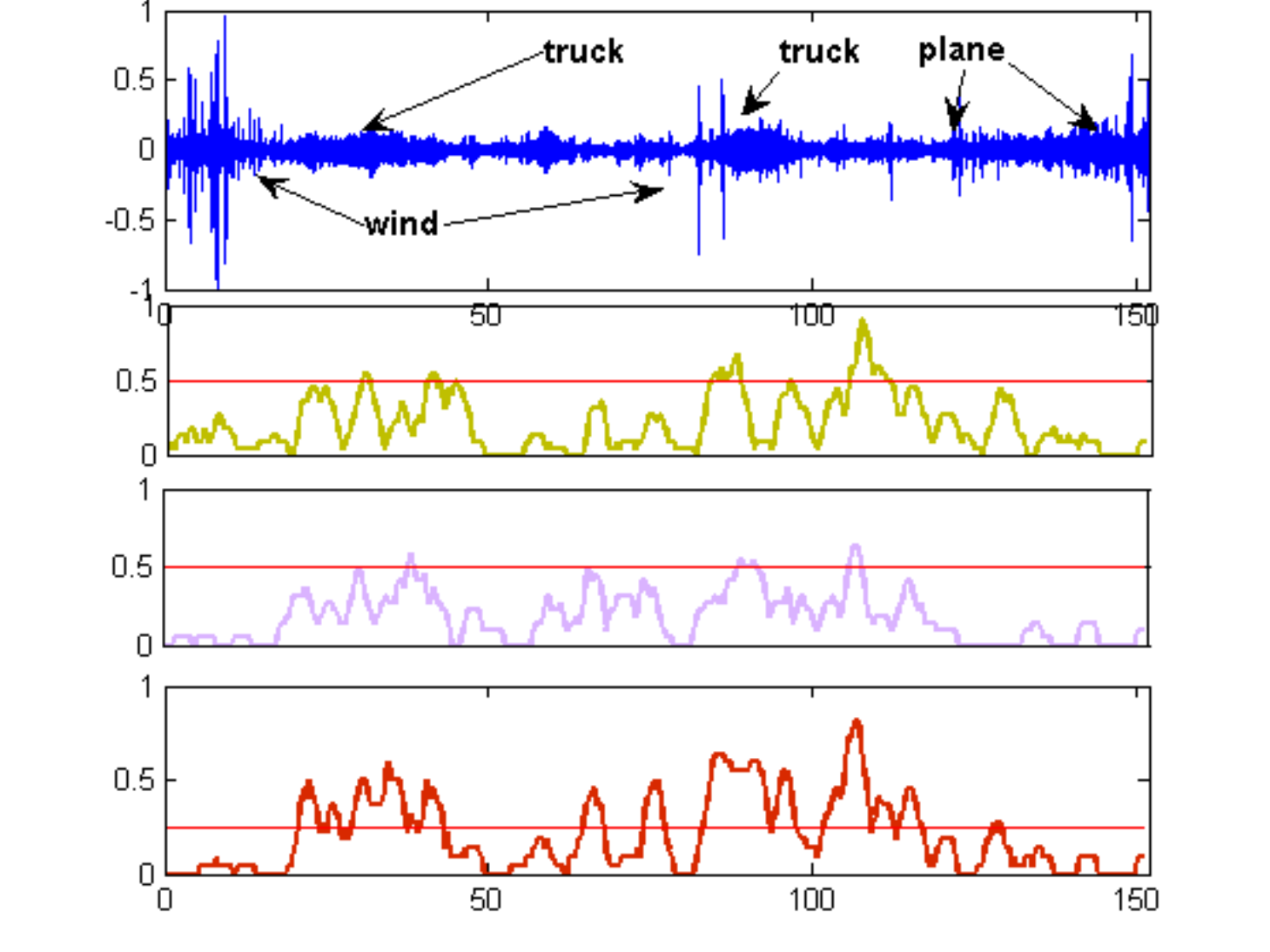}
\par\end{centering}

\caption{Results for test recording $\#9$. A sound of a truck was heard between
the $15^{th}$ and the $50^{th}$ seconds and also between the $80^{th}$
and the $110^{th}$ seconds of the recording. An airplane sound appeared
next. It lasted until the end of the recording. This recording was
not part of the training phase.}

\label{sig438} 
\end{figure}

\end{description}

\paragraph{Comments}
\begin{itemize}
\item The detection experiments demonstrate the relevance of our approach
to feature extraction. 
\item A combination of three schemes for feature extraction performs better
than any single scheme. 
\item Combining the MinDist and the CART classifiers significantly reduces
the number of false alarms. 
\item The algorithm produced satisfactory detection results even when the
conditions of the real signals essentially differ from the training
data and the surrounding conditions. When the conditions of the captured
signals are close to the training conditions, the detection is almost
perfect. 
\item The algorithm performs similarly on signal\ with sampling rates of
1000 SPS and 600 SPS. In a few cases, the results for SR of 1000 SPS
were significantly better than those for SR of 600 SPS.
\end{itemize}

\section{Conclusions and discussion}

\label{sec:s7} \label{sec:s5} We presented a robust algorithm that
detects the arrival of a vehicle of arbitrary type via the analysis
of its acoustic signature using an existing database of recorded and
processed acoustic signals for comparison.

To minimize the number of false alarms, we constructed an acoustic
signature of a certain vehicle using the distribution of the energies
among blocks which consist of its wavelet packet coefficients. This
distribution serves as an averaged version of the Fourier spectrum
of the signal. To reduce the dimensionality of the features sets,
we designed a scheme of random search for the near-optimal footprint
(RSNOFP), which proved to be an efficient tool for the extraction
of a small number of characteristic features of the objects to be
detected.

The following building blocks were used for the detection: (a) a classifier
that is based on the minimal distance (MinDist) from the reference
datasets and (b) the Classification and Regression Tree (CART) classifier.
These classifiers cross-validated one another. The detection process
is fast and can be implemented in real time.

This technology, which has many algorithmic variations, is generic
and can be used to solve a wide range of classification and detection
problems, which are based on acoustic processing such as process control,
and, more generally, for classification and detection of signals which
have near-periodic structure. Distinguishing between different vehicles
can also be achieved via this technology.

The successful results were obtained using a low sampling rate. Therefore,
the proposed method works well using very simple low-budget equipment.

\section{Appendix I: The wavelet and wavelet packet transforms}

\label{apendix1} Wavelet, in general, and wavelet packet, in particular,
transforms are widespread and have been described comprehensively
in the literature \cite{Daub92,wic,Mallat98}. Therefore, we restrict
ourselves to mention only relevant facts that are necessary to understand
the construction of the algorithm.

The output from the application of the \emph{wavelet transform} to
a signal $f$ of length $n=2^{J}$ is a set of $n$ correlated coefficients
of the signal with scaled and shifted versions of two basic waveforms
-- the father and mother wavelets. The transform is implemented through
iterated application of a conjugate pair of low-- ($L$) and high--
($H$) pass filters followed by downsampling. In the first decomposition
step, the filters are applied to $f$ and, after downsampling, the
result has two blocks $w_{0}^{1}$ and $w_{1}^{1}$ of the first scale
\c{s}, each of size $n/2$. These blocks consist of the correlation
coefficients of the signal with 2-sample shifts of the low frequency
father wavelet and the high frequency mother wavelet, respectively.
The block $w_{0}^{1}$ contains the coefficients necessary for the
reconstruction of the low-frequency component of the signal. Because
of the orthogonality of the filters, the energy ($l_{2}$ norm) of
the block $w_{0}^{1}$ is equal to that of the component $w_{0}^{1}$.
Similarly, the high frequency component $w_{1}^{1}$ can be reconstructed
from the block $w_{1}^{1}$. In this sense, each decomposition block
is linked to a certain half of the frequency domain of the signal.

While block $w_{1}^{1}$ is stored, the same procedure is applied
to block $w_{0}^{1}$ in order to generate the second level (scale)
of blocks $w_{0}^{2}$ and $w_{1}^{2}$ of size $n/4$. These blocks
consist of the correlation coefficients with 4-sample shifts of the
two times dilated versions of the father and mother wavelets. Their
spectra share the low frequency band previously occupied by the original
father wavelet. In a similar manner, $w_{0}^{2}$ is decomposed and
the procedure is repeated $m$ times. Finally, the signal $f$ is
transformed into a set of blocks $f\longrightarrow\{w_{0}^{m},\, w_{1}^{m},\, w_{1}^{m-1},\, w_{1}^{m-2},\ldots,\, w_{1}^{2},\, w_{1}^{1}\}$
up to the $m$-th decomposition level. This transform is orthogonal.
One block is remained at each level (scale) except for the last one.
Each block is related to a single waveform. Thus, the total number
of waveforms involved in the transform is $m+1$. Their spectra cover
the whole frequency domain and split it in a logarithmic form. Each
decomposition block is linked to a certain frequency band (not sharp)
and, since the transform is orthogonal, the $l_{2}$ norm of the coefficients
of the block is equal to the $l_{2}$ norm of the component of the
signal $f$ whose spectrum occupies this band.

Through the application of the \emph{wavelet packet} transform, many
more waveforms, namely, $2^{j}$ waveforms at the $j$-th decomposition
level are involved. The difference between the wavelet packet and
wavelet transforms begins in the second step of the decomposition.
Now both blocks $w_{0}^{1}$ and $w_{1}^{1}$ are stored at the first
level and at the same time both are processed by the pair of $L$
and $H$ filters, which generate four blocks $w_{0}^{2},\, w_{1}^{2},\, w_{2}^{2},\, w_{3}^{2}$
in the second level. These are the correlation coefficients of the
signal with 4-sample shifts of the four libraries of waveforms whose
spectra split the frequency domain into four parts. All of these blocks
are stored in the second level and transformed into eight blocks in
the third level, etc. The involved waveforms are well localized in
time and frequency domains. Their spectra form a refined partition
of the frequency domain (into $2^{j}$ parts in scale $j$). Correspondingly,
each block of the wavelet packet transform describes a certain frequency
band.

\begin{figure}[!h]
\begin{centering}
\includegraphics[width=0.85\columnwidth]{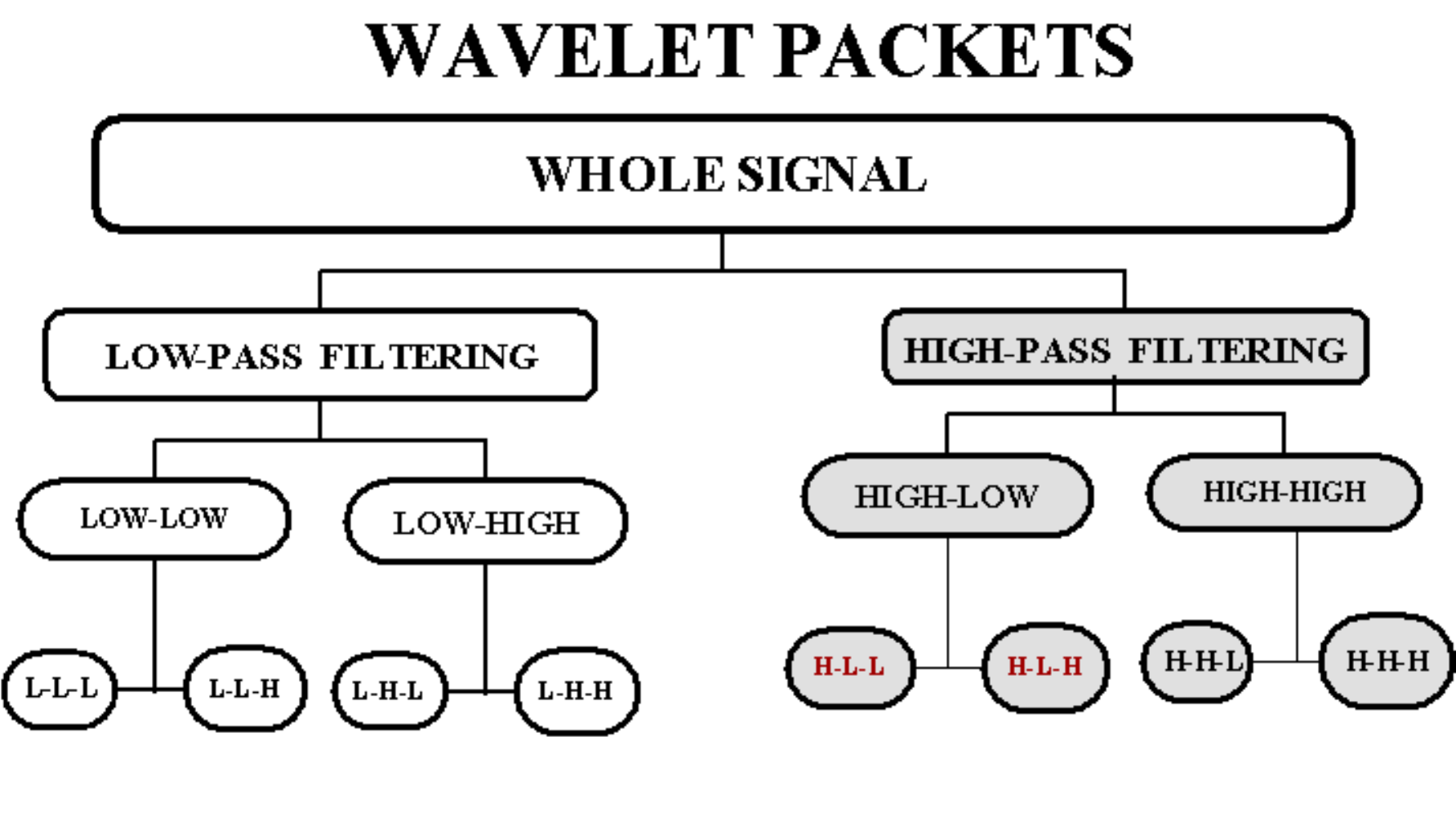}
\par\end{centering}

\caption{Flow of the wavelet packet decomposition.}

\label{WALSTRE4} 
\end{figure}

The flow of the wavelet packet transform is given by Fig. \ref{WALSTRE4}.
The partition of the frequency domain corresponds approximately to
the location of blocks in the diagram. 

\begin{figure}[!h]
\begin{centering}
\includegraphics[width=0.495\columnwidth]{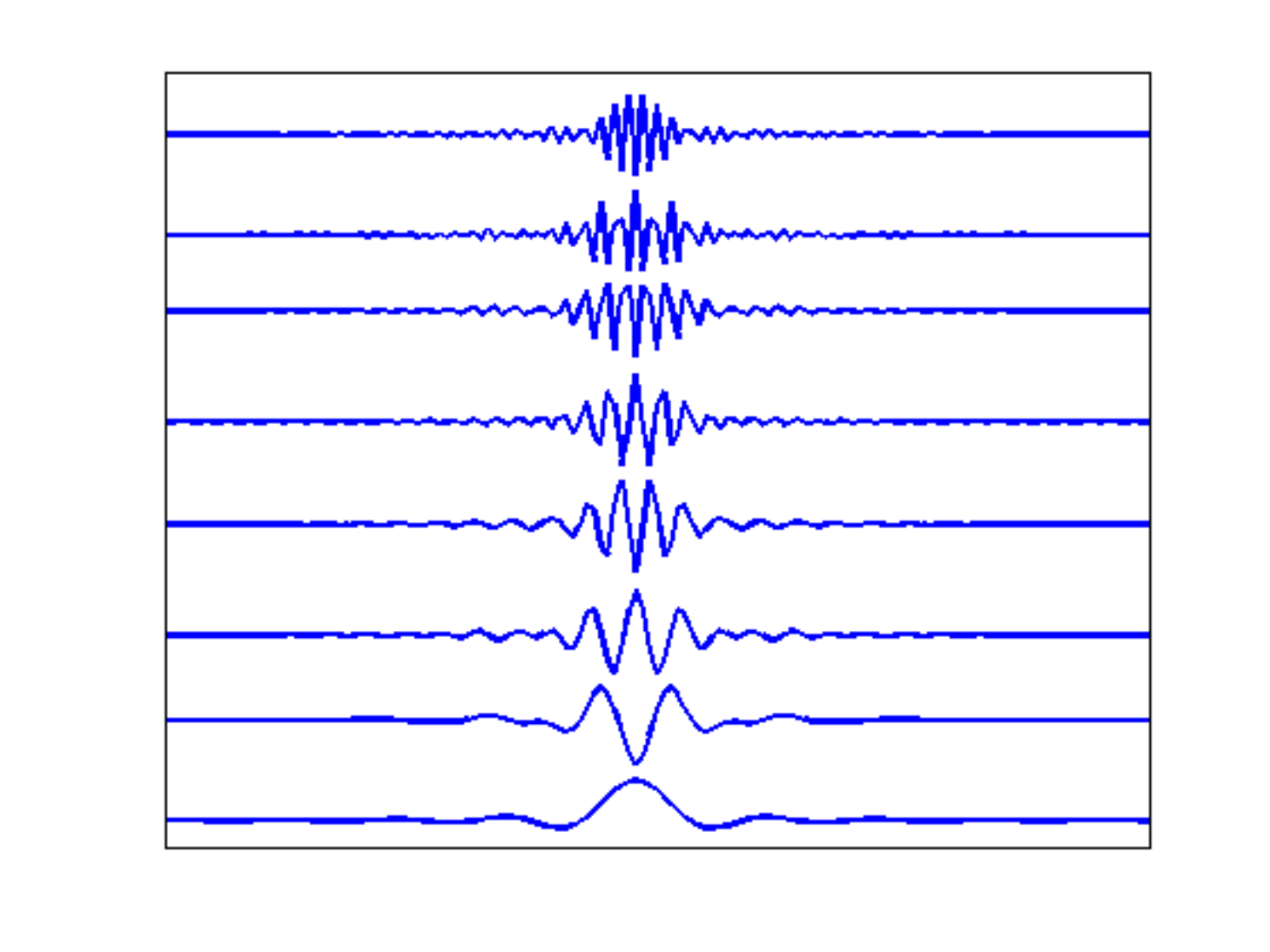}\includegraphics[width=0.495\columnwidth]{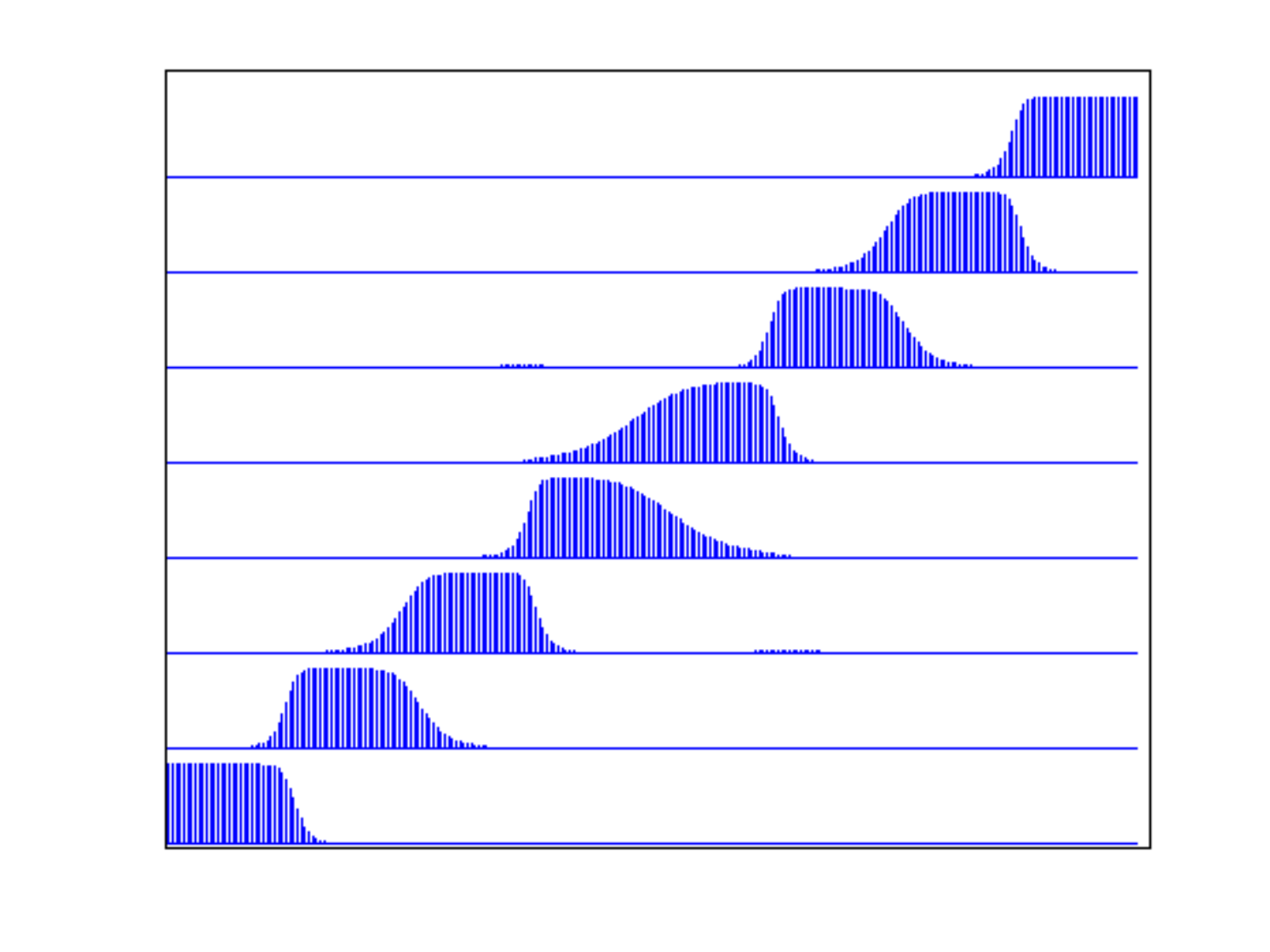}
\par\end{centering}

\caption{Wavelet packets derived from the spline of 6-th order after decomposition
into three scales (left) and their spectra (right).}

\label{spl6} 
\end{figure}

There are many wavelet packet libraries. They differ from each other
by their generating filters $L$ and $H$, the shape of the basic
waveforms and their frequency content. In Fig. \ref{spl6}, we display
the wavelet packets derived from the spline of 6-th order after decomposition
into three scales. While the splines do not have a compact support
in time domain, they are well localized. They produce perfect splitting
of the frequency domain (see Fig. \ref{spl6} right). 

\begin{figure}[!h]
\begin{centering}
\includegraphics[width=0.495\columnwidth]{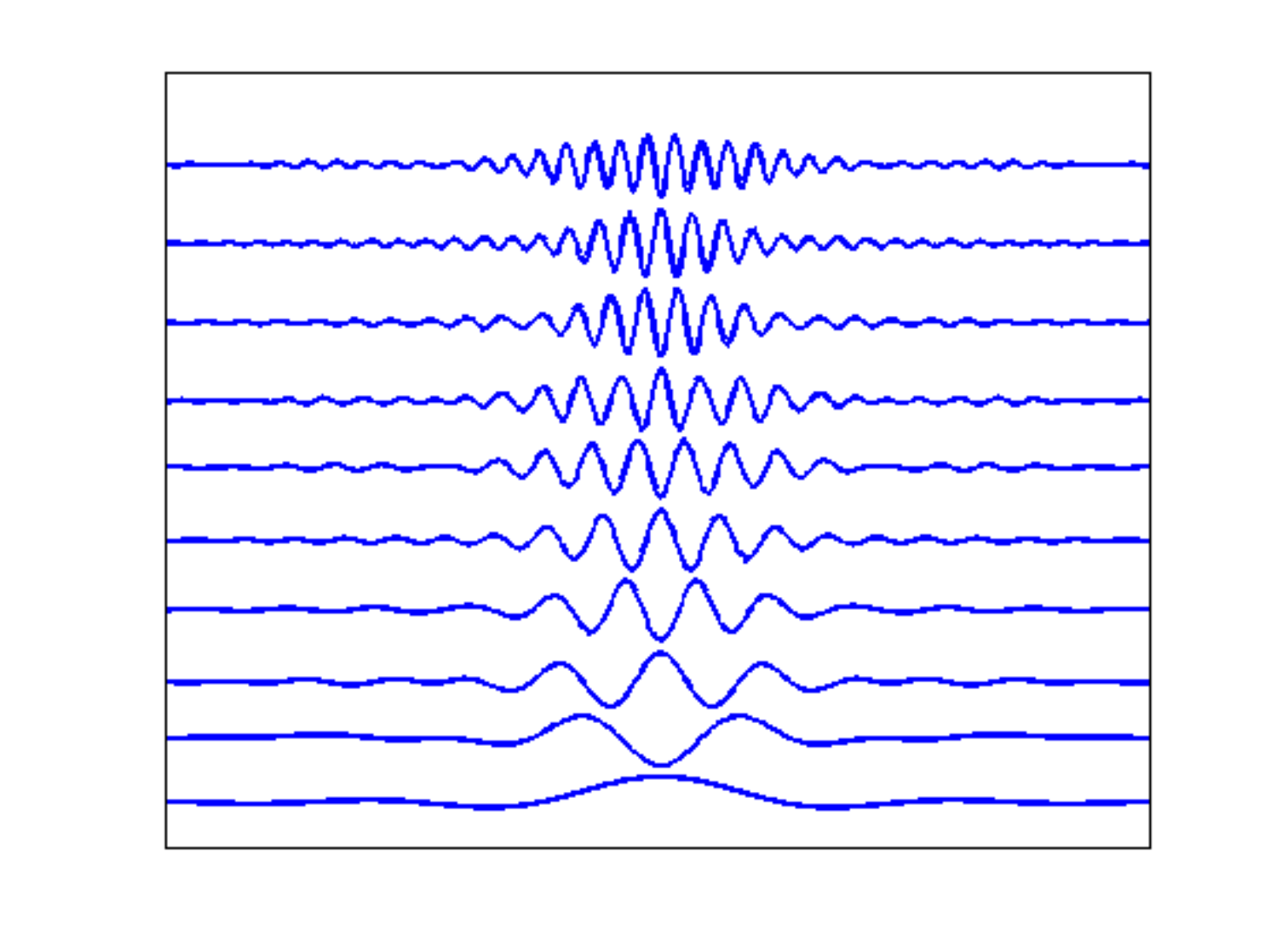}
\includegraphics[width=0.495\columnwidth]{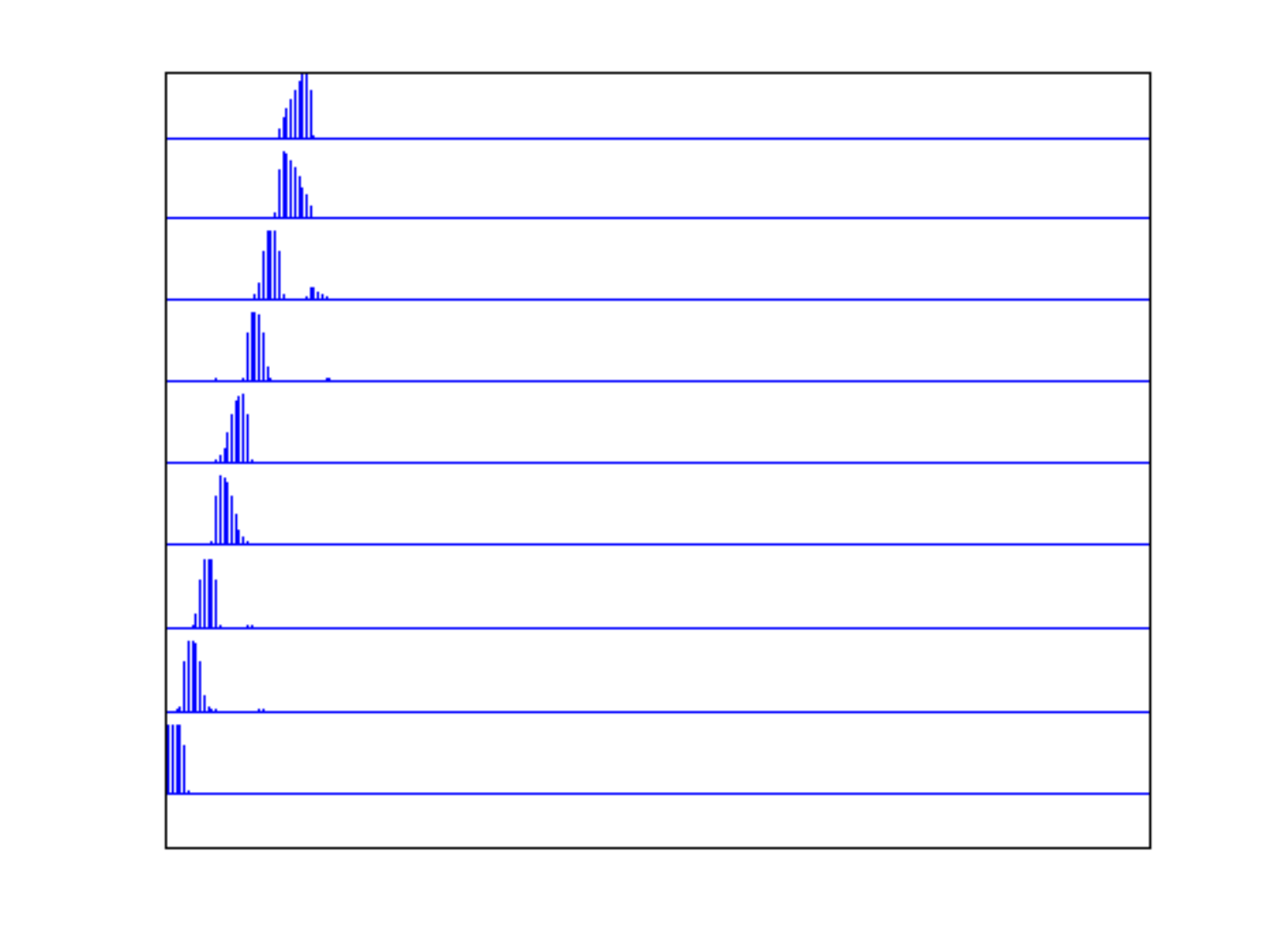}
\par\end{centering}

\caption{Wavelet packets derived from the spline of 6-Th order after decomposition
into six scales (left) and their spectra (right).}

\label{spl66} 
\end{figure}

There is a duality in the nature of the wavelet coefficients of a
certain block. On one hand, they indicate the presence of the corresponding
waveform in the signal and measure its contribution. On the other
hand, they evaluate the contents of the signal inside the related
frequency bands. One may argue that the wavelet packet transform bridges
the gap between time-domain and frequency-domain representations of
a signal. As one advances into coarser level (scale), a better frequency
resolution is seen at the expense of time domain resolution and vice
versa. In principle, the transform of a signal of length $n=2^{J}$
can be implemented up to the $J$-th decomposition level. At this
level there exist $n$ different waveforms, which are close to the
sine and cosine waves with multiple frequencies. In Fig. \ref{spl66},
we display a few wavelet packets derived from the spline of the 6-th
order after decomposition into six levels. The waveforms resemble
the windowed sine and cosine waves, whereas their spectra split the
Nyquist frequency domain into 64 bands. 

\section{Appendix II: The Random Search for a Near Optimal FootPrint (RSNOFP)
scheme}

\label{appendix2} 

As mentioned above, three versions of the RSNOFP scheme are used:
\begin{description}
\item [{RSNOFP~-~version~I:}]~

A random matrix $R_{1}$ of size $r\times\lambda$, where $r\ll\lambda$
(typically, $r=20$) is created. Entries of the matrix $R_{1}$ are
Gaussian random variables (see\cite{DT06} for other options to choose
$R_{1}$). The columns in the matrix are normalized. The matrix $B^{v}$
(defined in Section \ref{sec:ss32}) is multiplied by the matrix $R_{1}$.
As a result, we obtain a new matrix $C^{v}=B^{v}{\bf \cdot}R_{1}^{T}=\left(C_{i,j}^{v}\right)$
of size $M^{v}\times r$. Each row in $C^{v}$ is associated with
the corresponding slice from $A^{v}$. This procedure reduces the
dimensionality of the rows in $B^{v}$ by using the random projection
scheme that was described in Section \ref{sub:Local-methods}.

Next, we select the most significant columns (with respect to the
average $l_{1}$ norm) of the matrix $C^{v}$. Let 
\[
\vec{c}_{j}^{v}=\frac{1}{M^{v}}\sum_{i=1}^{M^{v}}|C_{i,j}^{v}|.
\]
be the average $l_{1}$ norm of the \emph{j}-Th column where $j=1,...,r$
and let $c^{v}=\left(c_{j}^{v}\right)_{j=1,...,r}$ be the vector
of all averages. We denote by $K$ the set of indices $k<r$ of the
largest coordinates of the vector $\vec{c}^{v}$ (typically, $k=12$).
Then, the columns, whose indices do not belong to $K$, are removed
from the matrix $C^{v}$ and the matrix $D^{v}$ of size $M^{v}\times k$
is obtained. This operation is equivalent to multiplication of $B^{v}$
with the matrix $H$ of size $k\times\lambda$, which is derived from
$R_{1}$ by removing the rows, whose indices do not belong to $K$.
Thus, the initial matrix $A^{v}$ consisting of the V-class slices,
whose size was, for example, $M^{v}\times1024$, is reduced to the
matrix $D^{v}$ of the \emph{random footprints} of the slices. The
size of $D^{v}$ is $M^{v}\times k$. To produce a similar reduction
of the matrix $A^{n}$ of N-class slices, we multiply the N-class
energy matrix $B^{n}$ with the matrix $H$. As a result, we obtain
the random footprints matrix $D^{n}=B^{n}{\bf \cdot}H$ of size $M^{n}\times k$.
We consider the coordinates of the $i$-th row of the matrices $D^{v}$
and $D^{n}$ as the set of $k$ characteristic features of the $i$-th
slice from the matrix $A^{v}$ and $A^{n}$, respectively.

Next, we calculate the Mahalanobis distances $\mu_{i}$, $i=1,\ldots,M^{v},$
from each row in the V-class matrix $D^{v}$ to the matrix $D^{n}$.
We denote by 
\[
\Delta=\frac{1}{M^{v}}\sum_{i=1}^{M^{v}}\mu_{i}
\]
the average of the sequence $\left\{ \mu_{i}\right\} $.

The value $\Delta$ is considered to be the distance between the feature
sets $D^{v}$ and $D^{n}$. The matrices $D^{v}$, $D^{n}$, $H$
and the value $\Delta$ are stored and we proceed to optimize the
features.

We repeat the above operations using a random matrix $R_{2}$, whose
structure is similar to the structure of the matrix $R_{1}$. As a
result, we obtain the feature matrices $D_{2}^{v}$ and $D_{2}^{n}$,
the random matrix $H_{2}$ and the distance value $\Delta_{2}$. The
distance value $\Delta_{2}$ is compared to the stored value $\Delta$.
Assume, $\Delta_{2}>\Delta$. This means that the features matrices
$D_{2}^{v}$ and $D_{2}^{n}$ are better separated from one another
than the stored matrices $D^{v}$ and $D^{n}$. In this case, we replace
the values in $D^{v}$, $D^{n}$, $H$ and $\Delta$ by the values
in $D_{2}^{v}$, $D_{2}^{n}$, $H_{2}$ and $\Delta_{2}$, respectively.
If $\Delta_{2}\leq\Delta$ then the values in $D^{v}$, $D^{n}$,
$H$ and $\Delta$ are left intact.

We iterate this procedures up to 500 times. In the end, we stored
the features matrices $D^{v}$ and $D^{n}$ such that the ``distance\textquotedbl{}
$\Delta$ between them among all the iterations is maximal. We have
stored the reduced random matrix $H$ and the pattern matrices $D^{v}$
and $D^{n}$, which will be used in the identification phase. These
items are denoted as $D_{rand}^{v}$, $D_{rand}^{n}$ and $H_{rand}$.

\item [{RSNOFP~-~Version~II:}] This version is similar, to some extent,
to Version I. The difference is that, instead of selecting the most
significant columns in the matrix $C^{v}$, we apply the Principal
Component Analysis (PCA) to this matrix. As a result, we obtain the
matrix $P=\left(P_{i,j}\right)$ of size $r\times r$. Each column
of $P$ contains the coefficients of one principal component. The
columns are arranged in decreasing component variance order. The size
of $P$ is reduced to $r\times k$ by retaining only the first $k$
columns 
\[
P_{k}=\left(P_{i,j}\right),\; i=1,...,r,\; j=1,...,k.
\]
 We obtain the feature matrix $D^{v}$ for the V-class by multiplying
$C^{v}$ by $P_{k}$:

\[
D^{v}=C^{v}{\bf \cdot}P_{k}=B^{v}{\bf \cdot}R_{1}^{T}{\bf \cdot}P_{k}=B^{v}{\bf \cdot}H,\mbox{ where }H=R_{1}^{T}{\bf \cdot}P_{k}.
\]
 The size of the matrix $\rho$ is $\lambda\times k$. Similarly,
we produce the feature matrix $D^{n}$ for the N-class: $D^{n}=B^{n}{\bf \cdot}H$.
Similarly to Version I, we measure the ``distance\textquotedbl{}
$\Delta$ between the feature sets $D^{v}$ and $D^{n}$. The matrices
$D^{v}$, $D^{n}$, $H$ and the value $\Delta$ are stored and we
proceed to optimization of the features, which is identical to Version
I. Finally, we stored the features matrices $D^{v}$ and $D^{n}$
and the matrix $H$. These items are denoted by $D_{pca}^{v}$, $D_{pca}^{n}$
and $H_{pca}$.

\item [{RSNOFP~-~version~III:}] This version differs from versions I
and II. Here, we do not multiply the energy matrix $B^{v}$ by a random
matrix but instead we perform a random permutation of the columns
and retain the first $r$ columns. Thus, we get the matrix $C^{v}$
of size $M^{v}\times r$. Note, that this transform can be presented
as the multiplication of the matrix $B^{v}$ by a matrix $T$ of size
$\lambda\times r$, $C^{v}=B^{v}{\bf \cdot}T$, where each column
consists of zeros except for one entry, which is equal to 1. \\
\textbf{Example:} Assume that the matrix $T$ is of size $4\times3$
and corresponds to the permutation $[1\,2\,3\,4]\to[3\,1\,4\,2]$
of the columns of a matrix of size $4\times4$ while retaining the
first three columns: 
\[
T=\left(\begin{array}{ccc}
0 & 1 & 0\\
0 & 0 & 0\\
1 & 0 & 0\\
0 & 0 & 1
\end{array}\right).
\]
 The other operations are similar to the operations in Version II.
We apply the PCA algorithm to the matrix $C^{v}$, which results in
the principal components matrix $P=\left(P_{i,j}\right)$ of size
$r\times r$ of coefficients of the principal components. The size
of $P$ is reduced to $r\times k$ by retaining only the first $k$
columns 
\[
P_{k}=\left(P_{i,j}\right),\; i=1,...,r,\; j=1,...,k.
\]
 We obtain the feature matrix $D^{v}$ for the V-class by multiplying
$C^{v}$ by $P_{k}$:

\[
D^{v}=C^{v}{\bf \cdot}P_{k}=B^{v}{\bf \cdot}T{\bf \cdot}P_{k}=B^{v}{\bf \cdot}H,\mbox{ where }H=T{\bf \cdot}P_{k}.
\]
 The size of the matrix $H$ is $\lambda\times k$. We construct,
in a similar manner, the feature matrix $D^{n}$ that corresponds
to the N-class: $D^{n}=B^{n}{\bf \cdot}H$. We measure the ``distance\textquotedbl{}
$\Delta$ between the sets of features $D^{v}$ and $D^{n}$. The
matrices $D^{v}$, $D^{n}$, $H$ and the value $\Delta$ are stored
and we proceed to optimize the features, using the same procedure
as in Versions I and II. Finally, the features matrices $D^{v}$ and
$D^{n}$ and the matrix $H$ are stored. They are denoted as $D_{perm}^{v}$,
$D_{perm}^{n}$ and $H_{perm}$. 

\end{description}
We illustrate the relations between the RSNOFP procedures (version
II) by the diagram in Fig. \ref{rsn}.

\noindent \begin{center}
\begin{figure}[!h]
\noindent \begin{centering}
\includegraphics[width=0.6\paperwidth]{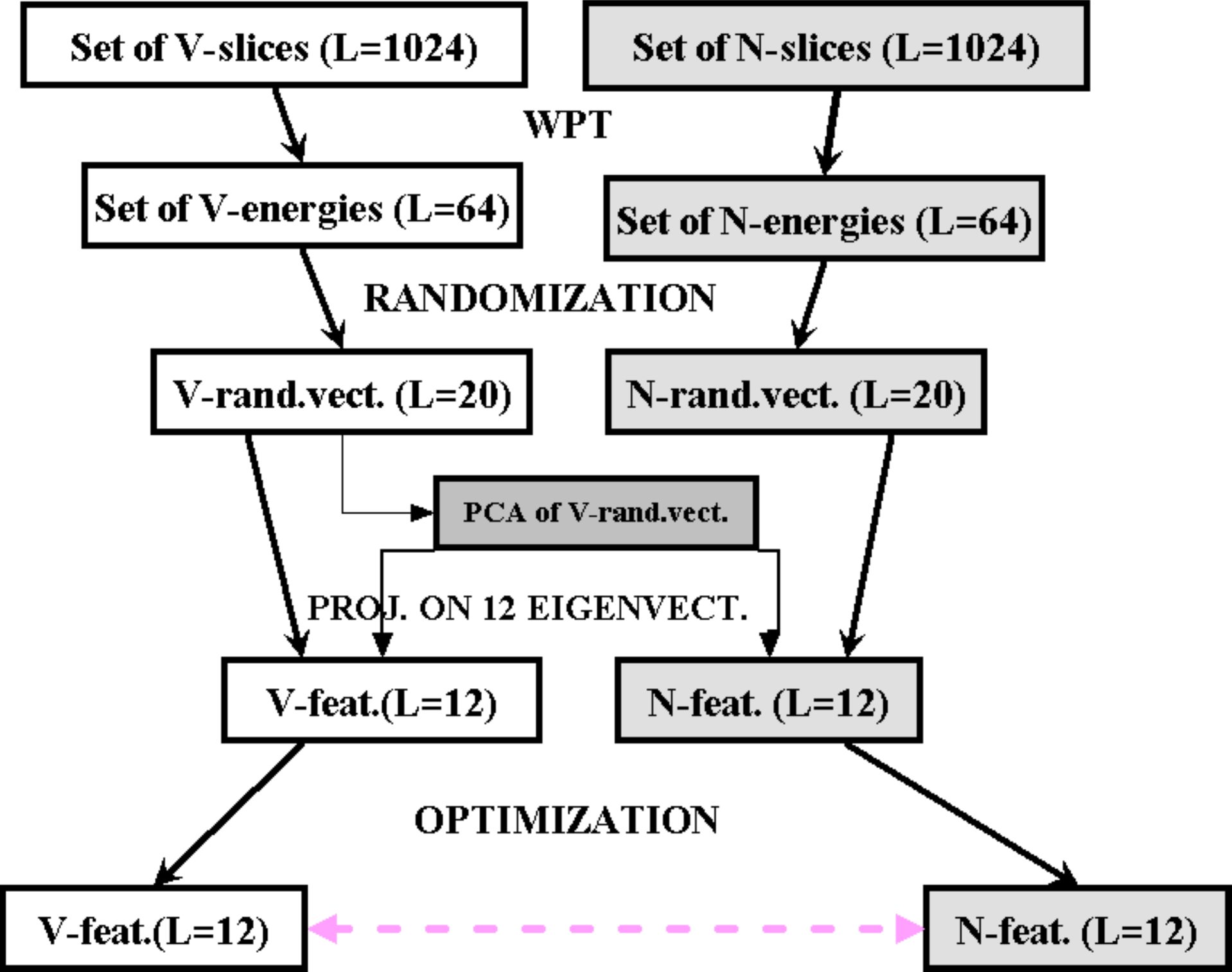}
\par\end{centering}

\caption{RSNOFP procedures (version II). WPT stands for wavelet packet transform.}

\label{rsn} 
\end{figure}

\par\end{center}

\chapter{Classification and Detection of High-Dimensional Medical Signals
via Dimensionality Reduction \label{cha:Classification-and-Detection-Vascular}}

In this chapter, we introduce a generic approach which clusters and
classifies datasets of acoustic signals. The proposed algorithms are
robust to noise and to diverse recording conditions in which the signals
were recorded. This methodology is able to detect different types
of events. 

Similarly to the algorithm in Chapter \ref{cha:Wavelet-based-acoustic},
the algorithms in this chapter consist of two steps: An offline training
(learning) step that clusters the data and a detection step that classifies
new data according to the clusters that were found in the first step.
Different classes of events can be added to the training set and consequently
to the detection phase. 

We present two algorithms for clustering and classification. They
apply a common feature extraction to the signals. However, the clustering
procedure they use is different. One uses the diffusion maps algorithm
(Chapter \ref{cha:Diffusion-Maps}) and the other uses principle component
analysis (Section \ref{sub:Global-methods}). The proposed algorithms
are generic and can be applied to various signal types for solving
different classification problems. We demonstrate the generality of
the algorithms on two different examples: detection of hypertension
in adolescents and detection of abnormalities in heart beats. The
robustness of the classification depends on the size and the quality
of the training set. 

This approach provides an alternative to the algorithm proposed in
Chapter \ref{cha:Automatic-identification-of} while sharing a few
common constructions.

\section{Introduction}

Clustering and classification of acoustic signals is a contemporary
problem. This suggests a generic approach to cluster and classify
datasets by discovering the geometric structure of the dataset. To
achieve this, a training step embeds the training data into a low-dimensional
space by using a dimensionality reduction scheme that preserves the
geometrical structure of the dataset. Given a new high-dimensional
data point for classification, it is first embedded into the low-dimensional
space and its classification is determined according to its neighboring
points in the low-dimensional space.

A preliminary step in the analysis of a signal is its decomposition
into overlapping subsequences we refer to as \emph{windows}. The resulting
windows are regarded as data points in a high dimensional space where
the dimensionality depends on the length of the subsequences. This
high dimensional data contains dominating features along with redundancies.
Consequently, the governing features need to be extracted while removing
the redundancies. This is achieved by applying dimensionality reduction
to the high dimensional data. The low dimensional space involves a
small number of free parameters that convey the dominating features
of the data. % This process is commonly known as feature extraction and uses, among others, dimensionality reduction methods. 

The feature extraction procedures that are used in this chapter share
the \emph{Spline Wavelet Packet Transform} (Appendix I of Chapter
\ref{cha:Automatic-identification-of}) as a common preliminary step.
However, they differ in the second step they use - one procedure uses
the Diffusion Maps (Chapter \ref{cha:Diffusion-Maps}) algorithm while
the second uses Principal Component Analysis (Section \ref{sub:Global-methods}). 

The classification of acoustic signals is obtained via a two-phase
process:
\begin{itemize}
\item \textbf{Learning phase} (also referred to as training phase) in which
data with a-priory classification is analyzed and features, which
characterize it, are extracted and stored. The output of this phase
is a separation of the input data into clusters, where each cluster
represents a different class. %The clusters in example 1 of the experimental results, are the class of moving vehicles and the class of background noise. The
learning phase is a one-time offline procedure.
\item \textbf{Detection phase} in which new data is classified using a feature
database of known acoustic signatures. The detection phase is done
in real-time (on-line) and the new data is classified according to
the cluster it is the closest to (the Euclidean distance is minimal). 
\end{itemize}
We illustrate the validity of these algorithms by using them to perform
two detection tasks - each of which takes as input a different type
of acoustic signals: 
\begin{enumerate}
\item \label{enu:Detection-Pulse}Detection of early symptoms of arterial
hypertension in adolescents.
\item \label{enu:Detection-of-cardio}Detection of cardio-vascular diseases.
\end{enumerate}
% In example XXX, the goal is to detect, with a minimal number of false detections, sounds that are emitted by specific moving vehicles, by separating the sounds of the vehicles from the background sounds. This example has obvious military and industrial applications (such as security, safety, surveillance, process control, etc.). This classification is complex because of the great variability in the surrounding conditions of the recordings such as velocity of the vehicle, the vehicle types, the physical condition of the vehicle, the distances between the vehicles and the recording device, the road type (asphalt, dirt), etc. 

In both experiments \ref{enu:Detection-Pulse} and \ref{enu:Detection-of-cardio}
the algorithm should produce a minimal number of false detections.
The classification of the data above is a difficult task since the
pulses can vary significantly from one person to another. Providing
an \emph{in-vivo} solution to these medical detection problems is
important since it enables to diagnose these diseases without the
need of an invasive procedure. Furthermore, it can can be used for
the diagnosis of other diseases by using acoustic signatures of the
body such as pulse, as in the examples at hand, blood flow for the
detection of various arterial diseases, abdominal sounds for the detection
of gastrological problem, etc. In addition, the classification can
be done in real-time.

Details of related works that tackle similar problems can be found
in Section \ref{related}.

The rest of the chapter is organized as follows: The mathematical
background on which the algorithm rely is given in Section \ref{sec:Mathematical-Background}.
In Section \ref{sec:Clustering-and-Classification} we present the
main algorithms. The results of the two experiments are given in Section
\ref{sec:Experimental-Results}. Finally, in Section \ref{sec:Conclusions}
we summarize the results and conclude this chapter.

\section{\label{sec:Mathematical-Background}Mathematical background}

The proposed algorithms use various mathematical tools:
\begin{itemize}
\item \textbf{Spline wavelet} - Similarly to Chapter \ref{cha:Automatic-identification-of},
we extract features from the acoustic signals by applying the wavelet
packet transform using spline wavelets of the sixth order (Battle-Lemarie
\cite{Daub92}). The collection of energies in the blocks of wavelet
packet coefficients constitutes the features of the signal. A description
of the wavelet and wavelet-packet transforms is given in Chapter \ref{cha:Automatic-identification-of}. 
\item \textbf{Principal component analysis} - This is one of the two dimensionality
reduction methods that are applied to the features after they were
extracted. A description of this technique is given in Section \ref{sub:Global-methods}. 
\item \textbf{Diffusion maps} - This is the second dimensionality reduction
method that is applied to the features after they were extracted.
A description of this technique is given in Chapter \ref{cha:Diffusion-Maps}. 
\item \textbf{Geometric harmonics} - This is an out-of-sample extension
scheme which we describe in details below. It is used to embed a new
data point into the low-dimensional space during the real-time detection
phase. The classification of the point is determined according to
the class of the cluster which is the closest to it.
\end{itemize}

\subsection{{\normalsize \label{sub:Geometric-Harmonic-(out}}Geometric Harmonic
(out-of-sample extension)}

The geometric harmonic \cite{LafonThesis04,CL_GH06,LKC06} provides
a way to extend a known function $f$ on the learning set to a new
point outside the sampling set, using both the target function and
the geometry of the training set. A specific function that we extend
is the low dimensional representation which is computed on the training
set. This function is extended to data points outside the training
set as part of the detection algorithm.

Let $\Omega$ be a dataset and let $\Xi_{t}$ be its diffusion embedding
map as defined in Eq. \ref{eq:DM_embedding_map} where $\mu_{l}$
and $\phi_{l}$ are the eigenvalues and eigenvectors, of the Gaussian
kernel with width $\sigma$ on the training data $\Omega$, respectively.
We denote by ${\normalcolor \overline{\Omega}}$ the new dataset that
includes the new data points to which we wish to extend $f$. $\sigma>0$
is the scale of extension. 

The kernel can be evaluated in the entire space $\mathbb{R}^{d}$.
Consequently, we get 
\[
\mu_{l}\phi_{l}(x)=\sum_{y\in\Omega}e^{-||x-y||^{2}/\sigma^{2}}\phi_{l}(y),x\in\Omega
\]
where it is possible to use any $x\in\mathbb{R}^{d}$ on the right-hand-side
of the identity.

The Nystr�m extension \cite{fowlkes04spectral} from $\Omega\: to\:\mathbb{R}^{d}$
of the eigenfunctions is given by:
\begin{equation}
\bar{\phi}_{l}(x)=\frac{1}{\mu_{l}}\sum_{y\in\Omega}e^{-||x-y||^{2}/\sigma^{2}}\phi_{l}(y),\: x\in R^{d}.\label{eq:Nys-ext-phi}
\end{equation}
Note that $\phi_{l}$ is being extended to a distance that is proportional
to the distance from the new point $x$ to the training set $\Omega$. 

Any function on the training set can be decomposed into
\begin{equation}
f(x)=\sum_{l}\left\langle \phi_{l},f\right\rangle \phi_{l}(x),\,\, x\in\Omega.
\end{equation}
The Nystr�m extension of $f$ on the rest of the space $\mathbb{R}^{d}$
is given by:

\begin{equation}
\bar{f}(x)=\sum_{l}\left\langle \phi_{l},f\right\rangle \bar{\phi}_{l}(x),\,\, x\in R^{d}.\label{eq:Ny-ext-f}
\end{equation}

The problem with this scheme is the choice of the kernel of extension.
In the equations above, the same kernel that was used for the diffusion
map embedding is used for the extension. Consequently, the functions
will be extended to a distance that is proportional to the width $\sigma$
of the Gaussian. However, when the diffusion embedding is computed,
one strives to use as small a scale as possible, since the diffusion
maps algorithm approximates the eigenvectors of the Laplace-Beltrami
operator on the manifold and thus allows to discover the geometry
of the underlying structure of the dataset. On the other hand, when
extending a function, e.g. the diffusion coordinates outside the training
set, one wants to be able to extend them as far as possible in order
to maximize their generalization power. From the above it follows
that the $\sigma$ of the kernel, which is used for extending, should
be as big as possible. This contradicts the strive for a small $\sigma$
during the embedding process. Furthermore, this scale should not be
the same for all functions we are trying to extend. Functions with
large variations on $\Omega$ should have a limited range of extension
since it is more difficult to predict their values. As a result, one
should adapt the scale of the extension to the function to be extended.
Thus, the geometric harmonic algorithm uses two different kernels
- one for the embedding with a small $\sigma$ and another for the
extension using a larger $\sigma$. We denote them by $\sigma_{embed}$
and $\sigma_{ext}$, respectively. The eigenfunctions of the the embedding
kernel constitute the functions we wish to extend and the eigenfunctions
of the extension kernel are the ones that are used in Eq. \ref{eq:Nys-ext-phi}.
A procedure for finding an appropriate $\sigma_{ext}$ is described
in \cite{LKC06}.

\section{\label{sec:Clustering-and-Classification}Clustering and classification
algorithms}

The classification algorithms for processing acoustic signals consist
of two phases: learning and detection. Two algorithms for clustering
and classification are presented. These algorithms use the same procedure
for feature extraction from a signal. However, they differ in the
dimensionality reduction algorithms they employ. One uses the diffusion
maps algorithm while the other uses PCA.

\subsection{Preparation of the recorded datasets}

The acoustics recordings are split into two sets: the training set
which is used in phase I and the test set which is used in phase II.

\paragraph*{\label{sub:PrepareLS}Preparation of the acoustic recordings for
the learning phase}

The input data for the learning phase consists of acoustic signals
given in a pulse-code modulation (PCM) format. The acoustic signals
may have different sizes and their classifications are known a-priory.
First, acoustics patterns are extracted from the input signals, where
each pattern represents a single acoustic event. These events are
then separated into classes, $c_{1},...,c_{k}$, where $k\in\mathbb{N}$
is the number of different classes (number of different types of events)
in the dataset. In examples \ref{enu:Detection-Pulse} and \ref{enu:Detection-of-cardio},
$c_{1},c_{2}$ are naturally determined as healthy and not healthy.
Specifically, %can be, for example, the classes of moving vehicles and background sounds, respectively,
in example \ref{enu:Detection-of-cardio} $c_{i}$ is either normal
heart beats or a vascular disease. 

The number of patterns is given by $n_{Ts}$. The training sample
set recordings, whose classification is known a-priory, is denoted
by $\Omega=\{s_{i}\}_{i=1}^{n_{Ts}}$, where each $s_{i}\in\Omega,$
$1\leq i\leq n_{Ts}$, represents an acoustic pattern from the training
set. Each signal $s_{i}$ from the training set $\Omega$ belongs
to a class $c_{j}$, $1\leq j\leq k$.

\paragraph*{\label{sub:PrepareTestSet}Preparation of the recordings for the
classification phase}

In this phase, a signal from the testing set is divided into short
segments. In the detection phase, each segment is processed and assigned
to the most suitable class.

\subsection{Learning phase{\normalsize \label{sub:Learning-phase} }}

This phase contains two parts. The first part applies a feature extraction
procedure to every signal and in the second part two methods of dimensionality
reduction are used for clustering and classification of the training
set to different classes.

\paragraph*{\label{sub:Part-I-:FE}Part I: Feature extraction}

In order to extract the requires features from each signal, the following
steps are employed:
\begin{description}
\item [{1.}] \textbf{Decomposition into windows:} Each acoustic signal
$s_{i}\in\Omega$ is decomposed is decomposed into windows of size
$l=2^{r}$, $r\in\mathbb{N}$, with overlapping of $\nu\%$. The result
is given by $\left\{ w_{j}\right\} _{j=1}^{n_{w}},\, w_{j}\in\mathbb{R}^{l}$.
We denote the classification of each window by $C\left(w_{j}\right)$.
\item [{2.}] \textbf{Application of spline wavelet:} We use the sixth order
spline wavelet packet. A spline wavelet is applied up to a scale $D\in\mathbb{N}$
on each window $w_{j}$. Typically, if $l=2^{10}=1024,$ then $D=6$
and if $l=2^{9}=512$ then $D=5$. The coefficients are taken from
the last scale $D$. This scale contains $l=2^{r}$ coefficients that
are arranged into $2^{D}$ blocks of length $2^{r-D}$. Each block
is associated with a certain frequency band. These bands form a near
uniform partition of the Nyquist frequency domain into $2^{D}$ parts.
The output of this step is a set of spline wavelet coefficients for
each window $w_{j}$. 
\item [{3.}] \textbf{Calculation of the energy:} We construct the acoustic
signature of a certain window using the distribution of energy among
its wavelet packet coefficients. The energy is calculated by normalized
sum of the coefficients in each block that was calculated in the previous
step. Consequently, this step produces for each window $w_{j}$ a
vector of size $2^{D}$ that contains the block energies of its wavelet
packet coefficients. This operation reduces the dimension by a factor
of $2^{r-D}$. 
\item [{4.}] \textbf{Averaging:} This step is applied in order to reduce
perturbations and noise. We calculate the average of every $\mu$
consecutive windows which are associated with the same signal in order
to receive a more robust signature. 
\end{description}

\paragraph*{Part II: Dimensionality reduction of the learning set}

After the features are extracted, we further reduce the dimensionality
of the feature training set by applying two methods of dimensionality
reduction: diffusion maps (Chapter \ref{cha:Diffusion-Maps}) and
PCA (Section \ref{sub:Global-methods}). This step also clusters the
data according to the a-priory classes. 

Let $F=\left\{ f_{j}\right\} _{j=1}^{n}$, where $f_{j}$ is a vector
of size $2^{D}$, be the features extracted from the signal windows
as described in Section \ref{sub:Part-I-:FE}. There are $n$ rows,
each of size $2^{D}$, where each row describes a fraction of a signal
whose classification is known. 

We apply the diffusion maps and PCA methods to $F$ and denote the
produced low-dimensional embedding by $F^{DM}=\left\{ f_{j}^{DM}\right\} _{j=1}^{n}$
and $F^{PCA}=\left\{ f_{j}^{PCA}\right\} _{j=1}^{n}$, respectively,
where $f_{j}^{DM}$ is of size $q$. Thus, we achieve a dimensionality
reduction from $2^{D}$ to $q$. Since the PCA algorithm involves
a step where the data is centered around the origin, we store the
center of gravity of $F$ for the classification phase.

\subsection{On-line classification phase}

The classification phase is done on-line. There is no need to wait
for the entire signal to be received. In order to classify a signal
at time $t$, the algorithm only needs the $\mu$ consecutive overlapping
windows of size $l$ with $\nu\%$ overlap that immediately precede
time $t$. $\mu$ and $l$ are the same as in the learning phase.
Given such a sequence, its classification is obtained by applying
the following step:
\begin{enumerate}
\item Feature extraction from the sequence giving a feature vector.
\item Embedding of the feature vector into the low-dimensional space. 
\item Assignment of each window in the tested recording to a specific class. 
\end{enumerate}
The feature extraction that is used in the classification uses the
same procedure as in the learning phase (Section \ref{sub:Learning-phase}).
However, the second step depends on the method that was used for the
dimensionality reduction. In case DM was used, the feature vector
is embedded using the Geometric Harmonic algorithm (Section \ref{sub:Geometric-Harmonic-(out}).
Specifically, the embedding space is extended to contain the feature
vector. If PCA was used, then the embedding of the feature vector
is obtained by projecting it on the principal components (after reducing
the center of gravity of the training set from the feature vector).
The third step is the same for both dimensionality reduction schemes
and is described below.

\paragraph*{Classification of a new window}

For each new embedded window (point), its $\delta$ nearest neighbors
in the embedding space are found%
\footnote{We used the TSTOOL software from http://www.dpi.physik.uni-goettingen.de/tstool/
to find the nearest neighbors.%
}, where $\delta\geq1$ is given as a parameter. The classification
is set according to the dominating class of the neighbors i.e. the
class to which the largest number of neighbors belong. The classification
probability is determined according to the ratio between the number
of neighbors from the dominating class and $\delta$.

\section{Experimental results\label{sec:Experimental-Results}}

We demonstrate the generality and the robustness of the algorithms
using two %three different examples of acoustics datasets. In all
the examples, we use the suggested algorithms with minor changes in
the algorithms' parameters which were determined empirically. The
algorithms' parameters are: $L$ - window size, $\nu$ - overlapping
percent, $\mu$ - number of windows to average, $D$ - the scale of
spline wavelet, $q$ - number of chosen eigenvalues in the DM algorithm.

The learning set was constructed as follows: we extracted from the
recordings, which were assigned to the learning phase, fragments -
each of which belongs to a certain known class.

The classification phase was tested on recordings that did not participate
in the training set construction. For every recording that was processed
in the classification phase, we provide graphs that include the following
parts: 
\begin{enumerate}
\item The original signal.
\item Graphs of the probabilities of the events that most likely took place
according to the output of the DM-based classification algorithm.
\item Graphs of the probabilities of the events that most likely took place
according to the output of the PCA-based classification algorithm. 
\end{enumerate}

\subsection{Experiment 1: Analyzing of radial artery pulse}

The signals that were used in this experiment consisted of the radial
artery pulse. This signals were captured at 200 seconds intervals
by an optoelectronic sensor at a sampling rate of 100 Hz.

The classes that have to be separated are: 
\begin{itemize}
\item Hypertension.
\item Nocturnal enuresis
\end{itemize}
The data were collected from different children (ages 9-14) in different
occasions. The training set contained 36 recordings and the detection
set contained 2 recordings that did not take part in the learning
phase. The following parameters were used in the learning and classification
phases: $L=1024$, $\nu=25\%$, $\mu=7$, $D=5$, $q=3$.

The justification for using the spline wavelet packet is: the contribution
of each oscillation inside the human body contains only a few dominating
bands. As the conditions are changed the configuration of these bands
may vary but the general disposition remains. From this we can assume
that the signature for the class of pulse signals which are related
to a certain disease, can be obtained as a combination of the inherent
energies in a set of blocks of the wavelet packet coefficients of
the decomposed signal.

\paragraph*{Results from the learning phase}

\begin{figure}[!h]
\begin{centering}
\includegraphics[width=0.99\columnwidth,height=7cm]{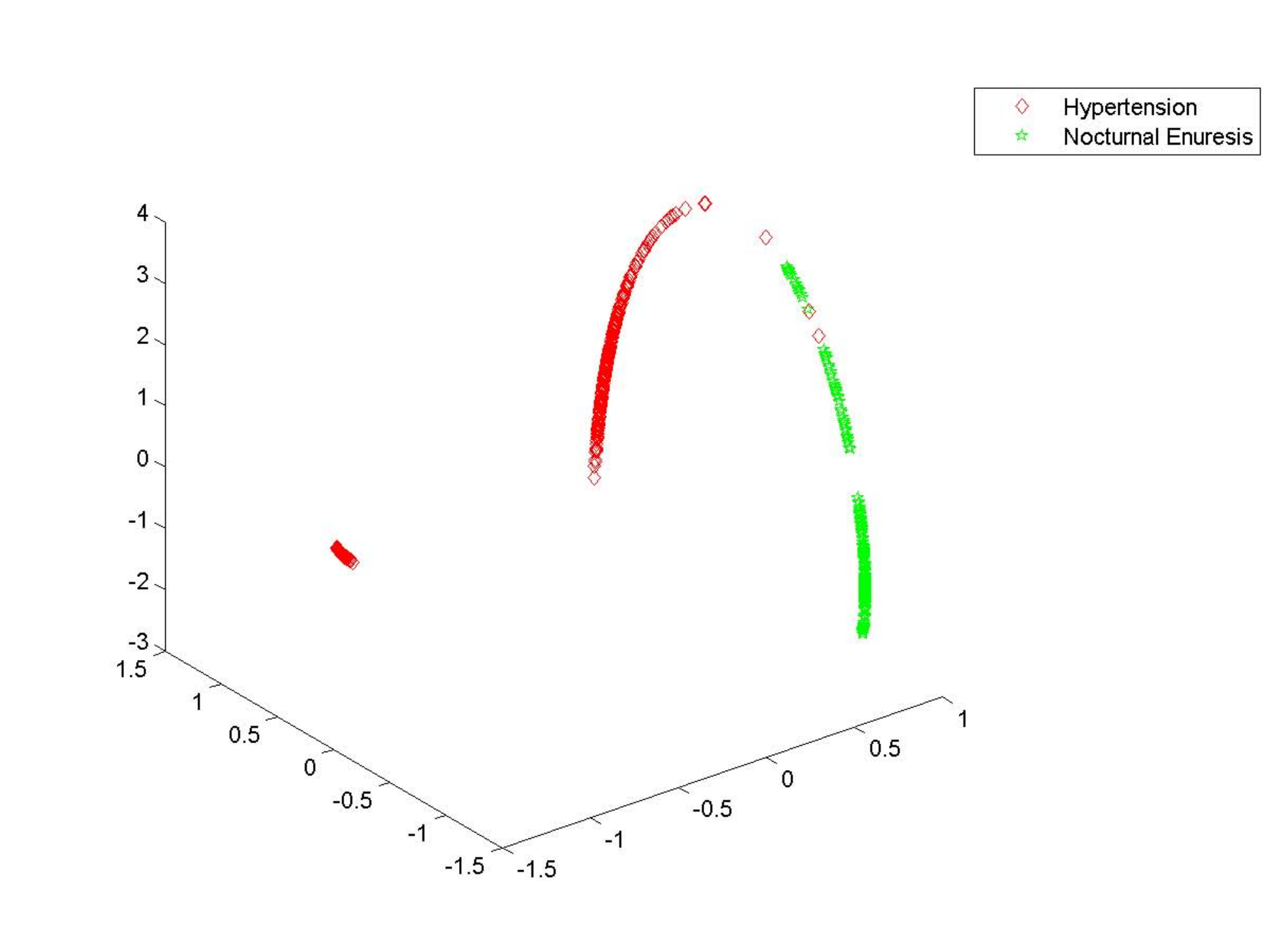}
\par\end{centering}

\caption{\label{fig:ExpII_ClustersDM}Clusters generated by the DM algorithm.
The plot is the data embedded into the diffusion space which is obtained
by the first three eigenvectors. }
\end{figure}

The clustering result from the DM algorithm is given in Fig. \ref{fig:ExpII_ClustersDM}.
The first three eigenvectors provide a complete separation into two
disjoint clusters.

\begin{figure}[!h]
\begin{centering}
\includegraphics[width=0.99\columnwidth,height=7cm]{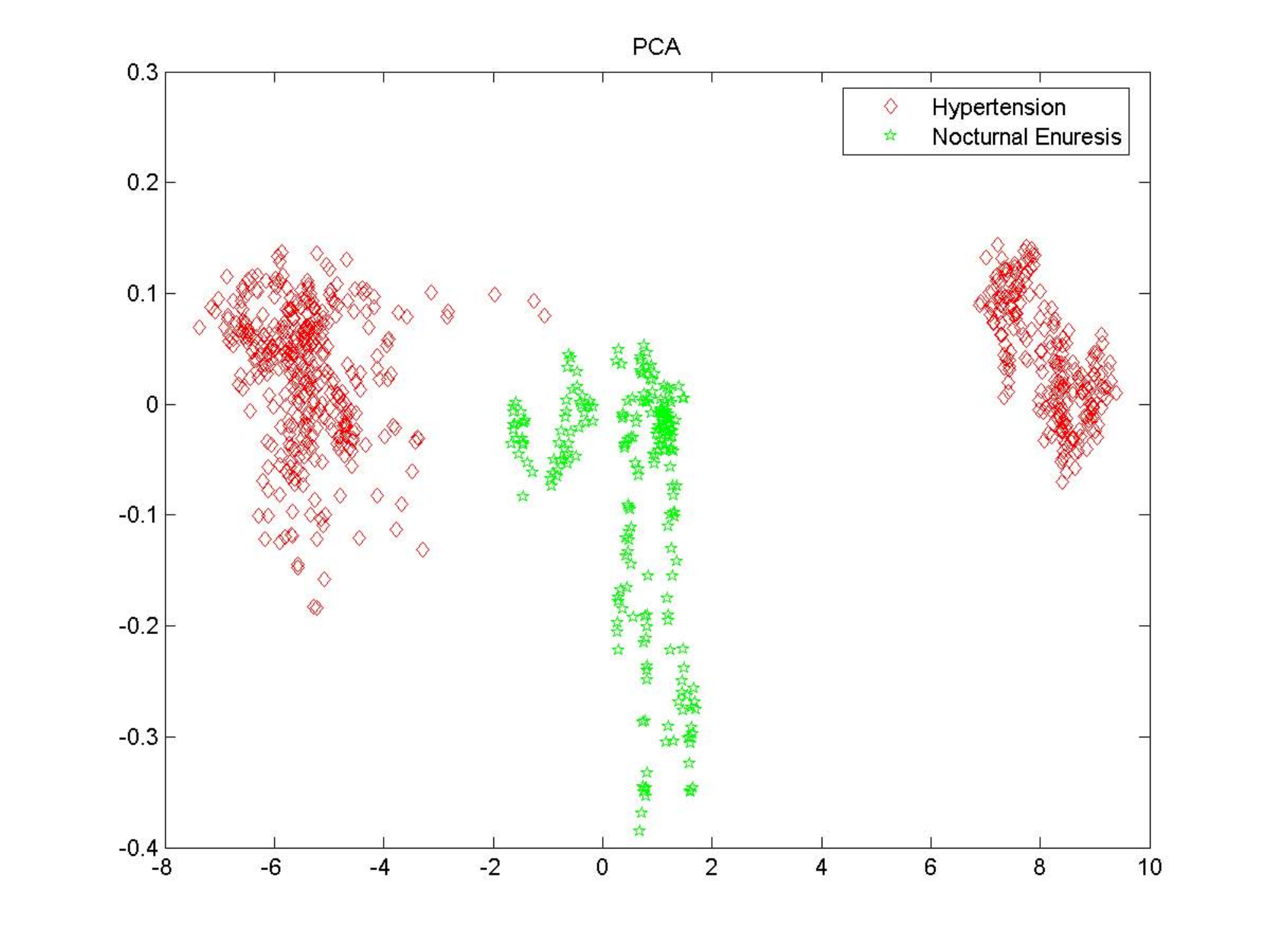}
\par\end{centering}

\caption{\label{fig:ExpII_ClustersPCA}Clusters generated by the PCA algorithm.
The plot is the data embedded into the space that is spanned by the
first two eigenvectors. }
\end{figure}

The clustering result from the PCA algorithm is given in Fig. \ref{fig:ExpII_ClustersPCA}.
The first two eigenvectors provide a complete separation between the
eigenvectors of the hypertension class (two clusters) and the eigenvectors
of the nocturnal enuresis class (two clusters).

\paragraph*{Results from the classification phase}

\begin{figure}[!h]
\begin{centering}
\includegraphics[width=0.99\columnwidth,height=8cm]{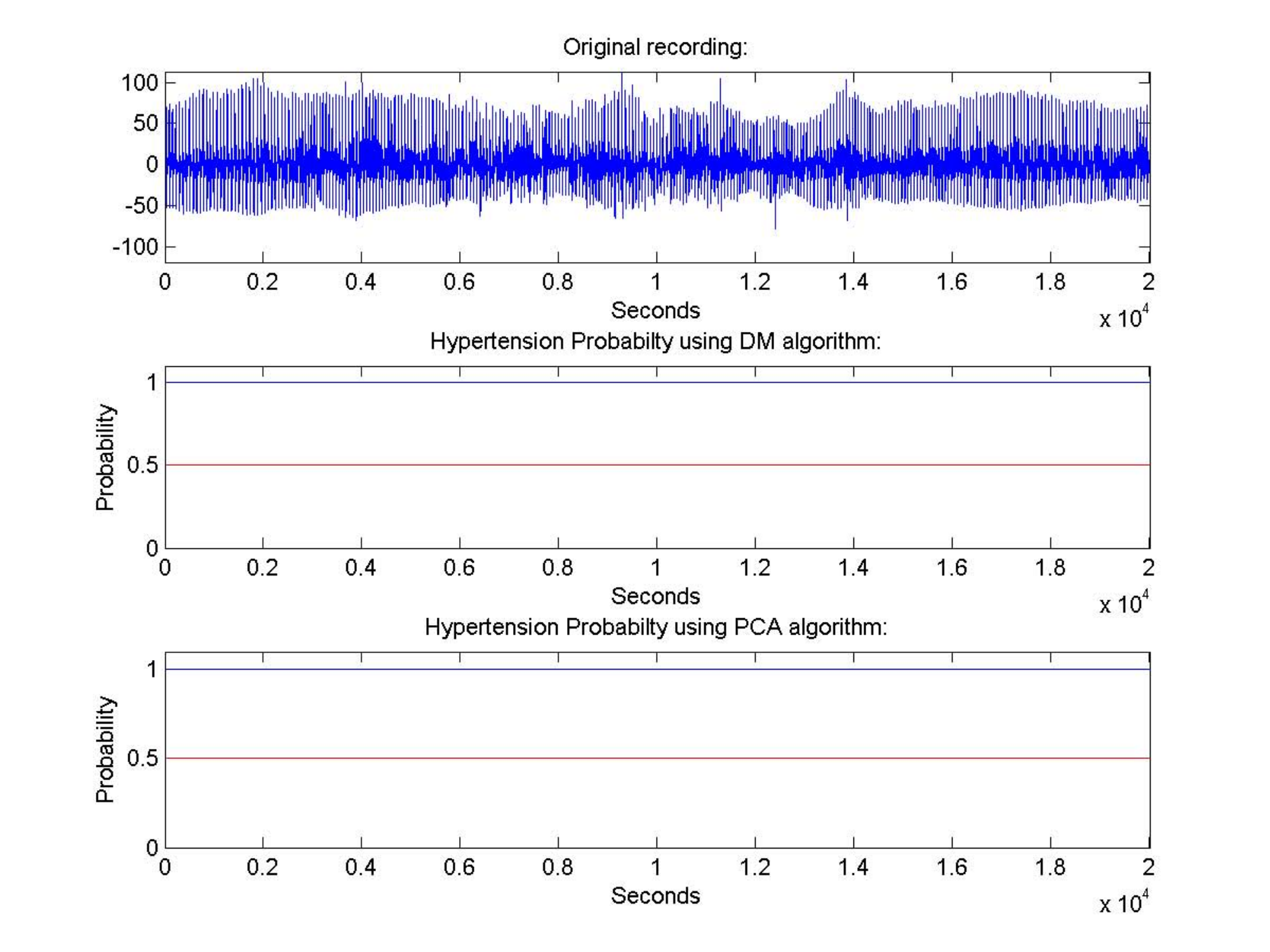}
\par\end{centering}

\caption{\label{fig:ExpII_HprK7}Classification results for a recording that
contains a hypertension disorder. Top: The original recording. Middle:
The blue line illustrates the probability of hypertension using the
DM algorithm. Bottom: The blue line illustrates the probability of
hypertension using the PCA algorithm. Both algorithm succeed in the
classification of this signal.}
\end{figure}

\begin{figure}[!h]
\begin{centering}
\includegraphics[width=0.99\columnwidth,height=10cm]{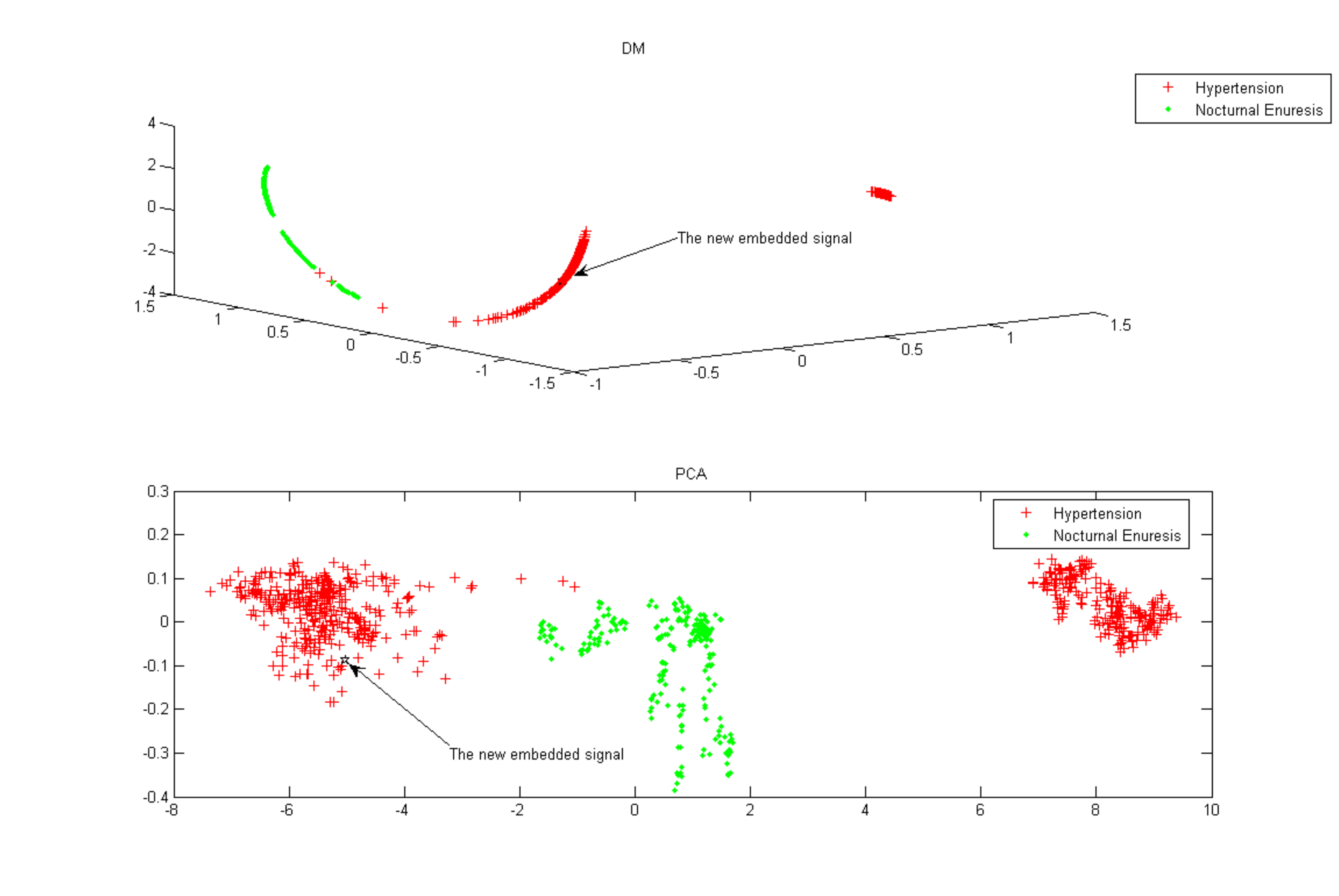}
\par\end{centering}

\caption{\label{fig:ExpIInewEmbed_HprK7}Embedding of a fraction of the signal
from Fig. \ref{fig:ExpII_HprK7} into the lower dimensional space.
Top: Embedding using the DM algorithm. Bottom: Embedding using the
PCA algorithm. }
\end{figure}

Figure \ref{fig:ExpII_HprK7} contains the results of a hypertension
signal that was not part of the training set. The blue line in the
two bottom plots represents the detection probability of the DM and
PCA algorithms of hypertension. It can be seen that they are both
equal to one throughout the recording and thereby illustrating accurate
classification since the signal belongs to a hypertensive patient.
In Fig. \ref{fig:ExpIInewEmbed_HprK7} we see a fraction of the new
signal embedded into the cluster that corresponds to the hypertension
class. 

\begin{figure}[!h]
\begin{centering}
\includegraphics[width=0.99\columnwidth,height=8cm]{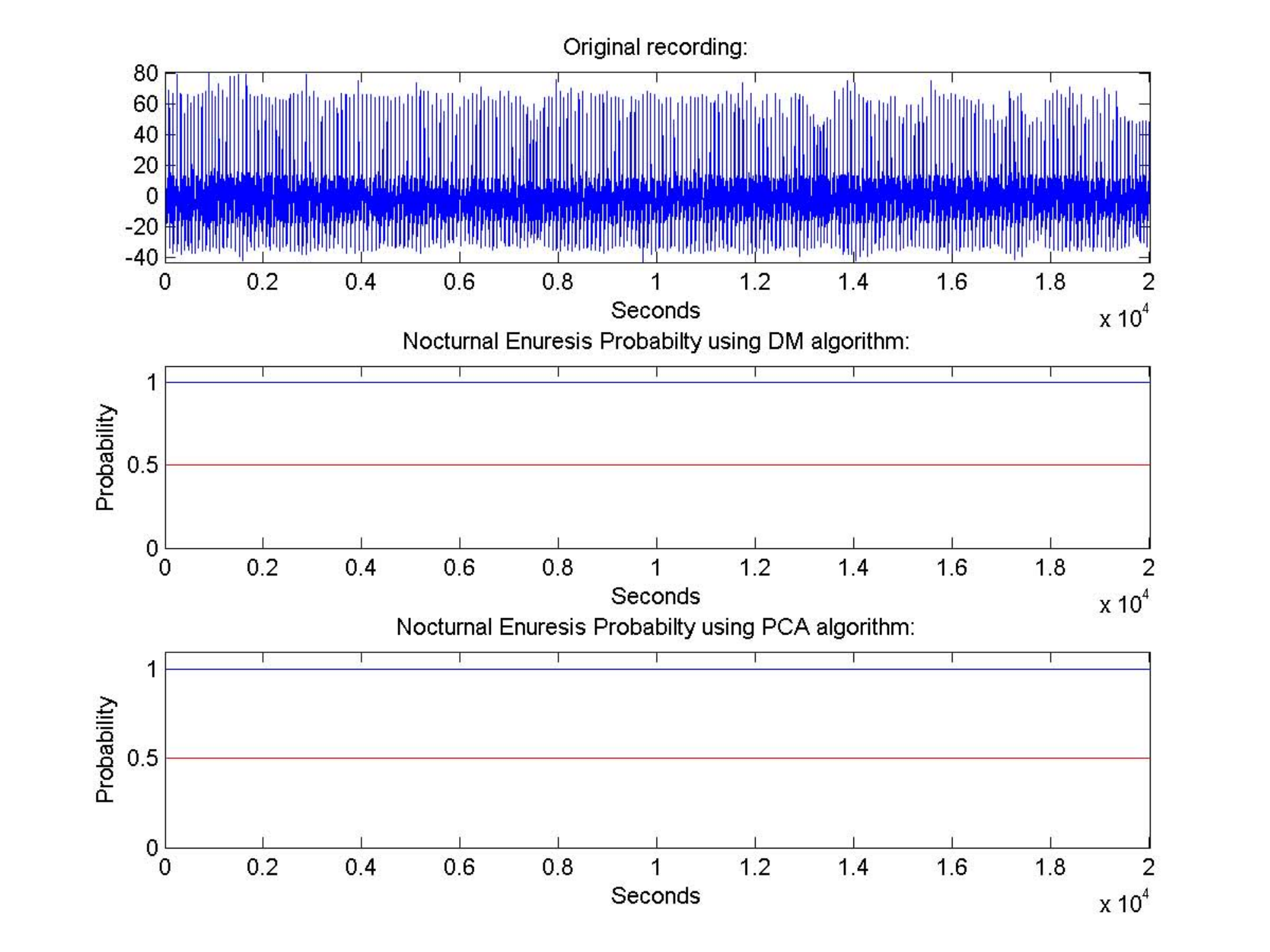}
\par\end{centering}

\caption{\label{fig:ExpII_EnK16}Classification results for a recording that
contains a nocturnal enuresis signal. Top: The original recording.
Middle: The probability of nocturnal enuresis using the DM algorithm.
Bottom: The probability of nocturnal enuresis using the PCA algorithm. }
\end{figure}

\begin{figure}[!h]
\begin{centering}
\includegraphics[width=0.99\columnwidth,height=12cm]{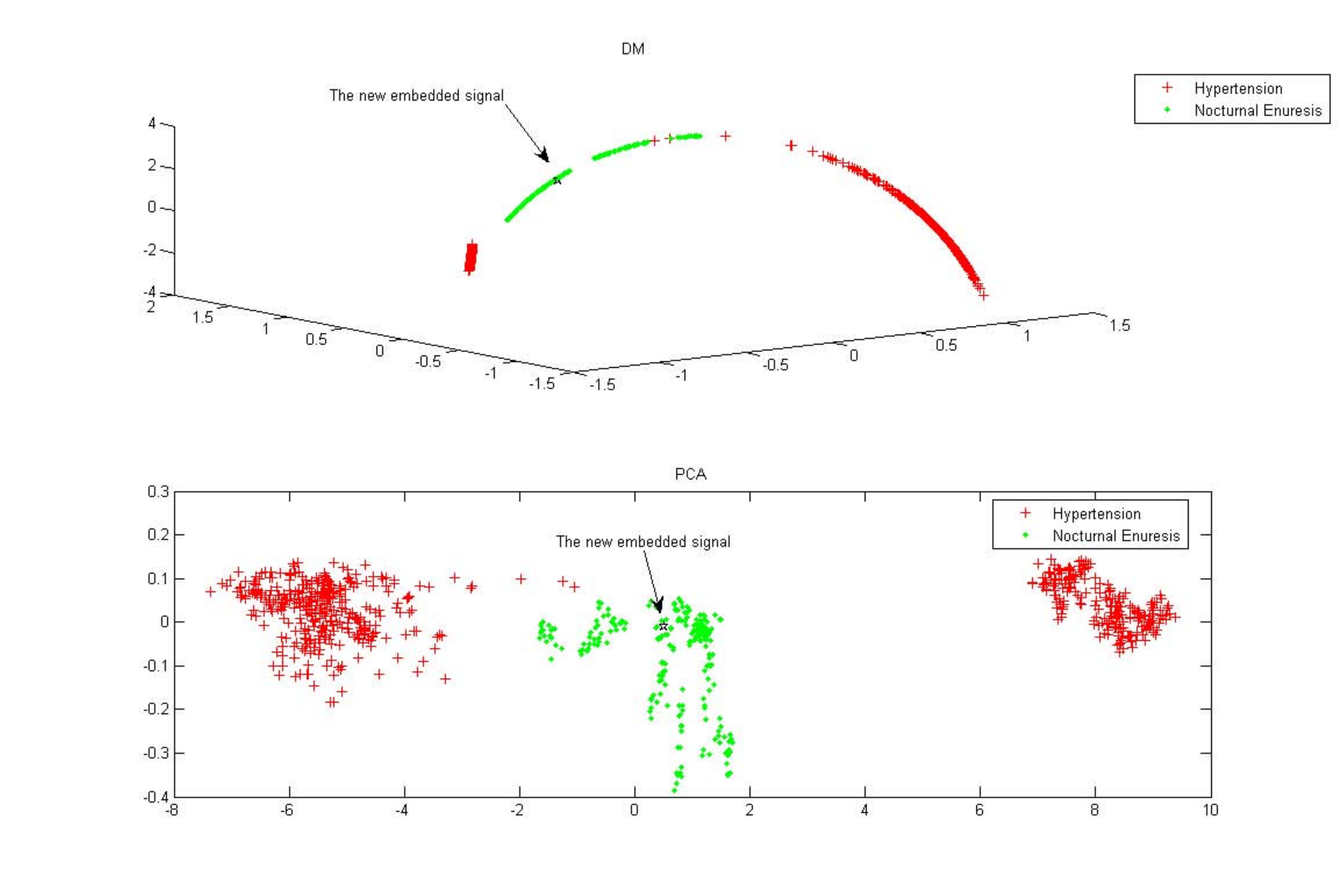}
\par\end{centering}

\caption{\label{fig:ExpIInewEmbed_Enk16}Embedding of a fraction of the signal
from Fig. \ref{fig:ExpII_EnK16} into a lower dimensional space. Top:
Embedding using the DM algorithm. Bottom: Embedding using the PCA
algorithm. }
\end{figure}

Figure \ref{fig:ExpII_EnK16} contains the classification results
of a nocturnal enuresis signal. We see that in both algorithms the
classification is accurate. In Fig. \ref{fig:ExpIInewEmbed_Enk16}
we see that a fraction of the new signal embedded into the cluster
that represents the nocturnal enuresis class.

\subsection{Experiment 2: Detection of a cardio vascular diseases}

The signals for this experiment were obtained using a pulse detector
working at sampling rates 22050 Hz and 11025 Hz. The signals were
downsampled to 2205 Hz. 

The classes in this experiment are: (a) normal heart beats and (b)
a cardio vascular disease. The data were collected from different
adults in different occasions. The learning sample set consisted of
7 recordings, 4 of them describe normal cardio behavior and 3 represent
a cardio vascular disorder. The detection set contained 2 recordings
that did not participate in the learning phase. The following parameters
were used in the learning and classification phases: $L=1024$, $\nu=75\%$,
$\mu=3$, $D=6$, $q=3$. These parameters were determined empirically.

\paragraph*{Results from the learning phase}

\begin{figure}[!h]
\begin{centering}
\includegraphics[width=0.99\columnwidth,height=8cm]{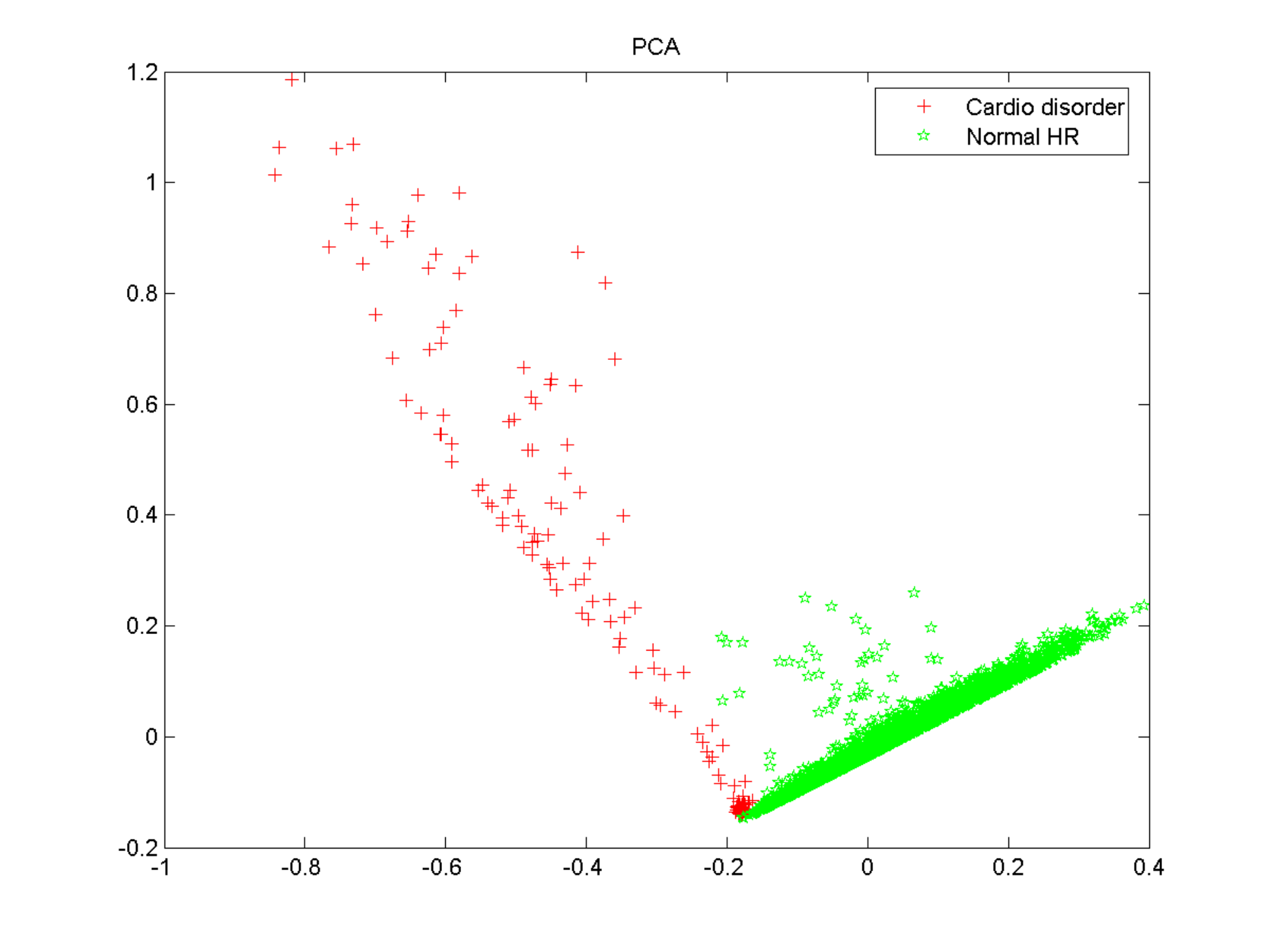}
\par\end{centering}

\caption{\label{fig:ExpIII_ClustersPCA}Clusters generated by the application
of the PCA algorithm. The plot is the data embedded into the space
 spanned by the first two eigenvectors. }
\end{figure}

\begin{figure}[!h]
\begin{centering}
\includegraphics[width=0.99\columnwidth,height=8cm]{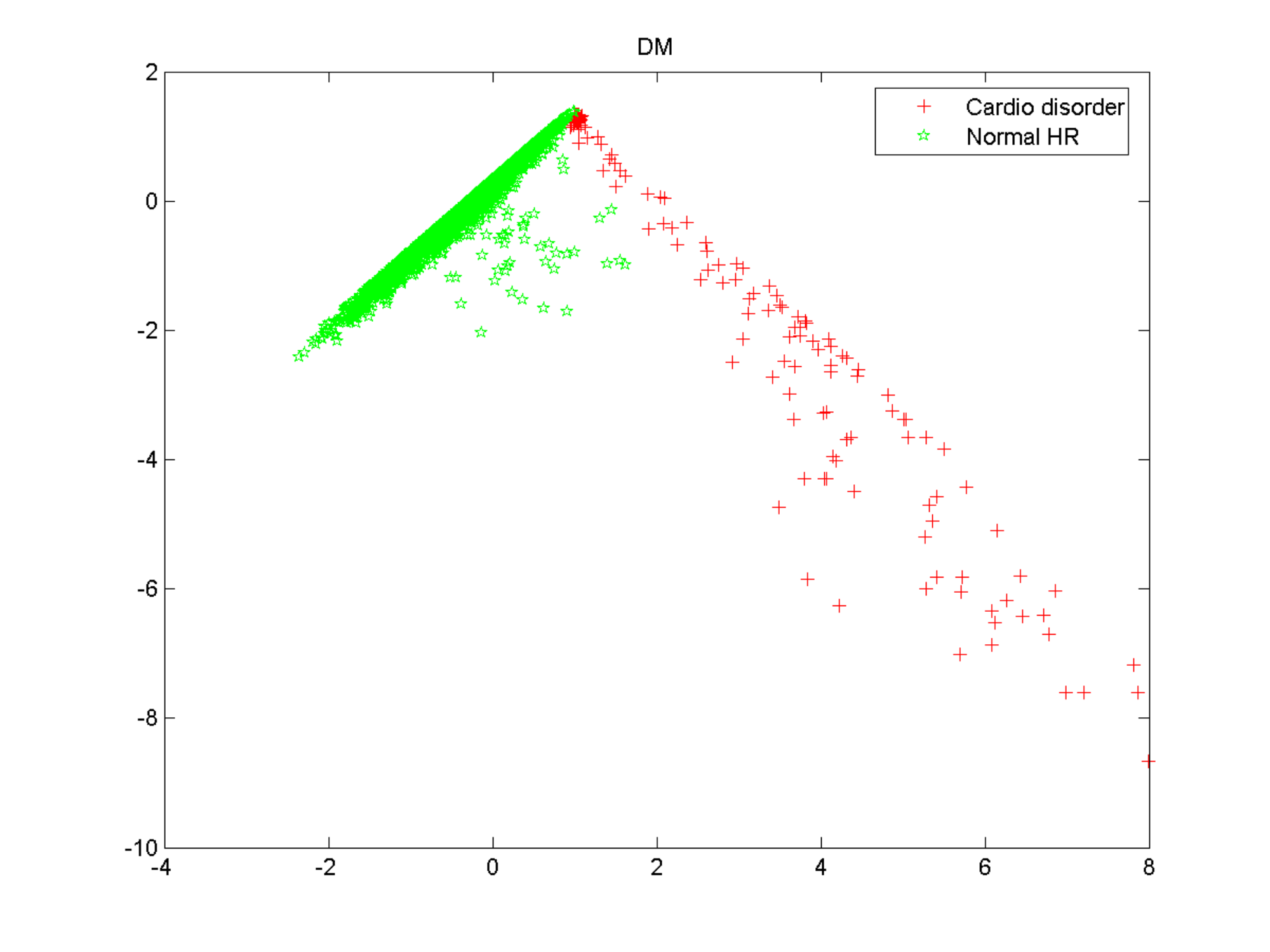}
\par\end{centering}

\caption{\label{fig:ExpIII_ClustersDM}Clusters generated by the application
of the DM algorithm. The plot is the data embedded into the space
spanned by the first two eigenvectors. }
\end{figure}

The clustering result of the PCA algorithm is given in Fig. \ref{fig:ExpIII_ClustersPCA}.
The first two eigenvectors provide a complete separation into two
disjoint clusters. The clustering result of the DM algorithm is given
in Fig. \ref{fig:ExpIII_ClustersDM}. The first two eigenvectors provide
a complete separation into two disjoint clusters. 

The clustering results obtained by both the PCA and the DM algorithms
in this example are similar indicating that the manifold is close
to linear in this case.
 
\paragraph*{Results from the classification phase}

\begin{flushleft}
Figure \ref{fig:ExpIII_Gardmr7} contains the classification results
of a cardio vascular disorder signal. DM and PCA provide clusters
that classify the data accurately.
\begin{figure}[!h]
\begin{centering}
\includegraphics[width=0.99\columnwidth,height=9cm]{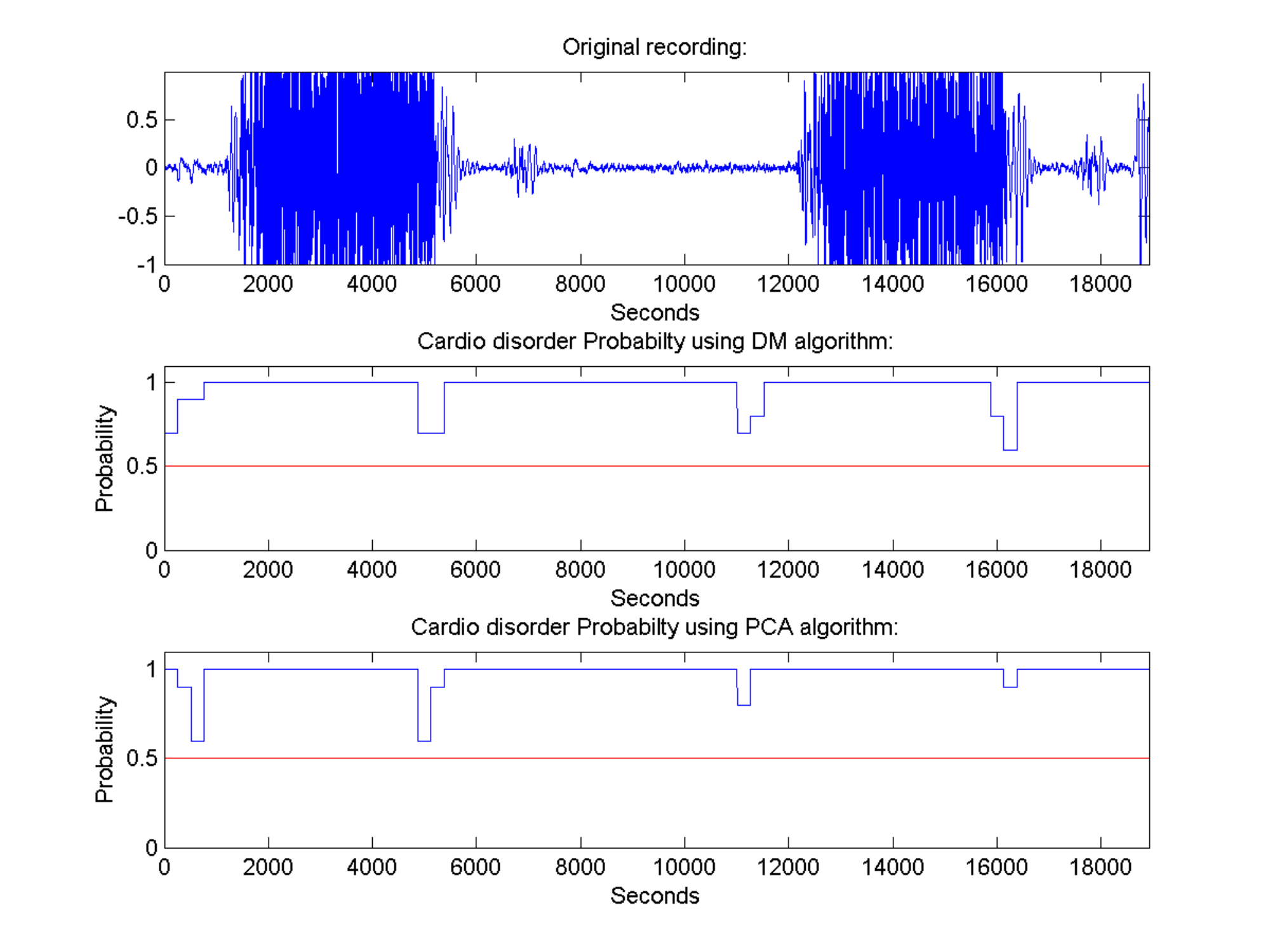}
\par\end{centering}

\caption{\label{fig:ExpIII_Gardmr7}Classification of a recording that contains
a cardio vascular disorder. Top: Original recording. Middle: The probability
for a cardio vascular disorder using the DM algorithm. Bottom: The
probability for a disorder using the PCA algorithm. }
\end{figure}
 
\par\end{flushleft}

\begin{flushleft}
Figure \ref{fig:ExpIII_N5_11} contains the classification results
of a normal heart beat signal. Both the DM and the PCA algorithms
classify the data accurately -- they detect the abnormal heart beats
and do not generate false detections.
\begin{figure}[!h]
\begin{centering}
\includegraphics[width=0.99\columnwidth,height=10cm]{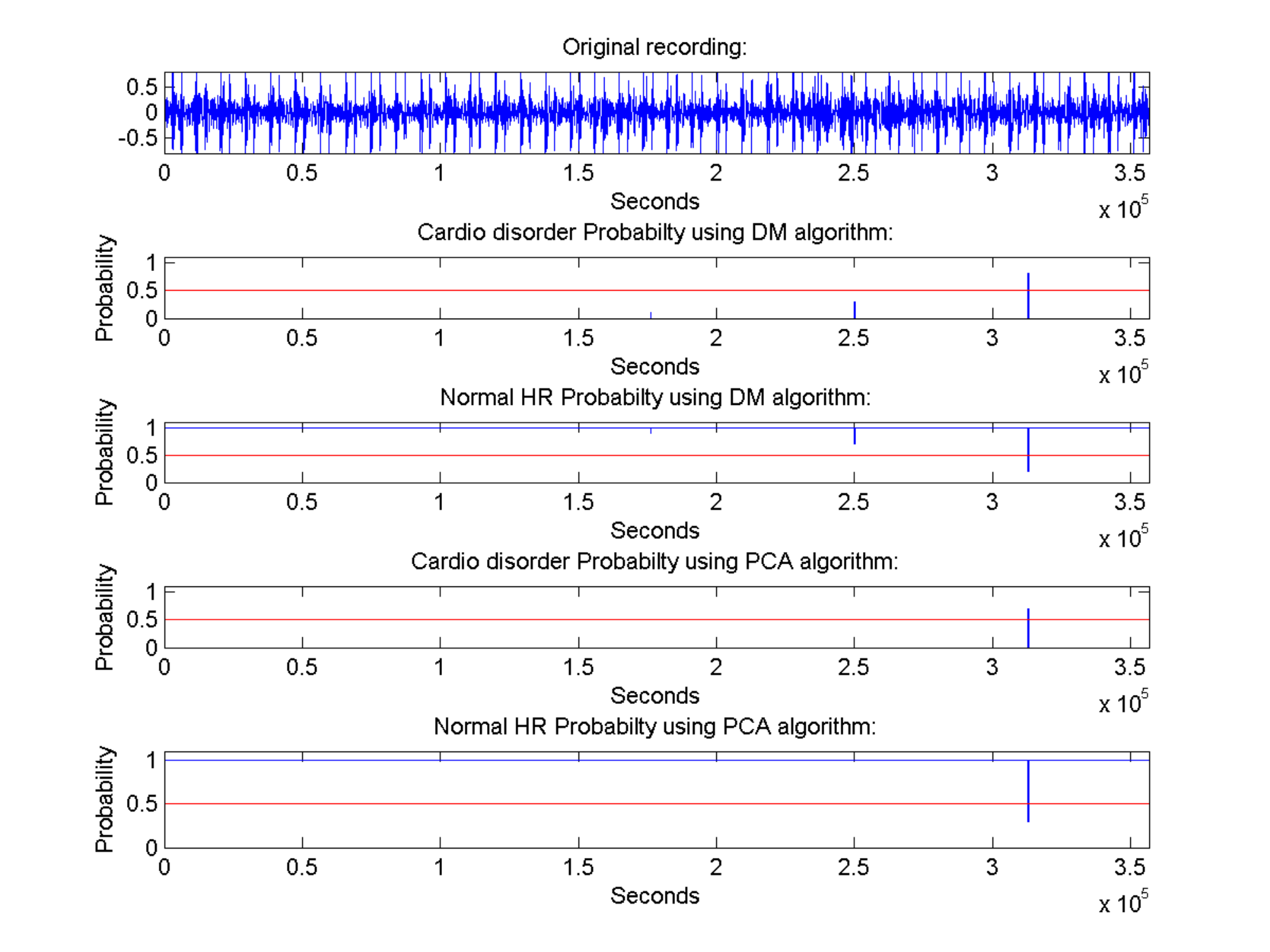}
\par\end{centering}

\caption{\label{fig:ExpIII_N5_11}Classification of a normal heart beat signal.
Top: Original recording. Second from top: The probability for a cardio
disorder using the DM algorithm. Third from top: The probability for
a normal cardio behavior using the DM algorithm. Fourth from top:
The probability for a cardio disorder using the PCA algorithm. Bottom:
The probability for a normal cardio behavior using the PCA algorithm. }
\end{figure}

\par\end{flushleft}

\section{\label{sec:Conclusions}Conclusions}

In this chapter, we introduced two algorithms for the detection of
different types of events according to their acoustic signatures.
Each algorithm has two phases. In both algorithms, dominating features
were extracted from every acoustic signal (Section \ref{sub:Learning-phase}).
In order to cluster different events, the features of the signals
were embedded into a lower dimensional space using either the DM algorithm
(Chapter \ref{cha:Diffusion-Maps}) or the PCA algorithm (Section
\ref{sub:Global-methods}). In the on-line classification phase of
new signals, acoustic dominating features were extracted from the
signal by employing similar steps to the those that were used in the
learning phase. The features of the new signal were embedded by using
either the geometric harmonic scheme (in the DM algorithm) or a simple
projection (in the PCA algorithm).

In the specific domain that was examined both algorithms perform very
well where the PCA algorithm has a slight time complexity advantage.
Nevertheless, when using the proposed schemes for other application
domains, the accuracy of the PCA algorithm might be inferior to that
of the DM algorithm due to its limiting capabilities in reducing the
dimensionality of data-sets with a non-linear structure.

\chapter{Final Conclusions}

In this thesis I introduced a novel method for dimensionality reduction
- diffusion bases. The method is based on the diffusion map dimensionality
reduction algorithm. I demonstrated the effectiveness of the method
for the segmentation of images that originated from three different
domains: hyper-spectral imagery, multi-contrast MRI and video. For
each domain, a different algorithm was tailored in which the diffusion
bases method was the key step that facilitated the generation of the
results. These results were shown (where applicable) to be competitive
with current state-of-the-art methods for the investigated problems.

In the second part of the thesis, I investigated dimensionality reduction
as a tool for solving problems from various domains. Specifically,
I used dimensionality reduction in order to uniquely characterize
materials using their spectral signatures. I demonstrated how dimensionality
reduction can be utilized for classification. I proposed an ensemble
algorithm which incorporates three different dimensionality reduction
techniques for the detection of vehicles. I also introduced two algorithms
that use PCA and DM as tools for classification of medical signals.
A clear conclusion can be made: \emph{dimensionality reduction proves
to be a key tool in classification tasks}. A question remains: ``Which
dimensionality reduction technique should be used ?''. The answer
depends on the problem domain. Nevertheless, ensemble methods which
apply several dimensionality reduction techniques might provide a
more general solution than the application of any single technique.

I hope this thesis will trigger more research on the integration of
dimensionality reduction for solving other problems in various other
domains.

\bibliographystyle{acm}
\bibliography{references}

\end{document}